\documentclass[letterpaper]{article}
\usepackage[preprint]{aaai2027}
\usepackage[hyphens]{url}
\usepackage{natbib}
\usepackage{graphicx}
\usepackage{booktabs}
\usepackage[table]{xcolor}
\usepackage{pifont}
\usepackage{tabularx}
\usepackage{longtable}
\usepackage{subcaption}
\usepackage{caption}
\usepackage{algorithm}
\usepackage{algcompatible}
\usepackage{algpseudocode}
\usepackage{adjustbox}
\usepackage{array}
\usepackage{multirow}
\usepackage{makecell}
\usepackage{dsfont}
\usepackage{amsmath,amssymb,amsthm}
\usepackage{mathtools}
\usepackage{collcell}
\usepackage{xparse}
\usepackage{xstring}
\usepackage[hidelinks,bookmarks=false,pdfencoding=auto]{hyperref}
\hypersetup{
  pdftitle={On the Few-Shot Adaptation of Vision-Language Models: A Few Gradient Signs Are All You Need},
  pdfauthor={Nairouz Mrabah; Ismail Ben Ayed; Mahdi S. Hosseini; Eric Granger}
}
\providecommand{\theHalgorithm}{\arabic{algorithm}}
\definecolor{LightGray}{RGB}{232,232,232}
\definecolor{mygreen}{RGB}{44,160,44}
\definecolor{myred}{RGB}{214,39,40}

\newtheorem{theorem}{Theorem}
\newtheorem{lemma}{Lemma}
\newtheorem{assumption}{Assumption}
\newtheorem{proposition}{Proposition}
\newtheorem{corollary}{Corollary}
\newcommand{\R}{\mathbb{R}}
\newcommand{\E}{\mathbb{E}}
\newcommand{\Prob}{\mathbb{P}}
\newcommand{\1}{\mathbf{1}}
\DeclareMathOperator{\sgn}{sgn}
\DeclareMathOperator{\Top}{Top}
\DeclareMathOperator{\supp}{supp}

\definecolor{SuppLinkBlue}{RGB}{0,70,140}
\newcommand{\suppanchor}[1]{%
  \leavevmode\hypertarget{suppnav.#1}{}%
  \label{suppnav:#1}%
}
\newcommand{\supplink}[2]{%
  \hyperlink{suppnav.#1}{\textcolor{SuppLinkBlue}{#2}}%
}

\newcommand{\pmdelimiter}{\space$\pm$\space}
\newcommand{\pmcell}[1]{%
  \IfSubStr{#1}{\pmdelimiter}{%
    \StrBefore{#1}{\pmdelimiter}[\pmcellmean]%
    \StrBehind{#1}{\pmdelimiter}[\pmcellstd]%
    \mbox{\pmcellmean{\scriptsize(\pmcellstd)}}%
  }{#1}%
}
\newcolumntype{d}{>{\collectcell\pmcell}c<{\endcollectcell}}

\title{On the Few-Shot Adaptation of Vision-Language Models:\\ A Few Gradient Signs Are All You Need}
\author{Nairouz Mrabah\textsuperscript{1}, Ismail Ben Ayed\textsuperscript{1},
Mahdi S. Hosseini\textsuperscript{2}, Eric Granger\textsuperscript{1}}
\affiliations{
\textsuperscript{1}Department of Systems Engineering, ETS Montreal, Canada\\
\textsuperscript{2}Department of Computer Science and Software Engineering, Concordia, Canada
}

\begin{document}
\pagestyle{plain}
\raggedbottom
\widowpenalty=10000
\maketitle
\thispagestyle{plain}
\alglanguage{algorithmic}
\begin{abstract}
The limited supervision available for the few-shot adaptation of vision-language models makes gradient estimation particularly noisy. We find that the gradient magnitudes are severely miscalibrated, while the signs of a small top-magnitude subset remain informative. This motivates \emph{SignSparseSGD} and \emph{DynSignSparseSGD}, two state-free optimizers that use gradient magnitudes only to select where to update and fixed-magnitude signs to decide how to update. The static variant fixes the support after the first minibatch. The dynamic variant refreshes it at every step. Across 11 datasets, three CLIP backbones, and five shot levels, the proposed methods are competitive with or outperform parameter-efficient and optimization-based baselines on average, while exhibiting markedly lower sensitivity to long training schedules. They introduce no task-specific modules, maintain no first- or second-moment optimizer states, and use one shared configuration across datasets, backbones, and shot levels. Our code is included in the supplementary materials and will be released upon publication.
\end{abstract}

%\footnote{Code is included in the suppl. materials and will be released upon publication.}
\section{Introduction}
\label{sec:introduction}

Foundation vision--language models (VLMs) such as CLIP~\cite{radford2021learning} support zero-shot recognition through aligned image and text representations. Few-shot adaptation remains difficult because support-set gradients are noisy, and optimization can quickly overfit the available examples~\cite{mrabah2025so,silva2024closer}. %Full-backbone adaptation also requires dense gradients and persistent optimizer states~\cite{zhao2024galore}. 
Parameter-efficient fine-tuning (PEFT) methods reduce the trainable set of parameters through prompts~\cite{zhou2022coop}, adapters~\cite{gao2024clipadapter}, low-rank factors~\cite{hu2022lora}, or parameter selection~\cite{nguyen2024saft}.

Despite their diversity, most PEFT methods inherit the same optimization recipe. They keep the gradient‑based update rule fixed, typically using Adam or SGD, and they focus on parameters or lightweight modules that should be updated. This implicitly assumes that the gradients in the few‑shot regime remain sufficiently reliable to drive learning, and that optimization noise can be handled by hyperparameter tuning. However, the structure of this few‑shot gradient noise is rarely characterized. In particular, we ask this question: is the discrepancy between a 1‑shot gradient and a higher‑shot estimate mainly a magnitude error, or does it frequently induce sign flips? Without answering this question, it is unclear whether magnitudes should be trusted for step scaling, whether sign information is reliable in few-shot adaptation, and how sparsity should be coupled with the update rule.

\begin{figure*}[t]
\centering
\includegraphics[width=\textwidth]{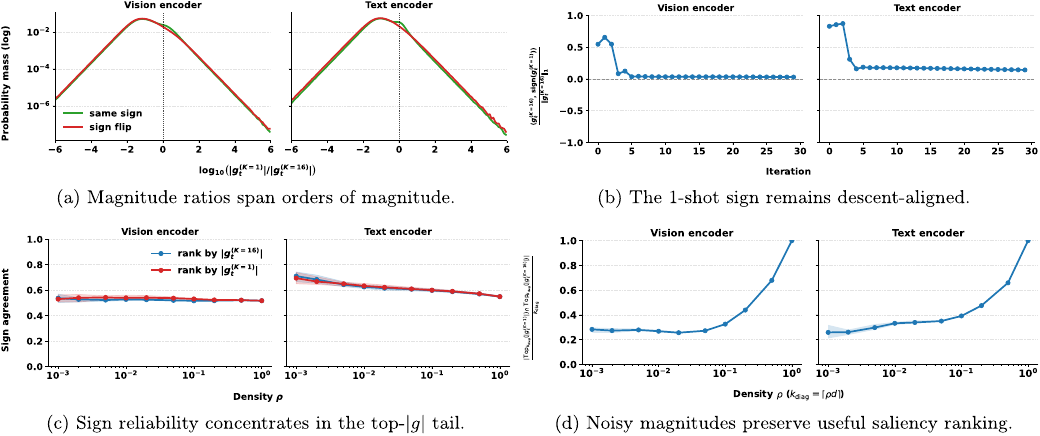}
%\caption{Few-shot gradient diagnostics for CLIP (ViT-B/16) on EuroSAT, comparing a 1-shot gradient with a 16-shot reference at matched iterates. Magnitudes are unreliable for scaling, while signs and top-magnitude ranking retain useful information. The supplementary material gives definitions, additional datasets, seeds, and checkpoints.}
\caption{Few-shot gradient noise in CLIP adaptation (ViT-B/16, EuroSAT, $K{=}1$ vs $K{=}16$; $K$ is the number of shots, $k_{\mathrm{diag}}=\lceil\rho d\rceil$ is the global diagnostic active-set size at nominal density $\rho$, and $d$ is the number of trainable coordinates).}
\label{fig:gradient_diagnostics}
\end{figure*}

To answer this question (\textcolor{blue}{\hyperlink{suppnav.noise}{Appendix~I}} provides further results and details), we assess the few-shot gradient noise by computing two gradients: a noisy 1-shot full-support gradient $g_t^{(K=1)}$ and a 16-shot full-support reference gradient $g_t^{(K=16)}$, evaluated on the same set of classes at each iteration $t$ for all parameters of CLIP (ViT-B/16). Fig.~\ref{fig:gradient_diagnostics}(a) shows that gradient magnitudes are highly unreliable, even when the sign matches. The coordinate-wise ratio $|g_{t,i}^{(K=1)}|/|g_{t,i}^{(K=16)}|$ spans orders of magnitude, indicating that magnitude noise can severely distort step sizes when raw gradients are used. In contrast, the gradient sign contains a descent-relevant signal, but not uniformly across coordinates. Fig.~\ref{fig:gradient_diagnostics}(b) reports a descent-relevant, magnitude-weighted sign reliability metric based on the alignment between $g_t^{(K=16)}$ and $\sgn\!\left(g_t^{(K=1)}\right)$. This metric remains positive over iterations, suggesting that a sign step driven by the 1-shot gradient is often aligned with a descent direction for the reference objective. Fig.~\ref{fig:gradient_diagnostics}(c) shows that sign agreement increases substantially when restricting to the top-$|g|$ tail (e.g., on EuroSAT from $\approx 0.58$ globally to $\approx 0.71$ on the top $0.1\%$ coordinates), and that selecting these coordinates using $|g_t^{(K=1)}|$ roughly tracks the oracle ranking by $|g_t^{(K=16)}|$. Finally, Fig.~\ref{fig:gradient_diagnostics}(d) suggests that noisy magnitudes remain informative as a saliency signal. The overlap between $\Top_{k_{\mathrm{diag}}}(|g_t^{(K=1)}|)$ and $\Top_{k_{\mathrm{diag}}}(|g_t^{(K=16)}|)$ is always above $20\%$ even for a gradient sparsity ratio $99.9\%$. 

These observations suggest a separation between \emph{saliency} and \emph{step scaling}. We use the gradient magnitude only to select a small blockwise active set, then update that set with fixed-magnitude signs. \textit{SignSparseSGD} fixes the support selected from the first minibatch. \textit{DynSignSparseSGD} refreshes it at every iteration. The dynamic variant can refine which parameters to update as training progresses to increase the learning capacity. Both methods maintain no first- or second-moment buffers. Unlike sparse Adam-style methods such as SAFT~\cite{nguyen2024saft} and SO~\cite{mrabah2025so}, they use neither moment-based scaling nor moment-based support retention. However, the proposed optimizers still require dense backpropagation, so their efficiency benefits occur at the optimizer stage rather than in backpropagation.

\textbf{Contributions.} \noindent\textbf{(1)} We show that gradient magnitude and gradient sign have different reliability profiles in few-shot adaptation. Although coordinate-wise magnitudes are too noisy for step scaling, their ranking remains informative for identifying salient parameters. The most salient coordinates also exhibit more reliable signs. These findings motivate separating \emph{where} to update from \emph{how} to update. \noindent\textbf{(2)} We introduce two state-free sign-sparse optimizers that implement this separation. SignSparseSGD selects a support set from the first minibatch and keeps it fixed throughout training. DynSignSparseSGD recomputes the support at every iteration to increase learning capacity. %, which allows the active coordinates to change as optimization progresses. 
Both methods update only the selected parameters using fixed-magnitude gradient signs. \noindent\textbf{(3)} A comprehensive evaluation of CLIP few-shot adaptation over 11 datasets, three backbones, five shot levels, and three random seeds. Under one shared configuration, DynSignSparseSGD yields the strongest performance in average accuracy across all backbone–shot settings. Both variants also resist long-horizon overfitting and transfer the training budget more reliably across datasets.

%\textbf{Contributions.} \noindent\textbf{(1)} We show that few-shot gradient magnitude and sign have different levels of reliability. Further diagnostic accounts of why few-shot gradient magnitudes remain useful for saliency ranking but are unreliable for coordinate-wise step scaling. These results motivate decoupling parameter selection from step scaling. \noindent\textbf{(2)} We introduce two state-free sign-sparse optimizers that use magnitude only for support selection and fixed signs for the update. These designs eliminate the moment terms while mitigating the noisy gradient magnitudes. \noindent\textbf{(3)} A comprehensive evaluation of CLIP few-shot adaptation over 11 datasets, three backbones, five shot levels, and three fixed seeds. Under one shared configuration for each optimization-based method, DynSignSparseSGD yields the strongest performance on average accuracy across all backbone–shot settings. Both variants also resist long-horizon overfitting and transfer the training budget more reliably across datasets. 

\section{Related Work}
\label{sec:related_work}

\paragraph{Parameter-efficient adaptation.}
Prompt methods, such as CoOp~\cite{zhou2022coop}, KgCoOp~\cite{yao2023kgcoop}, ProGrad~\cite{zhu2022prograd}, learn textual contexts instead of hand-crafted text template. %Adapter and cache methods add lightweight modules or support-feature memories (CLIP-Adapter~\cite{gao2024clipadapter}, TIP-Adapter~\cite{zhang2022tipadapter}, LP++~\cite{huang2024lpplusplus}). 
TIP-Adapter-F~\cite{zhang2022tipadapter} learns cache-key weights while keeping the CLIP encoders frozen. Training-free baselines build classifiers from frozen support features (GDA~\cite{wang2024hard}, ProKeR~\cite{bendou2025proker}). Low-rank reparametrization methods optimize factorized updates such as LoRA~\cite{hu2022lora}. Selection-based methods tune a prescribed subset, including LayerNorm parameters~\cite{ba2016layer} or the support-gradient mask of SAFT~\cite{nguyen2024saft}. These methods reduce the trainable set of parameters, but their performance can be sensitive to rank, prompt length, regularization, schedule, or validation-dependent choices~\cite{mrabah2025so,silva2024closer}.

\paragraph{Optimization-efficient training.}
GaLoRE~\cite{zhao2024galore} compresses gradients through periodically updated low-rank projections. FLoRA~\cite{hao2025flora} uses random low-rank projections and relates adapters to gradient compression. These methods were designed mainly for large language models. SO~\cite{mrabah2025so} targets few-shot VLM adaptation. It samples candidate coordinates, retains entries with large first moments, and maintains sparse Adam states. Our methods remove the moment buffers and their decay hyperparameters. Gradient magnitude determines only the active support, not the coordinate-wise step size. All these approaches still require dense backpropagation. Therefore, their savings concern optimizer storage rather than activation memory or gradient computation in backpropagation.

\begin{figure*}[t]
\centering
\includegraphics[width=\textwidth]{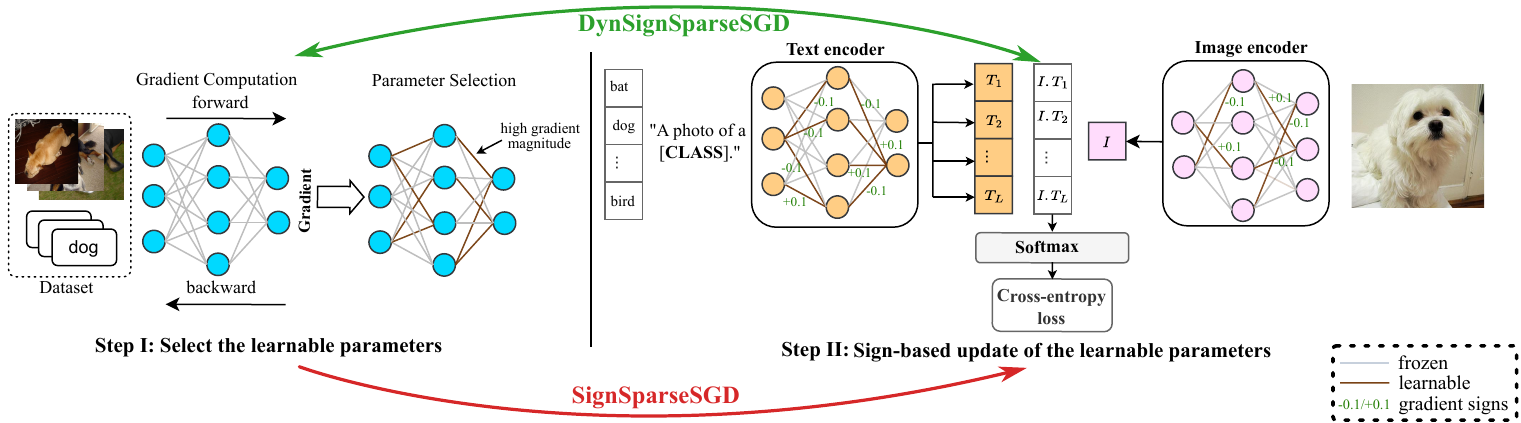}
\caption{The proposed optimizers first use the dense minibatch gradient to select a blockwise top-$|g|$ support and then update only those coordinates by fixed-magnitude signs. SignSparseSGD fixes the support after the first minibatch. DynSignSparseSGD refreshes it at every iteration. Both adapt the original CLIP parameters without adding task-specific modules.}
%\caption{{\color{myred}SignSparseSGD} and {\color{mygreen}DynSignSparseSGD} for few-shot adaptation of CLIP models. Gradients computed on the support set are used to select a sparse subset of learnable parameters (Step~\uppercase\expandafter{\romannumeral 1\relax}). These are updated with fixed-magnitude sign steps while the remaining weights are frozen (Step~\uppercase\expandafter{\romannumeral 2\relax}). {\color{myred}SignSparseSGD} uses a static sparse mask, whereas {\color{mygreen}DynSignSparseSGD} recomputes the mask at every iteration.}
\label{fig:method}
\end{figure*}

\paragraph{Sign-based optimization.}
SignSGD~\cite{bernstein2018signsgd} updates every coordinate with a fixed sign step. SignSGD with majority vote~\cite{bernstein2019signsgd} extends this rule to distributed learning to reduce communication. %relative to full-precision gradient exchange. 
However, dense sign updates perform poorly in our few-shot CLIP experiments (Fig.~\ref{fig:ablation} and detailed results in \textcolor{blue}{\hyperlink{suppnav.optimizer-comparison}{Appendix~O}}). \textcolor{blue}{\hyperlink{suppnav.update-normalization}{Appendix~L.2}} analyzes this failure. Dense fixed-magnitude steps induce excessive feature drift and amplify low-signal coordinates whose signs frequently flip. Our methods instead use gradient magnitudes to select a small salient support. All remaining coordinates receive zero update. This differs from dense SignSGD, which applies sign updates without sparse support selection. Worker aggregation belongs to its distributed majority-vote extension. It also differs from SAFT, which uses a fixed mask with Adam, and SO, which uses moment-based support retention. Sec.~\ref{sec:experiments} isolates the effects of sign, sparsity, support refresh, clipping, and gradient normalization.

\section{Proposed Sign-Sparse Optimization}
\label{sec:method}

\paragraph{Setup and notation.}
We adapt a pretrained CLIP model from a support set $D=\{z_{j_{\mathrm{data}}}=(x_{j_{\mathrm{data}}},y_{j_{\mathrm{data}}})\}_{j_{\mathrm{data}}=1}^{n}$, where $x_{j_{\mathrm{data}}}$ is an image and $y_{j_{\mathrm{data}}}\in\{1,\ldots,N\}$ is its class label. The set contains $K$ labeled images for each of $N$ classes, so $n=NK$. A disjoint query set is used only for evaluation. Let $\Theta\in\R^d$ concatenate the $d$ trainable parameters, and let $\Theta_t$ be the parameter vector at iteration $t$. For a minibatch $B_t\subseteq D$ of size $b_t=|B_t|$, the empirical risk and gradient are
\begin{equation}
 \mathcal L_{B_t}(\Theta)=\frac{1}{b_t}\sum_{z\in B_t}\ell(\Theta;z),
 \qquad
 g_t=\nabla_{\Theta}\mathcal L_{B_t}(\Theta_t)\in\R^d,
 \label{eq:minibatch_gradient_method}
\end{equation}
where $\ell(\Theta;z)$ is the cross-entropy loss and $g_t$ is the dense minibatch gradient.

\paragraph{Step I: blockwise parameter selection.}
The coordinate set $\{1,\ldots,d\}$ is partitioned by the $R$ trainable parameter tensors. Let $\{\mathcal I_r\}_{r=1}^{R}$ be this disjoint partition and $d_r=|\mathcal I_r|$. For a nominal density $\rho\in(0,1]$, the block budgets and selected support are
\begin{align}
 k_r&=\max\{1,\lfloor\rho d_r\rfloor\}, &
 S_{t,r}&=\Top_{k_r}(|g_{t,\mathcal I_r}|), \nonumber\\[-2pt]
 k_{\mathrm{step}}&=\sum_{r=1}^{R}k_r, &
 S_t&=\bigcup_{r=1}^{R}S_{t,r}.
 \label{eq:block_budgets}
\end{align}
Here, $g_{t,\mathcal I_r}$ is the restriction of $g_t$ to block $\mathcal I_r$, and $\Top_k(v)$ returns the indices of the $k$ largest entries of $v$ under a fixed tie rule. For a vector restricted to $\mathcal I_r$, these are the original coordinate indices in $\mathcal I_r$, so $S_{t,r}\subseteq\mathcal I_r$. Absolute values are elementwise. For any $S_{t}\subseteq\{1,\ldots,d\}$, let $m_{t}\in\{0,1\}^d$ be its mask. We set
\begin{equation}
 (m_t)_i=\1\{i\in S_{t}\},
 \;\;\;\; \widetilde g_t=m_t\odot g_t,
 \label{eq:sparse_gradient}
\end{equation}
where $\1\{\cdot\}$ is the indicator function and $\odot$ is elementwise multiplication. The per-block quota prevents one high-scale tensor from consuming the full budget. Because every block receives at least one coordinate, the realized density $k_{\mathrm{step}}/d$ may differ slightly from $\rho$. Equation~\eqref{eq:block_budgets} gives the refreshed supports used by DynSignSparseSGD. SignSparseSGD fixes the initial supports and uses $S_{t,r}=S_{0,r}$, $S_t=S_0$, and $m_t=m_0$ at every iteration.

\paragraph{Step II: fixed-magnitude sign update.}
The elementwise map $\sgn(a)$ equals $1$ for $a>0$, $-1$ for $a<0$, and $0$ for $a=0$. Let $\eta>0$ be the base learning rate and let $\alpha=0.1$ be the fixed sign multiplier used in all experiments. With $\lambda=\alpha \,\eta$, the signed sparse direction $s_t$ and update are
\begin{equation}
 s_t=\sgn(\widetilde g_t)=m_t\odot\sgn(g_t),
 \qquad
 \Theta_{t+1}=\Theta_t-\lambda \, s_t.
 \label{eq:sparse_sign_update}
\end{equation}
Each selected coordinate with a nonzero gradient moves by exactly $\lambda$. All other coordinates remain unchanged. Therefore, magnitude determines \emph{where} to update, while sign determines \emph{how} to move on the selected support.

\textcolor{blue}{Appendix~\ref{subsec:appE_sparsity_learning_rate}} shows that $\|\Theta_{t+1}-\Theta_t\|_2\le\lambda\sqrt{k_{\mathrm{step}}}$ and that, when average selected-coordinate curvature is comparable across densities, the explicit gap--curvature envelope gives $\lambda\propto k_{\mathrm{step}}^{-1/2}$. This law motivates sparsity-aware scaling. \textcolor{blue}{Appendix~\ref{sec:update_scale_measure}} provides evidence for calibrating $\alpha=0.1$ at $\rho=0.1\%$ and shows that $\alpha=1$ is overly large.

\begin{algorithm}[tb]
\caption{SignSparseSGD and DynSignSparseSGD}
\label{alg:signsparse}
\small
\begin{algorithmic}[1]
\REQUIRE Initial parameters $\Theta_0$, learning rate $\eta$, density ratio $\rho$, sign scale $\alpha$, block-specific parameter indices $\{\mathcal I_r\}_{r=1}^{R}$, iterations $T$, mode $q\in\{\textsc{static},\textsc{dynamic}\}$
\STATE $k_r\leftarrow\max\{1,\lfloor\rho \,|\mathcal I_r |\rfloor\}$ for all $r$
\FOR{$t=0,1,\ldots,T-1$}
  \STATE $g_t\leftarrow\nabla_{\Theta}\mathcal L_{B_t}(\Theta_t)$
  \IF{$q=\textsc{dynamic}$ or $t=0$}
    \STATE $S\leftarrow\bigcup_{r=1}^{R}\Top_{k_r}(|g_{t,\mathcal I_r}|)$
    \STATE $(m)_i\leftarrow\1\{i\in S\}$
  \ENDIF
  \STATE $\Theta_{t+1}\leftarrow\Theta_t-\alpha\,\eta\,m\odot\sgn(g_t)$
\ENDFOR
\end{algorithmic}
\end{algorithm}

\paragraph{Static versus dynamic support.}
SignSparseSGD uses $q=\textsc{static}$. It constructs $S_0$ from the first minibatch and uses $S_t=S_0$ and $m_t=m_0$ for all $T$ updates. Algorithm~\ref{alg:signsparse} writes the current support and mask as $S$ and $m$. DynSignSparseSGD uses $q=\textsc{dynamic}$ and reconstructs $S_t$ from the current gradient at every iteration. The two methods have the same per-step budget $k_{\mathrm{step}}$. Dynamic refresh lets coordinates enter and leave the active set as saliency changes. The static variant performs sign descent in one fixed coordinate subspace.

\begin{table*}[t]
\centering
\small
\setlength{\tabcolsep}{0.75mm}
\begin{tabular}{lrrrrrrrrrrrr}
\toprule
Method & IN & SUN & Air & Euro & Cars & Food & Pets & Flwr & Cal & DTD & UCF & Avg.\\
\midrule
CLIP & 66.7 & 62.6 & 23.9 & 42.2 & 65.5 & 85.2 & 88.3 & 67.4 & 93.3 & 44.4 & 65.2 & 64.1\\
LayerNorm & \textbf{68.7} & 66.7 & 25.5 & 49.6 & 66.7 & 85.7 & 89.4 & 69.9 & 93.5 & 45.9 & 67.8 & 66.3\\
SAFT & 66.1 & 67.2 & 27.4 & 68.9 & 64.8 & 84.1 & 89.9 & 80.7 & \underline{93.9} & 52.2 & 73.7 & 69.9\\
CoOp & 67.3 & 67.0 & 28.1 & 54.3 & 66.7 & 84.7 & 89.3 & 75.6 & 93.1 & 47.6 & 72.3 & 67.8\\
KgCoOp & \underline{68.6} & 67.5 & 25.7 & 57.4 & 65.0 & \underline{86.1} & 90.9 & 69.9 & 93.4 & 47.4 & 71.0 & 67.5\\
ProGrad & 65.4 & 65.7 & 24.0 & 56.1 & 63.6 & 83.2 & 89.2 & 75.3 & 92.0 & 48.0 & 71.5 & 66.7\\
%CLIP-Adapter & 67.6 & 64.3 & 24.1 & 45.6 & 66.2 & 85.4 & 89.2 & 68.6 & 93.4 & 45.0 & 66.3 & 65.1\\
%LP++ & 66.3 & 66.6 & 26.7 & 40.8 & 62.6 & 82.9 & 86.1 & 81.6 & 93.3 & 50.9 & 71.1 & 66.3\\
TIP-Adapter-F & 66.7 & 62.6 & 23.9 & 43.1 & 65.6 & 85.2 & 88.3 & 67.6 & 93.3 & 44.4 & 65.3 & 64.2\\
GDA & 67.0 & 65.0 & 24.4 & 46.9 & 66.4 & 85.3 & 88.3 & 75.1 & 93.3 & 49.5 & 66.5 & 66.2\\
ProKeR & 67.7 & 65.9 & 27.3 & 57.0 & 68.3 & 85.2 & 88.4 & 77.2 & 93.6 & 49.8 & 70.0 & 68.2\\
LoRA & 68.1 & \underline{69.6} & 28.9 & 65.7 & 69.7 & \textbf{86.2} & \underline{92.4} & 74.0 & \textbf{94.1} & 52.3 & 74.7 & 70.5\\
GaLoRE & 64.1 & 67.0 & 27.7 & 68.6 & 62.4 & 81.6 & 90.1 & 82.7 & 92.7 & 52.1 & 75.0 & 69.5\\
SO & 68.5 & 69.4 & \underline{30.0} & 69.6 & \underline{70.3} & 85.9 & \textbf{92.5} & 76.1 & 93.8 & 53.6 & 75.2 & 71.4\\
\rowcolor{LightGray} SignSparseSGD & 66.8 & 68.2 & 29.4 & \underline{71.2} & 69.1 & 83.8 & 91.3 & \underline{83.3} & 93.7 & \underline{54.3} & \underline{75.7} & \underline{71.5}\\
\rowcolor{LightGray} DynSignSparseSGD & 68.1 & \textbf{69.8} & \textbf{30.6} & \textbf{73.7} & \textbf{70.6} & 85.1 & \textbf{92.5} & \textbf{84.7} & 93.8 & \textbf{54.8} & \textbf{76.2} & \textbf{72.7}\\
\bottomrule
\end{tabular}
\caption{A2A 1-shot top-1 accuracy (\%) on 11 datasets with CLIP (ViT-B/16), averaged over three fixed random seeds.}
\label{tab:a2a_1shot}
\end{table*}

\paragraph{Optimizer state and scope.}
SAFT applies Adam on a fixed global mask. SO uses random candidates and moment-based support retention. Our methods instead select directly from blockwise gradient magnitudes and store no first- or second-moment buffers. %\textcolor{blue}{Appendix~E} separates the shared dense forward--backward cost from post-backpropagation overhead and quantifies wall-clock training time and idealized optimizer memory. 
\textcolor{blue}{\hyperlink{suppnav.geometry}{Appendix~F}} develops the blockwise steepest-descent geometry and trajectory-wise population bounds for both optimizers, exposing the roles of sparsity, curvature, finite-shot bias, minibatch noise, and static support miss. \textcolor{blue}{\hyperlink{suppnav.confidence}{Appendix~G}} interprets sparsification as confidence gating and identifies conditional regimes in which suppressing unreliable signs improves expected descent relative to dense SignSGD and Adam-style weighting. \textcolor{blue}{\hyperlink{suppnav.ranking}{Appendix~H}} separates magnitude calibration from ranking stability through gap-free selection-regret bounds and sufficient conditions for exact recovery of the population blockwise top-$k$ support.

%\paragraph{Optimizer state and scope.}
%SAFT applies Adam on a fixed global mask. SO uses random candidates and moment-based support retention. Our methods instead select directly from blockwise gradient magnitudes and store no first- or second-moment buffers. %\textcolor{blue}{Appendix~\ref{supp-sec:appX}} further analyzes the theoretical similarities between these methods and the proposed ones.

%Both variants still compute the dense gradient in Eq.~\eqref{eq:minibatch_gradient_method}. Therefore, both variants do not reduce activation memory or backpropagation cost. Their savings occur after backpropagation, where the update requires only the active indices and signs. The supplementary material provides implementation details and sparsity--step calibration.
\section{Results and Discussion}
\label{sec:experiments}

\paragraph{Protocol.} We evaluate VLM few-shot adaptation on 11 datasets: ImageNet (IN)~\cite{imagenet}, SUN397 (SUN)~\cite{sun397}, FGVC-Aircraft (Air)~\cite{aircraft}, EuroSAT (Euro)~\cite{eurosat}, Stanford Cars (Cars)~\cite{stanford_cars}, Food101 (Food)~\cite{food101}, Oxford Pets (Pets)~\cite{oxford_pets}, Flowers102 (Flwr)~\cite{flowers102}, Caltech101 (Cal)~\cite{caltech101}, DTD~\cite{dtd}, and UCF101 (UCF)~\cite{ucf101}. CLIP is used as a representative VLM with 3 vision backbones: ViT-B/32, ViT-B/16, and ViT-L/14. 

The experimental protocol covers two few-shot settings. In \emph{all-to-all} (A2A), we sample $K\!\in\!\{1,2,4,8,16\}$ labeled images per class from the full label set to form the support set and report top-1 accuracy on the test split. In \emph{base-to-new} (B2N), we generate one fixed class partition per dataset--seed pair by shuffling class labels and assigning half of the classes to \emph{Base} and the other half to \emph{New} ($50{:}50$). Each partition is saved and reused by all methods and backbones. During training, only \emph{Base} classes construct the support set. Evaluation is performed separately on \emph{Base} and \emph{New} classes, and HM denotes the harmonic mean of the two accuracies. All experiments use three fixed random seeds with identical support and query sets across methods.

The proposed optimizers are compared with representative baselines across different families: (i) Prompt-tuning: CoOp, KgCoOp, ProGrad; (ii) classifier adaptation: TIP-Adapter-F (trainable), GDA and ProKeR (training-free); (iii) Low-rank reparameterization: LoRA; (iv) Selection-based: LayerNorm, SAFT; (v) Optimization-based: GaLoRE, SO. For the ablation study, we introduce sign- and sparsity-based variants of Adam and SGD: SignAdam, SignSGD, SparseAdam (equivalent to SAFT), SparseSGD, and SignSparseAdam. We also include a comparison with stabilization techniques such as gradient clipping and normalization: GradNorm SignSGD, GradNorm SparseSGD, Local-clip SparseSGD, and Global-clip SparseSGD. Algorithms and implementation details of these optimizers are provided in \textcolor{blue}{\hyperlink{suppnav.algorithms}{Appendix~C}}.

For a fair comparison, the hyperparameters of the trainable methods are fixed across seeds, backbones, datasets, and shot levels, without dataset-specific tuning. The proposed methods and their ablation variants use a common configuration of 30 epochs, a batch size of 64, and a learning rate of $10^{-4}$. Other training-based methods follow the configuration of their official implementations, while GDA and ProKeR use their official training-free procedures. We highlight that several training-free methods leverage validation sets to fix the hyperparameters. Baselines and their limitations are discussed in \textcolor{blue}{\hyperlink{suppnav.related}{Appendix~B}}. Complete configurations of all methods are reported in \textcolor{blue}{\hyperlink{suppnav.hyperparameters}{Appendix~D}}. All experiments use the same hardware and software environments.  \textcolor{blue}{\hyperlink{suppnav.efficiency}{Appendix~E}} analyzes the training time and memory for the proposed optimizers and baselines, including the introduced sign‑ and sparsity‑based variants. \textcolor{blue}{\hyperlink{suppnav.sensitivity}{Appendix~M}} studies the sensitivity of our optimizers to the density ratio $\rho$.

\begin{table}[!htbp]
\centering
\small
\setlength{\tabcolsep}{2.0pt}
\begin{tabular}{llrrrrr}
\toprule
Backbone & Method & 1 & 2 & 4 & 8 & 16 \\
\midrule
\multirow{3}{*}{ViT-B/32}
 & SO & 66.7 & 69.1 & 71.5 & 75.2 & 78.5 \\
 & SignSparse & 66.4 & 69.6 & 71.8 & 74.4 & 77.4 \\
 & DynSignSparse & \textbf{67.9} & \textbf{71.0} & \textbf{73.4} & \textbf{76.2} & \textbf{79.0} \\
\midrule
\multirow{3}{*}{ViT-B/16}
 & SO & 71.4 & 74.2 & 76.6 & 79.5 & 82.2 \\
 & SignSparse & 71.5 & 74.2 & 76.6 & 78.8 & 81.5 \\
 & DynSignSparse & \textbf{72.7} & \textbf{75.4} & \textbf{77.8} & \textbf{80.0} & \textbf{82.8} \\
\midrule
\multirow{3}{*}{ViT-L/14}
 & SO & 77.5 & 79.8 & 81.9 & 84.4 & 86.6 \\
 & SignSparse & 77.8 & 80.6 & 82.3 & 84.4 & 86.4 \\
 & DynSignSparse & \textbf{78.5} & \textbf{81.1} & \textbf{83.1} & \textbf{85.1} & \textbf{87.1} \\
\bottomrule
\end{tabular}
\caption{A2A average top-1 accuracy (\%) over 11 datasets and 3 seeds for the strongest baseline and the proposed methods. Columns are the number of shots per class.}
\label{tab:a2a_summary}
\end{table}

\begin{figure*}[!t]
\centering
\includegraphics[width=\textwidth]{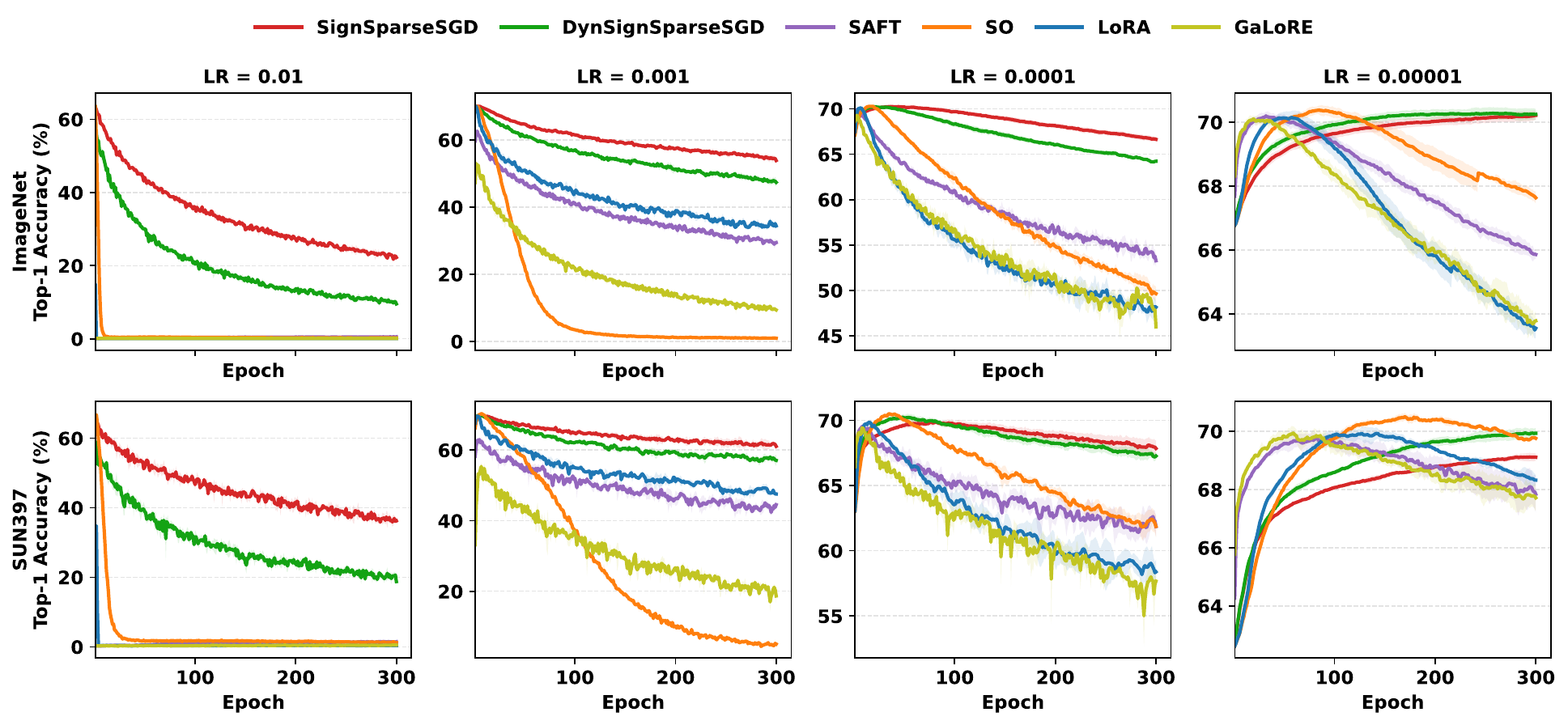}
%\caption{Test accuracy for 1-shot CLIP (ViT-B/16) adaptation on ImageNet and SUN397 under four learning rates. SignSparseSGD and DynSignSparseSGD degrade substantially less than LoRA, GaLoRE, SAFT, and SO under long training.}
\caption{Test accuracy for 1-shot CLIP (ViT-B/16) adaptation on ImageNet and SUN397 under four learning rates.} %SignSparseSGD and DynSignSparseSGD degrade substantially less than LoRA, GaLoRE, SAFT, and SO under long training.}
\label{fig:long_horizon}
\end{figure*}

\begin{figure*}[!t]
\centering
\includegraphics[width=0.78\textwidth]{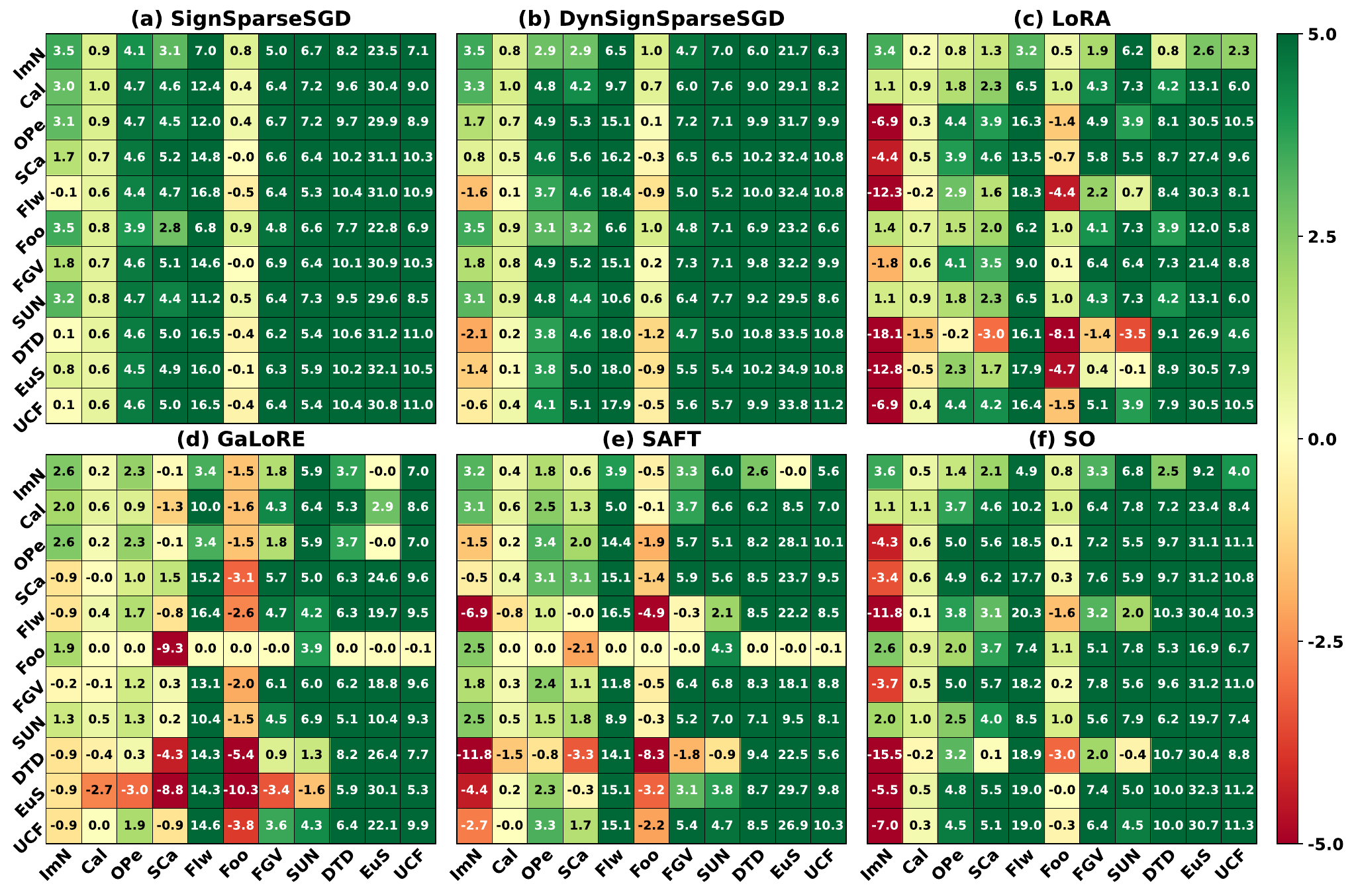}
\caption{Cross-dataset transfer of the selected training epoch in 1-shot adaptation with CLIP (ViT-B/16). Each entry is the test accuracy change relative to zero-shot CLIP when the epoch chosen on the row dataset is reused on the column dataset.} %Sign-sparse methods exhibit more reliable off-diagonal transfer.}
\label{fig:epoch_transfer}
\end{figure*}

\begin{figure*}[!t]
\centering
\includegraphics[width=\textwidth]{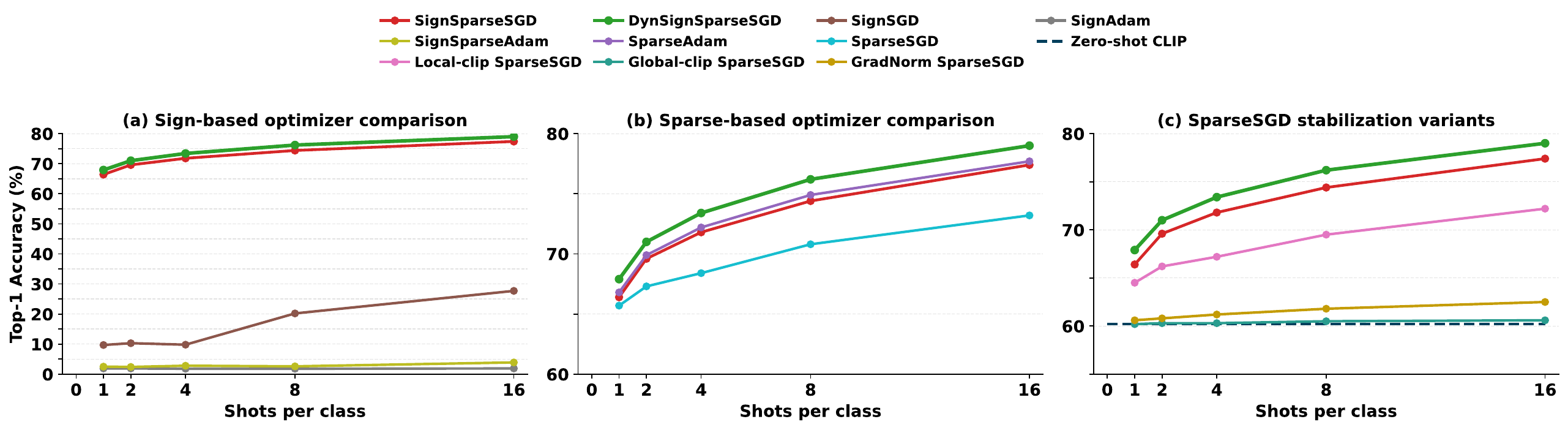}
\caption{Mean accuracy across 11 datasets with CLIP (ViT-B/32). Panels plot accuracy vs.\ shots per class: (a) sign-based optimizers: dense sign methods collapse, (b) sparse-based optimizers: sparsity alone does not make Adam/SGD recover the sign-sparse gains, (c) stabilization methods: clipping and gradient normalization do not recover the sign-sparse gains.}
\label{fig:ablation}
\end{figure*}

\paragraph{Main comparison.}
\emph{All-to-all (A2A).} Table~\ref{tab:a2a_1shot} reports the most data-constrained A2A setting with ViT-B/16. DynSignSparseSGD obtains the highest mean accuracy at 72.7\%, improving over SO by 1.3 points and LoRA by 2.2 points. It leads on seven datasets and ties the best Oxford Pets result. SignSparseSGD reaches 71.5\%, slightly above SO. Table~\ref{tab:a2a_summary} extends the comparison to all 15 backbone--shot combinations. DynSignSparseSGD has the highest mean in every setting and exceeds SO by 0.5--1.9 points. Its mean accuracy rises from 67.9/72.7/78.5 at 1 shot to 79.0/82.8/87.1 at 16 shots for ViT-B/32, ViT-B/16, and ViT-L/14, respectively. SignSparseSGD remains competitive but can trail at larger shot counts, which is consistent with a fixed initial support becoming less representative as optimization progresses. Complete per-dataset A2A tables with standard deviations are provided in \textcolor{blue}{\hyperlink{suppnav.peft-a2a}{Appendix~J.1}}.

\emph{Base-to-new (B2N).} Table~\ref{tab:b2n_16shot} reports the 16-shot B2N setting with ViT-B/16. DynSignSparseSGD obtains the highest \emph{Base}, \emph{New}, and HM accuracies at 88.1\%, 75.3\%, and 80.8\%. In the complete comparison of \textcolor{blue}{\hyperlink{suppnav.peft-b2n}{Appendix~J.2}}, its gains over the strongest baseline in each metric are 0.7 points on \emph{Base} over SO, 0.5 points on \emph{New} over LayerNorm, and 1.1 points in HM over BitFit. SignSparseSGD reaches 79.5\% HM, so support refresh adds 1.3 points. \textcolor{blue}{\hyperlink{suppnav.peft-b2n}{Appendix~J.2}} reports all five shot levels for ViT-B/32 and ViT-B/16. DynSignSparseSGD has the highest average HM in all 10 settings. Its HM increases from 72.8\% to 76.6\% on ViT-B/32 and from 76.9\% to 80.8\% on ViT-B/16, while its \emph{New}-class accuracy varies by less than one point across shots for each backbone.

\begin{table}[!htbp]
\centering
\small
\setlength{\tabcolsep}{4.2pt}
\renewcommand{\arraystretch}{0.96}
\begin{tabular}{lrrr}
\toprule
Method & Base & New & HM \\
\midrule
CLIP & 64.1 & 64.1 & 64.1 \\
LayerNorm & 79.3 & \underline{74.8} & 76.8 \\
SAFT & 87.2 & 74.1 & \underline{79.7} \\
CoOp & 85.2 & 69.0 & 75.5 \\
KgCoOp & 79.6 & 72.8 & 75.7 \\
ProGrad & 83.2 & 43.9 & 54.5 \\
LoRA & 87.1 & 35.0 & 46.7 \\
GaLoRE & 86.7 & 73.2 & 78.9 \\
SO & \underline{87.5} & 73.8 & 79.6 \\
\cmidrule(lr){1-4}
\rowcolor{LightGray} SignSparseSGD & 87.1 & 73.8 & 79.5 \\
\rowcolor{LightGray} DynSignSparseSGD & \textbf{88.1} & \textbf{75.3} & \textbf{80.8} \\
\bottomrule
\end{tabular}
\caption{B2N 16-shot Base, New, and harmonic-mean (HM) top-1 accuracy (\%) with CLIP (ViT-B/16), averaged over 11 datasets and three fixed random seeds.}
\label{tab:b2n_16shot}
\end{table}

\paragraph{Overfitting and budget transfer.}
Figure~\ref{fig:long_horizon} tests a much longer 300-epoch horizon. At the shared learning rate $10^{-4}$, the average peak-to-final drops over all 11 datasets are 0.93 and 1.85 points for SignSparseSGD and DynSignSparseSGD, versus 5.26 for SO, 6.06 for SAFT, 7.97 for LoRA, and 9.58 for GaLoRE. Our optimizers remain within five points of their peak on 11/11 and 10/11 datasets, respectively.

A validation-free method should also tolerate a training budget selected without target-dataset feedback. Figure~\ref{fig:epoch_transfer} selects the best epoch on one few-shot dataset (row) and reuses it on another (column). SignSparseSGD and DynSignSparseSGD yield predominantly positive off-diagonal changes relative to zero-shot CLIP. LoRA, GaLoRE, SAFT, and SO have substantially more negative cells, so a transient peak on one dataset is a poor universal stopping rule for those methods.

\paragraph{Effect of sign and sparse update.}
Fig.~\ref{fig:ablation} compares our optimizers with sign-only (SignAdam, SignSGD), sparse-only (SparseAdam, SparseSGD), and SignSparseAdam. The comparison also includes SparseSGD with stabilization techniques such as gradient normalization and clipping. First, dense sign methods collapse. In particular, SignAdam and SignSGD remain far below zero-shot even at 16-shot. These results indicate that sign-only updates applied to all coordinates are unstable. \textcolor{blue}{\hyperlink{suppnav.update-normalization}{Appendix~L.2}} analyzes in depth the collapse of SignSGD. Adding sparsity to SignAdam does not fix this collapse. Second, sparse updates for Adam and SGD yield strong performance. Compared with SparseAdam (SAFT), DynSignSparseSGD shows the best average results across shot levels, while SignSparseSGD remains competitive. Third, standard stabilization does not recover the gains of the proposed sign-sparse updates. Among the SparseSGD stabilization variants, local clipping is the strongest, but it remains consistently below SignSparseSGD and DynSignSparseSGD across shot levels and backbones. In contrast, global clipping and gradient normalization stay much closer to the zero-shot baseline. This suggests that the gain does not come only from stabilizing SparseSGD, but from combining sparsity with fixed-magnitude sign updates. Detailed results for the sign-based and sparse-based comparisons are reported in \textcolor{blue}{\hyperlink{suppnav.optimizer-comparison}{Appendix~O}}, and detailed results for the stabilization variants are reported in \textcolor{blue}{\hyperlink{suppnav.clipping}{Appendix~P}}. A separate matched-momentum comparison with a common static mask, coordinate step, and input stream is reported in \textcolor{blue}{\hyperlink{suppnav.optimizer-momentum}{Appendix~O.3}}.

\begin{figure}[!htbp]
\centering
\includegraphics[width=0.98\columnwidth]{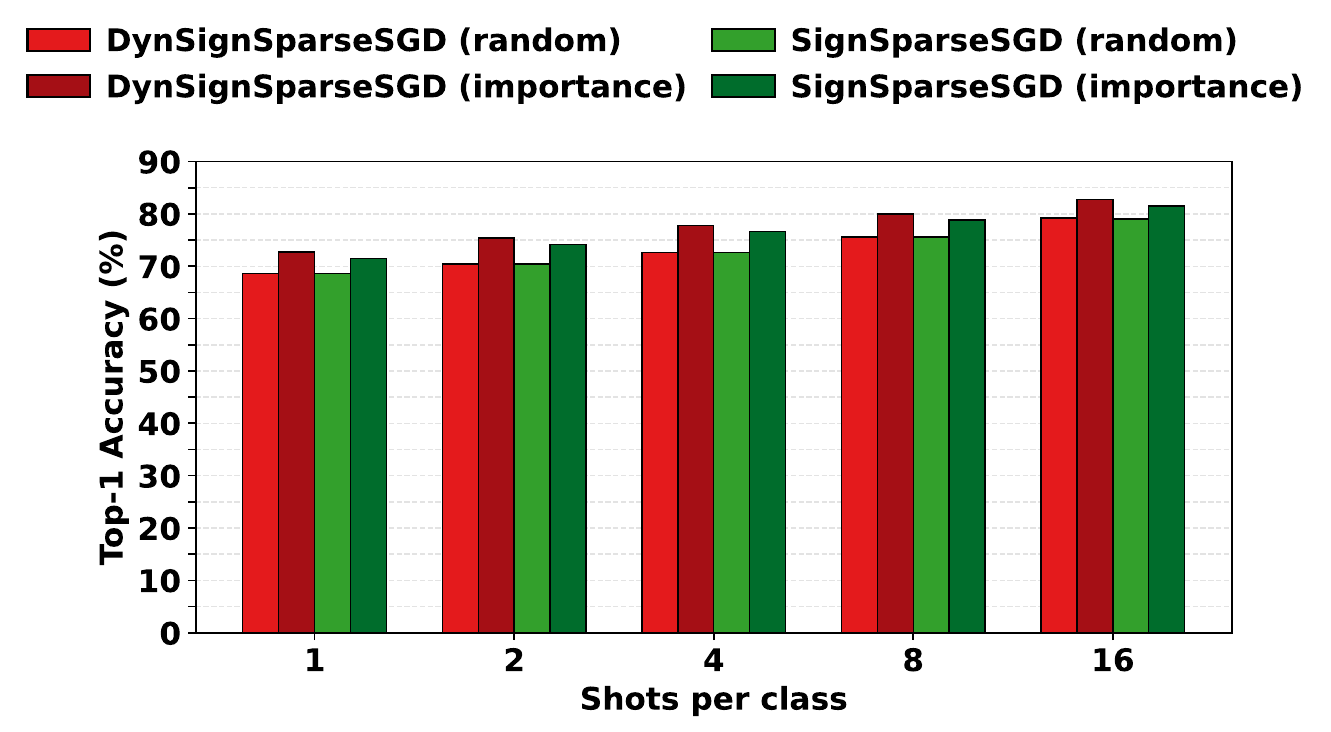}
\caption{Average top-1 accuracy over 11 datasets with CLIP (ViT-B/16): random versus top-$|g|$ coordinate selection.}
\label{fig:importance_selection}
\end{figure}

\begin{table}[!htbp]
\centering
\scriptsize
\setlength{\tabcolsep}{2.1pt}
\renewcommand{\arraystretch}{1.05}
\begin{tabular}{llrrrrrr}
\toprule
Method & Backbone & 1 & 2 & 4 & 8 & 16 & Overall \\
\midrule
\multirow{3}{*}{SignSparse}
 & ViT-B/16 & $+0.41$ & $+1.03$ & $+1.50$ & $+1.36$ & $+1.29$ & $+1.12$ \\
 & ViT-B/32 & $+0.44$ & $+1.24$ & $+1.50$ & $+1.08$ & $+1.24$ & $+1.10$ \\
 & ViT-L/14 & $+0.78$ & $+1.40$ & $+1.10$ & $+1.06$ & $+1.13$ & $+1.09$ \\
\midrule
\multirow{3}{*}{DynSignSparse}
 & ViT-B/16 & $+0.50$ & $+0.42$ & $+0.40$ & $+0.26$ & $+0.18$ & $+0.35$ \\
 & ViT-B/32 & $+0.95$ & $+0.92$ & $+0.37$ & $+0.44$ & $+0.26$ & $+0.59$ \\
 & ViT-L/14 & $+0.81$ & $+0.84$ & $+0.72$ & $+0.72$ & $+0.36$ & $+0.69$ \\
\bottomrule
\end{tabular}
\caption{Accuracy difference between tensor-wise and model-wise sparse-support selection, computed from the 11-dataset averages. %Positive values favor the tensor-wise $\Top_{k_r}$ rule. 
Columns 1--16 denote shots per class.}
\label{tab:selection_scope_delta}
\end{table}

\paragraph{Selection criteria.}
Fig.~\ref{fig:importance_selection} compares two ways to choose the parameters to update: (i) \emph{random} sampling and (ii) \emph{importance}-based selection using the largest gradient magnitudes. Importance-based selection consistently outperforms random sampling for both SignSparseSGD and DynSignSparseSGD across backbones and shots, with DynSignSparseSGD (importance) performing best overall. These results indicate that gradient magnitudes provide a reliable signal for identifying salient coordinates, while the sign-only update discards noisy magnitudes. Detailed results are in \textcolor{blue}{\hyperlink{suppnav.random-selection}{Appendix~Q}}.

\paragraph{Selection scope.}
The proposed optimizers apply $\Top_{k_r}$ independently to each trainable parameter tensor. We compare this tensor-wise rule with a model-wise alternative that applies one $\Top_k$ operation to all trainable parameters at the same nominal density. Table~\ref{tab:selection_scope_delta} reports tensor-wise minus model-wise accuracy. Tensor-wise selection improves every method--backbone--shot average. For the static method, the gain remains above one point from 2 to 16 shots. For the dynamic method, it is largest at 1--2 shots and generally narrows as the support set grows. This pattern is consistent with dynamic refresh partly correcting an initially concentrated model-wise allocation, whereas a static model-wise mask preserves that allocation throughout training. Full per-dataset and per-shot results are provided in \textcolor{blue}{\hyperlink{suppnav.selection-scope}{Appendix~R}}.

\begin{figure}[!htbp]
\centering
\includegraphics[width=0.9\columnwidth]{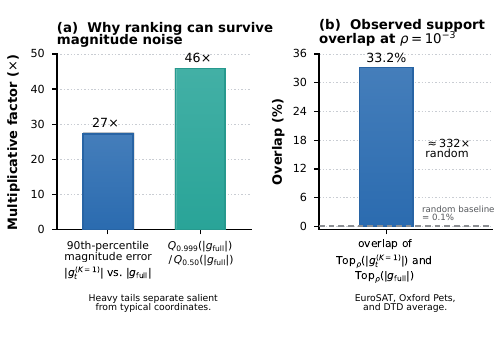}
\caption{Magnitude ranking under few-shot noise. We compare the 1-shot gradient $g_t^{(K=1)}$ with the full-training-set gradient $g_{\rm full}$ on EuroSAT, Oxford Pets, and DTD using CLIP (ViT-B/16), three seeds, and checkpoints 0, 10, and 30. (a) The 90th-percentile coordinate-wise magnitude discrepancy is $27\times$. The 99.9th percentile of $|g_{\rm full}|$ is $46\times$ its median. (b) At $\rho=10^{-3}$, the top-$0.1\%$ sets of $|g_t^{(K=1)}|$ and $|g_{\rm full}|$ overlap by $33.2\%$, compared with a $0.1\%$ random baseline.}
\label{fig:ranking_survives}
\end{figure}

\paragraph{Ranking survives magnitude noise.}
Fig.~\ref{fig:ranking_survives} compares the complete 1-shot gradient $g_t^{(K=1)}$ with the full-training-set gradient $g_{\rm full}$ at matched checkpoints. Fig.~\ref{fig:ranking_survives}~(a) shows a $27\times$ 90th-percentile coordinate-wise magnitude discrepancy. This confirms that $g_t^{(K=1)}$ is poorly calibrated for step scaling. However, the 99.9th percentile of $|g_{\rm full}|$ is $46\times$ its median. Thus, the threshold for the largest $0.1\%$ remains far above a typical coordinate. Fig.~\ref{fig:ranking_survives}~(b) verifies that the upper-tail ranking is still informative. At $\rho=10^{-3}$, the top-$0.1\%$ sets of $|g_t^{(K=1)}|$ and $|g_{\rm full}|$ overlap by $33.2\%$. Random sets of the same size overlap by only $0.1\%$ in expectation. Thus, the overlap is $332\times$ the random expectation. Magnitude remains useful for deciding \emph{where} to update, while the fixed sign step avoids using the noisy scale to decide \emph{how far} to move.

%\FloatBarrier

%\begin{samepage}
\paragraph{Limitations and practical recommendations.} This study is limited to CLIP-style image classification. Other architectures or modalities may require recalibration of the sign update scale. For a fixed short budget, DynSignSparseSGD is the stronger default because it follows evolving saliency. When selection cost or very long training is the primary concern, SignSparseSGD offers slightly lower late-epoch accuracy degradation.

\paragraph{Conclusion.} Few-shot gradients contain a useful distinction between saliency and scale. SignSparseSGD and DynSignSparseSGD use top-magnitude coordinates to allocate a tiny update budget, discard the noisy coordinate-wise scale, and move by fixed signs without maintaining moment buffers. Dynamic selection is the strongest fixed-budget default across the few-shot benchmark, while static selection provides the smallest long-horizon accuracy degradation. This motivates further investigation into whether parts of the backpropagation and gradient-based optimization pipeline can be simplified in this regime. A promising next step is to design mechanisms that identify salient parameters without relying on gradient computation. Another avenue is to develop approximate update rules that reliably estimate sign directions with a lower computational cost compared to standard backpropagation.
%\par
%\end{samepage}

% Overall, using magnitude for sparse support selection and fixed signs for the update provides strong few-shot accuracy, low optimizer-state storage, and improved schedule robustness

%This work suggests that some ingredients often treated as essential (e.g., exact gradient magnitudes and heavy optimizer states) may not always be necessary in the few-shot regime. Our results indicate that updating only a tiny fraction of parameters using their gradient signs is sufficient to achieve strong performance in few-shot adaptation. This motivates further investigation into whether parts of the backpropagation and gradient-based optimization pipeline can be simplified in this regime. A promising next step is to design mechanisms that identify salient parameters without relying on gradient computation. Another avenue is to develop approximate update rules that reliably estimate sign directions with a lower memory footprint and computational cost compared to standard backpropagation.

%

\FloatBarrier
\suppanchor{references}
\bibliography{references}

\clearpage
\onecolumn
\flushbottom
\widowpenalty=150
\setcounter{secnumdepth}{2}
\definecolor{LightGray}{RGB}{220,220,240}
\alglanguage{pseudocode}
\begin{center}
{\LARGE\bfseries Supplementary Material\par}
\vspace{0.5em}
{\large\bfseries On the Few-Shot Adaptation of Vision-Language Models:\\
A Few Gradient Signs Are All You Need\par}
\end{center}
\vspace{0.75em}
\appendix
\setcounter{equation}{0}
\renewcommand{\theequation}{S\arabic{equation}}
\renewcommand{\theHequation}{supp.\arabic{equation}}
\setcounter{figure}{0}
\renewcommand{\thefigure}{S\arabic{figure}}
\renewcommand{\theHfigure}{supp.\arabic{figure}}
\setcounter{table}{0}
\renewcommand{\thetable}{S\arabic{table}}
\renewcommand{\theHtable}{supp.\arabic{table}}
\setcounter{algorithm}{0}
\renewcommand{\thealgorithm}{S\arabic{algorithm}}
\renewcommand{\theHalgorithm}{supp.\arabic{algorithm}}
\section*{Table of Supplementary Contents}
\begingroup
\small
\setlength{\tabcolsep}{4pt}
\renewcommand{\arraystretch}{1.12}
\setlength{\LTpre}{0pt}
\setlength{\LTpost}{0pt}
\begin{longtable}{@{}>{\raggedright\arraybackslash}p{0.09\textwidth}>{\raggedright\arraybackslash}p{0.76\textwidth}>{\centering\arraybackslash}p{0.08\textwidth}@{}}
\toprule
\textbf{Appendix} & \textbf{Title} & \textbf{Page} \\
\midrule
\endfirsthead
\toprule
\textbf{Appendix} & \textbf{Title} & \textbf{Page} \\
\midrule
\endhead
\midrule
\multicolumn{3}{r}{\footnotesize Continued on the next page}\\
\endfoot
\bottomrule
\endlastfoot
A & \supplink{clip}{Few-shot Adaptation of CLIP} & \supplink{clip}{\pageref{suppnav:clip}} \\
B & \supplink{related}{Extended Related Work} & \supplink{related}{\pageref{suppnav:related}} \\
C & \supplink{algorithms}{Algorithms} & \supplink{algorithms}{\pageref{suppnav:algorithms}} \\
D & \supplink{hyperparameters}{Hyperparameters} & \supplink{hyperparameters}{\pageref{suppnav:hyperparameters}} \\
E & \supplink{efficiency}{Detailed Training-Time and Memory Analysis} & \supplink{efficiency}{\pageref{suppnav:efficiency}} \\
F & \supplink{geometry}{Geometry and Convergence Analysis} & \supplink{geometry}{\pageref{suppnav:geometry}} \\
F.1 & \supplink{geometry-oracle}{Sparse sign as a blockwise steepest-descent oracle} & \supplink{geometry-oracle}{\pageref{suppnav:geometry-oracle}} \\
F.2 & \supplink{sparsity-lr}{Sparsity--learning-rate coupling} & \supplink{sparsity-lr}{\pageref{suppnav:sparsity-lr}} \\
F.3 & \supplink{reachability}{Deterministic blockwise reachability} & \supplink{reachability}{\pageref{suppnav:reachability}} \\
F.4 & \supplink{bias-certificate}{Trajectory-wise finite-shot bias and a reference certificate} & \supplink{bias-certificate}{\pageref{suppnav:bias-certificate}} \\
F.5 & \supplink{descent-lemmas}{Auxiliary descent, support, and alignment lemmas} & \supplink{descent-lemmas}{\pageref{suppnav:descent-lemmas}} \\
F.6 & \supplink{population-bounds}{Trajectory-wise population bounds} & \supplink{population-bounds}{\pageref{suppnav:population-bounds}} \\
F.7 & \supplink{alignment}{Calibrated population alignment} & \supplink{alignment}{\pageref{suppnav:alignment}} \\
F.8 & \supplink{relative-errors}{Relative-error sufficient conditions} & \supplink{relative-errors}{\pageref{suppnav:relative-errors}} \\
G & \supplink{confidence}{Confidence Gating in the Few-Shot Regime} & \supplink{confidence}{\pageref{suppnav:confidence}} \\
G.1 & \supplink{confidence-exact}{Exact confidence-gated descent} & \supplink{confidence-exact}{\pageref{suppnav:confidence-exact}} \\
G.2 & \supplink{confidence-continuous}{Continuous confidence factors and binary gates} & \supplink{confidence-continuous}{\pageref{suppnav:confidence-continuous}} \\
G.3 & \supplink{confidence-oracle}{Descent-optimal blockwise supports} & \supplink{confidence-oracle}{\pageref{suppnav:confidence-oracle}} \\
G.4 & \supplink{confidence-regimes}{Explicit dense-versus-sparse SignSGD regimes} & \supplink{confidence-regimes}{\pageref{suppnav:confidence-regimes}} \\
G.5 & \supplink{confidence-adam}{Conditional advantage relative to Adam's confidence interpretation} & \supplink{confidence-adam}{\pageref{suppnav:confidence-adam}} \\
G.6 & \supplink{confidence-scope}{Scope and failure modes} & \supplink{confidence-scope}{\pageref{suppnav:confidence-scope}} \\
H & \supplink{ranking}{Magnitude Calibration and Blockwise Ranking Stability} & \supplink{ranking}{\pageref{suppnav:ranking}} \\
H.1 & \supplink{ranking-regret}{Gap-free support-selection regret} & \supplink{ranking-regret}{\pageref{suppnav:ranking-regret}} \\
H.2 & \supplink{ranking-recovery}{Gap-dependent exact support recovery} & \supplink{ranking-recovery}{\pageref{suppnav:ranking-recovery}} \\
H.3 & \supplink{ranking-interpretation}{Interpretation and limitations} & \supplink{ranking-interpretation}{\pageref{suppnav:ranking-interpretation}} \\
I & \supplink{noise}{Few-Shot Noise Characterization} & \supplink{noise}{\pageref{suppnav:noise}} \\
J & \supplink{peft}{SignSparseSGD \& DynSignSparseSGD vs PEFT/OET} & \supplink{peft}{\pageref{suppnav:peft}} \\
J.1 & \supplink{peft-a2a}{All-to-All Few-Shot Evaluation} & \supplink{peft-a2a}{\pageref{suppnav:peft-a2a}} \\
J.2 & \supplink{peft-b2n}{Base-to-New Few-Shot Evaluation} & \supplink{peft-b2n}{\pageref{suppnav:peft-b2n}} \\
K & \supplink{stability}{Stability Analysis} & \supplink{stability}{\pageref{suppnav:stability}} \\
L & \supplink{update-scale}{Update Scale} & \supplink{update-scale}{\pageref{suppnav:update-scale}} \\
L.1 & \supplink{update-measure}{Measuring Update Magnitude} & \supplink{update-measure}{\pageref{suppnav:update-measure}} \\
L.2 & \supplink{update-normalization}{Effect of Gradient Normalization} & \supplink{update-normalization}{\pageref{suppnav:update-normalization}} \\
M & \supplink{sensitivity}{Sensitivity Analysis} & \supplink{sensitivity}{\pageref{suppnav:sensitivity}} \\
N & \supplink{dyn-static}{DynSignSparseSGD vs SignSparseSGD} & \supplink{dyn-static}{\pageref{suppnav:dyn-static}} \\
O & \supplink{optimizer-comparison}{Sparse \& Sign Optimizer Comparison} & \supplink{optimizer-comparison}{\pageref{suppnav:optimizer-comparison}} \\
O.1 & \supplink{optimizer-raw}{Without Gradient Normalization} & \supplink{optimizer-raw}{\pageref{suppnav:optimizer-raw}} \\
O.2 & \supplink{optimizer-normalized}{With Gradient Normalization} & \supplink{optimizer-normalized}{\pageref{suppnav:optimizer-normalized}} \\
O.3 & \supplink{optimizer-momentum}{Matched Momentum Control} & \supplink{optimizer-momentum}{\pageref{suppnav:optimizer-momentum}} \\
P & \supplink{clipping}{Sign vs Gradient Clipping} & \supplink{clipping}{\pageref{suppnav:clipping}} \\
Q & \supplink{random-selection}{Randomness vs Gradient Magnitude} & \supplink{random-selection}{\pageref{suppnav:random-selection}} \\
R & \supplink{selection-scope}{Selection Scope} & \supplink{selection-scope}{\pageref{suppnav:selection-scope}} \\
S & \supplink{hardware_software}{Hardware and Software} & \supplink{hardware_software}{\pageref{suppnav:hardware_software}} \\
-- & \supplink{references}{References} & \supplink{references}{\pageref{suppnav:references}} \\
\end{longtable}
\endgroup
\clearpage
\section*{Summary of Notation}
\noindent The table records the notation used in the main paper and this supplementary material. Dimensions are symbolic when they depend on the CLIP architecture. Each index has one role, as specified below; its range is stated in the corresponding equation.
\vspace{0.5em}
\begingroup
\footnotesize
\setlength{\tabcolsep}{3pt}
\renewcommand{\arraystretch}{1.13}
\setlength{\LTpre}{0pt}
\setlength{\LTpost}{0pt}
\begin{longtable}{@{}>{\raggedright\arraybackslash}p{0.15\textwidth}>{\raggedright\arraybackslash}p{0.22\textwidth}>{\raggedright\arraybackslash}p{0.42\textwidth}>{\raggedright\arraybackslash}p{0.17\textwidth}@{}}
\toprule
\textbf{Symbol} & \textbf{Type / Domain / Dimension} & \textbf{Definition or Role} & \textbf{Canonical Definition Location} \\
\midrule
\endfirsthead
\toprule
\textbf{Symbol} & \textbf{Type / Domain / Dimension} & \textbf{Definition or Role} & \textbf{Canonical Definition Location} \\
\midrule
\endhead
\midrule
\multicolumn{4}{r}{\footnotesize Continued on the next page}\\
\endfoot
\bottomrule
\endlastfoot
\addlinespace[2pt]
\multicolumn{4}{@{}l}{\textbf{Indices, data, and objectives}}\\
\addlinespace[1pt]
$c,c^{\prime}$ & Class indices in $\{1,\ldots,N\}$ & Class index and secondary class index in softmax sums. & Appendix~\ref{app:clip_adaptation} \\
$i,i'$ & Coordinate indices in $\{1,\ldots,d\}$ & Parameter-coordinate index and secondary coordinate index in pairwise comparisons. & Main Sec.~\ref{sec:method}; Appendix~\ref{app:clip_adaptation} \\
$j$ & Order-statistic index & Rank position in an ordered magnitude or curvature sum. & Appendices~\ref{app:clip_adaptation} and~\ref{app:convergence} \\
$r$ & Block index in $\{1,\ldots,R\}$ & Index of a trainable parameter tensor or parameter block. & Main Eq.~\eqref{eq:block_budgets} \\
$t$ & Iteration index, usually $0\le t<T$ & Optimization iteration. & Main Algorithm~\ref{alg:signsparse} \\
$t'$ & Iteration index & Past-iteration index in trajectory sums. & Appendices~\ref{app:convergence} and~\ref{app:confidence_gating} \\
$j_{\mathrm{data}}$ & Example index & Position of an image or labeled example in a dataset. & Main Sec.~\ref{sec:method}; Appendix~\ref{app:clip_adaptation} \\
$j_{\mathrm{shot}}$ & Within-class example index & Position of a support example within class $c$. & Population setup underlying Appendix~\ref{app:convergence} \\
$j_{\mathrm{prompt}}$ & Prompt index in $\{1,\ldots,M_{\mathrm{prompt}}\}$ & Text-prompt position within a class. & Appendix~\ref{app:clip_adaptation} \\
$j_{\mathrm{obs}}$ & Observation index in $\{1,\ldots,T_{\mathrm{obs}}\}$ & Position of an idealized confidence observation. & Eq.~\eqref{eq:cg_confidence_observations} \\
$T$ & Positive integer & Optimization horizon or number of parameter updates. & Main Algorithm~\ref{alg:signsparse} \\
$T_{\mathrm{obs}}$ & Positive integer & Number of idealized confidence observations per coordinate. & Appendix~\ref{app:confidence_gating}, Eq.~\eqref{eq:cg_confidence_observations} \\
$N$ & Positive integer & Number of classes in a few-shot task. & Main Sec.~\ref{sec:method} \\
$K$ & Positive integer & Number of labeled support examples per class. & Main Sec.~\ref{sec:method} \\
$n=NK$ & Positive integer & Total number of labeled support examples. & Main Sec.~\ref{sec:method} \\
$D$ & Set of $n$ labeled examples & Few-shot support set; equivalently $\{z_{j_{\mathrm{data}}}\}_{j_{\mathrm{data}}=1}^{n}$ or the class--shot indexed form $\{z_{c,j_{\mathrm{shot}}}\}$. & Main Sec.~\ref{sec:method}; Appendix~\ref{app:clip_adaptation} \\
$D_{\mathrm{query}}$ & Dataset disjoint from $D$ & Query or test set used only for evaluation. & Appendix~\ref{app:clip_adaptation} \\
$z_{j_{\mathrm{data}}}=(x_{j_{\mathrm{data}}},y_{j_{\mathrm{data}}})$ & Labeled example & An image and its class label in the support set. & Main Sec.~\ref{sec:method} \\
$z_{c,j_{\mathrm{shot}}}$ & Random labeled example & The $j_{\mathrm{shot}}$th independent support example from class-conditional distribution $\mathcal P_c$. & Population setup underlying Appendix~\ref{app:convergence} \\
$x_{j_{\mathrm{data}}}$ & Image input & Image component of example $z_{j_{\mathrm{data}}}$; also used for a query image. & Main Sec.~\ref{sec:method}; Appendix~\ref{app:clip_adaptation} \\
$y_{j_{\mathrm{data}}}$ & Integer in $\{1,\ldots,N\}$ & Class label of image $x_{j_{\mathrm{data}}}$. & Appendix~\ref{app:clip_adaptation} \\
$\mathcal P_c$ & Probability distribution & Within-task class-conditional population distribution for class $c$. & Population setup underlying Assumption~\ref{assumption_1} \\
$B_t$ & Subset of $D$ & Optimization minibatch at iteration $t$. & Main Eq.~\eqref{eq:minibatch_gradient_method} \\
$b_t=|B_t|$ & Positive integer & Minibatch size at iteration $t$. & Main Eq.~\eqref{eq:minibatch_gradient_method} \\
$\ell(\Theta;z)$ & Scalar-valued loss & Per-example cross-entropy loss. & Main Eq.~\eqref{eq:minibatch_gradient_method}; Appendix~\ref{app:clip_adaptation} \\
$\mathcal L_{B_t}(\Theta)$ & $\mathbb R^d\to\mathbb R$ & Minibatch empirical loss, $b_t^{-1}\sum_{z\in B_t}\ell(\Theta;z)$. & Main Eq.~\eqref{eq:minibatch_gradient_method} \\
$f(\Theta)$ & $\mathbb R^d\to\mathbb R$ & Support-set empirical risk. & Appendix~\ref{app:clip_adaptation}; Eq.~\eqref{eq:population_gradient_decomposition} \\
$F(\Theta)$ & $\mathbb R^d\to\mathbb R$ & Balanced population risk over the class-conditional distributions. & Eq.~\eqref{eq:population_risk} \\
$F_\star$ & Real scalar & Lower bound satisfying $F(\Theta)\ge F_\star$ on the relevant trajectory envelope. & Assumption~\ref{assumption_1} \\
$D_{\mathrm{ref}}$ & Independent reference dataset & Reference set containing $K_{\mathrm{ref}}$ independent examples per class. & Appendix~\ref{app:convergence}, Proposition~\ref{prop:reference_gradient_certificate} \\
$K_{\mathrm{ref}}$ & Positive integer & Number of independent reference examples per class. & Appendix~\ref{app:convergence}, Proposition~\ref{prop:reference_gradient_certificate} \\
$n_{\mathrm{ref}}=NK_{\mathrm{ref}}$ & Positive integer & Total reference-set size. & Appendix~\ref{app:convergence}, Proposition~\ref{prop:reference_gradient_certificate} \\
$f_{\mathrm{ref}}$ & $\mathbb R^d\to\mathbb R$ & Empirical risk on $D_{\mathrm{ref}}$. & Appendix~\ref{app:convergence}, Proposition~\ref{prop:reference_gradient_certificate} \\
$\delta$ & Scalar in $(0,1)$ & Failure-probability level in the reference-gradient and ranking certificates. & Appendix~\ref{app:convergence}, Proposition~\ref{prop:reference_gradient_certificate} \\
\addlinespace[2pt]
\multicolumn{4}{@{}l}{\textbf{CLIP representations and predictions}}\\
\addlinespace[1pt]
$\Theta_{\mathrm{img}},\Theta_{\mathrm{txt}}$ & Architecture-dependent parameter collections & Image-encoder and text-encoder parameters. & Appendix~\ref{app:clip_adaptation} \\
$E_{\Theta_{\mathrm{img}}}$ & Image encoder into the shared embedding space & CLIP image encoder. & Appendix~\ref{app:clip_adaptation} \\
$H_{\Theta_{\mathrm{txt}}}$ & Text encoder into the shared embedding space & CLIP text encoder. & Appendix~\ref{app:clip_adaptation} \\
$x_{j_{\mathrm{data}}}^{\mathrm{txt}}$ & Text input & Text component paired with image $x_{j_{\mathrm{data}}}$ during CLIP pretraining. & Appendix~\ref{app:clip_adaptation} \\
$\phi_{j_{\mathrm{data}}}^{\mathrm{img}},\phi_{j_{\mathrm{data}}}^{\mathrm{txt}}$ & Normalized embedding vectors of the shared CLIP embedding dimension & Image and text embeddings. The symbol $\phi$ is reserved for embeddings; $e_D$ denotes finite-shot gradient bias. & Appendix~\ref{app:clip_adaptation} \\
$M_{\mathrm{prompt}}$ & Positive integer & Number of zero-shot text prompts per class. & Appendix~\ref{app:clip_adaptation} \\
$x_{j_{\mathrm{prompt}},c}^{\mathrm{prompt}}$ & Text prompt & Prompt $j_{\mathrm{prompt}}$ for class $c$. & Appendix~\ref{app:clip_adaptation} \\
$\bar\phi_c^{\mathrm{txt}}$ & Embedding vector & Unnormalized average of the prompt embeddings for class $c$. & Appendix~\ref{app:clip_adaptation} \\
$\phi_c^{\mathrm{txt}}$ & Unit-norm embedding vector & Normalized text prototype for class $c$. & Appendix~\ref{app:clip_adaptation} \\
$\tau$ & Positive real scalar & Softmax temperature used by CLIP or the specified baseline. & Appendix~\ref{app:clip_adaptation}, Eq.~\eqref{y_ic} \\
$\hat y_{j_{\mathrm{data}},c}(\Theta)$ & Scalar in $[0,1]$ & CLIP probability assigned to class $c$ for image $x_{j_{\mathrm{data}}}$. & Appendix~\ref{app:clip_adaptation}, Eq.~\eqref{y_ic} \\
\addlinespace[2pt]
\multicolumn{4}{@{}l}{\textbf{Core sign-sparse optimization}}\\
\addlinespace[1pt]
$\Theta\in\mathbb R^d$ & Parameter vector & Concatenation of all trainable CLIP parameters. & Main Sec.~\ref{sec:method} \\
$\Theta_t\in\mathbb R^d$ & Parameter vector & Trainable parameter vector at iteration $t$. & Main Sec.~\ref{sec:method} \\
$d$ & Positive integer & Total number of trainable coordinates and the ambient optimization dimension. & Main Sec.~\ref{sec:method} \\
$g_t\in\mathbb R^d$ & Dense vector & Dense minibatch gradient $\nabla_\Theta\mathcal L_{B_t}(\Theta_t)$. In the theory, $g_t=h_t+e_D(\Theta_t)+\xi_t$. & Main Eq.~\eqref{eq:minibatch_gradient_method}; Eq.~\eqref{eq:population_gradient_decomposition} \\
$R$ & Positive integer & Number of trainable parameter-tensor blocks. & Main Eq.~\eqref{eq:block_budgets} \\
$\mathcal I_r$ & Subset of $\{1,\ldots,d\}$ & Coordinate indices of trainable parameter block $r$; the blocks form a disjoint partition. & Main Eq.~\eqref{eq:block_budgets} \\
$d_r=|\mathcal I_r|$ & Positive integer & Dimension of block $r$. & Main Eq.~\eqref{eq:block_budgets} \\
$\rho\in(0,1]$ & Real scalar & Nominal per-block update density. & Main Eq.~\eqref{eq:block_budgets} \\
$k_r$ & Integer in $\{1,\ldots,d_r\}$ & Block budget $\max\{1,\lfloor\rho d_r\rfloor\}$. & Main Eq.~\eqref{eq:block_budgets} \\
$k_{\mathrm{step}}$ & Positive integer & Realized number of selected coordinates per update, $\sum_{r=1}^{R}k_r$. & Main Eq.~\eqref{eq:block_budgets} \\
$k_{\mathrm{diag}}$ & Positive integer & Global active-set size used only in the gradient diagnostics, $\lceil\rho d\rceil$. & Main Fig.~\ref{fig:gradient_diagnostics}; Appendix~\ref{app:fewshot_noise} \\
$\Top_k(v)$ & Set-valued operator returning $k$ indices & Indices of the $k$ largest entries of $v$ under the paper's fixed tie rule. & Main Eq.~\eqref{eq:block_budgets} \\
$S_{t,r}$ & Subset of $\mathcal I_r$ with $|S_{t,r}|=k_r$ & Current top-magnitude support in block $r$. & Main Eq.~\eqref{eq:block_budgets} \\
$S_t$ & Subset of $\{1,\ldots,d\}$ & Dynamic support $\bigcup_r S_{t,r}$ at iteration $t$. & Main Eq.~\eqref{eq:block_budgets} \\
$S$ & Subset of $\{1,\ldots,d\}$ & Fixed support used by a static sparse method; for SignSparseSGD, $S=S_0$. & Main Algorithm~\ref{alg:signsparse} \\
$m_t,m,m_S$ & Binary vectors in $\{0,1\}^d$ & Masks of $S_t$, $S$, or a generic support $S$, respectively. & Main Eq.~\eqref{eq:sparse_gradient} \\
$\mathbf 1\{\cdot\}$ & Indicator in $\{0,1\}$ & Indicator function used to construct masks and gates. & Main Eq.~\eqref{eq:sparse_gradient} \\
$\odot$ & Binary operator on equally shaped arrays & Elementwise multiplication. & Main Eq.~\eqref{eq:sparse_gradient} \\
$\oslash$ & Binary operator on equally shaped arrays & Elementwise division in the optimizer pseudocode. & Appendix~\ref{app:algorithms} \\
$\supp(v)$ & Subset of coordinate indices & Support of vector $v$. & Appendix~\ref{app:convergence}, Proposition~\ref{prop:sparse_steepest_descent} \\
$\widetilde g_t\in\mathbb R^d$ & Sparse vector & Masked gradient $m_t\odot g_t$ or $m\odot g_t$, according to the dynamic or static support. & Main Eq.~\eqref{eq:sparse_gradient} \\
$\sgn(a)$ & $\mathbb R\to\{-1,0,1\}$, applied elementwise & Ternary sign map with $\sgn(0)=0$. & Main Eq.~\eqref{eq:sparse_sign_update} \\
$s_t\in\{-1,0,1\}^d$ & Sparse signed direction & $\sgn(\widetilde g_t)=m_t\odot\sgn(g_t)$ with $m_t=m_0$ for the static variant. & Main Eq.~\eqref{eq:sparse_sign_update} \\
$\eta>0$ & Real scalar & Base learning rate. & Main Eq.~\eqref{eq:sparse_sign_update} \\
$\alpha>0$ & Real scalar & Sparse-sign multiplier; $\alpha=0.1$ in the reported experiments. & Main Eq.~\eqref{eq:sparse_sign_update} \\
$\lambda=\alpha\eta$ & Positive real scalar & Effective coordinate step of SignSparseSGD and DynSignSparseSGD. & Main Eq.~\eqref{eq:sparse_sign_update} \\
$\lambda_t>0$ & Positive real scalar & Iteration-dependent coordinate step in the theoretical analysis; $\lambda_t=\lambda$ is the constant-step algorithmic case. & Appendix~\ref{app:convergence} \\
$\Lambda_T$ & Positive real scalar & Cumulative step $\sum_{t=0}^{T-1}\lambda_t$. & Eq.~\eqref{eq:population_direction_schedule} \\
$q$ & Element of $\{\textsc{static},\textsc{dynamic}\}$ & Algorithm mode selecting a fixed or refreshed support. & Main Algorithm~\ref{alg:signsparse} \\
$g_t^{(K=1)},g_t^{(K=16)}$ & Vectors in $\mathbb R^d$ & Matched-iterate 1-shot gradient and 16-shot reference gradient used only in the noise diagnostics. & Main Fig.~\ref{fig:gradient_diagnostics}; Appendix~\ref{app:fewshot_noise} \\
$g_{\mathrm{full}}$ & Vector in $\mathbb R^d$ & Full-training-set reference gradient at the same checkpoint as $g_t^{(K=1)}$; distinct from the 16-shot reference. & Main Fig.~\ref{fig:ranking_survives} \\
\addlinespace[2pt]
\multicolumn{4}{@{}l}{\textbf{Optimizer states, clipping, and baseline hyperparameters}}\\
\addlinespace[1pt]
$u_t,v_t\in\mathbb R^d$ & First- and second-moment vectors & Adam-style first- and second-moment states; $u_t$ is the retained first-moment state in SignAdam and SignSparseAdam. & Appendix~\ref{app:algorithms} \\
$\hat u_t,\hat v_t\in\mathbb R^d$ & Bias-corrected moment vectors & Bias-corrected first and second moments. & Appendix~\ref{app:algorithms} \\
$\beta_1,\beta_2\in[0,1)$ & Real scalars & First- and second-moment decay coefficients. & Appendix~\ref{app:algorithms} \\
$\varepsilon>0$ & Real scalar & Denominator offset in Adam-style updates. & Appendix~\ref{app:algorithms} \\
$a_t\ge0$ & Real scalar & Online norm-matching factor used by GradNorm SignSGD or GradNorm SparseSGD. & Appendix~\ref{app:algorithms} \\
$c_{\mathrm{loc}}>0$ & Real scalar & Coordinatewise clipping threshold for Local-clip SparseSGD. & Appendix~\ref{app:algorithms} \\
$\operatorname{clip}(v;c_{\mathrm{loc}})$ & Vector with the same dimension as $v$ & Elementwise clamp of $v$ to $[-c_{\mathrm{loc}},c_{\mathrm{loc}}]$. & Appendix~\ref{app:algorithms} \\
$g_t^{\mathrm{loc}}\in\mathbb R^d$ & Sparse clipped vector & Coordinatewise-clipped sparse gradient. & Appendix~\ref{app:algorithms} \\
$c_{\mathrm{glob}}>0$ & Real scalar & Global $\ell_2$-norm clipping threshold. & Appendix~\ref{app:algorithms} \\
$a_t^{\mathrm{glob}}\in(0,1]$ & Real scalar & Global clipping factor $\min\{1,c_{\mathrm{glob}}/\|g_t\|_2\}$, with the stated zero-gradient convention. & Appendix~\ref{app:algorithms} \\
$g_t^{\mathrm{glob}},\widetilde g_t^{\mathrm{glob}}$ & Vectors in $\mathbb R^d$ & Globally clipped dense gradient and its masked sparse version. & Appendix~\ref{app:algorithms} \\
$E_{\mathrm{train}}$ & Positive integer & Number of training epochs for a reported method. & Appendix~\ref{app:hyperparameters} \\
$b_{\mathrm{train}}$ & Positive integer or full-batch & Training batch-size hyperparameter in the experimental protocol. & Appendix~\ref{app:hyperparameters} \\
$p_{\mathrm{flip}}\in[0,1]$ & Real scalar & Probability of random horizontal flipping in the shared augmentation pipeline. & Appendix~\ref{app:hyperparameters} \\
$M_{\mathrm{ctx}}$ & Positive integer & Number of trainable context tokens used by prompt-tuning baselines. & Appendix~\ref{app:hyperparameters} \\
$\zeta_1,\zeta_2$ & Real scalars & KgCoOp knowledge-regularization weight and ProGrad projection weight. & Appendix~\ref{app:hyperparameters} \\
$\iota_{\mathrm{TIP,mix}},\iota_{\mathrm{TIP,scale}}$ & Real scalars & TIP-Adapter-F cache-mixing and cache-scaling weights. & Appendix~\ref{app:hyperparameters} \\
$r_{\mathrm{LoRA}}$ & Positive integer & LoRA factor rank. & Appendix~\ref{app:hyperparameters} \\
$\alpha_{\mathrm{LoRA}}$ & Real scalar & LoRA-specific scaling factor; distinct from the sparse-sign multiplier $\alpha$. & Appendix~\ref{app:hyperparameters} \\
$r_{\mathrm{GaLoRE}}$ & Positive integer & GaLoRE projection rank. & Appendix~\ref{app:hyperparameters} \\
$p_{\mathrm{refresh}}$ & Positive integer & Support or projection refresh period in optimizer iterations. & Appendix~\ref{app:hyperparameters} \\
$s_{\mathrm{SO}}\in[0,1)$ & Real scalar & SO sparsity ratio, related to the paper's density by $\rho=1-s_{\mathrm{SO}}$. & Appendix~\ref{app:hyperparameters} \\
$W\in\mathbb R^{d_{\mathrm{out}}\times d_{\mathrm{in}}}$ & Matrix & Generic fully connected weight matrix used in the idealized memory analysis. & Appendix~\ref{app:efficiency} \\
$d_{\mathrm{in}},d_{\mathrm{out}}$ & Positive integers & Input and output dimensions of $W$, with $d_{\mathrm{out}}\le d_{\mathrm{in}}$. & Appendix~\ref{app:efficiency} \\
$r_{\mathrm{low}}$ & Positive integer & Generic low-rank dimension in the idealized memory comparison. & Appendix~\ref{app:efficiency} \\
\addlinespace[2pt]
\multicolumn{4}{@{}l}{\textbf{Sparse geometry and trajectory quantities}}\\
\addlinespace[1pt]
$|v|_{(j)}$ & Nonnegative scalar & $j$th largest coordinate magnitude of $v\in\mathbb R^d$. & Appendix~\ref{app:convergence}, \hyperlink{suppnav.geometry-oracle}{Sec.~F.1} \\
$\|v\|_{[k]}$ & Nonnegative scalar & Sum of the $k$ largest coordinate magnitudes of $v$. & Appendix~\ref{app:convergence}, \hyperlink{suppnav.geometry-oracle}{Sec.~F.1} \\
$\mathbf k=(k_1,\ldots,k_R)$ & Integer vector in $\mathbb N^R$ & Collection of blockwise selection budgets. & Appendix~\ref{app:convergence}, Proposition~\ref{prop:sparse_steepest_descent} \\
$\|v\|_{[\mathbf k]}$ & Nonnegative scalar & Blockwise top-$k$ mass $\sum_r\|v_{\mathcal I_r}\|_{[k_r]}$. & Appendix~\ref{app:convergence}, Proposition~\ref{prop:sparse_steepest_descent} \\
$\mathcal C_{d,k}$ & Subset of $\mathbb R^d$ & Feasible $\ell_\infty$/$\ell_1$ direction set $\{u:\|u\|_\infty\le1,\|u\|_1\le k\}$. & Appendix~\ref{app:convergence}, Proposition~\ref{prop:sparse_steepest_descent} \\
$Q_k(v)$ & Vector in $\{-1,0,1\}^d$ & Single-block top-$k$ sign oracle $m_{\Top_k(|v|)}\odot\sgn(v)$. & Appendix~\ref{app:convergence}, Proposition~\ref{prop:sparse_steepest_descent} \\
$\mathcal C_{\mathbf k}$ & Subset of $\mathbb R^d$ & Blockwise product of the sparse $\ell_\infty$ direction constraints. & Appendix~\ref{app:convergence}, Proposition~\ref{prop:sparse_steepest_descent} \\
$Q_{\mathbf k}(v)$ & Vector in $\{-1,0,1\}^d$ & Concatenated blockwise top-$k$ sign oracle. & Appendix~\ref{app:convergence}, Proposition~\ref{prop:sparse_steepest_descent} \\
$p\in[1,\infty)$ & Real scalar & Norm order in the sign-sparse step-norm identity. & Appendix~\ref{app:convergence}, Eq.~\eqref{eq:appE_step_lp_norm} \\
$\mathcal U_{\mathbf k}$ & Subset of $\{-1,0,1\}^d$ & Direction alphabet with at most $k_r$ nonzero entries in each block. & Eq.~\eqref{eq:population_direction_schedule} \\
$\mathcal R_T$ & Subset of $\mathbb R^d$ & Finite-horizon set reachable from $\Theta_0$ by admissible signed sparse steps. & Eq.~\eqref{eq:population_reachable_envelope} \\
$\mathcal A_T$ & Subset of $\mathbb R^d$ & Deterministic coordinatewise and blockwise $\ell_1$ adaptation envelope containing $\mathcal R_T$. & Eq.~\eqref{eq:population_reachable_envelope} \\
$h_t=\nabla F(\Theta_t)$ & Vector in $\mathbb R^d$ & Population gradient at iteration $t$. & Appendix~\ref{app:convergence}, Eq.~\eqref{eq:population_gradient_decomposition} \\
$e_D(\Theta)$ & Vector in $\mathbb R^d$ & Fixed finite-shot bias $\nabla f(\Theta)-\nabla F(\Theta)$ conditional on $D$. & Eq.~\eqref{eq:population_gradient_decomposition} \\
$\xi_t\in\mathbb R^d$ & Random vector & Minibatch and augmentation error around $\nabla f(\Theta_t)$. & Eq.~\eqref{eq:population_gradient_decomposition} \\
$L=(L_1,\ldots,L_d)\in\mathbb R_+^d$ & Nonnegative vector & Coefficients of the simultaneous-update quadratic upper bound for $F$ on $\mathcal A_T$. & Assumption~\ref{assumption_1}, Eq.~\eqref{eq:ass_population_coord_smooth} \\
$\|L_S\|_1$ & Nonnegative scalar & Curvature mass $\sum_{i\in S}L_i$ on support $S$. & Assumption~\ref{assumption_1}; Appendix~\ref{app:convergence} \\
$L_{r,(j)}$ & Nonnegative scalar & $j$th largest coordinatewise curvature in block $r$. & Eq.~\eqref{eq:population_curvature_budgets} \\
$\bar L_{\mathbf k}$ & Nonnegative scalar & Maximum blockwise selected curvature budget $\sum_r\sum_{j=1}^{k_r}L_{r,(j)}$. & Eq.~\eqref{eq:population_curvature_budgets} \\
$\underline\rho$ & Real scalar in $(0,1]$ & Minimum realized block density $\min_r k_r/d_r$. & Eq.~\eqref{eq:population_curvature_budgets} \\
$L_t^{\mathrm{sel}}$ & Nonnegative scalar & Realized selected curvature $\sum_{i\in S_t}L_i$, bounded by $\bar L_{\mathbf k}$. & Eq.~\eqref{eq:population_curvature_budgets} \\
$\nu\in\mathbb R_+^d$ & Nonnegative vector & Coordinatewise first-moment budgets for the minibatch/augmentation error. & Assumption~\ref{assumption_1} \\
$\gamma\in(0,1]$ & Real scalar & Batch-size decay exponent in $\mathbb E[|\xi_{t,i}|\mid D,\Theta_t]\le\nu_i/b_t^\gamma$. & Assumption~\ref{assumption_1} \\
$\mathcal E_{\mathrm{miss},t}(S)$ & Nonnegative scalar & Population top-$\mathbf k$ mass not captured by support $S$. & Eq.~\eqref{eq:static_support_missed_mass}; Appendix~\ref{app:convergence} \\
$w_0\in\mathbb R^d$ & Perturbation vector & Error perturbing the population gradient when a static support is selected at initialization. & Appendix~\ref{app:convergence}, Lemma~\ref{lem:appE:static_support_miss} \\
$\lambda_{S,T}^{\star},\lambda_{\mathbf k,T}^{\star}$ & Positive real scalars & Minimizers of the static and dynamic explicit gap--curvature envelopes. & Appendix~\ref{app:convergence}, Eq.~\eqref{eq:appE_envelope_minimizing_steps} \\
$L_{\mathrm{low}},L_{\mathrm{high}}$ & Positive real scalars & Budget-independent lower and upper bounds on average selected curvature. & Appendix~\ref{app:convergence}, Eq.~\eqref{eq:appE_comparable_average_curvature} \\
\makecell[l]{$\rho_{\mathrm{ref}},\alpha_{\mathrm{ref}}$\\$\lambda_{\mathrm{ref}},k_{\mathrm{step,ref}}$} & Reference density, multiplier, step, and integer budget & Reference configuration used for sparsity-aware rescaling. & Appendix~\ref{app:convergence}, \hyperlink{suppnav.sparsity-lr}{Sec.~F.2} \\
$A_{\mathrm{gap}},C_{\mathrm{curv}}$ & Positive real scalars & Generic objective-gap and curvature coefficients used to minimize the explicit envelope. & Appendix~\ref{app:convergence}, proof of Proposition~\ref{prop:sparsity_learning_rate_coupling} \\
$\varepsilon_{\mathrm{sharp}}>0$ & Real scalar & Perturbation parameter in the one-dimensional construction proving sharpness of the robust top-$k$ alignment factor. & Appendix~\ref{app:convergence}, Lemma~\ref{lem:appE:block_topk_robust_alignment} \\
$\mu_{\mathrm{sep}}>0$ & Real scalar & Separation parameter in the fixed-shot impossibility construction. & Appendix~\ref{app:convergence}, Proposition~\ref{prop:fixed_shot_no_go} \\
$\mathcal P_+,\mathcal P_-$ & Probability distributions & Paired one-dimensional populations in the fixed-shot impossibility construction. & Appendix~\ref{app:convergence}, Proposition~\ref{prop:fixed_shot_no_go} \\
$\widehat\theta$ & Real scalar & Common optimizer output on the coupled all-zero support event. & Appendix~\ref{app:convergence}, Proposition~\ref{prop:fixed_shot_no_go} \\
$\varsigma_{\mathrm{ref},T,i}$ & Nonnegative scalar & Coordinatewise sub-Gaussian proxy scale for the independent reference gradient. & Appendix~\ref{app:convergence}, Proposition~\ref{prop:reference_gradient_certificate} \\
$s_{\mathrm{mgf}}$ & Real scalar & Moment-generating-function argument in the reference-gradient condition. & Appendix~\ref{app:convergence}, Proposition~\ref{prop:reference_gradient_certificate} \\
$\mathsf r_{\mathrm{ref},T}(\delta)\in\mathbb R_+^d$ & Nonnegative vector & Coordinatewise high-probability reference-gradient radius. & Appendix~\ref{app:convergence}, Proposition~\ref{prop:reference_gradient_certificate} \\
\addlinespace[2pt]
\multicolumn{4}{@{}l}{\textbf{Alignment, confidence, and dense-versus-sparse comparisons}}\\
\addlinespace[1pt]
$\mathfrak a_t^{\mathrm{dyn}}$ & Real scalar & Conditional population alignment of the dynamic blockwise sign oracle. & Eq.~\eqref{eq:dynamic_population_alignment}; Appendix~\ref{app:convergence} \\
$\mathfrak a_t^{\mathrm{stat}}$ & Real scalar & Conditional population alignment of the fixed-support sign direction. & Appendix~\ref{app:convergence}, Lemma~\ref{lem:appE:selection_reliability} \\
$\mathfrak a_t^{\mathrm{dense}}$ & Real scalar & Dense confidence-weighted population-gradient alignment mass. & Appendix~\ref{app:confidence_gating}, Eq.~\eqref{eq:cg_dense_alignment_mass} \\
$\pi_{t,i}^{\mathrm{sel}}$ & Probability in $[0,1]$ & Conditional probability that coordinate $i$ belongs to the dynamic support $S_t$. & Appendix~\ref{app:convergence}, Lemma~\ref{lem:appE:selection_reliability} \\
$\chi_{t,i}^{\mathrm{sel}}$ & Real scalar in $[-1,1]$ & Conditional selected-sign reliability, given that coordinate $i$ is selected. & Appendix~\ref{app:convergence}, Lemma~\ref{lem:appE:selection_reliability} \\
$\chi_{t,i}^{\mathrm{stat}}$ & Real scalar in $[-1,1]$ & Conditional sign reliability for a coordinate of the fixed support. & Appendix~\ref{app:convergence}, Lemma~\ref{lem:appE:selection_reliability} \\
$\chi_{t,i}^{\mathrm{dense}}$ & Real scalar in $[-1,1]$ & Unconditional dense-sign reliability relative to $h_{t,i}$. & Appendix~\ref{app:confidence_gating}, Eq.~\eqref{eq:cg_dense_reliability} \\
$p_{t,i}^{\mathrm{corr}\mid\mathrm{sel}}$ & Probability in $[0,1]$ & Conditional probability of a correct stochastic-gradient sign after selection. & Appendix~\ref{app:convergence}, Lemma~\ref{lem:appE:selection_reliability} \\
$p_{t,i}^{\mathrm{corr}}$ & Probability in $[0,1]$ & Unconditional probability of a correct stochastic-gradient sign. & Appendix~\ref{app:confidence_gating}, Eq.~\eqref{eq:cg_unconditional_correctness} \\
$\psi_{\mathrm{dyn}},\psi_{\mathrm{stat}}$ & Functions $[0,\infty)\to[0,\infty)$ & Calibration lower bounds relating population alignment to blockwise gradient mass. & Theorem~\ref{thm:calibrated_population_alignment} \\
$\omega_{\mathrm{dyn}},\omega_{\mathrm{stat}}>0$ & Real scalars & Positive alignment coefficients in power-law or relative-error calibration conditions. & Theorem~\ref{thm:calibrated_population_alignment}; Corollary~\ref{cor:relative_calibration_stationarity} \\
$\vartheta_{\mathrm{dyn}},\vartheta_{\mathrm{stat}}\ge1$ & Real exponents & Power-law exponents in the calibrated-alignment lower bounds. & Theorem~\ref{thm:calibrated_population_alignment} \\
$\upsilon_{\mathrm{shot}}^{\mathrm{dyn}},\upsilon_{\mathrm{mb}}^{\mathrm{dyn}}$ & Nonnegative real scalars & Relative dynamic finite-shot and minibatch errors. & Corollary~\ref{cor:relative_calibration_stationarity} \\
$\upsilon_{\mathrm{cap}}^{\mathrm{stat}}$ & Real scalar in $(0,1]$ & Relative population-gradient mass captured by the static support. & Corollary~\ref{cor:relative_calibration_stationarity} \\
$\upsilon_{\mathrm{shot}}^{\mathrm{stat}},\upsilon_{\mathrm{mb}}^{\mathrm{stat}}$ & Nonnegative real scalars & Relative static finite-shot and minibatch errors. & Corollary~\ref{cor:relative_calibration_stationarity} \\
$\sigma>0$ & Real scalar & Standard deviation in the one-dimensional fixed-variance Gaussian example. & Appendix~\ref{app:convergence}, \hyperlink{suppnav.alignment}{Sec.~F.7} \\
$Z$ & Random scalar & Gaussian or counterexample noise variable, with its distribution specified locally. & Appendix~\ref{app:convergence} \\
$\Phi$ & Function $\mathbb R\to[0,1]$ & Standard normal cumulative distribution function. & Appendix~\ref{app:convergence}, \hyperlink{suppnav.alignment}{Sec.~F.7} \\
$B_{\mathrm{traj}}>0$ & Real scalar & Bound on the one-dimensional trajectory in the Gaussian alignment example. & Appendix~\ref{app:convergence}, \hyperlink{suppnav.alignment}{Sec.~F.7} \\
$\omega_{B_{\mathrm{traj}},\sigma}>0$ & Real scalar & Quadratic alignment coefficient on the bounded Gaussian trajectory. & Appendix~\ref{app:convergence}, \hyperlink{suppnav.alignment}{Sec.~F.7} \\
$\varpi\in[0,1]$ & Real scalar & Continuous or binary attenuation factor applied to one sign coordinate. & Appendix~\ref{app:confidence_gating}, Lemma~\ref{lem:cg_mse_gate} \\
$S^\circ$ & Deterministic blockwise support & Support selected before the current stochastic-gradient draw for oracle comparisons. & Appendix~\ref{app:confidence_gating}, Theorem~\ref{thm:cg_oracle_support} \\
$\varsigma_{t,i}^{\mathrm{noise}}>0$ & Real scalar & Coordinatewise Gaussian stochastic-gradient noise scale. & Appendix~\ref{app:confidence_gating}, Eq.~\eqref{eq:cg_gaussian_noise} \\
$\varsigma_{t,r}^{\mathrm{noise}}>0$ & Real scalar & Common Gaussian noise scale within block $r$ in the ordering result. & Appendix~\ref{app:confidence_gating}, Lemma~\ref{lem:cg_gaussian_ordering} \\
$\kappa_{t,\mathrm{align}}^{\mathrm{dyn}}$ & Positive real ratio & Dynamic-to-dense confidence-weighted alignment ratio. & Appendix~\ref{app:confidence_gating}, Eq.~\eqref{eq:cg_dynamic_capture_fractions} \\
$\kappa_{t,\mathrm{curv}}^{\mathrm{dyn}}$ & Real ratio in $(0,1]$ & Expected dynamic selected-curvature fraction. & Appendix~\ref{app:confidence_gating}, Eq.~\eqref{eq:cg_dynamic_capture_fractions} \\
$\lambda_{t,\mathrm{dyn}}^\star,\lambda_{t,\mathrm{dense}}^\star$ & Positive real scalars & Locally optimal coordinate steps for the dynamic sparse-sign and dense-sign quadratic envelopes. & Appendix~\ref{app:confidence_gating}, Eq.~\eqref{eq:cg_tuned_optimal_steps} \\
$L_{\mathrm{coord}}>0$ & Real scalar & Common coordinatewise curvature in the homogeneous-curvature comparison. & Appendix~\ref{app:confidence_gating}, Corollary~\ref{cor:cg_homogeneous_curvature_phase} \\
$\mathcal J^{\mathrm{flat}}$ & Subset of $\{1,\ldots,d\}$ & Fixed set of zero-population-gradient coordinates in the random-walk analysis. & Appendix~\ref{app:confidence_gating}, Proposition~\ref{prop:cg_random_walk} \\
$\mathcal G_{t,i}^{\mathrm{loc}}(\varpi)$ & Real scalar & Coordinate contribution to the expected local smoothness-model decrease under attenuation $\varpi$. & Appendix~\ref{app:confidence_gating}, Eq.~\eqref{eq:cg_local_gain} \\
$\mathcal J_{t,r}^{\mathrm{rel}},\mathcal J_t^{\mathrm{rel}}$ & Blockwise and global index sets & Idealized reliable-coordinate sets in the polarized confidence model. & Appendix~\ref{app:confidence_gating}, Lemma~\ref{lem:cg_short_horizon} \\
$\epsilon_{\mathrm{pol}}\in[0,1/2)$ & Real scalar & Polarization tolerance separating reliable and unreliable confidence factors. & Appendix~\ref{app:confidence_gating}, Eq.~\eqref{eq:cg_polarized_confidence} \\
$\mathsf Y_{s,i}^{\mathrm{conf}}$ & Gaussian random scalar & Idealized noisy observation of $\chi_{t,i}^{\mathrm{dense}}$. & Appendix~\ref{app:confidence_gating}, Eq.~\eqref{eq:cg_confidence_observations} \\
$\varsigma_{\mathrm{est}}>0$ & Real scalar & Standard deviation of an idealized confidence observation. & Appendix~\ref{app:confidence_gating}, Eq.~\eqref{eq:cg_confidence_observations} \\
$\overline{\mathsf Y}_{i}^{\mathrm{conf}}$ & Real scalar & Sample mean of the $T_{\mathrm{obs}}$ confidence observations for coordinate $i$. & Appendix~\ref{app:confidence_gating}, Lemma~\ref{lem:cg_short_horizon} \\
$\widehat S_{t,r}^{\mathrm{conf}},\widehat S_t^{\mathrm{conf}}$ & Estimated blockwise and global supports & Top-$k_r$ hard confidence estimates from the sample means. & Appendix~\ref{app:confidence_gating}, Eq.~\eqref{eq:cg_blockwise_confidence_gate} \\
$\widehat m_t^{\mathrm{conf}}$ & Binary vector in $\{0,1\}^d$ & Mask of the estimated confidence support. & Appendix~\ref{app:confidence_gating}, Eq.~\eqref{eq:cg_blockwise_confidence_gate} \\
\addlinespace[2pt]
\multicolumn{4}{@{}l}{\textbf{Magnitude ranking and diagnostic quantities}}\\
\addlinespace[1pt]
$S_{t,r}^{\mathrm{pop}},S_t^{\mathrm{pop}}$ & Blockwise and global supports & Population-gradient top-$k_r$ supports under the fixed tie rule. & Appendix~\ref{app:magnitude_ranking}, Eq.~\eqref{eq:rank_population_support} \\
$\operatorname{gap}_{t,r}^{\mathrm{rank}}$ & Nonnegative real scalar & Difference between the $k_r$th and $(k_r+1)$th ordered population-gradient magnitudes in block $r$. & Appendix~\ref{app:magnitude_ranking}, Eq.~\eqref{eq:rank_boundary_gap} \\
$p_{\mathrm{mom}}>0$ & Real scalar & Moment order used in the high-probability ranking-recovery bound. & Appendix~\ref{app:magnitude_ranking}, Corollary~\ref{cor:rank_probability} \\
$\mathfrak M_{t,i}^{\mathrm{mom}}\ge0$ & Real scalar & Conditional moment budget for coordinate $i$ of the total gradient perturbation. & Appendix~\ref{app:magnitude_ranking}, Eq.~\eqref{eq:rank_moment_condition} \\
$\Theta_{\mathrm{img},t}$ & Architecture-dependent vector & Visual-encoder parameter vector at iteration $t$. & Appendix~\ref{sec:update_scale}, \hyperlink{suppnav.update-measure}{Sec.~L.1} \\
$d_{\mathrm{img}}$ & Positive integer & Number of visual-encoder parameters. & Appendix~\ref{sec:update_scale}, \hyperlink{suppnav.update-measure}{Sec.~L.1} \\
$\Delta\Theta_{\mathrm{img},t}$ & Vector with the same dimension as $\Theta_{\mathrm{img},t}$ & One-step visual-encoder update $\Theta_{\mathrm{img},t+1}-\Theta_{\mathrm{img},t}$. & Appendix~\ref{sec:update_scale}, \hyperlink{suppnav.update-measure}{Sec.~L.1} \\
$r_t^{\mathrm{rel}}$ & Nonnegative real scalar & Relative visual-encoder update norm $\|\Delta\Theta_{\mathrm{img},t}\|_2/\|\Theta_{\mathrm{img},t}\|_2$. & Appendix~\ref{sec:update_scale}, \hyperlink{suppnav.update-measure}{Sec.~L.1} \\
$r_t^{\mathrm{abs}}$ & Nonnegative real scalar & Mean absolute visual-encoder update $d_{\mathrm{img}}^{-1}\|\Delta\Theta_{\mathrm{img},t}\|_1$. & Appendix~\ref{sec:update_scale}, \hyperlink{suppnav.update-measure}{Sec.~L.1} \\
$k_{\mathrm{img}}$ & Positive integer & Number of active visual-encoder coordinates in the update-scale diagnostic. & Appendix~\ref{sec:update_scale}, \hyperlink{suppnav.update-measure}{Sec.~L.1} \\
$\Delta\Theta_t^{\mathrm{method}}$ & Vector in $\mathbb R^d$ & Parameter update produced by the named comparison method at iteration $t$. & Appendix~\ref{sec:update_scale}, \hyperlink{suppnav.update-normalization}{Sec.~L.2} \\
\end{longtable}
\endgroup
\clearpage

\section{Few-shot Adaptation of CLIP}
\label{app:clip_adaptation}
\suppanchor{clip}

Vision--language models learn a shared embedding space for visual and textual inputs. CLIP~\cite{radford2021learning} uses an image encoder $E_{\Theta_{\mathrm{img}}}$ and a text encoder $H_{\Theta_{\mathrm{txt}}}$. The trainable components of $\Theta_{\mathrm{img}}$ and $\Theta_{\mathrm{txt}}$ are concatenated into the single vector $\Theta\in\R^d$ used in Sec.~\ref{sec:method} and Appendix~\ref{app:convergence}; $\Theta_t$ denotes this vector at iteration $t$. For a pretraining pair $(x_{j_{\mathrm{data}}},x_{j_{\mathrm{data}}}^{\mathrm{txt}})$, the normalized embeddings are
\begin{equation}
\phi_{j_{\mathrm{data}}}^{\mathrm{img}}=E_{\Theta_{\mathrm{img}}}(x_{j_{\mathrm{data}}}),
\qquad
\phi_{j_{\mathrm{data}}}^{\mathrm{txt}}=H_{\Theta_{\mathrm{txt}}}(x_{j_{\mathrm{data}}}^{\mathrm{txt}}).
\end{equation}
CLIP aligns paired embeddings with a symmetric InfoNCE loss~\cite{oord2018representation}.

\paragraph{Zero-shot inference.}
The classification problem has $N$ classes. Let $M_{\mathrm{prompt}}$ be the number of text prompts per class, and let $x_{j_{\mathrm{prompt}},c}^{\mathrm{prompt}}$ be prompt $j_{\mathrm{prompt}}\in\{1,\ldots,M_{\mathrm{prompt}}\}$ for class $c\in\{1,\ldots,N\}$. The normalized text prototype for class $c$ is
\begin{equation}
\bar\phi_c^{\mathrm{txt}}
=\frac{1}{M_{\mathrm{prompt}}}\sum_{j_{\mathrm{prompt}}=1}^{M_{\mathrm{prompt}}}
H_{\Theta_{\mathrm{txt}}}(x_{j_{\mathrm{prompt}},c}^{\mathrm{prompt}}),
\qquad
\phi_c^{\mathrm{txt}}=\frac{\bar\phi_c^{\mathrm{txt}}}{\|\bar\phi_c^{\mathrm{txt}}\|_2}.
\end{equation}
For a query image $x_{j_{\mathrm{data}}}$, let $\phi_{j_{\mathrm{data}}}^{\mathrm{img}}=E_{\Theta_{\mathrm{img}}}(x_{j_{\mathrm{data}}})$ and let $\tau>0$ be the softmax temperature. CLIP assigns class probability
\begin{equation}
\hat y_{j_{\mathrm{data}},c}(\Theta)
=
\frac{\exp\!\big((\phi_{j_{\mathrm{data}}}^{\mathrm{img}})^\top \phi_c^{\mathrm{txt}}/\tau\big)}
{\sum_{c'=1}^{N}\exp\!\big((\phi_{j_{\mathrm{data}}}^{\mathrm{img}})^\top \phi_{c'}^{\mathrm{txt}}/\tau\big)}.
\label{y_ic}
\end{equation}

\paragraph{Few-shot adaptation.}
Let $K$ be the number of labeled examples per class and let $n=NK$. The support set is
\begin{equation}
D=\{z_{j_{\mathrm{data}}}=(x_{j_{\mathrm{data}}},y_{j_{\mathrm{data}}})\}_{j_{\mathrm{data}}=1}^{n},
\end{equation}
where $y_{j_{\mathrm{data}}}\in\{1,\ldots,N\}$ is the class label of $x_{j_{\mathrm{data}}}$. We use $K\in\{1,2,4,8,16\}$ in the experiments. Let $D_{\mathrm{query}}$ be a disjoint set used only for evaluation. The per-example cross-entropy loss and support empirical risk are
\begin{equation}
\ell(\Theta;z_{j_{\mathrm{data}}})=-\sum_{c=1}^{N}\1\{y_{j_{\mathrm{data}}}=c\}\log\hat y_{j_{\mathrm{data}},c}(\Theta),
\qquad
f(\Theta)=\frac1n\sum_{z_{j_{\mathrm{data}}}\in D}\ell(\Theta;z_{j_{\mathrm{data}}}).
\end{equation}
For an optimization minibatch $B_t\subseteq D$, let $b_t=|B_t|$. The minibatch risk is
\begin{equation}
\mathcal L_{B_t}(\Theta)=\frac1{b_t}\sum_{z_{j_{\mathrm{data}}}\in B_t}\ell(\Theta;z_{j_{\mathrm{data}}}).
\end{equation}
Few-shot adaptation minimizes this support objective and evaluates the adapted model on $D_{\mathrm{query}}$.

\section{Extended Related Work}
\label{sec:extended_related_work}
\suppanchor{related}

\noindent \textbf{Parameter-efficient fine-tuning (PEFT). } These methods adapt VLMs by updating a small subset of parameters, and span four families.

\emph{Prompt tuning} introduces learnable or constrained textual context (CoOp~\cite{zhou2022coop}, KgCoOp~\cite{yao2023kgcoop}, and ProGrad~\cite{zhu2022prograd}). These methods improve over manual prompts but can underfit challenging transfers, are sensitive to prompt length and regularization strength~\cite{silva2024closer}, and require many training epochs (often more than 200 epochs).

\emph{Classifier adaptation} constructs task-specific classifiers from frozen CLIP features. TIP-Adapter-F~\cite{zhang2022tipadapter} initializes cache keys with support-image embeddings and uses their one-hot labels as values. It optimizes the cache-key weights while keeping the CLIP encoders frozen. The resulting cache-based predictions are combined with zero-shot CLIP logits. GDA~\cite{wang2024hard} estimates class means and a shared covariance matrix under a Gaussian class-conditional model. It then constructs a closed-form discriminant classifier and ensembles it with zero-shot CLIP. ProKeR~\cite{bendou2025proker} interprets cache-based adaptation through kernel regression. It derives a globally regularized proximal kernel-ridge solution in a reproducing kernel Hilbert space. GDA and ProKeR avoid iterative optimization. All three methods restrict adaptation to predictors over fixed CLIP representations. Therefore, they do not refine the CLIP backbone or address optimization under noisy few-shot gradients. This restriction limits their learning capacity. In contrast, our methods adapt the pretrained model through sign updates without first- or second-moment buffers.

\emph{Low-rank reparameterization} constrains updates to low-rank subspaces. For instance, LoRA~\cite{hu2022lora} inserts trainable low-rank factors into each weight and optimizes only these factors while freezing the backbone. DoRA~\cite{liu2024dora} decouples weight direction and magnitude to improve expressiveness. PiSSA~\cite{meng2024pissa} initializes the low-rank factors from principal singular vectors/values of the pretrained weights to ease optimization. Although effective in some regimes, performance depends strongly on rank and schedule choices and can peak early, then degrade in few-shot settings~\cite{mrabah2025so}.

\emph{Selection-based} strategies update preselected parameters while freezing the rest. LayerNorm~\cite{ba2016layer} tunes scale parameters. It restricts updates to affine adjustments and thus may lead to underfitting on challenging datasets. SAFT~\cite{nguyen2024saft} fixes a mask from support-set gradient magnitudes and trains only these coordinates.

In the validation-free protocol used in Sec.~\ref{sec:experiments}, several of these baselines exhibit sensitivity to the evaluation setting (all-to-all vs.\ base-to-new), backbone, dataset, and shot level.

\noindent \textbf{Optimization-efficient training (OET). } Recent work has sought to reduce the memory and computational burden by modifying the optimization procedure rather than the learnable parameters. The training cost is reduced by compressing either the gradient or the optimizer states. GaLore \cite{zhao2024galore} and FLoRA \cite{hao2025flora} use low-rank compression to project the gradient into subspaces of reduced dimension. While GaLore computes and stores the projection matrices using singular value decomposition, FLoRA generates these matrices randomly during training. Although these projections constrain the optimization space, the update is still performed for all parameters at each step, which may accelerate overfitting in few-shot settings. These techniques were proposed for training large language models (LLMs) and have rarely been investigated for VLM few-shot adaptation.

Unlike the low-rank OET methods, SO~\cite{mrabah2025so} is specifically proposed for few-shot adaptation and can be viewed as an Adam-style optimizer in which both the gradients and the first- and second-moments are sparsified. SO maintains a sparse, periodically refreshed update support. It samples a few gradient coordinates uniformly at random and retains the top first-moment values by magnitude. In contrast to our state-free sign-sparse updates, SO still depends on first- and second-moment estimates and their associated hyperparameters. Thus, it introduces additional factors of sensitivity and optimizer-state overhead.

\section{Algorithms}
\label{app:algorithms}
\suppanchor{algorithms}

\paragraph{Blockwise selection convention.}
Every sparse algorithm in this appendix uses the same per-parameter-tensor selection rule as Algorithm~\ref{alg:signsparse} and the released implementation. Unless stated otherwise, absolute values, powers, square roots, and divisions of vectors are elementwise; $\oslash$ explicitly denotes elementwise division. The trainable coordinates are partitioned into blocks $\{\mathcal I_r\}_{r=1}^{R}$, with $d_r=|\mathcal I_r|$ and $k_r=\max\{1,\lfloor\rho d_r\rfloor\}$. A static method forms $S=\bigcup_{r=1}^{R}\Top_{k_r}(|g_{0,\mathcal I_r}|)$ once. A dynamic method forms $S_t=\bigcup_{r=1}^{R}\Top_{k_r}(|g_{t,\mathcal I_r}|)$ at every iteration. The pseudocode below states these blockwise operations explicitly.

\paragraph{SignSGD.}
SignSGD replaces the full gradient $g_t$ with its elementwise sign, defined in Eq.~\eqref{eq:sparse_sign_update}. At each iteration, it applies a fixed step in the direction of $\sgn(g_t)$, ignoring gradient magnitudes. This state-free optimizer reduces the impact of scale but updates all parameters at every step. The baseline uses one minibatch gradient and does not aggregate signs across workers. Majority voting is a separate distributed extension.

\begin{algorithm}[h]
\caption{SignSGD}
\label{alg:signsgd}
\small
\begin{algorithmic}[1]
\Require Initial parameters $\Theta_{0}$, learning rate $\eta$
\For{each training iteration $t = 0,1,2,\dots$}
    \State Compute dense gradient $g_t \leftarrow \nabla_{\Theta}\mathcal{L}_{B_t}(\Theta_t)$
    \State $\Theta_{t+1} \leftarrow \Theta_t - \eta \, \sgn(g_t)$ \hfill (elementwise, Eq.~\eqref{eq:sparse_sign_update})
\EndFor
\end{algorithmic}
\end{algorithm}

\paragraph{SignAdam.}
SignAdam keeps Adam's first-moment estimate $u_t$ but discards the magnitude of the adaptive step. Instead, it updates parameters using the elementwise sign of the first moment $u_t$. This yields a sign-only variant of Adam that still requires storing the moment.

\begin{algorithm}[h]
\caption{SignAdam}
\label{alg:signadam}
\small
\begin{algorithmic}[1]
\Require Initial parameters $\Theta_{0}$, learning rate $\eta$,
         moment coefficient $\beta_1\in[0,1)$
\State Initialize $u_0 \leftarrow 0 \in \mathbb{R}^{d}$
\For{each training iteration $t = 0,1,2,\dots$}
    \State Compute dense gradient $g_t \leftarrow \nabla_{\Theta}\mathcal{L}_{B_t}(\Theta_t)$
    \State $u_{t+1} \leftarrow \beta_1 u_t + (1-\beta_1) g_t$
    \State $\Theta_{t+1} \leftarrow \Theta_t - \eta \, \sgn(u_{t+1})$ \hfill (elementwise, Eq.~\eqref{eq:sparse_sign_update})
\EndFor
\end{algorithmic}
\end{algorithm}

\paragraph{SparseSGD.}
SparseSGD applies SGD updates only to a fixed subset of coordinates selected once at initialization. Given a density $\rho$, it selects the $k_r$ largest gradient magnitudes within every parameter tensor, unions the resulting block supports into $S$, and constructs a static mask $m$. At every iteration, the optimizer uses the sparse gradient $\widetilde g_t = m \odot g_t$ and updates only the active coordinates, while the remaining parameters stay frozen.

\begin{algorithm}[h]
\caption{SparseSGD (static mask)}
\label{alg:sparsesgd}
\small
\begin{algorithmic}[1]
\Require Initial parameters $\Theta_{0}$, learning rate $\eta$, density $\rho$
\Require Blocks $\{\mathcal I_r\}_{r=1}^{R}$ with $d_r=|\mathcal I_r|$
\State Set $k_r \leftarrow \max\{1,\lfloor\rho d_r\rfloor\}$ for every $r$
\State Compute initial gradient $g_0 \leftarrow \nabla_{\Theta}\mathcal{L}_{B_0}(\Theta_0)$
\State $S \leftarrow \bigcup_{r=1}^{R}\Top_{k_r}(|g_{0,\mathcal I_r}|)$
\State Build binary mask $m \in \{0,1\}^{d}$ from $S$ using Eq.~\eqref{eq:sparse_gradient}
\For{each training iteration $t = 0,1,2,\dots$}
    \State Compute dense gradient $g_t \leftarrow \nabla_{\Theta}\mathcal{L}_{B_t}(\Theta_t)$
    \State $\widetilde g_t \leftarrow m \odot g_t$ \hfill (sparse gradient, Eq.~\eqref{eq:sparse_gradient})
    \State $\Theta_{t+1} \leftarrow \Theta_t - \eta \, \widetilde g_t$
\EndFor
\end{algorithmic}
\end{algorithm}

\paragraph{DynSparseSGD.}
DynSparseSGD re-applies the blockwise parameter-selection step at every iteration. Given density $\rho$, it selects the $k_r$ largest current-gradient magnitudes within each parameter tensor, unions them into $S_t$, builds the mask $m_t$, forms the sparse gradient $\widetilde g_t = m_t \odot g_t$, and performs an SGD step on this support.
The active subset, therefore, changes over time, increasing coverage of the parameter space while keeping each update sparse.

\begin{algorithm}[h]
\caption{DynSparseSGD (dynamic mask)}
\label{alg:dynsparsesgd}
\small
\begin{algorithmic}[1]
\Require Initial parameters $\Theta_{0}$, learning rate $\eta$, density $\rho$
\Require Blocks $\{\mathcal I_r\}_{r=1}^{R}$ with $d_r=|\mathcal I_r|$
\State Set $k_r \leftarrow \max\{1,\lfloor\rho d_r\rfloor\}$ for every $r$
\For{each training iteration $t = 0,1,2,\dots$}
    \State Compute dense gradient $g_t \leftarrow \nabla_{\Theta}\mathcal{L}_{B_t}(\Theta_t)$
    \State $S_t \leftarrow \bigcup_{r=1}^{R}\Top_{k_r}(|g_{t,\mathcal I_r}|)$
    \State Build binary mask $m_t \in \{0,1\}^{d}$ from $S_t$ using Eq.~\eqref{eq:sparse_gradient}
    \State $\widetilde g_t \leftarrow m_t \odot g_t$ \hfill (sparse gradient, Eq.~\eqref{eq:sparse_gradient})
    \State $\Theta_{t+1} \leftarrow \Theta_t - \eta \, \widetilde g_t$
\EndFor
\end{algorithmic}
\end{algorithm}

\paragraph{SparseAdam.}

SparseAdam is an Adam optimizer restricted to a fixed sparse support. Using the same blockwise initialization as SparseSGD, it selects a static mask $m$ from $g_0$ and maintains first and second moments $u_t, v_t$ over the selected coordinates. At each step, it computes the sparse gradient
$\widetilde g_t = m \odot g_t$, updates $u_t$ and $v_t$ using $\widetilde g_t$, and applies the Adam step only on the active coordinates.

\begin{algorithm}[h]
\caption{SparseAdam (static mask)}
\label{alg:sparseadam}
\small
\begin{algorithmic}[1]
\Require Initial parameters $\Theta_{0}$, learning rate $\eta$,
         moment coefficients $(\beta_1,\beta_2)\in[0,1)^2$, $\varepsilon>0$, density $\rho$
\Require Blocks $\{\mathcal I_r\}_{r=1}^{R}$ with $d_r=|\mathcal I_r|$
\State Set $k_r \leftarrow \max\{1,\lfloor\rho d_r\rfloor\}$ for every $r$
\State Compute initial gradient $g_0 \leftarrow \nabla_{\Theta}\mathcal{L}_{B_0}(\Theta_0)$
\State $S \leftarrow \bigcup_{r=1}^{R}\Top_{k_r}(|g_{0,\mathcal I_r}|)$
\State Build binary mask $m \in \{0,1\}^{d}$ from $S$ using Eq.~\eqref{eq:sparse_gradient}
\State Initialize $u_0 \leftarrow 0 \in \mathbb{R}^{d}$, $v_0 \leftarrow 0 \in \mathbb{R}^{d}$
\For{each training iteration $t = 0,1,2,\dots$}
    \State Compute dense gradient $g_t \leftarrow \nabla_{\Theta}\mathcal{L}_{B_t}(\Theta_t)$
    \State $\widetilde g_t \leftarrow m \odot g_t$ \hfill (sparse gradient, Eq.~\eqref{eq:sparse_gradient})
    \State $u_{t+1} \leftarrow \beta_1 u_t + (1-\beta_1)\, \widetilde g_t$
    \State $v_{t+1} \leftarrow \beta_2 v_t + (1-\beta_2)\, (\widetilde g_t \odot \widetilde g_t)$
    \State $\hat{u}_{t+1} \leftarrow u_{t+1} / (1-\beta_1^{t+1})$, \quad
           $\hat{v}_{t+1} \leftarrow v_{t+1} / (1-\beta_2^{t+1})$
    \State $\Theta_{t+1} \leftarrow \Theta_t - \eta \, \hat{u}_{t+1} \oslash \big(\sqrt{\hat{v}_{t+1}} + \varepsilon\big)$
\EndFor
\end{algorithmic}
\end{algorithm}

\begin{algorithm}[!h]
\caption{DynSparseAdam (dynamic mask)}
\label{alg:dynsparseadam}
\small
\begin{algorithmic}[1]
\Require Initial parameters $\Theta_{0}$, learning rate $\eta$,
         moment coefficients $(\beta_1,\beta_2)\in[0,1)^2$, $\varepsilon>0$, density $\rho$
\Require Blocks $\{\mathcal I_r\}_{r=1}^{R}$ with $d_r=|\mathcal I_r|$
\State Set $k_r \leftarrow \max\{1,\lfloor\rho d_r\rfloor\}$ for every $r$
\State Initialize $u_0 \leftarrow 0 \in \mathbb{R}^{d}$, $v_0 \leftarrow 0 \in \mathbb{R}^{d}$
\For{each training iteration $t = 0,1,2,\dots$}
    \State Compute dense gradient $g_t \leftarrow \nabla_{\Theta}\mathcal{L}_{B_t}(\Theta_t)$
    \State $S_t \leftarrow \bigcup_{r=1}^{R}\Top_{k_r}(|g_{t,\mathcal I_r}|)$
    \State Build binary mask $m_t \in \{0,1\}^{d}$ from $S_t$ using Eq.~\eqref{eq:sparse_gradient}
    \State $\widetilde g_t \leftarrow m_t \odot g_t$ \hfill (sparse gradient, Eq.~\eqref{eq:sparse_gradient})
    \State $u_{t+1} \leftarrow \beta_1 u_t + (1-\beta_1)\, \widetilde g_t$
    \State $v_{t+1} \leftarrow \beta_2 v_t + (1-\beta_2)\, (\widetilde g_t \odot \widetilde g_t)$
    \State $\hat{u}_{t+1} \leftarrow u_{t+1} / (1-\beta_1^{t+1})$, \quad
           $\hat{v}_{t+1} \leftarrow v_{t+1} / (1-\beta_2^{t+1})$
    \State $\Theta_{t+1} \leftarrow \Theta_t - \eta \, m_t \odot \Big(\hat{u}_{t+1} \oslash \big(\sqrt{\hat{v}_{t+1}} + \varepsilon\big)\Big)$
\EndFor
\end{algorithmic}
\end{algorithm}

\paragraph{DynSparseAdam.}
DynSparseAdam refreshes the blockwise sparse support at every iteration, combining Adam's adaptive moments with dynamic parameter selection. At step $t$, it selects within every parameter tensor, unions the selected coordinates into $S_t$, builds $m_t$, forms $\widetilde g_t = m_t \odot g_t$, updates the moments using this masked gradient, and applies the Adam step only on the current active coordinates. This variant is the dynamic counterpart of SparseAdam.

\paragraph{SignSparseAdam.}
SignSparseAdam combines Adam-type first-moment accumulation with a static sparse support and sign-only updates. A fixed blockwise mask $m$ is computed once from $g_0$ as in SparseSGD and SparseAdam. At each iteration, the optimizer updates a first-moment estimate $u_t$ using the sparse gradient $\widetilde g_t = m \odot g_t$ and applies the sign operator to the first-moment $u_t$. Only coordinates in the static support receive sign-based updates. All other parameters remain frozen.

\begin{algorithm}[h]
\caption{SignSparseAdam (static mask)}
\label{alg:signsparseadam}
\small
\begin{algorithmic}[1]
\Require Initial parameters $\Theta_{0}$, learning rate $\eta$,
         momentum $\beta_1 \in (0,1)$, density $\rho$
\Require Blocks $\{\mathcal I_r\}_{r=1}^{R}$ with $d_r=|\mathcal I_r|$
\State Set $k_r \leftarrow \max\{1,\lfloor\rho d_r\rfloor\}$ for every $r$
\State Compute initial gradient $g_0 \leftarrow \nabla_{\Theta}\mathcal{L}_{B_0}(\Theta_0)$
\State $S \leftarrow \bigcup_{r=1}^{R}\Top_{k_r}(|g_{0,\mathcal I_r}|)$
\State Build binary mask $m \in \{0,1\}^{d}$ from $S$ using Eq.~\eqref{eq:sparse_gradient}
\State Initialize $u_0 \leftarrow 0 \in \mathbb{R}^{d}$
\For{each training iteration $t = 0,1,2,\dots$}
    \State Compute dense gradient $g_t \leftarrow \nabla_{\Theta}\mathcal{L}_{B_t}(\Theta_t)$
    \State $\widetilde g_t \leftarrow m \odot g_t$ \hfill (sparse gradient, Eq.~\eqref{eq:sparse_gradient})
    \State $u_{t+1} \leftarrow \beta_1 u_t + (1-\beta_1)\, \widetilde g_t$
    \State $\Theta_{t+1} \leftarrow \Theta_t - \eta \, \sgn(u_{t+1})$ \hfill (elementwise, Eq.~\eqref{eq:sparse_sign_update})
\EndFor
\end{algorithmic}
\end{algorithm}

\paragraph{GradNorm SignSGD.}
GradNorm SignSGD is a norm-matched variant of SignSGD. At each iteration, it forms the sign vector $\sgn(g_t)$ and rescales it so that the $\ell_2$ norm of the sign update matches the $\ell_2$ norm of the dense SGD update computed from $g_t$. The direction therefore remains sign-only, but its overall magnitude follows the norm of the raw gradient.

\begin{algorithm}[h]
\caption{GradNorm SignSGD}
\label{alg:gradnormsignsgd}
\small
\begin{algorithmic}[1]
\Require Initial parameters $\Theta_{0}$, learning rate $\eta$
\For{each training iteration $t = 0,1,2,\dots$}
    \State Compute dense gradient $g_t \leftarrow \nabla_{\Theta}\mathcal{L}_{B_t}(\Theta_t)$
    \If{$\|\sgn(g_t)\|_2 = 0$}
        \State $a_t \leftarrow 0$
    \Else
        \State $a_t \leftarrow \|g_t\|_2 / \|\sgn(g_t)\|_2$ \hfill (norm-matching factor)
    \EndIf
    \State $\Theta_{t+1} \leftarrow \Theta_t - \eta \, a_t \, \sgn(g_t)$ \hfill (Eq.~\eqref{eq:sparse_sign_update})
\EndFor
\end{algorithmic}
\end{algorithm}

\paragraph{GradNorm SparseSGD.}
GradNorm SparseSGD is a norm-matched version of SparseSGD with a fixed blockwise sparse support. It uses the same static mask $m$ as SparseSGD, forms the sparse gradient $\widetilde g_t = m \odot g_t$, and rescales this sparse direction so that its $\ell_2$ norm matches that of the corresponding dense SGD update whenever $\|\widetilde g_t\|_2>0$. When $\widetilde g_t=0$, the algorithm uses $a_t=0$ and returns a zero update, which need not match a nonzero dense update. This preserves sparse updates while compensating for the reduction in update magnitude caused by masking.

\begin{algorithm}[h]
\caption{GradNorm SparseSGD (static mask)}
\label{alg:gradnormsparsesgd}
\small
\begin{algorithmic}[1]
\Require Initial parameters $\Theta_{0}$, learning rate $\eta$, density $\rho$
\Require Blocks $\{\mathcal I_r\}_{r=1}^{R}$ with $d_r=|\mathcal I_r|$
\State Set $k_r \leftarrow \max\{1,\lfloor\rho d_r\rfloor\}$ for every $r$
\State Compute initial gradient $g_0 \leftarrow \nabla_{\Theta}\mathcal{L}_{B_0}(\Theta_0)$
\State $S \leftarrow \bigcup_{r=1}^{R}\Top_{k_r}(|g_{0,\mathcal I_r}|)$
\State Build binary mask $m \in \{0,1\}^{d}$ from $S$ using Eq.~\eqref{eq:sparse_gradient}
\For{each training iteration $t = 0,1,2,\dots$}
    \State Compute dense gradient $g_t \leftarrow \nabla_{\Theta}\mathcal{L}_{B_t}(\Theta_t)$
    \State $\widetilde g_t \leftarrow m \odot g_t$ \hfill (sparse gradient, Eq.~\eqref{eq:sparse_gradient})
    \If{$\|\widetilde g_t\|_2 = 0$}
        \State $a_t \leftarrow 0$
    \Else
        \State $a_t \leftarrow \|g_t\|_2 / \|\widetilde g_t\|_2$ \hfill (norm-matching factor)
    \EndIf
    \State $\Theta_{t+1} \leftarrow \Theta_t - \eta \, a_t \, \widetilde g_t$
\EndFor
\end{algorithmic}
\end{algorithm}

\paragraph{Local-clip SparseSGD.}
Local-clip SparseSGD is a clipped version of SparseSGD with a fixed blockwise sparse support. After forming the sparse gradient $\widetilde g_t = m \odot g_t$, it clips each active coordinate independently to the interval $[-c_{\mathrm{loc}},c_{\mathrm{loc}}]$, where $c_{\mathrm{loc}}>0$. Formally, $\operatorname{clip}(v_i;c_{\mathrm{loc}})=\min\{\max\{v_i,-c_{\mathrm{loc}}\},c_{\mathrm{loc}}\}$ acts elementwise. This limits large coordinate-wise updates while retaining the magnitudes of smaller active gradients.

\begin{algorithm}[h]
\caption{Local-clip SparseSGD (static mask)}
\label{alg:localclipsparsesgd}
\small
\begin{algorithmic}[1]
\Require Initial parameters $\Theta_{0}$, learning rate $\eta$, density $\rho$, clipping threshold $c_{\mathrm{loc}}>0$
\Require Blocks $\{\mathcal I_r\}_{r=1}^{R}$ with $d_r=|\mathcal I_r|$
\State Set $k_r \leftarrow \max\{1,\lfloor\rho d_r\rfloor\}$ for every $r$
\State Compute initial gradient $g_0 \leftarrow \nabla_{\Theta}\mathcal{L}_{B_0}(\Theta_0)$
\State $S \leftarrow \bigcup_{r=1}^{R}\Top_{k_r}(|g_{0,\mathcal I_r}|)$
\State Build binary mask $m \in \{0,1\}^{d}$ from $S$ using Eq.~\eqref{eq:sparse_gradient}
\For{each training iteration $t = 0,1,2,\dots$}
    \State Compute dense gradient $g_t \leftarrow \nabla_{\Theta}\mathcal{L}_{B_t}(\Theta_t)$
    \State $\widetilde g_t \leftarrow m \odot g_t$ \hfill (sparse gradient, Eq.~\eqref{eq:sparse_gradient})
    \State $g_t^{\mathrm{loc}} \leftarrow \operatorname{clip}(\widetilde g_t;c_{\mathrm{loc}})$ \hfill (elementwise clamp)
    \State $\Theta_{t+1} \leftarrow \Theta_t - \eta \, g_t^{\mathrm{loc}}$
\EndFor
\end{algorithmic}
\end{algorithm}

\paragraph{Global-clip SparseSGD.}
Global-clip SparseSGD applies global $\ell_2$ clipping before the blockwise sparse SGD update. At each iteration, it rescales the dense gradient $g_t$ by a single factor so that its global $\ell_2$ norm does not exceed a threshold $c_{\mathrm{glob}}>0$, then applies the static mask $m$ and performs the SGD step on the clipped sparse gradient. This preserves the relative magnitudes of the dense gradient before sparsification while controlling the overall update size.

\begin{algorithm}[h]
\caption{Global-clip SparseSGD (static mask)}
\label{alg:globalclipsparsesgd}
\small
\begin{algorithmic}[1]
\Require Initial parameters $\Theta_{0}$, learning rate $\eta$, density $\rho$, norm threshold $c_{\mathrm{glob}}>0$
\Require Blocks $\{\mathcal I_r\}_{r=1}^{R}$ with $d_r=|\mathcal I_r|$
\State Set $k_r \leftarrow \max\{1,\lfloor\rho d_r\rfloor\}$ for every $r$
\State Compute initial gradient $g_0 \leftarrow \nabla_{\Theta}\mathcal{L}_{B_0}(\Theta_0)$
\State $S \leftarrow \bigcup_{r=1}^{R}\Top_{k_r}(|g_{0,\mathcal I_r}|)$
\State Build binary mask $m \in \{0,1\}^{d}$ from $S$ using Eq.~\eqref{eq:sparse_gradient}
\For{each training iteration $t = 0,1,2,\dots$}
    \State Compute dense gradient $g_t \leftarrow \nabla_{\Theta}\mathcal{L}_{B_t}(\Theta_t)$
    \If{$\|g_t\|_2 = 0$}
        \State $a_t^{\mathrm{glob}} \leftarrow 1$
    \Else
        \State $a_t^{\mathrm{glob}} \leftarrow \min\{1, c_{\mathrm{glob}} / \|g_t\|_2\}$ \hfill (global clipping factor)
    \EndIf
    \State $g_t^{\mathrm{glob}} \leftarrow a_t^{\mathrm{glob}} g_t$ \hfill (globally clipped dense gradient)
    \State $\widetilde g_t^{\mathrm{glob}} \leftarrow m \odot g_t^{\mathrm{glob}}$ \hfill (sparse clipped gradient)
    \State $\Theta_{t+1} \leftarrow \Theta_t - \eta \, \widetilde g_t^{\mathrm{glob}}$
\EndFor
\end{algorithmic}
\end{algorithm}

\section{Hyperparameters}
\label{app:hyperparameters}
\suppanchor{hyperparameters}

\noindent\textbf{Shared image processing pipeline.}
We use a single training-time image preprocessing pipeline for \emph{all} methods and datasets. The pipeline follows standard CLIP-style few-shot recipes based on random resized cropping and random horizontal flipping. This keeps the training protocol comparable across baselines.

\begin{table}[h]
\caption{Shared training-time data augmentation pipeline (applied uniformly across all methods).}
\centering
\label{tab:aug-pipeline}
\begin{tabular}{p{0.30\textwidth}l}
\toprule
Operation & Setting \\
\midrule
RandomResizedCrop & size $=224$, scale $=(0.08,\,1.0)$, interpolation $=$ bicubic \\
RandomHorizontalFlip & probability $p_{\mathrm{flip}}=0.5$ \\
\bottomrule
\end{tabular}
\end{table}

\noindent Unless otherwise stated, low-rank, selection-based, and optimization-based methods are applied to all layers of the vision and text Transformer blocks. For optimization-based PEFT methods, we follow the default optimizer of the corresponding implementation: Adam for LayerNorm, CoOp, ProGrad, TIP-Adapter-F, and LoRA, and SGD for KgCoOp. GDA and ProKeR are training-free: both construct a classifier from frozen support features and therefore use no optimizer, learning rate, or epoch schedule.

\noindent\textbf{LayerNorm.}
LayerNorm tuning updates only the affine scale and shift parameters in normalization layers. Its main hyperparameters are the learning rate $\eta$, the number of epochs $E_{\mathrm{train}}$, and the training batch size $b_{\mathrm{train}}$.

\begin{table}[!h]
\centering
\caption{Hyperparameters for LayerNorm.}
\label{tab:hp-layernorm}
\begin{tabular}{ll}
\toprule
Name (symbol) & Value \\
\midrule
Learning rate ($\eta$) & $1\times 10^{-4}$ \\
Epochs ($E_{\mathrm{train}}$) & $30$ \\
Batch size ($b_{\mathrm{train}}$) & $64$ \\
Optimizer & Adam \\
\bottomrule
\end{tabular}
\end{table}

\noindent\textbf{CoOp.}
CoOp learns a small set of continuous context tokens per class. To distinguish the trainable context length from the zero-shot prompt count $M_{\mathrm{prompt}}$ in Appendix~\ref{app:clip_adaptation}, we denote the context length by $M_{\mathrm{ctx}}$.

\begin{table}[!h]
\centering
\caption{Hyperparameters for CoOp.}
\label{tab:hp-coop}
\begin{tabular}{ll}
\toprule
Name (symbol) & Value \\
\midrule
Context length ($M_{\mathrm{ctx}}$) & $16$ \\
Learning rate ($\eta$) & $1\times 10^{-4}$ \\
Epochs ($E_{\mathrm{train}}$) & $200$ \\
Batch size ($b_{\mathrm{train}}$) & $64$ \\
Optimizer & Adam \\
\bottomrule
\end{tabular}
\end{table}

\noindent\textbf{KgCoOp.}
KgCoOp augments CoOp with a knowledge-guided regularizer of strength $\zeta_1$. We again denote the prompt context length by $M_{\mathrm{ctx}}$. Our configuration uses $M_{\mathrm{ctx}}=16$ and $\zeta_1=8.0$.

\begin{table}[!h]
\centering
\caption{Hyperparameters for KgCoOp.}
\label{tab:hp-kgcoop}
\begin{tabular}{ll}
\toprule
Name (symbol) & Value \\
\midrule
Context length ($M_{\mathrm{ctx}}$) & $16$ \\
Knowledge weight ($\zeta_1$) & $8.0$ \\
Learning rate ($\eta$) & $1\times 10^{-4}$ \\
Epochs ($E_{\mathrm{train}}$) & $200$ \\
Batch size ($b_{\mathrm{train}}$) & $64$ \\
Optimizer & SGD \\
\bottomrule
\end{tabular}
\end{table}

\noindent\textbf{ProGrad.}
ProGrad projects the prompt gradient onto a direction that preserves pretrained knowledge. Its method-specific hyperparameters are the context length $M_{\mathrm{ctx}}$, the projection weight $\zeta_2$, and the CLIP temperature $\tau$. Our configuration uses $M_{\mathrm{ctx}}=16$, $\zeta_2=1.0$, and $\tau=1.0$.

\begin{table}[!h]
\centering
\caption{Hyperparameters for ProGrad.}
\label{tab:hp-prograd}
\begin{tabular}{ll}
\toprule
Name (symbol) & Value \\
\midrule
Context length ($M_{\mathrm{ctx}}$) & $16$ \\
Projection weight ($\zeta_2$) & $1.0$ \\
Temperature ($\tau$) & $1.0$ \\
Learning rate ($\eta$) & $1\times 10^{-4}$ \\
Epochs ($E_{\mathrm{train}}$) & $200$ \\
Batch size ($b_{\mathrm{train}}$) & $64$ \\
Optimizer & Adam \\
\bottomrule
\end{tabular}
\end{table}

\noindent\textbf{GDA.}
GDA~\cite{wang2024hard} is training-free. It extracts normalized support features with the frozen CLIP image encoder over 10 augmented passes, estimates the class means and a shared precision matrix with the Kubokawa--Srivastava empirical-Bayes ridge estimator, and combines the resulting discriminant scores with the zero-shot CLIP logits. The selection rule is fixed across datasets, backbones, and shot levels: the ensemble coefficient is selected on the validation split from $\{10^{-4},10^{-3},10^{-2},10^{-1},1,10,10^2\}$. When validation samples are unavailable, the implementation uses $0.1$. The support-feature and evaluation batch sizes are 64 and 256, respectively. A numerical ridge of $10^{-6}$ is used only if the precision system is degenerate. For base-to-new evaluation, the feature-synthesis extension uses 64 nearest neighbors. GDA uses no optimizer, learning rate, or training epochs.

\begin{table}[!h]
\centering
\caption{Implementation settings for GDA. The selection rule and candidate set are shared across all experimental settings.}
\label{tab:hp-gda}
\begin{tabularx}{\textwidth}{@{}p{0.39\textwidth}>{\raggedright\arraybackslash}X@{}}
\toprule
Setting & Value \\
\midrule
Support augmentation passes & $10$ \\
Precision estimator & Kubokawa--Srivastava empirical-Bayes ridge \\
Ensemble-coefficient selection & Validation split \\
Candidate set & $\{10^{-4},10^{-3},10^{-2},10^{-1},1,10,10^2\}$ \\
Fallback ensemble coefficient & $0.1$ \\
Support / evaluation batch sizes  & $64$ / $256$ \\
Numerical ridge & $10^{-6}$, only for a degenerate precision system \\
Base-to-new nearest-neighbor count & $64$ \\
Optimization & Closed form; no optimizer, learning rate, or epochs \\
\bottomrule
\end{tabularx}
\end{table}

\noindent\textbf{TIP-Adapter-F.}
TIP-Adapter-F optimizes cache-key weights and blends cache-based scores with the frozen CLIP branch. We denote the cache-mixing and cache-scaling weights by $\iota_{\mathrm{TIP,mix}}$ and $\iota_{\mathrm{TIP,scale}}$, respectively. Our configuration uses $\iota_{\mathrm{TIP,mix}}=0.27$ and $\iota_{\mathrm{TIP,scale}}=20$.

\begin{table}[!h]
\centering
\caption{Hyperparameters for TIP-Adapter-F.}
\label{tab:hp-tipadapter}
\begin{tabular}{ll}
\toprule
Name (symbol) & Value \\
\midrule
Cache mix ($\iota_{\mathrm{TIP,mix}}$) & $0.27$ \\
Cache scale ($\iota_{\mathrm{TIP,scale}}$) & $20.0$ \\
Learning rate ($\eta$) & $5\times 10^{-4}$ \\
Epochs ($E_{\mathrm{train}}$) & $10$ \\
Batch size ($b_{\mathrm{train}}$) & $64$ \\
Optimizer & Adam \\
\bottomrule
\end{tabular}
\end{table}

\noindent\textbf{ProKeR.}
ProKeR~\cite{bendou2025proker} is training-free and solves proximal kernel ridge regression in closed form on frozen, normalized CLIP support features. We use the official ImageNet-transferred settings implemented in the repository. They depend only on the shot count $K$ and are kept unchanged across datasets and CLIP backbones. The support cache is constructed over 10 augmented passes with a feature-extraction batch size of 64, and predictions are evaluated in batches of 256. The optional diagonal jitter is set to zero. ProKeR uses no optimizer, learning rate, or training epochs.

\begin{table}[!h]
\centering
\caption{Shot-dependent ProKeR settings. The command-line names are reported to avoid introducing mathematical symbols that conflict with the notation of the paper.}
\label{tab:hp-proker}
\begin{tabular}{ccl}
\toprule
Shots ($K$) & \texttt{-{}-beta} & \texttt{-{}-lmbda} \\
\midrule
$1$  & $3.00$ & $0.05$ \\
$2$  & $2.60$ & $0.05$ \\
$4$  & $1.50$ & $0.05$ \\
$8$  & $1.66$ & $0.07$ \\
$16$ & $1.70$ & $0.10$ \\
\bottomrule
\end{tabular}
\end{table}

\noindent\textbf{LoRA.}
LoRA replaces a dense update by a low-rank reparameterization. We use the rank $r_{\mathrm{LoRA}}$ and a method-specific scaling factor $\alpha_{\mathrm{LoRA}}$ to distinguish it from the sparse sign scale $\alpha$ in Eq.~\eqref{eq:sparse_sign_update}. Following CLIP-LoRA, we set $r_{\mathrm{LoRA}}=2$ and $\alpha_{\mathrm{LoRA}}=1$.

\begin{table}[!h]
\centering
\caption{Hyperparameters for LoRA.}
\label{tab:hp-lora}
\begin{tabular}{ll}
\toprule
Name (symbol) & Value \\
\midrule
Rank ($r_{\mathrm{LoRA}}$) & $2$ \\
Scaling ($\alpha_{\mathrm{LoRA}}$) & $1$ \\
Learning rate ($\eta$) & $1\times 10^{-4}$ \\
Epochs ($E_{\mathrm{train}}$) & $30$ \\
Batch size ($b_{\mathrm{train}}$) & $64$ \\
Optimizer & Adam \\
\bottomrule
\end{tabular}
\end{table}

\noindent\textbf{GaLoRE.}
GaLoRE projects gradients onto a low-rank subspace that is refreshed periodically. Its main hyperparameters are the subspace rank $r_{\mathrm{GaLoRE}}$ and the refresh period $p_{\mathrm{refresh}}$, measured in optimizer iterations.

\begin{table}[!h]
\centering
\caption{Hyperparameters for GaLoRE.}
\label{tab:hp-galore}
\begin{tabular}{ll}
\toprule
Name (symbol) & Value \\
\midrule
Subspace rank ($r_{\mathrm{GaLoRE}}$) & $2$ \\
Refresh period ($p_{\mathrm{refresh}}$) & $10$ \\
Learning rate ($\eta$) & $1\times 10^{-4}$ \\
Epochs ($E_{\mathrm{train}}$) & $30$ \\
Batch size ($b_{\mathrm{train}}$) & $64$ \\
\bottomrule
\end{tabular}
\end{table}

\noindent\textbf{SO.}
SO is an Adam-style sparse optimizer that sparsifies both gradients and moment states on a periodically refreshed support. The support update combines uniform random sampling of gradient coordinates with retention of the largest first-moment entries by magnitude. The implementation exposes a sparsity ratio $s_{\mathrm{SO}}$, but to stay consistent with Sec.~\ref{sec:method} we report the equivalent density parameter $\rho = 1-s_{\mathrm{SO}}$. Our configuration uses $\rho=0.001$ ($0.1\%$), $(\beta_1,\beta_2)=(0.9,0.999)$, $\varepsilon=10^{-8}$, and a support-refresh period $p_{\mathrm{refresh}}=10$ iterations.

\begin{table}[!h]
\centering
\caption{Hyperparameters for SO.}
\label{tab:hp-so}
\begin{tabular}{ll}
\toprule
Name (symbol) & Value \\
\midrule
Density parameter ($\rho$) & $0.001$ ($0.1\%$) \\
Support update & uniform random + top first-moment \\
Refresh period ($p_{\mathrm{refresh}}$) & $10$ \\
First moment ($\beta_1$) & $0.9$ \\
Second moment ($\beta_2$) & $0.999$ \\
Epsilon ($\varepsilon$) & $1\times 10^{-8}$ \\
Learning rate ($\eta$) & $1\times 10^{-4}$ \\
Epochs ($E_{\mathrm{train}}$) & $30$ \\
Batch size ($b_{\mathrm{train}}$) & $64$ \\
\bottomrule
\end{tabular}
\end{table}

\noindent \textbf{Optimization baselines and sign/sparse variants.}
As stated in the main paper, all optimization-based methods reported in this appendix use the same training schedule: $E_{\mathrm{train}}=30$ epochs, $b_{\mathrm{train}}=64$, and $\eta=10^{-4}$. Following Appendix~\ref{app:algorithms}, we denote the sparse support by $S$ or $S_t$, the corresponding binary mask by $m$ or $m_t$, and the sparse gradient by $\widetilde g_t$. The standard map $\sgn$ in Eq.~\eqref{eq:sparse_sign_update} has unit nonzero entries. Dense sign baselines use the coordinate step $\eta$, whereas SignSparseSGD and DynSignSparseSGD use $\lambda=\alpha\eta$ with $\alpha=0.1$ as in Eq.~\eqref{eq:sparse_sign_update}. In our study, SAFT is equivalent to SparseAdam with a static top-$k$ support selected from the support-set gradient. Therefore, the hyperparameters in Table~\ref{tab:hp-sparseadam} also apply to SAFT. We also include DynSparseSGD and DynSparseAdam, the dynamic counterparts of SparseSGD and SparseAdam defined in Appendix~\ref{app:algorithms}, under the same shared schedule.

\begin{table}[!h]
\centering
\caption{Hyperparameters for SignSGD.}
\label{tab:hp-signsgd}
\begin{tabular}{ll}
\toprule
Name (symbol) & Value \\
\midrule
Learning rate ($\eta$) & $1\times 10^{-4}$ \\
Epochs ($E_{\mathrm{train}}$) & $30$ \\
Batch size ($b_{\mathrm{train}}$) & $64$ \\
\bottomrule
\end{tabular}
\end{table}

\begin{table}[!h]
\centering
\caption{Hyperparameters for SignAdam.}
\label{tab:hp-signadam}
\begin{tabular}{ll}
\toprule
Name (symbol) & Value \\
\midrule
First moment ($\beta_1$) & $0.9$ \\
Learning rate ($\eta$) & $1\times 10^{-4}$ \\
Epochs ($E_{\mathrm{train}}$) & $30$ \\
Batch size ($b_{\mathrm{train}}$) & $64$ \\
\bottomrule
\end{tabular}
\end{table}

\begin{table}[!h]
\centering
\caption{Hyperparameters for SparseSGD.}
\label{tab:hp-sparsesgd}
\begin{tabular}{ll}
\toprule
Name (symbol) & Value \\
\midrule
Density parameter ($\rho$) & $0.001$ ($0.1\%$) \\
Support $S$ / mask $m$ & top-$k$ from $|g_0|$ (static) \\
Learning rate ($\eta$) & $1\times 10^{-4}$ \\
Epochs ($E_{\mathrm{train}}$) & $30$ \\
Batch size ($b_{\mathrm{train}}$) & $64$ \\
\bottomrule
\end{tabular}
\end{table}

\begin{table}[!h]
\centering
\caption{Hyperparameters for SparseAdam.}
\label{tab:hp-sparseadam}
\begin{tabular}{ll}
\toprule
Name (symbol) & Value \\
\midrule
Density parameter ($\rho$) & $0.001$ ($0.1\%$) \\
Support $S$ / mask $m$ & top-$k$ from $|g_0|$ (static) \\
First moment ($\beta_1$) & $0.9$ \\
Second moment ($\beta_2$) & $0.999$ \\
Epsilon ($\varepsilon$) & $1\times 10^{-8}$ \\
Learning rate ($\eta$) & $1\times 10^{-4}$ \\
Epochs ($E_{\mathrm{train}}$) & $30$ \\
Batch size ($b_{\mathrm{train}}$) & $64$ \\
\bottomrule
\end{tabular}
\end{table}

\begin{table}[!h]
\centering
\caption{Hyperparameters for SignSparseAdam.}
\label{tab:hp-signsparseadam}
\begin{tabular}{ll}
\toprule
Name (symbol) & Value \\
\midrule
Density parameter ($\rho$) & $0.001$ ($0.1\%$) \\
Support $S$ / mask $m$ & top-$k$ from $|g_0|$ (static) \\
First moment ($\beta_1$) & $0.9$ \\
Learning rate ($\eta$) & $1\times 10^{-4}$ \\
Epochs ($E_{\mathrm{train}}$) & $30$ \\
Batch size ($b_{\mathrm{train}}$) & $64$ \\
\bottomrule
\end{tabular}
\end{table}

\begin{table}[!h]
\centering
\caption{Hyperparameters for DynSparseSGD.}
\label{tab:hp-dynsparsesgd}
\begin{tabular}{ll}
\toprule
Name (symbol) & Value \\
\midrule
Density parameter ($\rho$) & $0.001$ ($0.1\%$) \\
Support $S_t$ / mask $m_t$ & top-$k$ from $|g_t|$ (dynamic) \\
Learning rate ($\eta$) & $1\times 10^{-4}$ \\
Epochs ($E_{\mathrm{train}}$) & $30$ \\
Batch size ($b_{\mathrm{train}}$) & $64$ \\
\bottomrule
\end{tabular}
\end{table}

\begin{table}[!h]
\centering
\caption{Hyperparameters for DynSparseAdam.}
\label{tab:hp-dynsparseadam}
\begin{tabular}{ll}
\toprule
Name (symbol) & Value \\
\midrule
Density parameter ($\rho$) & $0.001$ ($0.1\%$) \\
Support $S_t$ / mask $m_t$ & top-$k$ from $|g_t|$ (dynamic) \\
First moment ($\beta_1$) & $0.9$ \\
Second moment ($\beta_2$) & $0.999$ \\
Epsilon ($\varepsilon$) & $1\times 10^{-8}$ \\
Learning rate ($\eta$) & $1\times 10^{-4}$ \\
Epochs ($E_{\mathrm{train}}$) & $30$ \\
Batch size ($b_{\mathrm{train}}$) & $64$ \\
\bottomrule
\end{tabular}
\end{table}

\noindent \textbf{SignSparseSGD.}
SignSparseSGD uses a fixed sparse support $S$ selected once from $g_0$, builds the corresponding static mask $m$, and applies the unit-valued sign operator of Eq.~\eqref{eq:sparse_sign_update} to the masked gradient $\widetilde g_t=m\odot g_t$. The separate scale $\alpha=0.1$ gives the effective coordinate step $\lambda=\alpha\eta$ in Eq.~\eqref{eq:sparse_sign_update}.

\begin{table}[!h]
\centering
\caption{Hyperparameters for SignSparseSGD.}
\label{tab:hp-signsparsesgd}
\begin{tabular}{ll}
\toprule
Name (symbol) & Value \\
\midrule
Density parameter ($\rho$) & $0.001$ ($0.1\%$) \\
Support $S$ / mask $m$ & top-$k$ from $|g_0|$ (static) \\
Learning rate ($\eta$) & $1\times 10^{-4}$ \\
Epochs ($E_{\mathrm{train}}$) & $30$ \\
Batch size ($b_{\mathrm{train}}$) & $64$ \\
\bottomrule
\end{tabular}
\end{table}

\noindent \textbf{DynSignSparseSGD.}
DynSignSparseSGD recomputes the active support $S_t$ at every iteration, builds the dynamic mask $m_t$, and applies the same sign-sparse update rule as SignSparseSGD. This preserves the state-free sign update while allowing the active coordinates to change over time.

\begin{table}[!h]
\centering
\caption{Hyperparameters for DynSignSparseSGD.}
\label{tab:hp-dynsignsparsesgd}
\begin{tabular}{ll}
\toprule
Name (symbol) & Value \\
\midrule
Density parameter ($\rho$) & $0.001$ ($0.1\%$) \\
Support $S_t$ / mask $m_t$ & top-$k$ from $|g_t|$ (dynamic) \\
Learning rate ($\eta$) & $1\times 10^{-4}$ \\
Epochs ($E_{\mathrm{train}}$) & $30$ \\
Batch size ($b_{\mathrm{train}}$) & $64$ \\
\bottomrule
\end{tabular}
\end{table}

\noindent \textbf{Stabilization techniques.}
These variants keep the same base update as SignSGD or SparseSGD, but modify the update magnitude through norm matching or clipping. For the two GradNorm variants, the rescaling factor $a_t$ is computed online, so it is part of the update rule rather than a tuned hyperparameter.

\noindent \textbf{GradNorm SignSGD.}
GradNorm SignSGD rescales the sign direction by the norm-matching factor $a_t$ defined in Algorithm~\ref{alg:gradnormsignsgd}. It introduces no additional tuned scalar beyond the shared training schedule.

\begin{table}[!h]
\centering
\caption{Hyperparameters for GradNorm SignSGD.}
\label{tab:hp-gradnormsignsgd}
\begin{tabular}{ll}
\toprule
Name (symbol) & Value \\
\midrule
Learning rate ($\eta$) & $1\times 10^{-4}$ \\
Epochs ($E_{\mathrm{train}}$) & $30$ \\
Batch size ($b_{\mathrm{train}}$) & $64$ \\
\bottomrule
\end{tabular}
\end{table}

\noindent \textbf{GradNorm SparseSGD.}
GradNorm SparseSGD uses the same static support as SparseSGD and rescales the sparse update by the norm-matching factor $a_t$ defined in Algorithm~\ref{alg:gradnormsparsesgd}.

\begin{table}[!h]
\centering
\caption{Hyperparameters for GradNorm SparseSGD.}
\label{tab:hp-gradnormsparsesgd}
\begin{tabular}{ll}
\toprule
Name (symbol) & Value \\
\midrule
Density parameter ($\rho$) & $0.001$ ($0.1\%$) \\
Support $S$ / mask $m$ & top-$k$ from $|g_0|$ (static) \\
Learning rate ($\eta$) & $1\times 10^{-4}$ \\
Epochs ($E_{\mathrm{train}}$) & $30$ \\
Batch size ($b_{\mathrm{train}}$) & $64$ \\
\bottomrule
\end{tabular}
\end{table}

\noindent \textbf{Local-clip SparseSGD.}
Local-clip SparseSGD keeps the same static support as SparseSGD and clips each active coordinate independently to the interval $[-c_{\mathrm{loc}},c_{\mathrm{loc}}]$. Our configuration uses $c_{\mathrm{loc}}=0.1$.

\begin{table}[!h]
\centering
\caption{Hyperparameters for Local-clip SparseSGD.}
\label{tab:hp-localclipsparsesgd}
\begin{tabular}{ll}
\toprule
Name (symbol) & Value \\
\midrule
Density parameter ($\rho$) & $0.001$ ($0.1\%$) \\
Support $S$ / mask $m$ & top-$k$ from $|g_0|$ (static) \\
Clipping threshold ($c_{\mathrm{loc}}$) & $0.1$ \\
Learning rate ($\eta$) & $1\times 10^{-4}$ \\
Epochs ($E_{\mathrm{train}}$) & $30$ \\
Batch size ($b_{\mathrm{train}}$) & $64$ \\
\bottomrule
\end{tabular}
\end{table}

\noindent \textbf{Global-clip SparseSGD.}
Global-clip SparseSGD applies global $\ell_2$ clipping with threshold $c_{\mathrm{glob}}$ before sparsification. Our configuration uses $c_{\mathrm{glob}}=0.1$.

\begin{table}[!h]
\centering
\caption{Hyperparameters for Global-clip SparseSGD.}
\label{tab:hp-globalclipsparsesgd}
\begin{tabular}{ll}
\toprule
Name (symbol) & Value \\
\midrule
Density parameter ($\rho$) & $0.001$ ($0.1\%$) \\
Support $S$ / mask $m$ & top-$k$ from $|g_0|$ (static) \\
Norm threshold ($c_{\mathrm{glob}}$) & $0.1$ \\
Learning rate ($\eta$) & $1\times 10^{-4}$ \\
Epochs ($E_{\mathrm{train}}$) & $30$ \\
Batch size ($b_{\mathrm{train}}$) & $64$ \\
\bottomrule
\end{tabular}
\end{table}

\section{Detailed Training-Time and Memory Analysis}
\label{app:efficiency}
\label{sec:memory_time}
\suppanchor{efficiency}

The proposed optimizers still execute the dense forward and backward passes used to obtain gradients. Consequently, they do not reduce gradient-computation cost, activation memory, or the transient buffers required by backpropagation. Their savings arise in the optimization stage after backpropagation: no first- or second-moment buffers are maintained, and the update is represented by the active coordinates and signs. The accounting below separates wall-clock training time from idealized optimizer-related storage. It is not a measurement of peak end-to-end GPU memory.

\begin{table}
  \centering
  \small
  \caption{Training time comparison on SUN397 for 16-shot classification with CLIP (ViT-B/16).}
  \label{tab:time_sun397}
  \begin{tabular}{l c}
    \toprule
    \textbf{Method} & \textbf{Training time} \\
    \midrule
    BitFit          & 18 min \\
    LayerNorm       & 17 min \\
    SAFT            & 21 min \\
    %Adapter         & 9 min \\
    TIP-Adapter-F     & 4 min \\
    %LP++            & 2 min \\
    CoOp            & 1 h 18 min \\
    KgCoOp          & 1 h 18 min \\
    ProGrad         & 2 h 08 min \\
    LoRA            & 19 min \\
    Adam            & 21 min \\
    GaLoRE          & 38 min \\
    SO              & 26 min \\
    \midrule
    SignSparseSGD   & 20 min \\
    DynSignSparseSGD& 21 min \\
    \bottomrule
  \end{tabular}
  
\end{table}

\textbf{Computational cost.} Table~\ref{tab:time_sun397} reports the measured training times available for SUN397 with CLIP (ViT-B/16) in the 16-shot setting. TIP-Adapter-F has the shortest reported training time, about 4 minutes, for optimizing cache-key weights with frozen CLIP encoders. Its timer excludes cache construction. %GDA and ProKeR are also training-free closed-form methods, but comparable SUN397 timing records were not available for this table; we therefore do not estimate their runtimes.

Among methods that still differentiate through the backbone, LayerNorm, LoRA, SAFT, dense SGD, Adam, SignSparseSGD, and DynSignSparseSGD all lie in a narrow range between 17 and 21 minutes. SO is slightly slower at 26 minutes because it maintains sparse first- and second-moment buffers and refreshes the active support during training. Even so, it remains much faster than GaLoRE and the prompt-learning baselines. These results indicate that, once the full CLIP backbone participates in the forward and backward pass, the dominant cost comes from the backbone computation rather than from the number of trainable parameters.

Prompt-tuning methods are much slower in wall-clock time. CoOp and KgCoOp each require about 1 h 18 min, and ProGrad requires about 2 h 08 min, mainly because these methods rely on longer training schedules. GaLoRE also introduces a clear overhead: its low-rank gradient projections increase runtime to about 38 minutes. In contrast, SignSparseSGD and DynSignSparseSGD match dense SGD almost exactly. Therefore, their sparse support selection and sign-only updates add negligible overhead in practice.

\textbf{Theoretical memory cost.} To complement the wall-clock comparison, we analyze the idealized optimization memory. We count only weights, gradients, and persistent optimizer states. Activations and backpropagation buffers are omitted because they are shared by all methods that backpropagate through the CLIP backbone.

We start with a single fully-connected layer with weight matrix $W \in \mathbb{R}^{d_{\mathrm{out}}\times d_{\mathrm{in}}}$, where $d_{\mathrm{in}}$ is the input dimension, $d_{\mathrm{out}}\leq d_{\mathrm{in}}$ is the output dimension, $r_{\mathrm{low}}$ is the rank used by low-rank methods, and $\rho \in (0,1]$ is the density of the sparse updates. Table~\ref{tab:conceptual_memory_signsparse} compares Adam, SparseSGD, SparseAdam, SignSparseSGD, SO, LoRA, and GaLoRE. SparseSGD and SignSparseSGD store only the active gradient values on the $\rho d_{\mathrm{out}}d_{\mathrm{in}}$ selected coordinates. SparseAdam adds two sparse moment buffers on the same support. SO is also sparse, but its dynamic support requires a value-index representation for the gradient and shared indices for the sparse moments. This yields $2d_{\mathrm{out}}d_{\mathrm{in}}\rho$ scalars for the sparse gradient and $3d_{\mathrm{out}}d_{\mathrm{in}}\rho$ scalars for persistent optimizer states. LoRA adds two rank-$r_{\mathrm{low}}$ factors, while GaLoRE keeps dense gradients and stores low-rank projection matrices.

\begin{table}[t]
\centering
\caption{Idealized memory requirements for a single fully-connected layer. The weight matrix is $W \in \mathbb{R}^{d_{\mathrm{out}}\times d_{\mathrm{in}}}$, with input dimension $d_{\mathrm{in}}$ and output dimension $d_{\mathrm{out}}\leq d_{\mathrm{in}}$. The rank and density ratio are $r_{\mathrm{low}}$ and $\rho$, respectively. We count the number of stored scalars for the weight parameters, the gradients of the trainable parameters, and the persistent optimizer states. Activations and backpropagation buffers are omitted because they are shared across methods.}
\resizebox{\textwidth}{!}{%
\begin{tabular}{lccccccccc}
\toprule
                   & Adam          & SparseSGD        & SparseAdam          & SignSparseSGD       & SO          & LoRA             & GaLoRE \\
\midrule
Weight             & $d_{\mathrm{out}}d_{\mathrm{in}}$ & $d_{\mathrm{out}}d_{\mathrm{in}}$ & $d_{\mathrm{out}}d_{\mathrm{in}}$ & $d_{\mathrm{out}}d_{\mathrm{in}}$ & $d_{\mathrm{out}}d_{\mathrm{in}}$ & $d_{\mathrm{out}}d_{\mathrm{in}} + d_{\mathrm{out}}r_{\mathrm{low}} + d_{\mathrm{in}}r_{\mathrm{low}}$ & $d_{\mathrm{out}}d_{\mathrm{in}}$ \\
Gradient           & $d_{\mathrm{out}}d_{\mathrm{in}}$ & $\rho d_{\mathrm{out}}d_{\mathrm{in}}$ & $\rho d_{\mathrm{out}}d_{\mathrm{in}}$ & $\rho d_{\mathrm{out}}d_{\mathrm{in}}$ & $2\rho d_{\mathrm{out}}d_{\mathrm{in}}$ & $d_{\mathrm{out}}r_{\mathrm{low}} + d_{\mathrm{in}}r_{\mathrm{low}}$ & $d_{\mathrm{out}}d_{\mathrm{in}}$ \\
Optimizer states   & $2d_{\mathrm{out}}d_{\mathrm{in}}$ & $0$ & $2\rho d_{\mathrm{out}}d_{\mathrm{in}}$ & $0$ & $3\rho d_{\mathrm{out}}d_{\mathrm{in}}$ & $2d_{\mathrm{out}}r_{\mathrm{low}} + 2d_{\mathrm{in}}r_{\mathrm{low}}$ & $d_{\mathrm{out}}r_{\mathrm{low}} + 2d_{\mathrm{in}}r_{\mathrm{low}}$ \\
\bottomrule
\end{tabular}}
\label{tab:conceptual_memory_signsparse}
\end{table}

From Table~\ref{tab:conceptual_memory_signsparse}, SGD stores the weight matrix and its dense gradient but no additional states. Adam keeps two full moment buffers of size $d_{\mathrm{out}}d_{\mathrm{in}}$ on top of the gradient, which yields a $3\times$ increase in optimizer-related memory compared with SGD. In contrast, the sparse variants reduce the gradient footprint to $d_{\mathrm{out}}d_{\mathrm{in}}\rho$ and confine their optimizer states to the active coordinates. In particular, SparseSGD and SignSparseSGD have the same optimization memory cost, since both avoid persistent moment buffers and maintain only the sparse gradient on the $\rho d_{\mathrm{out}}d_{\mathrm{in}}$ active entries (plus a negligible binary mask). So their optimizer-specific memory is essentially identical and strictly smaller than that of SparseAdam, which still carries $2d_{\mathrm{out}}d_{\mathrm{in}}\rho$ moment scalars. Low-rank approaches trade dense optimizer states for additional low-rank parameters.

\begin{table}[t]
\centering
\caption{Theoretical memory consumption for CLIP (ViT-B/16) when adapting all vision and text transformer blocks. We report the number of scalars (in millions) and the corresponding memory in MB, assuming 32-bit floats. Forward/backward activations, biases, and embedding layers are omitted since they are shared across all methods.}
\scalebox{0.65}{\begin{tabular}{lccccc}
\toprule
Method & Weight (\#Vars, MB) & Gradient (\#Vars, MB) & Opt.\ States (\#Vars, MB) & \#Trainable & Total Mem.\ (MB) \\
\midrule
Adam (Full FT) & $122.7$M, $468.1$ & $122.7$M, $468.1$ & $245.4$M, $936.2$ & $122.7$M & $1872.5$ \\
\midrule
LoRA ($r_{\mathrm{LoRA}}=2$)  & $123.2$M, $470.0$ & $0.5$M, $1.9$ & $1.0$M, $3.8$  & $0.5$M & $475.6$ \\
LoRA ($r_{\mathrm{LoRA}}=4$)  & $123.7$M, $471.9$ & $1.0$M, $3.8$ & $2.0$M, $7.5$  & $1.0$M & $483.1$ \\
LoRA ($r_{\mathrm{LoRA}}=8$)  & $124.7$M, $475.6$ & $2.0$M, $7.5$ & $3.9$M, $15.0$ & $2.0$M & $498.1$ \\
LoRA ($r_{\mathrm{LoRA}}=16$) & $126.6$M, $483.1$ & $3.9$M, $15.0$ & $7.9$M, $30.0$ & $3.9$M & $528.1$ \\
\midrule
GaLoRE ($r_{\mathrm{GaLoRE}}=2$)  & $122.7$M, $468.1$ & $122.7$M, $468.1$ & $0.7$M, $2.7$  & $122.7$M & $938.9$ \\
GaLoRE ($r_{\mathrm{GaLoRE}}=4$)  & $122.7$M, $468.1$ & $122.7$M, $468.1$ & $1.4$M, $5.4$  & $122.7$M & $941.6$ \\
GaLoRE ($r_{\mathrm{GaLoRE}}=8$)  & $122.7$M, $468.1$ & $122.7$M, $468.1$ & $2.8$M, $10.8$ & $122.7$M & $947.0$ \\
GaLoRE ($r_{\mathrm{GaLoRE}}=16$) & $122.7$M, $468.1$ & $122.7$M, $468.1$ & $5.7$M, $21.6$ & $122.7$M & $957.8$ \\
\midrule
SO ($\rho=0.01\%$) & $122.7$M, $468.1$ & $0.0$M, $0.1$ & $0.0$M, $0.1$ & $0.0$M & $468.4$ \\
SO ($\rho=0.1\%$)  & $122.7$M, $468.1$ & $0.2$M, $0.9$ & $0.4$M, $1.4$ & $0.1$M & $470.5$ \\
SO ($\rho=1\%$)    & $122.7$M, $468.1$ & $2.5$M, $9.4$ & $3.7$M, $14.0$ & $1.2$M & $491.5$ \\
SO ($\rho=10\%$)   & $122.7$M, $468.1$ & $24.5$M, $93.6$ & $36.8$M, $140.4$ & $12.3$M & $702.2$ \\
\midrule
SignSparseSGD ($\rho=0.01\%$) & $122.7$M, $468.1$ & $0.0$M, $0.0$ & $0.0$M, $0.0$ & $0.0$M & $468.2$ \\
SignSparseSGD ($\rho=0.1\%$)  & $122.7$M, $468.1$ & $0.1$M, $0.5$ & $0.0$M, $0.0$ & $0.1$M & $468.6$ \\
SignSparseSGD ($\rho=1\%$)    & $122.7$M, $468.1$ & $1.2$M, $4.7$ & $0.0$M, $0.0$ & $1.2$M & $472.8$ \\
SignSparseSGD ($\rho=10\%$)   & $122.7$M, $468.1$ & $12.3$M, $46.8$ & $0.0$M, $0.0$ & $12.3$M & $514.9$ \\
\bottomrule
\end{tabular}}
\label{tab:memory_clip}
\end{table}

Table~\ref{tab:memory_clip} extends the single-layer analysis to CLIP (ViT-B/16) when all 12 vision and 12 text transformer blocks are adapted. We count 32-bit weights, gradients, and optimizer states, and omit activations, biases, and embedding layers. The dense transformer weights contain $122.7$M parameters. SignSparseSGD, and likewise DynSignSparseSGD, keeps this dense backbone but adds only $k_{\mathrm{step}}$ active gradient entries and no persistent optimizer states. The table uses the idealized approximation $k_{\mathrm{step}}\approx\rho d$, which neglects the floor and minimum-one effects in Eq.~\eqref{eq:block_budgets}. SO also scales linearly with $\rho$, but its sparse moments increase the overhead to $5\rho d$ additional scalars in total. For example, at $\rho=1\%$, SO requires $491.5$ MB, compared with $472.8$ MB for SignSparseSGD and $1872.5$ MB for Adam. LoRA remains very compact at low rank because it optimizes only the added low-rank factors, whereas GaLoRE keeps the dense gradient and therefore remains close to $1$ GB even at small rank. Overall, among methods that preserve the original dense backbone parameterization, the sign-sparse optimizers have the smallest optimization-memory overhead, and DynSignSparseSGD has the same theoretical footprint as SignSparseSGD.

\section{Geometry and Convergence Analysis}
\label{app:convergence}
\suppanchor{geometry}

This appendix states every theorem, proposition, lemma, and corollary immediately before its proof. It proves the geometric and convergence results, together with auxiliary results and counterexamples. The notation $\Theta_t$, $D$, $F$, $f$, $e_D$, $\xi_t$, $g_t$, $s_t$, $S$, $S_t$, $m_S$, $\supp$, $\Top_k$, $\sgn$, $\eta$, $\alpha$, $\lambda$, $\lambda_t$, $\Lambda_T$, $\{\mathcal I_r\}_{r=1}^{R}$, $\mathbf{k}$, $k_{\mathrm{step}}$, $L$, $F_\star$, $\bar L_{\mathbf{k}}$, and $\underline\rho$ is defined in Sec.~\ref{sec:method} and this appendix, and summarized in the Summary of Notation at the beginning of this supplement. Each result below states its assumptions and conclusion before the proof. All sparse sets and norms use the per-parameter-tensor budgets in Eq.~\eqref{eq:block_budgets}.

\subsection{Sparse sign as a blockwise steepest-descent oracle}
\suppanchor{geometry-oracle}

\begin{proposition}[Sparse sign is an exact steepest-descent oracle]
\label{prop:sparse_steepest_descent}
For an ambient dimension $d\ge1$, a vector $v\in\R^d$, and a budget $k\in\{1,\ldots,d\}$, let $|v|_{(1)}\ge\cdots\ge|v|_{(d)}$ be its ordered magnitudes and let $\|v\|_{[k]}=\sum_{j=1}^{k}|v|_{(j)}$. The feasible direction set and sparse sign map are
\begin{equation*}
\mathcal C_{d,k}=\{u\in\R^d:\ \|u\|_\infty\le1,\ \|u\|_1\le k\},
\qquad
Q_k(v)=m_{\Top_k(|v|)}\odot\sgn(v).
\end{equation*}
Then
\begin{equation*}
Q_k(v)\in\arg\max_{u\in\mathcal C_{d,k}}\langle v,u\rangle,
\qquad
\max_{u\in\mathcal C_{d,k}}\langle v,u\rangle=\|v\|_{[k]}.
\end{equation*}
Now let $v\in\R^d$. For the block partition and budgets in Eq.~\eqref{eq:block_budgets}, let $\mathbf{k}=(k_1,\ldots,k_R)$. The blockwise quantities are
\begin{align*}
\mathcal C_{\mathbf{k}}
&=\{u\in\R^d:\ \|u_{\mathcal I_r}\|_\infty\le1,\ \|u_{\mathcal I_r}\|_1\le k_r\ \text{for every }r\},\\
(Q_{\mathbf{k}}(v))_{\mathcal I_r}
&=Q_{k_r}(v_{\mathcal I_r}),
\qquad
\|v\|_{[\mathbf{k}]}=\sum_{r=1}^{R}\|v_{\mathcal I_r}\|_{[k_r]}.
\end{align*}
Then
\begin{equation*}
Q_{\mathbf{k}}(v)\in\arg\max_{u\in\mathcal C_{\mathbf{k}}}\langle v,u\rangle,
\qquad
\max_{u\in\mathcal C_{\mathbf{k}}}\langle v,u\rangle=\|v\|_{[\mathbf{k}]}.
\end{equation*}
For a fixed support $S$,
\begin{equation*}
m_S\odot\sgn(v)
\in\arg\max_{\substack{\supp(u)\subseteq S\\\|u\|_\infty\le1}}\langle v,u\rangle,
\qquad
\max_{\substack{\supp(u)\subseteq S\\\|u\|_\infty\le1}}\langle v,u\rangle=\|v_S\|_1.
\end{equation*}
\end{proposition}

\begin{proof}
Let $d\ge1$, $v\in\R^d$, and $k\in\{1,\ldots,d\}$ be fixed. For any $u\in\mathcal C_{d,k}$,
\begin{equation*}
\langle v,u\rangle\le\sum_{i=1}^{d}|v_i||u_i|.
\end{equation*}
The numbers $a_i=|u_i|$ satisfy $0\le a_i\le1$ and $\sum_i a_i\le k$. The corresponding linear program assigns unit mass to coordinates with the $k$ largest magnitudes. Its value is $\|v\|_{[k]}$, and $Q_k(v)$ attains it. Ties only make the optimizer nonunique.

For the blockwise statement, let $v\in\R^d$ be fixed. The constraints and objective separate over the disjoint blocks:
\begin{equation*}
\max_{u\in\mathcal C_{\mathbf{k}}}\langle v,u\rangle
=\sum_{r=1}^{R}\max_{u_{\mathcal I_r}\in\mathcal C_{d_r,k_r}}
\langle v_{\mathcal I_r},u_{\mathcal I_r}\rangle
=\sum_{r=1}^{R}\|v_{\mathcal I_r}\|_{[k_r]}.
\end{equation*}
The concatenated vector $Q_{\mathbf{k}}(v)$ attains every blockwise maximum. For fixed $S$, coordinates outside $S$ are zero, and the remaining terms are maximized independently by $u_i=\sgn(v_i)$.
\end{proof}

\paragraph{Population setup and assumptions.}
For the theoretical analysis, the support examples $z_{c,j_{\mathrm{shot}}}\sim\mathcal P_c$ are independent across classes $c\in\{1,\ldots,N\}$ and shots $j_{\mathrm{shot}}\in\{1,\ldots,K\}$. The balanced population risk is
\begin{equation}
F(\Theta)=\frac1N\sum_{c=1}^{N}\E_{z\sim\mathcal P_c}[\ell(\Theta;z)].
\label{eq:population_risk}
\end{equation}
The empirical risk $f$ is the support average defined in Appendix~\ref{app:clip_adaptation}. Both risks are differentiable wherever their gradients are used. The finite-shot bias, residual stochastic error, and population gradient are
\begin{equation}
\begin{aligned}
e_D(\Theta)&=\nabla f(\Theta)-\nabla F(\Theta),\\
\xi_t&=g_t-\nabla f(\Theta_t),\qquad h_t=\nabla F(\Theta_t),\\
g_t&=h_t+e_D(\Theta_t)+\xi_t.
\end{aligned}
\label{eq:population_gradient_decomposition}
\end{equation}
Conditional on $D$, $e_D$ is a fixed function. The residual $\xi_t$ includes minibatch and augmentation randomness and need not have zero conditional mean. Thus the analysis does not assume an unbiased population-gradient oracle.

The initial pretrained vector $\Theta_0$ and positive steps $\{\lambda_t\}_{t\ge0}$ are deterministic relative to support sampling and optimizer randomness. The constant-step algorithm uses $\lambda_t=\lambda=\alpha\eta$. For a finite horizon $T\ge1$, the direction alphabet and cumulative step are
\begin{equation}
\begin{aligned}
\mathcal U_{\mathbf{k}}
&=\{s\in\{-1,0,1\}^d:\ \|s_{\mathcal I_r}\|_0\le k_r\ \text{for every }r\},\\
\Lambda_T&=\sum_{t=0}^{T-1}\lambda_t.
\end{aligned}
\label{eq:population_direction_schedule}
\end{equation}
The corresponding reachable set and deterministic adaptation envelope are
\begin{equation}
\begin{aligned}
\mathcal R_T
&=\bigcup_{t=0}^{T}\left\{\Theta_0-\sum_{t'=0}^{t-1}\lambda_{t'}s_{t'}:\ s_{t'}\in\mathcal U_{\mathbf{k}}\right\},\\
\mathcal A_T
&=\left\{\Theta:\ \|\Theta-\Theta_0\|_\infty\le\Lambda_T,\right.\\
&\hspace{39mm}\left.\| (\Theta-\Theta_0)_{\mathcal I_r}\|_1\le k_r\Lambda_T\ \text{for every }r\right\}.
\end{aligned}
\label{eq:population_reachable_envelope}
\end{equation}
The empty sum at $t=0$ is zero. These definitions use the block budgets and the blockwise norm from Proposition~\ref{prop:sparse_steepest_descent}.

\begin{assumption}[Population smoothness and additive minibatch control]
\label{assumption_1}
For the horizon under consideration, the following conditions hold.
\begin{enumerate}
\item A finite vector $L\in\R_+^d$ and a finite lower bound $F_\star\in\R$ satisfy $F(\Theta)\ge F_\star$ for every $\Theta\in\mathcal A_T$ and
\begin{equation}
F(\Theta')\le F(\Theta)
+\langle\nabla F(\Theta),\Theta'-\Theta\rangle
+\frac12\sum_{i=1}^{d}L_i(\Theta'_i-\Theta_i)^2
\label{eq:ass_population_coord_smooth}
\end{equation}
for every $\Theta,\Theta'\in\mathcal A_T$. This is a simultaneous-update quadratic upper bound, not only a collection of one-coordinate inequalities.
\item For the deterministic minibatch sizes $b_t\ge1$, a finite vector $\nu\in\R_+^d$ and $\gamma\in(0,1]$ satisfy
\begin{equation}
\E[|\xi_{t,i}|\mid D,\Theta_t]\le\frac{\nu_i}{b_t^\gamma},
\qquad i\in\{1,\ldots,d\},\quad t<T.
\label{eq:ass_minibatch_moment}
\end{equation}
\end{enumerate}
Conditional expectations in the statements below are assumed finite. Part (2) is used only where a batch-size error bound is invoked.
\end{assumption}

Writing $L_{r,(j)}$ for the $j$th largest entry of $L$ in block $r$, the selected-curvature quantities and minimum realized block density are
\begin{equation}
\begin{aligned}
L_t^{\mathrm{sel}}&=\sum_{i\in S_t}L_i,
&\bar L_{\mathbf{k}}&=\sum_{r=1}^{R}\sum_{j=1}^{k_r}L_{r,(j)},\\
\|L_S\|_1&=\sum_{i\in S}L_i,
&\underline\rho&=\min_{1\le r\le R}\frac{k_r}{d_r}.
\end{aligned}
\label{eq:population_curvature_budgets}
\end{equation}
In particular, $L_t^{\mathrm{sel}}\le\bar L_{\mathbf{k}}$. For any support $S$ with $|S\cap\mathcal I_r|=k_r$ in each block, its missed population-gradient mass is
\begin{equation}
\mathcal E_{\mathrm{miss},t}(S)
=\|h_t\|_{[\mathbf{k}]}-\|h_{t,S}\|_1\ge0.
\label{eq:static_support_missed_mass}
\end{equation}
The conditional population alignments are
\begin{equation}
\begin{aligned}
\mathfrak a_t^{\mathrm{dyn}}
&=\E[\langle h_t,Q_{\mathbf{k}}(g_t)\rangle\mid D,\Theta_t],\\
\mathfrak a_t^{\mathrm{stat}}
&=\E[\langle h_t,m_S\odot\sgn(g_t)\rangle\mid D,S,\Theta_t].
\end{aligned}
\label{eq:dynamic_population_alignment}
\end{equation}

\subsection{Sparsity--learning-rate coupling}
\label{subsec:appE_sparsity_learning_rate}
\suppanchor{sparsity-lr}

The steepest-descent interpretation also determines the norm of one sign-sparse step. For the constant-step specialization, the deterministic steps satisfy $\lambda_t=\lambda=\alpha\eta$. For either optimizer and every $p\in[1,\infty)$,
\begin{align}
\|\Theta_{t+1}-\Theta_t\|_p
&=\lambda\|s_t\|_0^{1/p}
\le\lambda k_{\mathrm{step}}^{1/p},
\label{eq:appE_step_lp_norm}\\
\|\Theta_{t+1}-\Theta_t\|_\infty
&=\lambda\1\{\|s_t\|_0>0\}
\le\lambda.
\label{eq:appE_step_linf_norm}
\end{align}
The equalities use the ternary convention in Eq.~\eqref{eq:sparse_sign_update}. In particular, a selected zero gradient contributes neither to $\|s_t\|_0$ nor to the update norm. If all $k_{\mathrm{step}}$ selected entries are nonzero, then
\begin{equation}
\|\Theta_{t+1}-\Theta_t\|_2=\lambda\sqrt{k_{\mathrm{step}}},
\qquad
\|\Theta_{t+1}-\Theta_t\|_1=\lambda k_{\mathrm{step}}.
\label{eq:appE_exact_step_norms}
\end{equation}
Thus, holding the coordinate step fixed while changing sparsity is not norm-neutral.

Theorem~\ref{thm:population_additive_bounds} is stated for arbitrary positive steps. For a constant step, its explicit objective-gap and curvature terms are bounded by
\begin{align}
&\frac{F(\Theta_0)-F_\star}{T\lambda}
+\frac{\lambda}{2}\|L_S\|_1
&&\text{for SignSparseSGD},
\label{eq:appE_static_gap_curvature_envelope}\\
&\frac{F(\Theta_0)-F_\star}{T\lambda}
+\frac{\lambda}{2}\bar L_{\mathbf{k}}
&&\text{for DynSignSparseSGD}.
\label{eq:appE_dynamic_gap_curvature_envelope}
\end{align}
The second line uses $L_t^{\mathrm{sel}}\le\bar L_{\mathbf{k}}$. The support-miss and stochastic-error terms are omitted only to isolate the explicit gap--curvature trade-off.

\begin{proposition}[Sparsity--learning-rate coupling of the explicit descent envelope]
\label{prop:sparsity_learning_rate_coupling}
The conditions are $T\ge1$, $F(\Theta_0)>F_\star$, $\|L_S\|_1>0$, and $\bar L_{\mathbf{k}}>0$. The unique positive minimizers of Eqs.~\eqref{eq:appE_static_gap_curvature_envelope} and~\eqref{eq:appE_dynamic_gap_curvature_envelope} are
\begin{equation}
\lambda_{S,T}^{\star}
=\sqrt{\frac{2\bigl(F(\Theta_0)-F_\star\bigr)}{T\|L_S\|_1}},
\qquad
\lambda_{\mathbf{k},T}^{\star}
=\sqrt{\frac{2\bigl(F(\Theta_0)-F_\star\bigr)}{T\bar L_{\mathbf{k}}}}.
\label{eq:appE_envelope_minimizing_steps}
\end{equation}
The comparison concerns a family of blockwise budgets for which there are constants $0<L_{\mathrm{low}}\le L_{\mathrm{high}}<\infty$, independent of the budget, such that
\begin{equation}
L_{\mathrm{low}}
\le\frac{\bar L_{\mathbf{k}}}{k_{\mathrm{step}}}
\le L_{\mathrm{high}}.
\label{eq:appE_comparable_average_curvature}
\end{equation}
Then
\begin{equation}
\sqrt{\frac{2\bigl(F(\Theta_0)-F_\star\bigr)}{T L_{\mathrm{high}}}}
\le
\lambda_{\mathbf{k},T}^{\star}\sqrt{k_{\mathrm{step}}}
\le
\sqrt{\frac{2\bigl(F(\Theta_0)-F_\star\bigr)}{T L_{\mathrm{low}}}}.
\label{eq:appE_l2_scale_bounds}
\end{equation}
Hence, $\lambda_{\mathbf{k},T}^{\star}\asymp k_{\mathrm{step}}^{-1/2}$ under a comparable average curvature budget. The same conclusion holds for the static method when $\|L_S\|_1/k_{\mathrm{step}}$ obeys Eq.~\eqref{eq:appE_comparable_average_curvature}.
\end{proposition}

\begin{proof}
For positive constants $A_{\mathrm{gap}}$ and $C_{\mathrm{curv}}$, the function $A_{\mathrm{gap}}/(T\lambda)+C_{\mathrm{curv}}\lambda/2$ on $\lambda>0$ has derivative $-A_{\mathrm{gap}}/(T\lambda^2)+C_{\mathrm{curv}}/2$ and strictly positive second derivative. Its unique minimizer is $\sqrt{2A_{\mathrm{gap}}/(TC_{\mathrm{curv}})}$. Substituting $A_{\mathrm{gap}}=F(\Theta_0)-F_\star$ and $C_{\mathrm{curv}}=\|L_S\|_1$ or $C_{\mathrm{curv}}=\bar L_{\mathbf{k}}$ gives Eq.~\eqref{eq:appE_envelope_minimizing_steps}. Equation~\eqref{eq:appE_l2_scale_bounds} follows by substituting Eq.~\eqref{eq:appE_comparable_average_curvature} into the dynamic minimizer and multiplying by $\sqrt{k_{\mathrm{step}}}$. The static statement is identical.
\end{proof}

For a reference configuration, let $\rho_{\mathrm{ref}}\in(0,1]$ be its nominal density, let $\alpha_{\mathrm{ref}}>0$ be its sign multiplier, let $\eta>0$ be the fixed base learning rate, let $\lambda_{\mathrm{ref}}=\alpha_{\mathrm{ref}}\eta$, and let $k_{\mathrm{step,ref}}$ denote its per-step budget. Three different comparison conventions are
\begin{align}
\text{fixed coordinate step:}\quad
&\lambda=\lambda_{\mathrm{ref}},
\label{eq:appE_fixed_coordinate_step}\\
\text{fixed $\ell_2$ update bound:}\quad
&\lambda=\lambda_{\mathrm{ref}}
\sqrt{\frac{k_{\mathrm{step,ref}}}{k_{\mathrm{step}}}},
\label{eq:appE_fixed_l2_step}\\
\text{fixed $\ell_1$ update bound:}\quad
&\lambda=\lambda_{\mathrm{ref}}
\frac{k_{\mathrm{step,ref}}}{k_{\mathrm{step}}}.
\label{eq:appE_fixed_l1_step}
\end{align}
The curvature-envelope scaling in Proposition~\ref{prop:sparsity_learning_rate_coupling} agrees with the fixed-$\ell_2$ convention when the average curvature budget is comparable. It does not imply the stronger fixed-$\ell_1$ scaling.

At the fixed base learning rate $\eta$, the same rescaling applies to $\alpha$ because $\lambda=\alpha\eta$. The exact blockwise fixed-$\ell_2$ rule is
\begin{equation}
\alpha
=\alpha_{\mathrm{ref}}
\sqrt{\frac{k_{\mathrm{step,ref}}}{k_{\mathrm{step}}}}.
\label{eq:appE_alpha_blockwise_rescaling}
\end{equation}
When the same parameter partition is used and the floor and minimum-one effects in Eq.~\eqref{eq:block_budgets} are negligible, $k_{\mathrm{step}}$ is approximately proportional to the nominal density. Equation~\eqref{eq:appE_alpha_blockwise_rescaling} then reduces to
\begin{equation}
\alpha
\approx\alpha_{\mathrm{ref}}
\sqrt{\frac{\rho_{\mathrm{ref}}}{\rho}}.
\label{eq:appE_alpha_density_rescaling}
\end{equation}
This is a relative rescaling law. It does not derive an absolute value for $\alpha$. Appendix~\ref{sec:update_scale} calibrates the reference pair $\rho_{\mathrm{ref}}=10^{-3}$ and $\alpha_{\mathrm{ref}}=0.1$ by matching the observed Euclidean update scale to dense SGD. Appendix~\ref{sec:sensitivity_analysis} intentionally holds $\eta$ and $\alpha$ fixed while varying density. That experiment is a fixed-coordinate-step sensitivity study, not an $\ell_2$-matched comparison.

Finally, Proposition~\ref{prop:sparsity_learning_rate_coupling} minimizes only the terms that depend explicitly on $\lambda$ in the displayed upper envelope. The iterates, support-miss terms, finite-shot bias, and minibatch errors can themselves change with $\lambda$. Therefore, the result is a principled scaling law and not an end-to-end claim that Eq.~\eqref{eq:appE_envelope_minimizing_steps} is the empirically optimal learning rate. Moreover, the inverse-square-root law need not hold when $\bar L_{\mathbf{k}}/k_{\mathrm{step}}$ changes substantially with sparsity.

\subsection{Deterministic blockwise reachability}
\suppanchor{reachability}

The sets $\mathcal R_T$ and $\mathcal A_T$ in Eq.~\eqref{eq:population_reachable_envelope} contain every admissible trajectory.

\begin{lemma}[Deterministic blockwise envelope]
\label{lem:appE:deterministic_envelope}
For either sign-sparse optimizer and every realization of its gradients,
\begin{equation}
\Theta_t\in\mathcal R_T\subseteq\mathcal A_T,
\qquad 0\le t\le T.
\end{equation}
\end{lemma}

\begin{proof}
Every update has the form $\Theta_{t+1}=\Theta_t-\lambda_ts_t$ with $s_t\in\mathcal U_{\mathbf{k}}$, so $\Theta_t\in\mathcal R_T$. For every coordinate,
\begin{equation}
|\Theta_{t,i}-\Theta_{0,i}|
\le\sum_{t'=0}^{t-1}\lambda_{t'}|s_{t',i}|
\le\Lambda_T.
\end{equation}
Within block $r$,
\begin{equation}
\|(\Theta_t-\Theta_0)_{\mathcal I_r}\|_1
\le\sum_{t'=0}^{t-1}\lambda_{t'}\|s_{t',\mathcal I_r}\|_1
\le k_r\Lambda_T.
\end{equation}
Thus $\Theta_t\in\mathcal A_T$ pathwise. Both sets are deterministic relative to the support sample.
\end{proof}

\subsection{Trajectory-wise finite-shot bias and a reference certificate}
\suppanchor{bias-certificate}

\begin{proposition}[No distribution-free exact guarantee at fixed shot]
\label{prop:fixed_shot_no_go}
For any finite $K$, no optimizer whose trajectory and output are measurable with respect to the support set and its internal randomness can guarantee population stationarity almost surely for every within-task distribution. This excludes both an almost-sure stationary output and an almost-sure vanishing weighted-average population-gradient norm. The impossibility already holds for $N=d=1$ and the squared loss.
\end{proposition}

\begin{proof}
The construction uses $N=d=1$ and $\ell(\theta;z)=(\theta-z)^2/2$. Let $\mu_{\mathrm{sep}}>0$ be fixed. Let $\mathcal P_+$ assign probability $1/2$ to each of $0$ and $2\mu_{\mathrm{sep}}$, and let $\mathcal P_-$ assign probability $1/2$ to each of $0$ and $-2\mu_{\mathrm{sep}}$. Their population gradients are
\begin{equation}
\nabla F_+(\theta)=\theta-\mu_{\mathrm{sep}},
\qquad
\nabla F_-(\theta)=\theta+\mu_{\mathrm{sep}}.
\end{equation}
For every finite $K$, the event that all $K$ support examples equal zero has probability $2^{-K}$ under both populations. On this event, the optimizer receives exactly the same support set under $\mathcal P_+$ and $\mathcal P_-$. Conditional on any common realization of its internal randomness, it therefore returns the same value $\widehat\theta$ in both problems. Yet
\begin{equation}
\max\{|\nabla F_+(\widehat\theta)|,
|\nabla F_-(\widehat\theta)|\}
=\max\{|\widehat\theta-\mu_{\mathrm{sep}}|,
|\widehat\theta+\mu_{\mathrm{sep}}|\}
\ge\mu_{\mathrm{sep}}.
\end{equation}
Thus the output cannot be population stationary for both distributions on an event of positive probability. The same coupling rules out weighted-average stationarity of the trajectory. For any deterministic positive weights $\lambda_t$ and $\Lambda_T=\sum_{t<T}\lambda_t$,
\begin{align}
&\frac1{\Lambda_T}\sum_{t<T}\lambda_t
|\nabla F_+(\Theta_t)|
+\frac1{\Lambda_T}\sum_{t<T}\lambda_t
|\nabla F_-(\Theta_t)|\nonumber\\
&\qquad=\frac1{\Lambda_T}\sum_{t<T}\lambda_t
\bigl(|\Theta_t-\mu_{\mathrm{sep}}|+|\Theta_t+\mu_{\mathrm{sep}}|\bigr)
\ge2\mu_{\mathrm{sep}}.
\label{eq:fixed_shot_weighted_no_go}
\end{align}
Hence both weighted averages cannot converge to zero on the common-support event. No distribution-free almost-sure fixed-shot guarantee of either type is possible.
\end{proof}

\begin{proposition}[Independent reference-gradient certificate]
\label{prop:reference_gradient_certificate}
Let $D_{\mathrm{ref}}$ contain $K_{\mathrm{ref}}$ independent reference examples per class, let $n_{\mathrm{ref}}=NK_{\mathrm{ref}}$, and let $f_{\mathrm{ref}}$ be its empirical risk. The reference set $D_{\mathrm{ref}}$ is assumed independent of $D$ and of the optimizer randomness. Let $\delta\in(0,1)$ and the nonnegative proxy scales $\varsigma_{\mathrm{ref},T,i}$ be fixed. The conditional law given $D$ and the realized trajectory $\Theta_{0:T-1}$ is assumed to satisfy, for every $t<T$, $i\in\{1,\ldots,d\}$, and $s_{\mathrm{mgf}}\in\R$,
\begin{equation*}
\E_{D_{\mathrm{ref}}}\!\left[
\exp\!\left(s_{\mathrm{mgf}}[\partial_i f_{\mathrm{ref}}(\Theta_t)-\partial_iF(\Theta_t)]\right)
\middle|D,\Theta_{0:T-1}\right]
\le
\exp\!\left(\frac{s_{\mathrm{mgf}}^2\varsigma_{\mathrm{ref},T,i}^2}{2n_{\mathrm{ref}}}\right).
\end{equation*}
The vector $\mathsf r_{\mathrm{ref},T}(\delta)\in\R_+^d$ is defined by
\begin{equation*}
(\mathsf r_{\mathrm{ref},T}(\delta))_i
=\varsigma_{\mathrm{ref},T,i}
\sqrt{\frac{2\log(2dT/\delta)}{n_{\mathrm{ref}}}}.
\end{equation*}
Then, conditionally on $D$ and the realized trajectory, with probability at least $1-\delta$ over $D_{\mathrm{ref}}$, simultaneously for every $t<T$,
\begin{equation}
\|e_D(\Theta_t)\|_{[\mathbf{k}]}
\le
\|\nabla f(\Theta_t)-\nabla f_{\mathrm{ref}}(\Theta_t)\|_{[\mathbf{k}]}
+\|\mathsf r_{\mathrm{ref},T}(\delta)\|_{[\mathbf{k}]}.
\label{eq:reference_gradient_certificate}
\end{equation}
The fixed-support version replaces $\|\cdot\|_{[\mathbf{k}]}$ by $v\mapsto\|v_S\|_1$.
\end{proposition}

\begin{proof}
The argument is conditional on $D$ and the complete realized optimizer trajectory. This trajectory is independent of $D_{\mathrm{ref}}$. For each $t<T$ and coordinate $i$, the stratified reference average is $\varsigma_{\mathrm{ref},T,i}/\sqrt{n_{\mathrm{ref}}}$-sub-Gaussian around $\partial_iF(\Theta_t)$. Hence
\begin{equation}
\Pr\!\left(
|\partial_i f_{\mathrm{ref}}(\Theta_t)-\partial_iF(\Theta_t)|
>(\mathsf r_{\mathrm{ref},T}(\delta))_i
\right)
\le\frac{\delta}{dT}.
\end{equation}
A union bound over $i\in\{1,\ldots,d\}$ and $t\in\{0,\ldots,T-1\}$ gives simultaneous coordinate control with probability at least $1-\delta$. Coordinatewise domination implies
\begin{equation}
\|\nabla f_{\mathrm{ref}}(\Theta_t)-\nabla F(\Theta_t)\|_{[\mathbf{k}]}
\le\|\mathsf r_{\mathrm{ref},T}(\delta)\|_{[\mathbf{k}]}.
\end{equation}
The triangle inequality applied to
\begin{equation}
e_D(\Theta_t)
=\nabla f(\Theta_t)-\nabla f_{\mathrm{ref}}(\Theta_t)
+\nabla f_{\mathrm{ref}}(\Theta_t)-\nabla F(\Theta_t)
\end{equation}
proves Eq.~\eqref{eq:reference_gradient_certificate}. Restriction to a fixed support gives the last statement.
\end{proof}

\subsection{Auxiliary descent, support, and alignment lemmas}
\suppanchor{descent-lemmas}
The following lemmas use the decomposition in Eq.~\eqref{eq:population_gradient_decomposition}, conditional on $D$.

\begin{lemma}[One-step population descent]
\label{lem:appE:one_step}
Under Assumption~\ref{assumption_1}(1), let $s\in\{-1,0,1\}^d$, let $\lambda>0$, and let $\Theta^+=\Theta-\lambda s$. The candidate iterates are assumed to satisfy $\Theta,\Theta^+\in\mathcal A_T$. Then
\begin{equation}
F(\Theta^+)\le F(\Theta)-\lambda\langle\nabla F(\Theta),s\rangle
+\frac{\lambda^2}{2}\sum_{i=1}^{d}L_i s_i^2.
\end{equation}
\end{lemma}
\begin{proof}
Equation~\eqref{eq:ass_population_coord_smooth} applies with $\Theta'=\Theta^+$.
\end{proof}

\begin{lemma}[Fixed-support robust sign alignment]
\label{lem:appE:fixed_support_alignment}
For every $a,b\in\R^d$ and support $S$,
\begin{equation}
\langle a,m_S\odot\sgn(a+b)\rangle
\ge\|a_S\|_1-2\|b_S\|_1.
\label{eq:appE:fixed_support_alignment}
\end{equation}
\end{lemma}
\begin{proof}
For $i\in S$, sign agreement gives $a_i\sgn(a_i+b_i)=|a_i|$. Under disagreement, including $a_i+b_i=0$, one has $|b_i|\ge|a_i|$ and
$a_i\sgn(a_i+b_i)\ge-|a_i|\ge|a_i|-2|b_i|$. Summing over $S$ proves the claim.
\end{proof}

\begin{lemma}[Static support missed mass]
\label{lem:appE:static_support_miss}
The support is assumed to satisfy $|S\cap\mathcal I_r|=k_r$ for every block. Then $\mathcal E_{\mathrm{miss},t}(S)\ge0$. If $w_0\in\R^d$ and
\begin{equation}
S=\bigcup_{r=1}^{R}\Top_{k_r}(|h_{0,\mathcal I_r}+w_{0,\mathcal I_r}|),
\end{equation}
then
\begin{equation}
\mathcal E_{\mathrm{miss},t}(S)
\le2\|h_t-h_0\|_{[\mathbf{k}]}+2\|w_0\|_{[\mathbf{k}]}.
\label{eq:static_support_drift_bound}
\end{equation}
\end{lemma}
\begin{proof}
Any support with $k_r$ entries per block captures no more absolute mass than the blockwise top-$k$ support. At $t=0$, oracle optimality and the triangle inequality give
\begin{equation}
\|h_0\|_{[\mathbf{k}]}-\|h_{0,S}\|_1\le2\|w_0\|_{[\mathbf{k}]}.
\end{equation}
Moreover,
\begin{align}
\mathcal E_{\mathrm{miss},t}(S)
&\le\mathcal E_{\mathrm{miss},0}(S)
+\|h_t-h_0\|_{[\mathbf{k}]}+\|(h_t-h_0)_S\|_1\\
&\le\mathcal E_{\mathrm{miss},0}(S)+2\|h_t-h_0\|_{[\mathbf{k}]}.
\end{align}
The claim follows by combining the bounds.
\end{proof}

\begin{proposition}[Necessity of static support capture]
\label{prop:static_support_no_go}
If a fixed support $S$ omits a coordinate $i\in\{1,\ldots,d\}$, no distribution-free theorem can guarantee full-dimensional stationarity for every smooth lower-bounded objective.
\end{proposition}
\begin{proof}
The construction uses $F(\Theta)=\Theta_i^2/2$, initialization $\Theta_{0,i}=1$, and an exact noiseless gradient oracle. Every static update is supported on $S$, so $\Theta_{t,i}=1$ for all $t$. Hence $\|\nabla F(\Theta_t)\|_\infty=1$. The failure is solely due to the omitted population-gradient mass.
\end{proof}

\begin{proposition}[Necessity of calibration under persistent sign noise]
\label{prop:persistent_noise_no_go}
Unbiasedness and a finite first moment of a fixed-variance gradient oracle do not by themselves imply positive population alignment for a sign update.
\end{proposition}
\begin{proof}
The construction uses $F(\theta)=\theta^2/2$ and $g=\theta+Z$, where $Z=9$ with probability $0.1$ and $Z=-1$ with probability $0.9$. Then $\E[Z]=0$, so $g$ is an unbiased gradient oracle. For every $\theta\in(0,1)$,
\begin{equation}
\E[\sgn(g)]=0.1-0.9=-0.8,
\qquad
\E[\theta\sgn(g)]=-0.8\theta<0.
\end{equation}
Therefore, the expected sign direction points uphill throughout $(0,1)$. A persistent-noise exact-stationarity result must impose sign calibration, symmetry, or another condition stronger than unbiasedness and finite first moment.
\end{proof}

\begin{lemma}[Blockwise top-$k$ captures gradient mass]
\label{lem:appE:topk_mass}
For every $a\in\R^d$,
\begin{equation}
\|a\|_{[\mathbf{k}]}\ge\underline\rho\,\|a\|_1.
\label{eq:appE:block_topk_mass}
\end{equation}
\end{lemma}
\begin{proof}
Within block $r$, the average of the largest $k_r$ magnitudes is at least the average of all $d_r$ magnitudes. Thus
$\|a_{\mathcal I_r}\|_{[k_r]}\ge(k_r/d_r)\|a_{\mathcal I_r}\|_1$. Summing over blocks proves the claim.
\end{proof}

\begin{lemma}[Robust blockwise top-$k$ alignment]
\label{lem:appE:block_topk_robust_alignment}
For all $a,b\in\R^d$,
\begin{equation}
\langle a,Q_{\mathbf{k}}(a+b)\rangle
\ge\|a\|_{[\mathbf{k}]}-2\|b\|_{[\mathbf{k}]}.
\label{eq:appE:block_topk_robust_alignment}
\end{equation}
The factor $2$ cannot be improved uniformly.
\end{lemma}
\begin{proof}
The blockwise oracle identity gives
$\langle a+b,Q_{\mathbf{k}}(a+b)\rangle=\|a+b\|_{[\mathbf{k}]}$.
Since $Q_{\mathbf{k}}(a+b)\in\mathcal C_{\mathbf{k}}$,
$|\langle b,Q_{\mathbf{k}}(a+b)\rangle|\le\|b\|_{[\mathbf{k}]}$.
The triangle inequality gives $\|a+b\|_{[\mathbf{k}]}\ge\|a\|_{[\mathbf{k}]}-\|b\|_{[\mathbf{k}]}$. Sharpness follows in one dimension from $a=1$, $b=-1-\varepsilon_{\mathrm{sharp}}$ with $\varepsilon_{\mathrm{sharp}}>0$, and the limit $\varepsilon_{\mathrm{sharp}}\downarrow0$.
\end{proof}

\begin{lemma}[Selection-probability and sign-reliability decomposition]
\label{lem:appE:selection_reliability}
Let $D$ and $t$ be fixed, with $h_t=\nabla F(\Theta_t)$. The conditional selection probability is
\begin{equation}
\pi_{t,i}^{\mathrm{sel}}
=\Pr(i\in S_t\mid D,\Theta_t).
\end{equation}
When $\pi_{t,i}^{\mathrm{sel}}>0$ and $h_{t,i}\ne0$, the conditional selected-sign reliability is
\begin{equation}
\chi_{t,i}^{\mathrm{sel}}
=\E[\sgn(h_{t,i})\sgn(g_{t,i})\mid D,\Theta_t,i\in S_t],
\end{equation}
and it is zero otherwise. Then
\begin{equation}
\mathfrak a_t^{\mathrm{dyn}}
=\sum_{i=1}^{d}|h_{t,i}|\pi_{t,i}^{\mathrm{sel}}\chi_{t,i}^{\mathrm{sel}}.
\label{eq:appE:selection_reliability_decomposition}
\end{equation}
For the static method, the conditional sign reliability is
\begin{equation}
\chi_{t,i}^{\mathrm{stat}}
=\E[\sgn(h_{t,i})\sgn(g_{t,i})\mid D,S,\Theta_t]
\end{equation}
when $i\in S$ and $h_{t,i}\ne0$, and $\chi_{t,i}^{\mathrm{stat}}=0$ otherwise. Then
\begin{equation}
\mathfrak a_t^{\mathrm{stat}}
=\sum_{i\in S}|h_{t,i}|\chi_{t,i}^{\mathrm{stat}}.
\end{equation}
For $\pi_{t,i}^{\mathrm{sel}}>0$ and $h_{t,i}\ne0$, if the selected stochastic gradient is almost surely nonzero, the conditional sign-correctness probability is
\begin{equation}
p_{t,i}^{\mathrm{corr}\mid\mathrm{sel}}
=\Pr\!\left(\sgn(g_{t,i})=\sgn(h_{t,i})\mid D,\Theta_t,i\in S_t\right).
\end{equation}
Then $\chi_{t,i}^{\mathrm{sel}}=2p_{t,i}^{\mathrm{corr}\mid\mathrm{sel}}-1$. When $\pi_{t,i}^{\mathrm{sel}}=0$ or $h_{t,i}=0$, both sides of the corresponding summand in Eq.~\eqref{eq:appE:selection_reliability_decomposition} are defined to be zero.
\end{lemma}
\begin{proof}
The dynamic direction coordinate is
$\1\{i\in S_t\}\sgn(g_{t,i})$. Multiplication by $h_{t,i}$, conditioning on selection, and summation yield the dynamic identity. The static identity is the same calculation with deterministic selection indicator $\1\{i\in S\}$.
\end{proof}

\subsection{Trajectory-wise population bounds}
\suppanchor{population-bounds}

\begin{theorem}[Trajectory-wise population bounds]
\label{thm:population_additive_bounds}
Under Assumption~\ref{assumption_1}(1), the bounds are conditional on $D$ and, for SignSparseSGD, also on its initial support $S$. Then
\begin{align}
\frac1{\Lambda_T}\sum_{t=0}^{T-1}\lambda_t
\E[\|\nabla F(\Theta_t)\|_{[\mathbf{k}]}\mid D,S]
\le{}&\frac{F(\Theta_0)-F_\star}{\Lambda_T}
+\frac{\|L_S\|_1}{2\Lambda_T}\sum_{t=0}^{T-1}\lambda_t^2\nonumber\\
&+\frac1{\Lambda_T}\sum_{t=0}^{T-1}\lambda_t
\E[\mathcal E_{\mathrm{miss},t}(S)\mid D,S]\nonumber\\
&+\frac2{\Lambda_T}\sum_{t=0}^{T-1}\lambda_t
\E[\|e_D(\Theta_t)_S\|_1+\|\xi_{t,S}\|_1\mid D,S].
\label{eq:signsparse_rate}
\end{align}
For DynSignSparseSGD,
\begin{align}
\frac1{\Lambda_T}\sum_{t=0}^{T-1}\lambda_t
\E[\|\nabla F(\Theta_t)\|_{[\mathbf{k}]}\mid D]
\le{}&\frac{F(\Theta_0)-F_\star}{\Lambda_T}
+\frac1{2\Lambda_T}\sum_{t=0}^{T-1}\lambda_t^2\E[L_t^{\mathrm{sel}}\mid D]\nonumber\\
&+\frac2{\Lambda_T}\sum_{t=0}^{T-1}\lambda_t
\E[\|e_D(\Theta_t)\|_{[\mathbf{k}]}+\|\xi_t\|_{[\mathbf{k}]}\mid D].
\label{eq:dynsignsparse_data_dependent_rate}
\end{align}
\end{theorem}

\begin{proof}
For the static method let $s_t=m_S\odot\sgn(g_t)$. Lemmas~\ref{lem:appE:one_step} and \ref{lem:appE:fixed_support_alignment}, with $b=e_D(\Theta_t)+\xi_t$, give
\begin{align}
F(\Theta_{t+1})
\le{}&F(\Theta_t)-\lambda_t\|h_{t,S}\|_1
+2\lambda_t\|e_D(\Theta_t)_S\|_1
+2\lambda_t\|\xi_{t,S}\|_1
+\frac{\lambda_t^2}{2}\|L_S\|_1.
\end{align}
The identity $\|h_{t,S}\|_1=\|h_t\|_{[\mathbf{k}]}-\mathcal E_{\mathrm{miss},t}(S)$ expresses the selected gradient mass. Taking conditional expectations, summing over $t$, and applying $F(\Theta_T)\ge F_\star$ yield the result after division by $\Lambda_T$.

For the dynamic method, Lemmas~\ref{lem:appE:one_step} and \ref{lem:appE:block_topk_robust_alignment} give
\begin{align}
F(\Theta_{t+1})
\le{}&F(\Theta_t)-\lambda_t\|h_t\|_{[\mathbf{k}]}
+2\lambda_t\|e_D(\Theta_t)\|_{[\mathbf{k}]}
+2\lambda_t\|\xi_t\|_{[\mathbf{k}]}
+\frac{\lambda_t^2}{2}L_t^{\mathrm{sel}}.
\end{align}
The same telescoping argument proves Eq.~\eqref{eq:dynsignsparse_data_dependent_rate}. Under Assumption~\ref{assumption_1}(2), the constant-step specialization uses $L_t^{\mathrm{sel}}\le\bar L_{\mathbf{k}}$ and
\begin{equation}
\E[\|\xi_t\|_{[\mathbf{k}]}\mid D]
\le\E[\|\xi_t\|_1\mid D]
\le\frac{\|\nu\|_1}{b_t^\gamma}.
\end{equation}

For completeness, Lemma~\ref{lem:appE:topk_mass} converts the dynamic bound to
\begin{align}
\frac1{\Lambda_T}\sum_{t<T}\lambda_t
\E[\|\nabla F(\Theta_t)\|_1\mid D]
\le\frac1{\underline\rho}\Bigg[&
\frac{F(\Theta_0)-F_\star}{\Lambda_T}
+\frac1{2\Lambda_T}\sum_{t<T}\lambda_t^2\E[L_t^{\mathrm{sel}}\mid D]
\nonumber\\
&+\frac2{\Lambda_T}\sum_{t<T}\lambda_t
\E[\|e_D(\Theta_t)\|_{[\mathbf{k}]}+\|\xi_t\|_{[\mathbf{k}]}\mid D]
\Bigg].
\label{eq:dynsignsparse_l1_corollary}
\end{align}
The inverse-density factor appears only in this norm conversion.

\end{proof}

\subsection{Calibrated population alignment}
\suppanchor{alignment}

\begin{theorem}[Exact weighted-average stationarity under calibrated population alignment]
\label{thm:calibrated_population_alignment}
Under Assumption~\ref{assumption_1}(1), the bounds are conditional on $D$. Let a function $\psi_{\mathrm{dyn}}:[0,\infty)\to[0,\infty)$ satisfy, almost surely for every $t<T$,
\begin{equation}
\mathfrak a_t^{\mathrm{dyn}}
\ge\psi_{\mathrm{dyn}}(\|\nabla F(\Theta_t)\|_{[\mathbf{k}]}).
\label{eq:dynamic_alignment_calibration}
\end{equation}
Then DynSignSparseSGD satisfies
\begin{equation}
\frac1{\Lambda_T}\sum_{t=0}^{T-1}\lambda_t
\E[\psi_{\mathrm{dyn}}(\|\nabla F(\Theta_t)\|_{[\mathbf{k}]})\mid D]
\le
\frac{F(\Theta_0)-F_\star}{\Lambda_T}
+\frac{\bar L_{\mathbf{k}}}{2\Lambda_T}\sum_{t=0}^{T-1}\lambda_t^2.
\label{eq:dynamic_alignment_rate}
\end{equation}
For SignSparseSGD, the conditioning additionally includes $S$. If $\psi_{\mathrm{stat}}:[0,\infty)\to[0,\infty)$ satisfies
\begin{equation*}
\mathfrak a_t^{\mathrm{stat}}
\ge\psi_{\mathrm{stat}}(\|\nabla F(\Theta_t)\|_{[\mathbf{k}]})
\end{equation*}
almost surely for every $t<T$, the analogous bound holds with $\psi_{\mathrm{stat}}$ and $\|L_S\|_1$. For the asymptotic conclusion, one finite vector $L$ and one finite lower bound $F_\star$ satisfy Assumption~\ref{assumption_1}(1) on $\bigcup_{T\ge1}\mathcal A_T$. The calibration functions and constants below are fixed along the infinite run, and the corresponding calibration inequality holds at every iteration. If $\psi_{\mathrm{dyn}}(x)\ge\omega_{\mathrm{dyn}}x^{\vartheta_{\mathrm{dyn}}}$ with $\omega_{\mathrm{dyn}}>0$ and $\vartheta_{\mathrm{dyn}}\ge1$, then the weighted average of $\E[\|\nabla F(\Theta_t)\|_{[\mathbf{k}]}\mid D]$ converges to zero whenever $\Lambda_T\to\infty$ and $\sum_{t<T}\lambda_t^2/\Lambda_T\to0$. The analogous conditional-expectation conclusion holds given $D,S$ for the static method under $\psi_{\mathrm{stat}}(x)\ge\omega_{\mathrm{stat}}x^{\vartheta_{\mathrm{stat}}}$ with $\omega_{\mathrm{stat}}>0$ and $\vartheta_{\mathrm{stat}}\ge1$. These are weighted-average conclusions in conditional expectation, not last-iterate or almost-sure convergence statements.
\end{theorem}

\begin{proof}
The argument is conditional on $D$ and $\Theta_t$. The one-step descent lemma and Eq.~\eqref{eq:dynamic_population_alignment} give
\begin{equation}
\E[F(\Theta_{t+1})\mid D,\Theta_t]
\le F(\Theta_t)-\lambda_t\mathfrak a_t^{\mathrm{dyn}}
+\frac{\lambda_t^2}{2}\E[L_t^{\mathrm{sel}}\mid D,\Theta_t].
\end{equation}
Equation~\eqref{eq:dynamic_alignment_calibration} and the bound $L_t^{\mathrm{sel}}\le\bar L_{\mathbf{k}}$ yield Eq.~\eqref{eq:dynamic_alignment_rate} after expectation over $\Theta_t$ and telescoping. The static proof substitutes $\mathfrak a_t^{\mathrm{stat}}$ and $\|L_S\|_1$.

If $\psi_{\mathrm{dyn}}(x)\ge\omega_{\mathrm{dyn}}x^{\vartheta_{\mathrm{dyn}}}$, Eq.~\eqref{eq:dynamic_alignment_rate} implies that the weighted average of $\E[\|\nabla F(\Theta_t)\|_{[\mathbf{k}]}^{\vartheta_{\mathrm{dyn}}}\mid D]$ vanishes under the stated step conditions. Jensen's inequality then gives
\begin{equation}
\frac1{\Lambda_T}\sum_{t<T}\lambda_t
\E[\|\nabla F(\Theta_t)\|_{[\mathbf{k}]}\mid D]
\le
\left(
\frac1{\Lambda_T}\sum_{t<T}\lambda_t
\E[\|\nabla F(\Theta_t)\|_{[\mathbf{k}]}^{\vartheta_{\mathrm{dyn}}}\mid D]
\right)^{1/\vartheta_{\mathrm{dyn}}},
\end{equation}
which also vanishes.

\end{proof}

\paragraph{Fixed-variance Gaussian example.}
Let $\sigma>0$, let $Z\sim\mathcal N(0,\sigma^2)$, and let $\Phi$ be the standard normal distribution function. For $F(\theta)=\theta^2/2$ and $g=\theta+Z$,
\begin{equation}
\E[\theta\sgn(g)]
=|\theta|\left\{2\Phi(|\theta|/\sigma)-1\right\}
\sim\sqrt{\frac{2}{\pi}}\frac{\theta^2}{\sigma}
\qquad(\theta\to0).
\end{equation}
For every $B_{\mathrm{traj}}>0$, continuity and positivity of
$x\mapsto\{2\Phi(x/\sigma)-1\}/x$ on $(0,B_{\mathrm{traj}}]$, together with its positive limit at zero, imply
\begin{equation}
|\theta|\{2\Phi(|\theta|/\sigma)-1\}
\ge\omega_{B_{\mathrm{traj}},\sigma}\theta^2,
\qquad |\theta|\le B_{\mathrm{traj}},
\end{equation}
for some $\omega_{B_{\mathrm{traj}},\sigma}>0$. Thus a fixed nonzero noise variance is compatible with vanishing weighted averages of expected gradient norms under a Robbins--Monro step schedule when the trajectory remains within one fixed bounded interval.

\subsection{Relative-error sufficient conditions}
\suppanchor{relative-errors}

\begin{corollary}[Relative-error and static-capture sufficient conditions]
\label{cor:relative_calibration_stationarity}
For the dynamic method, let $\upsilon_{\mathrm{shot}}^{\mathrm{dyn}}\ge0$ and $\upsilon_{\mathrm{mb}}^{\mathrm{dyn}}\ge0$. The following conditions are assumed to hold almost surely along the reachable trajectory:
\begin{equation*}
\|e_D(\Theta_t)\|_{[\mathbf{k}]}
\le\upsilon_{\mathrm{shot}}^{\mathrm{dyn}}\|\nabla F(\Theta_t)\|_{[\mathbf{k}]},
\qquad
\E[\|\xi_t\|_{[\mathbf{k}]}\mid D,\Theta_t]
\le\upsilon_{\mathrm{mb}}^{\mathrm{dyn}}\|\nabla F(\Theta_t)\|_{[\mathbf{k}]},
\end{equation*}
and let $\omega_{\mathrm{dyn}}=1-2\upsilon_{\mathrm{shot}}^{\mathrm{dyn}}-2\upsilon_{\mathrm{mb}}^{\mathrm{dyn}}>0$. Then the dynamic calibration holds with $\psi_{\mathrm{dyn}}(x)=\omega_{\mathrm{dyn}}x$.

For the static method, let $\upsilon_{\mathrm{cap}}^{\mathrm{stat}}\in(0,1]$, $\upsilon_{\mathrm{shot}}^{\mathrm{stat}}\ge0$, and $\upsilon_{\mathrm{mb}}^{\mathrm{stat}}\ge0$. The following conditions are assumed to hold almost surely:
\begin{equation}
\begin{aligned}
\|\nabla_SF(\Theta_t)\|_1
&\ge\upsilon_{\mathrm{cap}}^{\mathrm{stat}}\|\nabla F(\Theta_t)\|_{[\mathbf{k}]},\\
\|e_D(\Theta_t)_S\|_1
&\le\upsilon_{\mathrm{shot}}^{\mathrm{stat}}\|\nabla F(\Theta_t)\|_{[\mathbf{k}]},\\
\E[\|\xi_{t,S}\|_1\mid D,S,\Theta_t]
&\le\upsilon_{\mathrm{mb}}^{\mathrm{stat}}\|\nabla F(\Theta_t)\|_{[\mathbf{k}]}.
\end{aligned}
\label{eq:static_relative_errors}
\end{equation}
If $\omega_{\mathrm{stat}}=\upsilon_{\mathrm{cap}}^{\mathrm{stat}}-2\upsilon_{\mathrm{shot}}^{\mathrm{stat}}-2\upsilon_{\mathrm{mb}}^{\mathrm{stat}}>0$, then the static calibration holds with $\psi_{\mathrm{stat}}(x)=\omega_{\mathrm{stat}}x$. Under the horizon-uniform and step conditions of Theorem~\ref{thm:calibrated_population_alignment}, with the relative-error constants fixed along the run, the corresponding weighted averages of conditional expected blockwise top-$k$ population-gradient norms converge to zero.
\end{corollary}

\begin{proof}
For the dynamic method, Lemma~\ref{lem:appE:block_topk_robust_alignment} and conditional expectation give
\begin{align}
\mathfrak a_t^{\mathrm{dyn}}
&\ge\|\nabla F(\Theta_t)\|_{[\mathbf{k}]}
-2\|e_D(\Theta_t)\|_{[\mathbf{k}]}
-2\E[\|\xi_t\|_{[\mathbf{k}]}\mid D,\Theta_t]\\
&\ge\omega_{\mathrm{dyn}}\|\nabla F(\Theta_t)\|_{[\mathbf{k}]}.
\end{align}
For the static method, the first inequality in Eq.~\eqref{eq:static_relative_errors} is equivalent to
$\mathcal E_{\mathrm{miss},t}(S)\le(1-\upsilon_{\mathrm{cap}}^{\mathrm{stat}})\|\nabla F(\Theta_t)\|_{[\mathbf{k}]}$.
Thus the missed mass is controlled relatively rather than required to vanish as an external additive sequence. Lemma~\ref{lem:appE:fixed_support_alignment} and Eq.~\eqref{eq:static_relative_errors} yield
\begin{align}
\mathfrak a_t^{\mathrm{stat}}
&\ge\|\nabla_SF(\Theta_t)\|_1
-2\|e_D(\Theta_t)_S\|_1
-2\E[\|\xi_{t,S}\|_1\mid D,S,\Theta_t]\\
&\ge\omega_{\mathrm{stat}}\|\nabla F(\Theta_t)\|_{[\mathbf{k}]}.
\end{align}
Theorem~\ref{thm:calibrated_population_alignment} completes both conclusions.

\end{proof}

\begin{corollary}[Additive-error neighborhoods]
\label{cor:population_stationarity}
The assumptions of Theorem~\ref{thm:population_additive_bounds} hold along one infinite run. One finite vector $L$ and one finite lower bound $F_\star$ satisfy Assumption~\ref{assumption_1}(1) on $\bigcup_{T\ge1}\mathcal A_T$. The steps satisfy $\Lambda_T\to\infty$ and $\sum_{t<T}\lambda_t^2/\Lambda_T\to0$. For DynSignSparseSGD,
\begin{align}
\limsup_{T\to\infty}
\frac1{\Lambda_T}\sum_{t<T}\lambda_t
\E[\|\nabla F(\Theta_t)\|_{[\mathbf{k}]}\mid D]
\le{}&2\limsup_{T\to\infty}
\frac1{\Lambda_T}\sum_{t<T}\lambda_t
\E[\|e_D(\Theta_t)\|_{[\mathbf{k}]}+\|\xi_t\|_{[\mathbf{k}]}\mid D].
\label{eq:additive_dynamic_neighborhood}
\end{align}
For SignSparseSGD,
\begin{align}
\limsup_{T\to\infty}
\frac1{\Lambda_T}\sum_{t<T}\lambda_t
\E[\|\nabla F(\Theta_t)\|_{[\mathbf{k}]}\mid D,S]
\le{}&\limsup_{T\to\infty}
\frac1{\Lambda_T}\sum_{t<T}\lambda_t
\E[\mathcal E_{\mathrm{miss},t}(S)\mid D,S]\\
&+2\limsup_{T\to\infty}
\frac1{\Lambda_T}\sum_{t<T}\lambda_t
\E[\|e_D(\Theta_t)_S\|_1+\|\xi_{t,S}\|_1\mid D,S].
\label{eq:additive_static_neighborhood}
\end{align}
The weighted average of the corresponding conditional expected population-gradient norm converges to zero if the terms on the right-hand side vanish. Otherwise these inequalities quantify the additive-error neighborhood.
\end{corollary}

\begin{proof}
Equation~\eqref{eq:dynsignsparse_data_dependent_rate} applies with $L_t^{\mathrm{sel}}\le\bar L_{\mathbf{k}}$. The objective-gap and curvature terms vanish under the stated step conditions. Taking the limit superior gives Eq.~\eqref{eq:additive_dynamic_neighborhood}. The static result follows from Eq.~\eqref{eq:signsparse_rate}. If the additive terms do not vanish, Proposition~\ref{prop:static_support_no_go} and the fixed-bias examples show that no distribution-free exact statement is possible. The population-alignment theorem provides a separate exact route that does not require the absolute error magnitudes to vanish.
\end{proof}

\section{Confidence Gating in the Few-Shot Regime}
\label{app:confidence_gating}
\label{sec:appX}
\suppanchor{confidence}

This appendix gives a confidence-gating interpretation of SignSparseSGD and DynSignSparseSGD. It relates their binary masks to dense SignSGD~\cite{bernstein2018signsgd} and to the variance-adapted sign interpretation of Adam~\cite{kingma2014adam} according to work of~\cite{balles2018dissecting}. We remind the reader about some important notation summarized at the beginning of this supplement and defined in Appendix~\ref{app:convergence}. In particular, we condition on the support set $D$ and the current iterate $\Theta_t$, and write $h_t=\nabla F(\Theta_t)$. The coordinate step is $\lambda_t$. The constant step $\lambda=\alpha\eta$ in Eq.~\eqref{eq:sparse_sign_update} is its algorithmic special case. The central optimizer comparisons are conditional one-step comparisons of coordinate-wise smoothness upper bounds. They are complemented by a random-walk calculation and a stylized short-horizon estimation lemma. %None of the results asserts that one optimizer dominates another for every objective or noise distribution.

\subsection{Exact confidence-gated descent}
\suppanchor{confidence-exact}

Appendix~\ref{app:convergence} defines the dynamic selection probabilities $\pi_{t,i}^{\mathrm{sel}}$ and the conditional selected-sign reliabilities $\chi_{t,i}^{\mathrm{sel}}$. Their product retains the dependence between the top-$|g_t|$ mask and the sign computed from the same stochastic gradient. For comparison with dense SignSGD, the dense sign reliability is
\begin{equation}
\chi_{t,i}^{\mathrm{dense}}
=
\begin{cases}
\E[\sgn(h_{t,i})\sgn(g_{t,i})\mid D,\Theta_t], & h_{t,i}\ne0,\\
0, & h_{t,i}=0.
\end{cases}
\label{eq:cg_dense_reliability}
\end{equation}
When $h_{t,i}\ne0$ and $\Pr(g_{t,i}=0\mid D,\Theta_t)=0$, let
\begin{equation}
p_{t,i}^{\mathrm{corr}}
=\Pr\!\left(\sgn(g_{t,i})=\sgn(h_{t,i})\mid D,\Theta_t\right).
\label{eq:cg_unconditional_correctness}
\end{equation}
Then $\chi_{t,i}^{\mathrm{dense}}=2p_{t,i}^{\mathrm{corr}}-1$.

\begin{proposition}[Exact confidence-gated one-step bounds]
\label{prop:cg_exact_bounds}
Let $D$ and $\Theta_t$ be fixed. The coordinate-wise smoothness inequality in Eq.~\eqref{eq:ass_population_coord_smooth} is assumed to hold between $\Theta_t$ and each candidate next iterate considered below. Then DynSignSparseSGD satisfies
\begin{align}
\E[F(\Theta_{t+1})\mid D,\Theta_t]
\le{}&F(\Theta_t)
-\lambda_t\sum_{i=1}^{d}|h_{t,i}|\pi_{t,i}^{\mathrm{sel}}\chi_{t,i}^{\mathrm{sel}}
+\frac{\lambda_t^2}{2}\sum_{i=1}^{d}L_i\pi_{t,i}^{\mathrm{sel}}.
\label{eq:cg_dynamic_bound}
\end{align}
For SignSparseSGD, the conditioning additionally includes its fixed support $S$. Then
\begin{align}
\E[F(\Theta_{t+1})\mid D,S,\Theta_t]
\le{}&F(\Theta_t)
-\lambda_t\sum_{i\in S}|h_{t,i}|\chi_{t,i}^{\mathrm{stat}}
+\frac{\lambda_t^2}{2}\sum_{i\in S}L_i.
\label{eq:cg_static_bound}
\end{align}
Dense SignSGD with the same coordinate step satisfies
\begin{align}
\E[F(\Theta_{t+1})\mid D,\Theta_t]
\le{}&F(\Theta_t)
-\lambda_t\sum_{i=1}^{d}|h_{t,i}|\chi_{t,i}^{\mathrm{dense}}
+\frac{\lambda_t^2}{2}\sum_{i=1}^{d}L_i.
\label{eq:cg_dense_bound}
\end{align}
\end{proposition}

\begin{proof}
The coordinate-wise smoothness inequality in Eq.~\eqref{eq:ass_population_coord_smooth} applies to the candidate update. For the dynamic direction, $(s_t)_i=\1\{i\in S_t\}\sgn(g_{t,i})$. Therefore,
\begin{align*}
F(\Theta_{t+1})
\le F(\Theta_t)
-\lambda_t\sum_{i=1}^{d}h_{t,i}\1\{i\in S_t\}\sgn(g_{t,i})
+\frac{\lambda_t^2}{2}\sum_{i=1}^{d}L_i\1\{i\in S_t\}.
\end{align*}
Taking the conditional expectation and applying Lemma~\ref{lem:appE:selection_reliability} give Eq.~\eqref{eq:cg_dynamic_bound}. The same calculation with the deterministic mask $m_S$ gives Eq.~\eqref{eq:cg_static_bound}. Setting the mask to the all-ones vector gives Eq.~\eqref{eq:cg_dense_bound}. The curvature terms remain valid when a selected stochastic-gradient coordinate is zero because $\sgn(0)^2=0\le1$.
\end{proof}

Equation~\eqref{eq:cg_dynamic_bound} is the exact confidence-gating decomposition for the implemented same-gradient selector. The useful population-alignment mass is $\sum_i|h_{t,i}|\pi_{t,i}^{\mathrm{sel}}\chi_{t,i}^{\mathrm{sel}}$. The expected curvature cost is restricted to $\sum_iL_i\pi_{t,i}^{\mathrm{sel}}$. No independence between selection and sign correctness is assumed.

\subsection{Continuous confidence factors and binary gates}
\suppanchor{confidence-continuous}

The next lemma formalizes the connection to the sign-confidence factor of Balles and Hennig~\cite{balles2018dissecting}. It concerns directional estimation rather than objective decrease.

\begin{lemma}[Mean-square optimal confidence factor and binary threshold]
\label{lem:cg_mse_gate}
Let $i$ be a coordinate satisfying $h_{t,i}\ne0$, $\Pr(g_{t,i}=0\mid D,\Theta_t)=0$, and $p_{t,i}^{\mathrm{corr}}\ge1/2$. For a scalar attenuation $\varpi\in[0,1]$,
\begin{align}
&\E\!\left[
\left(\varpi\sgn(g_{t,i})-\sgn(h_{t,i})\right)^2
\middle|D,\Theta_t
\right]
\nonumber\\
&\qquad=
\left(\varpi-\chi_{t,i}^{\mathrm{dense}}\right)^2
+1-\left(\chi_{t,i}^{\mathrm{dense}}\right)^2.
\label{eq:cg_mse_identity}
\end{align}
The minimizer over $\varpi\in[0,1]$ is
$\varpi=\chi_{t,i}^{\mathrm{dense}}=2p_{t,i}^{\mathrm{corr}}-1$.
If $\varpi$ is restricted to $\{0,1\}$, updating is no worse than abstaining exactly when
\begin{equation}
p_{t,i}^{\mathrm{corr}}\ge\frac34
\quad\Longleftrightarrow\quad
\chi_{t,i}^{\mathrm{dense}}\ge\frac12.
\label{eq:cg_binary_threshold}
\end{equation}
Under an at-most-$k_r$ binary budget in block $r$, the mean-square optimal hard gate selects up to $k_r$ coordinates with the largest positive values of $2\chi_{t,i}^{\mathrm{dense}}-1$.
\end{lemma}

\begin{proof}
The random product $\sgn(g_{t,i})\sgn(h_{t,i})$ takes values in $\{-1,1\}$ and has conditional mean $\chi_{t,i}^{\mathrm{dense}}$. Expanding the square gives
\begin{equation*}
\varpi^2+1-2\varpi\chi_{t,i}^{\mathrm{dense}},
\end{equation*}
which yields Eq.~\eqref{eq:cg_mse_identity} by completing the square. The continuous minimizer follows because $\chi_{t,i}^{\mathrm{dense}}\in[0,1]$. The errors at $\varpi=0$ and $\varpi=1$ are $1$ and $2-2\chi_{t,i}^{\mathrm{dense}}$, respectively. Their comparison gives Eq.~\eqref{eq:cg_binary_threshold}. The blockwise statement follows by separability.
\end{proof}

Dense SignSGD fixes $\varpi=1$ for every coordinate. An ideal variance-adapted sign rule uses the continuous factor $\chi_{t,i}^{\mathrm{dense}}$. SignSparseSGD and DynSignSparseSGD instead use a binary factor: one on the active coordinates and zero elsewhere. Adam is related to the continuous interpretation, but it is not equal to the ideal factor. Using the bias-corrected first- and second-moment notation $\hat u_t$ and $\hat v_t$ from Appendix~\ref{app:algorithms}, and ignoring the denominator offset, a nonzero Adam coordinate can be written as
\begin{equation}
\frac{\hat u_{t,i}}{\sqrt{\hat v_{t,i}}}
=
\sgn(\hat u_{t,i})
\left(1+\frac{\hat v_{t,i}-\hat u_{t,i}^2}{\hat u_{t,i}^2}\right)^{-1/2}.
\label{eq:cg_adam_factorization}
\end{equation}
Thus Adam uses a smooth factor estimated from moving moments. When $\hat v_{t,i}\ge\hat u_{t,i}^2$, the bracketed term is a nonnegative relative-dispersion factor. Under a stationary population-moment interpretation, the factor decreases with relative variance and approximates sign confidence~\cite{balles2018dissecting}. Because Adam uses different exponential weights for its two moments, $\hat v_{t,i}-\hat u_{t,i}^2$ need not be a nonnegative variance estimate at every finite iteration. The hard gate avoids estimating and storing a continuous factor for every coordinate.

\begin{corollary}[First-step equivalence of bias-corrected Adam and dense SignSGD]
\label{cor:cg_adam_first_step}
Adam is initialized with $u_0=v_0=0$, coefficients $(\beta_1,\beta_2)\in[0,1)^2$, and zero denominator offset. Let $i$ be a coordinate with $g_{0,i}\ne0$. At the first update,
\begin{equation}
\hat u_{1,i}=g_{0,i},
\qquad
\hat v_{1,i}=g_{0,i}^2,
\qquad
\frac{\hat u_{1,i}}{\sqrt{\hat v_{1,i}}}=\sgn(g_{0,i}).
\label{eq:cg_adam_first_step}
\end{equation}
Hence, at an equal coordinate step, the first-step confidence allocation of bias-corrected Adam is identical to that of dense SignSGD. Adam has not yet accumulated information that can attenuate low-confidence coordinates, whereas SignSparseSGD and DynSignSparseSGD apply their blockwise hard gate immediately.
\end{corollary}

\begin{proof}
The first moment is $u_1=(1-\beta_1)g_0$ and the second moment is $v_1=(1-\beta_2)(g_0\odot g_0)$. Dividing by the corresponding bias-correction factors proves Eq.~\eqref{eq:cg_adam_first_step}.
\end{proof}

\subsection{Descent-optimal blockwise supports}
\suppanchor{confidence-oracle}

The mean-square criterion does not include gradient magnitude or curvature. The coordinate-wise smoothness model produces a different oracle.

\begin{theorem}[Oracle support for expected local descent]
\label{thm:cg_oracle_support}
Conditional on $D$ and $\Theta_t$, Eq.~\eqref{eq:ass_population_coord_smooth} holds between $\Theta_t$ and every candidate next iterate considered in the support comparison, with the same vector $L$. Let $S^\circ$ be a deterministic support selected before the current stochastic-gradient draw and satisfying $|S^\circ\cap\mathcal I_r|=k_r$ for every block $r$. The selected coordinates satisfy $\Pr(g_{t,i}=0\mid D,\Theta_t)=0$ for every $i\in S^\circ$. Then
\begin{align}
\E[F(\Theta_{t+1})\mid D,\Theta_t]
\le F(\Theta_t)
-\lambda_t\sum_{i\in S^\circ}
\left[
|h_{t,i}|\chi_{t,i}^{\mathrm{dense}}
-\frac{\lambda_t}{2}L_i
\right].
\label{eq:cg_oracle_support_bound}
\end{align}
Among supports with the same blockwise cardinalities, the best bound is obtained by selecting, within each block, the $k_r$ largest values of
\begin{equation}
|h_{t,i}|\chi_{t,i}^{\mathrm{dense}}-\frac{\lambda_t}{2}L_i.
\label{eq:cg_oracle_score}
\end{equation}
If at-most-$k_r$ supports are allowed, coordinates with a nonpositive value in Eq.~\eqref{eq:cg_oracle_score} should not be activated.
\end{theorem}

\begin{proof}
Because $S^\circ$ is fixed before the current draw, its selected-sign reliability equals $\chi_{t,i}^{\mathrm{dense}}$ on $S^\circ$. Equation~\eqref{eq:cg_static_bound} therefore gives Eq.~\eqref{eq:cg_oracle_support_bound}. The support optimization is separable across coordinates and blocks, which proves both selection statements.
\end{proof}

The score in Eq.~\eqref{eq:cg_oracle_score} combines population-gradient magnitude, sign confidence, and coordinate curvature. Raw top-magnitude selection is not universally optimal. The following specialization centers the conditional stochastic gradient at $h_t$. This centering is an additional assumption and does not follow from Eq.~\eqref{eq:population_gradient_decomposition}. If the minibatch gradient is conditionally centered at $\nabla f(\Theta_t)$, it requires $e_D(\Theta_t)=0$. Under common noise and curvature, the lemma establishes expected-magnitude ordering. Recovery from one observed gradient requires the separate gap condition below.

\begin{lemma}[Homoscedastic Gaussian confidence ordering]
\label{lem:cg_gaussian_ordering}
The conditional distribution given $D$ and $\Theta_t$ is assumed to satisfy
\begin{equation}
g_{t,i}\mid(D,\Theta_t)
\sim
\mathcal N\!\left(
 h_{t,i},
 (\varsigma_{t,i}^{\mathrm{noise}})^2
\right),
\qquad
\varsigma_{t,i}^{\mathrm{noise}}>0.
\label{eq:cg_gaussian_noise}
\end{equation}
Then, for $h_{t,i}\ne0$,
\begin{equation}
\chi_{t,i}^{\mathrm{dense}}
=
\operatorname{erf}\!\left(
\frac{|h_{t,i}|}{\sqrt2\,\varsigma_{t,i}^{\mathrm{noise}}}
\right).
\label{eq:cg_gaussian_reliability}
\end{equation}
If $\varsigma_{t,i}^{\mathrm{noise}}$ and $L_i$ are constant within block $r$, let $\varsigma_{t,r}^{\mathrm{noise}}$ denote the common noise scale. Then the oracle score in Eq.~\eqref{eq:cg_oracle_score} is strictly increasing in $|h_{t,i}|$ for $|h_{t,i}|>0$. Moreover, $\E[|g_{t,i}|\mid D,\Theta_t]$ is strictly increasing in $|h_{t,i}|$ under the same common noise scale.

For a block with $k_r<d_r$, the true-magnitude boundary is assumed to have a positive gap. If
\begin{equation}
\max_{i\in\mathcal I_r}|g_{t,i}-h_{t,i}|
<
\frac12\left(
|h_{t,\mathcal I_r}|_{(k_r)}-|h_{t,\mathcal I_r}|_{(k_r+1)}
\right),
\label{eq:cg_rank_recovery_condition}
\end{equation}
then
$\Top_{k_r}(|g_{t,\mathcal I_r}|)=\Top_{k_r}(|h_{t,\mathcal I_r}|)$.
\end{lemma}

\begin{proof}
For a positive $h_{t,i}$, the sign is correct exactly when $g_{t,i}>0$. Hence $p_{t,i}^{\mathrm{corr}}=\Phi(|h_{t,i}|/\varsigma_{t,i}^{\mathrm{noise}})$, and Eq.~\eqref{eq:cg_gaussian_reliability} follows from $2\Phi(x)-1=\operatorname{erf}(x/\sqrt2)$. For $a>0$ and a common scale $\varsigma_{t,r}^{\mathrm{noise}}$,
\begin{align*}
\frac{\mathrm d}{\mathrm da}
\left[
a\operatorname{erf}\!\left(\frac{a}{\sqrt2\,\varsigma_{t,r}^{\mathrm{noise}}}\right)
\right]
=
\operatorname{erf}\!\left(\frac{a}{\sqrt2\,\varsigma_{t,r}^{\mathrm{noise}}}\right)
+\sqrt{\frac2\pi}\frac{a}{\varsigma_{t,r}^{\mathrm{noise}}}
\exp\!\left(-\frac{a^2}{2(\varsigma_{t,r}^{\mathrm{noise}})^2}\right)>0.
\end{align*}
Subtracting the common curvature term preserves the ordering. Viewed as a scalar function of $a=|h_{t,i}|$, the folded-normal mean has derivative
$2\Phi(a/\varsigma_{t,r}^{\mathrm{noise}})-1>0$, which proves the expected-magnitude claim. Finally, Eq.~\eqref{eq:cg_rank_recovery_condition} prevents any coordinate below the top-$k_r$ boundary from overtaking a coordinate above it, which proves exact rank recovery.
\end{proof}

Under its conditional Gaussian model, this lemma separates step scaling from expected saliency ranking. A single observed magnitude can be badly calibrated as a numerical step size while its ordering remains informative because only perturbations that cross an order-statistic boundary change the mask. DynSignSparseSGD refreshes this one-sample proxy. SignSparseSGD estimates it once and can become stale. Strongly heteroscedastic noise breaks the ordering because Eq.~\eqref{eq:cg_gaussian_reliability} depends on signal-to-noise ratio rather than raw magnitude.

\subsection{Explicit dense-versus-sparse SignSGD regimes}
\label{subsec:cg_dense_sparse_regimes}
\suppanchor{confidence-regimes}

Subtracting Eqs.~\eqref{eq:cg_dynamic_bound} and~\eqref{eq:cg_dense_bound} gives an exact criterion for the actual dynamic gate.

\begin{corollary}[Sparse-versus-dense sign criterion]
\label{cor:cg_sparse_dense}
At iteration $t$, the DynSignSparseSGD upper bound in Eq.~\eqref{eq:cg_dynamic_bound} is no larger than the dense SignSGD upper bound in Eq.~\eqref{eq:cg_dense_bound} whenever
\begin{equation}
\sum_{i=1}^{d}|h_{t,i}|
\left(
\chi_{t,i}^{\mathrm{dense}}-\pi_{t,i}^{\mathrm{sel}}\chi_{t,i}^{\mathrm{sel}}
\right)
\le
\frac{\lambda_t}{2}\sum_{i=1}^{d}L_i\left(1-\pi_{t,i}^{\mathrm{sel}}\right).
\label{eq:cg_dynamic_dense_condition}
\end{equation}
For a deterministic support $S^\circ$ selected before the current draw, the condition reduces to
\begin{equation}
\sum_{i\notin S^\circ}
\left[
|h_{t,i}|\chi_{t,i}^{\mathrm{dense}}
-\frac{\lambda_t}{2}L_i
\right]
\le0.
\label{eq:cg_deterministic_dense_condition}
\end{equation}
Strict inequality gives a strict improvement of the corresponding upper bound.
\end{corollary}

\begin{proof}
The criterion follows by requiring the right-hand side of Eq.~\eqref{eq:cg_dynamic_bound} to be no greater than that of Eq.~\eqref{eq:cg_dense_bound} and rearranging. For a deterministic support, $\pi_{t,i}^{\mathrm{sel}}=\1\{i\in S^\circ\}$ and $\chi_{t,i}^{\mathrm{sel}}=\chi_{t,i}^{\mathrm{dense}}$ on the support.
\end{proof}

Equation~\eqref{eq:cg_dynamic_dense_condition} has two terms. Its left-hand side is the confidence-weighted population-gradient mass sacrificed by gating. Its right-hand side is the expected curvature cost saved by abstaining. Selection can also raise conditional reliability, so $\chi_{t,i}^{\mathrm{sel}}$ need not equal $\chi_{t,i}^{\mathrm{dense}}$. By Corollary~\ref{cor:cg_adam_first_step}, at a matched coordinate step the same criterion compares the sparse-sign gate with the first bias-corrected Adam update when the denominator offset is neglected.

For a deterministic support $S^\circ\subsetneq\{1,\ldots,d\}$, the dense right-hand side minus the sparse right-hand side is exactly
\begin{align}
&\lambda_t\bigl(d-|S^\circ|\bigr)
\left[
\frac{\lambda_t}{2\bigl(d-|S^\circ|\bigr)}
\sum_{i\notin S^\circ}L_i
-
\frac{1}{d-|S^\circ|}
\sum_{i\notin S^\circ}|h_{t,i}|\chi_{t,i}^{\mathrm{dense}}
\right].
\label{eq:cg_omitted_average_gap}
\end{align}
Thus, at a matched coordinate step, sparse sign descent is favored when the average omitted reliability-weighted population-gradient mass is at most $\lambda_t/2$ times the average omitted curvature. The bracketed per-coordinate advantage is accumulated over $d-|S^\circ|$ omitted coordinates. This accumulation is relevant for few-shot adaptation of a large model, where the number of weak coordinates can be very large.

The matched-step criterion does not account for the fact that sparsity changes the natural learning-rate scale. We next compare the best quadratic upper-bound envelope available to each direction at the current iterate. The dense confidence-weighted alignment is
\begin{equation}
\mathfrak a_t^{\mathrm{dense}}
=\sum_{i=1}^{d}|h_{t,i}|\chi_{t,i}^{\mathrm{dense}}.
\label{eq:cg_dense_alignment_mass}
\end{equation}
The dynamic expected selected curvature is already represented by
$\E[L_t^{\mathrm{sel}}\mid D,\Theta_t]=\sum_iL_i\pi_{t,i}^{\mathrm{sel}}$.
When the denominators below are positive, the dynamic alignment ratio and curvature fraction are
\begin{equation}
\kappa_{t,\mathrm{align}}^{\mathrm{dyn}}
=
\frac{\mathfrak a_t^{\mathrm{dyn}}}{\mathfrak a_t^{\mathrm{dense}}},
\qquad
\kappa_{t,\mathrm{curv}}^{\mathrm{dyn}}
=
\frac{\E[L_t^{\mathrm{sel}}\mid D,\Theta_t]}
{\|L\|_1}.
\label{eq:cg_dynamic_capture_fractions}
\end{equation}
The alignment ratio can exceed one because conditioning on dynamic selection can increase sign reliability. The curvature fraction lies in $(0,1]$ under the positivity conditions below.

\begin{theorem}[Locally tuned dense-versus-sparse sign regime]
\label{thm:cg_tuned_dense_sparse_regime}
Let $D$ and $\Theta_t$ be fixed. The positivity conditions are
\begin{equation}
\mathfrak a_t^{\mathrm{dense}}>0,
\qquad
\mathfrak a_t^{\mathrm{dyn}}>0,
\qquad
\|L\|_1>0,
\qquad
\E[L_t^{\mathrm{sel}}\mid D,\Theta_t]>0.
\label{eq:cg_tuned_regime_positivity}
\end{equation}
The conditional gradient law and the resulting quantities in Eqs.~\eqref{eq:cg_dense_alignment_mass} and~\eqref{eq:cg_dynamic_capture_fractions} remain fixed while the scalar coordinate step varies over nonnegative values. The unconstrained minimizers of the dynamic and dense quadratic upper bounds in Proposition~\ref{prop:cg_exact_bounds} are
\begin{equation}
\lambda_{t,\mathrm{dyn}}^\star
=
\frac{\mathfrak a_t^{\mathrm{dyn}}}
{\E[L_t^{\mathrm{sel}}\mid D,\Theta_t]},
\qquad
\lambda_{t,\mathrm{dense}}^\star
=
\frac{\mathfrak a_t^{\mathrm{dense}}}
{\|L\|_1}.
\label{eq:cg_tuned_optimal_steps}
\end{equation}
Equation~\eqref{eq:ass_population_coord_smooth} is assumed to hold almost surely along both candidate update segments for every coordinate step between zero and
$\max\{\lambda_{t,\mathrm{dyn}}^\star,\lambda_{t,\mathrm{dense}}^\star\}$.
The corresponding guaranteed decreases below $F(\Theta_t)$ are
\begin{equation}
\frac{(\mathfrak a_t^{\mathrm{dyn}})^2}
{2\E[L_t^{\mathrm{sel}}\mid D,\Theta_t]}
\qquad\text{and}\qquad
\frac{(\mathfrak a_t^{\mathrm{dense}})^2}
{2\|L\|_1},
\label{eq:cg_tuned_optimal_decreases}
\end{equation}
respectively. Hence, the locally tuned DynSignSparseSGD envelope is no worse than the locally tuned dense SignSGD envelope if and only if
\begin{equation}
\kappa_{t,\mathrm{align}}^{\mathrm{dyn}}
\ge
\sqrt{\kappa_{t,\mathrm{curv}}^{\mathrm{dyn}}}.
\label{eq:cg_tuned_dense_sparse_threshold}
\end{equation}
Strict inequality gives a strict envelope advantage.
\end{theorem}

\begin{proof}
For positive scalars $a$ and $c$, the quadratic function
$-\lambda a+\lambda^2c/2$ on $\lambda\ge0$ is minimized at $\lambda=a/c$, with minimum $-a^2/(2c)$. This identity applies to Eqs.~\eqref{eq:cg_dynamic_bound} and~\eqref{eq:cg_dense_bound}. Comparing the two decreases in Eq.~\eqref{eq:cg_tuned_optimal_decreases} gives
\begin{equation*}
\frac{(\mathfrak a_t^{\mathrm{dyn}})^2}
{\E[L_t^{\mathrm{sel}}\mid D,\Theta_t]}
\ge
\frac{(\mathfrak a_t^{\mathrm{dense}})^2}
{\|L\|_1},
\end{equation*}
which is equivalent to Eq.~\eqref{eq:cg_tuned_dense_sparse_threshold} under Eq.~\eqref{eq:cg_tuned_regime_positivity}.
\end{proof}

The positivity conditions isolate the nondegenerate finite-minimizer case. If an alignment is nonpositive, its quadratic envelope is minimized at the boundary $\lambda=0$. If a curvature denominator is zero while the corresponding alignment is positive, the coordinate-wise quadratic model has no finite minimizing step and cannot support a tuned-step comparison without an additional step constraint.

\begin{corollary}[Homogeneous-curvature phase boundary]
\label{cor:cg_homogeneous_curvature_phase}
The coordinates are assumed to share a common positive curvature $L_i=L_{\mathrm{coord}}>0$. Because the dynamic selector chooses exactly $k_{\mathrm{step}}$ coordinates,
\begin{equation}
\kappa_{t,\mathrm{curv}}^{\mathrm{dyn}}
=
\frac{k_{\mathrm{step}}}{d}.
\label{eq:cg_homogeneous_curvature_fraction}
\end{equation}
Under the positivity conditions of Theorem~\ref{thm:cg_tuned_dense_sparse_regime}, the locally tuned dynamic sparse-sign envelope is no worse than the locally tuned dense-sign envelope if and only if
\begin{equation}
\frac{\mathfrak a_t^{\mathrm{dyn}}}
{\mathfrak a_t^{\mathrm{dense}}}
\ge
\sqrt{\frac{k_{\mathrm{step}}}{d}}.
\label{eq:cg_homogeneous_phase_threshold}
\end{equation}
For a deterministic support $S^\circ$ selected before the current draw and satisfying $|S^\circ|=k_{\mathrm{step}}$, the same statement holds after replacing $\mathfrak a_t^{\mathrm{dyn}}$ by
$\sum_{i\in S^\circ}|h_{t,i}|\chi_{t,i}^{\mathrm{dense}}$.
\end{corollary}

\begin{proof}
Equation~\eqref{eq:block_budgets} gives $|S_t|=k_{\mathrm{step}}$ almost surely, so $\sum_i\pi_{t,i}^{\mathrm{sel}}=k_{\mathrm{step}}$. Substitution of the common curvature into Eq.~\eqref{eq:cg_dynamic_capture_fractions} gives the claim by Theorem~\ref{thm:cg_tuned_dense_sparse_regime}. The deterministic-support statement is the same calculation using Eq.~\eqref{eq:cg_oracle_support_bound}.
\end{proof}

Corollary~\ref{cor:cg_homogeneous_curvature_phase} gives an explicit phase boundary. If reliability-weighted alignment is diffuse, a support containing a fraction $k_{\mathrm{step}}/d$ of the coordinates attains approximately the same alignment ratio. For a proper sparse support,
$k_{\mathrm{step}}/d<\sqrt{k_{\mathrm{step}}/d}$, so the locally tuned dense envelope is favored in this diffuse approximation. If reliable alignment is concentrated, the sparse support can exceed the square-root threshold and the locally tuned sparse envelope is favored. At an exactly $10^{-3}$ realized density, the threshold is $\sqrt{10^{-3}}\approx0.0316$: attaining an alignment ratio above about $3.16\%$ is sufficient in the homogeneous-curvature local envelope.

This criterion refines the density discussion of SignSGD~\cite{bernstein2018signsgd}. The relevant few-shot signal is not the raw population-gradient magnitude alone. It is the reliability-weighted mass $|h_{t,i}|\chi_{t,i}^{\mathrm{dense}}$. Few-shot noise can make reliable alignment highly concentrated even when the raw gradient is not sparse. The locally tuned dense envelope is favored when this reliable mass is diffuse. The locally tuned sign-sparse envelope is favored when the selected coordinates retain a sufficiently large alignment ratio while removing a larger fraction of curvature. The same quadratic-envelope algebra applies to the SignSparseSGD bound in Eq.~\eqref{eq:cg_static_bound}. In that comparison, the static and dense reliabilities must be evaluated under the same conditioning on $D$, $S$, and $\Theta_t$. A stale static mask can reduce its alignment ratio, whereas dynamic refreshing can restore it.

Under the Gaussian model in Eq.~\eqref{eq:cg_gaussian_noise}, the inequality $\operatorname{erf}(x)\le2x/\sqrt\pi$ yields
\begin{equation}
|h_{t,i}|\chi_{t,i}^{\mathrm{dense}}
\le
\sqrt{\frac2\pi}\frac{|h_{t,i}|^2}{\varsigma_{t,i}^{\mathrm{noise}}}.
\label{eq:cg_low_snr_upper_bound}
\end{equation}
Therefore, a sufficient coordinate-wise condition for omitting coordinate $i$ in Eq.~\eqref{eq:cg_deterministic_dense_condition} is
\begin{equation}
\sqrt{\frac2\pi}\frac{|h_{t,i}|^2}{\varsigma_{t,i}^{\mathrm{noise}}}
\le
\frac{\lambda_t}{2}L_i.
\label{eq:cg_low_snr_omit_condition}
\end{equation}
Few-shot noise increases the set of coordinates satisfying Eq.~\eqref{eq:cg_low_snr_omit_condition}. In a large model, dense SignSGD still pays curvature on all of them. A sparse gate can remove an aggregate cost that scales with the number of low-confidence coordinates.

The same mechanism limits noise-driven parameter drift. The next proposition treats the exactly zero-signal limit.

\begin{proposition}[Zero-signal random-walk suppression]
\label{prop:cg_random_walk}
Let $\mathcal J^{\mathrm{flat}}\subseteq\{1,\ldots,d\}$ be a fixed set satisfying $h_{t,i}=0$ for every $i\in\mathcal J^{\mathrm{flat}}$ and every $t<T$. For the dynamic trajectory, conditional on $D$ and its past, each product $\1\{i\in S_t\}\sgn(g_{t,i})$ has mean zero on this set. Every selected stochastic-gradient coordinate in this set is almost surely nonzero. Thus the signed increments are martingale differences across iterations. Then
\begin{equation}
\E\!\left[
\| (\Theta_T-\Theta_0)_{\mathcal J^{\mathrm{flat}}}\|_2^2
\middle|D
\right]
=
\sum_{t=0}^{T-1}\lambda_t^2
\E\!\left[
\sum_{i\in\mathcal J^{\mathrm{flat}}}\pi_{t,i}^{\mathrm{sel}}
\middle|D
\right].
\label{eq:cg_sparse_random_walk}
\end{equation}
For the dense trajectory, the formula
$|\mathcal J^{\mathrm{flat}}|\sum_{t<T}\lambda_t^2$
requires the dense signs $\sgn(g_{t,i})$ to have zero conditional mean given $D$ and that trajectory's past, and to be almost surely nonzero on $\mathcal J^{\mathrm{flat}}$. For the static trajectory, the corresponding mean-square displacement conditional on $D,S$ equals
$|S\cap\mathcal J^{\mathrm{flat}}|\sum_{t<T}\lambda_t^2$
provided that the static signs on $S\cap\mathcal J^{\mathrm{flat}}$ have zero conditional mean given $D,S$ and that trajectory's past, and are almost surely nonzero. Neither condition follows from the dynamic-gate hypothesis. Without the respective nonzero-sign condition, each formula retains the factor $\sgn(g_{t,i})^2$ inside its expected per-step squared increment.
\end{proposition}

\begin{proof}
The squared displacement expands into within-time and cross-time terms. The martingale-difference assumption makes all cross-time terms vanish. The squared increment of coordinate $i$ at iteration $t$ is $\lambda_t^2\1\{i\in S_t\}\sgn(g_{t,i})^2$. On the stated coordinates the selected sign is nonzero almost surely, so the conditional second moment is $\lambda_t^2\pi_{t,i}^{\mathrm{sel}}$. Summation over coordinates and iterations and expectation over the trajectory conditional on $D$ yield the dynamic formula. The dense and static calculations use their own zero-conditional-mean assumptions to eliminate cross-time terms. Their deterministic gates and respective nonzero-sign assumptions then give the stated formulas.
\end{proof}

This proposition is a local mechanism rather than a generalization theorem. It shows why suppressing many uninformative signs can preserve a pretrained large model over a short adaptation horizon.

\subsection{Conditional advantage relative to Adam's confidence interpretation}
\suppanchor{confidence-adam}

For comparison of hard and continuous confidence factors, the coordinate contribution to the expected smoothness-model decrease at attenuation $\varpi\in[0,1]$ is
\begin{equation}
\mathcal G_{t,i}^{\mathrm{loc}}(\varpi)
=
\lambda_t|h_{t,i}|\chi_{t,i}^{\mathrm{dense}}\varpi
-\frac{\lambda_t^2}{2}L_i\varpi^2.
\label{eq:cg_local_gain}
\end{equation}
The ideal mean-square confidence rule uses $\varpi=\chi_{t,i}^{\mathrm{dense}}$. A selected hard gate uses $\varpi=1$, and an omitted coordinate uses $\varpi=0$.

\begin{proposition}[Hard gate versus ideal continuous confidence]
\label{prop:cg_hard_continuous}
The dense sign reliability is assumed to satisfy $\chi_{t,i}^{\mathrm{dense}}\in[0,1]$. Then
\begin{align}
\mathcal G_{t,i}^{\mathrm{loc}}(1)
-\mathcal G_{t,i}^{\mathrm{loc}}(\chi_{t,i}^{\mathrm{dense}})
={}&
\lambda_t(1-\chi_{t,i}^{\mathrm{dense}})
\left[
|h_{t,i}|\chi_{t,i}^{\mathrm{dense}}
-\frac{\lambda_tL_i}{2}(1+\chi_{t,i}^{\mathrm{dense}})
\right],
\label{eq:cg_selected_hard_advantage}\\
\mathcal G_{t,i}^{\mathrm{loc}}(0)
-\mathcal G_{t,i}^{\mathrm{loc}}(\chi_{t,i}^{\mathrm{dense}})
={}&
\lambda_t(\chi_{t,i}^{\mathrm{dense}})^2
\left[
\frac{\lambda_tL_i}{2}-|h_{t,i}|
\right].
\label{eq:cg_omitted_hard_advantage}
\end{align}
Consequently, a hard gate is no worse than the ideal continuous-confidence rule in the local model when every selected coordinate satisfies
\begin{equation}
|h_{t,i}|\chi_{t,i}^{\mathrm{dense}}
\ge
\frac{\lambda_tL_i}{2}(1+\chi_{t,i}^{\mathrm{dense}}),
\label{eq:cg_selected_polarized_condition}
\end{equation}
and every omitted coordinate satisfies
\begin{equation}
|h_{t,i}|\le\frac{\lambda_tL_i}{2}.
\label{eq:cg_omitted_polarized_condition}
\end{equation}
\end{proposition}

\begin{proof}
Substitution of $\varpi=1$, $\varpi=0$, and $\varpi=\chi_{t,i}^{\mathrm{dense}}$ into Eq.~\eqref{eq:cg_local_gain}, followed by subtraction and factorization, gives the two identities. The sign conditions in Eqs.~\eqref{eq:cg_selected_polarized_condition}--\eqref{eq:cg_omitted_polarized_condition} make the corresponding differences nonnegative.
\end{proof}

The conditions describe a polarized few-shot regime. A small set of salient coordinates has enough confidence-weighted signal to justify an unshrunk sign step. The remaining coordinates have so little local signal that even a continuously attenuated update is not worthwhile. This gives a conditional advantage over the ideal continuous confidence interpretation that motivates Adam. It does not imply universal dominance over the full Adam recursion, whose momentum sign, denominator offset, moment estimation, and nonstationarity are not represented by Eq.~\eqref{eq:cg_local_gain}.

A second distinction is statistical. Adam estimates continuous moment-dependent factors for every coordinate. A hard gate only needs to separate reliable from unreliable coordinates. The next stylized lemma isolates when this binary decision is easier over a short horizon.

\begin{lemma}[Polarized short-horizon confidence estimation]
\label{lem:cg_short_horizon}
For every block $r$, let $\mathcal J_{t,r}^{\mathrm{rel}}\subseteq\mathcal I_r$ satisfy $|\mathcal J_{t,r}^{\mathrm{rel}}|=k_r$, and let $\mathcal J_t^{\mathrm{rel}}=\bigcup_{r=1}^{R}\mathcal J_{t,r}^{\mathrm{rel}}$. Let $\epsilon_{\mathrm{pol}}\in[0,1/2)$ be fixed. The polarization conditions are
\begin{equation}
\chi_{t,i}^{\mathrm{dense}}\in[1-\epsilon_{\mathrm{pol}},1]
\quad(i\in\mathcal J_t^{\mathrm{rel}}),
\qquad
\chi_{t,i}^{\mathrm{dense}}\in[0,\epsilon_{\mathrm{pol}}]
\quad(i\notin\mathcal J_t^{\mathrm{rel}}).
\label{eq:cg_polarized_confidence}
\end{equation}
An idealized estimator is assumed to observe $T_{\mathrm{obs}}\ge1$ measurements
\begin{equation}
\mathsf Y_{j_{\mathrm{obs}},i}^{\mathrm{conf}}
\sim
\mathcal N\!\left(
\chi_{t,i}^{\mathrm{dense}},
\varsigma_{\mathrm{est}}^2
\right),
\qquad
j_{\mathrm{obs}}\in\{1,\ldots,T_{\mathrm{obs}}\},
\label{eq:cg_confidence_observations}
\end{equation}
with $\varsigma_{\mathrm{est}}>0$. For every coordinate, independence across $j_{\mathrm{obs}}$ is assumed; no independence across coordinates is required. Let
$\overline{\mathsf Y}_{i}^{\mathrm{conf}}
=T_{\mathrm{obs}}^{-1}\sum_{j_{\mathrm{obs}}=1}^{T_{\mathrm{obs}}}\mathsf Y_{j_{\mathrm{obs}},i}^{\mathrm{conf}}$.
The blockwise hard estimate is
\begin{equation}
\widehat S_{t,r}^{\mathrm{conf}}
=\Top_{k_r}(\overline{\mathsf Y}_{\mathcal I_r}^{\mathrm{conf}}),
\qquad
\widehat S_t^{\mathrm{conf}}
=\bigcup_{r=1}^{R}\widehat S_{t,r}^{\mathrm{conf}},
\qquad
\widehat m_t^{\mathrm{conf}}=m_{\widehat S_t^{\mathrm{conf}}}.
\label{eq:cg_blockwise_confidence_gate}
\end{equation}
Then
\begin{align}
\E\!\left[
\sum_{i=1}^{d}(\overline{\mathsf Y}_{i}^{\mathrm{conf}}-\chi_{t,i}^{\mathrm{dense}})^2
\right]
&=
\frac{d\varsigma_{\mathrm{est}}^2}{T_{\mathrm{obs}}},
\label{eq:cg_continuous_estimation_error}\\
\E\!\left[
\sum_{i=1}^{d}(\widehat m_{t,i}^{\mathrm{conf}}-\chi_{t,i}^{\mathrm{dense}})^2
\right]
&\le
 d\epsilon_{\mathrm{pol}}^2
+2d\exp\!\left(
-\frac{T_{\mathrm{obs}}(1/2-\epsilon_{\mathrm{pol}})^2}{2\varsigma_{\mathrm{est}}^2}
\right).
\label{eq:cg_binary_estimation_error}
\end{align}
The upper bound in Eq.~\eqref{eq:cg_binary_estimation_error} is smaller than the exact continuous-estimation error whenever
\begin{equation}
\epsilon_{\mathrm{pol}}^2
+
2\exp\!\left(
-\frac{T_{\mathrm{obs}}(1/2-\epsilon_{\mathrm{pol}})^2}{2\varsigma_{\mathrm{est}}^2}
\right)
<
\frac{\varsigma_{\mathrm{est}}^2}{T_{\mathrm{obs}}}.
\label{eq:cg_short_horizon_condition}
\end{equation}
\end{lemma}

\begin{proof}
Each sample mean $\overline{\mathsf Y}_{i}^{\mathrm{conf}}$ has variance $\varsigma_{\mathrm{est}}^2/T_{\mathrm{obs}}$, which proves Eq.~\eqref{eq:cg_continuous_estimation_error}. For a reliable coordinate, the event $\overline{\mathsf Y}_{i}^{\mathrm{conf}}<1/2$ requires a centered Gaussian variable with variance $\varsigma_{\mathrm{est}}^2/T_{\mathrm{obs}}$ to cross a margin of at least $1/2-\epsilon_{\mathrm{pol}}$. The same bound applies to the event $\overline{\mathsf Y}_{i}^{\mathrm{conf}}\ge1/2$ for an unreliable coordinate. Hence each threshold-crossing probability is at most
\begin{equation*}
\exp\!\left(
-\frac{T_{\mathrm{obs}}(1/2-\epsilon_{\mathrm{pol}})^2}{2\varsigma_{\mathrm{est}}^2}
\right).
\end{equation*}
Within a block, every reliable coordinate omitted by the top-$k_r$ rule is paired with an included unreliable coordinate. Such a pair requires at least one of its two coordinates to cross the threshold $1/2$. Therefore, the symmetric-difference size between $\widehat S_t^{\mathrm{conf}}$ and $\mathcal J_t^{\mathrm{rel}}$ is at most twice the number of threshold crossings. If a coordinate is classified on the correct side of the reliable-set partition, its squared error relative to $\chi_{t,i}^{\mathrm{dense}}$ is at most $\epsilon_{\mathrm{pol}}^2$. Otherwise, it is at most one. Summing these bounds and taking expectation proves Eq.~\eqref{eq:cg_binary_estimation_error}. Equation~\eqref{eq:cg_short_horizon_condition} compares the two bounds.
\end{proof}

Lemma~\ref{lem:cg_short_horizon} is not a lower bound for Adam or for optimal continuous confidence estimation. It compares the hard gate only with the unprojected sample mean specified in the lemma. Clipping to $[0,1]$ or adding structural priors can reduce continuous-estimation error. The Gaussian observation model is an analytically transparent abstraction that isolates the short-horizon mechanism: when confidence is blockwise polarized, recovering the reliable side of a margin can require fewer accurate observations than calibrating all $d$ continuous values. In large models, continuous estimation error accumulates over all coordinates. DynSignSparseSGD avoids persistent moment estimation and refreshes a blockwise hard gate from the current gradient. SignSparseSGD estimates the gate once, which lowers repeated selection cost but can introduce support staleness.

\subsection{Scope and failure modes}
\suppanchor{confidence-scope}

The confidence-gating interpretation is conditional. It identifies regimes rather than universal dominance.
\begin{itemize}
\item If useful high-confidence signal is dense and the coordinate step is small, most scores in Eq.~\eqref{eq:cg_oracle_score} are positive. Dense SignSGD can then be preferable because a sparse budget discards useful descent mass.
\item If noise scales vary strongly across coordinates, top-$|g_t|$ can select high-variance distractors. Confidence is ordered by signal-to-noise ratio in Eq.~\eqref{eq:cg_gaussian_reliability}, not by raw magnitude alone.
\item If the stochastic-gradient distribution is skewed enough that $p_{t,i}^{\mathrm{corr}}<1/2$, a positive sign factor points uphill on average. Appendix~\ref{app:convergence} gives such a counterexample.
\item If the horizon is long and the gradient statistics are sufficiently stationary, Adam can estimate useful continuous factors and can outperform a hard gate.
\item The static mask can become stale when the reliable set changes. The dynamic mask avoids permanent commitment but pays a new selection error at each iteration.
\end{itemize}

Dense SignSGD applies a fixed step to every coordinate. Adam estimates a continuous state for every coordinate. The proposed optimizers allocate a fixed-magnitude sign step only to a small blockwise support. This hard gate is advantageous when confidence is polarized, low-confidence coordinates are numerous, and the adaptation horizon is too short to justify dense continuous moment estimation.

\section{Magnitude Calibration and Blockwise Ranking Stability}
\label{app:magnitude_ranking}
\suppanchor{ranking}

Appendix~\ref{app:confidence_gating} analyzes the reliability of the sign after a coordinate has been selected. This appendix isolates the preceding selection question. Lemma~\ref{lem:cg_gaussian_ordering} states an exact rank-recovery condition inside a Gaussian ordering argument. We separate that condition from the Gaussian model, add a gap-free support-selection bound, and derive moment-based recovery probabilities. The results formalize why a stochastic-gradient magnitude can be poorly calibrated as a numerical step size while remaining useful for blockwise top-$k$ ranking. They match the per-parameter-tensor selection rule in Eq.~\eqref{eq:block_budgets}.

We remind the reader that $h_t=\nabla F(\Theta_t)$ as defined in Appendix~\ref{app:confidence_gating}. By Eq.~\eqref{eq:population_gradient_decomposition},
\begin{equation}
 g_t=h_t+e_D(\Theta_t)+\xi_t.
\label{eq:rank_gradient_decomposition}
\end{equation}
For every block, the population-ranking support is
\begin{equation}
S_{t,r}^{\mathrm{pop}}
=\Top_{k_r}(|h_{t,\mathcal I_r}|),
\qquad
S_t^{\mathrm{pop}}
=\bigcup_{r=1}^{R}S_{t,r}^{\mathrm{pop}}.
\label{eq:rank_population_support}
\end{equation}
The fixed tie convention from Sec.~\ref{sec:method} is used. Hence, 
$\|h_{t,S_t^{\mathrm{pop}}}\|_1=\|h_t\|_{[\mathbf{k}]}$. Although, the results are stated for the population gradient to match Appendix~\ref{app:convergence}, their deterministic parts hold with $h_t$ replaced by any fixed reference vector, including the higher-shot reference gradient used in Fig.~\ref{fig:gradient_diagnostics}.
The missed-mass functional $\mathcal E_{\mathrm{miss},t}(S)$ was defined in Eq.~\eqref{eq:static_support_missed_mass}. Evaluated at the dynamic support, it is the captured-population-gradient-mass regret
\begin{equation}
\mathcal E_{\mathrm{miss},t}(S_t)
=\|h_t\|_{[\mathbf{k}]}-\|h_{t,S_t}\|_1.
\label{eq:rank_mass_regret}
\end{equation}

\subsection{Gap-free support-selection regret}
\suppanchor{ranking-regret}

\begin{lemma}[Gap-free blockwise support-selection regret]
\label{lem:rank_gap_free}
For every iteration $t$ and every realization of the stochastic gradient,
\begin{align}
0\le \mathcal E_{\mathrm{miss},t}(S_t)
\le{}&
\|\big(e_D(\Theta_t)+\xi_t\big)_{S_t}\|_1
+
\|\big(e_D(\Theta_t)+\xi_t\big)_{S_t^{\mathrm{pop}}}\|_1
\label{eq:rank_gap_free_restricted}\\
\le{}&
2\|e_D(\Theta_t)+\xi_t\|_{[\mathbf{k}]}
\label{eq:rank_gap_free_topk}\\
\le{}&
2\|e_D(\Theta_t)\|_{[\mathbf{k}]}
+2\|\xi_t\|_{[\mathbf{k}]}.
\label{eq:rank_gap_free_decomposed}
\end{align}
Under Assumption~\ref{assumption_1}(2),
\begin{equation}
\E\!\left[
\mathcal E_{\mathrm{miss},t}(S_t)
\mid D,\Theta_t
\right]
\le
2\|e_D(\Theta_t)\|_{[\mathbf{k}]}
+
\frac{2\|\nu\|_1}{b_t^\gamma}.
\label{eq:rank_gap_free_expectation}
\end{equation}
The same statements apply to the initial support of SignSparseSGD by taking $t=0$ and replacing $S_t$ by the static support $S$.
\end{lemma}

\begin{proof}
The support $S_t$ maximizes the noisy $\ell_1$ mass separately in every block. Summing the blockwise optimality inequalities gives
\begin{equation}
\|g_{t,S_t}\|_1
\ge
\|g_{t,S_t^{\mathrm{pop}}}\|_1.
\label{eq:rank_noisy_support_optimality}
\end{equation}
The triangle inequality bounds the left-hand side of Eq.~\eqref{eq:rank_noisy_support_optimality}, while the reverse triangle inequality bounds its right-hand side. Equation~\eqref{eq:rank_gradient_decomposition} then yields
\begin{align*}
\|h_{t,S_t}\|_1
+\|\big(e_D(\Theta_t)+\xi_t\big)_{S_t}\|_1
\ge{}&
\|h_{t,S_t^{\mathrm{pop}}}\|_1
-\|\big(e_D(\Theta_t)+\xi_t\big)_{S_t^{\mathrm{pop}}}\|_1.
\end{align*}
Rearranging proves Eq.~\eqref{eq:rank_gap_free_restricted}. Nonnegativity follows because $S_t^{\mathrm{pop}}$ maximizes the population-gradient $\ell_1$ mass under the same blockwise budgets. Both $S_t$ and $S_t^{\mathrm{pop}}$ contain exactly $k_r$ coordinates from block $r$. Therefore, each restricted error norm is at most the corresponding blockwise top-$k$ norm. This proves Eq.~\eqref{eq:rank_gap_free_topk}. The triangle inequality for $\|\cdot\|_{[\mathbf{k}]}$ proves Eq.~\eqref{eq:rank_gap_free_decomposed}. Finally,
\begin{equation*}
\E[\|\xi_t\|_{[\mathbf{k}]}\mid D,\Theta_t]
\le
\E[\|\xi_t\|_1\mid D,\Theta_t]
\le
\frac{\|\nu\|_1}{b_t^\gamma},
\end{equation*}
which proves Eq.~\eqref{eq:rank_gap_free_expectation}.
\end{proof}

Lemma~\ref{lem:rank_gap_free} does not require a separation between adjacent ordered magnitudes. It controls the quantity that enters sparse descent: the population-gradient mass captured by the selected support. Exact support overlap can be a stricter and less informative diagnostic. For example, if many coordinates at the blockwise selection boundary have equal population-gradient magnitude, two supports can have little overlap while capturing the same mass. The regret in Eq.~\eqref{eq:rank_mass_regret} is then zero.

In the single-block special case with no finite-shot bias, Eq.~\eqref{eq:rank_gap_free_restricted} becomes
\begin{equation}
\|h_t\|_{[k]}-\|h_{t,S_t}\|_1
\le
\|\xi_{t,S_t}\|_1
+\|\xi_{t,S_t^{\mathrm{pop}}}\|_1
\le
2\|\xi_t\|_{[k]}.
\label{eq:rank_single_block_regret}
\end{equation}
Thus, the selection rule can be useful even when it does not recover every oracle coordinate exactly.

\subsection{Gap-dependent exact support recovery}
\suppanchor{ranking-recovery}

For every block satisfying $k_r<d_r$, the population rank-boundary gap is
\begin{equation}
\operatorname{gap}_{t,r}^{\mathrm{rank}}
=
|h_{t,\mathcal I_r}|_{(k_r)}
-
|h_{t,\mathcal I_r}|_{(k_r+1)}.
\label{eq:rank_boundary_gap}
\end{equation}
No boundary gap is required when $k_r=d_r$, because the whole block is selected.

\begin{lemma}[Gap-dependent exact blockwise support recovery]
\label{lem:rank_exact_recovery}
The rank-boundary gaps are assumed to satisfy $\operatorname{gap}_{t,r}^{\mathrm{rank}}>0$ for every block with $k_r<d_r$. If
\begin{equation}
2\|\big(e_D(\Theta_t)+\xi_t\big)_{\mathcal I_r}\|_\infty
<
\operatorname{gap}_{t,r}^{\mathrm{rank}}
\qquad
\text{for every }r\text{ with }k_r<d_r,
\label{eq:rank_exact_recovery_condition}
\end{equation}
then
\begin{equation}
S_t=S_t^{\mathrm{pop}}.
\label{eq:rank_exact_recovery_conclusion}
\end{equation}
\end{lemma}

\begin{proof}
Let $r$ be a fixed block with $k_r<d_r$. For every $i\in S_{t,r}^{\mathrm{pop}}$ and $i'\in\mathcal I_r\setminus S_{t,r}^{\mathrm{pop}}$,
\begin{equation}
|h_{t,i}|-|h_{t,i'}|
\ge
\operatorname{gap}_{t,r}^{\mathrm{rank}}.
\end{equation}
Using Eq.~\eqref{eq:rank_gradient_decomposition}, the reverse triangle inequality for coordinate $i$, and the triangle inequality for coordinate $i'$ gives
\begin{align}
|g_{t,i}|-|g_{t,i'}|
\ge{}&
|h_{t,i}|-|h_{t,i'}|
-2\|\big(e_D(\Theta_t)+\xi_t\big)_{\mathcal I_r}\|_\infty
\nonumber\\
>{}&0.
\label{eq:rank_pairwise_separation}
\end{align}
Every population top-$k_r$ coordinate has strictly larger noisy magnitude than every coordinate outside that set. Hence, $S_{t,r}=S_{t,r}^{\mathrm{pop}}$. The conclusion follows after taking the union over blocks. Blocks with $k_r=d_r$ agree automatically.
\end{proof}

The strict inequality in Eq.~\eqref{eq:rank_exact_recovery_condition} is necessary for a tie-independent guarantee. Equality can create a noisy tie at the selection boundary. The fixed tie convention may then select a different support. The condition is sufficient rather than necessary. A common positive rescaling of all coordinates preserves every ranking. By choosing the scale sufficiently far from one, the numerical magnitudes can be arbitrarily miscalibrated while Eq.~\eqref{eq:rank_exact_recovery_condition} fails.

\begin{corollary}[High-probability blockwise ranking recovery]
\label{cor:rank_probability}
The positive-gap conditions of Lemma~\ref{lem:rank_exact_recovery} are assumed to hold. Let $p_{\mathrm{mom}}>0$ be a fixed moment order. The nonnegative conditional moment budgets $\mathfrak M_{t,i}^{\mathrm{mom}}$ are assumed to satisfy
\begin{equation}
\E\!\left[
|e_D(\Theta_t)_i+\xi_{t,i}|^{p_{\mathrm{mom}}}
\mid D,\Theta_t
\right]
\le
\mathfrak M_{t,i}^{\mathrm{mom}}
\qquad
(i=1,\ldots,d).
\label{eq:rank_moment_condition}
\end{equation}
Then
\begin{align}
&\Prob\!\left(
S_t\ne S_t^{\mathrm{pop}}
\mid D,\Theta_t
\right)
\nonumber\\
&\quad\le
\min\!\left\{
1,
\sum_{\substack{1\le r\le R\\ k_r<d_r}}
\left(
\frac{2}{\operatorname{gap}_{t,r}^{\mathrm{rank}}}
\right)^{p_{\mathrm{mom}}}
\sum_{i\in\mathcal I_r}
\mathfrak M_{t,i}^{\mathrm{mom}}
\right\}.
\label{eq:rank_moment_probability}
\end{align}
No independence across coordinates is required. Under Assumption~\ref{assumption_1}(2), the first-moment specialization is
\begin{align}
&\Prob\!\left(
S_t\ne S_t^{\mathrm{pop}}
\mid D,\Theta_t
\right)
\nonumber\\
&\quad\le
\min\!\left\{
1,
\sum_{\substack{1\le r\le R\\ k_r<d_r}}
\frac{2}{\operatorname{gap}_{t,r}^{\mathrm{rank}}}
\left(
\|e_D(\Theta_t)_{\mathcal I_r}\|_1
+
\frac{\|\nu_{\mathcal I_r}\|_1}{b_t^\gamma}
\right)
\right\}.
\label{eq:rank_first_moment_probability}
\end{align}
\end{corollary}

\begin{proof}
If $S_t\ne S_t^{\mathrm{pop}}$, then Lemma~\ref{lem:rank_exact_recovery} implies that at least one block with $k_r<d_r$ satisfies
\begin{equation*}
\|\big(e_D(\Theta_t)+\xi_t\big)_{\mathcal I_r}\|_\infty
\ge
\frac12\operatorname{gap}_{t,r}^{\mathrm{rank}}.
\end{equation*}
A union bound over blocks and coordinates, conditional Markov's inequality, and Eq.~\eqref{eq:rank_moment_condition} give Eq.~\eqref{eq:rank_moment_probability}. For the first-moment specialization,
\begin{equation*}
\E[|e_D(\Theta_t)_i+\xi_{t,i}|\mid D,\Theta_t]
\le
|e_D(\Theta_t)_i|
+
\frac{\nu_i}{b_t^\gamma}.
\end{equation*}
Substitution with $p_{\mathrm{mom}}=1$ proves Eq.~\eqref{eq:rank_first_moment_probability}.
\end{proof}

For one block, zero finite-shot bias, and coordinatewise bounds
$\E[|\xi_{t,i}|^{p_{\mathrm{mom}}}\mid D,\Theta_t]
\le\mathfrak M_{t,i}^{\mathrm{mom}}$, Eq.~\eqref{eq:rank_moment_probability} reduces to
\begin{equation}
\Prob(S_t\ne S_t^{\mathrm{pop}}\mid D,\Theta_t)
\le
\min\!\left\{
1,
\frac{2^{p_{\mathrm{mom}}}}{
(\operatorname{gap}_{t,1}^{\mathrm{rank}})^{p_{\mathrm{mom}}}}
\sum_{i=1}^{d}\mathfrak M_{t,i}^{\mathrm{mom}}
\right\}.
\label{eq:rank_single_block_probability}
\end{equation}
This is the direct moment analogue of the deterministic order-statistic margin condition.

\subsection{Interpretation and limitations}
\suppanchor{ranking-interpretation}

Magnitude calibration and ranking stability are different properties. Calibration asks whether each observed magnitude is numerically close to its population counterpart. Ranking asks whether the perturbation is large enough to cross a blockwise order-statistic boundary. A large common scaling error can leave every support unchanged. Conversely, when a boundary gap approaches zero, a perturbation whose infinity norm also approaches zero can reverse the selected order.

Lemma~\ref{lem:rank_gap_free} and Lemma~\ref{lem:rank_exact_recovery} answer different questions. The gap-free result remains meaningful when exact recovery is impossible. It controls the descent-relevant mass lost by selection. The gap-dependent result provides exact support recovery when the top tail is separated. The probability bounds make this dependence explicit, but they can be loose in high dimension because they use coordinatewise Markov and union bounds. They should be interpreted as sufficient certificates rather than calibrated predictions of empirical overlap.

Finally, these results concern only where the optimizer spends its sparse update budget. They do not imply that every selected sign is correct. Appendix~\ref{app:confidence_gating} controls selected-sign reliability and its contribution to expected descent. Together, the two appendices formalize the design separation: noisy magnitudes are used for saliency ranking, while fixed-magnitude signs determine the coordinate steps.

\section{Few-Shot Noise Characterization}
\label{app:fewshot_noise}
\suppanchor{noise}

We complement Fig.~\ref{fig:gradient_diagnostics} by extending the same analysis from EuroSAT to Oxford-Pets. We retain the diagnostic notation from Sec.~\ref{sec:introduction}: $K$ is the number of shots per class, $d$ is the number of trainable coordinates, $\rho\in(0,1]$ is the nominal diagnostic density, and $k_{\mathrm{diag}}=\lceil\rho d\rceil$ is the global diagnostic active-set size. Its realized density is $k_{\mathrm{diag}}/d$. As in the introduction, we compare a noisy 1-shot gradient $g_t^{(K=1)}$ with a 16-shot reference gradient $g_t^{(K=16)}$ at each iteration $t$, on the same class subset, for all trainable parameters of CLIP (ViT-B/16). Panels (a)--(d) have the same interpretation as in Fig.~\ref{fig:gradient_diagnostics}. Panel (a) plots the conditional histogram of $\log_{10}(|g_t^{(K=1)}|/|g_t^{(K=16)}|)$ given sign agreement or sign flip. Panel (b) reports the magnitude-weighted sign reliability:
\[
\frac{\left\langle g_t^{(K=16)}, \sgn\!\left(g_t^{(K=1)}\right)\right\rangle}{\|g_t^{(K=16)}\|_1}.
\]
Panel (c) plots sign agreement in the top-$k_{\mathrm{diag}}$ coordinates, with ranking by $|g_t^{(K=16)}|$ (oracle) or by $|g_t^{(K=1)}|$ (realizable). Panel (d) plots the overlap between these two top-$k_{\mathrm{diag}}$ supports. These results show that the phenomenon in Fig.~\ref{fig:gradient_diagnostics} is not specific to EuroSAT. They also show that the strength of the signal depends on the dataset and on the encoder branch.

\begin{figure}[!h]
    \centering
    \begin{subfigure}[t]{0.485\linewidth}
        \centering
        \includegraphics[width=\linewidth]{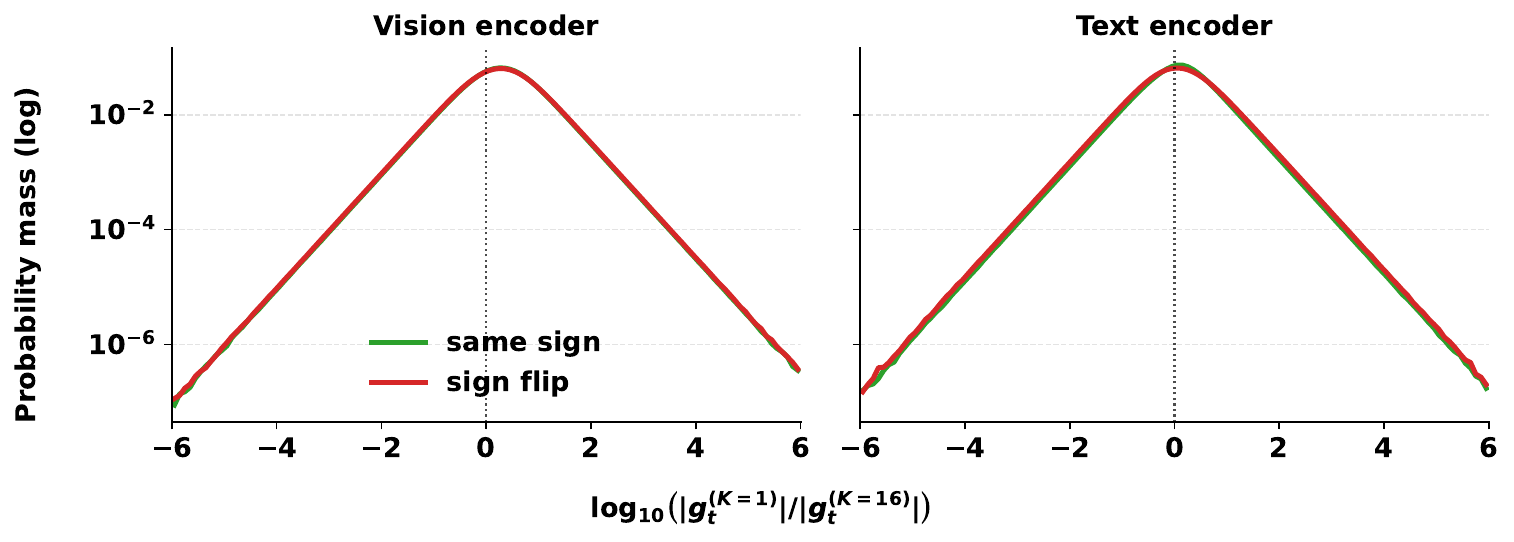}
        \caption{Noisy magnitudes.}
        \label{fig:fewshot_grad_noise_pets:a}
    \end{subfigure}\hfill
    \begin{subfigure}[t]{0.485\linewidth}
        \centering
        \includegraphics[width=\linewidth]{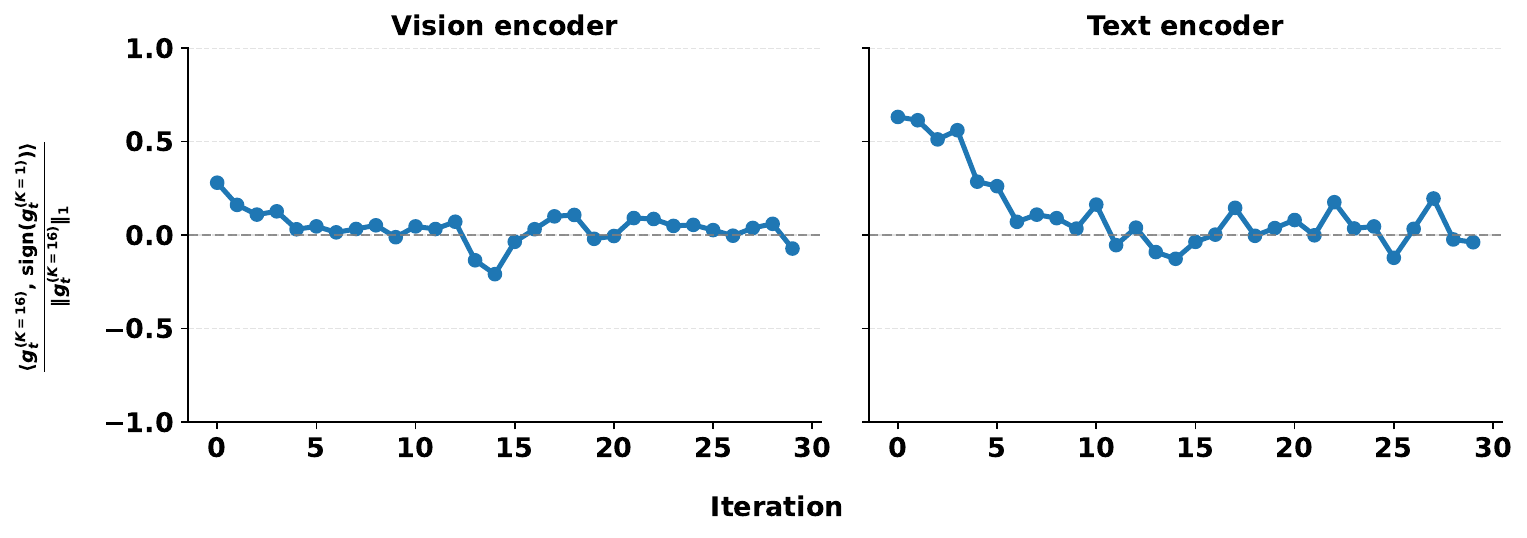}
        \caption{Descent-relevant sign.}
        \label{fig:fewshot_grad_noise_pets:b}
    \end{subfigure}

    \begin{subfigure}[t]{0.485\linewidth}
        \centering
        \includegraphics[width=\linewidth]{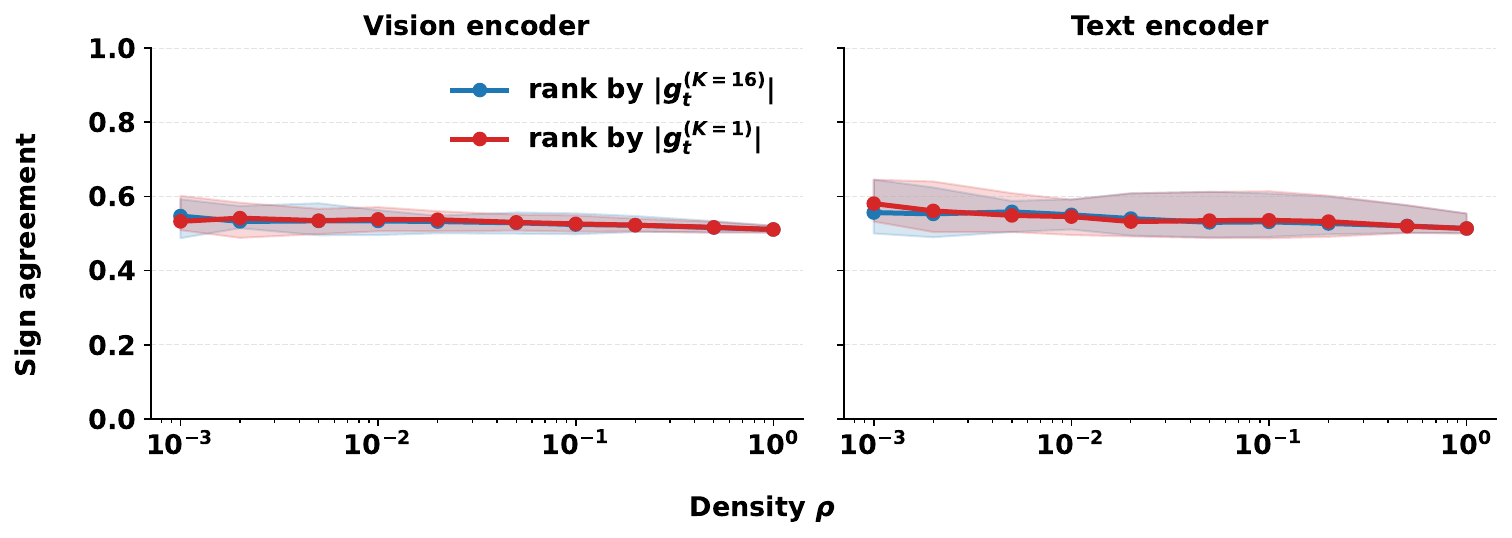}
        \caption{Sign concentrates in top-$|g|$.}
        \label{fig:fewshot_grad_noise_pets:c}
    \end{subfigure}\hfill
    \begin{subfigure}[t]{0.485\linewidth}
        \centering
        \includegraphics[width=\linewidth]{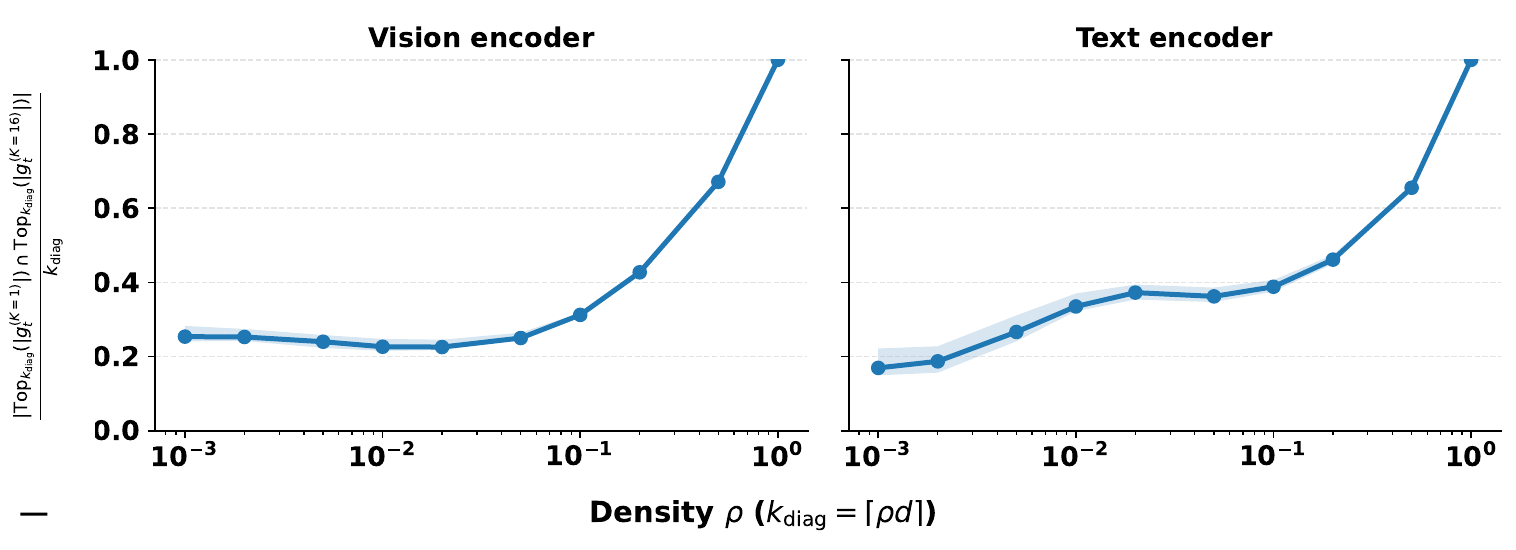}
        \caption{Noisy magnitudes as saliency.}
        \label{fig:fewshot_grad_noise_pets:d}
    \end{subfigure}
    \caption{Few-shot gradient noise in CLIP adaptation (ViT-B/16, Oxford-Pets, $K{=}1$ vs $K{=}16$; $K$ is the \# of shots, $k_{\mathrm{diag}}$ is the diagnostic active-set size, and $d$ is the total \# of trainable coordinates).}
    \label{fig:fewshot_grad_noise_pets}
\end{figure}

The same qualitative pattern as EuroSAT is present on Oxford-Pets. In Fig.~\ref{fig:fewshot_grad_noise_pets}(a), the same-sign and sign-flip ratio histograms overlap strongly in both encoders. Thus, even when the 1-shot and 16-shot gradients point in the same direction, the magnitude is often badly mis-scaled. In Fig.~\ref{fig:fewshot_grad_noise_pets}(b), the weighted sign correlation is clearly positive at the beginning of training, but it decays faster than on EuroSAT and becomes nearly zero near the end. This indicates that the sign still carries a useful descent signal, but mainly in the early phase of adaptation. Fig.~\ref{fig:fewshot_grad_noise_pets}(c) shows that this signal becomes clearer after restricting attention to salient coordinates. The median sign agreement rises from $0.509$ to $0.547$ in the vision encoder and from $0.531$ to $0.597$ in the text encoder when moving from the full gradient to the top $0.1\%$ coordinates ranked by $|g_t^{(K=1)}|$. Fig.~\ref{fig:fewshot_grad_noise_pets}(d) further shows that the noisy magnitude remains informative for support selection. At $\rho=10^{-3}$, the overlap between $\Top_{k_{\mathrm{diag}}}(|g_t^{(K=1)}|)$ and $\Top_{k_{\mathrm{diag}}}(|g_t^{(K=16)}|)$ is about $29.0\%$ in the vision encoder and $19.4\%$ in the text encoder. Therefore, on Oxford-Pets, the main issue is again the step scale, not the complete loss of directional information.

\begin{table}[!h]
\centering
\caption{Summary of the few-shot noise diagnostics for EuroSAT (Fig.~\ref{fig:gradient_diagnostics}) and Oxford-Pets (Fig.~\ref{fig:fewshot_grad_noise_pets}). Values are medians over iterations. The column ``Sign ($\rho=10^{-3}$)'' uses the realizable ranking by $|g_t^{(K=1)}|$.}
\label{tab:fewshot_noise_summary}
\footnotesize
\setlength{\tabcolsep}{3.5pt}
\begin{tabular}{@{}llcccc@{}}
\toprule
Dataset & Encoder &
\shortstack[c]{Sign\\($\rho=1$)} &
\shortstack[c]{Sign\\($\rho=10^{-3}$)} &
\shortstack[c]{Overlap\\($\rho=10^{-3}$)} &
\shortstack[c]{Weighted\\sign corr.} \\
\midrule
EuroSAT     & Vision & 0.516 & 0.529 & 0.283 & 0.034 \\
EuroSAT     & Text   & 0.548 & 0.694 & 0.258 & 0.167 \\
Oxford-Pets & Vision & 0.509 & 0.547 & 0.290 & 0.015 \\
Oxford-Pets & Text   & 0.531 & 0.597 & 0.194 & 0.097 \\
\bottomrule
\end{tabular}
\end{table}

Table~\ref{tab:fewshot_noise_summary} highlights the common pattern across EuroSAT and Oxford-Pets. In both datasets, the full-gradient sign agreement stays close to chance, which confirms that the raw 1-shot gradient is globally noisy. Yet the overlap between the noisy and reference top-$k_{\mathrm{diag}}$ supports remains between $19\%$ and $29\%$ at the extreme density $\rho=10^{-3}$. The magnitude is therefore unreliable as a step-size signal, but it remains informative as a saliency signal. This is exactly the role it plays in SignSparseSGD and DynSignSparseSGD. EuroSAT is the cleaner case, while Oxford-Pets shows the same behavior with weaker margins and a faster decay of the weighted sign correlation. The conclusion is the same as in Fig.~\ref{fig:gradient_diagnostics}: in the few-shot regime, it is more reliable to use the gradient magnitude to decide \emph{where} to update and to use a fixed-magnitude sign step to decide \emph{how} to update.

\section{SignSparseSGD \& DynSignSparseSGD vs PEFT/OET}
\label{app:peft_oet}
\suppanchor{peft}

This Appendix extends the main comparison with two additional PEFT baselines: BitFit~\cite{ben-zaken2022bitfit} and PiSSA~\cite{meng2024pissa}. BitFit tunes only the bias parameters, whereas PiSSA initializes the low-rank factors from the principal singular vectors and values of the pretrained weights to facilitate optimization. 

%This section covers two settings. In \emph{all-to-all} (A2A), we sample $K\!\in\!\{1,2,4,8,16\}$ labeled images per class from the full label set to form the support set and report top-1 accuracy on the corresponding test split. In \emph{base-to-new} (B2N), we generate one fixed class partition per dataset--seed pair by shuffling class labels and assigning half of the classes to \emph{Base} and the other half to \emph{New} ($50{:}50$). Each partition is saved and reused by all methods and backbones. During training, only \emph{Base} classes construct the support set. Evaluation is performed separately on \emph{Base} and \emph{New} classes, and HM denotes the harmonic mean of the two accuracies. All experiments use three fixed random seeds with identical support and query sets across methods.

\subsection{All-to-All Few-Shot Evaluation}
\label{app:a2a_additional}
\suppanchor{peft-a2a}

\begin{table}
\caption{Few-shot classification performance on 11 datasets with CLIP(ViT-B/32) backbone. Top-1 accuracy (\%) averaged over 3 random seeds.}
\label{tab:few_shot_ViTB32_with_aug}
\centering
\footnotesize
\setlength{\tabcolsep}{1.8pt}
\renewcommand{\arraystretch}{1.00}
\begin{adjustbox}{max totalsize={\textwidth}{0.94\textheight},center}
\begin{tabular}{@{}c l *{12}{d}@{}}
\toprule
Shots & Method & ImageNet & SUN & Aircr. & EuroSAT & Cars & Food & Pets & Flow. & Caltech & DTD & UCF & Avg. \\ \midrule
\multirow{1}{*}{0} & CLIP & 62.0 & 62.1 & 18.1 & 35.8 & 60.1 & 78.9 & 85.1 & 64.0 & 91.4 & 43.0 & 61.6 & 60.2 \\
\midrule
\multirow{15}{*}{1} & BitFit & 64.1 $\pm$ 0.3 & 66.8 $\pm$ 0.1 & 22.0 $\pm$ 0.4 & 63.0 $\pm$ 2.9 & 62.2 $\pm$ 0.3 & 80.3 $\pm$ 0.4 & 88.9 $\pm$ 0.2 & 70.2 $\pm$ 0.6 & 93.1 $\pm$ 0.2 & 49.9 $\pm$ 0.1 & 70.2 $\pm$ 0.1 & 66.4 $\pm$ 0.5 \\
& LayerNorm & 63.9 $\pm$ 0.1 & 64.3 $\pm$ 0.1 & 19.6 $\pm$ 0.0 & 40.3 $\pm$ 0.5 & 61.3 $\pm$ 0.2 & 79.8 $\pm$ 0.1 & 86.7 $\pm$ 0.1 & 65.8 $\pm$ 0.0 & 91.9 $\pm$ 0.1 & 44.4 $\pm$ 0.3 & 63.4 $\pm$ 0.1 & 62.0 $\pm$ 0.1 \\
& SAFT & 59.7 $\pm$ 0.7 & 64.4 $\pm$ 0.2 & 21.0 $\pm$ 1.0 & 66.2 $\pm$ 1.8 & 57.5 $\pm$ 1.0 & 78.0 $\pm$ 0.6 & 85.3 $\pm$ 1.1 & 73.3 $\pm$ 0.7 & 93.3 $\pm$ 0.7 & 47.6 $\pm$ 0.9 & 69.7 $\pm$ 0.6 & 65.1 $\pm$ 0.8 \\
\cmidrule(lr){2-14}
& CoOp & 62.4 $\pm$ 0.5 & 64.1 $\pm$ 0.2 & 21.4 $\pm$ 1.1 & 49.2 $\pm$ 4.4 & 60.4 $\pm$ 0.8 & 79.0 $\pm$ 1.2 & 85.9 $\pm$ 0.3 & 70.8 $\pm$ 0.9 & 91.3 $\pm$ 0.1 & 45.2 $\pm$ 1.4 & 68.0 $\pm$ 1.7 & 63.4 $\pm$ 1.1 \\
& KgCoOp & 63.5 $\pm$ 0.2 & 65.4 $\pm$ 0.4 & 20.6 $\pm$ 0.6 & 51.9 $\pm$ 0.9 & 60.4 $\pm$ 0.2 & 80.5 $\pm$ 0.0 & 88.5 $\pm$ 0.5 & 65.5 $\pm$ 0.6 & 92.2 $\pm$ 0.3 & 47.4 $\pm$ 1.8 & 66.6 $\pm$ 0.2 & 63.9 $\pm$ 0.5 \\
& ProGrad & 54.2 $\pm$ 4.9 & 61.9 $\pm$ 2.0 & 16.8 $\pm$ 1.8 & 50.7 $\pm$ 2.6 & 59.1 $\pm$ 1.7 & 77.0 $\pm$ 1.6 & 82.8 $\pm$ 4.5 & 69.0 $\pm$ 1.0 & 90.3 $\pm$ 0.8 & 45.1 $\pm$ 0.8 & 66.8 $\pm$ 1.5 & 61.2 $\pm$ 2.1 \\
\cmidrule(lr){2-14}
% & Adapter & 62.8 $\pm$ 0.1 & 63.0 $\pm$ 0.0 & 18.5 $\pm$ 0.2 & 39.0 $\pm$ 1.1 & 60.7 $\pm$ 0.1 & 79.3 $\pm$ 0.1 & 86.1 $\pm$ 0.1 & 64.7 $\pm$ 0.3 & 91.6 $\pm$ 0.0 & 43.7 $\pm$ 0.2 & 62.6 $\pm$ 0.3 & 61.1 $\pm$ 0.2 \\
% & LP++ & 60.9 $\pm$ 0.2 & 63.9 $\pm$ 0.6 & 20.7 $\pm$ 0.5 & 46.6 $\pm$ 1.2 & 54.9 $\pm$ 0.6 & 75.6 $\pm$ 0.6 & 83.3 $\pm$ 1.5 & 76.3 $\pm$ 1.1 & 91.0 $\pm$ 0.7 & 47.4 $\pm$ 1.4 & 66.6 $\pm$ 0.2 & 62.5 $\pm$ 0.8 \\
& TIP-Adapter-F & 62.1 $\pm$ 0.0 & 62.1 $\pm$ 0.0 & 18.2 $\pm$ 0.1 & 36.5 $\pm$ 0.4 & 60.2 $\pm$ 0.0 & 78.9 $\pm$ 0.0 & 85.1 $\pm$ 0.0 & 64.3 $\pm$ 0.0 & 91.4 $\pm$ 0.1 & 43.0 $\pm$ 0.0 & 61.7 $\pm$ 0.0 & 60.3 $\pm$ 0.1 \\
& GDA & 62.4 $\pm$ 0.0 & 63.8 $\pm$ 0.1 & 19.8 $\pm$ 0.3 & 47.6 $\pm$ 7.0 & 61.3 $\pm$ 0.8 & 78.5 $\pm$ 0.3 & 85.1 $\pm$ 0.1 & 72.1 $\pm$ 1.5 & 92.3 $\pm$ 0.1 & 44.7 $\pm$ 1.8 & 64.2 $\pm$ 0.4 & 62.9 $\pm$ 1.1 \\
& ProKeR & 62.7 $\pm$ 0.1 & 64.8 $\pm$ 0.1 & 21.3 $\pm$ 0.6 & 53.2 $\pm$ 2.9 & 62.6 $\pm$ 0.3 & 78.9 $\pm$ 0.2 & 85.0 $\pm$ 0.4 & 74.5 $\pm$ 1.4 & 92.4 $\pm$ 0.2 & 48.0 $\pm$ 0.3 & 66.1 $\pm$ 0.3 & 64.5 $\pm$ 0.6 \\
\cmidrule(lr){2-14}
& LoRA & 62.8 $\pm$ 0.6 & 67.4 $\pm$ 0.2 & 21.5 $\pm$ 0.2 & 58.2 $\pm$ 2.2 & 62.9 $\pm$ 0.2 & 80.8 $\pm$ 0.2 & 89.0 $\pm$ 0.3 & 68.2 $\pm$ 0.3 & 93.1 $\pm$ 0.1 & 49.7 $\pm$ 0.7 & 70.1 $\pm$ 0.2 & 65.8 $\pm$ 0.5 \\
& PiSSA & 60.7 $\pm$ 0.3 & 65.1 $\pm$ 0.2 & 22.2 $\pm$ 2.0 & 64.3 $\pm$ 6.5 & 62.1 $\pm$ 0.3 & 77.2 $\pm$ 0.7 & 86.4 $\pm$ 0.4 & 76.9 $\pm$ 1.5 & 92.9 $\pm$ 0.8 & 50.1 $\pm$ 1.2 & 70.6 $\pm$ 0.5 & 66.2 $\pm$ 1.3 \\
& GaLoRE & 55.9 $\pm$ 0.6 & 63.5 $\pm$ 0.3 & 20.2 $\pm$ 1.5 & 64.1 $\pm$ 8.1 & 54.4 $\pm$ 1.6 & 75.5 $\pm$ 0.7 & 85.2 $\pm$ 0.4 & 74.7 $\pm$ 0.9 & 91.1 $\pm$ 0.5 & 48.1 $\pm$ 2.1 & 70.5 $\pm$ 0.6 & 63.9 $\pm$ 1.6 \\
\cmidrule(lr){2-14}
& SO & 63.2 $\pm$ 0.5 & 67.4 $\pm$ 0.3 & 22.1 $\pm$ 0.2 & 62.5 $\pm$ 1.0 & 63.3 $\pm$ 0.2 & 80.5 $\pm$ 0.4 & 89.2 $\pm$ 0.5 & 70.8 $\pm$ 0.9 & 93.1 $\pm$ 0.3 & 51.0 $\pm$ 0.4 & 71.1 $\pm$ 0.2 & 66.7 $\pm$ 0.4 \\
& SignSparseSGD & 60.6 $\pm$ 0.6 & 65.3 $\pm$ 0.1 & 21.6 $\pm$ 0.5 & 66.0 $\pm$ 5.2 & 61.3 $\pm$ 1.1 & 77.9 $\pm$ 0.5 & 88.5 $\pm$ 0.4 & 75.0 $\pm$ 0.8 & 92.3 $\pm$ 0.5 & 50.9 $\pm$ 1.5 & 71.5 $\pm$ 0.3 & 66.4 $\pm$ 1.0 \\
& DynSignSparseSGD & 62.6 $\pm$ 0.3 & 67.0 $\pm$ 0.3 & 22.7 $\pm$ 0.8 & 69.1 $\pm$ 2.7 & 63.1 $\pm$ 0.4 & 79.7 $\pm$ 0.5 & 89.1 $\pm$ 0.2 & 77.7 $\pm$ 1.2 & 92.6 $\pm$ 0.7 & 51.9 $\pm$ 1.0 & 72.0 $\pm$ 0.4 & 67.9 $\pm$ 0.8 \\
\midrule
\multirow{15}{*}{2} & BitFit & 64.9 $\pm$ 0.2 & 68.2 $\pm$ 0.2 & 23.5 $\pm$ 0.4 & 68.6 $\pm$ 3.9 & 62.9 $\pm$ 0.2 & 80.6 $\pm$ 0.1 & 89.7 $\pm$ 0.6 & 74.0 $\pm$ 1.2 & 93.9 $\pm$ 0.3 & 53.7 $\pm$ 0.7 & 72.5 $\pm$ 0.7 & 68.4 $\pm$ 0.8 \\
& LayerNorm & 64.4 $\pm$ 0.0 & 65.4 $\pm$ 0.1 & 20.3 $\pm$ 0.2 & 41.7 $\pm$ 0.4 & 61.4 $\pm$ 0.1 & 80.5 $\pm$ 0.1 & 86.7 $\pm$ 0.1 & 66.6 $\pm$ 0.0 & 92.3 $\pm$ 0.1 & 46.4 $\pm$ 0.0 & 64.7 $\pm$ 0.1 & 62.8 $\pm$ 0.1 \\
& SAFT & 61.3 $\pm$ 0.1 & 65.2 $\pm$ 0.4 & 23.2 $\pm$ 0.6 & 78.2 $\pm$ 2.5 & 58.8 $\pm$ 0.0 & 78.8 $\pm$ 0.2 & 89.2 $\pm$ 0.2 & 78.6 $\pm$ 1.1 & 93.9 $\pm$ 0.3 & 53.8 $\pm$ 1.4 & 71.3 $\pm$ 0.2 & 68.4 $\pm$ 0.6 \\
\cmidrule(lr){2-14}
& CoOp & 63.4 $\pm$ 0.2 & 66.5 $\pm$ 0.1 & 22.3 $\pm$ 0.3 & 63.4 $\pm$ 3.4 & 61.6 $\pm$ 0.1 & 78.6 $\pm$ 1.4 & 88.8 $\pm$ 0.8 & 76.0 $\pm$ 0.9 & 93.3 $\pm$ 0.3 & 49.6 $\pm$ 1.4 & 70.8 $\pm$ 1.0 & 66.7 $\pm$ 0.9 \\
& KgCoOp & 63.8 $\pm$ 0.2 & 66.0 $\pm$ 0.3 & 20.9 $\pm$ 0.4 & 51.7 $\pm$ 1.0 & 60.8 $\pm$ 0.3 & 80.7 $\pm$ 0.2 & 88.3 $\pm$ 0.3 & 66.5 $\pm$ 0.9 & 92.7 $\pm$ 0.1 & 49.3 $\pm$ 1.0 & 68.1 $\pm$ 0.8 & 64.4 $\pm$ 0.5 \\
& ProGrad & 61.6 $\pm$ 1.2 & 65.2 $\pm$ 2.7 & 19.4 $\pm$ 3.3 & 61.9 $\pm$ 5.1 & 59.3 $\pm$ 2.2 & 74.7 $\pm$ 2.9 & 87.4 $\pm$ 1.2 & 75.0 $\pm$ 1.5 & 92.7 $\pm$ 0.6 & 48.1 $\pm$ 2.3 & 70.4 $\pm$ 0.7 & 65.1 $\pm$ 2.1 \\
\cmidrule(lr){2-14}
% & Adapter & 63.4 $\pm$ 0.0 & 64.3 $\pm$ 0.1 & 19.1 $\pm$ 0.1 & 39.5 $\pm$ 1.4 & 61.0 $\pm$ 0.2 & 79.8 $\pm$ 0.2 & 86.3 $\pm$ 0.1 & 65.2 $\pm$ 0.4 & 91.9 $\pm$ 0.1 & 44.7 $\pm$ 0.2 & 63.5 $\pm$ 0.1 & 61.7 $\pm$ 0.3 \\
% & LP++ & 63.1 $\pm$ 0.0 & 66.8 $\pm$ 0.2 & 22.2 $\pm$ 0.6 & 53.1 $\pm$ 2.2 & 58.1 $\pm$ 0.5 & 76.6 $\pm$ 0.7 & 84.8 $\pm$ 0.3 & 83.5 $\pm$ 1.0 & 92.6 $\pm$ 0.2 & 54.5 $\pm$ 0.2 & 69.9 $\pm$ 0.7 & 65.9 $\pm$ 0.6 \\
& TIP-Adapter-F & 62.1 $\pm$ 0.0 & 62.1 $\pm$ 0.0 & 18.2 $\pm$ 0.1 & 37.6 $\pm$ 0.2 & 60.3 $\pm$ 0.0 & 78.9 $\pm$ 0.0 & 85.1 $\pm$ 0.0 & 64.6 $\pm$ 0.1 & 91.4 $\pm$ 0.0 & 43.0 $\pm$ 0.1 & 61.8 $\pm$ 0.1 & 60.5 $\pm$ 0.1 \\
& GDA & 62.9 $\pm$ 0.1 & 66.1 $\pm$ 0.0 & 21.6 $\pm$ 0.4 & 68.5 $\pm$ 4.1 & 63.6 $\pm$ 0.1 & 78.9 $\pm$ 0.1 & 84.2 $\pm$ 1.3 & 84.8 $\pm$ 2.0 & 92.5 $\pm$ 1.0 & 51.7 $\pm$ 1.5 & 69.2 $\pm$ 0.7 & 67.6 $\pm$ 1.0 \\
& ProKeR & 63.4 $\pm$ 0.0 & 66.6 $\pm$ 0.1 & 22.7 $\pm$ 0.1 & 63.7 $\pm$ 1.6 & 64.8 $\pm$ 0.2 & 79.2 $\pm$ 0.1 & 85.5 $\pm$ 0.3 & 77.5 $\pm$ 0.7 & 93.6 $\pm$ 0.4 & 51.9 $\pm$ 1.1 & 68.2 $\pm$ 0.5 & 67.0 $\pm$ 0.5 \\
\cmidrule(lr){2-14}
& LoRA & 62.0 $\pm$ 0.2 & 67.8 $\pm$ 0.5 & 23.8 $\pm$ 0.4 & 65.9 $\pm$ 2.0 & 63.8 $\pm$ 0.6 & 80.4 $\pm$ 0.2 & 89.3 $\pm$ 0.9 & 75.7 $\pm$ 1.2 & 93.9 $\pm$ 0.5 & 55.1 $\pm$ 0.5 & 73.5 $\pm$ 0.8 & 68.3 $\pm$ 0.7 \\
& PiSSA & 61.1 $\pm$ 0.1 & 66.8 $\pm$ 0.3 & 24.9 $\pm$ 0.8 & 80.5 $\pm$ 1.2 & 63.7 $\pm$ 0.2 & 77.1 $\pm$ 0.5 & 88.1 $\pm$ 1.0 & 83.0 $\pm$ 1.3 & 94.0 $\pm$ 0.1 & 54.8 $\pm$ 1.0 & 74.0 $\pm$ 1.5 & 69.8 $\pm$ 0.7 \\
& GaLoRE & 57.8 $\pm$ 0.2 & 64.9 $\pm$ 0.2 & 22.6 $\pm$ 1.3 & 76.7 $\pm$ 0.8 & 60.6 $\pm$ 0.9 & 75.1 $\pm$ 0.5 & 85.8 $\pm$ 0.9 & 80.5 $\pm$ 0.7 & 92.3 $\pm$ 0.4 & 53.9 $\pm$ 0.8 & 73.7 $\pm$ 0.6 & 67.6 $\pm$ 0.7 \\
\cmidrule(lr){2-14}
& SO & 62.6 $\pm$ 0.2 & 67.6 $\pm$ 0.6 & 24.6 $\pm$ 0.3 & 69.3 $\pm$ 2.1 & 64.2 $\pm$ 0.4 & 80.1 $\pm$ 0.5 & 89.8 $\pm$ 1.0 & 78.7 $\pm$ 1.3 & 93.9 $\pm$ 0.7 & 55.0 $\pm$ 1.0 & 74.6 $\pm$ 0.6 & 69.1 $\pm$ 0.8 \\
& SignSparseSGD & 61.9 $\pm$ 0.3 & 66.9 $\pm$ 0.2 & 24.3 $\pm$ 0.6 & 79.1 $\pm$ 6.7 & 63.6 $\pm$ 0.1 & 78.0 $\pm$ 0.4 & 89.1 $\pm$ 0.4 & 78.7 $\pm$ 0.2 & 93.8 $\pm$ 0.3 & 56.4 $\pm$ 0.4 & 74.1 $\pm$ 0.8 & 69.6 $\pm$ 0.9 \\
& DynSignSparseSGD & 63.2 $\pm$ 0.1 & 68.2 $\pm$ 0.3 & 25.5 $\pm$ 0.6 & 79.8 $\pm$ 4.2 & 66.0 $\pm$ 0.1 & 79.5 $\pm$ 0.4 & 89.2 $\pm$ 0.3 & 83.8 $\pm$ 1.4 & 94.0 $\pm$ 0.3 & 57.1 $\pm$ 1.1 & 74.8 $\pm$ 0.6 & 71.0 $\pm$ 0.9 \\
\midrule
\multirow{15}{*}{4} & BitFit & 65.4 $\pm$ 0.1 & 69.9 $\pm$ 0.3 & 25.0 $\pm$ 0.1 & 72.6 $\pm$ 0.7 & 65.0 $\pm$ 0.4 & 80.9 $\pm$ 0.3 & 90.9 $\pm$ 0.1 & 78.6 $\pm$ 0.1 & 94.5 $\pm$ 0.2 & 57.1 $\pm$ 0.5 & 74.7 $\pm$ 0.1 & 70.4 $\pm$ 0.3 \\
& LayerNorm & 65.0 $\pm$ 0.0 & 66.7 $\pm$ 0.1 & 20.6 $\pm$ 0.1 & 42.4 $\pm$ 0.4 & 62.4 $\pm$ 0.2 & 80.9 $\pm$ 0.1 & 87.1 $\pm$ 0.1 & 67.2 $\pm$ 0.1 & 92.7 $\pm$ 0.1 & 47.7 $\pm$ 0.3 & 66.2 $\pm$ 0.1 & 63.5 $\pm$ 0.1 \\
& SAFT & 62.6 $\pm$ 0.4 & 66.0 $\pm$ 0.3 & 24.6 $\pm$ 0.5 & 83.1 $\pm$ 1.6 & 61.9 $\pm$ 0.4 & 79.4 $\pm$ 0.4 & 90.4 $\pm$ 0.1 & 83.4 $\pm$ 1.0 & 94.7 $\pm$ 0.2 & 56.7 $\pm$ 0.3 & 73.5 $\pm$ 0.3 & 70.6 $\pm$ 0.5 \\
\cmidrule(lr){2-14}
& CoOp & 64.5 $\pm$ 0.2 & 68.6 $\pm$ 0.2 & 23.4 $\pm$ 0.8 & 67.7 $\pm$ 1.2 & 64.2 $\pm$ 0.7 & 79.7 $\pm$ 0.4 & 89.7 $\pm$ 0.3 & 81.6 $\pm$ 1.3 & 93.7 $\pm$ 0.4 & 55.8 $\pm$ 0.3 & 74.1 $\pm$ 0.9 & 69.4 $\pm$ 0.6 \\
& KgCoOp & 64.2 $\pm$ 0.1 & 66.6 $\pm$ 0.3 & 21.1 $\pm$ 0.7 & 52.7 $\pm$ 2.6 & 61.1 $\pm$ 0.1 & 80.9 $\pm$ 0.1 & 88.8 $\pm$ 0.6 & 67.4 $\pm$ 0.7 & 92.9 $\pm$ 0.2 & 51.0 $\pm$ 0.4 & 68.9 $\pm$ 0.4 & 65.0 $\pm$ 0.6 \\
& ProGrad & 62.3 $\pm$ 1.4 & 68.9 $\pm$ 0.1 & 22.1 $\pm$ 1.0 & 67.1 $\pm$ 2.9 & 61.5 $\pm$ 1.3 & 79.1 $\pm$ 0.8 & 89.7 $\pm$ 0.6 & 77.0 $\pm$ 5.7 & 92.6 $\pm$ 0.7 & 54.3 $\pm$ 0.6 & 72.7 $\pm$ 0.5 & 67.9 $\pm$ 1.4 \\
\cmidrule(lr){2-14}
% & Adapter & 63.9 $\pm$ 0.1 & 65.8 $\pm$ 0.1 & 19.8 $\pm$ 0.1 & 40.6 $\pm$ 1.2 & 61.6 $\pm$ 0.0 & 80.2 $\pm$ 0.1 & 86.5 $\pm$ 0.2 & 66.0 $\pm$ 0.5 & 92.0 $\pm$ 0.1 & 46.7 $\pm$ 0.4 & 66.8 $\pm$ 0.1 & 62.7 $\pm$ 0.3 \\
% & LP++ & 65.2 $\pm$ 0.0 & 69.2 $\pm$ 0.4 & 24.6 $\pm$ 0.5 & 63.0 $\pm$ 2.5 & 62.6 $\pm$ 0.3 & 77.6 $\pm$ 0.2 & 85.8 $\pm$ 1.6 & 88.9 $\pm$ 0.4 & 93.2 $\pm$ 0.1 & 59.2 $\pm$ 1.0 & 72.0 $\pm$ 0.7 & 69.2 $\pm$ 0.7 \\
& TIP-Adapter-F & 62.2 $\pm$ 0.0 & 62.3 $\pm$ 0.0 & 18.6 $\pm$ 0.1 & 38.7 $\pm$ 0.3 & 60.4 $\pm$ 0.0 & 78.9 $\pm$ 0.0 & 85.2 $\pm$ 0.0 & 65.3 $\pm$ 0.1 & 91.5 $\pm$ 0.1 & 43.0 $\pm$ 0.0 & 61.9 $\pm$ 0.0 & 60.7 $\pm$ 0.1 \\
& GDA & 64.7 $\pm$ 0.1 & 70.1 $\pm$ 0.2 & 27.5 $\pm$ 0.8 & 72.5 $\pm$ 6.9 & 67.6 $\pm$ 0.3 & 79.5 $\pm$ 0.4 & 84.0 $\pm$ 3.4 & 91.2 $\pm$ 0.7 & 93.9 $\pm$ 0.3 & 58.0 $\pm$ 0.2 & 73.6 $\pm$ 1.4 & 71.1 $\pm$ 1.3 \\
& ProKeR & 64.7 $\pm$ 0.1 & 68.3 $\pm$ 0.2 & 25.0 $\pm$ 0.5 & 67.7 $\pm$ 2.8 & 67.7 $\pm$ 0.1 & 80.1 $\pm$ 0.1 & 86.0 $\pm$ 0.3 & 80.9 $\pm$ 0.2 & 94.2 $\pm$ 0.1 & 55.3 $\pm$ 0.2 & 70.7 $\pm$ 0.1 & 69.2 $\pm$ 0.4 \\
\cmidrule(lr){2-14}
& LoRA & 61.8 $\pm$ 0.5 & 69.2 $\pm$ 0.1 & 25.7 $\pm$ 0.4 & 65.9 $\pm$ 0.7 & 67.8 $\pm$ 0.9 & 80.5 $\pm$ 0.1 & 91.0 $\pm$ 0.2 & 84.9 $\pm$ 0.2 & 94.7 $\pm$ 0.2 & 59.9 $\pm$ 0.7 & 77.1 $\pm$ 0.2 & 70.8 $\pm$ 0.4 \\
& PiSSA & 61.9 $\pm$ 0.3 & 68.2 $\pm$ 0.3 & 28.0 $\pm$ 0.9 & 82.3 $\pm$ 0.7 & 67.9 $\pm$ 0.3 & 77.8 $\pm$ 0.0 & 89.0 $\pm$ 0.8 & 88.6 $\pm$ 1.0 & 94.4 $\pm$ 0.6 & 60.4 $\pm$ 0.4 & 76.6 $\pm$ 1.2 & 72.3 $\pm$ 0.6 \\
& GaLoRE & 58.7 $\pm$ 0.4 & 66.8 $\pm$ 0.4 & 26.5 $\pm$ 1.8 & 81.5 $\pm$ 0.4 & 64.8 $\pm$ 0.5 & 75.8 $\pm$ 0.9 & 87.9 $\pm$ 0.9 & 86.9 $\pm$ 0.1 & 93.9 $\pm$ 0.2 & 59.3 $\pm$ 1.2 & 76.1 $\pm$ 1.4 & 70.7 $\pm$ 0.7 \\
\cmidrule(lr){2-14}
& SO & 62.4 $\pm$ 0.3 & 69.0 $\pm$ 0.1 & 26.5 $\pm$ 0.1 & 70.7 $\pm$ 0.8 & 68.3 $\pm$ 0.8 & 80.2 $\pm$ 0.1 & 90.7 $\pm$ 0.2 & 86.7 $\pm$ 0.6 & 94.7 $\pm$ 0.3 & 60.1 $\pm$ 0.6 & 77.2 $\pm$ 0.4 & 71.5 $\pm$ 0.4 \\
& SignSparseSGD & 63.3 $\pm$ 0.4 & 69.0 $\pm$ 0.4 & 25.2 $\pm$ 1.1 & 82.6 $\pm$ 1.0 & 66.9 $\pm$ 0.3 & 78.2 $\pm$ 0.4 & 89.3 $\pm$ 0.4 & 85.8 $\pm$ 1.2 & 94.7 $\pm$ 0.4 & 59.1 $\pm$ 1.0 & 75.9 $\pm$ 0.7 & 71.8 $\pm$ 0.7 \\
& DynSignSparseSGD & 64.0 $\pm$ 0.2 & 69.9 $\pm$ 0.3 & 27.7 $\pm$ 0.6 & 83.9 $\pm$ 1.6 & 69.3 $\pm$ 0.1 & 79.6 $\pm$ 0.4 & 90.0 $\pm$ 0.4 & 88.9 $\pm$ 0.8 & 95.2 $\pm$ 0.2 & 61.1 $\pm$ 0.3 & 78.0 $\pm$ 0.9 & 73.4 $\pm$ 0.5 \\
\midrule
\multirow{15}{*}{8} & BitFit & 66.3 $\pm$ 0.1 & 71.2 $\pm$ 0.1 & 28.8 $\pm$ 0.1 & 81.1 $\pm$ 1.4 & 68.2 $\pm$ 0.1 & 81.0 $\pm$ 0.2 & 91.2 $\pm$ 0.2 & 84.8 $\pm$ 0.2 & 95.0 $\pm$ 0.1 & 61.2 $\pm$ 1.0 & 77.2 $\pm$ 0.8 & 73.3 $\pm$ 0.4 \\
& LayerNorm & 65.6 $\pm$ 0.0 & 68.1 $\pm$ 0.1 & 21.5 $\pm$ 0.1 & 47.7 $\pm$ 0.8 & 63.3 $\pm$ 0.3 & 81.8 $\pm$ 0.1 & 87.5 $\pm$ 0.2 & 69.2 $\pm$ 0.1 & 93.4 $\pm$ 0.1 & 51.0 $\pm$ 0.4 & 68.5 $\pm$ 0.1 & 65.2 $\pm$ 0.2 \\
& SAFT & 64.2 $\pm$ 0.2 & 67.7 $\pm$ 0.3 & 28.9 $\pm$ 0.2 & 88.0 $\pm$ 2.0 & 65.3 $\pm$ 0.4 & 79.4 $\pm$ 0.4 & 90.5 $\pm$ 0.3 & 87.4 $\pm$ 0.5 & 94.8 $\pm$ 0.5 & 60.9 $\pm$ 1.1 & 75.7 $\pm$ 0.5 & 73.0 $\pm$ 0.6 \\
\cmidrule(lr){2-14}
& CoOp & 65.7 $\pm$ 0.2 & 70.7 $\pm$ 0.0 & 27.1 $\pm$ 0.3 & 73.5 $\pm$ 1.9 & 66.8 $\pm$ 0.2 & 80.5 $\pm$ 0.4 & 90.5 $\pm$ 0.5 & 87.0 $\pm$ 1.1 & 94.3 $\pm$ 0.2 & 60.1 $\pm$ 1.2 & 77.0 $\pm$ 0.5 & 72.1 $\pm$ 0.6 \\
& KgCoOp & 64.5 $\pm$ 0.1 & 67.2 $\pm$ 0.1 & 22.2 $\pm$ 0.5 & 57.3 $\pm$ 3.0 & 61.6 $\pm$ 0.1 & 81.2 $\pm$ 0.1 & 89.3 $\pm$ 0.6 & 68.4 $\pm$ 0.7 & 93.1 $\pm$ 0.2 & 53.4 $\pm$ 1.5 & 70.2 $\pm$ 0.5 & 66.2 $\pm$ 0.7 \\
& ProGrad & 63.2 $\pm$ 1.8 & 70.8 $\pm$ 0.5 & 23.7 $\pm$ 1.5 & 73.6 $\pm$ 0.6 & 64.8 $\pm$ 3.0 & 80.7 $\pm$ 0.3 & 90.1 $\pm$ 0.7 & 84.3 $\pm$ 2.5 & 91.9 $\pm$ 2.5 & 59.1 $\pm$ 0.3 & 75.8 $\pm$ 1.2 & 70.7 $\pm$ 1.3 \\
\cmidrule(lr){2-14}
% & Adapter & 64.5 $\pm$ 0.1 & 67.5 $\pm$ 0.0 & 21.2 $\pm$ 0.3 & 44.5 $\pm$ 0.8 & 63.2 $\pm$ 0.3 & 81.2 $\pm$ 0.1 & 86.9 $\pm$ 0.2 & 69.2 $\pm$ 0.6 & 92.7 $\pm$ 0.1 & 50.3 $\pm$ 0.5 & 69.9 $\pm$ 0.4 & 64.6 $\pm$ 0.3 \\
% & LP++ & 66.2 $\pm$ 0.2 & 71.6 $\pm$ 0.1 & 27.7 $\pm$ 0.8 & 72.5 $\pm$ 2.7 & 68.7 $\pm$ 0.6 & 78.7 $\pm$ 0.4 & 87.2 $\pm$ 0.8 & 90.9 $\pm$ 0.4 & 94.0 $\pm$ 0.4 & 62.8 $\pm$ 0.6 & 76.4 $\pm$ 0.5 & 72.4 $\pm$ 0.7 \\
& TIP-Adapter-F & 62.5 $\pm$ 0.0 & 62.9 $\pm$ 0.1 & 19.2 $\pm$ 0.1 & 40.4 $\pm$ 0.4 & 60.9 $\pm$ 0.1 & 78.9 $\pm$ 0.0 & 85.4 $\pm$ 0.0 & 68.7 $\pm$ 0.2 & 91.7 $\pm$ 0.1 & 43.5 $\pm$ 0.1 & 62.1 $\pm$ 0.1 & 61.5 $\pm$ 0.1 \\
& GDA & 65.9 $\pm$ 0.1 & 71.8 $\pm$ 0.2 & 33.1 $\pm$ 0.7 & 83.4 $\pm$ 1.1 & 71.6 $\pm$ 0.5 & 80.1 $\pm$ 0.8 & 87.9 $\pm$ 0.2 & 93.8 $\pm$ 0.2 & 94.3 $\pm$ 0.2 & 63.3 $\pm$ 1.2 & 77.0 $\pm$ 1.2 & 74.7 $\pm$ 0.6 \\
& ProKeR & 65.1 $\pm$ 0.1 & 71.0 $\pm$ 0.2 & 28.8 $\pm$ 0.8 & 79.7 $\pm$ 1.8 & 72.5 $\pm$ 0.2 & 80.8 $\pm$ 0.1 & 87.0 $\pm$ 0.3 & 85.0 $\pm$ 0.1 & 94.7 $\pm$ 0.1 & 60.8 $\pm$ 1.1 & 75.0 $\pm$ 0.2 & 72.8 $\pm$ 0.5 \\
\cmidrule(lr){2-14}
& LoRA & 63.4 $\pm$ 0.1 & 70.6 $\pm$ 0.5 & 32.5 $\pm$ 1.4 & 81.0 $\pm$ 0.5 & 73.9 $\pm$ 0.8 & 79.9 $\pm$ 0.4 & 91.6 $\pm$ 0.5 & 91.1 $\pm$ 0.4 & 95.2 $\pm$ 0.3 & 63.3 $\pm$ 1.2 & 79.1 $\pm$ 0.4 & 74.7 $\pm$ 0.6 \\
& PiSSA & 63.3 $\pm$ 0.2 & 69.8 $\pm$ 0.4 & 34.8 $\pm$ 0.8 & 87.5 $\pm$ 2.2 & 74.1 $\pm$ 0.6 & 78.2 $\pm$ 0.2 & 90.0 $\pm$ 0.4 & 91.5 $\pm$ 0.5 & 94.9 $\pm$ 0.3 & 63.6 $\pm$ 1.2 & 78.5 $\pm$ 1.4 & 75.1 $\pm$ 0.7 \\
& GaLoRE & 60.3 $\pm$ 0.4 & 69.4 $\pm$ 0.2 & 31.8 $\pm$ 1.4 & 87.9 $\pm$ 1.1 & 71.8 $\pm$ 0.3 & 75.8 $\pm$ 0.3 & 89.8 $\pm$ 0.5 & 90.9 $\pm$ 0.8 & 94.3 $\pm$ 0.2 & 62.6 $\pm$ 0.7 & 79.1 $\pm$ 0.7 & 74.0 $\pm$ 0.6 \\
\cmidrule(lr){2-14}
& SO & 63.7 $\pm$ 0.1 & 70.6 $\pm$ 0.5 & 33.4 $\pm$ 1.1 & 85.0 $\pm$ 0.5 & 74.0 $\pm$ 0.8 & 79.6 $\pm$ 0.3 & 91.6 $\pm$ 0.5 & 91.6 $\pm$ 0.4 & 95.2 $\pm$ 0.2 & 63.2 $\pm$ 1.1 & 79.2 $\pm$ 0.3 & 75.2 $\pm$ 0.5 \\
& SignSparseSGD & 64.7 $\pm$ 0.1 & 70.9 $\pm$ 0.3 & 31.1 $\pm$ 0.8 & 85.9 $\pm$ 0.5 & 71.4 $\pm$ 0.5 & 79.0 $\pm$ 0.2 & 90.3 $\pm$ 0.1 & 89.3 $\pm$ 0.7 & 94.9 $\pm$ 0.3 & 62.3 $\pm$ 0.9 & 78.8 $\pm$ 0.7 & 74.4 $\pm$ 0.5 \\
& DynSignSparseSGD & 65.0 $\pm$ 0.2 & 71.9 $\pm$ 0.1 & 34.8 $\pm$ 0.6 & 88.4 $\pm$ 1.0 & 75.1 $\pm$ 0.2 & 79.8 $\pm$ 0.1 & 91.0 $\pm$ 0.3 & 92.5 $\pm$ 0.1 & 95.5 $\pm$ 0.3 & 64.5 $\pm$ 1.0 & 79.7 $\pm$ 0.6 & 76.2 $\pm$ 0.4 \\
\midrule
\multirow{15}{*}{16} & BitFit & 67.4 $\pm$ 0.1 & 73.0 $\pm$ 0.1 & 31.9 $\pm$ 0.3 & 87.6 $\pm$ 0.4 & 71.6 $\pm$ 0.3 & 81.9 $\pm$ 0.1 & 91.9 $\pm$ 0.1 & 90.2 $\pm$ 0.4 & 95.9 $\pm$ 0.1 & 66.4 $\pm$ 0.4 & 79.5 $\pm$ 0.4 & 76.1 $\pm$ 0.3 \\
& LayerNorm & 66.4 $\pm$ 0.1 & 69.8 $\pm$ 0.0 & 23.4 $\pm$ 0.1 & 55.9 $\pm$ 0.8 & 65.0 $\pm$ 0.1 & 82.2 $\pm$ 0.1 & 89.7 $\pm$ 0.1 & 73.0 $\pm$ 0.3 & 94.2 $\pm$ 0.1 & 55.2 $\pm$ 0.6 & 71.6 $\pm$ 0.2 & 67.9 $\pm$ 0.2 \\
& SAFT & 66.3 $\pm$ 0.2 & 69.2 $\pm$ 0.5 & 35.4 $\pm$ 0.3 & 92.3 $\pm$ 0.7 & 69.9 $\pm$ 0.2 & 79.7 $\pm$ 0.3 & 90.6 $\pm$ 0.1 & 91.2 $\pm$ 0.3 & 95.3 $\pm$ 0.1 & 65.3 $\pm$ 0.5 & 78.2 $\pm$ 0.4 & 75.8 $\pm$ 0.3 \\
\cmidrule(lr){2-14}
& CoOp & 66.7 $\pm$ 0.1 & 72.6 $\pm$ 0.1 & 28.6 $\pm$ 0.8 & 78.3 $\pm$ 2.2 & 71.2 $\pm$ 0.9 & 80.9 $\pm$ 0.0 & 90.1 $\pm$ 0.1 & 91.4 $\pm$ 0.5 & 94.7 $\pm$ 0.1 & 64.6 $\pm$ 0.3 & 78.9 $\pm$ 0.5 & 74.4 $\pm$ 0.5 \\
& KgCoOp & 64.8 $\pm$ 0.0 & 67.9 $\pm$ 0.0 & 22.7 $\pm$ 0.3 & 58.3 $\pm$ 1.0 & 62.3 $\pm$ 0.2 & 81.3 $\pm$ 0.2 & 89.6 $\pm$ 0.3 & 70.2 $\pm$ 0.9 & 93.5 $\pm$ 0.1 & 56.5 $\pm$ 1.1 & 71.2 $\pm$ 0.1 & 67.1 $\pm$ 0.4 \\
& ProGrad & 56.3 $\pm$ 4.2 & 72.5 $\pm$ 0.1 & 28.0 $\pm$ 1.0 & 80.6 $\pm$ 0.6 & 70.6 $\pm$ 0.2 & 80.9 $\pm$ 0.3 & 90.9 $\pm$ 0.3 & 88.8 $\pm$ 1.0 & 94.7 $\pm$ 0.3 & 63.3 $\pm$ 0.8 & 78.6 $\pm$ 0.3 & 73.2 $\pm$ 0.8 \\
\cmidrule(lr){2-14}
% & Adapter & 65.4 $\pm$ 0.1 & 69.5 $\pm$ 0.1 & 22.9 $\pm$ 0.4 & 49.9 $\pm$ 0.6 & 65.3 $\pm$ 0.1 & 81.5 $\pm$ 0.1 & 88.8 $\pm$ 0.2 & 74.5 $\pm$ 0.3 & 94.2 $\pm$ 0.1 & 56.4 $\pm$ 0.5 & 73.7 $\pm$ 0.2 & 67.5 $\pm$ 0.2 \\
% & LP++ & 67.2 $\pm$ 0.1 & 73.3 $\pm$ 0.1 & 31.3 $\pm$ 0.7 & 80.0 $\pm$ 2.4 & 72.7 $\pm$ 0.4 & 80.6 $\pm$ 0.1 & 88.3 $\pm$ 0.6 & 93.3 $\pm$ 0.7 & 94.9 $\pm$ 0.3 & 68.1 $\pm$ 0.3 & 78.9 $\pm$ 0.5 & 75.3 $\pm$ 0.6 \\
& TIP-Adapter-F & 63.7 $\pm$ 0.0 & 65.9 $\pm$ 0.1 & 21.4 $\pm$ 0.3 & 49.9 $\pm$ 0.7 & 64.3 $\pm$ 0.1 & 79.2 $\pm$ 0.1 & 85.8 $\pm$ 0.1 & 82.4 $\pm$ 0.2 & 92.0 $\pm$ 0.1 & 45.0 $\pm$ 0.3 & 64.5 $\pm$ 0.1 & 64.9 $\pm$ 0.2 \\
& GDA & 66.4 $\pm$ 0.1 & 73.4 $\pm$ 0.1 & 38.7 $\pm$ 0.5 & 87.0 $\pm$ 1.3 & 76.0 $\pm$ 0.2 & 80.6 $\pm$ 0.1 & 87.9 $\pm$ 0.5 & 96.3 $\pm$ 0.5 & 94.7 $\pm$ 0.1 & 68.6 $\pm$ 0.6 & 79.0 $\pm$ 0.5 & 77.1 $\pm$ 0.4 \\
& ProKeR & 66.9 $\pm$ 0.1 & 73.8 $\pm$ 0.2 & 35.3 $\pm$ 0.2 & 85.4 $\pm$ 1.5 & 77.9 $\pm$ 0.6 & 81.6 $\pm$ 0.1 & 88.5 $\pm$ 0.6 & 91.0 $\pm$ 0.2 & 95.5 $\pm$ 0.2 & 67.5 $\pm$ 0.2 & 78.7 $\pm$ 0.3 & 76.5 $\pm$ 0.4 \\
\cmidrule(lr){2-14}
& LoRA & 65.8 $\pm$ 0.3 & 72.6 $\pm$ 0.3 & 40.8 $\pm$ 0.6 & 90.4 $\pm$ 0.5 & 79.2 $\pm$ 0.1 & 80.1 $\pm$ 0.1 & 91.7 $\pm$ 0.1 & 94.8 $\pm$ 0.7 & 95.9 $\pm$ 0.3 & 67.9 $\pm$ 0.6 & 82.5 $\pm$ 0.5 & 78.4 $\pm$ 0.4 \\
& PiSSA & 65.5 $\pm$ 0.1 & 72.0 $\pm$ 0.2 & 40.7 $\pm$ 0.5 & 90.3 $\pm$ 0.5 & 79.0 $\pm$ 0.5 & 79.0 $\pm$ 0.3 & 89.9 $\pm$ 0.5 & 94.9 $\pm$ 0.7 & 95.5 $\pm$ 0.0 & 68.2 $\pm$ 0.0 & 81.4 $\pm$ 1.1 & 77.8 $\pm$ 0.4 \\
& GaLoRE & 63.0 $\pm$ 0.1 & 71.0 $\pm$ 0.1 & 40.7 $\pm$ 1.5 & 92.1 $\pm$ 0.6 & 76.6 $\pm$ 0.3 & 76.6 $\pm$ 0.6 & 89.5 $\pm$ 0.4 & 94.4 $\pm$ 0.5 & 95.0 $\pm$ 0.4 & 66.7 $\pm$ 1.0 & 80.6 $\pm$ 0.6 & 76.9 $\pm$ 0.6 \\
\cmidrule(lr){2-14}
& SO & 66.1 $\pm$ 0.4 & 72.6 $\pm$ 0.2 & 41.8 $\pm$ 0.6 & 90.6 $\pm$ 0.7 & 79.5 $\pm$ 0.2 & 80.0 $\pm$ 0.1 & 91.6 $\pm$ 0.1 & 94.8 $\pm$ 0.8 & 95.8 $\pm$ 0.2 & 68.4 $\pm$ 0.5 & 82.7 $\pm$ 0.3 & 78.5 $\pm$ 0.4 \\
& SignSparseSGD & 66.3 $\pm$ 0.2 & 73.1 $\pm$ 0.3 & 36.4 $\pm$ 0.9 & 90.7 $\pm$ 1.6 & 76.4 $\pm$ 0.6 & 79.3 $\pm$ 0.5 & 89.9 $\pm$ 1.0 & 93.8 $\pm$ 1.0 & 95.8 $\pm$ 0.0 & 67.5 $\pm$ 0.2 & 81.7 $\pm$ 0.4 & 77.4 $\pm$ 0.6 \\
& DynSignSparseSGD & 66.9 $\pm$ 0.2 & 73.5 $\pm$ 0.1 & 42.0 $\pm$ 1.1 & 92.0 $\pm$ 0.6 & 79.8 $\pm$ 0.4 & 80.3 $\pm$ 0.4 & 91.0 $\pm$ 0.3 & 95.7 $\pm$ 1.0 & 95.8 $\pm$ 0.0 & 69.2 $\pm$ 0.5 & 82.7 $\pm$ 0.5 & 79.0 $\pm$ 0.5 \\
\bottomrule
\end{tabular}
\end{adjustbox}
\end{table}
\begin{table}
\caption{Few-shot classification performance on 11 datasets with CLIP(ViT-B/16) backbone. Top-1 accuracy (\%) averaged over 3 random seeds.}
\label{tab:few_shot_ViTB16_with_aug}
\centering
\footnotesize
\setlength{\tabcolsep}{1.8pt}
\renewcommand{\arraystretch}{1.00}
\begin{adjustbox}{max totalsize={\textwidth}{0.94\textheight},center}
\begin{tabular}{@{}c l *{12}{d}@{}}
\toprule
Shots & Method & ImageNet & SUN & Aircr. & EuroSAT & Cars & Food & Pets & Flow. & Caltech & DTD & UCF & Avg. \\ \midrule
\multirow{1}{*}{0} & CLIP & 66.7 & 62.6 & 23.9 & 42.2 & 65.5 & 85.2 & 88.3 & 67.4 & 93.3 & 44.4 & 65.2 & 64.1 \\
\midrule
\multirow{15}{*}{1} & BitFit & 69.4 $\pm$ 0.2 & 69.3 $\pm$ 0.2 & 29.5 $\pm$ 0.3 & 72.0 $\pm$ 1.9 & 69.1 $\pm$ 0.4 & 85.5 $\pm$ 0.2 & 92.3 $\pm$ 0.3 & 74.8 $\pm$ 0.3 & 93.9 $\pm$ 0.2 & 52.1 $\pm$ 0.6 & 74.6 $\pm$ 0.4 & 71.1 $\pm$ 0.5 \\
& LayerNorm & 68.7 $\pm$ 0.1 & 66.7 $\pm$ 0.1 & 25.5 $\pm$ 0.2 & 49.6 $\pm$ 2.0 & 66.7 $\pm$ 0.2 & 85.7 $\pm$ 0.1 & 89.4 $\pm$ 0.1 & 69.9 $\pm$ 0.1 & 93.5 $\pm$ 0.1 & 45.9 $\pm$ 0.2 & 67.8 $\pm$ 0.2 & 66.3 $\pm$ 0.3 \\
& SAFT & 66.1 $\pm$ 0.1 & 67.2 $\pm$ 0.3 & 27.4 $\pm$ 0.8 & 68.9 $\pm$ 3.5 & 64.8 $\pm$ 0.7 & 84.1 $\pm$ 0.4 & 89.9 $\pm$ 0.9 & 80.7 $\pm$ 0.9 & 93.9 $\pm$ 0.2 & 52.2 $\pm$ 1.0 & 73.7 $\pm$ 0.4 & 69.9 $\pm$ 0.8 \\
\cmidrule(lr){2-14}
& CoOp & 67.3 $\pm$ 0.3 & 67.0 $\pm$ 0.6 & 28.1 $\pm$ 0.4 & 54.3 $\pm$ 6.9 & 66.7 $\pm$ 0.6 & 84.7 $\pm$ 0.5 & 89.3 $\pm$ 0.6 & 75.6 $\pm$ 1.6 & 93.1 $\pm$ 0.4 & 47.6 $\pm$ 1.6 & 72.3 $\pm$ 0.7 & 67.8 $\pm$ 1.3 \\
& KgCoOp & 68.6 $\pm$ 0.1 & 67.5 $\pm$ 0.4 & 25.7 $\pm$ 0.3 & 57.4 $\pm$ 2.7 & 65.0 $\pm$ 0.5 & 86.1 $\pm$ 0.1 & 90.9 $\pm$ 0.4 & 69.9 $\pm$ 0.4 & 93.4 $\pm$ 0.5 & 47.4 $\pm$ 1.0 & 71.0 $\pm$ 0.6 & 67.5 $\pm$ 0.6 \\
& ProGrad & 65.4 $\pm$ 1.8 & 65.7 $\pm$ 1.0 & 24.0 $\pm$ 2.6 & 56.1 $\pm$ 4.3 & 63.6 $\pm$ 4.0 & 83.2 $\pm$ 1.2 & 89.2 $\pm$ 1.9 & 75.3 $\pm$ 1.3 & 92.0 $\pm$ 1.3 & 48.0 $\pm$ 1.4 & 71.5 $\pm$ 0.7 & 66.7 $\pm$ 2.0 \\
\cmidrule(lr){2-14}
% & Adapter & 67.6 $\pm$ 0.1 & 64.3 $\pm$ 0.1 & 24.1 $\pm$ 0.2 & 45.6 $\pm$ 0.7 & 66.2 $\pm$ 0.2 & 85.4 $\pm$ 0.1 & 89.2 $\pm$ 0.1 & 68.6 $\pm$ 0.4 & 93.4 $\pm$ 0.1 & 45.0 $\pm$ 0.1 & 66.3 $\pm$ 0.2 & 65.1 $\pm$ 0.2 \\
% & LP++ & 66.3 $\pm$ 0.1 & 66.6 $\pm$ 0.2 & 26.7 $\pm$ 0.8 & 40.8 $\pm$ 2.4 & 62.6 $\pm$ 0.4 & 82.9 $\pm$ 1.3 & 86.1 $\pm$ 0.4 & 81.6 $\pm$ 0.8 & 93.3 $\pm$ 0.7 & 50.9 $\pm$ 1.2 & 71.1 $\pm$ 0.9 & 66.3 $\pm$ 0.8 \\
& TIP-Adapter-F & 66.7 $\pm$ 0.0 & 62.6 $\pm$ 0.0 & 23.9 $\pm$ 0.0 & 43.1 $\pm$ 0.1 & 65.6 $\pm$ 0.1 & 85.2 $\pm$ 0.0 & 88.3 $\pm$ 0.0 & 67.6 $\pm$ 0.1 & 93.3 $\pm$ 0.0 & 44.4 $\pm$ 0.0 & 65.3 $\pm$ 0.0 & 64.2 $\pm$ 0.0 \\
& GDA & 67.0 $\pm$ 0.2 & 65.0 $\pm$ 0.1 & 24.4 $\pm$ 0.7 & 46.9 $\pm$ 7.0 & 66.4 $\pm$ 0.5 & 85.3 $\pm$ 0.0 & 88.3 $\pm$ 0.0 & 75.1 $\pm$ 0.6 & 93.3 $\pm$ 0.0 & 49.5 $\pm$ 2.5 & 66.5 $\pm$ 2.1 & 66.2 $\pm$ 1.2 \\
& ProKeR & 67.7 $\pm$ 0.1 & 65.9 $\pm$ 0.1 & 27.3 $\pm$ 0.7 & 57.0 $\pm$ 4.0 & 68.4 $\pm$ 0.2 & 85.2 $\pm$ 0.2 & 88.4 $\pm$ 0.2 & 77.2 $\pm$ 0.7 & 93.6 $\pm$ 0.5 & 49.8 $\pm$ 0.6 & 70.0 $\pm$ 1.1 & 68.2 $\pm$ 0.7 \\
\cmidrule(lr){2-14}
& LoRA & 68.1 $\pm$ 0.6 & 69.6 $\pm$ 0.4 & 28.9 $\pm$ 0.3 & 65.7 $\pm$ 2.9 & 69.7 $\pm$ 0.1 & 86.2 $\pm$ 0.1 & 92.4 $\pm$ 0.1 & 74.0 $\pm$ 0.5 & 94.1 $\pm$ 0.1 & 52.3 $\pm$ 0.5 & 74.7 $\pm$ 0.4 & 70.5 $\pm$ 0.5 \\
& PiSSA & 66.0 $\pm$ 0.0 & 67.7 $\pm$ 0.7 & 29.9 $\pm$ 1.1 & 72.3 $\pm$ 3.0 & 69.7 $\pm$ 0.9 & 83.2 $\pm$ 0.7 & 90.5 $\pm$ 0.1 & 83.5 $\pm$ 0.5 & 93.4 $\pm$ 0.6 & 51.4 $\pm$ 0.8 & 75.4 $\pm$ 1.6 & 71.2 $\pm$ 0.9 \\
& GaLoRE & 64.1 $\pm$ 0.2 & 67.0 $\pm$ 0.3 & 27.7 $\pm$ 0.6 & 68.6 $\pm$ 9.3 & 62.4 $\pm$ 2.9 & 81.6 $\pm$ 0.8 & 90.1 $\pm$ 0.7 & 82.7 $\pm$ 1.0 & 92.7 $\pm$ 0.4 & 52.1 $\pm$ 0.6 & 75.0 $\pm$ 0.7 & 69.5 $\pm$ 1.6 \\
\cmidrule(lr){2-14}
& SO & 68.5 $\pm$ 0.4 & 69.4 $\pm$ 0.4 & 30.0 $\pm$ 0.5 & 69.6 $\pm$ 2.6 & 70.3 $\pm$ 0.2 & 85.9 $\pm$ 0.2 & 92.5 $\pm$ 0.1 & 76.1 $\pm$ 0.5 & 93.8 $\pm$ 0.2 & 53.6 $\pm$ 1.3 & 75.2 $\pm$ 0.6 & 71.4 $\pm$ 0.6 \\
& SignSparseSGD & 66.8 $\pm$ 0.1 & 68.2 $\pm$ 0.3 & 29.4 $\pm$ 1.3 & 71.2 $\pm$ 3.7 & 69.1 $\pm$ 0.6 & 83.8 $\pm$ 0.5 & 91.3 $\pm$ 0.3 & 83.3 $\pm$ 0.5 & 93.7 $\pm$ 0.2 & 54.3 $\pm$ 0.6 & 75.7 $\pm$ 0.3 & 71.5 $\pm$ 0.8 \\
& DynSignSparseSGD & 68.1 $\pm$ 0.1 & 69.8 $\pm$ 0.3 & 30.6 $\pm$ 1.2 & 73.7 $\pm$ 4.0 & 70.6 $\pm$ 0.4 & 85.1 $\pm$ 0.3 & 92.5 $\pm$ 0.2 & 84.7 $\pm$ 0.4 & 93.8 $\pm$ 0.1 & 54.8 $\pm$ 0.6 & 76.2 $\pm$ 0.4 & 72.7 $\pm$ 0.7 \\
\midrule
\multirow{15}{*}{2} & BitFit & 70.0 $\pm$ 0.1 & 70.5 $\pm$ 0.3 & 31.3 $\pm$ 1.0 & 78.0 $\pm$ 2.1 & 69.4 $\pm$ 0.5 & 85.7 $\pm$ 0.3 & 92.0 $\pm$ 0.8 & 79.8 $\pm$ 0.4 & 94.5 $\pm$ 0.3 & 56.5 $\pm$ 0.9 & 77.0 $\pm$ 0.4 & 73.2 $\pm$ 0.6 \\
& LayerNorm & 69.4 $\pm$ 0.0 & 68.0 $\pm$ 0.0 & 26.3 $\pm$ 0.2 & 52.4 $\pm$ 0.7 & 67.1 $\pm$ 0.1 & 86.1 $\pm$ 0.1 & 89.4 $\pm$ 0.0 & 71.2 $\pm$ 0.2 & 93.8 $\pm$ 0.0 & 48.1 $\pm$ 0.4 & 69.5 $\pm$ 0.1 & 67.4 $\pm$ 0.2 \\
& SAFT & 66.9 $\pm$ 0.3 & 68.8 $\pm$ 0.3 & 31.1 $\pm$ 0.9 & 78.7 $\pm$ 2.7 & 67.8 $\pm$ 0.2 & 84.8 $\pm$ 0.2 & 92.5 $\pm$ 0.6 & 85.0 $\pm$ 0.5 & 94.6 $\pm$ 0.1 & 57.4 $\pm$ 2.1 & 76.0 $\pm$ 0.3 & 73.1 $\pm$ 0.7 \\
\cmidrule(lr){2-14}
& CoOp & 68.5 $\pm$ 0.2 & 68.8 $\pm$ 0.4 & 30.9 $\pm$ 0.3 & 67.6 $\pm$ 2.3 & 68.1 $\pm$ 0.5 & 84.6 $\pm$ 0.3 & 91.0 $\pm$ 0.8 & 82.6 $\pm$ 2.2 & 94.1 $\pm$ 0.3 & 54.0 $\pm$ 1.2 & 74.9 $\pm$ 0.7 & 71.4 $\pm$ 0.8 \\
& KgCoOp & 68.9 $\pm$ 0.1 & 68.0 $\pm$ 0.4 & 26.3 $\pm$ 0.4 & 61.6 $\pm$ 2.9 & 65.6 $\pm$ 0.3 & 86.1 $\pm$ 0.1 & 91.3 $\pm$ 0.5 & 70.9 $\pm$ 0.3 & 93.8 $\pm$ 0.4 & 49.8 $\pm$ 0.5 & 72.3 $\pm$ 0.5 & 68.6 $\pm$ 0.6 \\
& ProGrad & 67.4 $\pm$ 0.6 & 69.0 $\pm$ 0.1 & 29.2 $\pm$ 1.3 & 66.7 $\pm$ 3.6 & 65.9 $\pm$ 1.3 & 83.9 $\pm$ 0.2 & 90.8 $\pm$ 0.3 & 80.8 $\pm$ 0.8 & 93.9 $\pm$ 0.3 & 53.3 $\pm$ 1.7 & 74.6 $\pm$ 0.6 & 70.5 $\pm$ 1.0 \\
\cmidrule(lr){2-14}
% & Adapter & 68.3 $\pm$ 0.0 & 66.6 $\pm$ 0.1 & 24.5 $\pm$ 0.1 & 46.5 $\pm$ 0.4 & 66.3 $\pm$ 0.1 & 85.6 $\pm$ 0.1 & 89.1 $\pm$ 0.1 & 70.1 $\pm$ 0.7 & 93.6 $\pm$ 0.1 & 46.0 $\pm$ 0.1 & 67.2 $\pm$ 0.2 & 65.8 $\pm$ 0.2 \\
% & LP++ & 68.4 $\pm$ 0.1 & 69.3 $\pm$ 0.2 & 29.0 $\pm$ 2.0 & 59.5 $\pm$ 6.4 & 66.2 $\pm$ 0.5 & 83.7 $\pm$ 0.5 & 88.0 $\pm$ 0.8 & 88.7 $\pm$ 0.6 & 93.8 $\pm$ 0.2 & 58.3 $\pm$ 1.1 & 73.4 $\pm$ 0.6 & 70.7 $\pm$ 1.2 \\
& TIP-Adapter-F & 66.8 $\pm$ 0.0 & 62.7 $\pm$ 0.0 & 24.0 $\pm$ 0.1 & 44.6 $\pm$ 0.3 & 65.6 $\pm$ 0.1 & 85.3 $\pm$ 0.0 & 88.3 $\pm$ 0.0 & 68.0 $\pm$ 0.1 & 93.3 $\pm$ 0.0 & 44.4 $\pm$ 0.0 & 65.3 $\pm$ 0.0 & 64.4 $\pm$ 0.1 \\
& GDA & 67.8 $\pm$ 0.1 & 67.5 $\pm$ 0.2 & 29.2 $\pm$ 0.7 & 71.8 $\pm$ 2.0 & 69.3 $\pm$ 0.5 & 85.3 $\pm$ 0.1 & 88.5 $\pm$ 0.2 & 88.9 $\pm$ 0.4 & 93.9 $\pm$ 0.8 & 57.1 $\pm$ 1.3 & 72.5 $\pm$ 0.9 & 72.0 $\pm$ 0.6 \\
& ProKeR & 68.4 $\pm$ 0.1 & 67.7 $\pm$ 0.2 & 29.3 $\pm$ 1.0 & 71.1 $\pm$ 1.9 & 70.7 $\pm$ 0.4 & 85.7 $\pm$ 0.1 & 87.5 $\pm$ 0.8 & 80.4 $\pm$ 0.2 & 94.1 $\pm$ 0.6 & 53.6 $\pm$ 0.7 & 72.0 $\pm$ 0.3 & 70.9 $\pm$ 0.6 \\
\cmidrule(lr){2-14}
& LoRA & 67.5 $\pm$ 0.3 & 70.7 $\pm$ 0.3 & 32.3 $\pm$ 0.6 & 75.2 $\pm$ 0.3 & 70.5 $\pm$ 0.3 & 85.7 $\pm$ 0.1 & 91.7 $\pm$ 0.8 & 82.4 $\pm$ 0.4 & 94.3 $\pm$ 0.5 & 58.0 $\pm$ 1.5 & 78.5 $\pm$ 0.8 & 73.3 $\pm$ 0.5 \\
& PiSSA & 66.7 $\pm$ 0.3 & 69.4 $\pm$ 0.3 & 33.9 $\pm$ 1.2 & 80.8 $\pm$ 1.4 & 71.3 $\pm$ 1.1 & 83.0 $\pm$ 0.8 & 91.1 $\pm$ 0.1 & 89.4 $\pm$ 0.5 & 94.8 $\pm$ 0.4 & 59.4 $\pm$ 0.9 & 78.8 $\pm$ 0.4 & 74.4 $\pm$ 0.7 \\
& GaLoRE & 64.3 $\pm$ 0.9 & 68.7 $\pm$ 0.1 & 31.6 $\pm$ 0.8 & 80.3 $\pm$ 1.6 & 70.5 $\pm$ 0.4 & 82.4 $\pm$ 0.4 & 88.8 $\pm$ 1.3 & 87.8 $\pm$ 0.4 & 93.6 $\pm$ 0.5 & 58.2 $\pm$ 1.0 & 77.6 $\pm$ 0.9 & 73.1 $\pm$ 0.7 \\
\cmidrule(lr){2-14}
& SO & 67.9 $\pm$ 0.2 & 70.7 $\pm$ 0.3 & 33.3 $\pm$ 0.8 & 78.8 $\pm$ 1.9 & 71.1 $\pm$ 0.3 & 85.3 $\pm$ 0.1 & 91.9 $\pm$ 1.0 & 85.0 $\pm$ 0.5 & 94.3 $\pm$ 0.4 & 58.6 $\pm$ 0.6 & 79.0 $\pm$ 0.8 & 74.2 $\pm$ 0.6 \\
& SignSparseSGD & 67.4 $\pm$ 0.4 & 69.7 $\pm$ 0.1 & 32.0 $\pm$ 1.4 & 79.5 $\pm$ 2.2 & 71.7 $\pm$ 0.2 & 84.1 $\pm$ 0.3 & 92.0 $\pm$ 0.3 & 87.2 $\pm$ 0.5 & 94.2 $\pm$ 0.3 & 59.5 $\pm$ 1.2 & 78.4 $\pm$ 1.0 & 74.2 $\pm$ 0.7 \\
& DynSignSparseSGD & 68.6 $\pm$ 0.2 & 71.0 $\pm$ 0.1 & 33.8 $\pm$ 2.1 & 82.0 $\pm$ 1.4 & 73.7 $\pm$ 0.3 & 85.1 $\pm$ 0.1 & 91.7 $\pm$ 0.4 & 89.6 $\pm$ 0.2 & 94.3 $\pm$ 0.4 & 60.1 $\pm$ 1.2 & 79.7 $\pm$ 0.9 & 75.4 $\pm$ 0.7 \\
\midrule
\multirow{15}{*}{4} & BitFit & 70.8 $\pm$ 0.0 & 72.2 $\pm$ 0.2 & 33.5 $\pm$ 0.1 & 78.7 $\pm$ 0.6 & 72.5 $\pm$ 0.3 & 86.3 $\pm$ 0.2 & 93.0 $\pm$ 0.2 & 85.1 $\pm$ 0.5 & 95.3 $\pm$ 0.3 & 60.9 $\pm$ 0.3 & 79.4 $\pm$ 0.5 & 75.2 $\pm$ 0.3 \\
& LayerNorm & 70.1 $\pm$ 0.0 & 69.2 $\pm$ 0.1 & 27.3 $\pm$ 0.2 & 53.4 $\pm$ 0.5 & 68.3 $\pm$ 0.1 & 86.4 $\pm$ 0.0 & 89.7 $\pm$ 0.1 & 72.7 $\pm$ 0.1 & 93.9 $\pm$ 0.2 & 49.9 $\pm$ 0.4 & 71.3 $\pm$ 0.1 & 68.4 $\pm$ 0.2 \\
& SAFT & 68.2 $\pm$ 0.2 & 69.5 $\pm$ 0.2 & 32.8 $\pm$ 0.9 & 83.6 $\pm$ 1.3 & 70.0 $\pm$ 0.2 & 85.5 $\pm$ 0.2 & 93.2 $\pm$ 0.4 & 88.7 $\pm$ 0.3 & 95.2 $\pm$ 0.3 & 61.6 $\pm$ 0.8 & 77.9 $\pm$ 0.9 & 75.1 $\pm$ 0.5 \\
\cmidrule(lr){2-14}
& CoOp & 69.6 $\pm$ 0.2 & 70.8 $\pm$ 0.3 & 32.3 $\pm$ 0.6 & 73.2 $\pm$ 2.0 & 71.4 $\pm$ 0.3 & 85.9 $\pm$ 0.4 & 92.6 $\pm$ 0.5 & 86.4 $\pm$ 1.1 & 94.9 $\pm$ 0.1 & 56.8 $\pm$ 1.3 & 76.9 $\pm$ 0.5 & 73.7 $\pm$ 0.7 \\
& KgCoOp & 69.2 $\pm$ 0.1 & 68.7 $\pm$ 0.3 & 27.3 $\pm$ 0.3 & 60.5 $\pm$ 4.0 & 66.1 $\pm$ 0.1 & 86.3 $\pm$ 0.1 & 91.8 $\pm$ 0.4 & 71.9 $\pm$ 0.5 & 94.3 $\pm$ 0.0 & 52.0 $\pm$ 0.4 & 73.1 $\pm$ 0.7 & 69.2 $\pm$ 0.6 \\
& ProGrad & 68.6 $\pm$ 0.3 & 71.1 $\pm$ 0.6 & 30.0 $\pm$ 0.8 & 68.5 $\pm$ 1.5 & 68.7 $\pm$ 3.4 & 85.6 $\pm$ 0.3 & 92.0 $\pm$ 0.5 & 85.4 $\pm$ 0.9 & 94.3 $\pm$ 0.2 & 57.5 $\pm$ 0.1 & 76.2 $\pm$ 0.4 & 72.5 $\pm$ 0.8 \\
\cmidrule(lr){2-14}
% & Adapter & 68.8 $\pm$ 0.1 & 68.2 $\pm$ 0.1 & 25.6 $\pm$ 0.1 & 47.8 $\pm$ 0.8 & 67.4 $\pm$ 0.1 & 85.8 $\pm$ 0.1 & 89.5 $\pm$ 0.1 & 71.1 $\pm$ 0.3 & 93.8 $\pm$ 0.0 & 47.6 $\pm$ 0.2 & 70.0 $\pm$ 0.3 & 66.9 $\pm$ 0.2 \\
% & LP++ & 70.1 $\pm$ 0.1 & 71.9 $\pm$ 0.2 & 32.6 $\pm$ 0.4 & 65.2 $\pm$ 4.8 & 71.2 $\pm$ 0.5 & 84.5 $\pm$ 0.2 & 89.9 $\pm$ 0.3 & 92.5 $\pm$ 1.2 & 94.8 $\pm$ 0.3 & 63.3 $\pm$ 1.3 & 76.5 $\pm$ 1.0 & 73.9 $\pm$ 0.9 \\
& TIP-Adapter-F & 66.9 $\pm$ 0.0 & 62.9 $\pm$ 0.0 & 24.1 $\pm$ 0.1 & 47.5 $\pm$ 0.5 & 65.9 $\pm$ 0.0 & 85.3 $\pm$ 0.0 & 88.3 $\pm$ 0.0 & 69.0 $\pm$ 0.1 & 93.4 $\pm$ 0.1 & 44.5 $\pm$ 0.1 & 65.3 $\pm$ 0.0 & 64.8 $\pm$ 0.1 \\
& GDA & 69.5 $\pm$ 0.1 & 71.9 $\pm$ 0.2 & 35.7 $\pm$ 0.7 & 78.2 $\pm$ 0.9 & 73.8 $\pm$ 0.4 & 85.5 $\pm$ 0.9 & 86.1 $\pm$ 0.7 & 94.8 $\pm$ 0.3 & 94.8 $\pm$ 0.2 & 62.5 $\pm$ 0.3 & 78.8 $\pm$ 0.4 & 75.6 $\pm$ 0.4 \\
& ProKeR & 69.6 $\pm$ 0.2 & 69.8 $\pm$ 0.3 & 32.5 $\pm$ 0.5 & 74.7 $\pm$ 1.0 & 73.7 $\pm$ 0.2 & 86.2 $\pm$ 0.1 & 89.0 $\pm$ 0.4 & 83.1 $\pm$ 0.0 & 95.0 $\pm$ 0.2 & 57.5 $\pm$ 0.6 & 74.3 $\pm$ 0.5 & 73.2 $\pm$ 0.4 \\
\cmidrule(lr){2-14}
& LoRA & 68.0 $\pm$ 0.1 & 71.9 $\pm$ 0.2 & 35.5 $\pm$ 0.5 & 73.5 $\pm$ 1.4 & 76.1 $\pm$ 0.3 & 86.0 $\pm$ 0.4 & 93.1 $\pm$ 0.1 & 89.4 $\pm$ 0.3 & 95.3 $\pm$ 0.4 & 63.3 $\pm$ 0.4 & 82.1 $\pm$ 0.5 & 75.8 $\pm$ 0.4 \\
& PiSSA & 67.5 $\pm$ 0.4 & 70.9 $\pm$ 0.5 & 36.3 $\pm$ 0.7 & 85.1 $\pm$ 2.1 & 76.1 $\pm$ 0.7 & 83.5 $\pm$ 0.9 & 91.9 $\pm$ 0.3 & 91.9 $\pm$ 0.0 & 95.6 $\pm$ 0.2 & 65.0 $\pm$ 0.5 & 80.6 $\pm$ 0.6 & 76.8 $\pm$ 0.6 \\
& GaLoRE & 65.6 $\pm$ 0.4 & 70.1 $\pm$ 0.4 & 37.7 $\pm$ 0.4 & 86.6 $\pm$ 1.1 & 74.7 $\pm$ 0.6 & 82.7 $\pm$ 0.2 & 91.6 $\pm$ 0.3 & 91.2 $\pm$ 0.3 & 94.7 $\pm$ 0.5 & 62.8 $\pm$ 1.0 & 80.1 $\pm$ 1.0 & 76.2 $\pm$ 0.6 \\
\cmidrule(lr){2-14}
& SO & 68.2 $\pm$ 0.2 & 71.8 $\pm$ 0.3 & 36.2 $\pm$ 0.6 & 78.3 $\pm$ 0.9 & 76.8 $\pm$ 0.2 & 85.9 $\pm$ 0.4 & 93.2 $\pm$ 0.2 & 90.8 $\pm$ 0.7 & 95.3 $\pm$ 0.4 & 63.7 $\pm$ 0.6 & 82.1 $\pm$ 0.7 & 76.6 $\pm$ 0.5 \\
& SignSparseSGD & 68.5 $\pm$ 0.3 & 71.9 $\pm$ 0.2 & 37.3 $\pm$ 0.7 & 81.5 $\pm$ 1.2 & 75.7 $\pm$ 0.7 & 84.7 $\pm$ 0.3 & 92.5 $\pm$ 0.2 & 91.4 $\pm$ 1.1 & 95.4 $\pm$ 0.1 & 62.6 $\pm$ 0.7 & 81.1 $\pm$ 1.1 & 76.6 $\pm$ 0.6 \\
& DynSignSparseSGD & 69.4 $\pm$ 0.2 & 72.8 $\pm$ 0.1 & 38.4 $\pm$ 1.1 & 84.1 $\pm$ 2.0 & 77.8 $\pm$ 0.3 & 85.4 $\pm$ 0.2 & 92.8 $\pm$ 0.3 & 93.2 $\pm$ 0.8 & 95.2 $\pm$ 0.1 & 64.5 $\pm$ 0.7 & 82.2 $\pm$ 1.2 & 77.8 $\pm$ 0.6 \\
\midrule
\multirow{15}{*}{8} & BitFit & 71.6 $\pm$ 0.2 & 73.8 $\pm$ 0.1 & 37.1 $\pm$ 0.9 & 84.1 $\pm$ 1.8 & 76.1 $\pm$ 0.6 & 86.5 $\pm$ 0.1 & 93.4 $\pm$ 0.1 & 90.6 $\pm$ 0.6 & 95.4 $\pm$ 0.2 & 64.4 $\pm$ 1.5 & 81.8 $\pm$ 0.5 & 77.7 $\pm$ 0.6 \\
& LayerNorm & 70.7 $\pm$ 0.1 & 70.7 $\pm$ 0.0 & 28.8 $\pm$ 0.1 & 60.2 $\pm$ 0.6 & 69.7 $\pm$ 0.1 & 87.1 $\pm$ 0.1 & 90.5 $\pm$ 0.2 & 74.5 $\pm$ 0.2 & 94.6 $\pm$ 0.1 & 52.6 $\pm$ 0.2 & 73.7 $\pm$ 0.2 & 70.3 $\pm$ 0.2 \\
& SAFT & 69.6 $\pm$ 0.3 & 70.7 $\pm$ 0.2 & 36.3 $\pm$ 0.9 & 88.9 $\pm$ 0.9 & 73.5 $\pm$ 0.5 & 85.7 $\pm$ 0.3 & 93.3 $\pm$ 0.1 & 91.8 $\pm$ 0.3 & 95.8 $\pm$ 0.2 & 65.4 $\pm$ 1.4 & 79.6 $\pm$ 0.5 & 77.3 $\pm$ 0.5 \\
\cmidrule(lr){2-14}
& CoOp & 70.6 $\pm$ 0.1 & 72.8 $\pm$ 0.4 & 34.8 $\pm$ 0.6 & 76.9 $\pm$ 1.3 & 74.0 $\pm$ 0.5 & 85.6 $\pm$ 0.5 & 92.6 $\pm$ 0.3 & 91.9 $\pm$ 0.3 & 95.0 $\pm$ 0.3 & 62.7 $\pm$ 1.4 & 79.6 $\pm$ 0.4 & 76.0 $\pm$ 0.6 \\
& KgCoOp & 69.5 $\pm$ 0.0 & 69.4 $\pm$ 0.2 & 28.0 $\pm$ 0.3 & 64.3 $\pm$ 2.4 & 66.7 $\pm$ 0.3 & 86.5 $\pm$ 0.1 & 92.2 $\pm$ 0.4 & 73.5 $\pm$ 0.7 & 94.2 $\pm$ 0.0 & 55.6 $\pm$ 0.3 & 73.9 $\pm$ 0.5 & 70.3 $\pm$ 0.5 \\
& ProGrad & 69.6 $\pm$ 0.4 & 73.3 $\pm$ 0.2 & 34.8 $\pm$ 0.2 & 76.7 $\pm$ 0.9 & 73.5 $\pm$ 1.9 & 86.1 $\pm$ 0.2 & 93.2 $\pm$ 0.3 & 89.9 $\pm$ 0.8 & 95.0 $\pm$ 0.2 & 61.3 $\pm$ 1.2 & 78.6 $\pm$ 0.5 & 75.6 $\pm$ 0.6 \\
\cmidrule(lr){2-14}
% & Adapter & 69.5 $\pm$ 0.1 & 69.7 $\pm$ 0.1 & 27.3 $\pm$ 0.2 & 54.0 $\pm$ 0.9 & 69.0 $\pm$ 0.1 & 86.4 $\pm$ 0.1 & 89.7 $\pm$ 0.2 & 72.7 $\pm$ 0.4 & 94.2 $\pm$ 0.2 & 50.5 $\pm$ 0.2 & 73.7 $\pm$ 0.2 & 68.8 $\pm$ 0.2 \\
% & LP++ & 71.2 $\pm$ 0.1 & 74.2 $\pm$ 0.2 & 35.8 $\pm$ 0.9 & 77.5 $\pm$ 0.1 & 75.5 $\pm$ 0.3 & 85.3 $\pm$ 0.1 & 90.1 $\pm$ 0.7 & 94.2 $\pm$ 0.6 & 95.2 $\pm$ 0.3 & 66.4 $\pm$ 1.5 & 79.9 $\pm$ 0.7 & 76.8 $\pm$ 0.5 \\
& TIP-Adapter-F & 67.4 $\pm$ 0.0 & 63.6 $\pm$ 0.1 & 24.7 $\pm$ 0.2 & 51.2 $\pm$ 0.3 & 66.7 $\pm$ 0.0 & 85.3 $\pm$ 0.0 & 88.6 $\pm$ 0.0 & 73.3 $\pm$ 0.2 & 93.5 $\pm$ 0.1 & 44.9 $\pm$ 0.1 & 65.7 $\pm$ 0.2 & 65.9 $\pm$ 0.1 \\
& GDA & 71.0 $\pm$ 0.2 & 74.0 $\pm$ 0.2 & 41.3 $\pm$ 0.8 & 85.7 $\pm$ 0.8 & 77.8 $\pm$ 0.5 & 86.4 $\pm$ 0.5 & 91.7 $\pm$ 0.3 & 96.1 $\pm$ 0.4 & 95.2 $\pm$ 0.2 & 67.5 $\pm$ 1.2 & 81.1 $\pm$ 0.8 & 78.9 $\pm$ 0.5 \\
& ProKeR & 70.2 $\pm$ 0.2 & 72.9 $\pm$ 0.3 & 37.1 $\pm$ 0.8 & 84.0 $\pm$ 1.2 & 78.3 $\pm$ 0.1 & 86.7 $\pm$ 0.0 & 90.2 $\pm$ 0.4 & 87.4 $\pm$ 0.2 & 95.4 $\pm$ 0.1 & 63.9 $\pm$ 1.2 & 78.6 $\pm$ 0.6 & 76.8 $\pm$ 0.5 \\
\cmidrule(lr){2-14}
& LoRA & 69.1 $\pm$ 0.2 & 73.3 $\pm$ 0.4 & 43.7 $\pm$ 0.5 & 84.6 $\pm$ 1.2 & 81.4 $\pm$ 0.2 & 85.3 $\pm$ 0.2 & 93.5 $\pm$ 0.1 & 94.3 $\pm$ 0.1 & 95.7 $\pm$ 0.1 & 67.0 $\pm$ 1.2 & 83.7 $\pm$ 0.2 & 79.2 $\pm$ 0.4 \\
& PiSSA & 69.0 $\pm$ 0.2 & 72.5 $\pm$ 0.2 & 43.5 $\pm$ 0.3 & 87.7 $\pm$ 0.9 & 82.0 $\pm$ 0.1 & 84.0 $\pm$ 0.0 & 93.4 $\pm$ 0.2 & 94.7 $\pm$ 0.5 & 95.5 $\pm$ 0.2 & 66.6 $\pm$ 1.4 & 82.8 $\pm$ 0.3 & 79.2 $\pm$ 0.4 \\
& GaLoRE & 66.5 $\pm$ 0.2 & 72.0 $\pm$ 0.6 & 43.6 $\pm$ 1.8 & 88.5 $\pm$ 2.2 & 80.1 $\pm$ 0.4 & 82.9 $\pm$ 0.2 & 92.6 $\pm$ 0.3 & 93.5 $\pm$ 1.0 & 95.3 $\pm$ 0.1 & 65.3 $\pm$ 0.8 & 82.5 $\pm$ 0.6 & 78.4 $\pm$ 0.7 \\
\cmidrule(lr){2-14}
& SO & 69.4 $\pm$ 0.1 & 73.0 $\pm$ 0.3 & 44.0 $\pm$ 0.9 & 86.7 $\pm$ 1.7 & 81.7 $\pm$ 0.2 & 85.1 $\pm$ 0.1 & 93.7 $\pm$ 0.0 & 94.6 $\pm$ 0.2 & 95.8 $\pm$ 0.3 & 67.1 $\pm$ 1.1 & 83.8 $\pm$ 0.2 & 79.5 $\pm$ 0.5 \\
& SignSparseSGD & 69.9 $\pm$ 0.3 & 73.7 $\pm$ 0.1 & 41.1 $\pm$ 1.7 & 84.7 $\pm$ 5.2 & 79.8 $\pm$ 0.7 & 84.8 $\pm$ 0.6 & 92.9 $\pm$ 0.4 & 94.0 $\pm$ 0.3 & 95.9 $\pm$ 0.2 & 66.5 $\pm$ 0.7 & 83.6 $\pm$ 0.6 & 78.8 $\pm$ 1.0 \\
& DynSignSparseSGD & 70.5 $\pm$ 0.4 & 74.5 $\pm$ 0.2 & 45.2 $\pm$ 0.7 & 85.2 $\pm$ 4.0 & 82.4 $\pm$ 0.5 & 85.3 $\pm$ 0.3 & 93.4 $\pm$ 0.1 & 95.0 $\pm$ 0.2 & 95.8 $\pm$ 0.1 & 68.0 $\pm$ 1.3 & 84.7 $\pm$ 0.2 & 80.0 $\pm$ 0.7 \\
\midrule
\multirow{15}{*}{16} & BitFit & 72.6 $\pm$ 0.0 & 75.5 $\pm$ 0.1 & 42.6 $\pm$ 0.3 & 88.5 $\pm$ 0.8 & 79.6 $\pm$ 0.2 & 87.0 $\pm$ 0.0 & 94.0 $\pm$ 0.2 & 94.5 $\pm$ 0.4 & 95.6 $\pm$ 0.3 & 68.3 $\pm$ 1.5 & 84.0 $\pm$ 0.1 & 80.2 $\pm$ 0.4 \\
& LayerNorm & 71.4 $\pm$ 0.0 & 72.2 $\pm$ 0.1 & 31.4 $\pm$ 0.2 & 67.8 $\pm$ 0.5 & 71.8 $\pm$ 0.1 & 87.4 $\pm$ 0.0 & 92.9 $\pm$ 0.1 & 78.2 $\pm$ 0.2 & 94.9 $\pm$ 0.0 & 56.5 $\pm$ 0.2 & 76.9 $\pm$ 0.1 & 72.8 $\pm$ 0.1 \\
& SAFT & 71.4 $\pm$ 0.1 & 72.2 $\pm$ 0.1 & 41.8 $\pm$ 0.2 & 93.1 $\pm$ 0.5 & 77.1 $\pm$ 0.1 & 85.9 $\pm$ 0.3 & 93.6 $\pm$ 0.4 & 95.1 $\pm$ 0.1 & 95.9 $\pm$ 0.2 & 69.6 $\pm$ 0.7 & 82.5 $\pm$ 0.2 & 79.8 $\pm$ 0.3 \\
\cmidrule(lr){2-14}
& CoOp & 71.7 $\pm$ 0.1 & 74.6 $\pm$ 0.3 & 38.4 $\pm$ 0.5 & 78.5 $\pm$ 0.5 & 77.9 $\pm$ 0.4 & 86.6 $\pm$ 0.2 & 93.4 $\pm$ 0.4 & 95.0 $\pm$ 0.6 & 95.2 $\pm$ 0.6 & 67.7 $\pm$ 0.5 & 83.3 $\pm$ 0.2 & 78.4 $\pm$ 0.4 \\
& KgCoOp & 69.7 $\pm$ 0.0 & 70.1 $\pm$ 0.2 & 29.1 $\pm$ 0.5 & 67.6 $\pm$ 1.1 & 67.4 $\pm$ 0.1 & 86.6 $\pm$ 0.1 & 92.6 $\pm$ 0.3 & 77.0 $\pm$ 1.0 & 94.3 $\pm$ 0.0 & 59.7 $\pm$ 1.1 & 75.5 $\pm$ 0.3 & 71.8 $\pm$ 0.4 \\
& ProGrad & 70.2 $\pm$ 1.8 & 74.7 $\pm$ 0.4 & 38.0 $\pm$ 0.1 & 79.6 $\pm$ 1.7 & 76.1 $\pm$ 1.8 & 86.5 $\pm$ 0.3 & 93.2 $\pm$ 0.3 & 93.0 $\pm$ 0.9 & 95.2 $\pm$ 0.4 & 65.3 $\pm$ 0.5 & 81.5 $\pm$ 0.4 & 77.6 $\pm$ 0.8 \\
\cmidrule(lr){2-14}
% & Adapter & 70.3 $\pm$ 0.1 & 71.8 $\pm$ 0.1 & 29.2 $\pm$ 0.5 & 60.2 $\pm$ 1.0 & 71.1 $\pm$ 0.1 & 86.8 $\pm$ 0.0 & 92.0 $\pm$ 0.0 & 80.3 $\pm$ 0.2 & 94.9 $\pm$ 0.1 & 57.7 $\pm$ 0.4 & 78.4 $\pm$ 0.3 & 72.1 $\pm$ 0.3 \\
% & LP++ & 72.1 $\pm$ 0.1 & 75.7 $\pm$ 0.1 & 39.7 $\pm$ 0.2 & 82.0 $\pm$ 0.8 & 78.9 $\pm$ 0.1 & 86.3 $\pm$ 0.3 & 91.9 $\pm$ 0.2 & 96.1 $\pm$ 0.2 & 95.8 $\pm$ 0.5 & 71.2 $\pm$ 0.8 & 82.7 $\pm$ 0.3 & 79.3 $\pm$ 0.3 \\
& TIP-Adapter-F & 68.9 $\pm$ 0.1 & 67.3 $\pm$ 0.1 & 28.7 $\pm$ 0.1 & 57.3 $\pm$ 0.3 & 70.9 $\pm$ 0.1 & 85.5 $\pm$ 0.1 & 89.2 $\pm$ 0.1 & 87.0 $\pm$ 0.3 & 93.7 $\pm$ 0.1 & 46.8 $\pm$ 0.3 & 68.2 $\pm$ 0.4 & 69.4 $\pm$ 0.2 \\
& GDA & 71.3 $\pm$ 0.1 & 75.7 $\pm$ 0.0 & 46.9 $\pm$ 0.3 & 86.5 $\pm$ 0.5 & 81.3 $\pm$ 0.4 & 86.5 $\pm$ 0.4 & 91.8 $\pm$ 0.3 & 97.6 $\pm$ 0.6 & 96.0 $\pm$ 0.0 & 72.1 $\pm$ 0.1 & 83.2 $\pm$ 0.3 & 80.8 $\pm$ 0.3 \\
& ProKeR & 71.9 $\pm$ 0.1 & 76.0 $\pm$ 0.0 & 44.5 $\pm$ 0.6 & 88.0 $\pm$ 0.4 & 83.0 $\pm$ 0.2 & 87.3 $\pm$ 0.2 & 92.5 $\pm$ 0.2 & 92.9 $\pm$ 0.4 & 96.0 $\pm$ 0.1 & 70.3 $\pm$ 0.5 & 82.7 $\pm$ 0.4 & 80.5 $\pm$ 0.3 \\
\cmidrule(lr){2-14}
& LoRA & 71.3 $\pm$ 0.2 & 74.8 $\pm$ 0.4 & 50.4 $\pm$ 1.0 & 90.3 $\pm$ 0.5 & 85.4 $\pm$ 0.3 & 85.8 $\pm$ 0.1 & 94.2 $\pm$ 0.3 & 97.0 $\pm$ 0.4 & 96.2 $\pm$ 0.2 & 71.1 $\pm$ 1.5 & 85.7 $\pm$ 0.5 & 82.0 $\pm$ 0.5 \\
& PiSSA & 70.9 $\pm$ 0.2 & 74.4 $\pm$ 0.2 & 51.3 $\pm$ 0.9 & 92.1 $\pm$ 1.0 & 85.6 $\pm$ 0.5 & 84.9 $\pm$ 0.2 & 93.4 $\pm$ 0.3 & 97.1 $\pm$ 0.1 & 96.1 $\pm$ 0.1 & 71.6 $\pm$ 0.8 & 85.5 $\pm$ 0.4 & 82.1 $\pm$ 0.4 \\
& GaLoRE & 68.7 $\pm$ 0.1 & 73.5 $\pm$ 0.2 & 50.4 $\pm$ 0.8 & 91.0 $\pm$ 1.6 & 83.7 $\pm$ 0.3 & 83.3 $\pm$ 0.4 & 92.9 $\pm$ 0.1 & 96.5 $\pm$ 0.4 & 95.7 $\pm$ 0.2 & 69.7 $\pm$ 0.3 & 85.4 $\pm$ 0.3 & 81.0 $\pm$ 0.4 \\
\cmidrule(lr){2-14}
& SO & 71.3 $\pm$ 0.2 & 74.7 $\pm$ 0.4 & 51.5 $\pm$ 1.1 & 90.6 $\pm$ 1.2 & 85.5 $\pm$ 0.3 & 85.7 $\pm$ 0.1 & 94.2 $\pm$ 0.1 & 97.2 $\pm$ 0.4 & 96.2 $\pm$ 0.1 & 71.1 $\pm$ 1.4 & 85.7 $\pm$ 0.4 & 82.2 $\pm$ 0.5 \\
& SignSparseSGD & 71.6 $\pm$ 0.1 & 75.0 $\pm$ 0.2 & 45.6 $\pm$ 0.9 & 91.2 $\pm$ 1.1 & 83.6 $\pm$ 0.5 & 85.7 $\pm$ 0.1 & 93.6 $\pm$ 0.3 & 96.8 $\pm$ 0.2 & 96.2 $\pm$ 0.2 & 71.2 $\pm$ 0.3 & 85.8 $\pm$ 0.0 & 81.5 $\pm$ 0.4 \\
& DynSignSparseSGD & 72.2 $\pm$ 0.1 & 75.7 $\pm$ 0.2 & 51.7 $\pm$ 0.6 & 91.8 $\pm$ 0.5 & 85.9 $\pm$ 0.3 & 86.1 $\pm$ 0.1 & 93.9 $\pm$ 0.1 & 97.7 $\pm$ 0.2 & 96.4 $\pm$ 0.3 & 72.2 $\pm$ 0.4 & 86.9 $\pm$ 0.1 & 82.8 $\pm$ 0.3 \\
\bottomrule
\end{tabular}
\end{adjustbox}
\end{table}
\begin{table}
\caption{Few-shot classification performance on 11 datasets with CLIP(ViT-L/14) backbone. Top-1 accuracy (\%) averaged over 3 random seeds.}
\label{tab:few_shot_ViTL14_with_aug}
\centering
\footnotesize
\setlength{\tabcolsep}{1.8pt}
\renewcommand{\arraystretch}{1.00}
\begin{adjustbox}{max totalsize={\textwidth}{0.94\textheight},center}
\begin{tabular}{@{}c l *{12}{d}@{}}
\toprule
Shots & Method & ImageNet & SUN & Aircr. & EuroSAT & Cars & Food & Pets & Flow. & Caltech & DTD & UCF & Avg. \\ \midrule
\multirow{1}{*}{0} & CLIP & 73.5 & 67.7 & 30.0 & 55.1 & 76.8 & 90.3 & 93.1 & 76.2 & 95.2 & 52.4 & 73.7 & 71.3 \\
\midrule
\multirow{15}{*}{1} & BitFit & 75.9 $\pm$ 0.4 & 73.6 $\pm$ 0.1 & 39.7 $\pm$ 0.9 & 71.4 $\pm$ 4.1 & 80.0 $\pm$ 0.1 & 90.9 $\pm$ 0.3 & 94.5 $\pm$ 0.1 & 83.2 $\pm$ 1.0 & 95.8 $\pm$ 0.2 & 59.5 $\pm$ 0.4 & 80.5 $\pm$ 0.3 & 76.8 $\pm$ 0.7 \\
& LayerNorm & 75.9 $\pm$ 0.1 & 71.2 $\pm$ 0.1 & 33.7 $\pm$ 0.1 & 61.9 $\pm$ 0.8 & 78.3 $\pm$ 0.2 & 90.9 $\pm$ 0.1 & 93.9 $\pm$ 0.1 & 78.7 $\pm$ 0.1 & 95.4 $\pm$ 0.2 & 54.1 $\pm$ 0.2 & 75.7 $\pm$ 0.3 & 73.6 $\pm$ 0.2 \\
& SAFT & 74.2 $\pm$ 0.6 & 73.7 $\pm$ 0.1 & 38.4 $\pm$ 1.6 & 74.2 $\pm$ 3.5 & 78.5 $\pm$ 0.0 & 89.4 $\pm$ 0.5 & 91.0 $\pm$ 1.9 & 89.1 $\pm$ 0.8 & 96.2 $\pm$ 0.8 & 59.6 $\pm$ 1.7 & 81.3 $\pm$ 0.8 & 76.9 $\pm$ 1.1 \\
\cmidrule(lr){2-14}
& CoOp & 71.2 $\pm$ 0.6 & 68.8 $\pm$ 0.6 & 37.2 $\pm$ 0.4 & 58.8 $\pm$ 3.3 & 78.1 $\pm$ 0.9 & 88.9 $\pm$ 0.7 & 93.4 $\pm$ 0.3 & 89.3 $\pm$ 1.0 & 94.5 $\pm$ 0.3 & 53.8 $\pm$ 1.6 & 77.6 $\pm$ 0.4 & 73.8 $\pm$ 0.9 \\
& KgCoOp & 75.6 $\pm$ 0.0 & 71.6 $\pm$ 0.2 & 34.1 $\pm$ 0.8 & 69.1 $\pm$ 2.0 & 77.7 $\pm$ 0.7 & 90.9 $\pm$ 0.2 & 93.8 $\pm$ 0.1 & 80.9 $\pm$ 0.5 & 95.3 $\pm$ 0.6 & 56.6 $\pm$ 0.4 & 76.8 $\pm$ 0.9 & 74.8 $\pm$ 0.6 \\
& ProGrad & 70.9 $\pm$ 0.6 & 69.2 $\pm$ 0.9 & 35.3 $\pm$ 1.0 & 59.3 $\pm$ 4.4 & 72.7 $\pm$ 6.5 & 88.6 $\pm$ 0.1 & 93.4 $\pm$ 0.2 & 86.2 $\pm$ 1.5 & 94.2 $\pm$ 1.2 & 54.8 $\pm$ 1.9 & 77.4 $\pm$ 0.7 & 72.9 $\pm$ 1.7 \\
\cmidrule(lr){2-14}
% & Adapter & 75.1 $\pm$ 0.1 & 70.4 $\pm$ 0.1 & 32.0 $\pm$ 0.2 & 59.0 $\pm$ 1.0 & 78.0 $\pm$ 0.1 & 90.6 $\pm$ 0.1 & 93.4 $\pm$ 0.1 & 77.0 $\pm$ 0.3 & 95.2 $\pm$ 0.1 & 53.6 $\pm$ 0.1 & 75.5 $\pm$ 0.3 & 72.7 $\pm$ 0.2 \\
% & LP++ & 73.4 $\pm$ 0.3 & 70.9 $\pm$ 0.4 & 36.7 $\pm$ 0.8 & 52.7 $\pm$ 6.2 & 73.6 $\pm$ 1.9 & 88.8 $\pm$ 0.6 & 91.5 $\pm$ 0.9 & 88.7 $\pm$ 1.7 & 95.7 $\pm$ 0.1 & 57.0 $\pm$ 0.8 & 78.9 $\pm$ 1.1 & 73.5 $\pm$ 1.3 \\
& TIP-Adapter-F & 73.5 $\pm$ 0.0 & 67.7 $\pm$ 0.0 & 30.1 $\pm$ 0.1 & 55.5 $\pm$ 0.2 & 76.9 $\pm$ 0.0 & 90.3 $\pm$ 0.0 & 93.1 $\pm$ 0.0 & 76.4 $\pm$ 0.0 & 95.2 $\pm$ 0.0 & 52.4 $\pm$ 0.0 & 73.7 $\pm$ 0.0 & 71.3 $\pm$ 0.0 \\
& GDA & 73.9 $\pm$ 0.3 & 70.2 $\pm$ 0.1 & 33.5 $\pm$ 1.7 & 63.1 $\pm$ 5.5 & 77.6 $\pm$ 0.3 & 90.3 $\pm$ 0.0 & 93.2 $\pm$ 0.2 & 87.7 $\pm$ 1.8 & 95.7 $\pm$ 0.4 & 54.9 $\pm$ 0.3 & 75.2 $\pm$ 2.2 & 74.1 $\pm$ 1.2 \\
& ProKeR & 74.4 $\pm$ 0.1 & 70.4 $\pm$ 0.1 & 36.5 $\pm$ 0.2 & 65.4 $\pm$ 3.7 & 78.9 $\pm$ 0.4 & 90.4 $\pm$ 0.1 & 93.7 $\pm$ 0.9 & 84.5 $\pm$ 0.7 & 95.4 $\pm$ 0.3 & 56.2 $\pm$ 0.4 & 76.7 $\pm$ 0.3 & 74.8 $\pm$ 0.6 \\
\cmidrule(lr){2-14}
& LoRA & 72.3 $\pm$ 0.5 & 73.6 $\pm$ 0.4 & 41.1 $\pm$ 1.2 & 69.0 $\pm$ 5.6 & 80.7 $\pm$ 0.2 & 90.8 $\pm$ 0.2 & 94.4 $\pm$ 0.1 & 86.6 $\pm$ 1.4 & 95.8 $\pm$ 0.5 & 59.0 $\pm$ 0.3 & 80.8 $\pm$ 0.2 & 76.7 $\pm$ 1.0 \\
& PiSSA & 72.2 $\pm$ 0.6 & 72.8 $\pm$ 0.5 & 39.2 $\pm$ 0.9 & 75.1 $\pm$ 1.1 & 78.4 $\pm$ 0.5 & 89.3 $\pm$ 0.5 & 93.4 $\pm$ 0.7 & 92.2 $\pm$ 0.8 & 95.7 $\pm$ 0.4 & 62.1 $\pm$ 0.6 & 82.9 $\pm$ 0.5 & 77.6 $\pm$ 0.6 \\
& GaLoRE & 72.5 $\pm$ 0.5 & 72.5 $\pm$ 0.5 & 39.6 $\pm$ 1.1 & 74.3 $\pm$ 2.8 & 74.2 $\pm$ 4.3 & 89.3 $\pm$ 0.4 & 92.9 $\pm$ 0.5 & 90.2 $\pm$ 2.0 & 95.5 $\pm$ 0.6 & 60.0 $\pm$ 0.9 & 82.1 $\pm$ 0.3 & 76.6 $\pm$ 1.2 \\
\cmidrule(lr){2-14}
& SO & 72.2 $\pm$ 0.5 & 73.5 $\pm$ 0.3 & 41.7 $\pm$ 1.4 & 72.9 $\pm$ 3.7 & 80.9 $\pm$ 0.3 & 90.7 $\pm$ 0.3 & 94.6 $\pm$ 0.1 & 88.7 $\pm$ 1.1 & 95.7 $\pm$ 0.4 & 60.2 $\pm$ 0.3 & 81.6 $\pm$ 0.2 & 77.5 $\pm$ 0.8 \\
& SignSparseSGD & 74.6 $\pm$ 0.5 & 73.3 $\pm$ 0.2 & 40.5 $\pm$ 1.7 & 71.7 $\pm$ 3.7 & 80.0 $\pm$ 0.3 & 90.2 $\pm$ 0.1 & 94.6 $\pm$ 0.2 & 91.5 $\pm$ 1.3 & 95.5 $\pm$ 0.3 & 61.7 $\pm$ 1.1 & 82.8 $\pm$ 0.5 & 77.8 $\pm$ 0.9 \\
& DynSignSparseSGD & 75.6 $\pm$ 0.2 & 74.1 $\pm$ 0.2 & 41.3 $\pm$ 0.9 & 73.8 $\pm$ 4.1 & 80.9 $\pm$ 0.3 & 90.7 $\pm$ 0.2 & 94.7 $\pm$ 0.3 & 92.3 $\pm$ 1.0 & 95.5 $\pm$ 0.2 & 61.5 $\pm$ 0.7 & 82.9 $\pm$ 0.8 & 78.5 $\pm$ 0.8 \\
\midrule
\multirow{15}{*}{2} & BitFit & 76.3 $\pm$ 0.2 & 75.0 $\pm$ 0.1 & 41.5 $\pm$ 1.1 & 76.5 $\pm$ 2.2 & 81.0 $\pm$ 0.2 & 91.1 $\pm$ 0.2 & 94.5 $\pm$ 0.3 & 87.8 $\pm$ 0.3 & 96.6 $\pm$ 0.7 & 64.0 $\pm$ 0.5 & 82.2 $\pm$ 0.6 & 78.8 $\pm$ 0.6 \\
& LayerNorm & 76.7 $\pm$ 0.1 & 72.6 $\pm$ 0.1 & 35.5 $\pm$ 0.3 & 63.5 $\pm$ 0.3 & 79.0 $\pm$ 0.1 & 91.1 $\pm$ 0.2 & 94.1 $\pm$ 0.1 & 80.7 $\pm$ 0.4 & 95.9 $\pm$ 0.0 & 55.7 $\pm$ 0.1 & 76.8 $\pm$ 0.2 & 74.7 $\pm$ 0.2 \\
& SAFT & 74.7 $\pm$ 0.3 & 75.1 $\pm$ 0.2 & 40.9 $\pm$ 1.4 & 85.1 $\pm$ 2.6 & 80.0 $\pm$ 0.2 & 90.7 $\pm$ 0.2 & 94.2 $\pm$ 0.2 & 92.7 $\pm$ 0.7 & 97.1 $\pm$ 0.3 & 66.8 $\pm$ 0.9 & 82.4 $\pm$ 0.8 & 80.0 $\pm$ 0.7 \\
\cmidrule(lr){2-14}
& CoOp & 73.3 $\pm$ 0.2 & 71.1 $\pm$ 0.3 & 39.6 $\pm$ 1.9 & 75.9 $\pm$ 1.0 & 79.2 $\pm$ 0.1 & 89.5 $\pm$ 0.8 & 94.1 $\pm$ 0.6 & 92.1 $\pm$ 0.5 & 95.9 $\pm$ 0.9 & 60.4 $\pm$ 1.1 & 81.0 $\pm$ 0.8 & 77.5 $\pm$ 0.7 \\
& KgCoOp & 76.0 $\pm$ 0.1 & 72.5 $\pm$ 0.2 & 35.2 $\pm$ 0.9 & 75.2 $\pm$ 1.8 & 78.1 $\pm$ 0.6 & 91.1 $\pm$ 0.1 & 94.4 $\pm$ 0.5 & 82.6 $\pm$ 0.3 & 95.8 $\pm$ 0.5 & 59.3 $\pm$ 0.5 & 78.0 $\pm$ 0.5 & 76.2 $\pm$ 0.5 \\
& ProGrad & 68.3 $\pm$ 7.6 & 71.9 $\pm$ 0.8 & 37.6 $\pm$ 1.8 & 75.4 $\pm$ 1.0 & 70.8 $\pm$ 11.9 & 89.5 $\pm$ 1.0 & 94.4 $\pm$ 0.6 & 90.1 $\pm$ 2.1 & 95.8 $\pm$ 0.6 & 58.4 $\pm$ 0.4 & 80.1 $\pm$ 0.3 & 75.6 $\pm$ 2.6 \\
\cmidrule(lr){2-14}
% & Adapter & 75.6 $\pm$ 0.1 & 72.0 $\pm$ 0.1 & 32.9 $\pm$ 0.5 & 60.9 $\pm$ 1.5 & 78.7 $\pm$ 0.1 & 90.7 $\pm$ 0.2 & 93.8 $\pm$ 0.1 & 78.9 $\pm$ 0.3 & 95.5 $\pm$ 0.2 & 55.0 $\pm$ 0.4 & 76.6 $\pm$ 0.2 & 73.7 $\pm$ 0.3 \\
% & LP++ & 75.2 $\pm$ 0.1 & 73.9 $\pm$ 0.1 & 39.3 $\pm$ 1.9 & 66.6 $\pm$ 4.8 & 76.2 $\pm$ 0.5 & 89.7 $\pm$ 0.4 & 92.1 $\pm$ 0.9 & 93.5 $\pm$ 0.8 & 96.8 $\pm$ 0.3 & 64.4 $\pm$ 0.8 & 80.3 $\pm$ 0.2 & 77.1 $\pm$ 1.0 \\
& TIP-Adapter-F & 73.5 $\pm$ 0.0 & 67.8 $\pm$ 0.0 & 30.0 $\pm$ 0.1 & 56.1 $\pm$ 0.4 & 77.1 $\pm$ 0.1 & 90.3 $\pm$ 0.0 & 93.1 $\pm$ 0.0 & 76.8 $\pm$ 0.1 & 95.2 $\pm$ 0.0 & 52.4 $\pm$ 0.0 & 73.8 $\pm$ 0.0 & 71.5 $\pm$ 0.1 \\
& GDA & 75.1 $\pm$ 0.0 & 72.9 $\pm$ 0.1 & 39.3 $\pm$ 1.1 & 76.8 $\pm$ 1.9 & 80.2 $\pm$ 0.5 & 90.1 $\pm$ 0.6 & 92.1 $\pm$ 1.4 & 95.3 $\pm$ 0.9 & 96.3 $\pm$ 0.1 & 59.0 $\pm$ 3.1 & 80.7 $\pm$ 0.8 & 78.0 $\pm$ 0.9 \\
& ProKeR & 75.0 $\pm$ 0.0 & 71.9 $\pm$ 0.2 & 38.4 $\pm$ 0.3 & 74.5 $\pm$ 0.5 & 80.5 $\pm$ 0.4 & 90.7 $\pm$ 0.0 & 93.7 $\pm$ 0.3 & 86.4 $\pm$ 0.6 & 96.5 $\pm$ 0.3 & 59.3 $\pm$ 1.0 & 78.3 $\pm$ 0.3 & 76.8 $\pm$ 0.4 \\
\cmidrule(lr){2-14}
& LoRA & 72.7 $\pm$ 0.1 & 74.1 $\pm$ 0.2 & 43.9 $\pm$ 1.0 & 75.3 $\pm$ 1.7 & 82.6 $\pm$ 0.4 & 90.5 $\pm$ 0.3 & 94.5 $\pm$ 0.8 & 93.3 $\pm$ 0.2 & 96.9 $\pm$ 0.5 & 65.3 $\pm$ 0.7 & 83.6 $\pm$ 0.7 & 79.3 $\pm$ 0.6 \\
& PiSSA & 72.2 $\pm$ 0.1 & 74.0 $\pm$ 0.2 & 41.6 $\pm$ 1.8 & 86.7 $\pm$ 2.7 & 81.6 $\pm$ 0.5 & 89.7 $\pm$ 0.4 & 92.9 $\pm$ 0.4 & 94.6 $\pm$ 0.9 & 96.6 $\pm$ 0.5 & 66.8 $\pm$ 0.4 & 84.0 $\pm$ 0.6 & 80.1 $\pm$ 0.8 \\
& GaLoRE & 73.0 $\pm$ 0.2 & 73.9 $\pm$ 0.3 & 40.7 $\pm$ 3.3 & 85.3 $\pm$ 2.2 & 80.7 $\pm$ 0.3 & 89.5 $\pm$ 0.3 & 93.5 $\pm$ 0.6 & 93.9 $\pm$ 0.5 & 96.1 $\pm$ 1.2 & 66.8 $\pm$ 0.9 & 83.9 $\pm$ 0.6 & 79.8 $\pm$ 0.9 \\
\cmidrule(lr){2-14}
& SO & 72.9 $\pm$ 0.2 & 73.9 $\pm$ 0.1 & 44.2 $\pm$ 1.2 & 78.4 $\pm$ 2.0 & 82.5 $\pm$ 0.5 & 90.5 $\pm$ 0.4 & 94.7 $\pm$ 0.6 & 93.5 $\pm$ 0.3 & 97.0 $\pm$ 0.5 & 66.5 $\pm$ 0.6 & 84.1 $\pm$ 0.6 & 79.8 $\pm$ 0.6 \\
& SignSparseSGD & 75.3 $\pm$ 0.2 & 74.9 $\pm$ 0.1 & 41.9 $\pm$ 1.9 & 84.4 $\pm$ 0.7 & 82.2 $\pm$ 0.4 & 90.5 $\pm$ 0.2 & 94.3 $\pm$ 0.4 & 94.3 $\pm$ 1.0 & 97.0 $\pm$ 0.3 & 67.6 $\pm$ 1.2 & 84.7 $\pm$ 0.7 & 80.6 $\pm$ 0.6 \\
& DynSignSparseSGD & 75.9 $\pm$ 0.3 & 75.6 $\pm$ 0.3 & 43.9 $\pm$ 2.6 & 82.6 $\pm$ 1.0 & 83.2 $\pm$ 0.6 & 90.8 $\pm$ 0.2 & 94.5 $\pm$ 0.3 & 95.2 $\pm$ 0.7 & 96.8 $\pm$ 0.6 & 68.5 $\pm$ 1.2 & 85.0 $\pm$ 0.6 & 81.1 $\pm$ 0.8 \\
\midrule
\multirow{15}{*}{4} & BitFit & 77.0 $\pm$ 0.1 & 76.5 $\pm$ 0.1 & 45.8 $\pm$ 0.5 & 78.8 $\pm$ 1.2 & 83.4 $\pm$ 0.2 & 91.2 $\pm$ 0.1 & 95.4 $\pm$ 0.3 & 93.8 $\pm$ 0.3 & 97.3 $\pm$ 0.3 & 66.0 $\pm$ 0.4 & 85.1 $\pm$ 0.6 & 80.9 $\pm$ 0.4 \\
& LayerNorm & 77.5 $\pm$ 0.1 & 74.4 $\pm$ 0.1 & 37.6 $\pm$ 0.2 & 64.2 $\pm$ 0.4 & 80.5 $\pm$ 0.0 & 91.4 $\pm$ 0.0 & 94.7 $\pm$ 0.2 & 82.4 $\pm$ 0.1 & 96.2 $\pm$ 0.1 & 57.9 $\pm$ 0.2 & 78.5 $\pm$ 0.2 & 75.9 $\pm$ 0.1 \\
& SAFT & 75.7 $\pm$ 0.3 & 75.8 $\pm$ 0.1 & 44.9 $\pm$ 0.7 & 87.7 $\pm$ 1.9 & 82.3 $\pm$ 0.5 & 91.1 $\pm$ 0.1 & 95.2 $\pm$ 0.2 & 95.5 $\pm$ 0.3 & 97.4 $\pm$ 0.3 & 68.4 $\pm$ 0.4 & 85.2 $\pm$ 0.5 & 81.7 $\pm$ 0.5 \\
\cmidrule(lr){2-14}
& CoOp & 74.6 $\pm$ 0.1 & 73.2 $\pm$ 0.3 & 43.1 $\pm$ 0.7 & 78.2 $\pm$ 3.8 & 82.5 $\pm$ 0.4 & 89.7 $\pm$ 0.2 & 94.1 $\pm$ 0.0 & 95.7 $\pm$ 0.2 & 96.5 $\pm$ 0.2 & 63.7 $\pm$ 0.9 & 82.4 $\pm$ 0.3 & 79.4 $\pm$ 0.6 \\
& KgCoOp & 76.2 $\pm$ 0.0 & 73.3 $\pm$ 0.0 & 36.6 $\pm$ 0.6 & 73.4 $\pm$ 2.0 & 78.4 $\pm$ 0.3 & 91.2 $\pm$ 0.1 & 94.9 $\pm$ 0.1 & 84.1 $\pm$ 0.6 & 96.5 $\pm$ 0.2 & 61.2 $\pm$ 1.2 & 79.0 $\pm$ 0.4 & 76.8 $\pm$ 0.5 \\
& ProGrad & 74.9 $\pm$ 0.3 & 74.2 $\pm$ 0.1 & 41.8 $\pm$ 0.5 & 78.0 $\pm$ 3.0 & 82.8 $\pm$ 0.2 & 90.2 $\pm$ 0.4 & 94.7 $\pm$ 0.4 & 93.2 $\pm$ 0.6 & 96.3 $\pm$ 0.2 & 63.6 $\pm$ 0.5 & 82.1 $\pm$ 0.7 & 79.2 $\pm$ 0.6 \\
\cmidrule(lr){2-14}
% & Adapter & 76.2 $\pm$ 0.1 & 74.0 $\pm$ 0.1 & 35.9 $\pm$ 0.5 & 61.0 $\pm$ 0.8 & 80.3 $\pm$ 0.1 & 90.8 $\pm$ 0.2 & 94.4 $\pm$ 0.1 & 83.3 $\pm$ 0.5 & 96.1 $\pm$ 0.1 & 57.2 $\pm$ 0.4 & 79.7 $\pm$ 0.3 & 75.3 $\pm$ 0.3 \\
% & LP++ & 76.7 $\pm$ 0.2 & 76.3 $\pm$ 0.2 & 43.8 $\pm$ 0.3 & 73.7 $\pm$ 1.0 & 79.9 $\pm$ 0.4 & 90.1 $\pm$ 0.2 & 93.5 $\pm$ 0.7 & 96.9 $\pm$ 0.2 & 96.7 $\pm$ 0.2 & 67.4 $\pm$ 0.3 & 83.1 $\pm$ 0.4 & 79.8 $\pm$ 0.4 \\
& TIP-Adapter-F & 73.8 $\pm$ 0.1 & 68.1 $\pm$ 0.1 & 30.6 $\pm$ 0.0 & 56.8 $\pm$ 0.3 & 77.4 $\pm$ 0.1 & 90.3 $\pm$ 0.0 & 93.2 $\pm$ 0.1 & 78.4 $\pm$ 0.0 & 95.2 $\pm$ 0.0 & 52.5 $\pm$ 0.0 & 73.9 $\pm$ 0.0 & 71.8 $\pm$ 0.1 \\
& GDA & 76.9 $\pm$ 0.1 & 76.2 $\pm$ 0.3 & 47.9 $\pm$ 0.6 & 80.2 $\pm$ 2.6 & 81.5 $\pm$ 1.3 & 91.2 $\pm$ 0.0 & 94.9 $\pm$ 0.2 & 98.1 $\pm$ 0.3 & 97.2 $\pm$ 0.1 & 67.6 $\pm$ 0.2 & 84.4 $\pm$ 0.3 & 81.5 $\pm$ 0.5 \\
& ProKeR & 76.1 $\pm$ 0.1 & 73.6 $\pm$ 0.1 & 42.6 $\pm$ 0.4 & 77.4 $\pm$ 2.3 & 82.9 $\pm$ 0.2 & 91.1 $\pm$ 0.0 & 94.6 $\pm$ 0.1 & 89.0 $\pm$ 0.2 & 97.0 $\pm$ 0.2 & 63.3 $\pm$ 0.6 & 80.5 $\pm$ 0.5 & 78.9 $\pm$ 0.4 \\
\cmidrule(lr){2-14}
& LoRA & 73.7 $\pm$ 0.3 & 75.6 $\pm$ 0.3 & 50.0 $\pm$ 0.6 & 75.6 $\pm$ 0.9 & 85.4 $\pm$ 0.7 & 90.6 $\pm$ 0.1 & 95.3 $\pm$ 0.2 & 97.0 $\pm$ 0.5 & 97.7 $\pm$ 0.3 & 68.8 $\pm$ 1.0 & 86.7 $\pm$ 0.2 & 81.5 $\pm$ 0.4 \\
& PiSSA & 73.2 $\pm$ 0.2 & 75.3 $\pm$ 0.0 & 49.0 $\pm$ 0.8 & 85.3 $\pm$ 0.7 & 84.3 $\pm$ 0.5 & 89.5 $\pm$ 0.3 & 94.7 $\pm$ 0.5 & 96.6 $\pm$ 0.1 & 96.8 $\pm$ 0.3 & 70.3 $\pm$ 0.3 & 87.0 $\pm$ 0.4 & 82.0 $\pm$ 0.4 \\
& GaLoRE & 73.7 $\pm$ 0.4 & 75.2 $\pm$ 0.2 & 48.3 $\pm$ 1.0 & 85.3 $\pm$ 0.8 & 83.0 $\pm$ 1.2 & 89.3 $\pm$ 0.4 & 94.0 $\pm$ 1.1 & 96.0 $\pm$ 1.5 & 97.2 $\pm$ 0.5 & 68.4 $\pm$ 0.9 & 87.0 $\pm$ 0.5 & 81.6 $\pm$ 0.8 \\
\cmidrule(lr){2-14}
& SO & 73.7 $\pm$ 0.3 & 75.3 $\pm$ 0.2 & 51.0 $\pm$ 0.4 & 79.9 $\pm$ 0.7 & 85.4 $\pm$ 0.2 & 90.4 $\pm$ 0.0 & 95.3 $\pm$ 0.2 & 97.1 $\pm$ 0.4 & 97.3 $\pm$ 0.3 & 68.8 $\pm$ 1.0 & 87.0 $\pm$ 0.6 & 81.9 $\pm$ 0.4 \\
& SignSparseSGD & 76.1 $\pm$ 0.2 & 76.2 $\pm$ 0.1 & 47.5 $\pm$ 0.1 & 85.0 $\pm$ 3.0 & 84.7 $\pm$ 0.7 & 90.6 $\pm$ 0.1 & 94.9 $\pm$ 0.0 & 96.5 $\pm$ 0.8 & 97.3 $\pm$ 0.1 & 69.4 $\pm$ 0.2 & 87.4 $\pm$ 1.0 & 82.3 $\pm$ 0.6 \\
& DynSignSparseSGD & 76.6 $\pm$ 0.1 & 77.0 $\pm$ 0.1 & 50.5 $\pm$ 0.6 & 85.4 $\pm$ 2.5 & 86.0 $\pm$ 0.9 & 91.1 $\pm$ 0.2 & 95.0 $\pm$ 0.1 & 97.0 $\pm$ 1.0 & 97.5 $\pm$ 0.3 & 70.8 $\pm$ 0.6 & 87.6 $\pm$ 0.7 & 83.1 $\pm$ 0.6 \\
\midrule
\multirow{15}{*}{8} & BitFit & 77.9 $\pm$ 0.1 & 78.1 $\pm$ 0.2 & 49.6 $\pm$ 0.6 & 84.0 $\pm$ 0.7 & 86.2 $\pm$ 0.4 & 91.4 $\pm$ 0.1 & 95.5 $\pm$ 0.4 & 96.6 $\pm$ 0.2 & 97.6 $\pm$ 0.4 & 70.7 $\pm$ 0.7 & 86.7 $\pm$ 0.3 & 83.1 $\pm$ 0.4 \\
& LayerNorm & 78.1 $\pm$ 0.1 & 76.1 $\pm$ 0.1 & 41.6 $\pm$ 0.2 & 68.2 $\pm$ 0.3 & 82.1 $\pm$ 0.1 & 91.9 $\pm$ 0.1 & 94.7 $\pm$ 0.1 & 85.9 $\pm$ 0.1 & 96.9 $\pm$ 0.2 & 61.4 $\pm$ 0.2 & 81.1 $\pm$ 0.4 & 78.0 $\pm$ 0.2 \\
& SAFT & 76.6 $\pm$ 0.1 & 76.8 $\pm$ 0.2 & 51.4 $\pm$ 0.6 & 90.3 $\pm$ 1.6 & 84.7 $\pm$ 0.4 & 91.5 $\pm$ 0.2 & 95.6 $\pm$ 0.2 & 96.9 $\pm$ 0.2 & 97.6 $\pm$ 0.2 & 72.1 $\pm$ 0.3 & 86.0 $\pm$ 0.3 & 83.6 $\pm$ 0.4 \\
\cmidrule(lr){2-14}
& CoOp & 76.2 $\pm$ 0.1 & 76.1 $\pm$ 0.2 & 48.1 $\pm$ 1.2 & 79.0 $\pm$ 1.0 & 84.7 $\pm$ 0.8 & 90.1 $\pm$ 0.3 & 94.8 $\pm$ 0.1 & 96.6 $\pm$ 0.5 & 96.4 $\pm$ 0.9 & 68.1 $\pm$ 0.5 & 85.1 $\pm$ 0.9 & 81.4 $\pm$ 0.6 \\
& KgCoOp & 76.4 $\pm$ 0.1 & 74.0 $\pm$ 0.3 & 37.9 $\pm$ 1.0 & 76.9 $\pm$ 1.7 & 79.0 $\pm$ 0.1 & 91.4 $\pm$ 0.1 & 94.7 $\pm$ 0.1 & 85.1 $\pm$ 0.8 & 96.6 $\pm$ 0.1 & 64.3 $\pm$ 0.6 & 80.6 $\pm$ 0.8 & 77.9 $\pm$ 0.5 \\
& ProGrad & 75.7 $\pm$ 0.8 & 76.5 $\pm$ 0.3 & 47.0 $\pm$ 0.8 & 82.0 $\pm$ 1.7 & 84.8 $\pm$ 0.2 & 90.1 $\pm$ 1.2 & 95.1 $\pm$ 0.3 & 95.0 $\pm$ 0.4 & 96.7 $\pm$ 0.5 & 67.9 $\pm$ 0.4 & 84.2 $\pm$ 0.4 & 81.4 $\pm$ 0.6 \\
\cmidrule(lr){2-14}
% & Adapter & 76.9 $\pm$ 0.1 & 76.2 $\pm$ 0.1 & 39.6 $\pm$ 0.3 & 64.6 $\pm$ 0.5 & 82.4 $\pm$ 0.2 & 91.3 $\pm$ 0.0 & 94.1 $\pm$ 0.3 & 90.4 $\pm$ 0.4 & 97.1 $\pm$ 0.3 & 61.6 $\pm$ 0.4 & 82.7 $\pm$ 0.2 & 77.9 $\pm$ 0.3 \\
% & LP++ & 77.5 $\pm$ 0.2 & 78.1 $\pm$ 0.3 & 47.6 $\pm$ 0.6 & 81.7 $\pm$ 1.2 & 84.0 $\pm$ 0.3 & 90.8 $\pm$ 0.1 & 93.6 $\pm$ 0.5 & 97.4 $\pm$ 0.2 & 96.8 $\pm$ 0.4 & 71.6 $\pm$ 0.5 & 85.3 $\pm$ 0.3 & 82.2 $\pm$ 0.4 \\
& TIP-Adapter-F & 74.6 $\pm$ 0.1 & 69.7 $\pm$ 0.1 & 32.9 $\pm$ 0.5 & 58.5 $\pm$ 0.2 & 80.0 $\pm$ 0.1 & 90.4 $\pm$ 0.0 & 93.2 $\pm$ 0.2 & 86.1 $\pm$ 0.3 & 95.4 $\pm$ 0.0 & 52.8 $\pm$ 0.1 & 74.2 $\pm$ 0.1 & 73.4 $\pm$ 0.1 \\
& GDA & 77.2 $\pm$ 0.2 & 76.8 $\pm$ 0.4 & 53.3 $\pm$ 0.8 & 87.2 $\pm$ 1.2 & 85.0 $\pm$ 0.2 & 91.0 $\pm$ 0.1 & 94.7 $\pm$ 0.1 & 98.9 $\pm$ 0.3 & 96.9 $\pm$ 0.4 & 72.0 $\pm$ 0.7 & 85.9 $\pm$ 0.7 & 83.5 $\pm$ 0.5 \\
& ProKeR & 76.5 $\pm$ 0.1 & 76.6 $\pm$ 0.2 & 48.5 $\pm$ 1.1 & 83.7 $\pm$ 0.5 & 86.4 $\pm$ 0.1 & 91.6 $\pm$ 0.1 & 94.9 $\pm$ 0.3 & 92.5 $\pm$ 0.2 & 97.1 $\pm$ 0.4 & 68.1 $\pm$ 0.8 & 83.3 $\pm$ 0.3 & 81.7 $\pm$ 0.4 \\
\cmidrule(lr){2-14}
& LoRA & 75.1 $\pm$ 0.1 & 77.0 $\pm$ 0.2 & 56.9 $\pm$ 0.8 & 86.5 $\pm$ 1.9 & 88.4 $\pm$ 0.2 & 90.5 $\pm$ 0.2 & 95.5 $\pm$ 0.2 & 98.0 $\pm$ 0.1 & 97.0 $\pm$ 0.5 & 72.4 $\pm$ 0.2 & 88.8 $\pm$ 0.5 & 84.2 $\pm$ 0.5 \\
& PiSSA & 74.1 $\pm$ 0.2 & 76.7 $\pm$ 0.4 & 55.7 $\pm$ 0.7 & 89.2 $\pm$ 1.8 & 88.1 $\pm$ 0.3 & 89.8 $\pm$ 0.3 & 94.7 $\pm$ 0.3 & 97.5 $\pm$ 0.1 & 96.9 $\pm$ 0.5 & 72.2 $\pm$ 0.5 & 88.6 $\pm$ 0.6 & 83.9 $\pm$ 0.5 \\
& GaLoRE & 74.7 $\pm$ 0.1 & 76.7 $\pm$ 0.2 & 54.5 $\pm$ 1.5 & 85.7 $\pm$ 4.6 & 86.8 $\pm$ 0.5 & 89.9 $\pm$ 0.2 & 94.9 $\pm$ 0.4 & 97.0 $\pm$ 0.6 & 96.9 $\pm$ 1.0 & 72.2 $\pm$ 0.7 & 88.3 $\pm$ 1.0 & 83.4 $\pm$ 1.0 \\
\cmidrule(lr){2-14}
& SO & 75.2 $\pm$ 0.3 & 77.0 $\pm$ 0.2 & 57.1 $\pm$ 0.6 & 88.2 $\pm$ 1.8 & 88.4 $\pm$ 0.4 & 90.6 $\pm$ 0.2 & 95.7 $\pm$ 0.1 & 98.1 $\pm$ 0.1 & 97.1 $\pm$ 0.6 & 72.5 $\pm$ 0.3 & 88.8 $\pm$ 0.2 & 84.4 $\pm$ 0.4 \\
& SignSparseSGD & 76.8 $\pm$ 0.2 & 77.5 $\pm$ 0.0 & 56.0 $\pm$ 1.2 & 89.2 $\pm$ 2.7 & 87.6 $\pm$ 0.3 & 90.7 $\pm$ 0.2 & 95.4 $\pm$ 0.2 & 97.3 $\pm$ 0.3 & 97.5 $\pm$ 0.6 & 72.0 $\pm$ 0.6 & 88.6 $\pm$ 0.5 & 84.4 $\pm$ 0.6 \\
& DynSignSparseSGD & 77.3 $\pm$ 0.1 & 78.3 $\pm$ 0.1 & 58.7 $\pm$ 0.6 & 88.9 $\pm$ 2.7 & 88.8 $\pm$ 0.2 & 91.2 $\pm$ 0.1 & 95.5 $\pm$ 0.3 & 97.8 $\pm$ 0.5 & 97.4 $\pm$ 0.5 & 73.1 $\pm$ 0.4 & 89.2 $\pm$ 0.4 & 85.1 $\pm$ 0.5 \\
\midrule
\multirow{15}{*}{16} & BitFit & 79.0 $\pm$ 0.1 & 79.5 $\pm$ 0.2 & 55.0 $\pm$ 0.4 & 90.4 $\pm$ 0.8 & 88.1 $\pm$ 0.2 & 91.7 $\pm$ 0.0 & 95.8 $\pm$ 0.2 & 97.9 $\pm$ 0.3 & 97.6 $\pm$ 0.2 & 75.0 $\pm$ 0.5 & 88.6 $\pm$ 0.5 & 85.3 $\pm$ 0.3 \\
& LayerNorm & 78.9 $\pm$ 0.0 & 77.8 $\pm$ 0.1 & 45.5 $\pm$ 0.2 & 72.9 $\pm$ 0.7 & 83.9 $\pm$ 0.2 & 92.0 $\pm$ 0.1 & 95.2 $\pm$ 0.2 & 91.7 $\pm$ 0.2 & 97.5 $\pm$ 0.1 & 66.3 $\pm$ 0.4 & 84.2 $\pm$ 0.4 & 80.5 $\pm$ 0.2 \\
& SAFT & 77.6 $\pm$ 0.2 & 77.7 $\pm$ 0.2 & 59.0 $\pm$ 0.4 & 94.1 $\pm$ 0.8 & 86.5 $\pm$ 0.2 & 91.5 $\pm$ 0.2 & 95.8 $\pm$ 0.1 & 97.9 $\pm$ 0.3 & 97.8 $\pm$ 0.2 & 74.9 $\pm$ 0.2 & 87.8 $\pm$ 0.2 & 85.5 $\pm$ 0.3 \\
\cmidrule(lr){2-14}
& CoOp & 77.8 $\pm$ 0.1 & 78.0 $\pm$ 0.1 & 53.1 $\pm$ 0.6 & 85.8 $\pm$ 0.8 & 87.4 $\pm$ 0.2 & 90.3 $\pm$ 0.4 & 94.6 $\pm$ 0.1 & 98.2 $\pm$ 0.4 & 97.1 $\pm$ 0.2 & 72.9 $\pm$ 1.2 & 86.7 $\pm$ 0.2 & 83.8 $\pm$ 0.4 \\
& KgCoOp & 76.7 $\pm$ 0.1 & 74.9 $\pm$ 0.2 & 39.9 $\pm$ 0.6 & 78.8 $\pm$ 2.1 & 79.4 $\pm$ 0.3 & 91.4 $\pm$ 0.0 & 94.8 $\pm$ 0.2 & 87.9 $\pm$ 0.6 & 96.9 $\pm$ 0.2 & 67.1 $\pm$ 0.1 & 81.9 $\pm$ 0.5 & 79.1 $\pm$ 0.4 \\
& ProGrad & 68.7 $\pm$ 11.0 & 78.1 $\pm$ 0.3 & 51.2 $\pm$ 0.1 & 85.4 $\pm$ 0.9 & 86.9 $\pm$ 0.4 & 90.8 $\pm$ 0.4 & 94.9 $\pm$ 0.2 & 96.7 $\pm$ 0.7 & 97.4 $\pm$ 0.1 & 71.5 $\pm$ 1.0 & 86.7 $\pm$ 0.4 & 82.6 $\pm$ 1.4 \\
\cmidrule(lr){2-14}
% & Adapter & 77.8 $\pm$ 0.1 & 77.9 $\pm$ 0.2 & 43.3 $\pm$ 0.4 & 72.2 $\pm$ 0.8 & 84.4 $\pm$ 0.0 & 91.6 $\pm$ 0.1 & 95.0 $\pm$ 0.1 & 96.4 $\pm$ 0.3 & 97.6 $\pm$ 0.1 & 68.6 $\pm$ 0.4 & 85.6 $\pm$ 0.2 & 80.9 $\pm$ 0.2 \\
% & LP++ & 78.4 $\pm$ 0.0 & 79.5 $\pm$ 0.1 & 52.2 $\pm$ 0.3 & 84.2 $\pm$ 0.8 & 86.4 $\pm$ 0.4 & 91.2 $\pm$ 0.1 & 94.7 $\pm$ 0.3 & 98.3 $\pm$ 0.1 & 97.4 $\pm$ 0.2 & 75.7 $\pm$ 0.7 & 87.2 $\pm$ 0.6 & 84.1 $\pm$ 0.3 \\
& TIP-Adapter-F & 76.0 $\pm$ 0.0 & 73.5 $\pm$ 0.1 & 41.6 $\pm$ 0.2 & 61.9 $\pm$ 0.3 & 84.2 $\pm$ 0.1 & 90.7 $\pm$ 0.0 & 93.7 $\pm$ 0.0 & 96.0 $\pm$ 0.1 & 95.5 $\pm$ 0.1 & 53.8 $\pm$ 0.1 & 75.7 $\pm$ 0.2 & 76.6 $\pm$ 0.1 \\
& GDA & 78.5 $\pm$ 0.0 & 79.9 $\pm$ 0.1 & 58.0 $\pm$ 0.5 & 90.0 $\pm$ 1.6 & 88.0 $\pm$ 0.2 & 91.7 $\pm$ 0.0 & 95.2 $\pm$ 0.8 & 99.4 $\pm$ 0.1 & 97.5 $\pm$ 0.3 & 75.5 $\pm$ 0.4 & 87.9 $\pm$ 0.3 & 85.6 $\pm$ 0.4 \\
& ProKeR & 78.0 $\pm$ 0.0 & 79.2 $\pm$ 0.1 & 56.1 $\pm$ 0.2 & 89.6 $\pm$ 1.0 & 88.9 $\pm$ 0.2 & 91.9 $\pm$ 0.1 & 95.1 $\pm$ 0.2 & 96.6 $\pm$ 0.1 & 97.5 $\pm$ 0.1 & 75.2 $\pm$ 0.2 & 86.7 $\pm$ 0.4 & 85.0 $\pm$ 0.2 \\
\cmidrule(lr){2-14}
& LoRA & 76.9 $\pm$ 0.2 & 78.7 $\pm$ 0.0 & 64.5 $\pm$ 1.0 & 92.0 $\pm$ 0.1 & 90.2 $\pm$ 0.3 & 90.8 $\pm$ 0.2 & 95.5 $\pm$ 0.0 & 98.9 $\pm$ 0.0 & 97.5 $\pm$ 0.3 & 76.2 $\pm$ 0.7 & 90.4 $\pm$ 0.1 & 86.5 $\pm$ 0.3 \\
& PiSSA & 75.5 $\pm$ 0.1 & 78.1 $\pm$ 0.2 & 63.4 $\pm$ 0.4 & 92.5 $\pm$ 0.5 & 89.9 $\pm$ 0.3 & 89.8 $\pm$ 0.1 & 95.0 $\pm$ 0.1 & 99.1 $\pm$ 0.2 & 97.2 $\pm$ 0.4 & 76.2 $\pm$ 0.7 & 90.3 $\pm$ 0.3 & 86.1 $\pm$ 0.3 \\
& GaLoRE & 75.9 $\pm$ 0.2 & 78.5 $\pm$ 0.2 & 61.4 $\pm$ 0.7 & 93.2 $\pm$ 1.0 & 89.0 $\pm$ 0.3 & 89.6 $\pm$ 0.3 & 94.9 $\pm$ 0.4 & 97.9 $\pm$ 0.6 & 96.8 $\pm$ 0.2 & 76.3 $\pm$ 0.4 & 89.0 $\pm$ 0.2 & 85.7 $\pm$ 0.4 \\
\cmidrule(lr){2-14}
& SO & 76.9 $\pm$ 0.1 & 78.5 $\pm$ 0.3 & 64.6 $\pm$ 0.5 & 92.5 $\pm$ 0.2 & 90.7 $\pm$ 0.1 & 90.8 $\pm$ 0.2 & 95.6 $\pm$ 0.1 & 98.9 $\pm$ 0.1 & 97.3 $\pm$ 0.1 & 76.6 $\pm$ 0.5 & 90.9 $\pm$ 0.2 & 86.6 $\pm$ 0.2 \\
& SignSparseSGD & 78.0 $\pm$ 0.1 & 79.3 $\pm$ 0.3 & 61.6 $\pm$ 0.9 & 92.7 $\pm$ 1.5 & 89.5 $\pm$ 0.3 & 90.8 $\pm$ 0.2 & 95.4 $\pm$ 0.3 & 98.5 $\pm$ 0.1 & 97.7 $\pm$ 0.0 & 76.5 $\pm$ 0.8 & 90.1 $\pm$ 0.2 & 86.4 $\pm$ 0.4 \\
& DynSignSparseSGD & 78.3 $\pm$ 0.1 & 79.6 $\pm$ 0.1 & 65.0 $\pm$ 0.3 & 93.4 $\pm$ 0.9 & 90.3 $\pm$ 0.1 & 91.2 $\pm$ 0.0 & 95.6 $\pm$ 0.1 & 98.9 $\pm$ 0.2 & 97.4 $\pm$ 0.3 & 77.8 $\pm$ 0.7 & 91.0 $\pm$ 0.2 & 87.1 $\pm$ 0.3 \\
\bottomrule
\end{tabular}
\end{adjustbox}
\end{table}

Tables~\ref{tab:few_shot_ViTB32_with_aug}, \ref{tab:few_shot_ViTB16_with_aug}, and \ref{tab:few_shot_ViTL14_with_aug} report the all-to-all few-shot results for CLIP with ViT-B/32, ViT-B/16, and ViT-L/14 backbones. DynSignSparseSGD attains the highest average accuracy at every shot count for all 3 backbones. On ViT-B/32, the average accuracy increases from 67.9\% at 1-shot to 79.0\% at 16-shot. On ViT-B/16, it reaches 72.7\%, 75.4\%, 77.8\%, 80.0\%, and 82.8\% for 1, 2, 4, 8, and 16 shots, respectively. On ViT-L/14, it attains 78.5\%, 81.1\%, 83.1\%, 85.1\%, and 87.1\% at the same shot counts. Therefore, the benefit of the dynamic sign-sparse update is not limited to the 1-shot regime emphasized in the main paper. It persists across the full range $K \in \{1,2,4,8,16\}$ and across all 15 backbone--shot settings. 

DynSignSparseSGD also improves over the static SignSparseSGD variant. Averaged over shot counts, the gain is approximately 1.6 points on ViT-B/32, 1.2 points on ViT-B/16, and 0.7 points on ViT-L/14. This pattern indicates that updating the active set during training is beneficial. The effect is more pronounced on the smaller backbones, where a fixed mask appears more restrictive. SignSparseSGD remains competitive and is often close to the strongest baseline. However, DynSignSparseSGD is consistently the stronger sign-sparse variant.

At the dataset level, DynSignSparseSGD is most consistently strong on SUN397, FGVC-Aircraft, Stanford-Cars, DTD, and UCF101. Across the 15 backbone--shot settings, it is best or tied for best in 10, 10, 13, 13, and 15 settings on these datasets, respectively. It also leads several EuroSAT settings, especially in the lower-shot ViT-B/32 and ViT-B/16 regimes. The updated training-free baselines exhibit complementary strengths. GDA is particularly strong on Flowers102, where it achieves the best result in 10 of the 15 settings, and it also leads SUN397 at 4 shots with ViT-B/32 and at 16 shots with ViT-L/14. ProKeR obtains the best SUN397 result at 16 shots for both ViT-B/32 and ViT-B/16. These dataset-specific wins do not translate into the strongest overall average, since both GDA and ProKeR remain below DynSignSparseSGD in every backbone--shot setting. The main remaining exceptions are ImageNet, where BitFit or LayerNorm is strongest throughout; Food101, where LayerNorm usually leads; Oxford-Pets, where SO is frequently strongest; and several EuroSAT and Caltech101 settings, where SAFT, PiSSA, or other baselines retain an advantage. Overall, these results show that a simple, state-free, dynamic sign-sparse optimizer can match or outperform specialized PEFT methods, optimization-based baselines, and strong training-free alternatives across a broad range of few-shot settings under a fixed protocol.

\subsection{Base-to-New Few-Shot Evaluation}
\label{app:b2n_additional}
\suppanchor{peft-b2n}

\begin{table}[p]
  \def\arraystretch{0.9}
  \centering
  \scriptsize
  \caption{Base-to-new 1-shot classification with CLIP(ViT-B/32). Top-1 accuracy (\%) averaged over 3 fixed random seeds.}
  \label{tab:b2n_peft_1shot_vit-b-32_with_aug}
  
  \begin{minipage}[t]{0.32\textwidth}
  \centering
\caption*{\textbf{Average across datasets.}}\vspace{-1em}
\setlength{\tabcolsep}{2.5pt}
\resizebox{\linewidth}{!}{%
\begin{tabular}{@{}lcc|c@{}}
\toprule
\textbf{Method} & \textbf{Base} & \textbf{Novel} & \textbf{HM} \\
\midrule
CLIP & 60.19 & 60.19 & 60.19 \\ 
BitFit & 75.17 & 70.06 & 72.27 \\
LayerNorm & 71.01 & 69.35 & 69.92 \\
SAFT & 74.73 & 68.83 & 71.41 \\
CoOp & 72.05 & 64.90 & 68.00 \\
KgCoOp & 73.26 & 68.58 & 70.44 \\
ProGrad & 67.25 & 39.85 & 47.36 \\
LoRA & 69.85 & 55.77 & 60.45 \\
PiSSA & 72.59 & 44.43 & 52.30 \\
GaLoRE & 72.52 & 67.47 & 69.71 \\
SO & 75.39 & 70.48 & 72.62 \\
\cmidrule(lr){1-4}
SignSparseSGD & 74.43 & 68.43 & 71.03 \\
DynSignSparseSGD & 76.19 & 70.24 & 72.84 \\
\bottomrule
\end{tabular}%
}
  \vspace{0.5em}
  \end{minipage}%
  \hfill
  \begin{minipage}[t]{0.32\textwidth}
  \centering
\caption*{\textbf{ImageNet}}\vspace{-1em}
\setlength{\tabcolsep}{2.5pt}
\resizebox{\linewidth}{!}{%
\begin{tabular}{@{}lcc|c@{}}
\toprule
\textbf{Method} & \textbf{Base} & \textbf{Novel} & \textbf{HM} \\
\midrule
CLIP & 62.04 & 62.04 & 62.04 \\ 
BitFit & 73.00 & 72.36 & 72.68 \\
LayerNorm & 72.44 & 72.55 & 72.49 \\
SAFT & 71.95 & 71.41 & 71.68 \\
CoOp & 70.23 & 68.97 & 69.59 \\
KgCoOp & 72.55 & 71.27 & 71.90 \\
ProGrad & 62.12 & 51.16 & 56.11 \\
LoRA & 65.80 & 61.78 & 63.72 \\
PiSSA & 64.48 & 52.83 & 58.07 \\
GaLoRE & 67.20 & 66.68 & 66.94 \\
SO & 72.72 & 72.22 & 72.47 \\
\cmidrule(lr){1-4}
SignSparseSGD & 70.44 & 69.50 & 69.96 \\
DynSignSparseSGD & 72.19 & 72.00 & 72.09 \\
\bottomrule
\end{tabular}%
}
  \vspace{0.5em}
  \end{minipage}%
  \hfill
  \begin{minipage}[t]{0.32\textwidth}
  \centering
\caption*{\textbf{Caltech101}}\vspace{-1em}
\setlength{\tabcolsep}{2.5pt}
\resizebox{\linewidth}{!}{%
\begin{tabular}{@{}lcc|c@{}}
\toprule
\textbf{Method} & \textbf{Base} & \textbf{Novel} & \textbf{HM} \\
\midrule
CLIP & 91.40 & 91.40 & 91.40 \\ 
BitFit & 96.77 & 95.51 & 96.13 \\
LayerNorm & 95.88 & 95.53 & 95.70 \\
SAFT & 96.45 & 95.46 & 95.94 \\
CoOp & 93.86 & 91.40 & 92.53 \\
KgCoOp & 95.43 & 95.44 & 95.43 \\
ProGrad & 92.27 & 74.86 & 82.56 \\
LoRA & 95.34 & 88.75 & 91.92 \\
PiSSA & 95.59 & 82.85 & 88.76 \\
GaLoRE & 95.81 & 95.29 & 95.54 \\
SO & 96.45 & 95.64 & 96.03 \\
\cmidrule(lr){1-4}
SignSparseSGD & 96.54 & 95.49 & 96.01 \\
DynSignSparseSGD & 96.63 & 95.67 & 96.14 \\
\bottomrule
\end{tabular}%
}
  \vspace{0.5em}
  \end{minipage}%
  \par\vspace{0.45em}
  \begin{minipage}[t]{0.32\textwidth}
  \centering
\caption*{\textbf{Oxford Flowers}}\vspace{-1em}
\setlength{\tabcolsep}{2.5pt}
\resizebox{\linewidth}{!}{%
\begin{tabular}{@{}lcc|c@{}}
\toprule
\textbf{Method} & \textbf{Base} & \textbf{Novel} & \textbf{HM} \\
\midrule
CLIP & 64.03 & 64.03 & 64.03 \\ 
BitFit & 77.43 & 74.27 & 75.74 \\
LayerNorm & 72.15 & 75.08 & 73.53 \\
SAFT & 82.57 & 69.13 & 75.13 \\
CoOp & 79.17 & 67.87 & 72.94 \\
KgCoOp & 72.87 & 74.49 & 73.60 \\
ProGrad & 76.23 & 47.72 & 58.60 \\
LoRA & 70.19 & 57.78 & 63.20 \\
PiSSA & 82.98 & 43.00 & 56.46 \\
GaLoRE & 84.01 & 68.74 & 75.59 \\
SO & 78.32 & 73.90 & 75.95 \\
\cmidrule(lr){1-4}
SignSparseSGD & 83.75 & 71.20 & 76.83 \\
DynSignSparseSGD & 84.43 & 73.17 & 78.34 \\
\bottomrule
\end{tabular}%
}
  \vspace{0.5em}
  \end{minipage}%
  \hfill
  \begin{minipage}[t]{0.32\textwidth}
  \centering
\caption*{\textbf{Oxford Pets}}\vspace{-1em}
\setlength{\tabcolsep}{2.5pt}
\resizebox{\linewidth}{!}{%
\begin{tabular}{@{}lcc|c@{}}
\toprule
\textbf{Method} & \textbf{Base} & \textbf{Novel} & \textbf{HM} \\
\midrule
CLIP & 85.06 & 85.06 & 85.06 \\ 
BitFit & 93.55 & 89.59 & 91.52 \\
LayerNorm & 93.34 & 88.53 & 90.83 \\
SAFT & 91.35 & 89.43 & 90.37 \\
CoOp & 91.84 & 87.06 & 89.37 \\
KgCoOp & 94.56 & 88.03 & 91.17 \\
ProGrad & 90.12 & 50.73 & 64.87 \\
LoRA & 91.91 & 77.78 & 84.13 \\
PiSSA & 90.79 & 61.49 & 73.06 \\
GaLoRE & 88.17 & 88.88 & 88.52 \\
SO & 93.76 & 89.89 & 91.77 \\
\cmidrule(lr){1-4}
SignSparseSGD & 92.54 & 89.50 & 90.99 \\
DynSignSparseSGD & 92.53 & 90.07 & 91.26 \\
\bottomrule
\end{tabular}%
}
  \vspace{0.5em}
  \end{minipage}%
  \hfill
  \begin{minipage}[t]{0.32\textwidth}
  \centering
\caption*{\textbf{Stanford Cars}}\vspace{-1em}
\setlength{\tabcolsep}{2.5pt}
\resizebox{\linewidth}{!}{%
\begin{tabular}{@{}lcc|c@{}}
\toprule
\textbf{Method} & \textbf{Base} & \textbf{Novel} & \textbf{HM} \\
\midrule
CLIP & 60.14 & 60.14 & 60.14 \\ 
BitFit & 73.52 & 69.72 & 71.55 \\
LayerNorm & 72.22 & 69.30 & 70.72 \\
SAFT & 72.01 & 66.62 & 69.20 \\
CoOp & 70.66 & 66.42 & 68.45 \\
KgCoOp & 70.82 & 69.11 & 69.94 \\
ProGrad & 63.09 & 39.03 & 48.12 \\
LoRA & 66.70 & 55.01 & 60.23 \\
PiSSA & 69.70 & 42.02 & 52.41 \\
GaLoRE & 70.46 & 64.60 & 67.39 \\
SO & 74.09 & 69.96 & 71.95 \\
\cmidrule(lr){1-4}
SignSparseSGD & 72.53 & 65.89 & 69.04 \\
DynSignSparseSGD & 74.52 & 68.28 & 71.26 \\
\bottomrule
\end{tabular}%
}
  \vspace{0.5em}
  \end{minipage}
\end{table}

\begin{table}[p]
  \ContinuedFloat
  \def\arraystretch{0.9}
  \centering
  \scriptsize
  \caption[]{Base-to-new 1-shot classification with CLIP(ViT-B/32). Top-1 accuracy (\%) averaged over 3 fixed random seeds. (continued)}
  \begin{minipage}[t]{0.32\textwidth}
  \centering
\caption*{\textbf{Food 101}}\vspace{-1em}
\setlength{\tabcolsep}{2.5pt}
\resizebox{\linewidth}{!}{%
\begin{tabular}{@{}lcc|c@{}}
\toprule
\textbf{Method} & \textbf{Base} & \textbf{Novel} & \textbf{HM} \\
\midrule
CLIP & 78.86 & 78.86 & 78.86 \\ 
BitFit & 84.39 & 85.19 & 84.78 \\
LayerNorm & 85.16 & 85.40 & 85.27 \\
SAFT & 82.42 & 83.96 & 83.18 \\
CoOp & 83.42 & 82.28 & 82.85 \\
KgCoOp & 85.55 & 85.26 & 85.39 \\
ProGrad & 75.45 & 51.06 & 58.65 \\
LoRA & 80.52 & 78.66 & 79.57 \\
PiSSA & 79.98 & 68.06 & 73.52 \\
GaLoRE & 80.75 & 82.66 & 81.69 \\
SO & 84.98 & 85.61 & 85.28 \\
\cmidrule(lr){1-4}
SignSparseSGD & 83.38 & 84.64 & 84.00 \\
DynSignSparseSGD & 84.34 & 85.35 & 84.84 \\
\bottomrule
\end{tabular}%
}
  \vspace{0.5em}
  \end{minipage}%
  \hfill
  \begin{minipage}[t]{0.32\textwidth}
  \centering
\caption*{\textbf{FGVC Aircraft}}\vspace{-1em}
\setlength{\tabcolsep}{2.5pt}
\resizebox{\linewidth}{!}{%
\begin{tabular}{@{}lcc|c@{}}
\toprule
\textbf{Method} & \textbf{Base} & \textbf{Novel} & \textbf{HM} \\
\midrule
CLIP & 18.06 & 18.06 & 18.06 \\ 
BitFit & 30.22 & 27.19 & 28.62 \\
LayerNorm & 27.04 & 28.70 & 27.81 \\
SAFT & 29.60 & 25.45 & 27.35 \\
CoOp & 28.14 & 20.65 & 23.73 \\
KgCoOp & 29.50 & 27.34 & 28.37 \\
ProGrad & 19.19 & 3.60 & 6.04 \\
LoRA & 25.46 & 14.97 & 18.74 \\
PiSSA & 28.62 & 6.91 & 11.03 \\
GaLoRE & 29.10 & 23.33 & 25.88 \\
SO & 32.12 & 28.20 & 30.04 \\
\cmidrule(lr){1-4}
SignSparseSGD & 28.52 & 24.39 & 26.29 \\
DynSignSparseSGD & 30.58 & 26.11 & 28.16 \\
\bottomrule
\end{tabular}%
}
  \vspace{0.5em}
  \end{minipage}%
  \hfill
  \begin{minipage}[t]{0.32\textwidth}
  \centering
\caption*{\textbf{SUN 397}}\vspace{-1em}
\setlength{\tabcolsep}{2.5pt}
\resizebox{\linewidth}{!}{%
\begin{tabular}{@{}lcc|c@{}}
\toprule
\textbf{Method} & \textbf{Base} & \textbf{Novel} & \textbf{HM} \\
\midrule
CLIP & 62.06 & 62.06 & 62.06 \\ 
BitFit & 76.87 & 74.28 & 75.54 \\
LayerNorm & 74.24 & 73.39 & 73.80 \\
SAFT & 76.08 & 73.03 & 74.51 \\
CoOp & 73.26 & 68.45 & 70.76 \\
KgCoOp & 74.62 & 72.26 & 73.42 \\
ProGrad & 66.26 & 48.71 & 55.93 \\
LoRA & 69.68 & 61.57 & 65.37 \\
PiSSA & 69.67 & 53.57 & 60.57 \\
GaLoRE & 72.69 & 70.70 & 71.67 \\
SO & 76.85 & 74.02 & 75.40 \\
\cmidrule(lr){1-4}
SignSparseSGD & 75.24 & 72.40 & 73.78 \\
DynSignSparseSGD & 76.77 & 73.77 & 75.23 \\
\bottomrule
\end{tabular}%
}
  \vspace{0.5em}
  \end{minipage}%
  \par\vspace{0.45em}
  \begin{minipage}[t]{0.32\textwidth}
  \centering
\caption*{\textbf{DTD}}\vspace{-1em}
\setlength{\tabcolsep}{2.5pt}
\resizebox{\linewidth}{!}{%
\begin{tabular}{@{}lcc|c@{}}
\toprule
\textbf{Method} & \textbf{Base} & \textbf{Novel} & \textbf{HM} \\
\midrule
CLIP & 42.97 & 42.97 & 42.97 \\ 
BitFit & 65.22 & 49.19 & 56.07 \\
LayerNorm & 56.24 & 51.66 & 53.67 \\
SAFT & 62.52 & 48.88 & 54.82 \\
CoOp & 57.09 & 42.59 & 48.75 \\
KgCoOp & 61.71 & 51.16 & 55.81 \\
ProGrad & 54.79 & 25.54 & 33.95 \\
LoRA & 58.78 & 37.38 & 45.57 \\
PiSSA & 60.55 & 25.12 & 35.44 \\
GaLoRE & 61.03 & 46.91 & 52.90 \\
SO & 65.02 & 51.66 & 57.55 \\
\cmidrule(lr){1-4}
SignSparseSGD & 64.49 & 49.19 & 55.74 \\
DynSignSparseSGD & 64.65 & 51.50 & 57.30 \\
\bottomrule
\end{tabular}%
}
  \vspace{0.5em}
  \end{minipage}%
  \hfill
  \begin{minipage}[t]{0.32\textwidth}
  \centering
\caption*{\textbf{EuroSAT}}\vspace{-1em}
\setlength{\tabcolsep}{2.5pt}
\resizebox{\linewidth}{!}{%
\begin{tabular}{@{}lcc|c@{}}
\toprule
\textbf{Method} & \textbf{Base} & \textbf{Novel} & \textbf{HM} \\
\midrule
CLIP & 35.81 & 35.81 & 35.81 \\ 
BitFit & 78.99 & 57.32 & 66.10 \\
LayerNorm & 62.35 & 48.74 & 53.68 \\
SAFT & 79.50 & 57.80 & 66.75 \\
CoOp & 69.63 & 48.09 & 56.56 \\
KgCoOp & 74.28 & 46.67 & 56.40 \\
ProGrad & 69.81 & 0.58 & 1.14 \\
LoRA & 74.00 & 16.40 & 26.01 \\
PiSSA & 79.15 & 3.15 & 5.89 \\
GaLoRE & 71.68 & 59.15 & 64.81 \\
SO & 78.02 & 58.24 & 66.24 \\
\cmidrule(lr){1-4}
SignSparseSGD & 73.65 & 55.10 & 62.41 \\
DynSignSparseSGD & 83.35 & 61.01 & 69.98 \\
\bottomrule
\end{tabular}%
}
  \vspace{0.5em}
  \end{minipage}%
  \hfill
  \begin{minipage}[t]{0.32\textwidth}
  \centering
\caption*{\textbf{UCF101}}\vspace{-1em}
\setlength{\tabcolsep}{2.5pt}
\resizebox{\linewidth}{!}{%
\begin{tabular}{@{}lcc|c@{}}
\toprule
\textbf{Method} & \textbf{Base} & \textbf{Novel} & \textbf{HM} \\
\midrule
CLIP & 61.64 & 61.64 & 61.64 \\ 
BitFit & 76.88 & 76.03 & 76.19 \\
LayerNorm & 70.02 & 74.02 & 71.65 \\
SAFT & 77.58 & 75.96 & 76.57 \\
CoOp & 75.29 & 70.13 & 72.44 \\
KgCoOp & 73.94 & 73.35 & 73.38 \\
ProGrad & 70.42 & 45.40 & 54.96 \\
LoRA & 69.98 & 63.43 & 66.44 \\
PiSSA & 76.94 & 49.72 & 60.10 \\
GaLoRE & 76.87 & 75.25 & 75.90 \\
SO & 76.92 & 75.95 & 76.16 \\
\cmidrule(lr){1-4}
SignSparseSGD & 77.60 & 75.43 & 76.32 \\
DynSignSparseSGD & 78.12 & 75.72 & 76.67 \\
\bottomrule
\end{tabular}%
}
  \vspace{0.5em}
  \end{minipage}
\end{table}

\begin{table}[p]
  \def\arraystretch{0.9}
  \centering
  \scriptsize
  \caption{Base-to-new 2-shot classification with CLIP(ViT-B/32). Top-1 accuracy (\%) averaged over 3 fixed random seeds.}
  \label{tab:b2n_peft_2shot_vit-b-32_with_aug}
  \begin{minipage}[t]{0.32\textwidth}
  \centering
\caption*{\textbf{Average across datasets.}}\vspace{-1em}
\setlength{\tabcolsep}{2.5pt}
\resizebox{\linewidth}{!}{%
\begin{tabular}{@{}lcc|c@{}}
\toprule
\textbf{Method} & \textbf{Base} & \textbf{Novel} & \textbf{HM} \\
\midrule
CLIP & 60.19 & 60.19 & 60.19 \\ 
BitFit & 76.64 & 71.40 & 73.73 \\
LayerNorm & 71.58 & 69.64 & 70.35 \\
SAFT & 77.53 & 70.26 & 73.47 \\
CoOp & 74.59 & 65.08 & 68.89 \\
KgCoOp & 73.44 & 68.58 & 70.55 \\
ProGrad & 70.82 & 43.36 & 51.13 \\
LoRA & 72.62 & 53.45 & 59.78 \\
PiSSA & 75.68 & 40.49 & 49.80 \\
GaLoRE & 75.34 & 67.81 & 71.05 \\
SO & 77.26 & 71.08 & 73.81 \\
\cmidrule(lr){1-4}
SignSparseSGD & 77.42 & 69.42 & 72.89 \\
DynSignSparseSGD & 78.42 & 70.77 & 74.13 \\
\bottomrule
\end{tabular}%
}
  \vspace{0.5em}
  \end{minipage}%
  \hfill
  \begin{minipage}[t]{0.32\textwidth}
  \centering
\caption*{\textbf{ImageNet}}\vspace{-1em}
\setlength{\tabcolsep}{2.5pt}
\resizebox{\linewidth}{!}{%
\begin{tabular}{@{}lcc|c@{}}
\toprule
\textbf{Method} & \textbf{Base} & \textbf{Novel} & \textbf{HM} \\
\midrule
CLIP & 62.04 & 62.04 & 62.04 \\ 
BitFit & 73.38 & 72.60 & 72.98 \\
LayerNorm & 72.98 & 72.93 & 72.95 \\
SAFT & 72.25 & 71.03 & 71.63 \\
CoOp & 71.31 & 69.59 & 70.44 \\
KgCoOp & 72.56 & 71.27 & 71.91 \\
ProGrad & 63.18 & 44.17 & 51.98 \\
LoRA & 67.54 & 55.57 & 60.97 \\
PiSSA & 66.29 & 46.93 & 54.95 \\
GaLoRE & 67.15 & 66.21 & 66.67 \\
SO & 72.40 & 71.25 & 71.82 \\
\cmidrule(lr){1-4}
SignSparseSGD & 71.00 & 69.29 & 70.13 \\
DynSignSparseSGD & 72.44 & 71.75 & 72.09 \\
\bottomrule
\end{tabular}%
}
  \vspace{0.5em}
  \end{minipage}%
  \hfill
  \begin{minipage}[t]{0.32\textwidth}
  \centering
\caption*{\textbf{Caltech101}}\vspace{-1em}
\setlength{\tabcolsep}{2.5pt}
\resizebox{\linewidth}{!}{%
\begin{tabular}{@{}lcc|c@{}}
\toprule
\textbf{Method} & \textbf{Base} & \textbf{Novel} & \textbf{HM} \\
\midrule
CLIP & 91.40 & 91.40 & 91.40 \\ 
BitFit & 97.31 & 95.88 & 96.58 \\
LayerNorm & 96.07 & 95.55 & 95.81 \\
SAFT & 97.19 & 95.23 & 96.19 \\
CoOp & 96.42 & 92.57 & 94.42 \\
KgCoOp & 95.91 & 95.44 & 95.67 \\
ProGrad & 94.73 & 79.98 & 86.65 \\
LoRA & 95.81 & 88.35 & 91.93 \\
PiSSA & 96.66 & 81.37 & 88.35 \\
GaLoRE & 96.11 & 95.33 & 95.72 \\
SO & 97.35 & 95.72 & 96.52 \\
\cmidrule(lr){1-4}
SignSparseSGD & 97.07 & 95.62 & 96.34 \\
DynSignSparseSGD & 97.33 & 95.69 & 96.50 \\
\bottomrule
\end{tabular}%
}
  \vspace{0.5em}
  \end{minipage}%
  \par\vspace{0.45em}
  \begin{minipage}[t]{0.32\textwidth}
  \centering
\caption*{\textbf{Oxford Flowers}}\vspace{-1em}
\setlength{\tabcolsep}{2.5pt}
\resizebox{\linewidth}{!}{%
\begin{tabular}{@{}lcc|c@{}}
\toprule
\textbf{Method} & \textbf{Base} & \textbf{Novel} & \textbf{HM} \\
\midrule
CLIP & 64.03 & 64.03 & 64.03 \\ 
BitFit & 79.33 & 74.79 & 76.88 \\
LayerNorm & 72.46 & 75.00 & 73.64 \\
SAFT & 88.06 & 71.44 & 78.82 \\
CoOp & 83.93 & 69.72 & 76.07 \\
KgCoOp & 73.30 & 74.49 & 73.85 \\
ProGrad & 74.78 & 50.88 & 60.48 \\
LoRA & 70.79 & 56.94 & 62.86 \\
PiSSA & 88.09 & 38.15 & 52.90 \\
GaLoRE & 87.55 & 71.24 & 78.53 \\
SO & 81.82 & 75.04 & 78.18 \\
\cmidrule(lr){1-4}
SignSparseSGD & 88.23 & 71.22 & 78.78 \\
DynSignSparseSGD & 89.51 & 74.42 & 81.23 \\
\bottomrule
\end{tabular}%
}
  \vspace{0.5em}
  \end{minipage}%
  \hfill
  \begin{minipage}[t]{0.32\textwidth}
  \centering
\caption*{\textbf{Oxford Pets}}\vspace{-1em}
\setlength{\tabcolsep}{2.5pt}
\resizebox{\linewidth}{!}{%
\begin{tabular}{@{}lcc|c@{}}
\toprule
\textbf{Method} & \textbf{Base} & \textbf{Novel} & \textbf{HM} \\
\midrule
CLIP & 85.06 & 85.06 & 85.06 \\ 
BitFit & 94.17 & 89.94 & 92.00 \\
LayerNorm & 93.23 & 88.63 & 90.83 \\
SAFT & 92.81 & 89.87 & 91.29 \\
CoOp & 92.49 & 87.33 & 89.83 \\
KgCoOp & 94.72 & 88.03 & 91.24 \\
ProGrad & 91.60 & 63.86 & 74.88 \\
LoRA & 91.62 & 76.77 & 83.45 \\
PiSSA & 91.37 & 53.24 & 66.79 \\
GaLoRE & 91.95 & 89.29 & 90.59 \\
SO & 93.78 & 89.52 & 91.59 \\
\cmidrule(lr){1-4}
SignSparseSGD & 93.65 & 89.85 & 91.70 \\
DynSignSparseSGD & 94.06 & 90.17 & 92.06 \\
\bottomrule
\end{tabular}%
}
  \vspace{0.5em}
  \end{minipage}%
  \hfill
  \begin{minipage}[t]{0.32\textwidth}
  \centering
\caption*{\textbf{Stanford Cars}}\vspace{-1em}
\setlength{\tabcolsep}{2.5pt}
\resizebox{\linewidth}{!}{%
\begin{tabular}{@{}lcc|c@{}}
\toprule
\textbf{Method} & \textbf{Base} & \textbf{Novel} & \textbf{HM} \\
\midrule
CLIP & 60.14 & 60.14 & 60.14 \\ 
BitFit & 74.08 & 68.76 & 71.30 \\
LayerNorm & 72.75 & 69.51 & 71.08 \\
SAFT & 74.72 & 66.67 & 70.46 \\
CoOp & 71.47 & 64.86 & 67.99 \\
KgCoOp & 71.28 & 69.11 & 70.16 \\
ProGrad & 67.93 & 33.74 & 45.00 \\
LoRA & 71.33 & 48.22 & 57.47 \\
PiSSA & 73.59 & 35.92 & 48.24 \\
GaLoRE & 65.74 & 59.98 & 62.72 \\
SO & 75.97 & 68.78 & 72.18 \\
\cmidrule(lr){1-4}
SignSparseSGD & 73.48 & 65.01 & 68.96 \\
DynSignSparseSGD & 75.40 & 67.13 & 71.01 \\
\bottomrule
\end{tabular}%
}
  \vspace{0.5em}
  \end{minipage}
\end{table}

\begin{table}[p]
  \ContinuedFloat
  \def\arraystretch{0.9}
  \centering
  \scriptsize
  \caption[]{Base-to-new 2-shot classification with CLIP(ViT-B/32). Top-1 accuracy (\%) averaged over 3 fixed random seeds. (continued)}
  \begin{minipage}[t]{0.32\textwidth}
  \centering
\caption*{\textbf{Food 101}}\vspace{-1em}
\setlength{\tabcolsep}{2.5pt}
\resizebox{\linewidth}{!}{%
\begin{tabular}{@{}lcc|c@{}}
\toprule
\textbf{Method} & \textbf{Base} & \textbf{Novel} & \textbf{HM} \\
\midrule
CLIP & 78.86 & 78.86 & 78.86 \\ 
BitFit & 85.60 & 86.24 & 85.91 \\
LayerNorm & 85.41 & 85.78 & 85.58 \\
SAFT & 83.81 & 85.38 & 84.58 \\
CoOp & 84.55 & 84.57 & 84.55 \\
KgCoOp & 85.48 & 85.26 & 85.36 \\
ProGrad & 81.75 & 67.00 & 73.40 \\
LoRA & 82.44 & 78.03 & 80.17 \\
PiSSA & 81.34 & 64.03 & 71.65 \\
GaLoRE & 81.73 & 83.62 & 82.66 \\
SO & 85.90 & 86.47 & 86.18 \\
\cmidrule(lr){1-4}
SignSparseSGD & 84.33 & 85.45 & 84.88 \\
DynSignSparseSGD & 85.41 & 86.25 & 85.82 \\
\bottomrule
\end{tabular}%
}
  \vspace{0.5em}
  \end{minipage}%
  \hfill
  \begin{minipage}[t]{0.32\textwidth}
  \centering
\caption*{\textbf{FGVC Aircraft}}\vspace{-1em}
\setlength{\tabcolsep}{2.5pt}
\resizebox{\linewidth}{!}{%
\begin{tabular}{@{}lcc|c@{}}
\toprule
\textbf{Method} & \textbf{Base} & \textbf{Novel} & \textbf{HM} \\
\midrule
CLIP & 18.06 & 18.06 & 18.06 \\ 
BitFit & 32.16 & 29.44 & 30.72 \\
LayerNorm & 27.72 & 29.04 & 28.33 \\
SAFT & 33.33 & 28.42 & 30.64 \\
CoOp & 29.92 & 22.03 & 25.35 \\
KgCoOp & 29.48 & 27.34 & 28.37 \\
ProGrad & 27.86 & 7.25 & 10.79 \\
LoRA & 26.30 & 15.37 & 19.35 \\
PiSSA & 31.84 & 5.67 & 9.53 \\
GaLoRE & 30.74 & 24.53 & 27.28 \\
SO & 33.37 & 29.84 & 31.50 \\
\cmidrule(lr){1-4}
SignSparseSGD & 31.65 & 26.70 & 28.95 \\
DynSignSparseSGD & 33.63 & 28.36 & 30.77 \\
\bottomrule
\end{tabular}%
}
  \vspace{0.5em}
  \end{minipage}%
  \hfill
  \begin{minipage}[t]{0.32\textwidth}
  \centering
\caption*{\textbf{SUN 397}}\vspace{-1em}
\setlength{\tabcolsep}{2.5pt}
\resizebox{\linewidth}{!}{%
\begin{tabular}{@{}lcc|c@{}}
\toprule
\textbf{Method} & \textbf{Base} & \textbf{Novel} & \textbf{HM} \\
\midrule
CLIP & 62.06 & 62.06 & 62.06 \\ 
BitFit & 77.53 & 74.76 & 76.10 \\
LayerNorm & 74.82 & 73.94 & 74.37 \\
SAFT & 77.00 & 73.52 & 75.20 \\
CoOp & 76.01 & 71.37 & 73.60 \\
KgCoOp & 75.69 & 72.26 & 73.93 \\
ProGrad & 68.97 & 51.95 & 59.25 \\
LoRA & 72.20 & 58.40 & 64.56 \\
PiSSA & 72.79 & 50.25 & 59.45 \\
GaLoRE & 74.63 & 71.66 & 73.10 \\
SO & 77.81 & 74.32 & 76.01 \\
\cmidrule(lr){1-4}
SignSparseSGD & 76.64 & 72.46 & 74.47 \\
DynSignSparseSGD & 77.63 & 73.81 & 75.65 \\
\bottomrule
\end{tabular}%
}
  \vspace{0.5em}
  \end{minipage}%
  \par\vspace{0.45em}
  \begin{minipage}[t]{0.32\textwidth}
  \centering
\caption*{\textbf{DTD}}\vspace{-1em}
\setlength{\tabcolsep}{2.5pt}
\resizebox{\linewidth}{!}{%
\begin{tabular}{@{}lcc|c@{}}
\toprule
\textbf{Method} & \textbf{Base} & \textbf{Novel} & \textbf{HM} \\
\midrule
CLIP & 42.97 & 42.97 & 42.97 \\ 
BitFit & 67.59 & 51.43 & 58.31 \\
LayerNorm & 57.05 & 51.31 & 53.85 \\
SAFT & 70.53 & 50.35 & 58.73 \\
CoOp & 64.17 & 43.87 & 52.10 \\
KgCoOp & 61.31 & 51.16 & 55.67 \\
ProGrad & 61.07 & 24.69 & 34.78 \\
LoRA & 63.04 & 37.00 & 46.45 \\
PiSSA & 68.48 & 21.76 & 32.85 \\
GaLoRE & 68.56 & 49.15 & 57.16 \\
SO & 68.52 & 53.55 & 60.03 \\
\cmidrule(lr){1-4}
SignSparseSGD & 70.97 & 51.00 & 59.30 \\
DynSignSparseSGD & 70.25 & 53.01 & 60.38 \\
\bottomrule
\end{tabular}%
}
  \vspace{0.5em}
  \end{minipage}%
  \hfill
  \begin{minipage}[t]{0.32\textwidth}
  \centering
\caption*{\textbf{EuroSAT}}\vspace{-1em}
\setlength{\tabcolsep}{2.5pt}
\resizebox{\linewidth}{!}{%
\begin{tabular}{@{}lcc|c@{}}
\toprule
\textbf{Method} & \textbf{Base} & \textbf{Novel} & \textbf{HM} \\
\midrule
CLIP & 35.81 & 35.81 & 35.81 \\ 
BitFit & 82.25 & 64.56 & 72.16 \\
LayerNorm & 63.11 & 49.69 & 54.62 \\
SAFT & 82.36 & 64.55 & 72.29 \\
CoOp & 72.53 & 39.79 & 49.94 \\
KgCoOp & 72.27 & 46.67 & 55.62 \\
ProGrad & 73.20 & 3.48 & 5.99 \\
LoRA & 80.20 & 15.89 & 24.53 \\
PiSSA & 81.80 & 2.78 & 5.27 \\
GaLoRE & 84.59 & 58.90 & 69.38 \\
SO & 82.20 & 60.50 & 69.33 \\
\cmidrule(lr){1-4}
SignSparseSGD & 83.89 & 60.08 & 69.76 \\
DynSignSparseSGD & 85.73 & 61.19 & 71.18 \\
\bottomrule
\end{tabular}%
}
  \vspace{0.5em}
  \end{minipage}%
  \hfill
  \begin{minipage}[t]{0.32\textwidth}
  \centering
\caption*{\textbf{UCF101}}\vspace{-1em}
\setlength{\tabcolsep}{2.5pt}
\resizebox{\linewidth}{!}{%
\begin{tabular}{@{}lcc|c@{}}
\toprule
\textbf{Method} & \textbf{Base} & \textbf{Novel} & \textbf{HM} \\
\midrule
CLIP & 61.64 & 61.64 & 61.64 \\ 
BitFit & 79.64 & 76.96 & 78.07 \\
LayerNorm & 71.76 & 74.61 & 72.83 \\
SAFT & 80.73 & 76.44 & 78.34 \\
CoOp & 77.64 & 70.13 & 73.48 \\
KgCoOp & 75.82 & 73.35 & 74.28 \\
ProGrad & 73.99 & 49.94 & 59.22 \\
LoRA & 77.52 & 57.43 & 65.86 \\
PiSSA & 80.27 & 45.26 & 57.79 \\
GaLoRE & 80.00 & 75.96 & 77.72 \\
SO & 80.78 & 76.89 & 78.58 \\
\cmidrule(lr){1-4}
SignSparseSGD & 80.67 & 76.93 & 78.56 \\
DynSignSparseSGD & 81.26 & 76.72 & 78.70 \\
\bottomrule
\end{tabular}%
}
  \vspace{0.5em}
  \end{minipage}
\end{table}

\begin{table}[p]
  \def\arraystretch{0.9}
  \centering
  \scriptsize
  \caption{Base-to-new 4-shot classification with CLIP(ViT-B/32). Top-1 accuracy (\%) averaged over 3 fixed random seeds.}
  \label{tab:b2n_peft_4shot_vit-b-32_with_aug}
  \begin{minipage}[t]{0.32\textwidth}
  \centering
\caption*{\textbf{Average across datasets.}}\vspace{-1em}
\setlength{\tabcolsep}{2.5pt}
\resizebox{\linewidth}{!}{%
\begin{tabular}{@{}lcc|c@{}}
\toprule
\textbf{Method} & \textbf{Base} & \textbf{Novel} & \textbf{HM} \\
\midrule
CLIP & 60.19 & 60.19 & 60.19 \\ 
BitFit & 78.97 & 71.10 & 74.57 \\
LayerNorm & 72.47 & 69.97 & 70.97 \\
SAFT & 79.72 & 69.61 & 73.94 \\
CoOp & 77.17 & 66.34 & 70.84 \\
KgCoOp & 74.48 & 68.58 & 70.97 \\
ProGrad & 73.71 & 43.36 & 51.83 \\
LoRA & 77.26 & 46.09 & 55.68 \\
PiSSA & 78.96 & 36.67 & 46.88 \\
GaLoRE & 78.29 & 68.55 & 72.71 \\
SO & 79.84 & 70.85 & 74.78 \\
\cmidrule(lr){1-4}
SignSparseSGD & 79.65 & 69.30 & 73.72 \\
DynSignSparseSGD & 80.77 & 70.64 & 74.98 \\
\bottomrule
\end{tabular}%
}
  \vspace{0.5em}
  \end{minipage}%
  \hfill
  \begin{minipage}[t]{0.32\textwidth}
  \centering
\caption*{\textbf{ImageNet}}\vspace{-1em}
\setlength{\tabcolsep}{2.5pt}
\resizebox{\linewidth}{!}{%
\begin{tabular}{@{}lcc|c@{}}
\toprule
\textbf{Method} & \textbf{Base} & \textbf{Novel} & \textbf{HM} \\
\midrule
CLIP & 62.04 & 62.04 & 62.04 \\ 
BitFit & 73.91 & 72.54 & 73.22 \\
LayerNorm & 73.45 & 73.22 & 73.33 \\
SAFT & 73.19 & 70.83 & 71.98 \\
CoOp & 72.64 & 70.29 & 71.44 \\
KgCoOp & 72.88 & 71.27 & 72.07 \\
ProGrad & 67.25 & 44.22 & 53.35 \\
LoRA & 70.26 & 43.05 & 53.38 \\
PiSSA & 68.49 & 39.06 & 49.74 \\
GaLoRE & 68.08 & 65.97 & 67.01 \\
SO & 72.21 & 70.07 & 71.12 \\
\cmidrule(lr){1-4}
SignSparseSGD & 71.82 & 69.22 & 70.50 \\
DynSignSparseSGD & 72.85 & 71.57 & 72.20 \\
\bottomrule
\end{tabular}%
}
  \vspace{0.5em}
  \end{minipage}%
  \hfill
  \begin{minipage}[t]{0.32\textwidth}
  \centering
\caption*{\textbf{Caltech101}}\vspace{-1em}
\setlength{\tabcolsep}{2.5pt}
\resizebox{\linewidth}{!}{%
\begin{tabular}{@{}lcc|c@{}}
\toprule
\textbf{Method} & \textbf{Base} & \textbf{Novel} & \textbf{HM} \\
\midrule
CLIP & 91.40 & 91.40 & 91.40 \\ 
BitFit & 97.44 & 95.52 & 96.47 \\
LayerNorm & 96.29 & 95.63 & 95.96 \\
SAFT & 97.02 & 95.40 & 96.20 \\
CoOp & 96.43 & 93.67 & 95.01 \\
KgCoOp & 96.13 & 95.44 & 95.78 \\
ProGrad & 95.41 & 76.84 & 85.05 \\
LoRA & 97.34 & 83.83 & 90.08 \\
PiSSA & 97.18 & 77.61 & 86.25 \\
GaLoRE & 96.88 & 94.86 & 95.85 \\
SO & 97.63 & 95.62 & 96.61 \\
\cmidrule(lr){1-4}
SignSparseSGD & 97.19 & 95.39 & 96.28 \\
DynSignSparseSGD & 97.30 & 95.67 & 96.47 \\
\bottomrule
\end{tabular}%
}
  \vspace{0.5em}
  \end{minipage}%
  \par\vspace{0.45em}
  \begin{minipage}[t]{0.32\textwidth}
  \centering
\caption*{\textbf{Oxford Flowers}}\vspace{-1em}
\setlength{\tabcolsep}{2.5pt}
\resizebox{\linewidth}{!}{%
\begin{tabular}{@{}lcc|c@{}}
\toprule
\textbf{Method} & \textbf{Base} & \textbf{Novel} & \textbf{HM} \\
\midrule
CLIP & 64.03 & 64.03 & 64.03 \\ 
BitFit & 85.68 & 73.19 & 78.87 \\
LayerNorm & 73.63 & 74.95 & 74.21 \\
SAFT & 92.15 & 69.94 & 79.48 \\
CoOp & 87.87 & 69.73 & 77.70 \\
KgCoOp & 73.98 & 74.49 & 74.16 \\
ProGrad & 79.65 & 54.54 & 64.75 \\
LoRA & 87.23 & 45.77 & 59.88 \\
PiSSA & 93.05 & 36.87 & 52.65 \\
GaLoRE & 92.15 & 70.35 & 79.72 \\
SO & 91.67 & 72.90 & 81.15 \\
\cmidrule(lr){1-4}
SignSparseSGD & 91.40 & 71.84 & 80.40 \\
DynSignSparseSGD & 93.11 & 73.50 & 82.12 \\
\bottomrule
\end{tabular}%
}
  \vspace{0.5em}
  \end{minipage}%
  \hfill
  \begin{minipage}[t]{0.32\textwidth}
  \centering
\caption*{\textbf{Oxford Pets}}\vspace{-1em}
\setlength{\tabcolsep}{2.5pt}
\resizebox{\linewidth}{!}{%
\begin{tabular}{@{}lcc|c@{}}
\toprule
\textbf{Method} & \textbf{Base} & \textbf{Novel} & \textbf{HM} \\
\midrule
CLIP & 85.06 & 85.06 & 85.06 \\ 
BitFit & 95.36 & 90.12 & 92.65 \\
LayerNorm & 93.44 & 88.79 & 91.00 \\
SAFT & 94.52 & 89.50 & 91.93 \\
CoOp & 94.17 & 86.86 & 90.34 \\
KgCoOp & 94.76 & 88.03 & 91.26 \\
ProGrad & 92.69 & 67.23 & 77.74 \\
LoRA & 93.07 & 75.45 & 83.23 \\
PiSSA & 93.53 & 53.21 & 67.73 \\
GaLoRE & 92.71 & 88.62 & 90.61 \\
SO & 95.14 & 90.03 & 92.50 \\
\cmidrule(lr){1-4}
SignSparseSGD & 93.85 & 89.80 & 91.75 \\
DynSignSparseSGD & 94.24 & 90.33 & 92.22 \\
\bottomrule
\end{tabular}%
}
  \vspace{0.5em}
  \end{minipage}%
  \hfill
  \begin{minipage}[t]{0.32\textwidth}
  \centering
\caption*{\textbf{Stanford Cars}}\vspace{-1em}
\setlength{\tabcolsep}{2.5pt}
\resizebox{\linewidth}{!}{%
\begin{tabular}{@{}lcc|c@{}}
\toprule
\textbf{Method} & \textbf{Base} & \textbf{Novel} & \textbf{HM} \\
\midrule
CLIP & 60.14 & 60.14 & 60.14 \\ 
BitFit & 74.80 & 68.71 & 71.61 \\
LayerNorm & 72.94 & 69.68 & 71.26 \\
SAFT & 75.73 & 66.45 & 70.78 \\
CoOp & 73.72 & 68.04 & 70.75 \\
KgCoOp & 71.87 & 69.11 & 70.45 \\
ProGrad & 66.95 & 45.25 & 53.86 \\
LoRA & 72.88 & 44.11 & 54.92 \\
PiSSA & 75.20 & 32.72 & 45.57 \\
GaLoRE & 75.10 & 64.52 & 69.39 \\
SO & 76.92 & 70.38 & 73.50 \\
\cmidrule(lr){1-4}
SignSparseSGD & 75.63 & 64.82 & 69.80 \\
DynSignSparseSGD & 78.86 & 68.37 & 73.23 \\
\bottomrule
\end{tabular}%
}
  \vspace{0.5em}
  \end{minipage}
\end{table}

\begin{table}[p]
  \ContinuedFloat
  \def\arraystretch{0.9}
  \centering
  \scriptsize
  \caption[]{Base-to-new 4-shot classification with CLIP(ViT-B/32). Top-1 accuracy (\%) averaged over 3 fixed random seeds. (continued)}
  \begin{minipage}[t]{0.32\textwidth}
  \centering
\caption*{\textbf{Food 101}}\vspace{-1em}
\setlength{\tabcolsep}{2.5pt}
\resizebox{\linewidth}{!}{%
\begin{tabular}{@{}lcc|c@{}}
\toprule
\textbf{Method} & \textbf{Base} & \textbf{Novel} & \textbf{HM} \\
\midrule
CLIP & 78.86 & 78.86 & 78.86 \\ 
BitFit & 85.95 & 86.33 & 86.13 \\
LayerNorm & 85.78 & 86.01 & 85.89 \\
SAFT & 84.27 & 85.00 & 84.63 \\
CoOp & 84.68 & 83.81 & 84.24 \\
KgCoOp & 85.72 & 85.26 & 85.48 \\
ProGrad & 83.20 & 64.11 & 72.30 \\
LoRA & 83.39 & 70.37 & 76.33 \\
PiSSA & 82.00 & 58.59 & 68.34 \\
GaLoRE & 80.53 & 83.30 & 81.88 \\
SO & 85.62 & 86.57 & 86.09 \\
\cmidrule(lr){1-4}
SignSparseSGD & 83.54 & 84.03 & 83.78 \\
DynSignSparseSGD & 84.69 & 85.75 & 85.22 \\
\bottomrule
\end{tabular}%
}
  \vspace{0.5em}
  \end{minipage}%
  \hfill
  \begin{minipage}[t]{0.32\textwidth}
  \centering
\caption*{\textbf{FGVC Aircraft}}\vspace{-1em}
\setlength{\tabcolsep}{2.5pt}
\resizebox{\linewidth}{!}{%
\begin{tabular}{@{}lcc|c@{}}
\toprule
\textbf{Method} & \textbf{Base} & \textbf{Novel} & \textbf{HM} \\
\midrule
CLIP & 18.06 & 18.06 & 18.06 \\ 
BitFit & 34.65 & 30.22 & 32.28 \\
LayerNorm & 28.40 & 29.70 & 29.02 \\
SAFT & 37.71 & 28.84 & 32.67 \\
CoOp & 33.71 & 24.87 & 28.53 \\
KgCoOp & 30.72 & 27.34 & 28.92 \\
ProGrad & 31.06 & 4.40 & 7.57 \\
LoRA & 32.75 & 6.50 & 10.84 \\
PiSSA & 36.81 & 2.36 & 4.38 \\
GaLoRE & 35.07 & 26.16 & 29.96 \\
SO & 36.63 & 29.88 & 32.90 \\
\cmidrule(lr){1-4}
SignSparseSGD & 37.35 & 28.80 & 32.52 \\
DynSignSparseSGD & 38.99 & 30.02 & 33.92 \\
\bottomrule
\end{tabular}%
}
  \vspace{0.5em}
  \end{minipage}%
  \hfill
  \begin{minipage}[t]{0.32\textwidth}
  \centering
\caption*{\textbf{SUN 397}}\vspace{-1em}
\setlength{\tabcolsep}{2.5pt}
\resizebox{\linewidth}{!}{%
\begin{tabular}{@{}lcc|c@{}}
\toprule
\textbf{Method} & \textbf{Base} & \textbf{Novel} & \textbf{HM} \\
\midrule
CLIP & 62.06 & 62.06 & 62.06 \\ 
BitFit & 79.14 & 74.23 & 76.58 \\
LayerNorm & 76.23 & 74.49 & 75.34 \\
SAFT & 78.50 & 72.56 & 75.39 \\
CoOp & 77.67 & 68.64 & 72.86 \\
KgCoOp & 76.38 & 72.26 & 74.25 \\
ProGrad & 74.24 & 44.14 & 55.27 \\
LoRA & 76.36 & 46.00 & 57.39 \\
PiSSA & 75.89 & 42.03 & 54.10 \\
GaLoRE & 76.33 & 71.28 & 73.70 \\
SO & 78.88 & 72.87 & 75.73 \\
\cmidrule(lr){1-4}
SignSparseSGD & 78.22 & 72.22 & 75.08 \\
DynSignSparseSGD & 79.24 & 73.47 & 76.23 \\
\bottomrule
\end{tabular}%
}
  \vspace{0.5em}
  \end{minipage}%
  \par\vspace{0.45em}
  \begin{minipage}[t]{0.32\textwidth}
  \centering
\caption*{\textbf{DTD}}\vspace{-1em}
\setlength{\tabcolsep}{2.5pt}
\resizebox{\linewidth}{!}{%
\begin{tabular}{@{}lcc|c@{}}
\toprule
\textbf{Method} & \textbf{Base} & \textbf{Novel} & \textbf{HM} \\
\midrule
CLIP & 42.97 & 42.97 & 42.97 \\ 
BitFit & 71.62 & 50.58 & 59.28 \\
LayerNorm & 59.58 & 51.97 & 55.37 \\
SAFT & 73.43 & 50.66 & 59.94 \\
CoOp & 69.73 & 45.10 & 54.66 \\
KgCoOp & 64.09 & 51.16 & 56.80 \\
ProGrad & 62.52 & 28.36 & 38.85 \\
LoRA & 70.97 & 22.26 & 33.79 \\
PiSSA & 74.19 & 17.79 & 28.66 \\
GaLoRE & 73.31 & 50.23 & 59.59 \\
SO & 73.15 & 50.93 & 60.04 \\
\cmidrule(lr){1-4}
SignSparseSGD & 74.32 & 51.16 & 60.59 \\
DynSignSparseSGD & 75.44 & 52.93 & 62.21 \\
\bottomrule
\end{tabular}%
}
  \vspace{0.5em}
  \end{minipage}%
  \hfill
  \begin{minipage}[t]{0.32\textwidth}
  \centering
\caption*{\textbf{EuroSAT}}\vspace{-1em}
\setlength{\tabcolsep}{2.5pt}
\resizebox{\linewidth}{!}{%
\begin{tabular}{@{}lcc|c@{}}
\toprule
\textbf{Method} & \textbf{Base} & \textbf{Novel} & \textbf{HM} \\
\midrule
CLIP & 35.81 & 35.81 & 35.81 \\ 
BitFit & 88.93 & 64.58 & 74.81 \\
LayerNorm & 64.05 & 50.43 & 55.46 \\
SAFT & 88.80 & 60.39 & 71.83 \\
CoOp & 77.92 & 46.65 & 57.92 \\
KgCoOp & 75.76 & 46.67 & 56.59 \\
ProGrad & 80.42 & 2.20 & 4.06 \\
LoRA & 84.14 & 21.16 & 31.99 \\
PiSSA & 90.43 & 1.81 & 3.54 \\
GaLoRE & 89.21 & 63.45 & 73.95 \\
SO & 88.24 & 63.70 & 73.95 \\
\cmidrule(lr){1-4}
SignSparseSGD & 89.97 & 58.93 & 71.14 \\
DynSignSparseSGD & 90.18 & 59.22 & 71.43 \\
\bottomrule
\end{tabular}%
}
  \vspace{0.5em}
  \end{minipage}%
  \hfill
  \begin{minipage}[t]{0.32\textwidth}
  \centering
\caption*{\textbf{UCF101}}\vspace{-1em}
\setlength{\tabcolsep}{2.5pt}
\resizebox{\linewidth}{!}{%
\begin{tabular}{@{}lcc|c@{}}
\toprule
\textbf{Method} & \textbf{Base} & \textbf{Novel} & \textbf{HM} \\
\midrule
CLIP & 61.64 & 61.64 & 61.64 \\ 
BitFit & 81.15 & 76.08 & 78.32 \\
LayerNorm & 73.36 & 74.83 & 73.79 \\
SAFT & 81.60 & 76.14 & 78.54 \\
CoOp & 80.34 & 72.05 & 75.77 \\
KgCoOp & 76.98 & 73.35 & 74.86 \\
ProGrad & 77.41 & 45.66 & 57.34 \\
LoRA & 81.44 & 48.53 & 60.70 \\
PiSSA & 81.80 & 41.29 & 54.71 \\
GaLoRE & 81.84 & 75.26 & 78.12 \\
SO & 82.17 & 76.44 & 78.96 \\
\cmidrule(lr){1-4}
SignSparseSGD & 82.81 & 76.06 & 79.05 \\
DynSignSparseSGD & 83.58 & 76.25 & 79.52 \\
\bottomrule
\end{tabular}%
}
  \vspace{0.5em}
  \end{minipage}
\end{table}

\begin{table}[p]
  \def\arraystretch{0.9}
  \centering
  \scriptsize
  \caption{Base-to-new 8-shot classification with CLIP(ViT-B/32). Top-1 accuracy (\%) averaged over 3 fixed random seeds.}
  \label{tab:b2n_peft_8shot_vit-b-32_with_aug}
  \begin{minipage}[t]{0.32\textwidth}
  \centering
\caption*{\textbf{Average across datasets.}}\vspace{-1em}
\setlength{\tabcolsep}{2.5pt}
\resizebox{\linewidth}{!}{%
\begin{tabular}{@{}lcc|c@{}}
\toprule
\textbf{Method} & \textbf{Base} & \textbf{Novel} & \textbf{HM} \\
\midrule
CLIP & 60.19 & 60.19 & 60.19 \\ 
BitFit & 80.67 & 70.81 & 75.09 \\
LayerNorm & 73.48 & 70.31 & 71.63 \\
SAFT & 81.63 & 69.66 & 74.73 \\
CoOp & 79.54 & 66.23 & 71.67 \\
KgCoOp & 75.16 & 68.58 & 71.29 \\
ProGrad & 77.12 & 38.91 & 48.43 \\
LoRA & 80.50 & 36.59 & 47.40 \\
PiSSA & 80.86 & 31.65 & 42.03 \\
GaLoRE & 80.60 & 67.75 & 73.10 \\
SO & 81.47 & 70.21 & 75.07 \\
\cmidrule(lr){1-4}
SignSparseSGD & 81.33 & 69.34 & 74.46 \\
DynSignSparseSGD & 82.52 & 71.17 & 76.03 \\
\bottomrule
\end{tabular}%
}
  \vspace{0.5em}
  \end{minipage}%
  \hfill
  \begin{minipage}[t]{0.32\textwidth}
  \centering
\caption*{\textbf{ImageNet}}\vspace{-1em}
\setlength{\tabcolsep}{2.5pt}
\resizebox{\linewidth}{!}{%
\begin{tabular}{@{}lcc|c@{}}
\toprule
\textbf{Method} & \textbf{Base} & \textbf{Novel} & \textbf{HM} \\
\midrule
CLIP & 62.04 & 62.04 & 62.04 \\ 
BitFit & 74.73 & 72.56 & 73.63 \\
LayerNorm & 74.13 & 73.65 & 73.88 \\
SAFT & 74.07 & 70.96 & 72.48 \\
CoOp & 73.77 & 69.64 & 71.64 \\
KgCoOp & 73.32 & 71.27 & 72.28 \\
ProGrad & 70.65 & 39.97 & 50.83 \\
LoRA & 72.20 & 30.14 & 42.51 \\
PiSSA & 70.83 & 30.29 & 42.42 \\
GaLoRE & 68.99 & 64.59 & 66.72 \\
SO & 72.62 & 68.25 & 70.37 \\
\cmidrule(lr){1-4}
SignSparseSGD & 73.10 & 68.71 & 70.83 \\
DynSignSparseSGD & 73.84 & 71.22 & 72.50 \\
\bottomrule
\end{tabular}%
}
  \vspace{0.5em}
  \end{minipage}%
  \hfill
  \begin{minipage}[t]{0.32\textwidth}
  \centering
\caption*{\textbf{Caltech101}}\vspace{-1em}
\setlength{\tabcolsep}{2.5pt}
\resizebox{\linewidth}{!}{%
\begin{tabular}{@{}lcc|c@{}}
\toprule
\textbf{Method} & \textbf{Base} & \textbf{Novel} & \textbf{HM} \\
\midrule
CLIP & 91.40 & 91.40 & 91.40 \\ 
BitFit & 97.56 & 95.64 & 96.59 \\
LayerNorm & 96.49 & 95.68 & 96.08 \\
SAFT & 97.47 & 95.60 & 96.52 \\
CoOp & 96.89 & 93.20 & 95.01 \\
KgCoOp & 96.31 & 95.44 & 95.87 \\
ProGrad & 96.67 & 75.23 & 84.61 \\
LoRA & 97.39 & 80.37 & 88.06 \\
PiSSA & 97.57 & 75.63 & 85.21 \\
GaLoRE & 96.88 & 95.45 & 96.15 \\
SO & 97.43 & 95.75 & 96.58 \\
\cmidrule(lr){1-4}
SignSparseSGD & 97.22 & 95.55 & 96.37 \\
DynSignSparseSGD & 97.62 & 95.92 & 96.75 \\
\bottomrule
\end{tabular}%
}
  \vspace{0.5em}
  \end{minipage}%
  \par\vspace{0.45em}
  \begin{minipage}[t]{0.32\textwidth}
  \centering
\caption*{\textbf{Oxford Flowers}}\vspace{-1em}
\setlength{\tabcolsep}{2.5pt}
\resizebox{\linewidth}{!}{%
\begin{tabular}{@{}lcc|c@{}}
\toprule
\textbf{Method} & \textbf{Base} & \textbf{Novel} & \textbf{HM} \\
\midrule
CLIP & 64.03 & 64.03 & 64.03 \\ 
BitFit & 90.52 & 72.33 & 80.32 \\
LayerNorm & 74.98 & 75.50 & 75.15 \\
SAFT & 93.70 & 70.28 & 80.26 \\
CoOp & 91.76 & 68.04 & 78.06 \\
KgCoOp & 75.30 & 74.49 & 74.82 \\
ProGrad & 87.69 & 46.12 & 59.50 \\
LoRA & 93.54 & 35.93 & 51.64 \\
PiSSA & 94.43 & 31.19 & 46.51 \\
GaLoRE & 93.59 & 69.46 & 79.71 \\
SO & 94.53 & 71.87 & 81.58 \\
\cmidrule(lr){1-4}
SignSparseSGD & 93.48 & 68.45 & 78.99 \\
DynSignSparseSGD & 94.98 & 72.37 & 82.10 \\
\bottomrule
\end{tabular}%
}
  \vspace{0.5em}
  \end{minipage}%
  \hfill
  \begin{minipage}[t]{0.32\textwidth}
  \centering
\caption*{\textbf{Oxford Pets}}\vspace{-1em}
\setlength{\tabcolsep}{2.5pt}
\resizebox{\linewidth}{!}{%
\begin{tabular}{@{}lcc|c@{}}
\toprule
\textbf{Method} & \textbf{Base} & \textbf{Novel} & \textbf{HM} \\
\midrule
CLIP & 85.06 & 85.06 & 85.06 \\ 
BitFit & 95.06 & 89.77 & 92.33 \\
LayerNorm & 93.29 & 89.09 & 91.10 \\
SAFT & 94.39 & 89.48 & 91.86 \\
CoOp & 94.88 & 89.84 & 92.28 \\
KgCoOp & 94.89 & 88.03 & 91.32 \\
ProGrad & 94.15 & 65.13 & 76.66 \\
LoRA & 94.67 & 65.68 & 77.36 \\
PiSSA & 93.91 & 51.32 & 66.05 \\
GaLoRE & 93.51 & 89.47 & 91.44 \\
SO & 95.34 & 90.03 & 92.60 \\
\cmidrule(lr){1-4}
SignSparseSGD & 94.24 & 89.16 & 91.62 \\
DynSignSparseSGD & 95.00 & 90.25 & 92.55 \\
\bottomrule
\end{tabular}%
}
  \vspace{0.5em}
  \end{minipage}%
  \hfill
  \begin{minipage}[t]{0.32\textwidth}
  \centering
\caption*{\textbf{Stanford Cars}}\vspace{-1em}
\setlength{\tabcolsep}{2.5pt}
\resizebox{\linewidth}{!}{%
\begin{tabular}{@{}lcc|c@{}}
\toprule
\textbf{Method} & \textbf{Base} & \textbf{Novel} & \textbf{HM} \\
\midrule
CLIP & 60.14 & 60.14 & 60.14 \\ 
BitFit & 78.05 & 68.82 & 73.13 \\
LayerNorm & 73.89 & 70.25 & 72.01 \\
SAFT & 80.40 & 66.52 & 72.79 \\
CoOp & 76.46 & 66.24 & 70.97 \\
KgCoOp & 72.31 & 69.11 & 70.66 \\
ProGrad & 72.54 & 24.46 & 36.54 \\
LoRA & 79.74 & 26.49 & 39.71 \\
PiSSA & 79.92 & 23.74 & 36.59 \\
GaLoRE & 78.69 & 62.04 & 69.37 \\
SO & 81.11 & 66.97 & 73.36 \\
\cmidrule(lr){1-4}
SignSparseSGD & 80.33 & 65.03 & 71.87 \\
DynSignSparseSGD & 82.56 & 67.56 & 74.31 \\
\bottomrule
\end{tabular}%
}
  \vspace{0.5em}
  \end{minipage}
\end{table}

\begin{table}[p]
  \ContinuedFloat
  \def\arraystretch{0.9}
  \centering
  \scriptsize
  \caption[]{Base-to-new 8-shot classification with CLIP(ViT-B/32). Top-1 accuracy (\%) averaged over 3 fixed random seeds. (continued)}
  \begin{minipage}[t]{0.32\textwidth}
  \centering
\caption*{\textbf{Food 101}}\vspace{-1em}
\setlength{\tabcolsep}{2.5pt}
\resizebox{\linewidth}{!}{%
\begin{tabular}{@{}lcc|c@{}}
\toprule
\textbf{Method} & \textbf{Base} & \textbf{Novel} & \textbf{HM} \\
\midrule
CLIP & 78.86 & 78.86 & 78.86 \\ 
BitFit & 86.28 & 86.37 & 86.32 \\
LayerNorm & 86.34 & 86.43 & 86.38 \\
SAFT & 84.45 & 85.13 & 84.78 \\
CoOp & 85.78 & 84.59 & 85.18 \\
KgCoOp & 86.14 & 85.26 & 85.69 \\
ProGrad & 84.55 & 63.29 & 72.31 \\
LoRA & 84.86 & 61.55 & 71.34 \\
PiSSA & 82.55 & 52.87 & 64.42 \\
GaLoRE & 82.89 & 83.65 & 83.27 \\
SO & 85.96 & 86.44 & 86.20 \\
\cmidrule(lr){1-4}
SignSparseSGD & 84.50 & 84.75 & 84.62 \\
DynSignSparseSGD & 85.41 & 86.29 & 85.85 \\
\bottomrule
\end{tabular}%
}
  \vspace{0.5em}
  \end{minipage}%
  \hfill
  \begin{minipage}[t]{0.32\textwidth}
  \centering
\caption*{\textbf{FGVC Aircraft}}\vspace{-1em}
\setlength{\tabcolsep}{2.5pt}
\resizebox{\linewidth}{!}{%
\begin{tabular}{@{}lcc|c@{}}
\toprule
\textbf{Method} & \textbf{Base} & \textbf{Novel} & \textbf{HM} \\
\midrule
CLIP & 18.06 & 18.06 & 18.06 \\ 
BitFit & 37.15 & 30.84 & 33.70 \\
LayerNorm & 29.12 & 29.98 & 29.53 \\
SAFT & 38.71 & 30.76 & 34.27 \\
CoOp & 36.13 & 26.71 & 30.64 \\
KgCoOp & 32.14 & 27.34 & 29.54 \\
ProGrad & 32.74 & 3.28 & 5.48 \\
LoRA & 38.91 & 3.18 & 5.83 \\
PiSSA & 40.29 & 1.88 & 3.48 \\
GaLoRE & 40.31 & 25.79 & 31.35 \\
SO & 40.85 & 31.10 & 35.30 \\
\cmidrule(lr){1-4}
SignSparseSGD & 38.47 & 27.64 & 32.13 \\
DynSignSparseSGD & 42.57 & 29.94 & 35.10 \\
\bottomrule
\end{tabular}%
}
  \vspace{0.5em}
  \end{minipage}%
  \hfill
  \begin{minipage}[t]{0.32\textwidth}
  \centering
\caption*{\textbf{SUN 397}}\vspace{-1em}
\setlength{\tabcolsep}{2.5pt}
\resizebox{\linewidth}{!}{%
\begin{tabular}{@{}lcc|c@{}}
\toprule
\textbf{Method} & \textbf{Base} & \textbf{Novel} & \textbf{HM} \\
\midrule
CLIP & 62.06 & 62.06 & 62.06 \\ 
BitFit & 80.58 & 74.00 & 77.13 \\
LayerNorm & 77.55 & 75.04 & 76.26 \\
SAFT & 80.08 & 72.39 & 76.01 \\
CoOp & 79.42 & 69.43 & 74.07 \\
KgCoOp & 77.02 & 72.26 & 74.56 \\
ProGrad & 77.40 & 36.60 & 49.32 \\
LoRA & 79.22 & 31.68 & 45.25 \\
PiSSA & 78.17 & 34.18 & 47.55 \\
GaLoRE & 78.10 & 70.57 & 74.13 \\
SO & 80.39 & 72.39 & 76.16 \\
\cmidrule(lr){1-4}
SignSparseSGD & 80.15 & 72.15 & 75.92 \\
DynSignSparseSGD & 80.69 & 73.71 & 77.02 \\
\bottomrule
\end{tabular}%
}
  \vspace{0.5em}
  \end{minipage}%
  \par\vspace{0.45em}
  \begin{minipage}[t]{0.32\textwidth}
  \centering
\caption*{\textbf{DTD}}\vspace{-1em}
\setlength{\tabcolsep}{2.5pt}
\resizebox{\linewidth}{!}{%
\begin{tabular}{@{}lcc|c@{}}
\toprule
\textbf{Method} & \textbf{Base} & \textbf{Novel} & \textbf{HM} \\
\midrule
CLIP & 42.97 & 42.97 & 42.97 \\ 
BitFit & 75.24 & 49.31 & 59.56 \\
LayerNorm & 62.64 & 52.47 & 57.02 \\
SAFT & 77.58 & 49.85 & 60.69 \\
CoOp & 73.03 & 43.25 & 54.13 \\
KgCoOp & 66.91 & 51.16 & 57.92 \\
ProGrad & 68.92 & 23.15 & 34.54 \\
LoRA & 77.46 & 14.24 & 24.01 \\
PiSSA & 77.74 & 12.46 & 21.45 \\
GaLoRE & 76.85 & 51.20 & 61.44 \\
SO & 77.33 & 50.62 & 61.17 \\
\cmidrule(lr){1-4}
SignSparseSGD & 77.98 & 49.96 & 60.90 \\
DynSignSparseSGD & 78.82 & 52.51 & 63.02 \\
\bottomrule
\end{tabular}%
}
  \vspace{0.5em}
  \end{minipage}%
  \hfill
  \begin{minipage}[t]{0.32\textwidth}
  \centering
\caption*{\textbf{EuroSAT}}\vspace{-1em}
\setlength{\tabcolsep}{2.5pt}
\resizebox{\linewidth}{!}{%
\begin{tabular}{@{}lcc|c@{}}
\toprule
\textbf{Method} & \textbf{Base} & \textbf{Novel} & \textbf{HM} \\
\midrule
CLIP & 35.81 & 35.81 & 35.81 \\ 
BitFit & 88.15 & 63.23 & 73.63 \\
LayerNorm & 64.42 & 49.64 & 55.16 \\
SAFT & 92.54 & 61.04 & 73.52 \\
CoOp & 83.93 & 48.46 & 61.23 \\
KgCoOp & 74.03 & 46.67 & 56.00 \\
ProGrad & 82.90 & 0.94 & 1.82 \\
LoRA & 83.09 & 14.84 & 23.22 \\
PiSSA & 90.73 & 0.28 & 0.55 \\
GaLoRE & 91.97 & 58.12 & 71.10 \\
SO & 85.85 & 62.74 & 72.42 \\
\cmidrule(lr){1-4}
SignSparseSGD & 90.34 & 66.52 & 76.60 \\
DynSignSparseSGD & 91.06 & 66.25 & 76.59 \\
\bottomrule
\end{tabular}%
}
  \vspace{0.5em}
  \end{minipage}%
  \hfill
  \begin{minipage}[t]{0.32\textwidth}
  \centering
\caption*{\textbf{UCF101}}\vspace{-1em}
\setlength{\tabcolsep}{2.5pt}
\resizebox{\linewidth}{!}{%
\begin{tabular}{@{}lcc|c@{}}
\toprule
\textbf{Method} & \textbf{Base} & \textbf{Novel} & \textbf{HM} \\
\midrule
CLIP & 61.64 & 61.64 & 61.64 \\ 
BitFit & 84.10 & 76.02 & 79.69 \\
LayerNorm & 75.40 & 75.73 & 75.32 \\
SAFT & 84.59 & 74.24 & 78.83 \\
CoOp & 82.91 & 69.08 & 75.20 \\
KgCoOp & 78.34 & 73.35 & 75.53 \\
ProGrad & 80.12 & 49.82 & 61.14 \\
LoRA & 84.47 & 38.38 & 52.51 \\
PiSSA & 83.31 & 34.28 & 48.12 \\
GaLoRE & 84.81 & 74.96 & 79.38 \\
SO & 84.71 & 76.15 & 79.99 \\
\cmidrule(lr){1-4}
SignSparseSGD & 84.79 & 74.87 & 79.25 \\
DynSignSparseSGD & 85.14 & 76.81 & 80.58 \\
\bottomrule
\end{tabular}%
}
  \vspace{0.5em}
  \end{minipage}
\end{table}

\begin{table}[p]
  \def\arraystretch{0.9}
  \centering
  \scriptsize
  \caption{Base-to-new 16-shot classification with CLIP(ViT-B/32). Top-1 accuracy (\%) averaged over 3 fixed random seeds.}
  \label{tab:b2n_peft_16shot_vit-b-32_with_aug}
  \begin{minipage}[t]{0.32\textwidth}
  \centering
\caption*{\textbf{Average across datasets.}}\vspace{-1em}
\setlength{\tabcolsep}{2.5pt}
\resizebox{\linewidth}{!}{%
\begin{tabular}{@{}lcc|c@{}}
\toprule
\textbf{Method} & \textbf{Base} & \textbf{Novel} & \textbf{HM} \\
\midrule
CLIP & 60.19 & 60.19 & 60.19 \\ 
BitFit & 83.09 & 70.21 & 75.61 \\
LayerNorm & 75.62 & 70.82 & 72.90 \\
SAFT & 83.81 & 69.49 & 75.45 \\
CoOp & 81.63 & 64.90 & 71.51 \\
KgCoOp & 76.36 & 68.58 & 71.79 \\
ProGrad & 78.84 & 42.32 & 51.91 \\
LoRA & 83.92 & 28.18 & 38.90 \\
PiSSA & 83.84 & 27.73 & 38.47 \\
GaLoRE & 82.83 & 68.13 & 74.21 \\
SO & 84.17 & 69.79 & 75.80 \\
\cmidrule(lr){1-4}
SignSparseSGD & 83.45 & 69.17 & 75.14 \\
DynSignSparseSGD & 84.68 & 70.75 & 76.57 \\
\bottomrule
\end{tabular}%
}
  \vspace{0.5em}
  \end{minipage}%
  \hfill
  \begin{minipage}[t]{0.32\textwidth}
  \centering
\caption*{\textbf{ImageNet}}\vspace{-1em}
\setlength{\tabcolsep}{2.5pt}
\resizebox{\linewidth}{!}{%
\begin{tabular}{@{}lcc|c@{}}
\toprule
\textbf{Method} & \textbf{Base} & \textbf{Novel} & \textbf{HM} \\
\midrule
CLIP & 62.04 & 62.04 & 62.04 \\ 
BitFit & 75.89 & 72.31 & 74.06 \\
LayerNorm & 74.93 & 73.88 & 74.40 \\
SAFT & 75.54 & 70.87 & 73.12 \\
CoOp & 75.21 & 69.13 & 72.05 \\
KgCoOp & 73.52 & 71.27 & 72.38 \\
ProGrad & 72.78 & 43.92 & 54.70 \\
LoRA & 74.30 & 23.82 & 36.05 \\
PiSSA & 73.03 & 23.73 & 35.79 \\
GaLoRE & 71.44 & 64.35 & 67.71 \\
SO & 74.33 & 68.00 & 71.02 \\
\cmidrule(lr){1-4}
SignSparseSGD & 74.60 & 68.80 & 71.58 \\
DynSignSparseSGD & 75.16 & 71.04 & 73.04 \\
\bottomrule
\end{tabular}%
}
  \vspace{0.5em}
  \end{minipage}%
  \hfill
  \begin{minipage}[t]{0.32\textwidth}
  \centering
\caption*{\textbf{Caltech101}}\vspace{-1em}
\setlength{\tabcolsep}{2.5pt}
\resizebox{\linewidth}{!}{%
\begin{tabular}{@{}lcc|c@{}}
\toprule
\textbf{Method} & \textbf{Base} & \textbf{Novel} & \textbf{HM} \\
\midrule
CLIP & 91.40 & 91.40 & 91.40 \\ 
BitFit & 97.74 & 95.73 & 96.72 \\
LayerNorm & 96.94 & 95.82 & 96.37 \\
SAFT & 97.67 & 95.33 & 96.48 \\
CoOp & 97.30 & 91.98 & 94.56 \\
KgCoOp & 96.85 & 95.44 & 96.14 \\
ProGrad & 97.08 & 80.79 & 88.17 \\
LoRA & 97.74 & 76.00 & 85.51 \\
PiSSA & 97.51 & 73.79 & 84.01 \\
GaLoRE & 97.26 & 95.25 & 96.24 \\
SO & 97.84 & 95.51 & 96.66 \\
\cmidrule(lr){1-4}
SignSparseSGD & 97.62 & 95.25 & 96.42 \\
DynSignSparseSGD & 97.76 & 95.57 & 96.65 \\
\bottomrule
\end{tabular}%
}
  \vspace{0.5em}
  \end{minipage}%
  \par\vspace{0.45em}
  \begin{minipage}[t]{0.32\textwidth}
  \centering
\caption*{\textbf{Oxford Flowers}}\vspace{-1em}
\setlength{\tabcolsep}{2.5pt}
\resizebox{\linewidth}{!}{%
\begin{tabular}{@{}lcc|c@{}}
\toprule
\textbf{Method} & \textbf{Base} & \textbf{Novel} & \textbf{HM} \\
\midrule
CLIP & 64.03 & 64.03 & 64.03 \\ 
BitFit & 95.14 & 69.31 & 80.08 \\
LayerNorm & 77.66 & 75.60 & 76.55 \\
SAFT & 96.17 & 68.93 & 80.28 \\
CoOp & 95.66 & 64.01 & 76.54 \\
KgCoOp & 78.46 & 74.49 & 76.35 \\
ProGrad & 88.20 & 52.50 & 65.58 \\
LoRA & 96.81 & 29.25 & 44.66 \\
PiSSA & 97.29 & 29.72 & 45.25 \\
GaLoRE & 96.44 & 68.01 & 79.73 \\
SO & 97.05 & 70.43 & 81.55 \\
\cmidrule(lr){1-4}
SignSparseSGD & 95.60 & 68.66 & 79.82 \\
DynSignSparseSGD & 97.20 & 71.88 & 82.58 \\
\bottomrule
\end{tabular}%
}
  \vspace{0.5em}
  \end{minipage}%
  \hfill
  \begin{minipage}[t]{0.32\textwidth}
  \centering
\caption*{\textbf{Oxford Pets}}\vspace{-1em}
\setlength{\tabcolsep}{2.5pt}
\resizebox{\linewidth}{!}{%
\begin{tabular}{@{}lcc|c@{}}
\toprule
\textbf{Method} & \textbf{Base} & \textbf{Novel} & \textbf{HM} \\
\midrule
CLIP & 85.06 & 85.06 & 85.06 \\ 
BitFit & 95.38 & 89.55 & 92.36 \\
LayerNorm & 94.01 & 89.43 & 91.63 \\
SAFT & 94.75 & 89.80 & 92.17 \\
CoOp & 95.04 & 89.82 & 92.35 \\
KgCoOp & 95.23 & 88.03 & 91.48 \\
ProGrad & 94.31 & 69.14 & 79.56 \\
LoRA & 94.84 & 50.38 & 65.58 \\
PiSSA & 94.76 & 43.20 & 59.10 \\
GaLoRE & 94.69 & 89.52 & 92.01 \\
SO & 95.27 & 90.16 & 92.63 \\
\cmidrule(lr){1-4}
SignSparseSGD & 94.76 & 89.89 & 92.25 \\
DynSignSparseSGD & 95.10 & 90.37 & 92.66 \\
\bottomrule
\end{tabular}%
}
  \vspace{0.5em}
  \end{minipage}%
  \hfill
  \begin{minipage}[t]{0.32\textwidth}
  \centering
\caption*{\textbf{Stanford Cars}}\vspace{-1em}
\setlength{\tabcolsep}{2.5pt}
\resizebox{\linewidth}{!}{%
\begin{tabular}{@{}lcc|c@{}}
\toprule
\textbf{Method} & \textbf{Base} & \textbf{Novel} & \textbf{HM} \\
\midrule
CLIP & 60.14 & 60.14 & 60.14 \\ 
BitFit & 80.67 & 67.18 & 73.31 \\
LayerNorm & 75.15 & 70.21 & 72.58 \\
SAFT & 83.40 & 64.49 & 72.73 \\
CoOp & 80.47 & 64.70 & 71.71 \\
KgCoOp & 72.61 & 69.11 & 70.81 \\
ProGrad & 76.16 & 28.91 & 41.18 \\
LoRA & 84.47 & 17.40 & 28.84 \\
PiSSA & 84.54 & 18.91 & 30.89 \\
GaLoRE & 82.79 & 62.26 & 71.07 \\
SO & 84.84 & 66.13 & 74.32 \\
\cmidrule(lr){1-4}
SignSparseSGD & 83.05 & 64.37 & 72.52 \\
DynSignSparseSGD & 85.54 & 66.30 & 74.70 \\
\bottomrule
\end{tabular}%
}
  \vspace{0.5em}
  \end{minipage}
\end{table}

\begin{table}[p]
  \ContinuedFloat
  \def\arraystretch{0.9}
  \centering
  \scriptsize
  \caption[]{Base-to-new 16-shot classification with CLIP(ViT-B/32). Top-1 accuracy (\%) averaged over 3 fixed random seeds. (continued)}
  \begin{minipage}[t]{0.32\textwidth}
  \centering
\caption*{\textbf{Food 101}}\vspace{-1em}
\setlength{\tabcolsep}{2.5pt}
\resizebox{\linewidth}{!}{%
\begin{tabular}{@{}lcc|c@{}}
\toprule
\textbf{Method} & \textbf{Base} & \textbf{Novel} & \textbf{HM} \\
\midrule
CLIP & 78.86 & 78.86 & 78.86 \\ 
BitFit & 86.70 & 86.37 & 86.53 \\
LayerNorm & 86.91 & 87.02 & 86.96 \\
SAFT & 85.21 & 84.65 & 84.93 \\
CoOp & 86.46 & 85.58 & 86.01 \\
KgCoOp & 86.44 & 85.26 & 85.84 \\
ProGrad & 84.86 & 64.21 & 72.62 \\
LoRA & 85.65 & 41.80 & 56.09 \\
PiSSA & 84.38 & 44.83 & 58.55 \\
GaLoRE & 82.44 & 82.90 & 82.67 \\
SO & 85.63 & 85.71 & 85.67 \\
\cmidrule(lr){1-4}
SignSparseSGD & 85.04 & 83.96 & 84.49 \\
DynSignSparseSGD & 86.07 & 86.11 & 86.09 \\
\bottomrule
\end{tabular}%
}
  \vspace{0.5em}
  \end{minipage}%
  \hfill
  \begin{minipage}[t]{0.32\textwidth}
  \centering
\caption*{\textbf{FGVC Aircraft}}\vspace{-1em}
\setlength{\tabcolsep}{2.5pt}
\resizebox{\linewidth}{!}{%
\begin{tabular}{@{}lcc|c@{}}
\toprule
\textbf{Method} & \textbf{Base} & \textbf{Novel} & \textbf{HM} \\
\midrule
CLIP & 18.06 & 18.06 & 18.06 \\ 
BitFit & 42.43 & 30.08 & 35.19 \\
LayerNorm & 31.60 & 30.40 & 30.96 \\
SAFT & 47.15 & 30.96 & 37.37 \\
CoOp & 39.23 & 22.95 & 28.94 \\
KgCoOp & 32.67 & 27.34 & 29.74 \\
ProGrad & 36.03 & 1.68 & 3.06 \\
LoRA & 47.96 & 1.12 & 2.13 \\
PiSSA & 49.78 & 1.52 & 2.90 \\
GaLoRE & 49.08 & 27.57 & 35.28 \\
SO & 48.46 & 31.28 & 38.01 \\
\cmidrule(lr){1-4}
SignSparseSGD & 45.85 & 28.73 & 35.30 \\
DynSignSparseSGD & 49.74 & 31.96 & 38.91 \\
\bottomrule
\end{tabular}%
}
  \vspace{0.5em}
  \end{minipage}%
  \hfill
  \begin{minipage}[t]{0.32\textwidth}
  \centering
\caption*{\textbf{SUN 397}}\vspace{-1em}
\setlength{\tabcolsep}{2.5pt}
\resizebox{\linewidth}{!}{%
\begin{tabular}{@{}lcc|c@{}}
\toprule
\textbf{Method} & \textbf{Base} & \textbf{Novel} & \textbf{HM} \\
\midrule
CLIP & 62.06 & 62.06 & 62.06 \\ 
BitFit & 81.85 & 74.13 & 77.79 \\
LayerNorm & 78.93 & 75.47 & 77.15 \\
SAFT & 81.29 & 72.41 & 76.58 \\
CoOp & 81.20 & 68.51 & 74.32 \\
KgCoOp & 77.76 & 72.26 & 74.91 \\
ProGrad & 78.99 & 41.43 & 54.28 \\
LoRA & 80.98 & 23.92 & 36.89 \\
PiSSA & 80.22 & 26.22 & 39.52 \\
GaLoRE & 79.70 & 70.66 & 74.90 \\
SO & 81.33 & 71.61 & 76.15 \\
\cmidrule(lr){1-4}
SignSparseSGD & 80.98 & 71.95 & 76.19 \\
DynSignSparseSGD & 81.72 & 73.54 & 77.40 \\
\bottomrule
\end{tabular}%
}
  \vspace{0.5em}
  \end{minipage}%
  \par\vspace{0.45em}
  \begin{minipage}[t]{0.32\textwidth}
  \centering
\caption*{\textbf{DTD}}\vspace{-1em}
\setlength{\tabcolsep}{2.5pt}
\resizebox{\linewidth}{!}{%
\begin{tabular}{@{}lcc|c@{}}
\toprule
\textbf{Method} & \textbf{Base} & \textbf{Novel} & \textbf{HM} \\
\midrule
CLIP & 42.97 & 42.97 & 42.97 \\ 
BitFit & 79.27 & 47.22 & 59.06 \\
LayerNorm & 67.59 & 52.20 & 58.82 \\
SAFT & 80.39 & 47.99 & 60.00 \\
CoOp & 76.49 & 42.98 & 54.99 \\
KgCoOp & 71.10 & 51.16 & 59.47 \\
ProGrad & 71.38 & 27.08 & 39.07 \\
LoRA & 80.11 & 10.96 & 19.28 \\
PiSSA & 80.68 & 12.46 & 21.56 \\
GaLoRE & 79.59 & 50.89 & 61.99 \\
SO & 79.87 & 49.23 & 60.82 \\
\cmidrule(lr){1-4}
SignSparseSGD & 80.43 & 49.65 & 61.33 \\
DynSignSparseSGD & 80.56 & 51.81 & 63.00 \\
\bottomrule
\end{tabular}%
}
  \vspace{0.5em}
  \end{minipage}%
  \hfill
  \begin{minipage}[t]{0.32\textwidth}
  \centering
\caption*{\textbf{EuroSAT}}\vspace{-1em}
\setlength{\tabcolsep}{2.5pt}
\resizebox{\linewidth}{!}{%
\begin{tabular}{@{}lcc|c@{}}
\toprule
\textbf{Method} & \textbf{Base} & \textbf{Novel} & \textbf{HM} \\
\midrule
CLIP & 35.81 & 35.81 & 35.81 \\ 
BitFit & 93.40 & 65.69 & 77.13 \\
LayerNorm & 69.75 & 52.78 & 59.35 \\
SAFT & 93.51 & 64.75 & 76.47 \\
CoOp & 86.39 & 45.61 & 59.45 \\
KgCoOp & 75.74 & 46.67 & 56.47 \\
ProGrad & 85.12 & 5.65 & 10.53 \\
LoRA & 92.99 & 3.26 & 6.24 \\
PiSSA & 93.55 & 0.68 & 1.35 \\
GaLoRE & 90.71 & 63.12 & 74.43 \\
SO & 93.78 & 65.27 & 76.90 \\
\cmidrule(lr){1-4}
SignSparseSGD & 93.48 & 66.60 & 77.67 \\
DynSignSparseSGD & 95.16 & 63.59 & 76.05 \\
\bottomrule
\end{tabular}%
}
  \vspace{0.5em}
  \end{minipage}%
  \hfill
  \begin{minipage}[t]{0.32\textwidth}
  \centering
\caption*{\textbf{UCF101}}\vspace{-1em}
\setlength{\tabcolsep}{2.5pt}
\resizebox{\linewidth}{!}{%
\begin{tabular}{@{}lcc|c@{}}
\toprule
\textbf{Method} & \textbf{Base} & \textbf{Novel} & \textbf{HM} \\
\midrule
CLIP & 61.64 & 61.64 & 61.64 \\ 
BitFit & 85.54 & 74.76 & 79.47 \\
LayerNorm & 78.40 & 76.25 & 77.09 \\
SAFT & 86.85 & 74.19 & 79.82 \\
CoOp & 84.52 & 68.64 & 75.65 \\
KgCoOp & 79.54 & 73.35 & 76.06 \\
ProGrad & 82.28 & 50.24 & 62.26 \\
LoRA & 87.32 & 32.10 & 46.62 \\
PiSSA & 86.54 & 29.93 & 44.21 \\
GaLoRE & 86.95 & 74.91 & 80.29 \\
SO & 87.48 & 74.34 & 80.12 \\
\cmidrule(lr){1-4}
SignSparseSGD & 86.56 & 72.96 & 78.96 \\
DynSignSparseSGD & 87.42 & 76.04 & 81.15 \\
\bottomrule
\end{tabular}%
}
  \vspace{0.5em}
  \end{minipage}
\end{table}

\begin{table}[p]
  \def\arraystretch{0.9}
  \centering
  \scriptsize
  \caption{Base-to-new 1-shot classification with CLIP(ViT-B/16). Top-1 accuracy (\%) averaged over 3 fixed random seeds.}
  \label{tab:b2n_peft_1shot_vit-b-16_with_aug}
  \begin{minipage}[t]{0.32\textwidth}
  \centering
\caption*{\textbf{Average across datasets.}}\vspace{-1em}
\setlength{\tabcolsep}{2.5pt}
\resizebox{\linewidth}{!}{%
\begin{tabular}{@{}lcc|c@{}}
\toprule
\textbf{Method} & \textbf{Base} & \textbf{Novel} & \textbf{HM} \\
\midrule
CLIP & 64.07 & 64.07 & 64.07 \\ 
BitFit & 78.63 & 74.59 & 76.44 \\
LayerNorm & 74.30 & 73.56 & 73.70 \\
SAFT & 78.44 & 74.07 & 76.00 \\
CoOp & 75.69 & 68.87 & 71.78 \\
KgCoOp & 75.28 & 72.84 & 73.81 \\
ProGrad & 71.12 & 44.46 & 52.70 \\
LoRA & 73.72 & 61.61 & 66.07 \\
PiSSA & 76.26 & 49.66 & 57.64 \\
GaLoRE & 77.29 & 72.18 & 74.49 \\
SO & 79.06 & 74.57 & 76.57 \\
\cmidrule(lr){1-4}
SignSparseSGD & 78.86 & 73.97 & 76.17 \\
DynSignSparseSGD & 79.78 & 74.47 & 76.86 \\
\bottomrule
\end{tabular}%
}
  \vspace{0.5em}
  \end{minipage}%
  \hfill
  \begin{minipage}[t]{0.32\textwidth}
  \centering
\caption*{\textbf{ImageNet}}\vspace{-1em}
\setlength{\tabcolsep}{2.5pt}
\resizebox{\linewidth}{!}{%
\begin{tabular}{@{}lcc|c@{}}
\toprule
\textbf{Method} & \textbf{Base} & \textbf{Novel} & \textbf{HM} \\
\midrule
CLIP & 66.72 & 66.72 & 66.72 \\ 
BitFit & 77.83 & 77.24 & 77.53 \\
LayerNorm & 76.83 & 76.94 & 76.88 \\
SAFT & 77.11 & 76.78 & 76.94 \\
CoOp & 75.43 & 74.10 & 74.76 \\
KgCoOp & 76.95 & 75.71 & 76.32 \\
ProGrad & 64.97 & 54.32 & 59.16 \\
LoRA & 71.14 & 67.08 & 69.05 \\
PiSSA & 70.19 & 58.37 & 63.73 \\
GaLoRE & 74.09 & 74.19 & 74.14 \\
SO & 77.29 & 76.69 & 76.99 \\
\cmidrule(lr){1-4}
SignSparseSGD & 75.81 & 75.19 & 75.50 \\
DynSignSparseSGD & 76.98 & 76.75 & 76.86 \\
\bottomrule
\end{tabular}%
}
  \vspace{0.5em}
  \end{minipage}%
  \hfill
  \begin{minipage}[t]{0.32\textwidth}
  \centering
\caption*{\textbf{Caltech101}}\vspace{-1em}
\setlength{\tabcolsep}{2.5pt}
\resizebox{\linewidth}{!}{%
\begin{tabular}{@{}lcc|c@{}}
\toprule
\textbf{Method} & \textbf{Base} & \textbf{Novel} & \textbf{HM} \\
\midrule
CLIP & 93.31 & 93.31 & 93.31 \\ 
BitFit & 97.22 & 96.89 & 97.05 \\
LayerNorm & 96.94 & 96.65 & 96.78 \\
SAFT & 96.90 & 96.92 & 96.90 \\
CoOp & 95.68 & 93.65 & 94.60 \\
KgCoOp & 96.56 & 96.54 & 96.54 \\
ProGrad & 93.21 & 79.50 & 85.73 \\
LoRA & 95.91 & 91.38 & 93.59 \\
PiSSA & 96.20 & 84.57 & 90.00 \\
GaLoRE & 96.28 & 96.54 & 96.40 \\
SO & 97.32 & 96.89 & 97.10 \\
\cmidrule(lr){1-4}
SignSparseSGD & 96.87 & 96.66 & 96.75 \\
DynSignSparseSGD & 97.32 & 96.66 & 96.98 \\
\bottomrule
\end{tabular}%
}
  \vspace{0.5em}
  \end{minipage}%
  \par\vspace{0.45em}
  \begin{minipage}[t]{0.32\textwidth}
  \centering
\caption*{\textbf{Oxford Flowers}}\vspace{-1em}
\setlength{\tabcolsep}{2.5pt}
\resizebox{\linewidth}{!}{%
\begin{tabular}{@{}lcc|c@{}}
\toprule
\textbf{Method} & \textbf{Base} & \textbf{Novel} & \textbf{HM} \\
\midrule
CLIP & 67.40 & 67.40 & 67.40 \\ 
BitFit & 82.51 & 76.68 & 79.48 \\
LayerNorm & 73.25 & 77.98 & 75.49 \\
SAFT & 88.19 & 75.53 & 81.33 \\
CoOp & 85.24 & 76.76 & 80.74 \\
KgCoOp & 74.02 & 77.87 & 75.85 \\
ProGrad & 80.88 & 40.27 & 53.60 \\
LoRA & 74.39 & 68.35 & 71.09 \\
PiSSA & 88.31 & 53.65 & 66.65 \\
GaLoRE & 88.31 & 73.92 & 80.42 \\
SO & 84.04 & 76.95 & 80.30 \\
\cmidrule(lr){1-4}
SignSparseSGD & 89.11 & 76.12 & 82.08 \\
DynSignSparseSGD & 89.99 & 76.40 & 82.61 \\
\bottomrule
\end{tabular}%
}
  \vspace{0.5em}
  \end{minipage}%
  \hfill
  \begin{minipage}[t]{0.32\textwidth}
  \centering
\caption*{\textbf{Oxford Pets}}\vspace{-1em}
\setlength{\tabcolsep}{2.5pt}
\resizebox{\linewidth}{!}{%
\begin{tabular}{@{}lcc|c@{}}
\toprule
\textbf{Method} & \textbf{Base} & \textbf{Novel} & \textbf{HM} \\
\midrule
CLIP & 88.25 & 88.25 & 88.25 \\ 
BitFit & 95.86 & 91.15 & 93.43 \\
LayerNorm & 94.95 & 91.50 & 93.14 \\
SAFT & 94.07 & 90.79 & 92.38 \\
CoOp & 95.28 & 90.81 & 92.95 \\
KgCoOp & 96.44 & 91.25 & 93.75 \\
ProGrad & 93.81 & 65.11 & 76.46 \\
LoRA & 94.44 & 84.27 & 89.03 \\
PiSSA & 94.32 & 72.20 & 81.76 \\
GaLoRE & 94.36 & 90.59 & 92.43 \\
SO & 96.05 & 91.68 & 93.79 \\
\cmidrule(lr){1-4}
SignSparseSGD & 95.19 & 91.09 & 93.08 \\
DynSignSparseSGD & 95.34 & 91.77 & 93.50 \\
\bottomrule
\end{tabular}%
}
  \vspace{0.5em}
  \end{minipage}%
  \hfill
  \begin{minipage}[t]{0.32\textwidth}
  \centering
\caption*{\textbf{Stanford Cars}}\vspace{-1em}
\setlength{\tabcolsep}{2.5pt}
\resizebox{\linewidth}{!}{%
\begin{tabular}{@{}lcc|c@{}}
\toprule
\textbf{Method} & \textbf{Base} & \textbf{Novel} & \textbf{HM} \\
\midrule
CLIP & 65.51 & 65.51 & 65.51 \\ 
BitFit & 78.66 & 75.90 & 77.26 \\
LayerNorm & 76.38 & 75.71 & 76.05 \\
SAFT & 78.86 & 74.44 & 76.58 \\
CoOp & 76.02 & 73.14 & 74.54 \\
KgCoOp & 74.96 & 75.69 & 75.32 \\
ProGrad & 64.88 & 41.63 & 50.56 \\
LoRA & 71.21 & 61.98 & 66.26 \\
PiSSA & 76.37 & 49.94 & 60.37 \\
GaLoRE & 76.26 & 72.11 & 74.13 \\
SO & 78.99 & 75.89 & 77.41 \\
\cmidrule(lr){1-4}
SignSparseSGD & 78.82 & 74.44 & 76.56 \\
DynSignSparseSGD & 79.49 & 75.31 & 77.34 \\
\bottomrule
\end{tabular}%
}
  \vspace{0.5em}
  \end{minipage}
\end{table}

\begin{table}[p]
  \ContinuedFloat
  \def\arraystretch{0.9}
  \centering
  \scriptsize
  \caption[]{Base-to-new 1-shot classification with CLIP(ViT-B/16). Top-1 accuracy (\%) averaged over 3 fixed random seeds. (continued)}
  \begin{minipage}[t]{0.32\textwidth}
  \centering
\caption*{\textbf{Food 101}}\vspace{-1em}
\setlength{\tabcolsep}{2.5pt}
\resizebox{\linewidth}{!}{%
\begin{tabular}{@{}lcc|c@{}}
\toprule
\textbf{Method} & \textbf{Base} & \textbf{Novel} & \textbf{HM} \\
\midrule
CLIP & 85.23 & 85.23 & 85.23 \\ 
BitFit & 89.71 & 89.66 & 89.68 \\
LayerNorm & 89.93 & 90.32 & 90.11 \\
SAFT & 88.12 & 88.58 & 88.35 \\
CoOp & 88.80 & 86.64 & 87.70 \\
KgCoOp & 90.50 & 90.23 & 90.36 \\
ProGrad & 85.26 & 58.73 & 69.14 \\
LoRA & 86.55 & 83.84 & 85.17 \\
PiSSA & 85.07 & 74.05 & 79.17 \\
GaLoRE & 86.99 & 87.88 & 87.43 \\
SO & 90.07 & 90.06 & 90.06 \\
\cmidrule(lr){1-4}
SignSparseSGD & 88.68 & 89.08 & 88.88 \\
DynSignSparseSGD & 89.74 & 90.01 & 89.87 \\
\bottomrule
\end{tabular}%
}
  \vspace{0.5em}
  \end{minipage}%
  \hfill
  \begin{minipage}[t]{0.32\textwidth}
  \centering
\caption*{\textbf{FGVC Aircraft}}\vspace{-1em}
\setlength{\tabcolsep}{2.5pt}
\resizebox{\linewidth}{!}{%
\begin{tabular}{@{}lcc|c@{}}
\toprule
\textbf{Method} & \textbf{Base} & \textbf{Novel} & \textbf{HM} \\
\midrule
CLIP & 23.91 & 23.91 & 23.91 \\ 
BitFit & 39.45 & 37.32 & 38.35 \\
LayerNorm & 34.81 & 36.48 & 35.60 \\
SAFT & 39.83 & 34.20 & 36.79 \\
CoOp & 37.97 & 30.60 & 33.87 \\
KgCoOp & 37.79 & 35.16 & 36.40 \\
ProGrad & 33.67 & 5.76 & 9.69 \\
LoRA & 30.96 & 21.55 & 25.41 \\
PiSSA & 35.67 & 9.29 & 14.74 \\
GaLoRE & 37.83 & 31.10 & 34.11 \\
SO & 40.25 & 37.80 & 38.99 \\
\cmidrule(lr){1-4}
SignSparseSGD & 38.47 & 35.00 & 36.63 \\
DynSignSparseSGD & 40.69 & 35.20 & 37.75 \\
\bottomrule
\end{tabular}%
}
  \vspace{0.5em}
  \end{minipage}%
  \hfill
  \begin{minipage}[t]{0.32\textwidth}
  \centering
\caption*{\textbf{SUN 397}}\vspace{-1em}
\setlength{\tabcolsep}{2.5pt}
\resizebox{\linewidth}{!}{%
\begin{tabular}{@{}lcc|c@{}}
\toprule
\textbf{Method} & \textbf{Base} & \textbf{Novel} & \textbf{HM} \\
\midrule
CLIP & 62.56 & 62.56 & 62.56 \\ 
BitFit & 78.57 & 75.45 & 76.97 \\
LayerNorm & 76.11 & 74.66 & 75.37 \\
SAFT & 78.35 & 74.79 & 76.52 \\
CoOp & 75.38 & 70.75 & 72.97 \\
KgCoOp & 75.61 & 72.40 & 73.96 \\
ProGrad & 67.27 & 50.46 & 57.52 \\
LoRA & 71.98 & 63.80 & 67.64 \\
PiSSA & 72.18 & 55.70 & 62.87 \\
GaLoRE & 76.38 & 73.61 & 74.96 \\
SO & 79.05 & 75.59 & 77.27 \\
\cmidrule(lr){1-4}
SignSparseSGD & 77.50 & 74.02 & 75.71 \\
DynSignSparseSGD & 78.78 & 75.41 & 77.05 \\
\bottomrule
\end{tabular}%
%
}
  \vspace{0.5em}
  \end{minipage}
\end{table}

Tables~\ref{tab:b2n_peft_1shot_vit-b-32_with_aug}--\ref{tab:b2n_peft_16shot_vit-b-32_with_aug} and \ref{tab:b2n_peft_1shot_vit-b-16_with_aug}--\ref{tab:b2n_peft_16shot_vit-b-16_with_aug} report the B2N results for CLIP with ViT-B/32 and ViT-B/16. DynSignSparseSGD attains the highest average harmonic mean (HM) in all 10 backbone--shot settings. On ViT-B/32, its HM increases from 72.84\% at 1 shot to 76.57\% at 16 shots. Its average gain over the strongest baseline at each shot level is 0.49 points. On ViT-B/16, its HM increases from 76.86\% to 80.79\%, with an average gain of 0.60 points. In the 16-shot ViT-B/16 setting highlighted in the main paper, DynSignSparseSGD obtains 88.12\% \emph{Base}, 75.30\% \emph{New}, and 80.79\% HM. The strongest baseline values in the three respective columns are 87.47\% \emph{Base} for SO, 74.79\% \emph{New} for LayerNorm, and 79.72\% HM for BitFit.

Refreshing the support consistently improves the static variant. Averaged over the five shot levels, DynSignSparseSGD improves over SignSparseSGD by 1.46 HM points on ViT-B/32 and 0.88 points on ViT-B/16. At 16 shots, it attains the highest per-dataset HM on 6 of 11 datasets with ViT-B/32 and on 8 of 11 datasets with ViT-B/16. At 1 shot, SO is the strongest non-sign-sparse baseline in average HM for both backbones. DynSignSparseSGD remains higher, reaching 72.84\% versus 72.62\% on ViT-B/32 and 76.86\% versus 76.57\% on ViT-B/16.

The \emph{New}-class results are stable across shot levels. For DynSignSparseSGD, the range of the average \emph{New} accuracy is 0.93 points on ViT-B/32 and 0.83 points on ViT-B/16, while the \emph{Base} accuracy increases as more labeled examples become available. SignSparseSGD follows the same general pattern. GaLoRE is also comparatively stable, but its average HM remains below both sign-sparse methods at every shot level. In contrast, LoRA and PiSSA increasingly favor the base classes. On ViT-B/16, LoRA decreases from 61.61\% \emph{New} and 66.07\% HM at 1 shot to 35.03\% \emph{New} and 46.65\% HM at 16 shots. PiSSA decreases from 49.66\% \emph{New} and 57.64\% HM to 33.34\% and 44.70\%, respectively. Under the shared protocol, DynSignSparseSGD therefore combines increasing \emph{Base} accuracy with stable \emph{New} accuracy and attains the highest average HM throughout the evaluated sweep.

\section{Stability Analysis}
\label{app:stability}
\suppanchor{stability}

\begin{figure}[!t]
  \centering
  \includegraphics[width=0.94\textwidth,height=0.80\textheight,keepaspectratio]{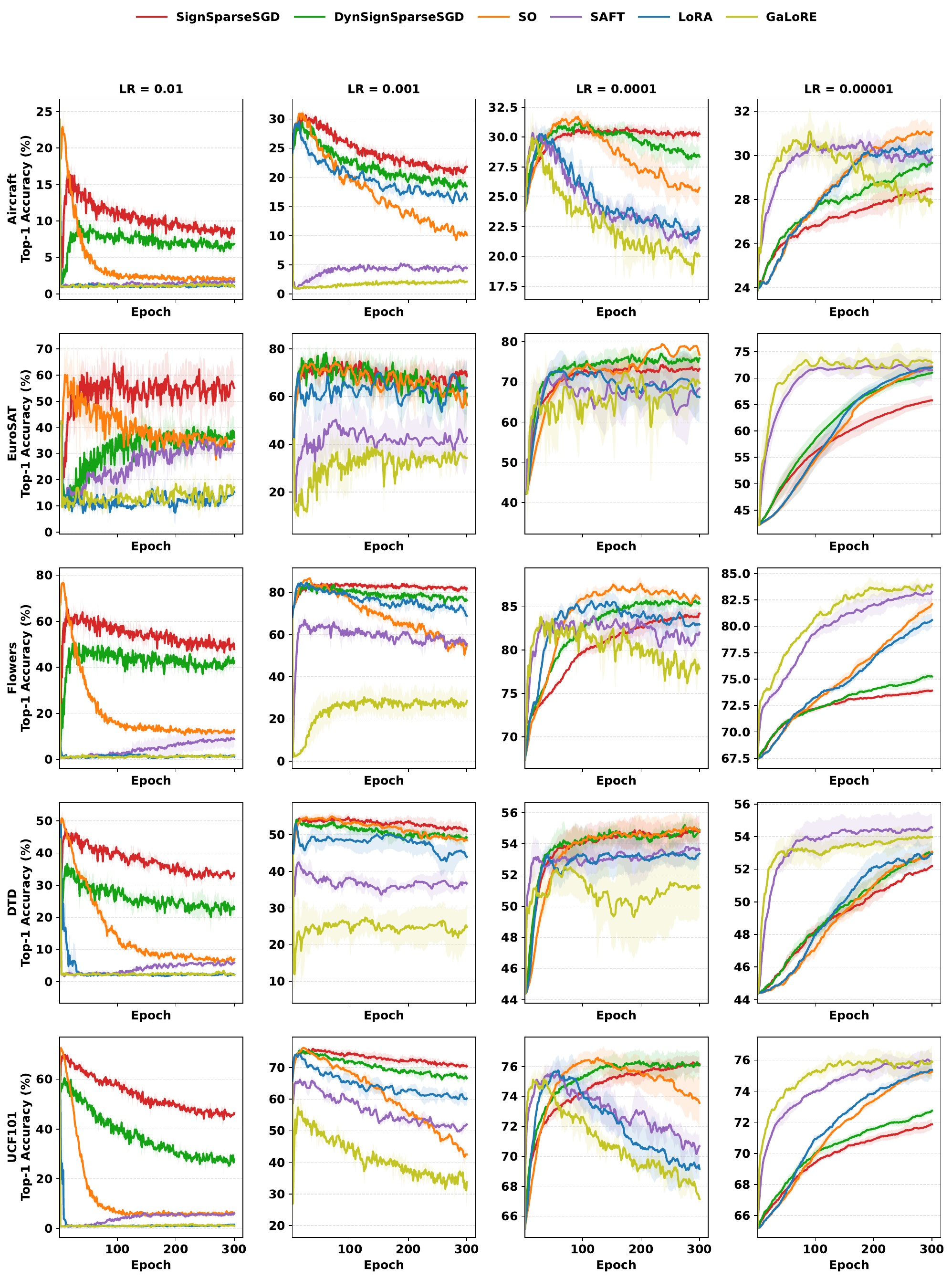}
  \caption{Test accuracy (mean $\pm$ std over 3 seeds) over training epochs for SignSparseSGD, DynSignSparseSGD, LoRA, GaLoRE, SAFT, and SO in a 1-shot setting on five datasets (Aircraft, EuroSAT, Flowers, DTD, and UCF101) using CLIP (ViT-B/16). Each column corresponds to one learning rate: $10^{-2}$, $10^{-3}$, $10^{-4}$, and $10^{-5}$.}
  \label{fig:stability_group1_with_aug}
\end{figure}

\begin{figure}[!t]
  \centering
  \includegraphics[width=0.94\textwidth,height=0.80\textheight,keepaspectratio]{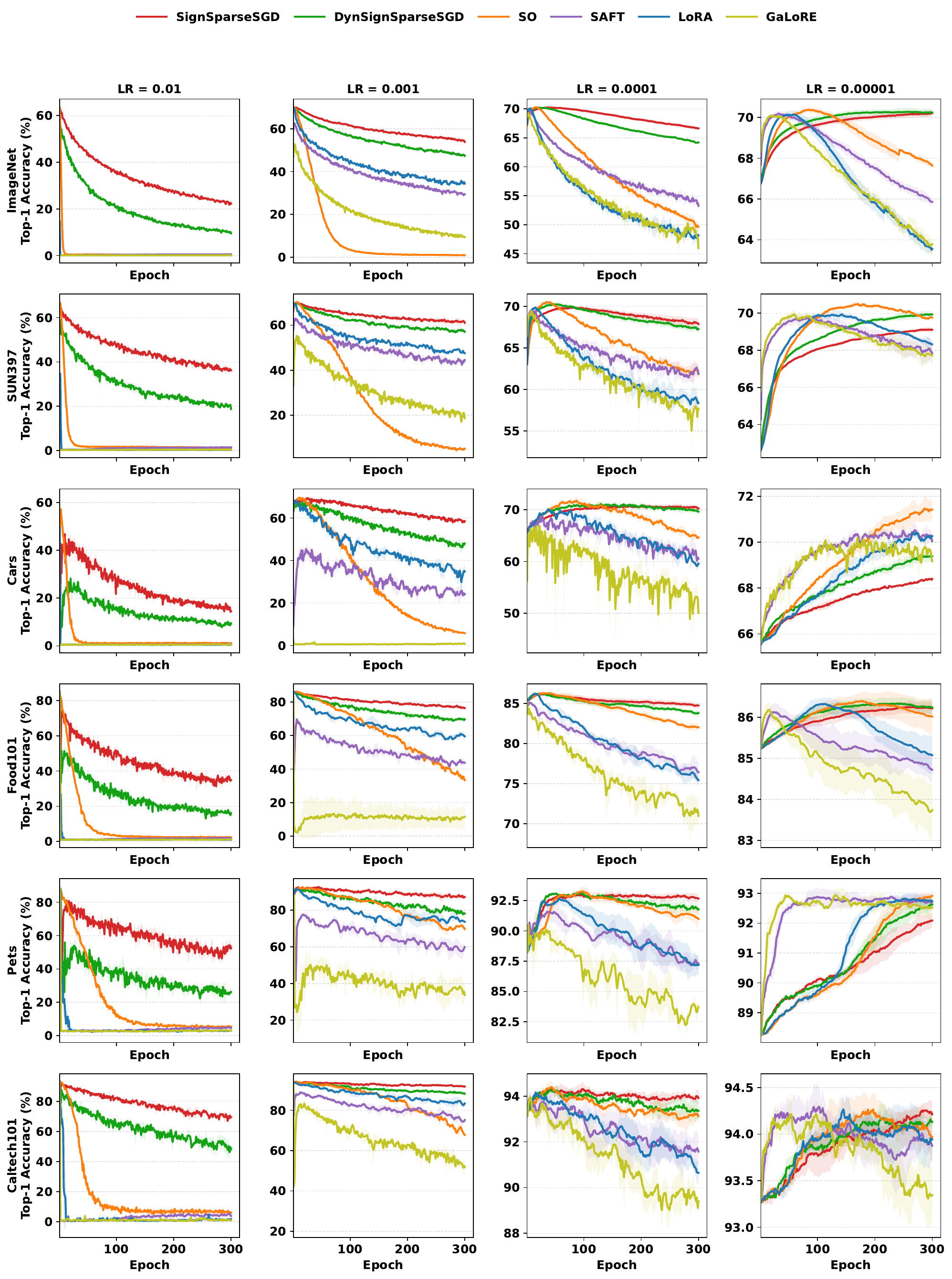}
  \caption{Test accuracy (mean $\pm$ std over 3 seeds) over training epochs for SignSparseSGD, DynSignSparseSGD, LoRA, GaLoRE, SAFT, and SO in a 1-shot setting on six datasets (ImageNet, SUN397, Cars, Food101, Pets, and Caltech101) using CLIP (ViT-B/16). Each column corresponds to one learning rate: $10^{-2}$, $10^{-3}$, $10^{-4}$, and $10^{-5}$.}
  \label{fig:stability_group2_with_aug}
\end{figure}

To complement Fig.~\ref{fig:long_horizon} in the main paper, Fig.~\ref{fig:stability_group1_with_aug} and Fig.~\ref{fig:stability_group2_with_aug} extend the 1-shot CLIP (ViT-B/16) analysis from ImageNet and SUN397 to all 11 datasets. We plot test accuracy (mean $\pm$ std over 3 fixed seeds) for SignSparseSGD, DynSignSparseSGD, LoRA, GaLoRE, SAFT, and SO under four learning rates ($10^{-2}$, $10^{-3}$, $10^{-4}$, and $10^{-5}$). Unlike the fixed 30-epoch budget used in the main experiments, these runs are continued up to 300 epochs in order to expose late-epoch overfitting directly. The goal of this appendix is therefore not to redefine the training protocol of Sec.~\ref{sec:experiments}, but to visualize how sensitive each method is to the training budget once optimization is allowed to continue for much longer.

\begin{table}
\centering
\caption{Average peak-to-end accuracy drop (percentage points) across the 11 datasets in Fig.~\ref{fig:stability_group1_with_aug} and Fig.~\ref{fig:stability_group2_with_aug}. Lower is better.}
\label{tab:stability_drop_summary}
\setlength{\tabcolsep}{4.5pt}
\begin{tabular}{lcccc}
\toprule
Method & $10^{-2}$ & $10^{-3}$ & $10^{-4}$ & $10^{-5}$ \\
\midrule
SignSparseSGD    & 26.68 & 7.30  & 0.93 & 0.01 \\
DynSignSparseSGD & 35.42 & 12.75 & 1.85 & 0.03 \\
LoRA             & 53.21 & 18.60 & 7.97 & 0.92 \\
GaLoRE           & 43.81 & 33.74 & 9.58 & 1.61 \\
SAFT             & 40.33 & 20.16 & 6.06 & 0.97 \\
SO               & 59.92 & 37.39 & 5.26 & 0.39 \\
\bottomrule
\end{tabular}
\end{table}

\paragraph{Overall trend.} The main trend at $10^{-4}$ is consistent across the 11 datasets. The sign-sparse optimizers exhibit the smallest peak-to-end degradation. Here, the peak-to-end drop is defined as the difference between the best test accuracy reached during training and the final accuracy at epoch 300. Averaged over the 11 datasets, this drop is $0.93$ points for SignSparseSGD and $1.85$ points for DynSignSparseSGD. The corresponding values are $5.26$ for SO, $6.06$ for SAFT, $7.97$ for LoRA, and $9.58$ for GaLoRE, as reported in Tab.~\ref{tab:stability_drop_summary}. SignSparseSGD remains within 5 points of its peak on all 11 datasets, and DynSignSparseSGD does so on 10 out of 11 datasets. The corresponding counts are 7 out of 11 for SO, 6 out of 11 for SAFT, and 3 out of 11 for LoRA and GaLoRE. The standard deviation bands are also small. At $10^{-4}$, the mean per-epoch standard deviation is $0.55$ for both sign-sparse methods, compared with $1.14$ for LoRA, $1.21$ for SAFT, and $1.69$ for GaLoRE. SO has similar variance ($0.56$) but larger late-epoch accuracy degradation. These observations are consistent with reduced sensitivity to prolonged training.

\paragraph{Dataset-level behavior.} The difference is most pronounced on the more difficult datasets in Fig.~\ref{fig:stability_group2_with_aug}. On ImageNet at $10^{-4}$, the peak-to-end drop is $3.61$ points for SignSparseSGD and $6.02$ points for DynSignSparseSGD. The corresponding drops are $21.98$ for LoRA, $23.31$ for GaLoRE, $16.66$ for SAFT, and $20.65$ for SO. On SUN397, the drops are $2.03$ and $2.99$ for the two sign-sparse methods, compared with $11.47$, $11.76$, $7.68$, and $8.65$ for LoRA, GaLoRE, SAFT, and SO, respectively. On Cars, the drops are $0.47$ and $1.44$ for SignSparseSGD and DynSignSparseSGD, while the baselines lose between $7.09$ and $16.92$ points. A similar pattern is observed on the datasets in Fig.~\ref{fig:stability_group1_with_aug}, although the absolute gap is smaller on the easier benchmarks. On Aircraft at $10^{-4}$, the peak-to-end drop is $0.55$ for SignSparseSGD and $2.88$ for DynSignSparseSGD, compared with $5.97$ for SO and approximately $8$--$10$ points for LoRA, GaLoRE, and SAFT. On UCF101, the corresponding drops are $0.12$ and $0.32$, compared with $3.02$ for SO, $4.86$ for SAFT, $6.58$ for LoRA, and $8.01$ for GaLoRE. On EuroSAT, Flowers, and DTD, several methods remain competitive, but the sign-sparse curves still show smaller late-epoch degradation. Therefore, the stability trend is not restricted to a single dataset family.

\paragraph{Static versus dynamic support.} The comparison between SignSparseSGD and DynSignSparseSGD is also informative. At $10^{-4}$ and epoch 30, DynSignSparseSGD attains slightly higher average accuracy across the 11 datasets ($71.03\%$ vs.\ $70.01\%$). This is consistent with the fixed-budget protocol used in the main paper. By epoch 300, SignSparseSGD becomes slightly better on average ($72.25\%$ vs.\ $71.88\%$). From epoch 30 to epoch 300, SignSparseSGD improves on 7 of 11 datasets and does not incur a drop larger than 5 points. DynSignSparseSGD improves on 5 of 11 datasets and exceeds a 5-point drop only on ImageNet. The same comparison yields 7 of 11, 8 of 11, and 5 of 11 datasets with drops larger than 5 points for LoRA, GaLoRE, and SAFT, respectively. These results suggest that the dynamic mask is slightly preferable under a short fixed budget, while the static mask yields slightly smaller degradation under extended training.

\paragraph{Effect of the learning rate.} The learning-rate sweep supports the same interpretation. At $10^{-5}$, all methods train slowly, and late-epoch degradation is limited. The mean peak epoch is 292 for SignSparseSGD and 280 for DynSignSparseSGD, which suggests under-training rather than overfitting. At $10^{-3}$, late-epoch degradation becomes more visible, but the sign-sparse methods remain the most stable. In this regime, SignSparseSGD is the only method whose final average accuracy across the 11 datasets remains above the zero-shot baseline, with an average margin of $+1.62$ points. At $10^{-2}$, all methods operate outside the useful regime, but SignSparseSGD still attains the highest final average accuracy and the smallest average peak-to-end drop. Overall, the long-horizon curves are consistent with the interpretation in the main paper: sign-sparse updates reduce sensitivity to both training duration and learning rate, and they mitigate overfitting relative to LoRA, GaLoRE, SAFT, and SO.

\section{Update Scale}
\label{sec:update_scale}
\suppanchor{update-scale}

\subsection{Measuring Update Magnitude}
\label{sec:update_scale_measure}
\suppanchor{update-measure}

In this Appendix, we analyze the size of the parameter updates produced by related optimization techniques. For a 1-shot run on DTD, EuroSAT, FGVC-Aircraft, and UCF101 with CLIP (ViT-B/16), we log only the visual-encoder update. Let $\Theta_{\mathrm{img},t}$ denote the visual-encoder parameters at iteration $t$ and let $d_{\mathrm{img}}$ be their dimension. We use
\begin{equation}
\Delta \Theta_{\mathrm{img},t}
=\Theta_{\mathrm{img},t+1}-\Theta_{\mathrm{img},t},
\qquad
r_t^{\mathrm{rel}}
=\frac{\|\Delta \Theta_{\mathrm{img},t}\|_2}{\|\Theta_{\mathrm{img},t}\|_2},
\qquad
r_t^{\mathrm{abs}}
=\frac{1}{d_{\mathrm{img}}}\|\Delta \Theta_{\mathrm{img},t}\|_1,
\end{equation}
together with the visual-encoder update density. Here, $r_t^{\mathrm{rel}}$ is the most relevant quantity for comparing dense and sparse methods because it measures the actual drift of the visual-encoder parameter vector.

\begin{figure}[!h]
  \centering
  \begin{subfigure}[t]{0.49\textwidth}
    \centering
    \includegraphics[width=\linewidth]{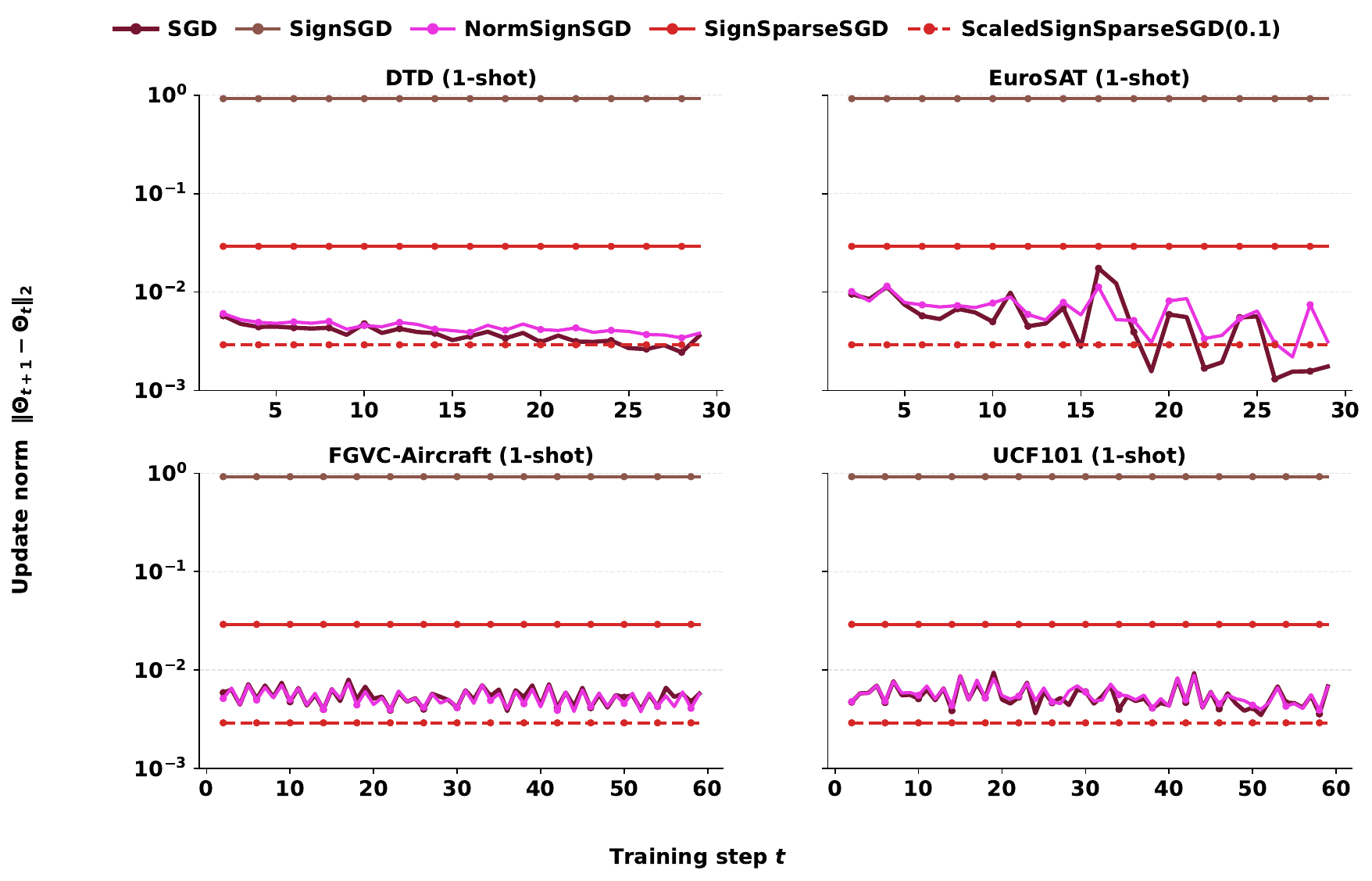}
    \caption{Absolute update norm $\|\Delta\Theta_{\mathrm{img},t}\|_2$.}
    \label{fig:update_scale_abs}
  \end{subfigure}
  \hfill
  \begin{subfigure}[t]{0.49\textwidth}
    \centering
    \includegraphics[width=\linewidth]{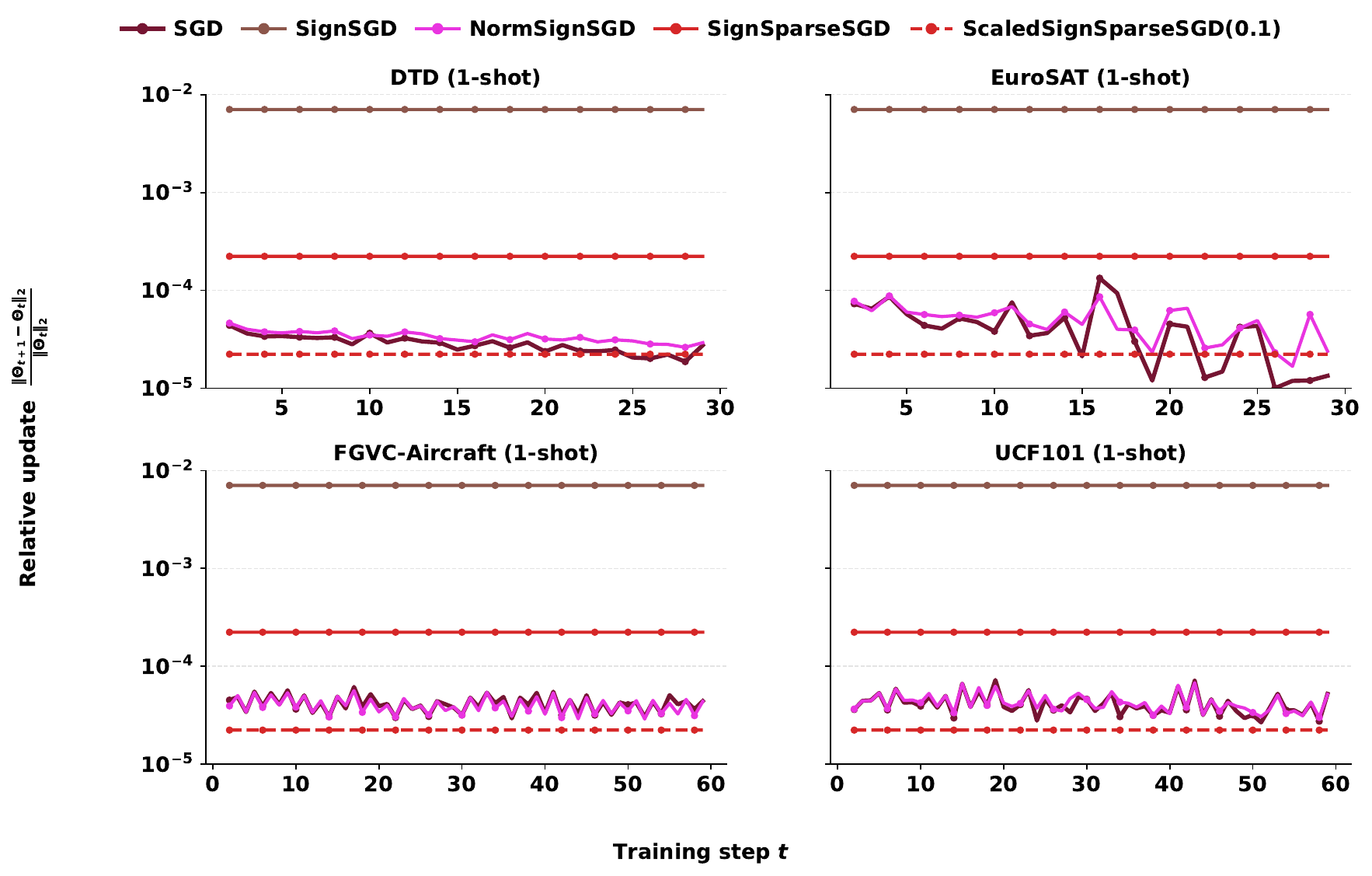}
    \caption{Relative update $r_t^{\mathrm{rel}}$.}
    \label{fig:update_scale_rel}
  \end{subfigure}

  \vspace{0.4em}

  \begin{subfigure}[t]{0.49\textwidth}
    \centering
    \includegraphics[width=\linewidth]{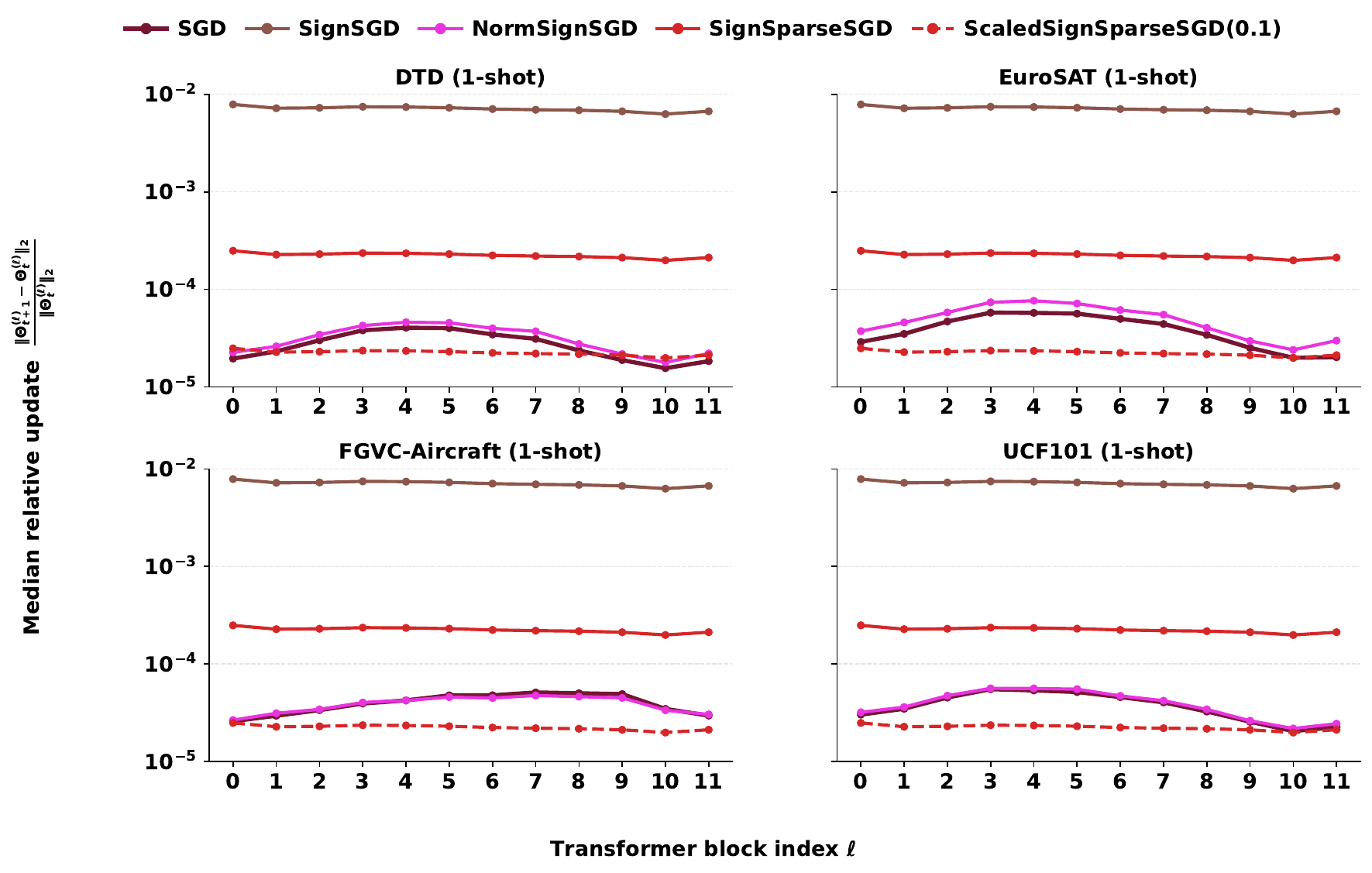}
    \caption{Median per-block relative update across training.}
    \label{fig:update_scale_blocks}
  \end{subfigure}
  \hfill
  \begin{subfigure}[t]{0.49\textwidth}
    \centering
    \includegraphics[width=\linewidth]{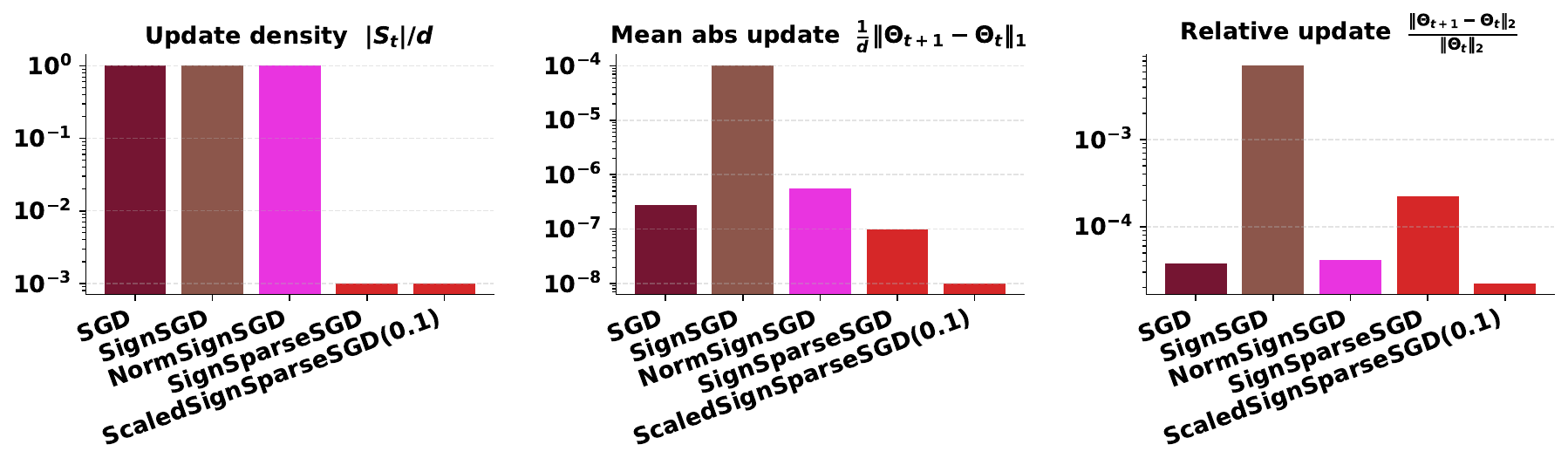}
    \caption{Summary over the four datasets.}
    \label{fig:update_scale_summary}
  \end{subfigure}
  \caption{Update-scale diagnostics on the visual encoder of CLIP (ViT-B/16) for the 1-shot classification. Dense methods have update density $1$, while sparse methods use $\rho=0.1\%$. The curve ``ScaledSignSparseSGD(0.1)'' corresponds exactly to the SignSparseSGD used in the main paper. The curve ``SignSparseSGD'' uses nonzero sign magnitude $\pm 1$ and is shown only as an ablation of the scaling factor.}
  \label{fig:update_scale_diagnostics}
\end{figure}

To avoid ambiguity, the curve labeled ``ScaledSignSparseSGD(0.1)'' in Fig.~\ref{fig:update_scale_diagnostics} is exactly the optimizer used in the main paper: it uses the unit-valued sign direction from Eq.~\eqref{eq:sparse_sign_update} and the separate scale $\alpha=0.1$ from Eq.~\eqref{eq:sparse_sign_update}. The curve labeled ``SignSparseSGD'' is a diagnostic variant with $\alpha=1$. It is included only to isolate the role of the sparse sign scale.

\begin{figure}[!h]
  \centering
  \resizebox{0.98\textwidth}{!}{\includegraphics{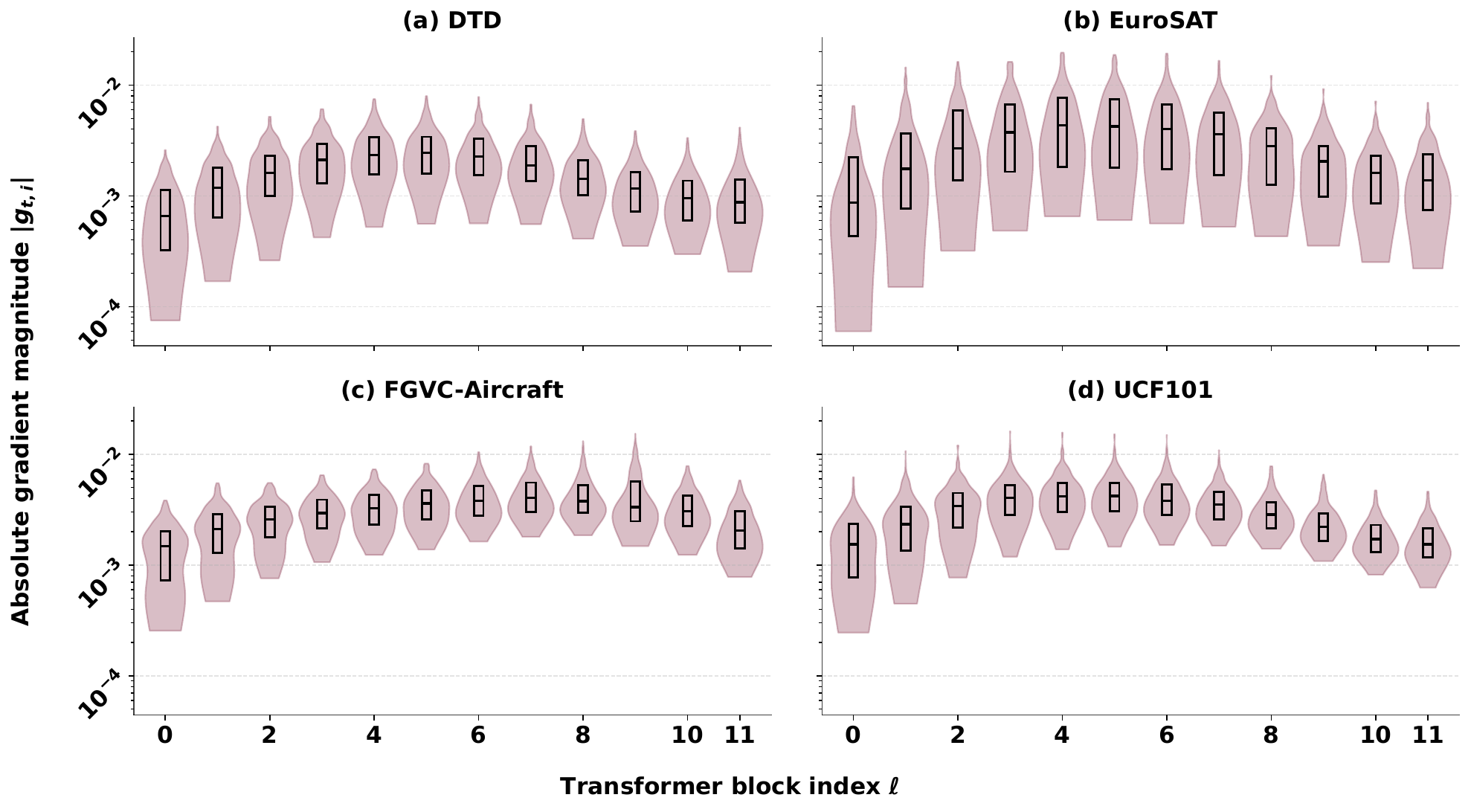}}
  \caption{Absolute gradient magnitude $|g_{t,i}|$ across the 12 visual Transformer blocks of CLIP (ViT-B/16) under dense SGD in the 1-shot regime. Each violin aggregates the logged absolute gradients.}
  \label{fig:update_scale_gradients}
\end{figure}

Fig.~\ref{fig:update_scale_gradients} shows two stable facts. First, the gradients are small in absolute value. Most of the mass lies between $10^{-4}$ and $10^{-2}$, and the medians are typically in the low $10^{-3}$ range. With the shared learning rate $\eta=10^{-4}$, dense SGD therefore produces coordinate-wise parameter changes that are usually in the $10^{-7}$ to $10^{-6}$ range. Second, the scale is not uniform across depth. On all four datasets, the median gradient magnitude increases from the first blocks to the middle of the visual Transformer, then decreases again near the output blocks. Across datasets, the median is roughly $6.6\times 10^{-4}$ to $1.5\times 10^{-3}$ at block $0$, peaks around blocks $4$--$7$ at about $2.4\times 10^{-3}$ to $4.3\times 10^{-3}$, and falls back to about $8.8\times 10^{-4}$ to $2.1\times 10^{-3}$ at block $11$.

This figure characterizes the scale of the \emph{raw gradients}, not the scale of the final optimizer step. This distinction is important for SignSparseSGD. In dense SGD, the update magnitude is directly proportional to the gradient magnitude. In SignSparseSGD, however, the update is obtained after two additional operations: only a sparse subset of coordinates is kept, and each selected coordinate is assigned a fixed sign magnitude. Therefore, the relevant quantity is not a typical coordinate-wise gradient value, but the \emph{overall model update} after sparsification and sign scaling. For this reason, we next measure the whole-model update scale induced by each optimizer.

Fig.~\ref{fig:update_scale_diagnostics}(a) and Fig.~\ref{fig:update_scale_diagnostics}(b) show that SignSGD operates at a much larger scale than dense SGD under the same learning rate. This is expected. Dense SignSGD updates every visual parameter by $\pm \eta$. Since the visual encoder has about $d_{\mathrm{img}} \approx 8.49\times 10^{7}$ trainable parameters, its update norm is fixed at
\begin{equation}
\|\Delta \Theta_{\mathrm{img},t}\|_2 \approx \eta \sqrt{d_{\mathrm{img}}} \approx 9.2\times 10^{-1},
\end{equation}
which is exactly the flat high-magnitude curve observed in Fig.~\ref{fig:update_scale_abs}. In contrast, dense SGD stays in the $10^{-3}$ regime. Averaged over the four datasets, the median update norm is about $4.9\times 10^{-3}$ for SGD, while it is about $9.2\times 10^{-1}$ for SignSGD. The same gap appears in relative scale. The median relative update is about $3.8\times 10^{-5}$ for SGD and about $7.0\times 10^{-3}$ for SignSGD. SignSGD therefore changes the parameter vector by roughly two orders of magnitude more than SGD at every step. This explains why it is too aggressive under the shared learning-rate budget.

Sparsity alone reduces the sign-update scale, but it does not solve the problem completely. With $\rho=0.1\%$, the sparse support contains about $k_{\mathrm{img}} \approx 8.49\times 10^{4}$ active visual parameters, so the unscaled sparse sign step satisfies:
\begin{equation}
\|\Delta \Theta_{\mathrm{img},t}\|_2 \approx \eta \sqrt{k_{\mathrm{img}}} \approx 2.9\times 10^{-2}.
\end{equation}
This is much smaller than SignSGD, but it is still too large compared with SGD. In Fig.~\ref{fig:update_scale_abs}, the diagnostic variant ``SignSparseSGD'' with nonzero sign value $\pm 1$ is flat at about $2.9\times 10^{-2}$, which remains about $5.9\times$ above the SGD median. The same ordering appears in Fig.~\ref{fig:update_scale_rel}, where the median relative update is about $2.2\times 10^{-4}$ for the unscaled sparse sign variant, compared with about $3.8\times 10^{-5}$ for SGD. So, sparsity helps, but the sign magnitude is still too large.

Multiplying the nonzero sign value by $0.1$ moves the sparse sign update into the correct regime. For ``ScaledSignSparseSGD(0.1)'', which is the actual SignSparseSGD used in the main paper, the update norm becomes:
\begin{equation}
\|\Delta \Theta_{\mathrm{img},t}\|_2 \approx 0.1\,\eta \sqrt{k_{\mathrm{img}}} \approx 2.9\times 10^{-3}.
\end{equation}
This is now in the same $10^{-3}$ band as dense SGD and GradNorm SignSGD. In relative terms, the average median update is about $2.2\times 10^{-5}$ for ScaledSignSparseSGD(0.1), $3.8\times 10^{-5}$ for SGD, and $4.2\times 10^{-5}$ for GradNorm SignSGD. The scaled sparse sign update is therefore slightly conservative, but it is clearly comparable to dense SGD. By contrast, the $\pm 1$ sparse sign update remains in the $10^{-4}$ band, and dense SignSGD remains in the $10^{-2}$ band.

Fig.~\ref{fig:update_scale_diagnostics}(c) gives the same conclusion at the block level. SGD and GradNorm SignSGD display a middle-layer hump, which mirrors the gradient profile of Fig.~\ref{fig:update_scale_gradients}. Dense SignSGD is almost flat across blocks because each layer receives a dense fixed-magnitude sign step. The sparse sign variants are also flatter than SGD because the support size is fixed by $\rho$ in every block. The key difference is the vertical scale. The unscaled sparse sign variant stays consistently above SGD in every block, while ScaledSignSparseSGD(0.1) moves to the same $10^{-5}$ band as SGD. Thus, the role of the factor $0.1$ is not local to one dataset or one layer. It corrects the update scale across the whole visual Transformer.

Fig.~\ref{fig:update_scale_diagnostics}(d) also clarifies which metric should be matched. Because sparse methods modify only $0.1\%$ of the coordinates, the mean absolute visual-encoder update $r_t^{\mathrm{abs}}=d_{\mathrm{img}}^{-1}\|\Delta \Theta_{\mathrm{img},t}\|_1$ is strongly diluted by zeros. This is why the unscaled sparse sign variant can have a \emph{smaller} mean absolute update than dense SGD, about $1.0\times 10^{-7}$ versus $2.7\times 10^{-7}$, while still producing a much larger relative $\ell_2$ displacement, about $2.2\times 10^{-4}$ versus $3.8\times 10^{-5}$. Therefore, the average absolute change per parameter is not the right notion of scale under extreme sparsity. The relevant quantity is the relative update $r_t^{\mathrm{rel}}$, which measures how far the model moves in parameter space.

Overall, these diagnostics calibrate the reference scale $\alpha_{\mathrm{ref}}=0.1$ in Eq.~\eqref{eq:sparse_sign_update} at $\rho_{\mathrm{ref}}=0.1\%$. They provide the absolute reference point that Proposition~\ref{prop:sparsity_learning_rate_coupling} cannot determine from smoothness alone. The proposition supplies the relative rescaling law across support budgets, while the measurements here show that the chosen reference point brings the sparse sign update into the same operating regime as dense SGD. Using $\alpha=1$ would keep the optimizer in an overly large-step regime. This is why the main paper uses the scaled version and why the curve ``ScaledSignSparseSGD(0.1)'' should be interpreted as the real SignSparseSGD throughout the paper.

\subsection{Effect of Gradient Normalization}
\label{effect_gradient_normalization}
\suppanchor{update-normalization}

We now test directly whether the behavior of SignSGD and SparseSGD under the shared learning-rate budget of Sec.~\ref{sec:experiments} is mainly caused by the scale of the update. Let $\Delta\Theta_t^{\mathrm{method}}$ denote the parameter update produced by the indicated method at iteration $t$. Dense SGD uses $\Delta\Theta_t^{\mathrm{SGD}}=-\eta g_t$, so $\|\Delta\Theta_t^{\mathrm{SGD}}\|_2=\eta\|g_t\|_2$. In contrast, raw SignSGD uses $\sgn(g_t)$, and raw SparseSGD uses $\widetilde g_t=m\odot g_t$. Their update norms are
\begin{equation}
\|\Delta\Theta_t^{\mathrm{SignSGD}}\|_2=\eta\|\sgn(g_t)\|_2,
\qquad
\|\Delta\Theta_t^{\mathrm{SparseSGD}}\|_2=\eta\|\widetilde g_t\|_2.
\end{equation}
These two quantities are not matched to $\eta \|g_t\|_2$ in general. GradNorm SignSGD and GradNorm SparseSGD, defined in Algorithms~\ref{alg:gradnormsignsgd} and \ref{alg:gradnormsparsesgd}, multiply the sign or sparse direction by a factor $a_t$ so that the resulting update norm matches the dense SGD norm when the direction is nonzero. A zero sparse direction instead produces a zero update. They are therefore useful reference baselines. They keep the same direction as SignSGD or SparseSGD, but remove the scale mismatch.

This point is important in few-shot CLIP adaptation. The pretrained model already provides a strong representation, so useful adaptation must stay within a narrow update budget. If the update is too large, the model drifts too quickly away from the pretrained solution. If the update is too small, the optimizer uses the 30-epoch budget inefficiently. The figures below isolate these two cases. All runs use 1-shot CLIP (ViT-B/16), three seeds, and 300 training epochs in order to visualize both convergence speed and late-epoch accuracy degradation.

\begin{figure}[!t]
  \centering
  \includegraphics[width=0.94\textwidth,height=0.80\textheight,keepaspectratio]{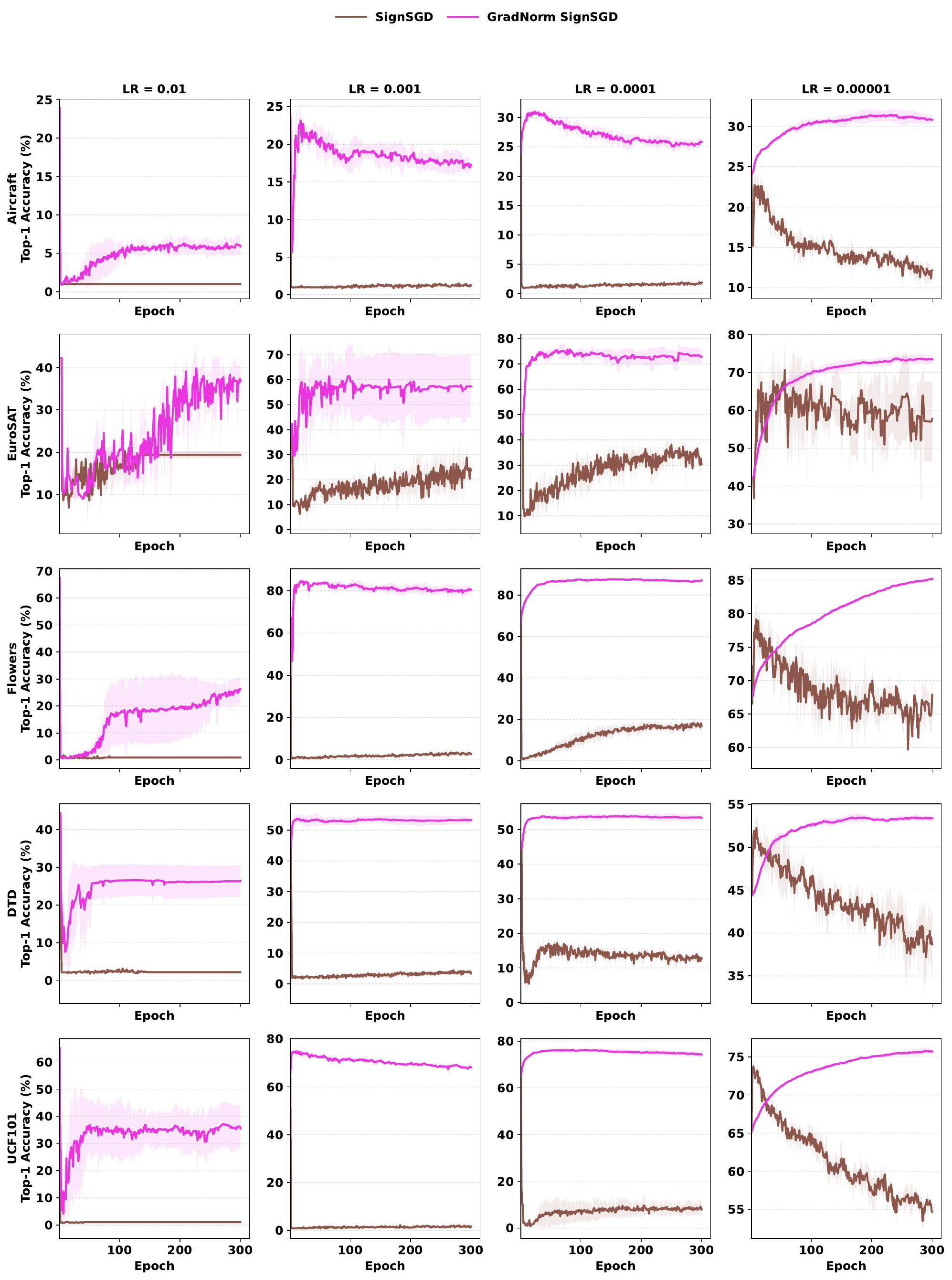}
  \caption{Test accuracy (mean $\pm$ std over 3 seeds) over epochs for SignSGD and GradNorm SignSGD in a 1-shot setting on five datasets (Aircraft, EuroSAT, Flowers, DTD, and UCF101) using CLIP (ViT-B/16). Each column corresponds to one learning rate: $10^{-2}$, $10^{-3}$, $10^{-4}$, and $10^{-5}$.}
  \label{fig:update_scale_sign_group1_with_aug}
\end{figure}

\begin{figure}[!t]
  \centering
  \includegraphics[width=0.94\textwidth,height=0.80\textheight,keepaspectratio]{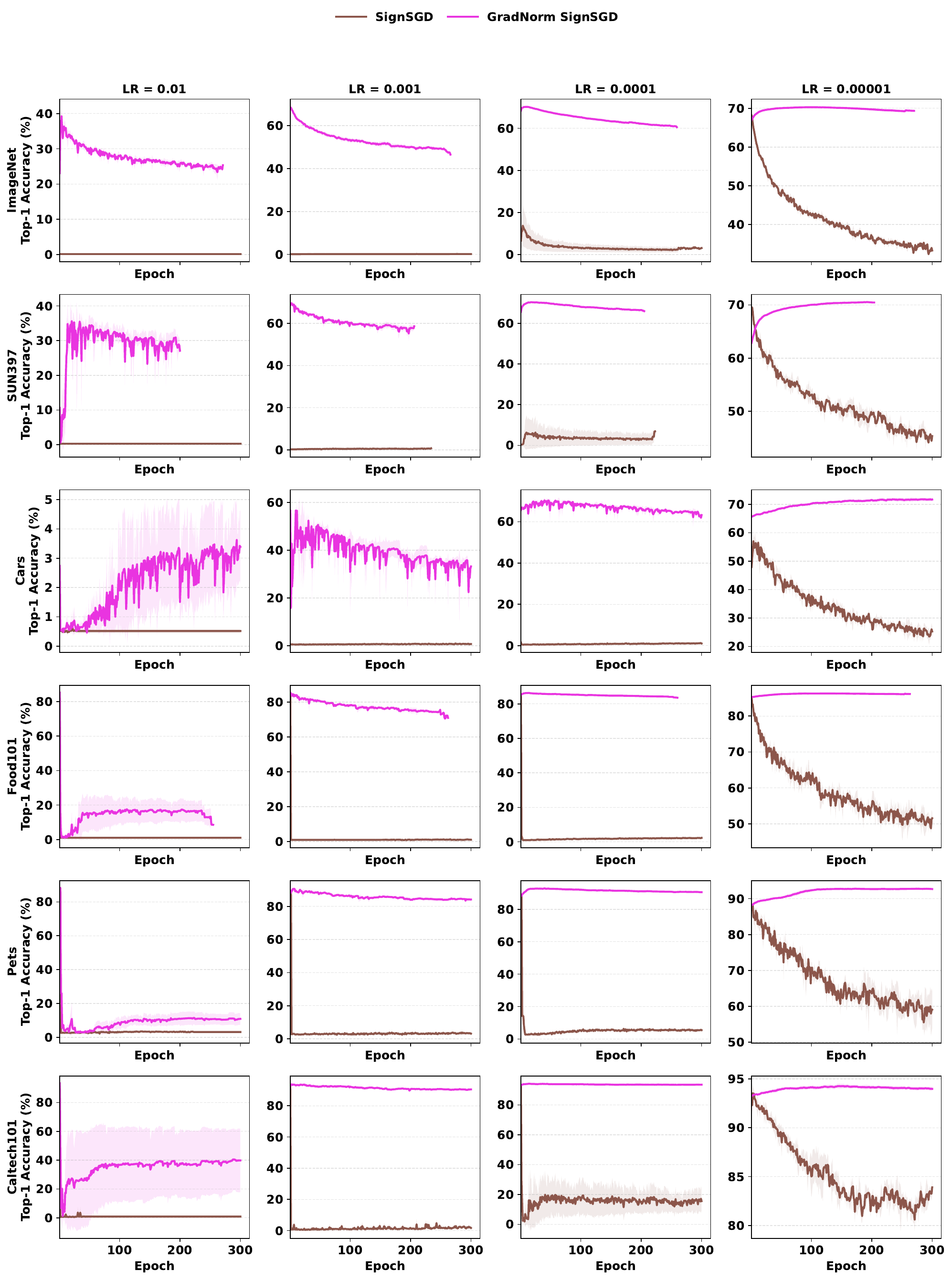}
  \caption{Test accuracy (mean $\pm$ std over 3 seeds) over epochs for SignSGD and GradNorm SignSGD in a 1-shot setting on six datasets (ImageNet, SUN397, Cars, Food101, Pets, and Caltech101) using CLIP (ViT-B/16). Each column corresponds to one learning rate: $10^{-2}$, $10^{-3}$, $10^{-4}$, and $10^{-5}$.}
  \label{fig:update_scale_sign_group2_with_aug}
\end{figure}

\paragraph{SignSGD versus GradNorm SignSGD.}
Figs.~\ref{fig:update_scale_sign_group1_with_aug} and \ref{fig:update_scale_sign_group2_with_aug} show that dense SignSGD collapses because its raw update is too large. Under Eq.~\eqref{eq:sparse_sign_update}, the nonzero entries of its direction $\sgn(g_t)$ have unit magnitude. When most coordinates are nonzero, $\|\sgn(g_t)\|_2$ is therefore approximately $\sqrt{d}$. The step norm is then driven mainly by the parameter dimension, not by the current gradient magnitude. In few-shot adaptation, where $\|g_t\|_2$ can be small and highly variable, this produces steps that are often much larger than the corresponding dense SGD step at the same learning rate. The result is rapid drift away from the pretrained initialization and a sharp drop in accuracy.

The curves make this failure mode explicit. At the main learning rate $\eta=10^{-4}$, raw SignSGD reaches only $6.1\%$ average accuracy at epoch 30 across the 11 datasets, and only $9.5\%$ at epoch 300. The norm-matched baseline removes this collapse. GradNorm SignSGD reaches $72.5\%$ on average at epoch 30 and $70.1\%$ at epoch 300. Across the 44 dataset--learning-rate settings in Figs.~\ref{fig:update_scale_sign_group1_with_aug} and \ref{fig:update_scale_sign_group2_with_aug}, GradNorm SignSGD achieves a higher final accuracy than SignSGD in all 44 cases. At $\eta=10^{-4}$, raw SignSGD peaks at epoch 1 on 9 of the 11 datasets, and by epoch 4 on ImageNet. The norm-matched variant instead follows a standard optimization trajectory and continues to improve well beyond the first few iterations. This shows that the collapse of dense SignSGD in the main ablation is primarily a scale problem.

The same interpretation also explains the other columns. At $\eta=10^{-2}$, both methods operate outside the useful regime, but GradNorm SignSGD still remains above raw SignSGD on almost every dataset. At $\eta=10^{-5}$, training is slower, yet the same ordering persists. Therefore, the sign direction itself is not the main cause of failure. Once the global norm is corrected, dense sign updates become meaningful baselines. They remain weaker than the sign-sparse variants discussed later in Appendix~\ref{app:sparse_sign_normalized}, but the immediate collapse disappears.

\begin{figure}[!t]
  \centering
  \includegraphics[width=0.94\textwidth,height=0.80\textheight,keepaspectratio]{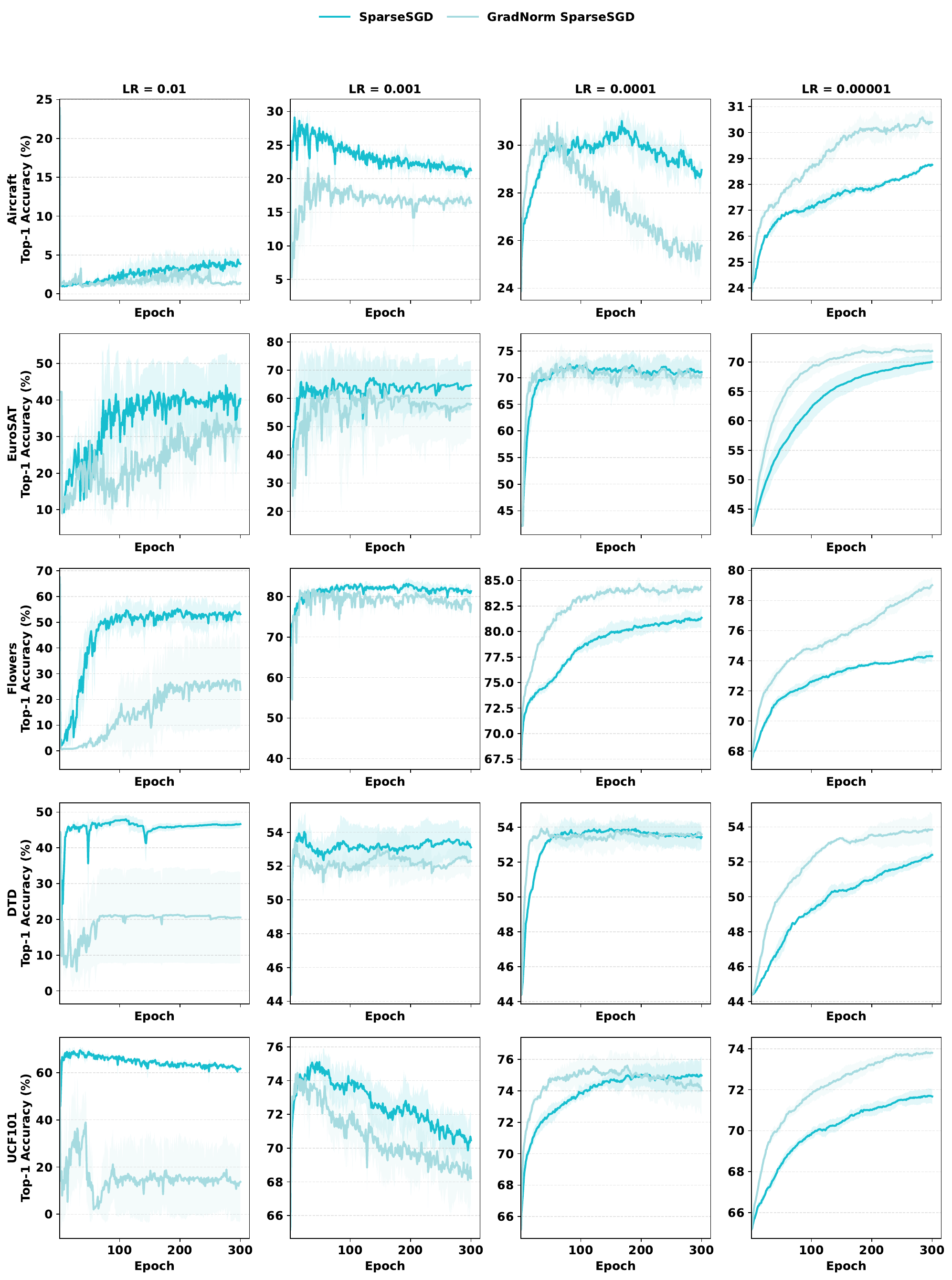}
  \caption{Test accuracy (mean $\pm$ std over 3 seeds) over epochs for SparseSGD and GradNorm SparseSGD in a 1-shot setting on five datasets (Aircraft, EuroSAT, Flowers, DTD, and UCF101) using CLIP (ViT-B/16). Each column corresponds to one learning rate: $10^{-2}$, $10^{-3}$, $10^{-4}$, and $10^{-5}$.}
  \label{fig:update_scale_sparse_group1_with_aug}
\end{figure}

\begin{figure}[!t]
  \centering
  \includegraphics[width=0.94\textwidth,height=0.80\textheight,keepaspectratio]{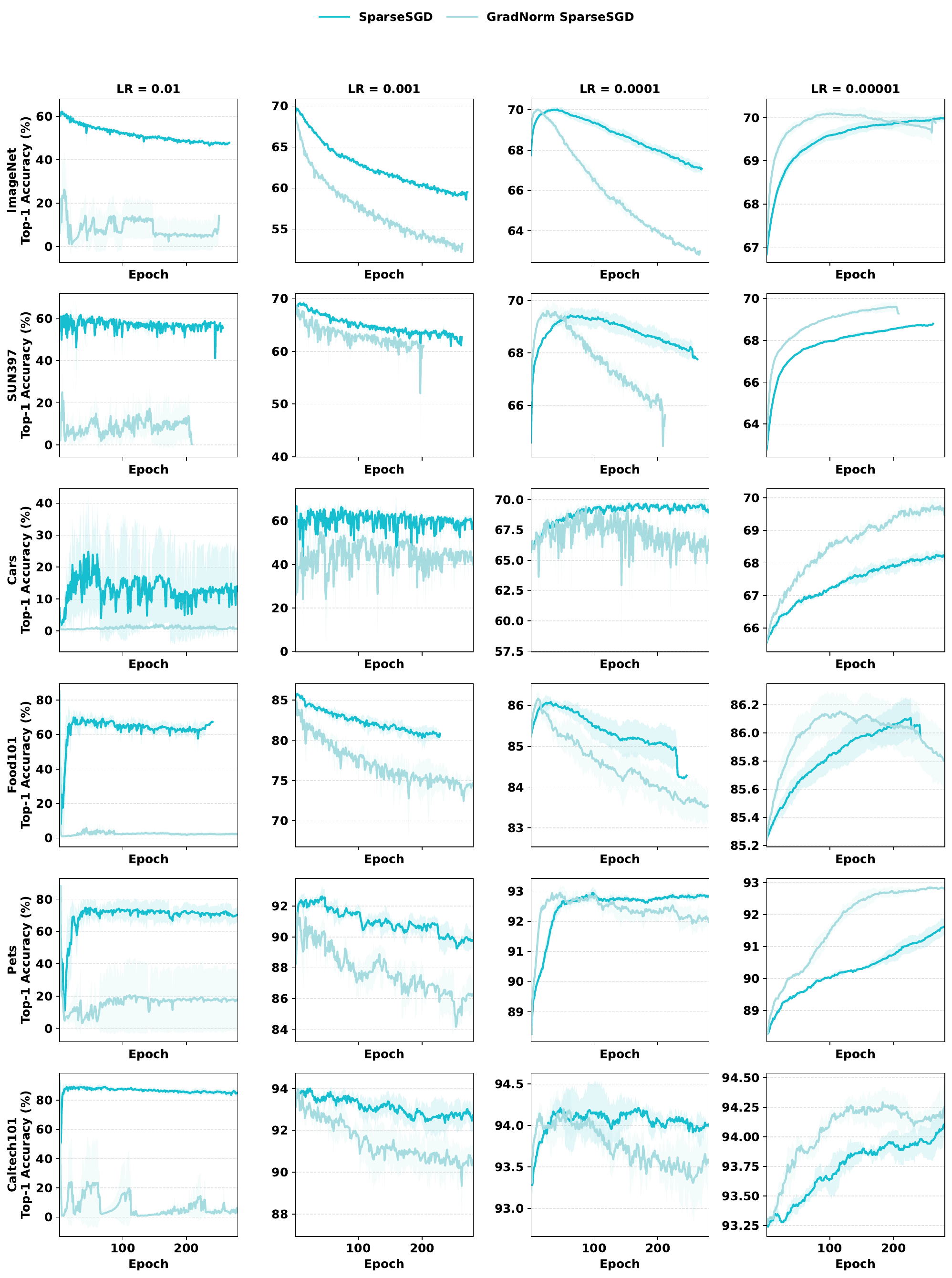}
  \caption{Test accuracy (mean $\pm$ std over 3 seeds) over epochs for SparseSGD and GradNorm SparseSGD in a 1-shot setting on six datasets (ImageNet, SUN397, Cars, Food101, Pets, and Caltech101) using CLIP (ViT-B/16). Each column corresponds to one learning rate: $10^{-2}$, $10^{-3}$, $10^{-4}$, and $10^{-5}$.}
  \label{fig:update_scale_sparse_group2_with_aug}
\end{figure}

\paragraph{SparseSGD versus GradNorm SparseSGD.}
Figs.~\ref{fig:update_scale_sparse_group1_with_aug} and \ref{fig:update_scale_sparse_group2_with_aug} reveal a different effect for SparseSGD. Here the issue is not collapse, but under-scaling. Since $\widetilde g_t = m \odot g_t$ keeps only the active coordinates, we always have $\|\widetilde g_t\|_2 \leq \|g_t\|_2$. Raw SparseSGD therefore applies a smaller global step than dense SGD under the same learning rate. In practice, this behaves like a smaller effective learning rate. The optimizer remains stable, but it moves more slowly.

This is most visible at $\eta=10^{-4}$, which is also the learning rate used in the main optimizer ablation. At this value, GradNorm SparseSGD reaches $71.5\%$ average accuracy at epoch 30, slightly above $70.4\%$ for SparseSGD. More importantly, the norm-matched sparse baseline reaches its average peak much earlier, at about 74 epochs instead of about 120 for SparseSGD. At $\eta=10^{-4}$, GradNorm SparseSGD reaches its peak earlier on 9 of the 11 datasets, and it has higher epoch-30 accuracy on 7 of them. The final accuracies remain close, with $70.1\%$ for GradNorm SparseSGD and $71.3\%$ for SparseSGD. This indicates that the main effect of gradient normalization is to accelerate convergence, not to change the eventual accuracy ceiling by a large margin.

The high-learning-rate columns also support this interpretation. At $\eta=10^{-2}$ and, to a lesser extent, at $\eta=10^{-3}$, raw SparseSGD often looks stronger than its norm-matched counterpart. This does not contradict the scale argument. In those regimes, the smaller raw sparse step acts as implicit damping because the nominal learning rate is already too large. GradNorm removes this damping by restoring the dense-SGD scale. The informative comparison is therefore the one at $\eta=10^{-4}$, where the learning rate matches the protocol of Sec.~\ref{sec:experiments}. At $\eta=10^{-5}$, both sparse methods train slowly, and GradNorm SparseSGD again reaches useful accuracy earlier on most datasets.

\paragraph{Interpretation.}
These four figures clarify two distinct failure modes. Dense SignSGD fails because its raw sign step does not shrink with $\|g_t\|_2$. SparseSGD is slow because masking reduces the update norm and makes the optimizer effectively conservative. GradNorm is useful precisely because it isolates update scale from update direction. It keeps the sign rule or the sparse support fixed and matches the dense SGD norm when the direction is nonzero. A zero sparse direction produces a zero update. In that sense, it is a reference baseline rather than a new tuned method.

This interpretation is also consistent with the design choice in Sec.~\ref{sec:method}. The separate scale $\alpha$ in Eq.~\eqref{eq:sparse_sign_update} must be calibrated under sparsity. If that magnitude is too large, the sign update collapses as in dense SignSGD. If the sparse update norm is too small, optimization becomes unnecessarily slow, as in raw SparseSGD. Appendix~\ref{app:sparse_sign_normalized} therefore uses the norm-matched variants as the fair reference points when comparing sign-only and sparse-only baselines against SignSparseSGD and DynSignSparseSGD. The main conclusion does not change. Correcting the scale makes SignSGD and SparseSGD fairer baselines, but it does not recover the full accuracy--stability trade-off of the proposed optimizers.

\section{Sensitivity Analysis}
\label{sec:sensitivity_analysis}
\suppanchor{sensitivity}

\begin{figure}[!h]
  \centering
  \resizebox{0.9\textwidth}{!}{\includegraphics{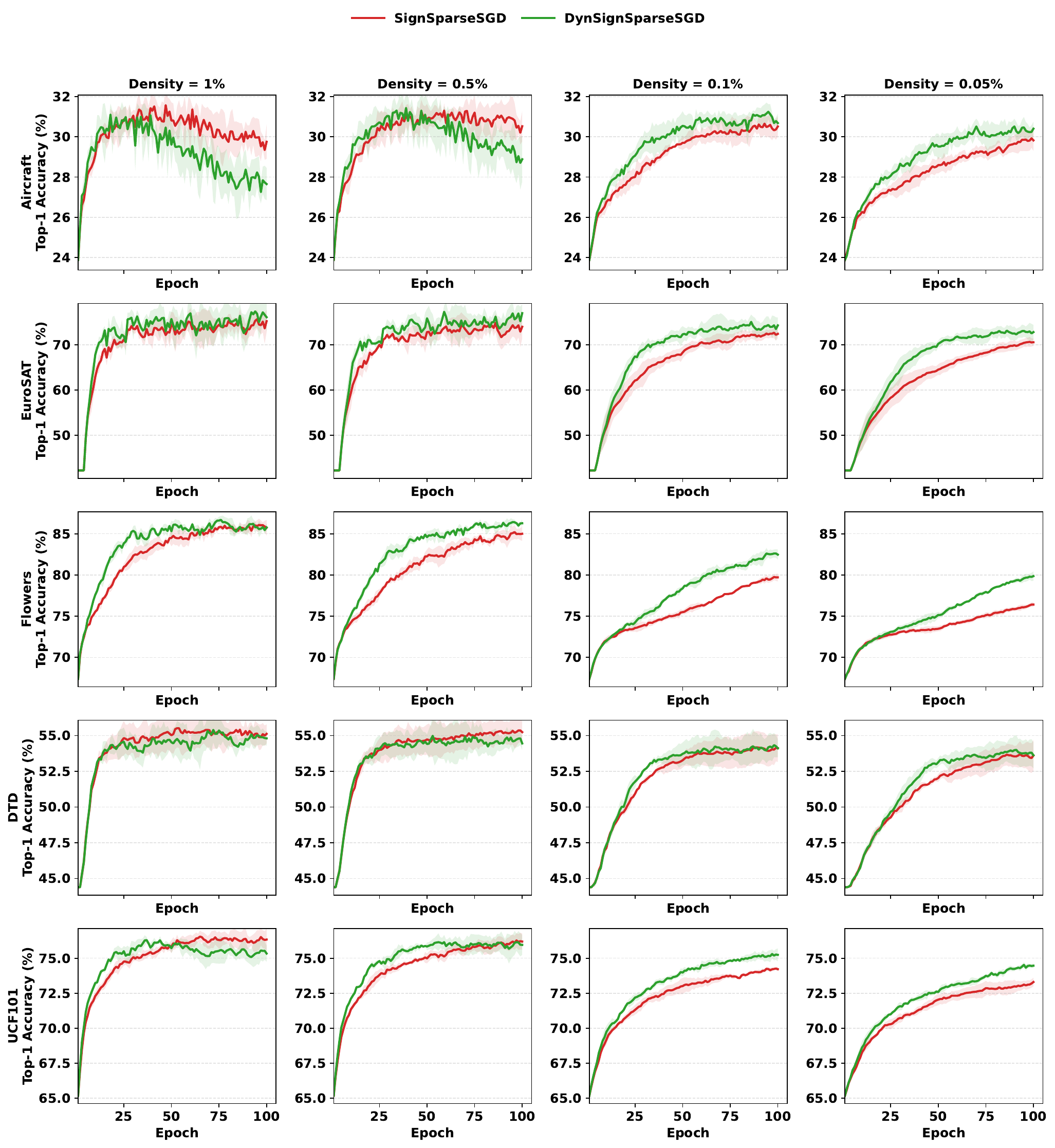}}
  \caption{Test accuracy (mean $\pm$ std over 3 seeds) over training epochs for SignSparseSGD, and DynSignSparseSGD in a 1-shot setting on five datasets (Aircraft, EuroSAT, Flowers, DTD, and UCF101) using CLIP (ViT-B/16) for 4 density ratios (1\%, 0.5\%, 0.1\%, 0.01\%). The learning rate is fixed to $10^{-4}$ and the batch size is equal to 64.}
  \label{fig:fewshot_curves_sparsity_1shot_5datasets}
\end{figure}

\begin{figure}[!h]
  \centering
  \resizebox{0.9\textwidth}{!}{\includegraphics{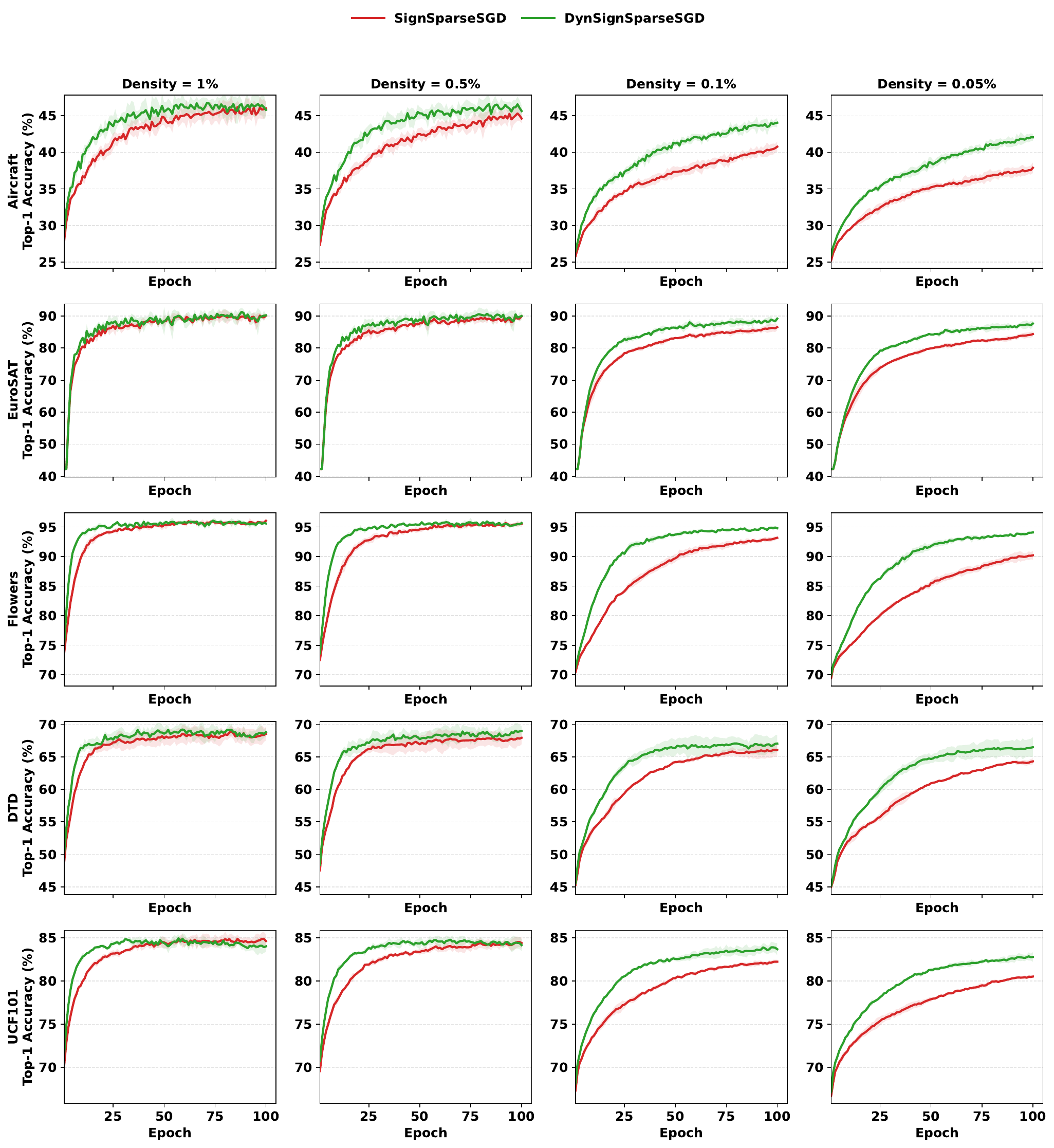}}
  \caption{Test accuracy (mean $\pm$ std over 3 seeds) over training epochs for SignSparseSGD, and DynSignSparseSGD in an 8-shot setting on five datasets (Aircraft, EuroSAT, Flowers, DTD, and UCF101) using CLIP (ViT-B/16) for 4 density ratios (1\%, 0.5\%, 0.1\%, 0.01\%). The learning rate is fixed to $10^{-4}$ and the batch size is 64.}
  \label{fig:fewshot_curves_sparsity_8shot_5datasets}
\end{figure}

As shown in Fig.~\ref{fig:fewshot_curves_sparsity_1shot_5datasets} and Fig.~\ref{fig:fewshot_curves_sparsity_8shot_5datasets}, varying the density ratio over nearly two orders of magnitude leaves the behaviour of both SignSparseSGD and DynSignSparseSGD remarkably stable across datasets and shot levels, with only minor changes in the final test accuracy and in the shape of the learning curves. In the challenging 1-shot regime, all four densities converge to similar plateaus on Aircraft, EuroSAT, Flowers, DTD, and UCF101. In the 8-shot regime, this spread becomes even narrower, and the curves for different densities almost overlap for most datasets, indicating that both optimizers can be deployed with a coarse choice of density ratio without sacrificing performance.

A consistent pattern across datasets is that denser masks (e.g., 1\% and 0.5\%) reach their plateau in fewer epochs, whereas highly sparse configurations (e.g., 0.1\% and 0.05\%) require more iterations to converge but do so in a smooth and monotonic fashion, with reduced oscillations and no signs of late-epoch overfitting. This effect is clearly visible on Aircraft and DTD in the 1-shot setting, where dense masks rise quickly and then exhibit small fluctuations around the plateau, while very sparse masks climb more gradually yet end at a comparable or slightly higher accuracy. On datasets such as EuroSAT, Flowers, and UCF101, the curves for all densities quickly saturate close to the same level. So, the main impact of sparsity is on the speed rather than the quality of convergence. In the 8-shot setting, additional supervision accelerates convergence for every density, and even the most aggressive sparsity reaches a near-optimal plateau well before the training budget is exhausted, which suggests that sign-sparse optimization can exploit extra labeled examples without requiring denser updates.

Comparing the two optimizers, DynSignSparseSGD consistently matches or exceeds the performance of SignSparseSGD for all densities and datasets, with the advantage being most pronounced in the high-sparsity regime where only a tiny fraction of parameters are updated at each step. In several cases, such as Aircraft and DTD at low density, the static mask of SignSparseSGD leads to a visible accuracy gap that persists throughout training, whereas the dynamic mask of DynSignSparseSGD enables the optimizer to track informative directions over time and recover a plateau that is close to the one achieved at higher density. On the other datasets, the gains are smaller but still measurable, and DynSignSparseSGD remains at least as stable as the static variant. Taken together, these observations indicate that the proposed sign-sparse optimizers are not overly sensitive to the exact density ratio, and that increasing sparsity mainly slows down rather than harms convergence. Furthermore, the dynamic variant provides an additional safety margin by maintaining stable performance over a wide range of sparsity levels with a fixed learning rate and batch size.

\section{DynSignSparseSGD vs SignSparseSGD}
\label{app:dyn_vs_static}
\suppanchor{dyn-static}

\begin{table}[!h]
\caption{Ablation on the dynamic selection process in few-shot classification with CLIP (ViT-B/32) backbone. We compare SignSparseSGD and DynSignSparseSGD. Top-1 accuracy (\%) averaged over 3 fixed random seeds.}
\label{tab:ablation_sign_vs_dyn_ViTB32_with_aug}
\centering
\footnotesize
\setlength{\tabcolsep}{1.8pt}
\renewcommand{\arraystretch}{1.00}
\begin{adjustbox}{max totalsize={\textwidth}{0.95\textheight},center}
\begin{tabular}{@{}c l *{12}{d}@{}}
\toprule
Shots & Method & ImageNet & SUN & Aircr. & EuroSAT & Cars & Food & Pets & Flow. & Caltech & DTD & UCF & Avg. \\ \midrule
\multirow{1}{*}{0} & CLIP & 62.0 & 62.1 & 18.1 & 35.8 & 60.1 & 78.9 & 85.1 & 64.0 & 91.4 & 43.0 & 61.6 & 60.2 \\
\midrule
\multirow{2}{*}{1} & SignSparseSGD & 60.6 $\pm$ 0.6 & 65.3 $\pm$ 0.1 & 21.6 $\pm$ 0.5 & 66.0 $\pm$ 5.2 & 61.3 $\pm$ 1.1 & 77.9 $\pm$ 0.5 & 88.5 $\pm$ 0.4 & 75.0 $\pm$ 0.8 & 92.3 $\pm$ 0.5 & 50.9 $\pm$ 1.5 & 71.5 $\pm$ 0.3 & 66.4 $\pm$ 1.0 \\
& DynSignSparseSGD & 62.6 $\pm$ 0.3 & 67.0 $\pm$ 0.3 & 22.7 $\pm$ 0.8 & 69.1 $\pm$ 2.7 & 63.1 $\pm$ 0.4 & 79.7 $\pm$ 0.5 & 89.1 $\pm$ 0.2 & 77.7 $\pm$ 1.2 & 92.6 $\pm$ 0.7 & 51.9 $\pm$ 1.0 & 72.0 $\pm$ 0.4 & 67.9 $\pm$ 0.8 \\
\midrule
\multirow{2}{*}{2} & SignSparseSGD & 61.9 $\pm$ 0.3 & 66.9 $\pm$ 0.2 & 24.3 $\pm$ 0.6 & 79.1 $\pm$ 6.7 & 63.6 $\pm$ 0.1 & 78.0 $\pm$ 0.4 & 89.1 $\pm$ 0.4 & 78.7 $\pm$ 0.2 & 93.8 $\pm$ 0.3 & 56.4 $\pm$ 0.4 & 74.1 $\pm$ 0.8 & 69.6 $\pm$ 0.9 \\
& DynSignSparseSGD & 63.2 $\pm$ 0.1 & 68.2 $\pm$ 0.3 & 25.5 $\pm$ 0.6 & 79.8 $\pm$ 4.2 & 66.0 $\pm$ 0.1 & 79.5 $\pm$ 0.4 & 89.2 $\pm$ 0.3 & 83.8 $\pm$ 1.4 & 94.0 $\pm$ 0.3 & 57.1 $\pm$ 1.1 & 74.8 $\pm$ 0.6 & 71.0 $\pm$ 0.9 \\
\midrule
\multirow{2}{*}{4} & SignSparseSGD & 63.3 $\pm$ 0.4 & 69.0 $\pm$ 0.4 & 25.2 $\pm$ 1.1 & 82.6 $\pm$ 1.0 & 66.9 $\pm$ 0.3 & 78.2 $\pm$ 0.4 & 89.3 $\pm$ 0.4 & 85.8 $\pm$ 1.2 & 94.7 $\pm$ 0.4 & 59.1 $\pm$ 1.0 & 75.9 $\pm$ 0.7 & 71.8 $\pm$ 0.7 \\
& DynSignSparseSGD & 64.0 $\pm$ 0.2 & 69.9 $\pm$ 0.3 & 27.7 $\pm$ 0.6 & 83.9 $\pm$ 1.6 & 69.3 $\pm$ 0.1 & 79.6 $\pm$ 0.4 & 90.0 $\pm$ 0.4 & 88.9 $\pm$ 0.8 & 95.2 $\pm$ 0.2 & 61.1 $\pm$ 0.3 & 78.0 $\pm$ 0.9 & 73.4 $\pm$ 0.5 \\
\midrule
\multirow{2}{*}{8} & SignSparseSGD & 64.7 $\pm$ 0.1 & 70.9 $\pm$ 0.3 & 31.1 $\pm$ 0.8 & 85.9 $\pm$ 0.5 & 71.4 $\pm$ 0.5 & 79.0 $\pm$ 0.2 & 90.3 $\pm$ 0.1 & 89.3 $\pm$ 0.7 & 94.9 $\pm$ 0.3 & 62.3 $\pm$ 0.9 & 78.8 $\pm$ 0.7 & 74.4 $\pm$ 0.5 \\
& DynSignSparseSGD & 65.0 $\pm$ 0.2 & 71.9 $\pm$ 0.1 & 34.8 $\pm$ 0.6 & 88.4 $\pm$ 1.0 & 75.1 $\pm$ 0.2 & 79.8 $\pm$ 0.1 & 91.0 $\pm$ 0.3 & 92.5 $\pm$ 0.1 & 95.5 $\pm$ 0.3 & 64.5 $\pm$ 1.0 & 79.7 $\pm$ 0.6 & 76.2 $\pm$ 0.4 \\
\midrule
\multirow{2}{*}{16} & SignSparseSGD & 66.3 $\pm$ 0.2 & 73.1 $\pm$ 0.3 & 36.4 $\pm$ 0.9 & 90.7 $\pm$ 1.6 & 76.4 $\pm$ 0.6 & 79.3 $\pm$ 0.5 & 89.9 $\pm$ 1.0 & 93.8 $\pm$ 1.0 & 95.8 $\pm$ 0.0 & 67.5 $\pm$ 0.2 & 81.7 $\pm$ 0.4 & 77.4 $\pm$ 0.6 \\
& DynSignSparseSGD & 66.9 $\pm$ 0.2 & 73.5 $\pm$ 0.1 & 42.0 $\pm$ 1.1 & 92.0 $\pm$ 0.6 & 79.8 $\pm$ 0.4 & 80.3 $\pm$ 0.4 & 91.0 $\pm$ 0.3 & 95.7 $\pm$ 1.0 & 95.8 $\pm$ 0.0 & 69.2 $\pm$ 0.5 & 82.7 $\pm$ 0.5 & 79.0 $\pm$ 0.5 \\
\bottomrule
\end{tabular}
\end{adjustbox}
\end{table}

\begin{table}[!h]
\caption{Ablation on the dynamic selection process in few-shot classification with CLIP (ViT-B/16) backbone. We compare SignSparseSGD and DynSignSparseSGD. Top-1 accuracy (\%) averaged over 3 fixed random seeds.}
\label{tab:ablation_sign_vs_dyn_ViTB16_with_aug}
\centering
\footnotesize
\setlength{\tabcolsep}{1.8pt}
\renewcommand{\arraystretch}{1.00}
\begin{adjustbox}{max totalsize={\textwidth}{0.95\textheight},center}
\begin{tabular}{@{}c l *{12}{d}@{}}
\toprule
Shots & Method & ImageNet & SUN & Aircr. & EuroSAT & Cars & Food & Pets & Flow. & Caltech & DTD & UCF & Avg. \\ \midrule
\multirow{1}{*}{0} & CLIP & 66.7 & 62.6 & 23.9 & 42.2 & 65.5 & 85.2 & 88.3 & 67.4 & 93.3 & 44.4 & 65.2 & 64.1 \\
\midrule
\multirow{2}{*}{1} & SignSparseSGD & 66.8 $\pm$ 0.1 & 68.2 $\pm$ 0.3 & 29.4 $\pm$ 1.3 & 71.2 $\pm$ 3.7 & 69.1 $\pm$ 0.6 & 83.8 $\pm$ 0.5 & 91.3 $\pm$ 0.3 & 83.3 $\pm$ 0.5 & 93.7 $\pm$ 0.2 & 54.3 $\pm$ 0.6 & 75.7 $\pm$ 0.3 & 71.5 $\pm$ 0.8 \\
& DynSignSparseSGD & 68.1 $\pm$ 0.1 & 69.8 $\pm$ 0.3 & 30.6 $\pm$ 1.2 & 73.7 $\pm$ 4.0 & 70.6 $\pm$ 0.4 & 85.1 $\pm$ 0.3 & 92.5 $\pm$ 0.2 & 84.7 $\pm$ 0.4 & 93.8 $\pm$ 0.1 & 54.8 $\pm$ 0.6 & 76.2 $\pm$ 0.4 & 72.7 $\pm$ 0.7 \\
\midrule
\multirow{2}{*}{2} & SignSparseSGD & 67.4 $\pm$ 0.4 & 69.7 $\pm$ 0.1 & 32.0 $\pm$ 1.4 & 79.5 $\pm$ 2.2 & 71.7 $\pm$ 0.2 & 84.1 $\pm$ 0.3 & 92.0 $\pm$ 0.3 & 87.2 $\pm$ 0.5 & 94.2 $\pm$ 0.3 & 59.5 $\pm$ 1.2 & 78.4 $\pm$ 1.0 & 74.2 $\pm$ 0.7 \\
& DynSignSparseSGD & 68.6 $\pm$ 0.2 & 71.0 $\pm$ 0.1 & 33.8 $\pm$ 2.1 & 82.0 $\pm$ 1.4 & 73.7 $\pm$ 0.3 & 85.1 $\pm$ 0.1 & 91.7 $\pm$ 0.4 & 89.6 $\pm$ 0.2 & 94.3 $\pm$ 0.4 & 60.1 $\pm$ 1.2 & 79.7 $\pm$ 0.9 & 75.4 $\pm$ 0.7 \\
\midrule
\multirow{2}{*}{4} & SignSparseSGD & 68.5 $\pm$ 0.3 & 71.9 $\pm$ 0.2 & 37.3 $\pm$ 0.7 & 81.5 $\pm$ 1.2 & 75.7 $\pm$ 0.7 & 84.7 $\pm$ 0.3 & 92.5 $\pm$ 0.2 & 91.4 $\pm$ 1.1 & 95.4 $\pm$ 0.1 & 62.6 $\pm$ 0.7 & 81.1 $\pm$ 1.1 & 76.6 $\pm$ 0.6 \\
& DynSignSparseSGD & 69.4 $\pm$ 0.2 & 72.8 $\pm$ 0.1 & 38.4 $\pm$ 1.1 & 84.1 $\pm$ 2.0 & 77.8 $\pm$ 0.3 & 85.4 $\pm$ 0.2 & 92.8 $\pm$ 0.3 & 93.2 $\pm$ 0.8 & 95.2 $\pm$ 0.1 & 64.5 $\pm$ 0.7 & 82.2 $\pm$ 1.2 & 77.8 $\pm$ 0.6 \\
\midrule
\multirow{2}{*}{8} & SignSparseSGD & 69.9 $\pm$ 0.3 & 73.7 $\pm$ 0.1 & 41.1 $\pm$ 1.7 & 84.7 $\pm$ 5.2 & 79.8 $\pm$ 0.7 & 84.8 $\pm$ 0.6 & 92.9 $\pm$ 0.4 & 94.0 $\pm$ 0.3 & 95.9 $\pm$ 0.2 & 66.5 $\pm$ 0.7 & 83.6 $\pm$ 0.6 & 78.8 $\pm$ 1.0 \\
& DynSignSparseSGD & 70.5 $\pm$ 0.4 & 74.5 $\pm$ 0.2 & 45.2 $\pm$ 0.7 & 85.2 $\pm$ 4.0 & 82.4 $\pm$ 0.5 & 85.3 $\pm$ 0.3 & 93.4 $\pm$ 0.1 & 95.0 $\pm$ 0.2 & 95.8 $\pm$ 0.1 & 68.0 $\pm$ 1.3 & 84.7 $\pm$ 0.2 & 80.0 $\pm$ 0.7 \\
\midrule
\multirow{2}{*}{16} & SignSparseSGD & 71.6 $\pm$ 0.1 & 75.0 $\pm$ 0.2 & 45.6 $\pm$ 0.9 & 91.2 $\pm$ 1.1 & 83.6 $\pm$ 0.5 & 85.7 $\pm$ 0.1 & 93.6 $\pm$ 0.3 & 96.8 $\pm$ 0.2 & 96.2 $\pm$ 0.2 & 71.2 $\pm$ 0.3 & 85.8 $\pm$ 0.0 & 81.5 $\pm$ 0.4 \\
& DynSignSparseSGD & 72.2 $\pm$ 0.1 & 75.7 $\pm$ 0.2 & 51.7 $\pm$ 0.6 & 91.8 $\pm$ 0.5 & 85.9 $\pm$ 0.3 & 86.1 $\pm$ 0.1 & 93.9 $\pm$ 0.1 & 97.7 $\pm$ 0.2 & 96.4 $\pm$ 0.3 & 72.2 $\pm$ 0.4 & 86.9 $\pm$ 0.1 & 82.8 $\pm$ 0.3 \\
\bottomrule
\end{tabular}
\end{adjustbox}
\end{table}

\begin{table}[!h]
\caption{Ablation on the dynamic selection process in the few-shot classification with CLIP (ViT-L/14) backbone. We compare SignSparseSGD and DynSignSparseSGD. Top-1 accuracy (\%) averaged over 3 fixed random seeds.}
\label{tab:ablation_sign_vs_dyn_ViTL14_with_aug}
\centering
\footnotesize
\setlength{\tabcolsep}{1.8pt}
\renewcommand{\arraystretch}{1.00}
\begin{adjustbox}{max totalsize={\textwidth}{0.95\textheight},center}
\begin{tabular}{@{}c l *{12}{d}@{}}
\toprule
Shots & Method & ImageNet & SUN & Aircr. & EuroSAT & Cars & Food & Pets & Flow. & Caltech & DTD & UCF & Avg. \\ \midrule
\multirow{1}{*}{0} & CLIP & 73.5 & 67.7 & 30.0 & 55.1 & 76.8 & 90.3 & 93.1 & 76.2 & 95.2 & 52.4 & 73.7 & 71.3 \\
\midrule
\multirow{2}{*}{1} & SignSparseSGD & 74.6 $\pm$ 0.5 & 73.3 $\pm$ 0.2 & 40.5 $\pm$ 1.7 & 71.7 $\pm$ 3.7 & 80.0 $\pm$ 0.3 & 90.2 $\pm$ 0.1 & 94.6 $\pm$ 0.2 & 91.5 $\pm$ 1.3 & 95.5 $\pm$ 0.3 & 61.7 $\pm$ 1.1 & 82.8 $\pm$ 0.5 & 77.8 $\pm$ 0.9 \\
& DynSignSparseSGD & 75.6 $\pm$ 0.2 & 74.1 $\pm$ 0.2 & 41.3 $\pm$ 0.9 & 73.8 $\pm$ 4.1 & 80.9 $\pm$ 0.3 & 90.7 $\pm$ 0.2 & 94.7 $\pm$ 0.3 & 92.3 $\pm$ 1.0 & 95.5 $\pm$ 0.2 & 61.5 $\pm$ 0.7 & 82.9 $\pm$ 0.8 & 78.5 $\pm$ 0.8 \\
\midrule
\multirow{2}{*}{2} & SignSparseSGD & 75.3 $\pm$ 0.2 & 74.9 $\pm$ 0.1 & 41.9 $\pm$ 1.9 & 84.4 $\pm$ 0.7 & 82.2 $\pm$ 0.4 & 90.5 $\pm$ 0.2 & 94.3 $\pm$ 0.4 & 94.3 $\pm$ 1.0 & 97.0 $\pm$ 0.3 & 67.6 $\pm$ 1.2 & 84.7 $\pm$ 0.7 & 80.6 $\pm$ 0.6 \\
& DynSignSparseSGD & 75.9 $\pm$ 0.3 & 75.6 $\pm$ 0.3 & 43.9 $\pm$ 2.6 & 82.6 $\pm$ 1.0 & 83.2 $\pm$ 0.6 & 90.8 $\pm$ 0.2 & 94.5 $\pm$ 0.3 & 95.2 $\pm$ 0.7 & 96.8 $\pm$ 0.6 & 68.5 $\pm$ 1.2 & 85.0 $\pm$ 0.6 & 81.1 $\pm$ 0.8 \\
\midrule
\multirow{2}{*}{4} & SignSparseSGD & 76.1 $\pm$ 0.2 & 76.2 $\pm$ 0.1 & 47.5 $\pm$ 0.1 & 85.0 $\pm$ 3.0 & 84.7 $\pm$ 0.7 & 90.6 $\pm$ 0.1 & 94.9 $\pm$ 0.0 & 96.5 $\pm$ 0.8 & 97.3 $\pm$ 0.1 & 69.4 $\pm$ 0.2 & 87.4 $\pm$ 1.0 & 82.3 $\pm$ 0.6 \\
& DynSignSparseSGD & 76.6 $\pm$ 0.1 & 77.0 $\pm$ 0.1 & 50.5 $\pm$ 0.6 & 85.4 $\pm$ 2.5 & 86.0 $\pm$ 0.9 & 91.1 $\pm$ 0.2 & 95.0 $\pm$ 0.1 & 97.0 $\pm$ 1.0 & 97.5 $\pm$ 0.3 & 70.8 $\pm$ 0.6 & 87.6 $\pm$ 0.7 & 83.1 $\pm$ 0.6 \\
\midrule
\multirow{2}{*}{8} & SignSparseSGD & 76.8 $\pm$ 0.2 & 77.5 $\pm$ 0.0 & 56.0 $\pm$ 1.2 & 89.2 $\pm$ 2.7 & 87.6 $\pm$ 0.3 & 90.7 $\pm$ 0.2 & 95.4 $\pm$ 0.2 & 97.3 $\pm$ 0.3 & 97.5 $\pm$ 0.6 & 72.0 $\pm$ 0.6 & 88.6 $\pm$ 0.5 & 84.4 $\pm$ 0.6 \\
& DynSignSparseSGD & 77.3 $\pm$ 0.1 & 78.3 $\pm$ 0.1 & 58.7 $\pm$ 0.6 & 88.9 $\pm$ 2.7 & 88.8 $\pm$ 0.2 & 91.2 $\pm$ 0.1 & 95.5 $\pm$ 0.3 & 97.8 $\pm$ 0.5 & 97.4 $\pm$ 0.5 & 73.1 $\pm$ 0.4 & 89.2 $\pm$ 0.4 & 85.1 $\pm$ 0.5 \\
\midrule
\multirow{2}{*}{16} & SignSparseSGD & 78.0 $\pm$ 0.1 & 79.3 $\pm$ 0.3 & 61.6 $\pm$ 0.9 & 92.7 $\pm$ 1.5 & 89.5 $\pm$ 0.3 & 90.8 $\pm$ 0.2 & 95.4 $\pm$ 0.3 & 98.5 $\pm$ 0.1 & 97.7 $\pm$ 0.0 & 76.5 $\pm$ 0.8 & 90.1 $\pm$ 0.2 & 86.4 $\pm$ 0.4 \\
& DynSignSparseSGD & 78.3 $\pm$ 0.1 & 79.6 $\pm$ 0.1 & 65.0 $\pm$ 0.3 & 93.4 $\pm$ 0.9 & 90.3 $\pm$ 0.1 & 91.2 $\pm$ 0.0 & 95.6 $\pm$ 0.1 & 98.9 $\pm$ 0.2 & 97.4 $\pm$ 0.3 & 77.8 $\pm$ 0.7 & 91.0 $\pm$ 0.2 & 87.1 $\pm$ 0.3 \\
\bottomrule
\end{tabular}
\end{adjustbox}
\end{table}

Tab.~\ref{tab:ablation_sign_vs_dyn_ViTB32_with_aug}, Tab.~\ref{tab:ablation_sign_vs_dyn_ViTB16_with_aug}, and Tab.~\ref{tab:ablation_sign_vs_dyn_ViTL14_with_aug} compare a fixed sparse support (SignSparseSGD) with a refreshed sparse support (DynSignSparseSGD). This is a controlled comparison. Both methods use the same sign-sparse update and remain state-free. The only difference is whether the active coordinates are selected once at initialization or recomputed at every iteration.

Across all three backbones, refreshing the support consistently improves the average accuracy over the 11 datasets. On ViT-B/32, DynSignSparseSGD improves over SignSparseSGD by $+1.5$, $+1.4$, $+1.6$, $+1.8$, and $+1.6$ points from 1-shot to 16-shot, for a mean gain of about $+1.6$ points. On ViT-B/16, the gain is $+1.2$ points at 1, 2, 4, and 8 shots, and $+1.3$ points at 16 shots, which gives a mean gain of about $+1.2$ points. On ViT-L/14, the gain is smaller but remains positive on average at every shot level: $+0.7$, $+0.5$, $+0.8$, $+0.7$, and $+0.7$ points. The same trend appears at both ends of the few-shot regime. For example, the average accuracy on ViT-B/32 rises from $66.4\%$ to $67.9\%$ at 1-shot and from $77.4\%$ to $79.0\%$ at 16-shot. On ViT-B/16, it rises from $71.5\%$ to $72.7\%$ at 1-shot and from $81.5\%$ to $82.8\%$ at 16-shot. On ViT-L/14, it rises from $77.8\%$ to $78.5\%$ at 1-shot and from $86.4\%$ to $87.1\%$ at 16-shot.

The improvement is broad rather than driven by a few outliers. Across the 165 backbone--shot--dataset comparisons, DynSignSparseSGD matches or improves SignSparseSGD in 156 cases, with 154 strict gains. The clearest and most recurrent gains appear on Aircraft, Cars, Flowers, EuroSAT, and DTD. These gains are especially visible on ViT-B/32 and ViT-B/16, where dynamic selection often brings several additional points on fine-grained or harder transfer benchmarks. A few isolated reversals remain, mostly on Caltech and in a small number of ViT-L/14 settings, but they do not change the overall pattern.

This behavior is consistent with the motivation of DynSignSparseSGD. In SignSparseSGD, the support is fixed from the first gradient snapshot. In the few-shot regime, this snapshot can be noisy and may overemphasize coordinates that are only temporarily salient. DynSignSparseSGD avoids this commitment by recomputing the top-$k$ set from the current gradient at each iteration. The optimizer can therefore move its sparse budget toward coordinates that remain informative as training evolves, while preserving the robustness of sign updates.

Overall, these tables isolate the value of dynamic support selection. The gain does not come from additional optimizer states or from a different update rule. It comes from updating the right coordinates at the right time. Under the same sparsity budget and the same sign-based step, a dynamic support is consistently more effective than a static one for few-shot CLIP adaptation.

\section{Sparse \& Sign Optimizer Comparison}
\label{app:sparse_sign_comparison}
\suppanchor{optimizer-comparison}

\subsection{Without Gradient Normalization}
\label{app:sparse_sign_raw}
\label{without_gradient_normlization}
\suppanchor{optimizer-raw}
\begin{table}
\caption{Few-shot classification performance on 11 datasets with CLIP (ViT-B/32) backbone. We compare SignSparseSGD and DynSignSparseSGD against multiple optimizer families: sign-based optimizers (SignSGD, SignAdam), sparsity-based optimizers (SparseSGD, SparseAdam), sign-sparse optimizers (SignSparseAdam, SignSparseSGD), and dynamic-sparse optimizers (DynSparseAdam, DynSparseSGD, DynSignSparseSGD). Top-1 accuracy (\%) averaged over fixed 3 seeds. Top-1 accuracy (\%) averaged over 3 seeds.}
\label{tab:ablation_optim_ViTB32_with_aug}
\centering
\footnotesize
\setlength{\tabcolsep}{1.8pt}
\renewcommand{\arraystretch}{1.00}
\begin{adjustbox}{max totalsize={\textwidth}{0.95\textheight},center}
\begin{tabular}{@{}c l *{12}{d}@{}}
\toprule
Shots & Method & ImageNet & SUN & Aircr. & EuroSAT & Cars & Food & Pets & Flow. & Caltech & DTD & UCF & Avg. \\ \midrule
\multirow{1}{*}{0} & CLIP & 62.0 & 62.1 & 18.1 & 35.8 & 60.1 & 78.9 & 85.1 & 64.0 & 91.4 & 43.0 & 61.6 & 60.2 \\
\midrule
\multirow{9}{*}{1} & SignSGD & 4.2 $\pm$ 2.8 & 12.7 $\pm$ 1.8 & 1.0 $\pm$ 0.1 & 20.6 $\pm$ 1.2 & 0.5 $\pm$ 0.0 & 1.5 $\pm$ 0.6 & 5.5 $\pm$ 1.6 & 2.6 $\pm$ 0.8 & 30.1 $\pm$ 20.3 & 15.2 $\pm$ 2.2 & 8.0 $\pm$ 8.6 & 9.2 $\pm$ 3.6 \\
& SignAdam & 0.1 $\pm$ 0.0 & 0.2 $\pm$ 0.0 & 0.9 $\pm$ 0.1 & 10.6 $\pm$ 0.7 & 0.5 $\pm$ 0.1 & 1.0 $\pm$ 0.1 & 2.4 $\pm$ 0.1 & 1.1 $\pm$ 0.4 & 1.0 $\pm$ 0.6 & 2.1 $\pm$ 0.1 & 0.9 $\pm$ 0.1 & 1.9 $\pm$ 0.2 \\
\cmidrule(lr){2-14}
& SparseSGD & 64.3 $\pm$ 0.2 & 66.1 $\pm$ 0.1 & 21.7 $\pm$ 0.4 & 60.6 $\pm$ 5.3 & 61.1 $\pm$ 0.5 & 80.0 $\pm$ 0.3 & 88.5 $\pm$ 0.7 & 69.0 $\pm$ 0.8 & 93.0 $\pm$ 0.4 & 49.1 $\pm$ 0.3 & 67.7 $\pm$ 0.4 & 65.6 $\pm$ 0.9 \\
& SparseAdam & 59.7 $\pm$ 0.7 & 64.4 $\pm$ 0.2 & 21.0 $\pm$ 1.0 & 66.2 $\pm$ 1.8 & 57.5 $\pm$ 1.0 & 78.0 $\pm$ 0.6 & 85.3 $\pm$ 1.1 & 73.3 $\pm$ 0.7 & 93.3 $\pm$ 0.7 & 47.6 $\pm$ 0.9 & 69.7 $\pm$ 0.6 & 65.1 $\pm$ 0.8 \\
\cmidrule(lr){2-14}
& SignSparseAdam & 0.1 $\pm$ 0.0 & 0.2 $\pm$ 0.0 & 1.0 $\pm$ 0.1 & 15.7 $\pm$ 7.8 & 0.5 $\pm$ 0.1 & 1.0 $\pm$ 0.0 & 2.9 $\pm$ 0.3 & 0.9 $\pm$ 0.4 & 0.6 $\pm$ 0.1 & 2.1 $\pm$ 0.0 & 1.1 $\pm$ 0.1 & 2.4 $\pm$ 0.8 \\
& SignSparseSGD & 60.6 $\pm$ 0.6 & 65.3 $\pm$ 0.1 & 21.6 $\pm$ 0.5 & 66.0 $\pm$ 5.2 & 61.3 $\pm$ 1.1 & 77.9 $\pm$ 0.5 & 88.5 $\pm$ 0.4 & 75.0 $\pm$ 0.8 & 92.3 $\pm$ 0.5 & 50.9 $\pm$ 1.5 & 71.5 $\pm$ 0.3 & 66.4 $\pm$ 1.0 \\
\cmidrule(lr){2-14}
& DynSparseAdam & 58.8 $\pm$ 0.5 & 64.2 $\pm$ 0.4 & 21.7 $\pm$ 1.0 & 65.0 $\pm$ 2.7 & 60.0 $\pm$ 0.6 & 77.1 $\pm$ 0.6 & 88.1 $\pm$ 0.3 & 75.1 $\pm$ 1.5 & 92.1 $\pm$ 0.7 & 50.0 $\pm$ 2.1 & 71.4 $\pm$ 0.5 & 65.8 $\pm$ 1.0 \\
& DynSparseSGD & 64.1 $\pm$ 0.4 & 66.8 $\pm$ 0.1 & 22.0 $\pm$ 0.1 & 64.6 $\pm$ 3.8 & 61.4 $\pm$ 0.2 & 80.7 $\pm$ 0.3 & 89.6 $\pm$ 0.3 & 70.9 $\pm$ 0.2 & 92.7 $\pm$ 0.5 & 50.1 $\pm$ 0.5 & 68.8 $\pm$ 0.2 & 66.5 $\pm$ 0.6 \\
& DynSignSparseSGD & 62.6 $\pm$ 0.3 & 67.0 $\pm$ 0.3 & 22.7 $\pm$ 0.8 & 69.1 $\pm$ 2.7 & 63.1 $\pm$ 0.4 & 79.7 $\pm$ 0.5 & 89.1 $\pm$ 0.2 & 77.7 $\pm$ 1.2 & 92.6 $\pm$ 0.7 & 51.9 $\pm$ 1.0 & 72.0 $\pm$ 0.4 & 67.9 $\pm$ 0.8 \\
\midrule
\multirow{9}{*}{2} & SignSGD & 6.2 $\pm$ 3.9 & 14.7 $\pm$ 2.2 & 1.2 $\pm$ 0.2 & 15.6 $\pm$ 5.6 & 0.8 $\pm$ 0.1 & 1.2 $\pm$ 0.1 & 2.6 $\pm$ 0.7 & 7.0 $\pm$ 2.2 & 1.0 $\pm$ 0.2 & 11.8 $\pm$ 3.8 & 2.2 $\pm$ 0.3 & 5.8 $\pm$ 1.8 \\
& SignAdam & 0.1 $\pm$ 0.0 & 0.2 $\pm$ 0.1 & 1.1 $\pm$ 0.1 & 10.4 $\pm$ 1.6 & 0.5 $\pm$ 0.1 & 1.0 $\pm$ 0.0 & 2.7 $\pm$ 0.1 & 0.8 $\pm$ 0.3 & 0.6 $\pm$ 0.2 & 2.0 $\pm$ 0.1 & 1.0 $\pm$ 0.1 & 1.9 $\pm$ 0.2 \\
\cmidrule(lr){2-14}
& SparseSGD & 64.8 $\pm$ 0.1 & 66.9 $\pm$ 0.2 & 23.3 $\pm$ 0.2 & 68.6 $\pm$ 2.9 & 62.3 $\pm$ 0.2 & 80.6 $\pm$ 0.2 & 89.6 $\pm$ 0.5 & 72.0 $\pm$ 1.2 & 93.3 $\pm$ 0.2 & 52.8 $\pm$ 1.0 & 69.2 $\pm$ 0.7 & 67.6 $\pm$ 0.7 \\
& SparseAdam & 61.3 $\pm$ 0.1 & 65.2 $\pm$ 0.4 & 23.2 $\pm$ 0.6 & 78.2 $\pm$ 2.5 & 58.8 $\pm$ 0.0 & 78.8 $\pm$ 0.2 & 89.2 $\pm$ 0.2 & 78.6 $\pm$ 1.1 & 93.9 $\pm$ 0.3 & 53.8 $\pm$ 1.4 & 71.3 $\pm$ 0.2 & 68.4 $\pm$ 0.6 \\
\cmidrule(lr){2-14}
& SignSparseAdam & 0.1 $\pm$ 0.0 & 0.3 $\pm$ 0.1 & 0.8 $\pm$ 0.2 & 13.8 $\pm$ 2.7 & 0.4 $\pm$ 0.1 & 1.0 $\pm$ 0.0 & 2.9 $\pm$ 0.2 & 1.1 $\pm$ 0.3 & 0.9 $\pm$ 0.4 & 1.9 $\pm$ 0.4 & 0.8 $\pm$ 0.1 & 2.2 $\pm$ 0.4 \\
& SignSparseSGD & 61.9 $\pm$ 0.3 & 66.9 $\pm$ 0.2 & 24.3 $\pm$ 0.6 & 79.1 $\pm$ 6.7 & 63.6 $\pm$ 0.1 & 78.0 $\pm$ 0.4 & 89.1 $\pm$ 0.4 & 78.7 $\pm$ 0.2 & 93.8 $\pm$ 0.3 & 56.4 $\pm$ 0.4 & 74.1 $\pm$ 0.8 & 69.6 $\pm$ 0.9 \\
\cmidrule(lr){2-14}
& DynSparseAdam & 59.8 $\pm$ 0.2 & 66.4 $\pm$ 0.4 & 24.2 $\pm$ 0.6 & 80.4 $\pm$ 1.9 & 62.7 $\pm$ 1.0 & 77.0 $\pm$ 0.2 & 87.3 $\pm$ 1.1 & 81.5 $\pm$ 1.4 & 93.1 $\pm$ 0.5 & 55.2 $\pm$ 0.7 & 73.9 $\pm$ 0.6 & 69.2 $\pm$ 0.8 \\
& DynSparseSGD & 64.7 $\pm$ 0.0 & 68.2 $\pm$ 0.2 & 23.2 $\pm$ 0.7 & 73.4 $\pm$ 4.7 & 63.5 $\pm$ 0.3 & 80.5 $\pm$ 0.3 & 90.1 $\pm$ 0.7 & 75.8 $\pm$ 1.2 & 93.2 $\pm$ 0.4 & 54.1 $\pm$ 0.4 & 71.1 $\pm$ 0.8 & 68.9 $\pm$ 0.9 \\
& DynSignSparseSGD & 63.2 $\pm$ 0.1 & 68.2 $\pm$ 0.3 & 25.5 $\pm$ 0.6 & 79.8 $\pm$ 4.2 & 66.0 $\pm$ 0.1 & 79.5 $\pm$ 0.4 & 89.2 $\pm$ 0.3 & 83.8 $\pm$ 1.4 & 94.0 $\pm$ 0.3 & 57.1 $\pm$ 1.1 & 74.8 $\pm$ 0.6 & 71.0 $\pm$ 0.9 \\
\midrule
\multirow{9}{*}{4} & SignSGD & 13.5 $\pm$ 0.5 & 22.9 $\pm$ 1.5 & 1.5 $\pm$ 0.1 & 16.5 $\pm$ 4.0 & 0.9 $\pm$ 0.0 & 1.3 $\pm$ 0.1 & 3.3 $\pm$ 0.8 & 13.1 $\pm$ 5.4 & 18.4 $\pm$ 23.7 & 20.3 $\pm$ 2.2 & 3.0 $\pm$ 0.8 & 10.4 $\pm$ 3.6 \\
& SignAdam & 0.1 $\pm$ 0.0 & 0.3 $\pm$ 0.0 & 1.1 $\pm$ 0.1 & 9.8 $\pm$ 1.1 & 0.5 $\pm$ 0.1 & 1.0 $\pm$ 0.1 & 2.7 $\pm$ 0.2 & 0.8 $\pm$ 0.3 & 0.8 $\pm$ 0.2 & 2.2 $\pm$ 0.1 & 1.0 $\pm$ 0.2 & 1.8 $\pm$ 0.2 \\
\cmidrule(lr){2-14}
& SparseSGD & 65.6 $\pm$ 0.1 & 68.0 $\pm$ 0.1 & 25.0 $\pm$ 0.0 & 72.1 $\pm$ 1.4 & 63.8 $\pm$ 0.2 & 80.9 $\pm$ 0.1 & 90.5 $\pm$ 0.5 & 75.8 $\pm$ 0.8 & 94.1 $\pm$ 0.2 & 53.5 $\pm$ 0.5 & 71.2 $\pm$ 0.2 & 69.1 $\pm$ 0.4 \\
& SparseAdam & 62.6 $\pm$ 0.4 & 66.0 $\pm$ 0.3 & 24.6 $\pm$ 0.5 & 83.1 $\pm$ 1.6 & 61.9 $\pm$ 0.4 & 79.4 $\pm$ 0.4 & 90.4 $\pm$ 0.1 & 83.4 $\pm$ 1.0 & 94.7 $\pm$ 0.2 & 56.7 $\pm$ 0.3 & 73.5 $\pm$ 0.3 & 70.6 $\pm$ 0.5 \\
\cmidrule(lr){2-14}
& SignSparseAdam & 0.1 $\pm$ 0.0 & 0.3 $\pm$ 0.1 & 1.4 $\pm$ 0.2 & 22.3 $\pm$ 2.2 & 0.5 $\pm$ 0.0 & 1.0 $\pm$ 0.0 & 2.8 $\pm$ 0.2 & 1.6 $\pm$ 0.3 & 1.5 $\pm$ 0.1 & 2.4 $\pm$ 0.5 & 0.8 $\pm$ 0.1 & 3.2 $\pm$ 0.3 \\
& SignSparseSGD & 63.3 $\pm$ 0.4 & 69.0 $\pm$ 0.4 & 25.2 $\pm$ 1.1 & 82.6 $\pm$ 1.0 & 66.9 $\pm$ 0.3 & 78.2 $\pm$ 0.4 & 89.3 $\pm$ 0.4 & 85.8 $\pm$ 1.2 & 94.7 $\pm$ 0.4 & 59.1 $\pm$ 1.0 & 75.9 $\pm$ 0.7 & 71.8 $\pm$ 0.7 \\
\cmidrule(lr){2-14}
& DynSparseAdam & 60.8 $\pm$ 0.2 & 68.1 $\pm$ 0.4 & 26.7 $\pm$ 1.1 & 85.8 $\pm$ 0.7 & 67.1 $\pm$ 0.4 & 77.0 $\pm$ 0.1 & 88.7 $\pm$ 0.4 & 86.9 $\pm$ 0.7 & 94.4 $\pm$ 0.3 & 59.8 $\pm$ 0.4 & 76.3 $\pm$ 0.8 & 72.0 $\pm$ 0.5 \\
& DynSparseSGD & 65.8 $\pm$ 0.2 & 69.9 $\pm$ 0.3 & 25.7 $\pm$ 0.5 & 74.1 $\pm$ 0.9 & 65.9 $\pm$ 0.4 & 81.0 $\pm$ 0.2 & 91.2 $\pm$ 0.1 & 81.3 $\pm$ 0.2 & 94.3 $\pm$ 0.2 & 56.3 $\pm$ 0.7 & 73.5 $\pm$ 0.3 & 70.8 $\pm$ 0.4 \\
& DynSignSparseSGD & 64.0 $\pm$ 0.2 & 69.9 $\pm$ 0.3 & 27.7 $\pm$ 0.6 & 83.9 $\pm$ 1.6 & 69.3 $\pm$ 0.1 & 79.6 $\pm$ 0.4 & 90.0 $\pm$ 0.4 & 88.9 $\pm$ 0.8 & 95.2 $\pm$ 0.2 & 61.1 $\pm$ 0.3 & 78.0 $\pm$ 0.9 & 73.4 $\pm$ 0.5 \\
\midrule
\multirow{9}{*}{8} & SignSGD & 7.1 $\pm$ 8.3 & 31.1 $\pm$ 0.2 & 1.9 $\pm$ 0.3 & 27.5 $\pm$ 2.7 & 1.1 $\pm$ 0.2 & 2.2 $\pm$ 0.4 & 16.5 $\pm$ 15.8 & 41.4 $\pm$ 2.6 & 30.4 $\pm$ 29.4 & 16.7 $\pm$ 5.7 & 3.7 $\pm$ 0.1 & 16.3 $\pm$ 6.0 \\
& SignAdam & 0.1 $\pm$ 0.0 & 0.3 $\pm$ 0.0 & 1.0 $\pm$ 0.1 & 10.1 $\pm$ 0.8 & 0.4 $\pm$ 0.1 & 1.0 $\pm$ 0.0 & 2.7 $\pm$ 0.3 & 1.3 $\pm$ 0.3 & 0.7 $\pm$ 0.2 & 2.0 $\pm$ 0.2 & 0.8 $\pm$ 0.2 & 1.9 $\pm$ 0.2 \\
\cmidrule(lr){2-14}
& SparseSGD & 66.3 $\pm$ 0.1 & 69.3 $\pm$ 0.1 & 28.5 $\pm$ 0.8 & 81.1 $\pm$ 1.4 & 65.4 $\pm$ 0.6 & 81.0 $\pm$ 0.3 & 91.0 $\pm$ 0.3 & 80.8 $\pm$ 0.3 & 94.7 $\pm$ 0.1 & 57.3 $\pm$ 0.5 & 73.3 $\pm$ 0.1 & 71.7 $\pm$ 0.4 \\
& SparseAdam & 64.2 $\pm$ 0.2 & 67.7 $\pm$ 0.3 & 28.9 $\pm$ 0.2 & 88.0 $\pm$ 2.0 & 65.3 $\pm$ 0.4 & 79.4 $\pm$ 0.4 & 90.5 $\pm$ 0.3 & 87.4 $\pm$ 0.5 & 94.8 $\pm$ 0.5 & 60.9 $\pm$ 1.1 & 75.7 $\pm$ 0.5 & 73.0 $\pm$ 0.6 \\
\cmidrule(lr){2-14}
& SignSparseAdam & 0.1 $\pm$ 0.0 & 0.3 $\pm$ 0.0 & 1.9 $\pm$ 0.1 & 19.2 $\pm$ 7.0 & 0.5 $\pm$ 0.1 & 1.0 $\pm$ 0.1 & 3.2 $\pm$ 0.7 & 1.7 $\pm$ 1.1 & 1.3 $\pm$ 0.6 & 3.3 $\pm$ 0.3 & 1.0 $\pm$ 0.1 & 3.0 $\pm$ 0.9 \\
& SignSparseSGD & 64.7 $\pm$ 0.1 & 70.9 $\pm$ 0.3 & 31.1 $\pm$ 0.8 & 85.9 $\pm$ 0.5 & 71.4 $\pm$ 0.5 & 79.0 $\pm$ 0.2 & 90.3 $\pm$ 0.1 & 89.3 $\pm$ 0.7 & 94.9 $\pm$ 0.3 & 62.3 $\pm$ 0.9 & 78.8 $\pm$ 0.7 & 74.4 $\pm$ 0.5 \\
\cmidrule(lr){2-14}
& DynSparseAdam & 62.0 $\pm$ 0.2 & 69.8 $\pm$ 0.4 & 33.7 $\pm$ 0.3 & 88.8 $\pm$ 0.7 & 73.3 $\pm$ 0.5 & 77.6 $\pm$ 0.2 & 90.0 $\pm$ 0.5 & 90.9 $\pm$ 0.2 & 94.5 $\pm$ 0.2 & 62.6 $\pm$ 0.4 & 79.4 $\pm$ 0.7 & 74.8 $\pm$ 0.4 \\
& DynSparseSGD & 66.7 $\pm$ 0.1 & 71.3 $\pm$ 0.0 & 29.8 $\pm$ 0.7 & 83.3 $\pm$ 2.0 & 68.4 $\pm$ 0.4 & 81.5 $\pm$ 0.1 & 91.6 $\pm$ 0.3 & 86.0 $\pm$ 0.6 & 94.8 $\pm$ 0.1 & 60.8 $\pm$ 0.8 & 75.9 $\pm$ 0.7 & 73.6 $\pm$ 0.5 \\
& DynSignSparseSGD & 65.0 $\pm$ 0.2 & 71.9 $\pm$ 0.1 & 34.8 $\pm$ 0.6 & 88.4 $\pm$ 1.0 & 75.1 $\pm$ 0.2 & 79.8 $\pm$ 0.1 & 91.0 $\pm$ 0.3 & 92.5 $\pm$ 0.1 & 95.5 $\pm$ 0.3 & 64.5 $\pm$ 1.0 & 79.7 $\pm$ 0.6 & 76.2 $\pm$ 0.4 \\
\midrule
\multirow{9}{*}{16} & SignSGD & 24.8 $\pm$ 0.6 & 17.0 $\pm$ 14.0 & 2.4 $\pm$ 0.4 & 33.0 $\pm$ 3.3 & 2.2 $\pm$ 0.1 & 2.5 $\pm$ 0.8 & 3.5 $\pm$ 0.4 & 40.8 $\pm$ 14.7 & 73.5 $\pm$ 1.5 & 29.8 $\pm$ 9.3 & 9.8 $\pm$ 2.3 & 21.7 $\pm$ 4.3 \\
& SignAdam & 0.1 $\pm$ 0.0 & 0.3 $\pm$ 0.0 & 0.8 $\pm$ 0.1 & 11.4 $\pm$ 0.9 & 0.5 $\pm$ 0.1 & 1.0 $\pm$ 0.0 & 2.7 $\pm$ 0.1 & 1.3 $\pm$ 0.4 & 0.9 $\pm$ 0.1 & 2.0 $\pm$ 0.2 & 1.2 $\pm$ 0.1 & 2.0 $\pm$ 0.2 \\
\cmidrule(lr){2-14}
& SparseSGD & 67.1 $\pm$ 0.1 & 70.8 $\pm$ 0.1 & 31.9 $\pm$ 1.0 & 88.2 $\pm$ 0.5 & 68.2 $\pm$ 0.2 & 81.5 $\pm$ 0.1 & 91.4 $\pm$ 0.2 & 86.5 $\pm$ 0.1 & 95.0 $\pm$ 0.2 & 62.0 $\pm$ 0.6 & 76.4 $\pm$ 0.2 & 74.4 $\pm$ 0.3 \\
& SparseAdam & 66.3 $\pm$ 0.2 & 69.2 $\pm$ 0.5 & 35.4 $\pm$ 0.3 & 92.3 $\pm$ 0.7 & 69.9 $\pm$ 0.2 & 79.7 $\pm$ 0.3 & 90.6 $\pm$ 0.1 & 91.2 $\pm$ 0.3 & 95.3 $\pm$ 0.1 & 65.3 $\pm$ 0.5 & 78.2 $\pm$ 0.4 & 75.8 $\pm$ 0.3 \\
\cmidrule(lr){2-14}
& SignSparseAdam & 0.1 $\pm$ 0.0 & 0.3 $\pm$ 0.0 & 2.0 $\pm$ 0.3 & 29.0 $\pm$ 6.5 & 0.5 $\pm$ 0.0 & 1.0 $\pm$ 0.0 & 2.9 $\pm$ 0.3 & 3.3 $\pm$ 2.0 & 2.2 $\pm$ 2.1 & 3.1 $\pm$ 0.5 & 1.6 $\pm$ 0.8 & 4.2 $\pm$ 1.1 \\
& SignSparseSGD & 66.3 $\pm$ 0.2 & 73.1 $\pm$ 0.3 & 36.4 $\pm$ 0.9 & 90.7 $\pm$ 1.6 & 76.4 $\pm$ 0.6 & 79.3 $\pm$ 0.5 & 89.9 $\pm$ 1.0 & 93.8 $\pm$ 1.0 & 95.8 $\pm$ 0.0 & 67.5 $\pm$ 0.2 & 81.7 $\pm$ 0.4 & 77.4 $\pm$ 0.6 \\
\cmidrule(lr){2-14}
& DynSparseAdam & 64.0 $\pm$ 0.1 & 71.9 $\pm$ 0.0 & 41.8 $\pm$ 1.1 & 91.6 $\pm$ 1.4 & 78.1 $\pm$ 0.2 & 78.1 $\pm$ 0.7 & 89.5 $\pm$ 0.6 & 94.6 $\pm$ 1.2 & 95.2 $\pm$ 0.2 & 67.1 $\pm$ 0.4 & 81.9 $\pm$ 0.9 & 77.6 $\pm$ 0.6 \\
& DynSparseSGD & 67.9 $\pm$ 0.0 & 73.2 $\pm$ 0.0 & 33.7 $\pm$ 0.1 & 88.4 $\pm$ 0.6 & 72.8 $\pm$ 0.1 & 82.0 $\pm$ 0.2 & 91.6 $\pm$ 0.2 & 91.9 $\pm$ 0.5 & 95.5 $\pm$ 0.1 & 66.9 $\pm$ 0.6 & 79.4 $\pm$ 0.3 & 76.7 $\pm$ 0.2 \\
& DynSignSparseSGD & 66.9 $\pm$ 0.2 & 73.5 $\pm$ 0.1 & 42.0 $\pm$ 1.1 & 92.0 $\pm$ 0.6 & 79.8 $\pm$ 0.4 & 80.3 $\pm$ 0.4 & 91.0 $\pm$ 0.3 & 95.7 $\pm$ 1.0 & 95.8 $\pm$ 0.0 & 69.2 $\pm$ 0.5 & 82.7 $\pm$ 0.5 & 79.0 $\pm$ 0.5 \\
\bottomrule
\end{tabular}
\end{adjustbox}
\end{table}
\begin{table}
\caption{Few-shot classification performance on 11 datasets using the CLIP (ViT-B/16) backbone. We compare SignSparseSGD and DynSignSparseSGD against multiple optimizer families: sign-based optimizers (SignSGD, SignAdam), sparsity-based optimizers (SparseSGD, SparseAdam), sign-sparse optimizers (SignSparseAdam, SignSparseSGD), and dynamic-sparse optimizers (DynSparseAdam, DynSparseSGD, DynSignSparseSGD). Top-1 accuracy (\%) averaged over fixed 3 seeds.}
\label{tab:ablation_optim_ViTB16_with_aug}
\centering
\footnotesize
\setlength{\tabcolsep}{1.8pt}
\renewcommand{\arraystretch}{1.00}
\begin{adjustbox}{max totalsize={\textwidth}{0.95\textheight},center}
\begin{tabular}{@{}c l *{12}{d}@{}}
\toprule
Shots & Method & ImageNet & SUN & Aircr. & EuroSAT & Cars & Food & Pets & Flow. & Caltech & DTD & UCF & Avg. \\ \midrule
\multirow{1}{*}{0} & CLIP & 66.7 & 62.6 & 23.9 & 42.2 & 65.5 & 85.2 & 88.3 & 67.4 & 93.3 & 44.4 & 65.2 & 64.1 \\
\midrule
\multirow{9}{*}{1} & SignSGD & 8.9 $\pm$ 1.0 & 4.5 $\pm$ 5.2 & 1.1 $\pm$ 0.2 & 22.9 $\pm$ 3.9 & 0.6 $\pm$ 0.1 & 1.4 $\pm$ 0.4 & 3.2 $\pm$ 0.4 & 3.1 $\pm$ 1.0 & 28.3 $\pm$ 17.5 & 13.0 $\pm$ 5.2 & 1.0 $\pm$ 0.2 & 8.0 $\pm$ 3.2 \\
& SignAdam & 0.1 $\pm$ 0.0 & 0.2 $\pm$ 0.0 & 1.1 $\pm$ 0.0 & 9.5 $\pm$ 0.1 & 0.4 $\pm$ 0.0 & 1.0 $\pm$ 0.0 & 2.7 $\pm$ 0.1 & 1.1 $\pm$ 0.1 & 0.5 $\pm$ 0.2 & 2.2 $\pm$ 0.2 & 1.1 $\pm$ 0.2 & 1.8 $\pm$ 0.1 \\
\cmidrule(lr){2-14}
& SparseSGD & 70.0 $\pm$ 0.1 & 68.6 $\pm$ 0.2 & 28.9 $\pm$ 0.3 & 65.2 $\pm$ 1.0 & 68.0 $\pm$ 0.3 & 85.6 $\pm$ 0.5 & 91.3 $\pm$ 0.4 & 73.6 $\pm$ 0.2 & 94.3 $\pm$ 0.1 & 51.8 $\pm$ 0.6 & 71.3 $\pm$ 0.1 & 69.9 $\pm$ 0.4 \\
& SparseAdam & 66.1 $\pm$ 0.1 & 67.2 $\pm$ 0.3 & 27.4 $\pm$ 0.8 & 68.9 $\pm$ 3.5 & 64.8 $\pm$ 0.7 & 84.1 $\pm$ 0.4 & 89.9 $\pm$ 0.9 & 80.7 $\pm$ 0.9 & 93.9 $\pm$ 0.2 & 52.2 $\pm$ 1.0 & 73.7 $\pm$ 0.4 & 69.9 $\pm$ 0.8 \\
\cmidrule(lr){2-14}
& SignSparseAdam & 0.1 $\pm$ 0.0 & 0.3 $\pm$ 0.1 & 1.1 $\pm$ 0.1 & 13.2 $\pm$ 2.0 & 0.5 $\pm$ 0.0 & 1.0 $\pm$ 0.0 & 2.7 $\pm$ 0.6 & 0.9 $\pm$ 0.3 & 1.4 $\pm$ 0.6 & 2.1 $\pm$ 0.2 & 0.7 $\pm$ 0.5 & 2.2 $\pm$ 0.4 \\
& SignSparseSGD & 66.8 $\pm$ 0.1 & 68.2 $\pm$ 0.3 & 29.4 $\pm$ 1.3 & 71.2 $\pm$ 3.7 & 69.1 $\pm$ 0.6 & 83.8 $\pm$ 0.5 & 91.3 $\pm$ 0.3 & 83.3 $\pm$ 0.5 & 93.7 $\pm$ 0.2 & 54.3 $\pm$ 0.6 & 75.7 $\pm$ 0.3 & 71.5 $\pm$ 0.8 \\
\cmidrule(lr){2-14}
& DynSparseAdam & 65.4 $\pm$ 0.1 & 67.9 $\pm$ 0.4 & 29.4 $\pm$ 1.1 & 74.0 $\pm$ 4.4 & 68.4 $\pm$ 1.0 & 83.2 $\pm$ 0.7 & 91.5 $\pm$ 0.5 & 84.1 $\pm$ 0.3 & 93.4 $\pm$ 0.1 & 53.6 $\pm$ 0.1 & 75.9 $\pm$ 0.8 & 71.5 $\pm$ 0.9 \\
& DynSparseSGD & 69.9 $\pm$ 0.1 & 69.5 $\pm$ 0.2 & 30.0 $\pm$ 0.5 & 73.0 $\pm$ 1.6 & 67.8 $\pm$ 0.7 & 86.0 $\pm$ 0.1 & 92.3 $\pm$ 0.3 & 75.8 $\pm$ 0.5 & 94.1 $\pm$ 0.2 & 52.9 $\pm$ 0.6 & 72.8 $\pm$ 0.4 & 71.3 $\pm$ 0.5 \\
& DynSignSparseSGD & 68.1 $\pm$ 0.1 & 69.8 $\pm$ 0.3 & 30.6 $\pm$ 1.2 & 73.7 $\pm$ 4.0 & 70.6 $\pm$ 0.4 & 85.1 $\pm$ 0.3 & 92.5 $\pm$ 0.2 & 84.7 $\pm$ 0.4 & 93.8 $\pm$ 0.1 & 54.8 $\pm$ 0.6 & 76.2 $\pm$ 0.4 & 72.7 $\pm$ 0.7 \\
\midrule
\multirow{9}{*}{2} & SignSGD & 8.6 $\pm$ 5.8 & 20.7 $\pm$ 1.0 & 1.6 $\pm$ 0.1 & 22.9 $\pm$ 5.1 & 0.7 $\pm$ 0.1 & 1.3 $\pm$ 0.4 & 3.5 $\pm$ 0.6 & 3.2 $\pm$ 0.6 & 0.9 $\pm$ 0.1 & 8.7 $\pm$ 6.0 & 1.9 $\pm$ 0.3 & 6.7 $\pm$ 1.8 \\
& SignAdam & 0.1 $\pm$ 0.0 & 0.3 $\pm$ 0.0 & 1.0 $\pm$ 0.1 & 10.0 $\pm$ 1.0 & 0.6 $\pm$ 0.1 & 1.1 $\pm$ 0.1 & 2.7 $\pm$ 0.1 & 0.8 $\pm$ 0.3 & 1.2 $\pm$ 0.6 & 1.9 $\pm$ 0.1 & 0.8 $\pm$ 0.1 & 1.8 $\pm$ 0.2 \\
\cmidrule(lr){2-14}
& SparseSGD & 70.5 $\pm$ 0.1 & 69.7 $\pm$ 0.1 & 31.2 $\pm$ 0.4 & 73.3 $\pm$ 0.9 & 69.1 $\pm$ 0.2 & 86.0 $\pm$ 0.3 & 92.2 $\pm$ 0.5 & 76.3 $\pm$ 0.7 & 94.4 $\pm$ 0.1 & 54.3 $\pm$ 1.3 & 73.4 $\pm$ 0.6 & 71.9 $\pm$ 0.5 \\
& SparseAdam & 66.9 $\pm$ 0.3 & 68.8 $\pm$ 0.3 & 31.1 $\pm$ 0.9 & 78.7 $\pm$ 2.7 & 67.8 $\pm$ 0.2 & 84.8 $\pm$ 0.2 & 92.5 $\pm$ 0.6 & 85.0 $\pm$ 0.5 & 94.6 $\pm$ 0.1 & 57.4 $\pm$ 2.1 & 76.0 $\pm$ 0.3 & 73.1 $\pm$ 0.7 \\
\cmidrule(lr){2-14}
& SignSparseAdam & 0.1 $\pm$ 0.0 & 0.4 $\pm$ 0.0 & 0.9 $\pm$ 0.1 & 15.0 $\pm$ 2.0 & 0.6 $\pm$ 0.1 & 1.0 $\pm$ 0.0 & 3.0 $\pm$ 0.2 & 1.4 $\pm$ 0.4 & 0.6 $\pm$ 0.1 & 3.2 $\pm$ 0.3 & 1.1 $\pm$ 0.3 & 2.5 $\pm$ 0.3 \\
& SignSparseSGD & 67.4 $\pm$ 0.4 & 69.7 $\pm$ 0.1 & 32.0 $\pm$ 1.4 & 79.5 $\pm$ 2.2 & 71.7 $\pm$ 0.2 & 84.1 $\pm$ 0.3 & 92.0 $\pm$ 0.3 & 87.2 $\pm$ 0.5 & 94.2 $\pm$ 0.3 & 59.5 $\pm$ 1.2 & 78.4 $\pm$ 1.0 & 74.2 $\pm$ 0.7 \\
\cmidrule(lr){2-14}
& DynSparseAdam & 65.7 $\pm$ 0.4 & 69.3 $\pm$ 0.3 & 32.6 $\pm$ 1.4 & 83.1 $\pm$ 1.6 & 71.9 $\pm$ 0.5 & 83.5 $\pm$ 0.3 & 91.4 $\pm$ 0.7 & 88.6 $\pm$ 0.7 & 93.7 $\pm$ 0.4 & 59.8 $\pm$ 0.2 & 78.7 $\pm$ 1.0 & 74.4 $\pm$ 0.7 \\
& DynSparseSGD & 70.4 $\pm$ 0.1 & 70.9 $\pm$ 0.1 & 31.2 $\pm$ 1.4 & 77.8 $\pm$ 2.5 & 69.8 $\pm$ 0.5 & 86.2 $\pm$ 0.1 & 92.5 $\pm$ 0.3 & 81.8 $\pm$ 0.2 & 94.3 $\pm$ 0.3 & 57.2 $\pm$ 0.8 & 75.6 $\pm$ 0.6 & 73.4 $\pm$ 0.6 \\
& DynSignSparseSGD & 68.6 $\pm$ 0.2 & 71.0 $\pm$ 0.1 & 33.8 $\pm$ 2.1 & 82.0 $\pm$ 1.4 & 73.7 $\pm$ 0.3 & 85.1 $\pm$ 0.1 & 91.7 $\pm$ 0.4 & 89.6 $\pm$ 0.2 & 94.3 $\pm$ 0.4 & 60.1 $\pm$ 1.2 & 79.7 $\pm$ 0.9 & 75.4 $\pm$ 0.7 \\
\midrule
\multirow{9}{*}{4} & SignSGD & 17.8 $\pm$ 0.3 & 27.7 $\pm$ 0.8 & 1.6 $\pm$ 0.2 & 23.8 $\pm$ 3.6 & 0.9 $\pm$ 0.1 & 1.5 $\pm$ 0.6 & 3.7 $\pm$ 0.7 & 7.0 $\pm$ 3.0 & 7.4 $\pm$ 5.1 & 12.1 $\pm$ 8.5 & 3.0 $\pm$ 0.2 & 9.7 $\pm$ 2.1 \\
& SignAdam & 0.1 $\pm$ 0.0 & 0.3 $\pm$ 0.0 & 0.9 $\pm$ 0.1 & 9.6 $\pm$ 1.3 & 0.4 $\pm$ 0.0 & 1.0 $\pm$ 0.1 & 2.8 $\pm$ 0.2 & 0.7 $\pm$ 0.2 & 1.0 $\pm$ 0.6 & 2.5 $\pm$ 0.5 & 1.0 $\pm$ 0.0 & 1.8 $\pm$ 0.3 \\
\cmidrule(lr){2-14}
& SparseSGD & 71.0 $\pm$ 0.1 & 71.1 $\pm$ 0.2 & 33.6 $\pm$ 0.6 & 75.6 $\pm$ 1.2 & 70.9 $\pm$ 0.2 & 86.6 $\pm$ 0.2 & 93.0 $\pm$ 0.1 & 80.7 $\pm$ 0.4 & 94.7 $\pm$ 0.3 & 56.5 $\pm$ 0.5 & 75.1 $\pm$ 0.2 & 73.5 $\pm$ 0.4 \\
& SparseAdam & 68.2 $\pm$ 0.2 & 69.5 $\pm$ 0.2 & 32.8 $\pm$ 0.9 & 83.6 $\pm$ 1.3 & 70.0 $\pm$ 0.2 & 85.5 $\pm$ 0.2 & 93.2 $\pm$ 0.4 & 88.7 $\pm$ 0.3 & 95.2 $\pm$ 0.3 & 61.6 $\pm$ 0.8 & 77.9 $\pm$ 0.9 & 75.1 $\pm$ 0.5 \\
\cmidrule(lr){2-14}
& SignSparseAdam & 0.1 $\pm$ 0.0 & 0.4 $\pm$ 0.1 & 1.5 $\pm$ 0.2 & 19.4 $\pm$ 0.7 & 0.5 $\pm$ 0.0 & 1.0 $\pm$ 0.0 & 2.8 $\pm$ 0.1 & 1.1 $\pm$ 0.3 & 0.9 $\pm$ 0.5 & 2.6 $\pm$ 0.3 & 1.7 $\pm$ 0.3 & 2.9 $\pm$ 0.2 \\
& SignSparseSGD & 68.5 $\pm$ 0.3 & 71.9 $\pm$ 0.2 & 37.3 $\pm$ 0.7 & 81.5 $\pm$ 1.2 & 75.7 $\pm$ 0.7 & 84.7 $\pm$ 0.3 & 92.5 $\pm$ 0.2 & 91.4 $\pm$ 1.1 & 95.4 $\pm$ 0.1 & 62.6 $\pm$ 0.7 & 81.1 $\pm$ 1.1 & 76.6 $\pm$ 0.6 \\
\cmidrule(lr){2-14}
& DynSparseAdam & 66.8 $\pm$ 0.2 & 70.6 $\pm$ 0.2 & 37.4 $\pm$ 1.1 & 84.7 $\pm$ 1.9 & 76.4 $\pm$ 0.2 & 84.2 $\pm$ 0.2 & 91.9 $\pm$ 0.3 & 91.8 $\pm$ 0.5 & 94.8 $\pm$ 0.4 & 63.2 $\pm$ 0.0 & 81.1 $\pm$ 0.9 & 76.6 $\pm$ 0.5 \\
& DynSparseSGD & 71.2 $\pm$ 0.1 & 72.5 $\pm$ 0.2 & 34.4 $\pm$ 1.0 & 78.3 $\pm$ 0.9 & 73.3 $\pm$ 0.1 & 86.8 $\pm$ 0.0 & 93.2 $\pm$ 0.1 & 87.1 $\pm$ 0.5 & 94.9 $\pm$ 0.3 & 58.8 $\pm$ 0.4 & 77.8 $\pm$ 0.1 & 75.3 $\pm$ 0.3 \\
& DynSignSparseSGD & 69.4 $\pm$ 0.2 & 72.8 $\pm$ 0.1 & 38.4 $\pm$ 1.1 & 84.1 $\pm$ 2.0 & 77.8 $\pm$ 0.3 & 85.4 $\pm$ 0.2 & 92.8 $\pm$ 0.3 & 93.2 $\pm$ 0.8 & 95.2 $\pm$ 0.1 & 64.5 $\pm$ 0.7 & 82.2 $\pm$ 1.2 & 77.8 $\pm$ 0.6 \\
\midrule
\multirow{9}{*}{8} & SignSGD & 24.6 $\pm$ 0.5 & 25.1 $\pm$ 12.9 & 1.9 $\pm$ 0.1 & 19.2 $\pm$ 6.9 & 1.3 $\pm$ 0.2 & 1.8 $\pm$ 0.7 & 3.0 $\pm$ 0.5 & 23.4 $\pm$ 4.3 & 26.5 $\pm$ 26.6 & 14.8 $\pm$ 13.8 & 7.0 $\pm$ 0.9 & 13.5 $\pm$ 6.1 \\
& SignAdam & 0.1 $\pm$ 0.0 & 0.2 $\pm$ 0.0 & 0.9 $\pm$ 0.1 & 8.6 $\pm$ 1.7 & 0.5 $\pm$ 0.0 & 1.0 $\pm$ 0.0 & 2.6 $\pm$ 0.1 & 0.9 $\pm$ 0.2 & 0.8 $\pm$ 0.1 & 2.3 $\pm$ 0.1 & 0.9 $\pm$ 0.1 & 1.7 $\pm$ 0.2 \\
\cmidrule(lr){2-14}
& SparseSGD & 71.5 $\pm$ 0.0 & 72.1 $\pm$ 0.1 & 37.4 $\pm$ 0.6 & 82.3 $\pm$ 0.3 & 72.9 $\pm$ 0.2 & 86.9 $\pm$ 0.3 & 93.0 $\pm$ 0.2 & 84.6 $\pm$ 0.2 & 95.3 $\pm$ 0.1 & 59.8 $\pm$ 0.7 & 77.3 $\pm$ 0.3 & 75.7 $\pm$ 0.3 \\
& SparseAdam & 69.6 $\pm$ 0.3 & 70.7 $\pm$ 0.2 & 36.3 $\pm$ 0.9 & 88.9 $\pm$ 0.9 & 73.5 $\pm$ 0.5 & 85.7 $\pm$ 0.3 & 93.3 $\pm$ 0.1 & 91.8 $\pm$ 0.3 & 95.8 $\pm$ 0.2 & 65.4 $\pm$ 1.4 & 79.6 $\pm$ 0.5 & 77.3 $\pm$ 0.5 \\
\cmidrule(lr){2-14}
& SignSparseAdam & 0.1 $\pm$ 0.0 & 0.3 $\pm$ 0.1 & 1.5 $\pm$ 0.4 & 19.8 $\pm$ 3.8 & 0.6 $\pm$ 0.2 & 1.1 $\pm$ 0.1 & 3.6 $\pm$ 0.9 & 2.6 $\pm$ 1.1 & 0.8 $\pm$ 0.2 & 3.1 $\pm$ 0.7 & 2.1 $\pm$ 0.2 & 3.2 $\pm$ 0.7 \\
& SignSparseSGD & 69.9 $\pm$ 0.3 & 73.7 $\pm$ 0.1 & 41.1 $\pm$ 1.7 & 84.7 $\pm$ 5.2 & 79.8 $\pm$ 0.7 & 84.8 $\pm$ 0.6 & 92.9 $\pm$ 0.4 & 94.0 $\pm$ 0.3 & 95.9 $\pm$ 0.2 & 66.5 $\pm$ 0.7 & 83.6 $\pm$ 0.6 & 78.8 $\pm$ 1.0 \\
\cmidrule(lr){2-14}
& DynSparseAdam & 68.0 $\pm$ 0.3 & 73.0 $\pm$ 0.3 & 44.7 $\pm$ 0.8 & 86.7 $\pm$ 2.5 & 81.2 $\pm$ 0.2 & 83.8 $\pm$ 0.3 & 92.7 $\pm$ 0.3 & 94.1 $\pm$ 0.6 & 95.8 $\pm$ 0.2 & 66.2 $\pm$ 1.4 & 83.8 $\pm$ 0.3 & 79.1 $\pm$ 0.6 \\
& DynSparseSGD & 72.1 $\pm$ 0.2 & 73.7 $\pm$ 0.2 & 39.1 $\pm$ 0.2 & 83.3 $\pm$ 1.9 & 76.3 $\pm$ 0.4 & 87.0 $\pm$ 0.1 & 93.4 $\pm$ 0.1 & 91.4 $\pm$ 0.2 & 95.1 $\pm$ 0.2 & 63.7 $\pm$ 0.6 & 79.9 $\pm$ 0.1 & 77.7 $\pm$ 0.4 \\
& DynSignSparseSGD & 70.5 $\pm$ 0.4 & 74.5 $\pm$ 0.2 & 45.2 $\pm$ 0.7 & 85.2 $\pm$ 4.0 & 82.4 $\pm$ 0.5 & 85.3 $\pm$ 0.3 & 93.4 $\pm$ 0.1 & 95.0 $\pm$ 0.2 & 95.8 $\pm$ 0.1 & 68.0 $\pm$ 1.3 & 84.7 $\pm$ 0.2 & 80.0 $\pm$ 0.7 \\
\midrule
\multirow{9}{*}{16} & SignSGD & 31.7 $\pm$ 0.4 & 29.3 $\pm$ 16.0 & 2.5 $\pm$ 0.8 & 32.2 $\pm$ 1.3 & 2.3 $\pm$ 0.2 & 3.1 $\pm$ 0.3 & 4.0 $\pm$ 0.4 & 35.7 $\pm$ 2.0 & 21.7 $\pm$ 13.8 & 11.0 $\pm$ 1.7 & 8.4 $\pm$ 1.1 & 16.5 $\pm$ 3.4 \\
& SignAdam & 0.1 $\pm$ 0.0 & 0.2 $\pm$ 0.0 & 1.1 $\pm$ 0.1 & 9.1 $\pm$ 0.9 & 0.5 $\pm$ 0.1 & 1.0 $\pm$ 0.0 & 2.7 $\pm$ 0.0 & 1.7 $\pm$ 0.8 & 0.7 $\pm$ 0.2 & 2.1 $\pm$ 0.4 & 0.9 $\pm$ 0.1 & 1.8 $\pm$ 0.2 \\
\cmidrule(lr){2-14}
& SparseSGD & 72.4 $\pm$ 0.0 & 73.4 $\pm$ 0.0 & 41.5 $\pm$ 0.2 & 86.9 $\pm$ 0.5 & 75.7 $\pm$ 0.1 & 87.3 $\pm$ 0.1 & 93.5 $\pm$ 0.2 & 90.6 $\pm$ 0.6 & 95.6 $\pm$ 0.1 & 64.5 $\pm$ 0.5 & 80.4 $\pm$ 0.4 & 78.3 $\pm$ 0.2 \\
& SparseAdam & 71.4 $\pm$ 0.1 & 72.2 $\pm$ 0.1 & 41.8 $\pm$ 0.2 & 93.1 $\pm$ 0.5 & 77.1 $\pm$ 0.1 & 85.9 $\pm$ 0.3 & 93.6 $\pm$ 0.4 & 95.1 $\pm$ 0.1 & 95.9 $\pm$ 0.2 & 69.6 $\pm$ 0.7 & 82.5 $\pm$ 0.2 & 79.8 $\pm$ 0.3 \\
\cmidrule(lr){2-14}
& SignSparseAdam & 0.1 $\pm$ 0.0 & 0.3 $\pm$ 0.1 & 1.5 $\pm$ 0.4 & 24.5 $\pm$ 4.4 & 0.5 $\pm$ 0.1 & 1.0 $\pm$ 0.0 & 3.7 $\pm$ 0.5 & 6.0 $\pm$ 4.1 & 2.4 $\pm$ 2.1 & 3.4 $\pm$ 0.4 & 4.0 $\pm$ 0.8 & 4.3 $\pm$ 1.2 \\
& SignSparseSGD & 71.6 $\pm$ 0.1 & 75.0 $\pm$ 0.2 & 45.6 $\pm$ 0.9 & 91.2 $\pm$ 1.1 & 83.6 $\pm$ 0.5 & 85.7 $\pm$ 0.1 & 93.6 $\pm$ 0.3 & 96.8 $\pm$ 0.2 & 96.2 $\pm$ 0.2 & 71.2 $\pm$ 0.3 & 85.8 $\pm$ 0.0 & 81.5 $\pm$ 0.4 \\
\cmidrule(lr){2-14}
& DynSparseAdam & 69.5 $\pm$ 0.1 & 74.2 $\pm$ 0.3 & 51.2 $\pm$ 0.3 & 91.3 $\pm$ 0.7 & 84.9 $\pm$ 0.2 & 84.6 $\pm$ 0.1 & 93.2 $\pm$ 0.3 & 96.1 $\pm$ 0.7 & 96.0 $\pm$ 0.1 & 71.7 $\pm$ 0.2 & 85.9 $\pm$ 0.7 & 81.7 $\pm$ 0.3 \\
& DynSparseSGD & 73.1 $\pm$ 0.1 & 75.5 $\pm$ 0.0 & 44.3 $\pm$ 0.1 & 88.7 $\pm$ 1.0 & 79.9 $\pm$ 0.3 & 87.3 $\pm$ 0.2 & 93.9 $\pm$ 0.1 & 95.3 $\pm$ 0.1 & 95.6 $\pm$ 0.1 & 69.3 $\pm$ 0.8 & 83.5 $\pm$ 0.2 & 80.6 $\pm$ 0.3 \\
& DynSignSparseSGD & 72.2 $\pm$ 0.1 & 75.7 $\pm$ 0.2 & 51.7 $\pm$ 0.6 & 91.8 $\pm$ 0.5 & 85.9 $\pm$ 0.3 & 86.1 $\pm$ 0.1 & 93.9 $\pm$ 0.1 & 97.7 $\pm$ 0.2 & 96.4 $\pm$ 0.3 & 72.2 $\pm$ 0.4 & 86.9 $\pm$ 0.1 & 82.8 $\pm$ 0.3 \\
\bottomrule
\end{tabular}
\end{adjustbox}
\end{table}
\begin{table}
\caption{Few-shot classification performance on 11 datasets with CLIP (ViT-L/14) backbone. We compare SignSparseSGD and DynSignSparseSGD against multiple optimizer families: sign-based optimizers (SignSGD, SignAdam), sparsity-based optimizers (SparseSGD, SparseAdam), sign-sparse optimizers (SignSparseAdam, SignSparseSGD), and dynamic-sparse optimizers (DynSparseAdam, DynSparseSGD, DynSignSparseSGD). Top-1 accuracy (\%) averaged over fixed 3 seeds. Top-1 accuracy (\%) averaged over 3 seeds.}
\label{tab:ablation_optim_ViTL14_with_aug}
\centering
\footnotesize
\setlength{\tabcolsep}{1.8pt}
\renewcommand{\arraystretch}{1.00}
\begin{adjustbox}{max totalsize={\textwidth}{0.95\textheight},center}
\begin{tabular}{@{}c l *{12}{d}@{}}
\toprule
Shots & Method & ImageNet & SUN & Aircr. & EuroSAT & Cars & Food & Pets & Flow. & Caltech & DTD & UCF & Avg. \\ \midrule
\multirow{1}{*}{0} & CLIP & 73.5 & 67.7 & 30.0 & 55.1 & 76.8 & 90.3 & 93.1 & 76.2 & 95.2 & 52.4 & 73.7 & 71.3 \\
\midrule
\multirow{9}{*}{1} & SignSGD & 14.0 $\pm$ 1.5 & 23.7 $\pm$ 1.6 & 1.2 $\pm$ 0.3 & 17.5 $\pm$ 2.0 & 0.7 $\pm$ 0.1 & 5.8 $\pm$ 6.3 & 21.8 $\pm$ 5.3 & 3.2 $\pm$ 0.2 & 52.7 $\pm$ 12.2 & 21.4 $\pm$ 4.9 & 34.6 $\pm$ 5.0 & 17.9 $\pm$ 3.6 \\
& SignAdam & 0.1 $\pm$ 0.0 & 0.2 $\pm$ 0.0 & 1.0 $\pm$ 0.2 & 11.1 $\pm$ 1.0 & 0.5 $\pm$ 0.0 & 1.0 $\pm$ 0.0 & 2.7 $\pm$ 0.1 & 0.9 $\pm$ 0.1 & 0.9 $\pm$ 0.2 & 2.1 $\pm$ 0.1 & 1.1 $\pm$ 0.2 & 2.0 $\pm$ 0.2 \\
\cmidrule(lr){2-14}
& SparseSGD & 75.9 $\pm$ 0.0 & 71.7 $\pm$ 0.1 & 36.7 $\pm$ 1.0 & 66.9 $\pm$ 2.2 & 78.5 $\pm$ 0.1 & 90.9 $\pm$ 0.1 & 94.2 $\pm$ 0.1 & 80.6 $\pm$ 0.2 & 95.5 $\pm$ 0.5 & 56.4 $\pm$ 1.0 & 77.7 $\pm$ 0.0 & 75.0 $\pm$ 0.5 \\
& SparseAdam & 74.2 $\pm$ 0.6 & 73.7 $\pm$ 0.1 & 38.4 $\pm$ 1.6 & 74.2 $\pm$ 3.5 & 78.5 $\pm$ 0.0 & 89.4 $\pm$ 0.5 & 91.0 $\pm$ 1.9 & 89.1 $\pm$ 0.8 & 96.2 $\pm$ 0.8 & 59.6 $\pm$ 1.7 & 81.3 $\pm$ 0.8 & 76.9 $\pm$ 1.1 \\
\cmidrule(lr){2-14}
& SignSparseAdam & 0.1 $\pm$ 0.0 & 0.3 $\pm$ 0.0 & 1.0 $\pm$ 0.0 & 15.7 $\pm$ 5.3 & 0.5 $\pm$ 0.1 & 1.0 $\pm$ 0.1 & 2.6 $\pm$ 0.3 & 0.6 $\pm$ 0.1 & 1.8 $\pm$ 1.2 & 2.1 $\pm$ 0.1 & 1.0 $\pm$ 0.1 & 2.4 $\pm$ 0.7 \\
& SignSparseSGD & 74.6 $\pm$ 0.5 & 73.3 $\pm$ 0.2 & 40.5 $\pm$ 1.7 & 71.7 $\pm$ 3.7 & 80.0 $\pm$ 0.3 & 90.2 $\pm$ 0.1 & 94.6 $\pm$ 0.2 & 91.5 $\pm$ 1.3 & 95.5 $\pm$ 0.3 & 61.7 $\pm$ 1.1 & 82.8 $\pm$ 0.5 & 77.8 $\pm$ 0.9 \\
\cmidrule(lr){2-14}
& DynSparseAdam & 73.4 $\pm$ 0.4 & 73.0 $\pm$ 0.5 & 39.8 $\pm$ 1.6 & 72.8 $\pm$ 3.3 & 79.2 $\pm$ 0.4 & 90.0 $\pm$ 0.2 & 94.0 $\pm$ 0.4 & 91.4 $\pm$ 1.2 & 95.7 $\pm$ 0.6 & 61.0 $\pm$ 0.8 & 82.5 $\pm$ 1.0 & 77.5 $\pm$ 1.0 \\
& DynSparseSGD & 76.3 $\pm$ 0.0 & 72.6 $\pm$ 0.1 & 38.1 $\pm$ 0.6 & 70.8 $\pm$ 2.5 & 79.3 $\pm$ 0.2 & 91.1 $\pm$ 0.1 & 94.3 $\pm$ 0.1 & 81.4 $\pm$ 0.5 & 95.7 $\pm$ 0.3 & 57.6 $\pm$ 0.9 & 78.2 $\pm$ 0.2 & 75.9 $\pm$ 0.5 \\
& DynSignSparseSGD & 75.6 $\pm$ 0.2 & 74.1 $\pm$ 0.2 & 41.3 $\pm$ 0.9 & 73.8 $\pm$ 4.1 & 80.9 $\pm$ 0.3 & 90.7 $\pm$ 0.2 & 94.7 $\pm$ 0.3 & 92.3 $\pm$ 1.0 & 95.5 $\pm$ 0.2 & 61.5 $\pm$ 0.7 & 82.9 $\pm$ 0.8 & 78.5 $\pm$ 0.8 \\
\midrule
\multirow{9}{*}{2} & SignSGD & 17.6 $\pm$ 1.1 & 28.1 $\pm$ 4.1 & 1.2 $\pm$ 0.2 & 17.0 $\pm$ 5.6 & 0.7 $\pm$ 0.2 & 1.5 $\pm$ 0.4 & 3.6 $\pm$ 1.1 & 4.7 $\pm$ 0.3 & 57.5 $\pm$ 4.8 & 23.8 $\pm$ 6.5 & 26.9 $\pm$ 17.6 & 16.6 $\pm$ 3.8 \\
& SignAdam & 0.1 $\pm$ 0.0 & 0.2 $\pm$ 0.0 & 1.0 $\pm$ 0.2 & 11.1 $\pm$ 1.4 & 0.5 $\pm$ 0.1 & 1.0 $\pm$ 0.1 & 2.8 $\pm$ 0.1 & 1.0 $\pm$ 0.2 & 0.6 $\pm$ 0.1 & 2.0 $\pm$ 0.1 & 0.9 $\pm$ 0.2 & 1.9 $\pm$ 0.2 \\
\cmidrule(lr){2-14}
& SparseSGD & 76.5 $\pm$ 0.1 & 72.9 $\pm$ 0.1 & 38.7 $\pm$ 0.4 & 69.4 $\pm$ 1.7 & 79.4 $\pm$ 0.1 & 91.0 $\pm$ 0.2 & 94.4 $\pm$ 0.1 & 82.6 $\pm$ 0.6 & 95.8 $\pm$ 0.2 & 59.5 $\pm$ 0.1 & 78.6 $\pm$ 0.4 & 76.3 $\pm$ 0.4 \\
& SparseAdam & 74.7 $\pm$ 0.3 & 75.1 $\pm$ 0.2 & 40.9 $\pm$ 1.4 & 85.1 $\pm$ 2.6 & 80.0 $\pm$ 0.2 & 90.7 $\pm$ 0.2 & 94.2 $\pm$ 0.2 & 92.7 $\pm$ 0.7 & 97.1 $\pm$ 0.3 & 66.8 $\pm$ 0.9 & 82.4 $\pm$ 0.8 & 80.0 $\pm$ 0.7 \\
\cmidrule(lr){2-14}
& SignSparseAdam & 0.1 $\pm$ 0.0 & 0.3 $\pm$ 0.0 & 1.0 $\pm$ 0.0 & 12.5 $\pm$ 1.0 & 0.5 $\pm$ 0.1 & 1.0 $\pm$ 0.0 & 2.7 $\pm$ 0.0 & 1.5 $\pm$ 1.1 & 0.8 $\pm$ 0.4 & 2.0 $\pm$ 0.1 & 0.9 $\pm$ 0.0 & 2.1 $\pm$ 0.2 \\
& SignSparseSGD & 75.3 $\pm$ 0.2 & 74.9 $\pm$ 0.1 & 41.9 $\pm$ 1.9 & 84.4 $\pm$ 0.7 & 82.2 $\pm$ 0.4 & 90.5 $\pm$ 0.2 & 94.3 $\pm$ 0.4 & 94.3 $\pm$ 1.0 & 97.0 $\pm$ 0.3 & 67.6 $\pm$ 1.2 & 84.7 $\pm$ 0.7 & 80.6 $\pm$ 0.6 \\
\cmidrule(lr){2-14}
& DynSparseAdam & 73.5 $\pm$ 0.2 & 74.3 $\pm$ 0.4 & 41.7 $\pm$ 2.9 & 85.1 $\pm$ 1.0 & 81.3 $\pm$ 0.8 & 89.6 $\pm$ 0.1 & 93.3 $\pm$ 0.7 & 94.3 $\pm$ 0.5 & 96.7 $\pm$ 0.4 & 67.2 $\pm$ 1.4 & 84.2 $\pm$ 0.5 & 80.1 $\pm$ 0.8 \\
& DynSparseSGD & 77.1 $\pm$ 0.1 & 74.1 $\pm$ 0.1 & 40.5 $\pm$ 0.6 & 72.3 $\pm$ 1.8 & 80.2 $\pm$ 0.2 & 91.3 $\pm$ 0.2 & 94.4 $\pm$ 0.0 & 84.0 $\pm$ 0.3 & 96.1 $\pm$ 0.2 & 61.4 $\pm$ 0.3 & 79.9 $\pm$ 0.2 & 77.4 $\pm$ 0.4 \\
& DynSignSparseSGD & 75.9 $\pm$ 0.3 & 75.6 $\pm$ 0.3 & 43.9 $\pm$ 2.6 & 82.6 $\pm$ 1.0 & 83.2 $\pm$ 0.6 & 90.8 $\pm$ 0.2 & 94.5 $\pm$ 0.3 & 95.2 $\pm$ 0.7 & 96.8 $\pm$ 0.6 & 68.5 $\pm$ 1.2 & 85.0 $\pm$ 0.6 & 81.1 $\pm$ 0.8 \\
\midrule
\multirow{9}{*}{4} & SignSGD & 24.3 $\pm$ 1.1 & 35.7 $\pm$ 1.1 & 1.5 $\pm$ 0.1 & 15.2 $\pm$ 3.7 & 1.0 $\pm$ 0.2 & 1.1 $\pm$ 0.1 & 2.9 $\pm$ 0.4 & 17.4 $\pm$ 5.3 & 70.9 $\pm$ 4.8 & 32.5 $\pm$ 3.2 & 33.9 $\pm$ 20.2 & 21.5 $\pm$ 3.6 \\
& SignAdam & 0.1 $\pm$ 0.0 & 0.3 $\pm$ 0.0 & 1.0 $\pm$ 0.1 & 10.1 $\pm$ 1.3 & 0.5 $\pm$ 0.1 & 1.0 $\pm$ 0.0 & 2.7 $\pm$ 0.3 & 0.8 $\pm$ 0.1 & 0.7 $\pm$ 0.1 & 2.1 $\pm$ 0.2 & 0.9 $\pm$ 0.1 & 1.8 $\pm$ 0.2 \\
\cmidrule(lr){2-14}
& SparseSGD & 77.2 $\pm$ 0.1 & 74.2 $\pm$ 0.1 & 41.8 $\pm$ 0.7 & 70.4 $\pm$ 1.5 & 80.4 $\pm$ 0.1 & 91.3 $\pm$ 0.1 & 94.6 $\pm$ 0.2 & 84.4 $\pm$ 0.3 & 96.4 $\pm$ 0.1 & 60.8 $\pm$ 0.1 & 80.4 $\pm$ 0.0 & 77.4 $\pm$ 0.3 \\
& SparseAdam & 75.7 $\pm$ 0.3 & 75.8 $\pm$ 0.1 & 44.9 $\pm$ 0.7 & 87.7 $\pm$ 1.9 & 82.3 $\pm$ 0.5 & 91.1 $\pm$ 0.1 & 95.2 $\pm$ 0.2 & 95.5 $\pm$ 0.3 & 97.4 $\pm$ 0.3 & 68.4 $\pm$ 0.4 & 85.2 $\pm$ 0.5 & 81.7 $\pm$ 0.5 \\
\cmidrule(lr){2-14}
& SignSparseAdam & 0.1 $\pm$ 0.0 & 0.3 $\pm$ 0.0 & 1.0 $\pm$ 0.2 & 12.0 $\pm$ 3.0 & 0.5 $\pm$ 0.1 & 1.0 $\pm$ 0.0 & 2.8 $\pm$ 0.0 & 1.4 $\pm$ 0.2 & 0.5 $\pm$ 0.1 & 2.2 $\pm$ 0.0 & 0.8 $\pm$ 0.1 & 2.0 $\pm$ 0.3 \\
& SignSparseSGD & 76.1 $\pm$ 0.2 & 76.2 $\pm$ 0.1 & 47.5 $\pm$ 0.1 & 85.0 $\pm$ 3.0 & 84.7 $\pm$ 0.7 & 90.6 $\pm$ 0.1 & 94.9 $\pm$ 0.0 & 96.5 $\pm$ 0.8 & 97.3 $\pm$ 0.1 & 69.4 $\pm$ 0.2 & 87.4 $\pm$ 1.0 & 82.3 $\pm$ 0.6 \\
\cmidrule(lr){2-14}
& DynSparseAdam & 74.4 $\pm$ 0.2 & 75.6 $\pm$ 0.3 & 48.7 $\pm$ 1.1 & 84.7 $\pm$ 0.6 & 84.0 $\pm$ 0.7 & 89.9 $\pm$ 0.3 & 94.6 $\pm$ 0.4 & 96.0 $\pm$ 1.0 & 97.2 $\pm$ 0.3 & 69.3 $\pm$ 0.5 & 87.4 $\pm$ 0.9 & 82.0 $\pm$ 0.6 \\
& DynSparseSGD & 77.7 $\pm$ 0.1 & 75.4 $\pm$ 0.1 & 43.5 $\pm$ 0.1 & 72.3 $\pm$ 0.5 & 81.8 $\pm$ 0.1 & 91.5 $\pm$ 0.0 & 94.9 $\pm$ 0.2 & 88.0 $\pm$ 0.5 & 96.4 $\pm$ 0.3 & 63.0 $\pm$ 0.2 & 82.0 $\pm$ 0.2 & 78.8 $\pm$ 0.2 \\
& DynSignSparseSGD & 76.6 $\pm$ 0.1 & 77.0 $\pm$ 0.1 & 50.5 $\pm$ 0.6 & 85.4 $\pm$ 2.5 & 86.0 $\pm$ 0.9 & 91.1 $\pm$ 0.2 & 95.0 $\pm$ 0.1 & 97.0 $\pm$ 1.0 & 97.5 $\pm$ 0.3 & 70.8 $\pm$ 0.6 & 87.6 $\pm$ 0.7 & 83.1 $\pm$ 0.6 \\
\midrule
\multirow{9}{*}{8} & SignSGD & 31.6 $\pm$ 0.4 & 42.1 $\pm$ 1.4 & 2.1 $\pm$ 0.1 & 27.7 $\pm$ 6.5 & 1.2 $\pm$ 0.1 & 14.8 $\pm$ 15.0 & 4.9 $\pm$ 2.3 & 28.6 $\pm$ 13.2 & 82.1 $\pm$ 0.9 & 39.1 $\pm$ 2.9 & 60.4 $\pm$ 1.6 & 30.4 $\pm$ 4.0 \\
& SignAdam & 0.1 $\pm$ 0.0 & 0.2 $\pm$ 0.0 & 1.1 $\pm$ 0.0 & 10.9 $\pm$ 0.9 & 0.5 $\pm$ 0.1 & 1.0 $\pm$ 0.0 & 2.8 $\pm$ 0.1 & 1.0 $\pm$ 0.3 & 1.0 $\pm$ 0.2 & 2.3 $\pm$ 0.4 & 0.9 $\pm$ 0.2 & 2.0 $\pm$ 0.2 \\
\cmidrule(lr){2-14}
& SparseSGD & 77.8 $\pm$ 0.1 & 75.7 $\pm$ 0.2 & 45.6 $\pm$ 0.3 & 78.3 $\pm$ 0.3 & 81.8 $\pm$ 0.1 & 91.5 $\pm$ 0.1 & 94.4 $\pm$ 0.3 & 88.9 $\pm$ 0.7 & 96.8 $\pm$ 0.1 & 64.4 $\pm$ 0.7 & 81.5 $\pm$ 0.1 & 79.7 $\pm$ 0.3 \\
& SparseAdam & 76.6 $\pm$ 0.1 & 76.8 $\pm$ 0.2 & 51.4 $\pm$ 0.6 & 90.3 $\pm$ 1.6 & 84.7 $\pm$ 0.4 & 91.5 $\pm$ 0.2 & 95.6 $\pm$ 0.2 & 96.9 $\pm$ 0.2 & 97.6 $\pm$ 0.2 & 72.1 $\pm$ 0.3 & 86.0 $\pm$ 0.3 & 83.6 $\pm$ 0.4 \\
\cmidrule(lr){2-14}
& SignSparseAdam & 0.1 $\pm$ 0.0 & 0.3 $\pm$ 0.0 & 1.4 $\pm$ 0.3 & 14.0 $\pm$ 2.3 & 0.5 $\pm$ 0.0 & 1.0 $\pm$ 0.0 & 2.7 $\pm$ 0.0 & 2.0 $\pm$ 1.1 & 1.0 $\pm$ 0.1 & 2.2 $\pm$ 0.0 & 0.9 $\pm$ 0.2 & 2.4 $\pm$ 0.4 \\
& SignSparseSGD & 76.8 $\pm$ 0.2 & 77.5 $\pm$ 0.0 & 56.0 $\pm$ 1.2 & 89.2 $\pm$ 2.7 & 87.6 $\pm$ 0.3 & 90.7 $\pm$ 0.2 & 95.4 $\pm$ 0.2 & 97.3 $\pm$ 0.3 & 97.5 $\pm$ 0.6 & 72.0 $\pm$ 0.6 & 88.6 $\pm$ 0.5 & 84.4 $\pm$ 0.6 \\
\cmidrule(lr){2-14}
& DynSparseAdam & 74.9 $\pm$ 0.3 & 76.8 $\pm$ 0.2 & 56.4 $\pm$ 1.3 & 88.7 $\pm$ 2.2 & 87.6 $\pm$ 0.3 & 90.0 $\pm$ 0.3 & 94.8 $\pm$ 0.1 & 97.3 $\pm$ 0.1 & 97.0 $\pm$ 0.7 & 71.8 $\pm$ 0.5 & 88.7 $\pm$ 0.5 & 84.0 $\pm$ 0.6 \\
& DynSparseSGD & 78.4 $\pm$ 0.1 & 76.9 $\pm$ 0.1 & 47.6 $\pm$ 0.3 & 80.9 $\pm$ 0.6 & 83.5 $\pm$ 0.3 & 92.0 $\pm$ 0.1 & 94.9 $\pm$ 0.2 & 92.9 $\pm$ 0.2 & 97.2 $\pm$ 0.2 & 68.2 $\pm$ 0.6 & 84.0 $\pm$ 0.2 & 81.5 $\pm$ 0.3 \\
& DynSignSparseSGD & 77.3 $\pm$ 0.1 & 78.3 $\pm$ 0.1 & 58.7 $\pm$ 0.6 & 88.9 $\pm$ 2.7 & 88.8 $\pm$ 0.2 & 91.2 $\pm$ 0.1 & 95.5 $\pm$ 0.3 & 97.8 $\pm$ 0.5 & 97.4 $\pm$ 0.5 & 73.1 $\pm$ 0.4 & 89.2 $\pm$ 0.4 & 85.1 $\pm$ 0.5 \\
\midrule
\multirow{9}{*}{16} & SignSGD & 39.0 $\pm$ 0.3 & 48.6 $\pm$ 1.8 & 3.2 $\pm$ 0.7 & 35.0 $\pm$ 1.4 & 1.9 $\pm$ 0.4 & 4.2 $\pm$ 1.0 & 4.8 $\pm$ 0.4 & 66.7 $\pm$ 7.5 & 86.1 $\pm$ 1.9 & 47.7 $\pm$ 2.7 & 63.4 $\pm$ 2.2 & 36.4 $\pm$ 1.8 \\
& SignAdam & 0.1 $\pm$ 0.0 & 0.3 $\pm$ 0.0 & 1.1 $\pm$ 0.1 & 9.3 $\pm$ 0.1 & 0.5 $\pm$ 0.0 & 1.0 $\pm$ 0.0 & 2.7 $\pm$ 0.2 & 1.0 $\pm$ 0.5 & 0.5 $\pm$ 0.1 & 2.1 $\pm$ 0.4 & 1.0 $\pm$ 0.1 & 1.8 $\pm$ 0.1 \\
\cmidrule(lr){2-14}
& SparseSGD & 78.4 $\pm$ 0.1 & 77.1 $\pm$ 0.1 & 49.0 $\pm$ 0.4 & 83.1 $\pm$ 0.7 & 83.6 $\pm$ 0.1 & 91.8 $\pm$ 0.1 & 95.0 $\pm$ 0.2 & 92.6 $\pm$ 0.4 & 97.4 $\pm$ 0.2 & 68.5 $\pm$ 0.5 & 83.9 $\pm$ 0.1 & 81.9 $\pm$ 0.3 \\
& SparseAdam & 77.6 $\pm$ 0.2 & 77.7 $\pm$ 0.2 & 59.0 $\pm$ 0.4 & 94.1 $\pm$ 0.8 & 86.5 $\pm$ 0.2 & 91.5 $\pm$ 0.2 & 95.8 $\pm$ 0.1 & 97.9 $\pm$ 0.3 & 97.8 $\pm$ 0.2 & 74.9 $\pm$ 0.2 & 87.8 $\pm$ 0.2 & 85.5 $\pm$ 0.3 \\
\cmidrule(lr){2-14}
& SignSparseAdam & 0.1 $\pm$ 0.0 & 0.2 $\pm$ 0.0 & 1.0 $\pm$ 0.0 & 19.7 $\pm$ 1.6 & 0.5 $\pm$ 0.0 & 1.0 $\pm$ 0.0 & 2.7 $\pm$ 0.0 & 1.6 $\pm$ 0.5 & 2.2 $\pm$ 2.2 & 2.9 $\pm$ 0.6 & 1.0 $\pm$ 0.1 & 3.0 $\pm$ 0.5 \\
& SignSparseSGD & 78.0 $\pm$ 0.1 & 79.3 $\pm$ 0.3 & 61.6 $\pm$ 0.9 & 92.7 $\pm$ 1.5 & 89.5 $\pm$ 0.3 & 90.8 $\pm$ 0.2 & 95.4 $\pm$ 0.3 & 98.5 $\pm$ 0.1 & 97.7 $\pm$ 0.0 & 76.5 $\pm$ 0.8 & 90.1 $\pm$ 0.2 & 86.4 $\pm$ 0.4 \\
\cmidrule(lr){2-14}
& DynSparseAdam & 76.1 $\pm$ 0.1 & 78.5 $\pm$ 0.2 & 63.6 $\pm$ 0.4 & 92.8 $\pm$ 0.6 & 89.3 $\pm$ 0.5 & 90.1 $\pm$ 0.1 & 95.1 $\pm$ 0.2 & 98.2 $\pm$ 0.5 & 97.2 $\pm$ 0.3 & 76.3 $\pm$ 0.2 & 90.0 $\pm$ 0.5 & 86.1 $\pm$ 0.3 \\
& DynSparseSGD & 79.2 $\pm$ 0.0 & 78.6 $\pm$ 0.0 & 52.6 $\pm$ 0.1 & 85.6 $\pm$ 0.9 & 85.5 $\pm$ 0.1 & 92.1 $\pm$ 0.1 & 95.3 $\pm$ 0.2 & 96.9 $\pm$ 0.2 & 97.3 $\pm$ 0.1 & 72.7 $\pm$ 0.3 & 86.2 $\pm$ 0.2 & 83.8 $\pm$ 0.2 \\
& DynSignSparseSGD & 78.3 $\pm$ 0.1 & 79.6 $\pm$ 0.1 & 65.0 $\pm$ 0.3 & 93.4 $\pm$ 0.9 & 90.3 $\pm$ 0.1 & 91.2 $\pm$ 0.0 & 95.6 $\pm$ 0.1 & 98.9 $\pm$ 0.2 & 97.4 $\pm$ 0.3 & 77.8 $\pm$ 0.7 & 91.0 $\pm$ 0.2 & 87.1 $\pm$ 0.3 \\
\bottomrule
\end{tabular}
\end{adjustbox}
\end{table}

Tables~\ref{tab:ablation_optim_ViTB32_with_aug}, \ref{tab:ablation_optim_ViTB16_with_aug}, and \ref{tab:ablation_optim_ViTL14_with_aug} complement Fig.~\ref{fig:ablation} by reporting per-dataset accuracies for dense sign-based optimizers (SignSGD, SignAdam), sparse optimizers (SparseSGD, SparseAdam/SAFT), sign-sparse optimizers (SignSparseSGD, SignSparse Adam), and dynamic sparse variants (DynSparseSGD, DynSparseAdam, DynSignSparseSGD) on three CLIP backbones. In this subsection, all methods are compared under the same training schedule and without gradient normalization. The tables therefore reflect the raw behavior of each update rule under a shared learning-rate budget. Under this setting, DynSignSparseSGD achieves the best average accuracy at every shot level on ViT-B/32, ViT-B/16, and ViT-L/14, while remaining sparse and state-free. At 16-shot, it reaches $79.0\%$, $82.8\%$, and $87.1\%$ on the three backbones, respectively. Relative to SignSparseSGD, the average gain is about $+1.6$, $+1.2$, and $+0.7$ points. Relative to DynSparseAdam, the gain is about $+1.6$, $+1.1$, and $+1.0$ points.

A first observation concerns the dense sign baselines. In this unnormalized regime, SignSGD, SignAdam, and SignSparseAdam remain below the corresponding zero-shot CLIP baseline for every reported backbone--shot--dataset configuration. In subsection~\ref{with_gradient_normlization}, we show that this collapse in performance is due to the update scaling. This point is important for the next subsection, where we revisit SignSGD and SparseSGD with gradient normalization.

%The sparse baselines are much stronger.
SparseSGD is above zero-shot CLIP in every reported case, and SparseAdam, which is equivalent to SAFT in our implementation, is also competitive on average. SignSparseSGD further improves over SparseSGD in the average columns on all three backbones and at all shot levels, which suggests that sign updates become effective once they are restricted to a small set of informative coordinates. The remaining gap between these sparse baselines and DynSignSparseSGD should still be read here as an unnormalized comparison, not as a definitive statement about SparseSGD under a different update scaling.

Within the sparse family, the main effect comes from combining sign updates with dynamic support selection. DynSignSparseSGD matches or improves SignSparseSGD in 156 of the 165 backbone--shot--dataset comparisons, and it also matches or improves DynSparseAdam in 156 cases. The gains are especially visible on difficult datasets such as Aircraft, DTD, Flowers, Cars, and UCF. For example, at 16-shot on ViT-B/32, DynSignSparseSGD reaches $42.0$ on Aircraft and $69.2$ on DTD, compared with $36.4$ and $67.5$ for SignSparseSGD and $41.8$ and $67.1$ for DynSparseAdam. On ViT-B/16, the corresponding numbers are $51.7$ and $72.2$. On ViT-L/14, they are $65.0$ and $77.8$. There are a few isolated reversals, for example on ImageNet or EuroSAT in some settings, but the average columns show a stable overall ordering.

Overall, these tables identify the strongest unnormalized optimizer in this comparison. The best performance comes from combining sign updates, sparsity, and dynamic support selection. Appendix~\ref{sec:update_scale} shows that the update magnitude is important. Appendix~\ref{app:sparse_sign_normalized} addresses this problem with gradient normalization.

\subsection{With Gradient Normalization}
\label{app:sparse_sign_normalized}
\label{with_gradient_normlization}
\suppanchor{optimizer-normalized}

To isolate the role of the learning scale, we revisit dense SignSGD and SparseSGD with the norm-matched variants defined in Appendix~\ref{app:algorithms}. Appendix~\ref{sec:update_scale} (Learning Scale) shows that, under a shared learning rate, the raw SignSGD and SparseSGD updates are not matched in magnitude to the corresponding dense SGD step. We therefore compare GradNorm SignSGD, GradNorm SparseSGD, SignSparseSGD, and DynSignSparseSGD on 1-shot CLIP (ViT-B/16) across all 11 datasets and four learning rates. Unlike the fixed 30-epoch budget used in Sec.~\ref{sec:experiments} and Appendix~\ref{app:sparse_sign_raw}, these runs are extended to 300 epochs in order to visualize convergence speed and late-epoch accuracy degradation directly. The most informative regime is $\eta=10^{-4}$, which is also the learning rate used in the main optimizer ablation. At $\eta=10^{-2}$ all methods operate outside the useful regime, and at $\eta=10^{-5}$ all methods train slowly.

\begin{figure}[!h]
  \centering
  \includegraphics[width=0.94\textwidth,height=0.80\textheight,keepaspectratio]{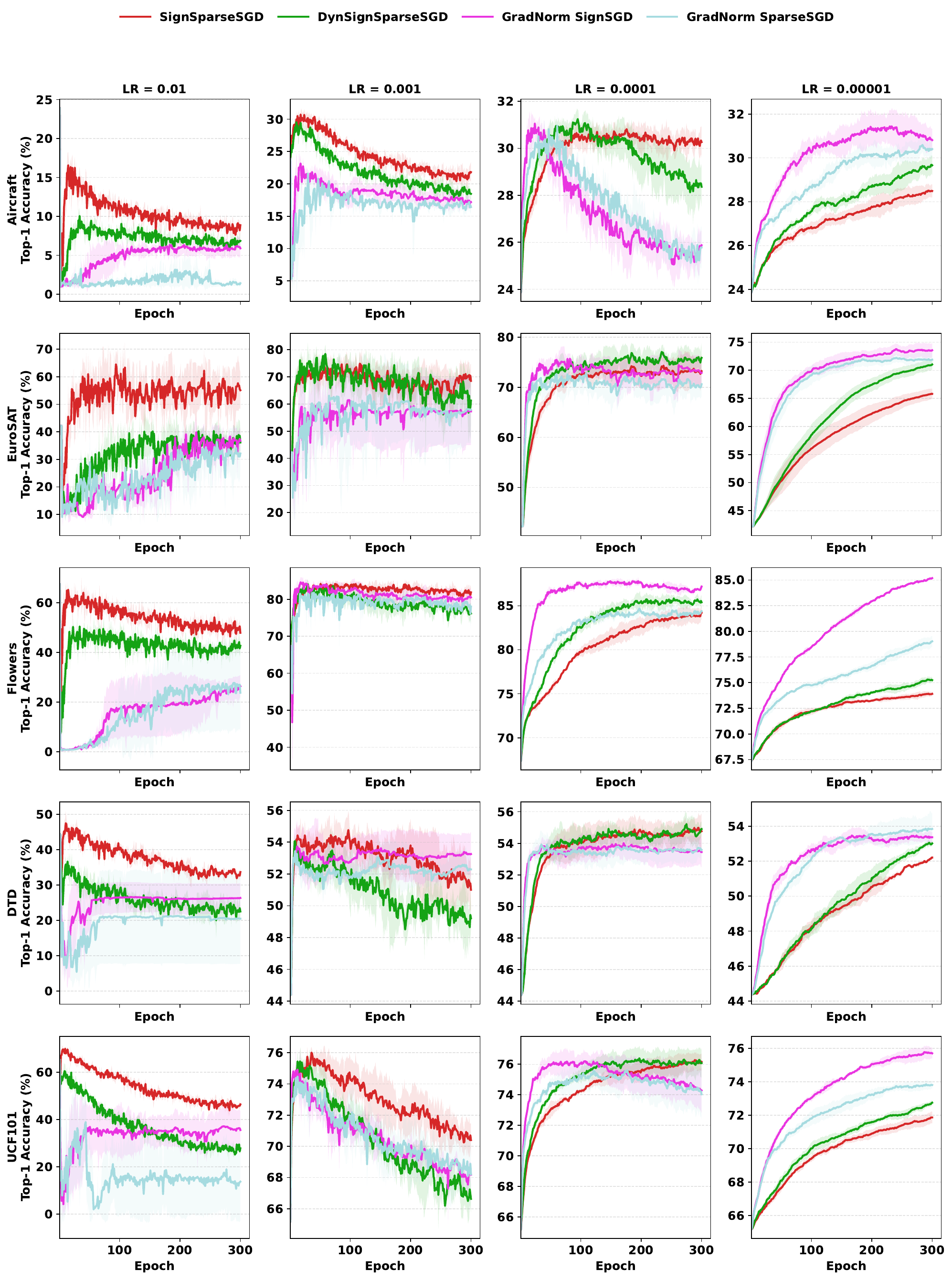}
  \caption{Test accuracy (mean $\pm$ std over 3 seeds) over epochs for SignSparseSGD, DynSignSparseSGD, GradNorm SignSGD, and GradNorm SparseSGD in a 1-shot setting on 5 datasets (Aircraft, EuroSAT, Flowers, DTD, and UCF) using CLIP (ViT-B/16). Each column corresponds to one learning rate: $10^{-2}$, $10^{-3}$, $10^{-4}$, and $10^{-5}$.}
  \label{fig:gradnorm_comparison_group1_with_aug}
\end{figure}

\begin{figure}[!h]
  \centering
  \includegraphics[width=0.94\textwidth,height=0.80\textheight,keepaspectratio]{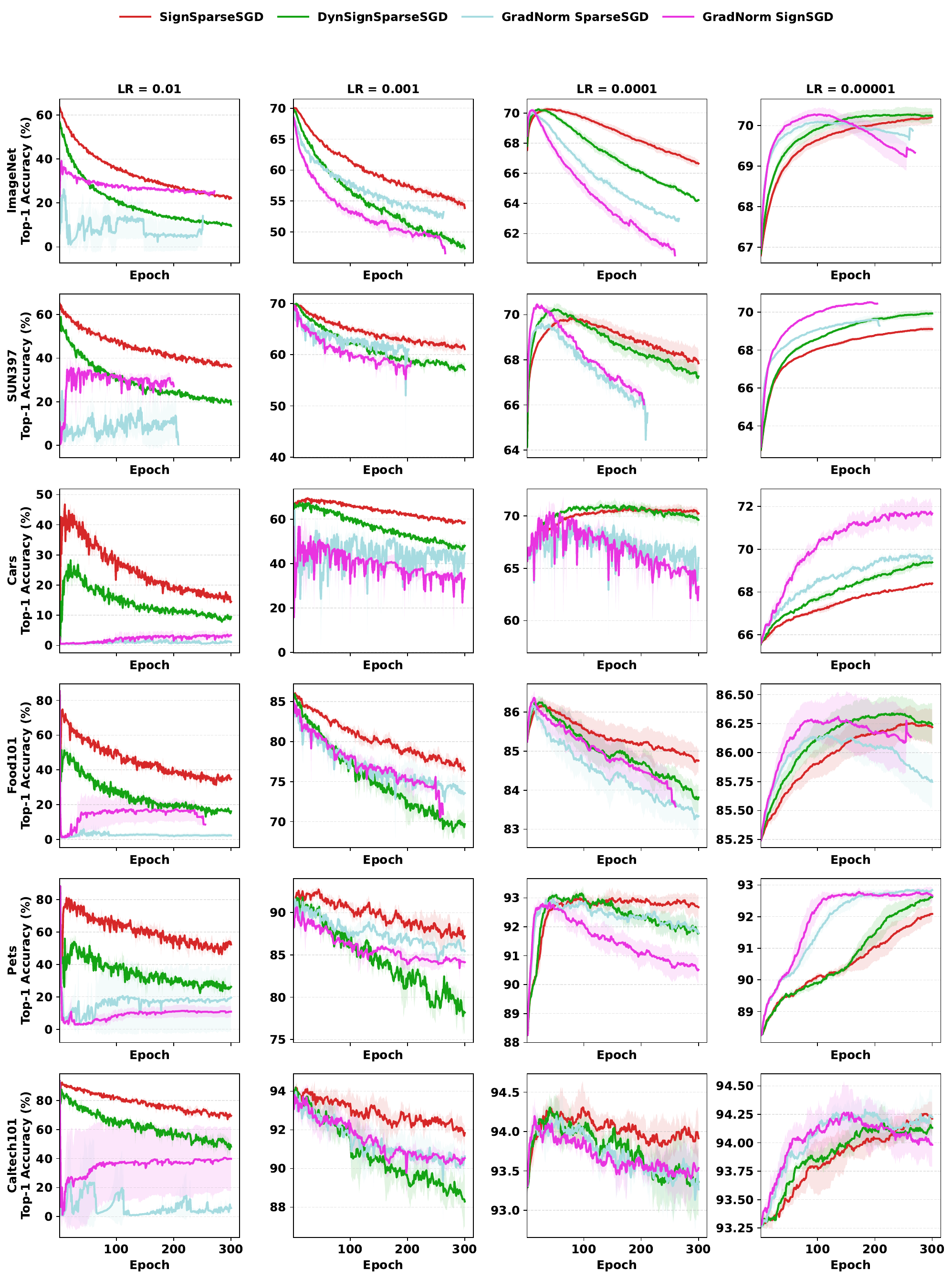}
  \caption{Test accuracy (mean $\pm$ std over 3 seeds) over epochs for SignSparseSGD, DynSignSparseSGD, GradNorm SignSGD, and GradNorm SparseSGD in a 1-shot setting on 6 datasets (ImageNet, SUN, Cars, Food101, Pets, and Caltech) using CLIP (ViT-B/16). Each column corresponds to one learning rate: $10^{-2}$, $10^{-3}$, $10^{-4}$, and $10^{-5}$.}
  \label{fig:gradnorm_comparison_group2_with_aug}
\end{figure}

A first observation is that gradient normalization fixes the collapse of dense SignSGD. This is exactly the point made in Appendix~\ref{sec:update_scale}. After norm matching, the dense sign baseline no longer fails immediately. At $\eta=10^{-4}$, GradNorm SignSGD reaches $72.5\%$ average accuracy at epoch 30, compared with only $6.1\%$ for SignSGD. Across the 44 dataset--learning-rate settings covered by Fig.~\ref{fig:gradnorm_comparison_group1_with_aug} and Fig.~\ref{fig:gradnorm_comparison_group2_with_aug}, GradNorm SignSGD achieves a higher final accuracy than SignSGD in all 44 cases. The comparison between SignSGD and GradNorm SignSGD should therefore be read together with Appendix~\ref{sec:update_scale}: once the update norm is corrected, dense sign directions become meaningful baselines, even though they remain less robust than the sign-sparse variants.

A second observation is that GradNorm SparseSGD is not as slow as SparseSGD. This is again consistent with Appendix~\ref{sec:update_scale}. At $\eta=10^{-4}$, GradNorm SparseSGD reaches $71.5\%$ average accuracy by epoch 30, slightly above SparseSGD at $70.4\%$. Its average peak is also reached substantially earlier, at about 74 epochs instead of about 120 for SparseSGD. The final accuracies remain close, with $70.1\%$ for GradNorm SparseSGD and $71.3\%$ for SparseSGD, but the norm-matched sparse baseline reaches its useful regime faster. This is the fairer comparison when the goal is to separate sparsity from the scale of the update.

The main conclusion, however, does not change. Even after correcting the learning scale, the sign-sparse optimizers remain stronger on the accuracy--stability trade-off. At $\eta=10^{-4}$, DynSignSparseSGD attains the highest average peak accuracy across the 11 datasets at $73.7\%$. SignSparseSGD attains the highest final average accuracy at $72.3\%$ and the smallest average peak-to-end drop at $0.93$ points. The corresponding final averages for GradNorm SignSGD and GradNorm SparseSGD are $70.1\%$ and $70.1\%$, with larger peak-to-end drops of $3.47$ and $2.64$ points. From epoch 30 to the end of training at $\eta=10^{-4}$, SignSparseSGD gains $+2.2$ points on average and DynSignSparseSGD gains $+0.9$, whereas GradNorm SignSGD and GradNorm SparseSGD lose $2.4$ and $1.3$ points. At the dataset level, the two sign-sparse methods achieve the best final accuracy on 10 of the 11 datasets and the best peak accuracy on 8 of them. SignSparseSGD also has the smallest late-epoch drop on all 11 datasets.

The gains are especially visible on Aircraft, EuroSAT, Cars, ImageNet, and UCF101 at $\eta=10^{-4}$. On Aircraft, SignSparseSGD and DynSignSparseSGD reach $30.3$ and $28.4$, compared with $25.9$ and $25.8$ for GradNorm SignSGD and GradNorm SparseSGD. On Cars, the corresponding numbers are $70.2$ and $69.6$, compared with $63.2$ and $66.0$. On ImageNet, they are $66.6$ and $64.2$, compared with $60.5$ and $63.0$. EuroSAT and UCF101 show the same trend. There are a few isolated reversals (e.g., Flowers), but they do not change the overall ordering.

Overall, gradient normalization makes SignSGD and SparseSGD fairer baselines, which is important for the comparison. However, it does not remove the advantage of combining sign updates with sparsity. The strongest curves still come from the sign-sparse family. DynSignSparseSGD gives the highest peak accuracy, and SignSparseSGD gives the most stable long-horizon behavior.

\subsection{Matched Momentum Control}
\label{app:matched_momentum}
\suppanchor{optimizer-momentum}

This experiment compares instantaneous-gradient and first-moment signs with a common static mask and coordinate-step magnitude. Both variants adapt CLIP (ViT-B/16) on ImageNet and SUN397 with one labeled example per class. Each dataset is evaluated with seeds 1, 2, and 3 at both $\lambda=10^{-5}$ and $\lambda=10^{-4}$. All runs use 30 epochs, batch size 64, nominal density $\rho=0.001$, and the crop/flip augmentation pipeline. The resulting comparison contains 24 training runs.

\paragraph{Matched updates.}
Both variants use the SignSparseSGD parameter scope in the image and text encoders. Within each paired run, they share the pretrained parameters, eligible tensors, and initial tensor-wise top-magnitude mask $m$. The mask remains fixed throughout training. On each successful update, the instantaneous variant applies $-\lambda m\odot\sgn(g_t)$. The momentum variant forms the first moment $u_{t+1}=\beta_1u_t+(1-\beta_1)(m\odot g_t)$ from $u_0=0$ and applies $-\lambda m\odot\sgn(u_{t+1})$. Both variants use $\lambda$ directly as the coordinate coefficient. These values correspond to $\eta=10^{-4}$ and $10^{-3}$ in SignSparseSGD with $\alpha=0.1$. Neither variant uses second-moment normalization, weight decay, an amplitude-changing bias correction, a learning-rate schedule, or warmup.

The pair receives the same materialized image and label stream, including identical minibatch order and augmentation draws. Both models use FP32 stored weights with automatic mixed precision and gradient scaling. A shared skip decision omits both updates whenever either model has a nonfinite gradient. Thus, successful-step counts agree within each pair. The comparison follows this coupled protocol rather than two independently skipped training runs. The final checkpoint is always the fixed epoch-30 checkpoint. Test accuracy does not select checkpoints or stopping times. Evaluation uses the original CLIP top-1 routine with unscaled cosine scores.

\begin{table}[t]
\centering
\small
\setlength{\tabcolsep}{6pt}
\begin{tabular}{llrrr}
\toprule
Dataset & Coordinate step $\lambda$ & Instantaneous sign & Momentum sign & Difference (pp) \\
\midrule
ImageNet & $10^{-5}$ & $70.131\pm0.186$ & $69.089\pm0.334$ & $+1.041\pm0.148$ \\
ImageNet & $10^{-4}$ & $66.568\pm0.598$ & $59.756\pm0.447$ & $+6.812\pm0.184$ \\
SUN397 & $10^{-5}$ & $69.214\pm0.223$ & $69.634\pm0.377$ & $-0.420\pm0.204$ \\
SUN397 & $10^{-4}$ & $67.924\pm0.053$ & $63.345\pm0.782$ & $+4.579\pm0.777$ \\
\bottomrule
\end{tabular}
\caption{Matched momentum control for one-shot CLIP (ViT-B/16) at epoch 30. Accuracy is reported in percent as mean $\pm$ sample standard deviation over three seeds. The difference is instantaneous-sign minus momentum-sign accuracy in percentage points; its deviation is computed from the three within-seed differences. Differences are computed before rounding. The two methods share the static mask, input stream, coordinate step, and successful-step budget.}
\label{tab:matched_momentum}
\end{table}

\paragraph{Results.}
Table~\ref{tab:matched_momentum} shows higher mean accuracy for instantaneous signs in three of the four dataset--step settings. On ImageNet, the gains are 1.041 and 6.812 percentage points at $\lambda=10^{-5}$ and $10^{-4}$, respectively. On SUN397, momentum improves accuracy by 0.420 points at $\lambda=10^{-5}$, whereas instantaneous signs improve accuracy by 4.579 points at $\lambda=10^{-4}$. These results show that the effect of momentum depends on the dataset and coordinate step. They do not establish a uniform disadvantage of momentum. The comparison is separate from the unnormalized optimizer results in Appendix~\ref{app:sparse_sign_raw}. Its conclusions concern the two datasets, one-shot ViT-B/16, and the fixed 30-epoch budget; no matched 300-epoch extension is included.

\section{Sign vs Gradient Clipping}
\label{sec:sign_vs_clipping}
\label{sec:sign_vs_gradient_clipping}
\suppanchor{clipping}

\begin{table}[!h]
\caption{Few-shot classification performance (shots 1, 2, 4, 8, 16) on 11 datasets with CLIP(ViT-B/32) backbone. We compare SignSparseSGD and DynSignSparseSGD against two stabilised variants: Local-clip SparseSGD and Global-clip SparseSGD. Top-1 accuracy (\%) averaged over 3 seeds.}
\label{tab:ablation_signsparsesgd_clip_ViTB32_with_aug}
\centering
\footnotesize
\setlength{\tabcolsep}{1.8pt}
\renewcommand{\arraystretch}{0.90}
\begin{adjustbox}{max totalsize={\textwidth}{0.95\textheight},center}
\begin{tabular}{@{}c l *{12}{d}@{}}
\toprule
Shots & Method & ImageNet & SUN & Aircr. & EuroSAT & Cars & Food & Pets & Flow. & Caltech & DTD & UCF & Avg. \\ \midrule
\multirow{1}{*}{0} & CLIP & 62.0 & 62.1 & 18.1 & 35.8 & 60.1 & 78.9 & 85.1 & 64.0 & 91.4 & 43.0 & 61.6 & 60.2 \\
\midrule
\multirow{4}{*}{1} 
& Local-clip SparseSGD & 64.7 $\pm$ 0.2 & 66.0 $\pm$ 0.1 & 21.1 $\pm$ 0.2 & 52.0 $\pm$ 0.6 & 62.1 $\pm$ 0.1 & 80.8 $\pm$ 0.3 & 87.7 $\pm$ 0.2 & 67.9 $\pm$ 0.7 & 92.5 $\pm$ 0.0 & 48.7 $\pm$ 0.3 & 66.2 $\pm$ 0.3 & 64.5 $\pm$ 0.3 \\
& Global-clip SparseSGD & 62.1 $\pm$ 0.0 & 62.1 $\pm$ 0.0 & 18.2 $\pm$ 0.0 & 35.9 $\pm$ 0.0 & 60.2 $\pm$ 0.0 & 78.9 $\pm$ 0.0 & 85.1 $\pm$ 0.0 & 64.1 $\pm$ 0.0 & 91.4 $\pm$ 0.0 & 42.8 $\pm$ 0.0 & 61.7 $\pm$ 0.0 & 60.2 $\pm$ 0.0 \\
\cmidrule(lr){2-14}
& SignSparseSGD & 60.6 $\pm$ 0.6 & 65.3 $\pm$ 0.1 & 21.6 $\pm$ 0.5 & 66.0 $\pm$ 5.2 & 61.3 $\pm$ 1.1 & 77.9 $\pm$ 0.5 & 88.5 $\pm$ 0.4 & 75.0 $\pm$ 0.8 & 92.3 $\pm$ 0.5 & 50.9 $\pm$ 1.5 & 71.5 $\pm$ 0.3 & 66.4 $\pm$ 1.0 \\
& DynSignSparseSGD & 62.6 $\pm$ 0.3 & 67.0 $\pm$ 0.3 & 22.7 $\pm$ 0.8 & 69.1 $\pm$ 2.7 & 63.1 $\pm$ 0.4 & 79.7 $\pm$ 0.5 & 89.1 $\pm$ 0.2 & 77.7 $\pm$ 1.2 & 92.6 $\pm$ 0.7 & 51.9 $\pm$ 1.0 & 72.0 $\pm$ 0.4 & 67.9 $\pm$ 0.8 \\
\midrule
\multirow{4}{*}{2} 
& Local-clip SparseSGD & 65.2 $\pm$ 0.1 & 67.1 $\pm$ 0.1 & 21.5 $\pm$ 0.4 & 58.9 $\pm$ 4.0 & 62.8 $\pm$ 0.1 & 81.5 $\pm$ 0.0 & 89.3 $\pm$ 0.6 & 69.5 $\pm$ 0.6 & 92.9 $\pm$ 0.1 & 51.7 $\pm$ 0.3 & 67.4 $\pm$ 0.2 & 66.2 $\pm$ 0.6 \\
& Global-clip SparseSGD & 62.2 $\pm$ 0.0 & 62.2 $\pm$ 0.0 & 18.2 $\pm$ 0.1 & 35.9 $\pm$ 0.0 & 60.3 $\pm$ 0.0 & 78.9 $\pm$ 0.0 & 85.1 $\pm$ 0.0 & 64.1 $\pm$ 0.0 & 91.4 $\pm$ 0.0 & 42.9 $\pm$ 0.0 & 61.7 $\pm$ 0.1 & 60.3 $\pm$ 0.0 \\
\cmidrule(lr){2-14}
& SignSparseSGD & 61.9 $\pm$ 0.3 & 66.9 $\pm$ 0.2 & 24.3 $\pm$ 0.6 & 79.1 $\pm$ 6.7 & 63.6 $\pm$ 0.1 & 78.0 $\pm$ 0.4 & 89.1 $\pm$ 0.4 & 78.7 $\pm$ 0.2 & 93.8 $\pm$ 0.3 & 56.4 $\pm$ 0.4 & 74.1 $\pm$ 0.8 & 69.6 $\pm$ 0.9 \\
& DynSignSparseSGD & 63.2 $\pm$ 0.1 & 68.2 $\pm$ 0.3 & 25.5 $\pm$ 0.6 & 79.8 $\pm$ 4.2 & 66.0 $\pm$ 0.1 & 79.5 $\pm$ 0.4 & 89.2 $\pm$ 0.3 & 83.8 $\pm$ 1.4 & 94.0 $\pm$ 0.3 & 57.1 $\pm$ 1.1 & 74.8 $\pm$ 0.6 & 71.0 $\pm$ 0.9 \\
\midrule
\multirow{4}{*}{4} 
& Local-clip SparseSGD & 65.8 $\pm$ 0.1 & 68.2 $\pm$ 0.1 & 22.8 $\pm$ 0.4 & 59.5 $\pm$ 0.3 & 63.8 $\pm$ 0.3 & 81.8 $\pm$ 0.2 & 90.3 $\pm$ 0.2 & 72.8 $\pm$ 0.4 & 93.6 $\pm$ 0.1 & 52.1 $\pm$ 0.4 & 68.7 $\pm$ 0.5 & 67.2 $\pm$ 0.2 \\
& Global-clip SparseSGD & 62.4 $\pm$ 0.0 & 62.5 $\pm$ 0.1 & 18.2 $\pm$ 0.0 & 35.9 $\pm$ 0.0 & 60.3 $\pm$ 0.0 & 78.9 $\pm$ 0.0 & 85.1 $\pm$ 0.0 & 64.3 $\pm$ 0.0 & 91.4 $\pm$ 0.0 & 42.9 $\pm$ 0.1 & 61.8 $\pm$ 0.0 & 60.3 $\pm$ 0.0 \\
\cmidrule(lr){2-14}
& SignSparseSGD & 63.3 $\pm$ 0.4 & 69.0 $\pm$ 0.4 & 25.2 $\pm$ 1.1 & 82.6 $\pm$ 1.0 & 66.9 $\pm$ 0.3 & 78.2 $\pm$ 0.4 & 89.3 $\pm$ 0.4 & 85.8 $\pm$ 1.2 & 94.7 $\pm$ 0.4 & 59.1 $\pm$ 1.0 & 75.9 $\pm$ 0.7 & 71.8 $\pm$ 0.7 \\
& DynSignSparseSGD & 64.0 $\pm$ 0.2 & 69.9 $\pm$ 0.3 & 27.7 $\pm$ 0.6 & 83.9 $\pm$ 1.6 & 69.3 $\pm$ 0.1 & 79.6 $\pm$ 0.4 & 90.0 $\pm$ 0.4 & 88.9 $\pm$ 0.8 & 95.2 $\pm$ 0.2 & 61.1 $\pm$ 0.3 & 78.0 $\pm$ 0.9 & 73.4 $\pm$ 0.5 \\
\midrule
\multirow{4}{*}{8} 
& Local-clip SparseSGD & 66.3 $\pm$ 0.1 & 69.6 $\pm$ 0.1 & 25.1 $\pm$ 0.5 & 67.6 $\pm$ 2.0 & 65.3 $\pm$ 0.2 & 82.1 $\pm$ 0.1 & 91.1 $\pm$ 0.2 & 76.6 $\pm$ 0.4 & 94.1 $\pm$ 0.1 & 55.4 $\pm$ 0.7 & 71.2 $\pm$ 0.1 & 69.5 $\pm$ 0.4 \\
& Global-clip SparseSGD & 62.7 $\pm$ 0.0 & 62.8 $\pm$ 0.1 & 18.3 $\pm$ 0.1 & 36.0 $\pm$ 0.0 & 60.3 $\pm$ 0.0 & 79.0 $\pm$ 0.0 & 85.2 $\pm$ 0.0 & 64.4 $\pm$ 0.0 & 91.5 $\pm$ 0.0 & 42.9 $\pm$ 0.0 & 61.9 $\pm$ 0.0 & 60.5 $\pm$ 0.0 \\
\cmidrule(lr){2-14}
& SignSparseSGD & 64.7 $\pm$ 0.1 & 70.9 $\pm$ 0.3 & 31.1 $\pm$ 0.8 & 85.9 $\pm$ 0.5 & 71.4 $\pm$ 0.5 & 79.0 $\pm$ 0.2 & 90.3 $\pm$ 0.1 & 89.3 $\pm$ 0.7 & 94.9 $\pm$ 0.3 & 62.3 $\pm$ 0.9 & 78.8 $\pm$ 0.7 & 74.4 $\pm$ 0.5 \\
& DynSignSparseSGD & 65.0 $\pm$ 0.2 & 71.9 $\pm$ 0.1 & 34.8 $\pm$ 0.6 & 88.4 $\pm$ 1.0 & 75.1 $\pm$ 0.2 & 79.8 $\pm$ 0.1 & 91.0 $\pm$ 0.3 & 92.5 $\pm$ 0.1 & 95.5 $\pm$ 0.3 & 64.5 $\pm$ 1.0 & 79.7 $\pm$ 0.6 & 76.2 $\pm$ 0.4 \\
\midrule
\multirow{4}{*}{16} 
& Local-clip SparseSGD & 67.2 $\pm$ 0.1 & 71.0 $\pm$ 0.1 & 27.0 $\pm$ 0.3 & 74.4 $\pm$ 0.7 & 67.3 $\pm$ 0.3 & 82.4 $\pm$ 0.1 & 91.7 $\pm$ 0.2 & 82.7 $\pm$ 0.7 & 94.7 $\pm$ 0.1 & 60.8 $\pm$ 0.2 & 74.6 $\pm$ 0.2 & 72.2 $\pm$ 0.3 \\
& Global-clip SparseSGD & 63.2 $\pm$ 0.1 & 63.2 $\pm$ 0.0 & 18.6 $\pm$ 0.0 & 36.3 $\pm$ 0.0 & 60.3 $\pm$ 0.0 & 79.1 $\pm$ 0.0 & 85.4 $\pm$ 0.0 & 64.4 $\pm$ 0.1 & 91.5 $\pm$ 0.0 & 43.2 $\pm$ 0.1 & 61.9 $\pm$ 0.0 & 60.6 $\pm$ 0.0 \\
\cmidrule(lr){2-14}
& SignSparseSGD & 66.3 $\pm$ 0.2 & 73.1 $\pm$ 0.3 & 36.4 $\pm$ 0.9 & 90.7 $\pm$ 1.6 & 76.4 $\pm$ 0.6 & 79.3 $\pm$ 0.5 & 89.9 $\pm$ 1.0 & 93.8 $\pm$ 1.0 & 95.8 $\pm$ 0.0 & 67.5 $\pm$ 0.2 & 81.7 $\pm$ 0.4 & 77.4 $\pm$ 0.6 \\
& DynSignSparseSGD & 66.9 $\pm$ 0.2 & 73.5 $\pm$ 0.1 & 42.0 $\pm$ 1.1 & 92.0 $\pm$ 0.6 & 79.8 $\pm$ 0.4 & 80.3 $\pm$ 0.4 & 91.0 $\pm$ 0.3 & 95.7 $\pm$ 1.0 & 95.8 $\pm$ 0.0 & 69.2 $\pm$ 0.5 & 82.7 $\pm$ 0.5 & 79.0 $\pm$ 0.5 \\
\bottomrule
\end{tabular}
\end{adjustbox}
\end{table}

\begin{table}[!h]
\caption{Few-shot classification performance (shots 1, 2, 4, 8, 16) on 11 datasets with CLIP(ViT-B/16) backbone. We compare SignSparseSGD and DynSignSparseSGD against two stabilised variants: Local-clip SparseSGD and Global-clip SparseSGD. Top-1 accuracy (\%) averaged over 3 seeds.}
\label{tab:ablation_signsparsesgd_clip_ViTB16_with_aug}
\centering
\footnotesize
\setlength{\tabcolsep}{1.8pt}
\renewcommand{\arraystretch}{0.90}
\begin{adjustbox}{max totalsize={\textwidth}{0.95\textheight},center}
\begin{tabular}{@{}c l *{12}{d}@{}}
\toprule
Shots & Method & ImageNet & SUN & Aircr. & EuroSAT & Cars & Food & Pets & Flow. & Caltech & DTD & UCF & Avg. \\ \midrule
\multirow{1}{*}{0} & CLIP & 66.7 & 62.6 & 23.9 & 42.2 & 65.5 & 85.2 & 88.3 & 67.4 & 93.3 & 44.4 & 65.2 & 64.1 \\
\midrule
\multirow{4}{*}{1} 
& Local-clip SparseSGD & 69.9 $\pm$ 0.1 & 68.5 $\pm$ 0.1 & 27.4 $\pm$ 0.3 & 63.4 $\pm$ 1.9 & 67.7 $\pm$ 0.1 & 86.2 $\pm$ 0.1 & 90.4 $\pm$ 0.1 & 73.6 $\pm$ 0.3 & 94.1 $\pm$ 0.1 & 50.6 $\pm$ 0.6 & 70.6 $\pm$ 0.2 & 69.3 $\pm$ 0.3 \\
& Global-clip SparseSGD & 66.8 $\pm$ 0.0 & 62.7 $\pm$ 0.0 & 24.0 $\pm$ 0.0 & 42.3 $\pm$ 0.0 & 65.6 $\pm$ 0.0 & 85.3 $\pm$ 0.0 & 88.3 $\pm$ 0.0 & 67.5 $\pm$ 0.0 & 93.3 $\pm$ 0.0 & 44.4 $\pm$ 0.0 & 65.2 $\pm$ 0.0 & 64.1 $\pm$ 0.0 \\
\cmidrule(lr){2-14}
& SignSparseSGD & 66.8 $\pm$ 0.1 & 68.2 $\pm$ 0.3 & 29.4 $\pm$ 1.3 & 71.2 $\pm$ 3.7 & 69.1 $\pm$ 0.6 & 83.8 $\pm$ 0.5 & 91.3 $\pm$ 0.3 & 83.3 $\pm$ 0.5 & 93.7 $\pm$ 0.2 & 54.3 $\pm$ 0.6 & 75.7 $\pm$ 0.3 & 71.5 $\pm$ 0.8 \\
& DynSignSparseSGD & 68.1 $\pm$ 0.1 & 69.8 $\pm$ 0.3 & 30.6 $\pm$ 1.2 & 73.7 $\pm$ 4.0 & 70.6 $\pm$ 0.4 & 85.1 $\pm$ 0.3 & 92.5 $\pm$ 0.2 & 84.7 $\pm$ 0.4 & 93.8 $\pm$ 0.1 & 54.8 $\pm$ 0.6 & 76.2 $\pm$ 0.4 & 72.7 $\pm$ 0.7 \\
\midrule
\multirow{4}{*}{2} 
& Local-clip SparseSGD & 70.4 $\pm$ 0.1 & 69.4 $\pm$ 0.0 & 29.2 $\pm$ 0.4 & 67.4 $\pm$ 0.9 & 68.6 $\pm$ 0.2 & 86.6 $\pm$ 0.2 & 91.7 $\pm$ 0.7 & 74.0 $\pm$ 0.3 & 94.1 $\pm$ 0.0 & 53.0 $\pm$ 1.0 & 72.1 $\pm$ 0.2 & 70.6 $\pm$ 0.4 \\
& Global-clip SparseSGD & 66.8 $\pm$ 0.0 & 62.8 $\pm$ 0.0 & 24.0 $\pm$ 0.0 & 42.3 $\pm$ 0.0 & 65.6 $\pm$ 0.0 & 85.3 $\pm$ 0.0 & 88.3 $\pm$ 0.0 & 67.5 $\pm$ 0.0 & 93.3 $\pm$ 0.0 & 44.4 $\pm$ 0.0 & 65.2 $\pm$ 0.0 & 64.1 $\pm$ 0.0 \\
\cmidrule(lr){2-14}
& SignSparseSGD & 67.4 $\pm$ 0.4 & 69.7 $\pm$ 0.1 & 32.0 $\pm$ 1.4 & 79.5 $\pm$ 2.2 & 71.7 $\pm$ 0.2 & 84.1 $\pm$ 0.3 & 92.0 $\pm$ 0.3 & 87.2 $\pm$ 0.5 & 94.2 $\pm$ 0.3 & 59.5 $\pm$ 1.2 & 78.4 $\pm$ 1.0 & 74.2 $\pm$ 0.7 \\
& DynSignSparseSGD & 68.6 $\pm$ 0.2 & 71.0 $\pm$ 0.1 & 33.8 $\pm$ 2.1 & 82.0 $\pm$ 1.4 & 73.7 $\pm$ 0.3 & 85.1 $\pm$ 0.1 & 91.7 $\pm$ 0.4 & 89.6 $\pm$ 0.2 & 94.3 $\pm$ 0.4 & 60.1 $\pm$ 1.2 & 79.7 $\pm$ 0.9 & 75.4 $\pm$ 0.7 \\
\midrule
\multirow{4}{*}{4} 
& Local-clip SparseSGD & 70.9 $\pm$ 0.1 & 70.8 $\pm$ 0.1 & 31.0 $\pm$ 0.3 & 69.7 $\pm$ 0.5 & 70.2 $\pm$ 0.2 & 86.8 $\pm$ 0.1 & 92.9 $\pm$ 0.1 & 77.3 $\pm$ 0.3 & 94.5 $\pm$ 0.1 & 54.3 $\pm$ 0.5 & 73.6 $\pm$ 0.1 & 72.0 $\pm$ 0.2 \\
& Global-clip SparseSGD & 67.0 $\pm$ 0.0 & 63.0 $\pm$ 0.0 & 24.1 $\pm$ 0.0 & 42.3 $\pm$ 0.0 & 65.6 $\pm$ 0.0 & 85.3 $\pm$ 0.0 & 88.3 $\pm$ 0.0 & 67.6 $\pm$ 0.1 & 93.3 $\pm$ 0.0 & 44.4 $\pm$ 0.0 & 65.2 $\pm$ 0.0 & 64.2 $\pm$ 0.0 \\
\cmidrule(lr){2-14}
& SignSparseSGD & 68.5 $\pm$ 0.3 & 71.9 $\pm$ 0.2 & 37.3 $\pm$ 0.7 & 81.5 $\pm$ 1.2 & 75.7 $\pm$ 0.7 & 84.7 $\pm$ 0.3 & 92.5 $\pm$ 0.2 & 91.4 $\pm$ 1.1 & 95.4 $\pm$ 0.1 & 62.6 $\pm$ 0.7 & 81.1 $\pm$ 1.1 & 76.6 $\pm$ 0.6 \\
& DynSignSparseSGD & 69.4 $\pm$ 0.2 & 72.8 $\pm$ 0.1 & 38.4 $\pm$ 1.1 & 84.1 $\pm$ 2.0 & 77.8 $\pm$ 0.3 & 85.4 $\pm$ 0.2 & 92.8 $\pm$ 0.3 & 93.2 $\pm$ 0.8 & 95.2 $\pm$ 0.1 & 64.5 $\pm$ 0.7 & 82.2 $\pm$ 1.2 & 77.8 $\pm$ 0.6 \\
\midrule
\multirow{4}{*}{8} 
& Local-clip SparseSGD & 71.5 $\pm$ 0.0 & 71.9 $\pm$ 0.0 & 33.9 $\pm$ 0.5 & 75.2 $\pm$ 0.4 & 71.8 $\pm$ 0.4 & 87.1 $\pm$ 0.0 & 93.0 $\pm$ 0.1 & 81.7 $\pm$ 0.1 & 94.9 $\pm$ 0.1 & 57.2 $\pm$ 0.7 & 75.9 $\pm$ 0.4 & 74.0 $\pm$ 0.2 \\
& Global-clip SparseSGD & 67.2 $\pm$ 0.0 & 63.4 $\pm$ 0.1 & 24.1 $\pm$ 0.0 & 42.5 $\pm$ 0.0 & 65.7 $\pm$ 0.0 & 85.3 $\pm$ 0.0 & 88.3 $\pm$ 0.0 & 67.7 $\pm$ 0.1 & 93.3 $\pm$ 0.0 & 44.4 $\pm$ 0.1 & 65.3 $\pm$ 0.0 & 64.3 $\pm$ 0.0 \\
\cmidrule(lr){2-14}
& SignSparseSGD & 69.9 $\pm$ 0.3 & 73.7 $\pm$ 0.1 & 41.1 $\pm$ 1.7 & 84.7 $\pm$ 5.2 & 79.8 $\pm$ 0.7 & 84.8 $\pm$ 0.6 & 92.9 $\pm$ 0.4 & 94.0 $\pm$ 0.3 & 95.9 $\pm$ 0.2 & 66.5 $\pm$ 0.7 & 83.6 $\pm$ 0.6 & 78.8 $\pm$ 1.0 \\
& DynSignSparseSGD & 70.5 $\pm$ 0.4 & 74.5 $\pm$ 0.2 & 45.2 $\pm$ 0.7 & 85.2 $\pm$ 4.0 & 82.4 $\pm$ 0.5 & 85.3 $\pm$ 0.3 & 93.4 $\pm$ 0.1 & 95.0 $\pm$ 0.2 & 95.8 $\pm$ 0.1 & 68.0 $\pm$ 1.3 & 84.7 $\pm$ 0.2 & 80.0 $\pm$ 0.7 \\
\midrule
\multirow{4}{*}{16} 
& Local-clip SparseSGD & 72.3 $\pm$ 0.1 & 73.5 $\pm$ 0.0 & 36.1 $\pm$ 0.3 & 80.4 $\pm$ 1.0 & 74.7 $\pm$ 0.1 & 87.4 $\pm$ 0.2 & 93.7 $\pm$ 0.0 & 87.9 $\pm$ 0.5 & 95.1 $\pm$ 0.1 & 62.1 $\pm$ 0.4 & 79.4 $\pm$ 0.2 & 76.6 $\pm$ 0.3 \\
& Global-clip SparseSGD & 67.6 $\pm$ 0.0 & 64.3 $\pm$ 0.1 & 24.2 $\pm$ 0.1 & 42.6 $\pm$ 0.0 & 65.8 $\pm$ 0.1 & 85.4 $\pm$ 0.0 & 88.4 $\pm$ 0.0 & 68.0 $\pm$ 0.0 & 93.3 $\pm$ 0.0 & 44.4 $\pm$ 0.1 & 65.7 $\pm$ 0.0 & 64.5 $\pm$ 0.0 \\
\cmidrule(lr){2-14}
& SignSparseSGD & 71.6 $\pm$ 0.1 & 75.0 $\pm$ 0.2 & 45.6 $\pm$ 0.9 & 91.2 $\pm$ 1.1 & 83.6 $\pm$ 0.5 & 85.7 $\pm$ 0.1 & 93.6 $\pm$ 0.3 & 96.8 $\pm$ 0.2 & 96.2 $\pm$ 0.2 & 71.2 $\pm$ 0.3 & 85.8 $\pm$ 0.0 & 81.5 $\pm$ 0.4 \\
& DynSignSparseSGD & 72.2 $\pm$ 0.1 & 75.7 $\pm$ 0.2 & 51.7 $\pm$ 0.6 & 91.8 $\pm$ 0.5 & 85.9 $\pm$ 0.3 & 86.1 $\pm$ 0.1 & 93.9 $\pm$ 0.1 & 97.7 $\pm$ 0.2 & 96.4 $\pm$ 0.3 & 72.2 $\pm$ 0.4 & 86.9 $\pm$ 0.1 & 82.8 $\pm$ 0.3 \\
\bottomrule
\end{tabular}
\end{adjustbox}
\end{table}

\begin{table}[!h]
\caption{Few-shot classification performance (shots 1, 2, 4, 8, 16) on 11 datasets with CLIP(ViT-L/14) backbone. We compare SignSparseSGD and DynSignSparseSGD against two stabilised variants: Local-clip SparseSGD and Global-clip SparseSGD. Top-1 accuracy (\%) averaged over 3 seeds.}
\label{tab:ablation_signsparsesgd_clip_ViTL14_with_aug}
\centering
\footnotesize
\setlength{\tabcolsep}{1.8pt}
\renewcommand{\arraystretch}{1.00}
\begin{adjustbox}{max totalsize={\textwidth}{0.95\textheight},center}
\begin{tabular}{@{}c l *{12}{d}@{}}
\toprule
Shots & Method & ImageNet & SUN & Aircr. & EuroSAT & Cars & Food & Pets & Flow. & Caltech & DTD & UCF & Avg. \\ \midrule
\multirow{1}{*}{0} & CLIP & 73.5 & 67.7 & 30.0 & 55.1 & 76.8 & 90.3 & 93.1 & 76.2 & 95.2 & 52.4 & 73.7 & 71.3 \\
\midrule
\multirow{4}{*}{1} 
& Local-clip SparseSGD& 75.9 $\pm$ 0.0 & 71.3 $\pm$ 0.1 & 35.8 $\pm$ 0.2 & 65.6 $\pm$ 2.6 & 78.5 $\pm$ 0.1 & 90.9 $\pm$ 0.0 & 94.2 $\pm$ 0.1 & 80.4 $\pm$ 0.1 & 95.5 $\pm$ 0.3 & 56.3 $\pm$ 0.3 & 76.9 $\pm$ 0.3 & 74.7 $\pm$ 0.4 \\
& Global-clip SparseSGD & 73.5 $\pm$ 0.0 & 67.7 $\pm$ 0.0 & 30.1 $\pm$ 0.0 & 55.2 $\pm$ 0.0 & 76.9 $\pm$ 0.0 & 90.3 $\pm$ 0.0 & 93.1 $\pm$ 0.0 & 76.2 $\pm$ 0.0 & 95.2 $\pm$ 0.0 & 52.4 $\pm$ 0.0 & 73.7 $\pm$ 0.0 & 71.3 $\pm$ 0.0 \\
\cmidrule(lr){2-14}
& SignSparseSGD & 74.6 $\pm$ 0.5 & 73.3 $\pm$ 0.2 & 40.5 $\pm$ 1.7 & 71.7 $\pm$ 3.7 & 80.0 $\pm$ 0.3 & 90.2 $\pm$ 0.1 & 94.6 $\pm$ 0.2 & 91.5 $\pm$ 1.3 & 95.5 $\pm$ 0.3 & 61.7 $\pm$ 1.1 & 82.8 $\pm$ 0.5 & 77.8 $\pm$ 0.9 \\
& DynSignSparseSGD & 75.6 $\pm$ 0.2 & 74.1 $\pm$ 0.2 & 41.3 $\pm$ 0.9 & 73.8 $\pm$ 4.1 & 80.9 $\pm$ 0.3 & 90.7 $\pm$ 0.2 & 94.7 $\pm$ 0.3 & 92.3 $\pm$ 1.0 & 95.5 $\pm$ 0.2 & 61.5 $\pm$ 0.7 & 82.9 $\pm$ 0.8 & 78.5 $\pm$ 0.8 \\
\midrule
\multirow{4}{*}{2} 
& Local-clip SparseSGD& 76.5 $\pm$ 0.1 & 72.4 $\pm$ 0.0 & 37.4 $\pm$ 0.5 & 67.4 $\pm$ 0.3 & 78.9 $\pm$ 0.1 & 91.1 $\pm$ 0.2 & 94.4 $\pm$ 0.1 & 81.5 $\pm$ 0.5 & 95.7 $\pm$ 0.1 & 57.9 $\pm$ 0.3 & 78.0 $\pm$ 0.6 & 75.5 $\pm$ 0.2 \\
& Global-clip SparseSGD & 73.5 $\pm$ 0.0 & 67.8 $\pm$ 0.0 & 30.1 $\pm$ 0.0 & 55.2 $\pm$ 0.0 & 76.9 $\pm$ 0.0 & 90.3 $\pm$ 0.0 & 93.1 $\pm$ 0.0 & 76.2 $\pm$ 0.0 & 95.2 $\pm$ 0.0 & 52.4 $\pm$ 0.0 & 73.7 $\pm$ 0.0 & 71.3 $\pm$ 0.0 \\
\cmidrule(lr){2-14}
& SignSparseSGD & 75.3 $\pm$ 0.2 & 74.9 $\pm$ 0.1 & 41.9 $\pm$ 1.9 & 84.4 $\pm$ 0.7 & 82.2 $\pm$ 0.4 & 90.5 $\pm$ 0.2 & 94.3 $\pm$ 0.4 & 94.3 $\pm$ 1.0 & 97.0 $\pm$ 0.3 & 67.6 $\pm$ 1.2 & 84.7 $\pm$ 0.7 & 80.6 $\pm$ 0.6 \\
& DynSignSparseSGD & 75.9 $\pm$ 0.3 & 75.6 $\pm$ 0.3 & 43.9 $\pm$ 2.6 & 82.6 $\pm$ 1.0 & 83.2 $\pm$ 0.6 & 90.8 $\pm$ 0.2 & 94.5 $\pm$ 0.3 & 95.2 $\pm$ 0.7 & 96.8 $\pm$ 0.6 & 68.5 $\pm$ 1.2 & 85.0 $\pm$ 0.6 & 81.1 $\pm$ 0.8 \\
\midrule
\multirow{4}{*}{4} 
& Local-clip SparseSGD& 77.1 $\pm$ 0.1 & 73.8 $\pm$ 0.1 & 39.5 $\pm$ 0.2 & 68.6 $\pm$ 0.4 & 80.1 $\pm$ 0.1 & 91.4 $\pm$ 0.1 & 94.7 $\pm$ 0.1 & 83.1 $\pm$ 0.6 & 96.2 $\pm$ 0.1 & 59.7 $\pm$ 0.3 & 79.2 $\pm$ 0.2 & 76.7 $\pm$ 0.2 \\
& Global-clip SparseSGD & 73.6 $\pm$ 0.0 & 67.9 $\pm$ 0.0 & 30.1 $\pm$ 0.0 & 55.2 $\pm$ 0.0 & 76.9 $\pm$ 0.0 & 90.3 $\pm$ 0.0 & 93.1 $\pm$ 0.0 & 76.2 $\pm$ 0.0 & 95.2 $\pm$ 0.0 & 52.4 $\pm$ 0.0 & 73.7 $\pm$ 0.0 & 71.3 $\pm$ 0.0 \\
\cmidrule(lr){2-14}
& SignSparseSGD & 76.1 $\pm$ 0.2 & 76.2 $\pm$ 0.1 & 47.5 $\pm$ 0.1 & 85.0 $\pm$ 3.0 & 84.7 $\pm$ 0.7 & 90.6 $\pm$ 0.1 & 94.9 $\pm$ 0.0 & 96.5 $\pm$ 0.8 & 97.3 $\pm$ 0.1 & 69.4 $\pm$ 0.2 & 87.4 $\pm$ 1.0 & 82.3 $\pm$ 0.6 \\
& DynSignSparseSGD & 76.6 $\pm$ 0.1 & 77.0 $\pm$ 0.1 & 50.5 $\pm$ 0.6 & 85.4 $\pm$ 2.5 & 86.0 $\pm$ 0.9 & 91.1 $\pm$ 0.2 & 95.0 $\pm$ 0.1 & 97.0 $\pm$ 1.0 & 97.5 $\pm$ 0.3 & 70.8 $\pm$ 0.6 & 87.6 $\pm$ 0.7 & 83.1 $\pm$ 0.6 \\
\midrule
\multirow{4}{*}{8} 
& Local-clip SparseSGD& 77.7 $\pm$ 0.1 & 75.2 $\pm$ 0.0 & 42.3 $\pm$ 0.6 & 73.2 $\pm$ 0.5 & 81.5 $\pm$ 0.1 & 91.8 $\pm$ 0.1 & 94.6 $\pm$ 0.2 & 86.0 $\pm$ 0.6 & 96.5 $\pm$ 0.2 & 62.6 $\pm$ 0.4 & 81.1 $\pm$ 0.5 & 78.4 $\pm$ 0.3 \\
& Global-clip SparseSGD & 73.7 $\pm$ 0.0 & 68.1 $\pm$ 0.0 & 30.2 $\pm$ 0.1 & 55.3 $\pm$ 0.0 & 76.9 $\pm$ 0.0 & 90.3 $\pm$ 0.0 & 93.1 $\pm$ 0.0 & 76.2 $\pm$ 0.0 & 95.2 $\pm$ 0.0 & 52.4 $\pm$ 0.1 & 73.8 $\pm$ 0.0 & 71.4 $\pm$ 0.0 \\
\cmidrule(lr){2-14}
& SignSparseSGD & 76.8 $\pm$ 0.2 & 77.5 $\pm$ 0.0 & 56.0 $\pm$ 1.2 & 89.2 $\pm$ 2.7 & 87.6 $\pm$ 0.3 & 90.7 $\pm$ 0.2 & 95.4 $\pm$ 0.2 & 97.3 $\pm$ 0.3 & 97.5 $\pm$ 0.6 & 72.0 $\pm$ 0.6 & 88.6 $\pm$ 0.5 & 84.4 $\pm$ 0.6 \\
& DynSignSparseSGD & 77.3 $\pm$ 0.1 & 78.3 $\pm$ 0.1 & 58.7 $\pm$ 0.6 & 88.9 $\pm$ 2.7 & 88.8 $\pm$ 0.2 & 91.2 $\pm$ 0.1 & 95.5 $\pm$ 0.3 & 97.8 $\pm$ 0.5 & 97.4 $\pm$ 0.5 & 73.1 $\pm$ 0.4 & 89.2 $\pm$ 0.4 & 85.1 $\pm$ 0.5 \\
\midrule
\multirow{4}{*}{16} 
& Local-clip SparseSGD& 78.4 $\pm$ 0.0 & 76.7 $\pm$ 0.0 & 45.9 $\pm$ 0.4 & 78.0 $\pm$ 1.4 & 83.0 $\pm$ 0.1 & 92.0 $\pm$ 0.1 & 95.0 $\pm$ 0.1 & 90.6 $\pm$ 0.5 & 97.0 $\pm$ 0.1 & 67.7 $\pm$ 0.5 & 83.3 $\pm$ 0.2 & 80.7 $\pm$ 0.3 \\
& Global-clip SparseSGD & 73.9 $\pm$ 0.0 & 68.3 $\pm$ 0.1 & 30.2 $\pm$ 0.1 & 55.5 $\pm$ 0.0 & 76.9 $\pm$ 0.0 & 90.3 $\pm$ 0.0 & 93.2 $\pm$ 0.0 & 76.3 $\pm$ 0.0 & 95.2 $\pm$ 0.0 & 52.4 $\pm$ 0.1 & 73.9 $\pm$ 0.0 & 71.5 $\pm$ 0.0 \\
\cmidrule(lr){2-14}
& SignSparseSGD & 78.0 $\pm$ 0.1 & 79.3 $\pm$ 0.3 & 61.6 $\pm$ 0.9 & 92.7 $\pm$ 1.5 & 89.5 $\pm$ 0.3 & 90.8 $\pm$ 0.2 & 95.4 $\pm$ 0.3 & 98.5 $\pm$ 0.1 & 97.7 $\pm$ 0.0 & 76.5 $\pm$ 0.8 & 90.1 $\pm$ 0.2 & 86.4 $\pm$ 0.4 \\
& DynSignSparseSGD & 78.3 $\pm$ 0.1 & 79.6 $\pm$ 0.1 & 65.0 $\pm$ 0.3 & 93.4 $\pm$ 0.9 & 90.3 $\pm$ 0.1 & 91.2 $\pm$ 0.0 & 95.6 $\pm$ 0.1 & 98.9 $\pm$ 0.2 & 97.4 $\pm$ 0.3 & 77.8 $\pm$ 0.7 & 91.0 $\pm$ 0.2 & 87.1 $\pm$ 0.3 \\
\bottomrule
\end{tabular}
\end{adjustbox}
\end{table}

Tables~\ref{tab:ablation_signsparsesgd_clip_ViTB32_with_aug}, \ref{tab:ablation_signsparsesgd_clip_ViTB16_with_aug}, and \ref{tab:ablation_signsparsesgd_clip_ViTL14_with_aug} complement Fig.~\ref{fig:ablation} by reporting per-dataset accuracies for the clipping-based SparseSGD baselines defined in Appendix~\ref{app:algorithms}. As in Sec.~\ref{sec:experiments}, all methods use the same training schedule: 30 epochs, batch size 64, and learning rate $10^{-4}$. The goal is to test whether standard clipping can explain the gains of the proposed sign-sparse optimizers.

A first observation is that global clipping is too conservative in this setting. Across the three backbones, Global-clip SparseSGD stays almost identical to the zero-shot CLIP baseline. On ViT-B/16, its average accuracy moves from the zero-shot value of $64.1\%$ to only $64.1$--$64.5\%$ from 1-shot to 16-shot. The corresponding ranges are $60.2$--$60.6\%$ for ViT-B/32 and $71.3$--$71.5\%$ for ViT-L/14. This matches Fig.~\ref{fig:ablation}: global clipping remains much closer to zero-shot CLIP than to the sign-sparse methods.

A second observation is that local clipping is clearly stronger than global clipping, but it still does not recover the gains of sign-sparse updates. Local-clip SparseSGD improves over Global-clip SparseSGD in all 165 backbone--shot--dataset comparisons. In the average columns, the mean gap between the two clipping baselines is $7.2$ points over the 15 backbone--shot configurations. This gap grows with the number of shots, from $4.3$ points at 1-shot to $11.0$ points at 16-shot when averaged over the three backbones.

Even the stronger local-clipping baseline remains below the sign-sparse update methods in every average column. Averaged over the 15 backbone--shot configurations, SignSparseSGD improves over Local-clip SparseSGD by $4.4$ points and DynSignSparseSGD by $5.5$ points. DynSignSparseSGD is the best method in all 15 average columns. The gap also widens with the number of shots. Averaged over the 3 backbones (ViT-B/32, ViT-B/16 and ViT-L/14), SignSparseSGD and DynSignSparseSGD improve over Local-clip SparseSGD by $+2.4$ and $+3.5$ points at 1-shot, and by $+5.3$ and $+6.5$ points at 16-shot.

The advantage is not uniform on every dataset. Local clipping remains competitive on ImageNet, Food101, Pets, and Caltech. It is even better than both sign-sparse methods on ImageNet and Food101 in all 15 backbone--shot configurations, but these gains do not change the overall ranking. However, the sign-sparse methods are much stronger on Aircraft, EuroSAT, DTD, Flowers, Cars, and UCF101. For example, at 16-shot on ViT-B/32, DynSignSparseSGD improves over Local-clip SparseSGD by $+15.0$ points on Aircraft, $+17.6$ on EuroSAT, and $+8.4$ on DTD. Similar gaps appear on ViT-B/16 and ViT-L/14.

Overall, these tables support the interpretation of the main paper. Clipping can stabilize SparseSGD, especially in its local variant, but clipping alone does not explain the gains of SignSparseSGD and DynSignSparseSGD. The benefit comes from combining sparsity with fixed-magnitude sign updates, and it is strongest when the support is refreshed during training. Gradient magnitude helps choose where to update, but standard clipping is not enough to determine how those coordinates should be updated.

\section{Randomness vs Gradient Magnitude}
\label{app:random_vs_magnitude}
\suppanchor{random-selection}

\begin{table}
\caption{Ablation on connection selection in few-shot classification with CLIP (ViT-B/32) backbone. We compare SignSparseSGD and DynSignSparseSGD under random gradient pruning (Rand) and importance-based gradient pruning (Impt). Top-1 accuracy (\%) averaged over 3 fixed random seeds.}
\label{tab:ablation_selection_ViTB32_with_aug}
\centering
\footnotesize
\setlength{\tabcolsep}{1.8pt}
\renewcommand{\arraystretch}{1.00}
\begin{adjustbox}{max totalsize={\textwidth}{0.95\textheight},center}
\begin{tabular}{@{}c l *{14}{d}@{}}
\toprule
Shots & Method & \multicolumn{2}{c}{Selection} & ImageNet & SUN & Aircr. & EuroSAT & Cars & Food & Pets & Flow. & Caltech & DTD & UCF & Avg. \\ 
\cmidrule(lr){3-4}
 &  & Rand & Impt &  &  &  &  &  &  &  &  &  &  &  &  \\ \midrule
\multirow{1}{*}{0} & CLIP &  &  & 62.0 & 62.1 & 18.1 & 35.8 & 60.1 & 78.9 & 85.1 & 64.0 & 91.4 & 43.0 & 61.6 & 60.2 \\
\midrule
\multirow{4}{*}{1} & SignSparseSGD & $\checkmark$ & $\times$ & 64.9 $\pm$ 0.2 & 66.3 $\pm$ 0.0 & 20.8 $\pm$ 0.3 & 47.0 $\pm$ 1.1 & 62.5 $\pm$ 0.1 & 80.6 $\pm$ 0.1 & 87.2 $\pm$ 0.0 & 68.3 $\pm$ 0.2 & 92.5 $\pm$ 0.1 & 46.7 $\pm$ 0.4 & 65.6 $\pm$ 0.2 & 63.8 $\pm$ 0.2 \\
& SignSparseSGD & $\times$ & $\checkmark$ & 60.6 $\pm$ 0.6 & 65.3 $\pm$ 0.1 & 21.6 $\pm$ 0.5 & 66.0 $\pm$ 5.2 & 61.3 $\pm$ 1.1 & 77.9 $\pm$ 0.5 & 88.5 $\pm$ 0.4 & 75.0 $\pm$ 0.8 & 92.3 $\pm$ 0.5 & 50.9 $\pm$ 1.5 & 71.5 $\pm$ 0.3 & 66.4 $\pm$ 1.0 \\
\cmidrule(lr){2-16}
& DynSignSparseSGD & $\checkmark$ & $\times$ & 65.0 $\pm$ 0.2 & 66.2 $\pm$ 0.1 & 20.8 $\pm$ 0.1 & 47.0 $\pm$ 1.0 & 62.5 $\pm$ 0.2 & 80.7 $\pm$ 0.1 & 87.3 $\pm$ 0.1 & 68.4 $\pm$ 0.3 & 92.5 $\pm$ 0.1 & 46.7 $\pm$ 0.6 & 65.7 $\pm$ 0.1 & 63.9 $\pm$ 0.2 \\
& DynSignSparseSGD & $\times$ & $\checkmark$ & 62.6 $\pm$ 0.3 & 67.0 $\pm$ 0.3 & 22.7 $\pm$ 0.8 & 69.1 $\pm$ 2.7 & 63.1 $\pm$ 0.4 & 79.7 $\pm$ 0.5 & 89.1 $\pm$ 0.2 & 77.7 $\pm$ 1.2 & 92.6 $\pm$ 0.7 & 51.9 $\pm$ 1.0 & 72.0 $\pm$ 0.4 & 67.9 $\pm$ 0.8 \\
\midrule
\multirow{4}{*}{2} & SignSparseSGD & $\checkmark$ & $\times$ & 65.6 $\pm$ 0.1 & 67.7 $\pm$ 0.2 & 21.8 $\pm$ 0.1 & 50.9 $\pm$ 1.1 & 63.5 $\pm$ 0.1 & 81.4 $\pm$ 0.1 & 87.7 $\pm$ 0.2 & 70.3 $\pm$ 0.8 & 92.9 $\pm$ 0.1 & 50.1 $\pm$ 0.4 & 67.7 $\pm$ 0.2 & 65.4 $\pm$ 0.3 \\
& SignSparseSGD & $\times$ & $\checkmark$ & 61.9 $\pm$ 0.3 & 66.9 $\pm$ 0.2 & 24.3 $\pm$ 0.6 & 79.1 $\pm$ 6.7 & 63.6 $\pm$ 0.1 & 78.0 $\pm$ 0.4 & 89.1 $\pm$ 0.4 & 78.7 $\pm$ 0.2 & 93.8 $\pm$ 0.3 & 56.4 $\pm$ 0.4 & 74.1 $\pm$ 0.8 & 69.6 $\pm$ 0.9 \\
\cmidrule(lr){2-16}
& DynSignSparseSGD & $\checkmark$ & $\times$ & 65.7 $\pm$ 0.0 & 67.8 $\pm$ 0.1 & 21.7 $\pm$ 0.1 & 51.0 $\pm$ 0.9 & 63.4 $\pm$ 0.3 & 81.4 $\pm$ 0.1 & 87.7 $\pm$ 0.3 & 70.4 $\pm$ 0.6 & 92.9 $\pm$ 0.2 & 50.1 $\pm$ 0.4 & 67.9 $\pm$ 0.3 & 65.5 $\pm$ 0.3 \\
& DynSignSparseSGD & $\times$ & $\checkmark$ & 63.2 $\pm$ 0.1 & 68.2 $\pm$ 0.3 & 25.5 $\pm$ 0.6 & 79.8 $\pm$ 4.2 & 66.0 $\pm$ 0.1 & 79.5 $\pm$ 0.4 & 89.2 $\pm$ 0.3 & 83.8 $\pm$ 1.4 & 94.0 $\pm$ 0.3 & 57.1 $\pm$ 1.1 & 74.8 $\pm$ 0.6 & 71.0 $\pm$ 0.9 \\
\midrule
\multirow{4}{*}{4} & SignSparseSGD & $\checkmark$ & $\times$ & 66.3 $\pm$ 0.0 & 69.5 $\pm$ 0.2 & 23.4 $\pm$ 0.3 & 53.2 $\pm$ 0.5 & 65.3 $\pm$ 0.1 & 81.7 $\pm$ 0.1 & 89.7 $\pm$ 0.3 & 74.6 $\pm$ 0.1 & 94.2 $\pm$ 0.1 & 52.5 $\pm$ 0.6 & 69.8 $\pm$ 0.3 & 67.3 $\pm$ 0.2 \\
& SignSparseSGD & $\times$ & $\checkmark$ & 63.3 $\pm$ 0.4 & 69.0 $\pm$ 0.4 & 25.2 $\pm$ 1.1 & 82.6 $\pm$ 1.0 & 66.9 $\pm$ 0.3 & 78.2 $\pm$ 0.4 & 89.3 $\pm$ 0.4 & 85.8 $\pm$ 1.2 & 94.7 $\pm$ 0.4 & 59.1 $\pm$ 1.0 & 75.9 $\pm$ 0.7 & 71.8 $\pm$ 0.7 \\
\cmidrule(lr){2-16}
& DynSignSparseSGD & $\checkmark$ & $\times$ & 66.6 $\pm$ 0.1 & 69.7 $\pm$ 0.1 & 23.4 $\pm$ 0.3 & 53.2 $\pm$ 0.4 & 65.5 $\pm$ 0.1 & 81.8 $\pm$ 0.1 & 89.6 $\pm$ 0.3 & 74.6 $\pm$ 0.1 & 94.3 $\pm$ 0.0 & 52.6 $\pm$ 0.4 & 69.7 $\pm$ 0.4 & 67.4 $\pm$ 0.2 \\
& DynSignSparseSGD & $\times$ & $\checkmark$ & 64.0 $\pm$ 0.2 & 69.9 $\pm$ 0.3 & 27.7 $\pm$ 0.6 & 83.9 $\pm$ 1.6 & 69.3 $\pm$ 0.1 & 79.6 $\pm$ 0.4 & 90.0 $\pm$ 0.4 & 88.9 $\pm$ 0.8 & 95.2 $\pm$ 0.2 & 61.1 $\pm$ 0.3 & 78.0 $\pm$ 0.9 & 73.4 $\pm$ 0.5 \\
\midrule
\multirow{4}{*}{8} & SignSparseSGD & $\checkmark$ & $\times$ & 67.2 $\pm$ 0.1 & 71.4 $\pm$ 0.2 & 26.5 $\pm$ 0.2 & 64.1 $\pm$ 1.4 & 67.8 $\pm$ 0.4 & 82.3 $\pm$ 0.0 & 91.2 $\pm$ 0.2 & 81.4 $\pm$ 0.2 & 94.6 $\pm$ 0.2 & 57.4 $\pm$ 0.7 & 73.3 $\pm$ 0.1 & 70.6 $\pm$ 0.3 \\
& SignSparseSGD & $\times$ & $\checkmark$ & 64.7 $\pm$ 0.1 & 70.9 $\pm$ 0.3 & 31.1 $\pm$ 0.8 & 85.9 $\pm$ 0.5 & 71.4 $\pm$ 0.5 & 79.0 $\pm$ 0.2 & 90.3 $\pm$ 0.1 & 89.3 $\pm$ 0.7 & 94.9 $\pm$ 0.3 & 62.3 $\pm$ 0.9 & 78.8 $\pm$ 0.7 & 74.4 $\pm$ 0.5 \\
\cmidrule(lr){2-16}
& DynSignSparseSGD & $\checkmark$ & $\times$ & 67.6 $\pm$ 0.0 & 71.5 $\pm$ 0.1 & 26.4 $\pm$ 0.3 & 63.9 $\pm$ 1.6 & 68.1 $\pm$ 0.3 & 82.3 $\pm$ 0.1 & 91.2 $\pm$ 0.3 & 81.4 $\pm$ 0.1 & 94.7 $\pm$ 0.1 & 57.2 $\pm$ 0.5 & 73.3 $\pm$ 0.0 & 70.7 $\pm$ 0.3 \\
& DynSignSparseSGD & $\times$ & $\checkmark$ & 65.0 $\pm$ 0.2 & 71.9 $\pm$ 0.1 & 34.8 $\pm$ 0.6 & 88.4 $\pm$ 1.0 & 75.1 $\pm$ 0.2 & 79.8 $\pm$ 0.1 & 91.0 $\pm$ 0.3 & 92.5 $\pm$ 0.1 & 95.5 $\pm$ 0.3 & 64.5 $\pm$ 1.0 & 79.7 $\pm$ 0.6 & 76.2 $\pm$ 0.4 \\
\midrule
\multirow{4}{*}{16} & SignSparseSGD & $\checkmark$ & $\times$ & 68.2 $\pm$ 0.1 & 73.5 $\pm$ 0.1 & 29.2 $\pm$ 0.7 & 74.2 $\pm$ 1.5 & 71.6 $\pm$ 0.2 & 82.7 $\pm$ 0.0 & 91.8 $\pm$ 0.2 & 88.5 $\pm$ 0.4 & 95.3 $\pm$ 0.2 & 63.0 $\pm$ 0.6 & 77.4 $\pm$ 0.3 & 74.1 $\pm$ 0.4 \\
& SignSparseSGD & $\times$ & $\checkmark$ & 66.3 $\pm$ 0.2 & 73.1 $\pm$ 0.3 & 36.4 $\pm$ 0.9 & 90.7 $\pm$ 1.6 & 76.4 $\pm$ 0.6 & 79.3 $\pm$ 0.5 & 89.9 $\pm$ 1.0 & 93.8 $\pm$ 1.0 & 95.8 $\pm$ 0.0 & 67.5 $\pm$ 0.2 & 81.7 $\pm$ 0.4 & 77.4 $\pm$ 0.6 \\
\cmidrule(lr){2-16}
& DynSignSparseSGD & $\checkmark$ & $\times$ & 69.0 $\pm$ 0.1 & 73.8 $\pm$ 0.1 & 29.4 $\pm$ 0.5 & 74.2 $\pm$ 1.3 & 71.9 $\pm$ 0.1 & 82.7 $\pm$ 0.1 & 91.8 $\pm$ 0.1 & 88.9 $\pm$ 0.1 & 95.3 $\pm$ 0.1 & 62.5 $\pm$ 0.5 & 77.3 $\pm$ 0.2 & 74.3 $\pm$ 0.3 \\
& DynSignSparseSGD & $\times$ & $\checkmark$ & 66.9 $\pm$ 0.2 & 73.5 $\pm$ 0.1 & 42.0 $\pm$ 1.1 & 92.0 $\pm$ 0.6 & 79.8 $\pm$ 0.4 & 80.3 $\pm$ 0.4 & 91.0 $\pm$ 0.3 & 95.7 $\pm$ 1.0 & 95.8 $\pm$ 0.0 & 69.2 $\pm$ 0.5 & 82.7 $\pm$ 0.5 & 79.0 $\pm$ 0.5 \\
\bottomrule
\end{tabular}
\end{adjustbox}
\end{table}
\begin{table}
\caption{Ablation on connection selection in few-shot classification with CLIP (ViT-B/16) backbone. We compare SignSparseSGD and DynSignSparseSGD under random gradient pruning (Rand) and importance-based gradient pruning (Impt). Top-1 accuracy (\%) averaged over 3 fixed random seeds.}
\label{tab:ablation_selection_ViTB16_with_aug}
\centering
\footnotesize
\setlength{\tabcolsep}{1.8pt}
\renewcommand{\arraystretch}{1.00}
\begin{adjustbox}{max totalsize={\textwidth}{0.95\textheight},center}
\begin{tabular}{@{}c l *{14}{d}@{}}
\toprule
Shots & Method & \multicolumn{2}{c}{Selection} & ImageNet & SUN & Aircr. & EuroSAT & Cars & Food & Pets & Flow. & Caltech & DTD & UCF & Avg. \\ 
\cmidrule(lr){3-4}
 &  & Rand & Impt &  &  &  &  &  &  &  &  &  &  &  &  \\ \midrule
\multirow{1}{*}{0} & CLIP &  &  & 66.7 & 62.6 & 23.9 & 42.2 & 65.5 & 85.2 & 88.3 & 67.4 & 93.3 & 44.4 & 65.2 & 64.1 \\
\midrule
\multirow{4}{*}{1} & SignSparseSGD & $\checkmark$ & $\times$ & 70.0 $\pm$ 0.1 & 68.8 $\pm$ 0.1 & 27.6 $\pm$ 0.2 & 57.9 $\pm$ 1.2 & 68.2 $\pm$ 0.1 & 86.1 $\pm$ 0.1 & 89.9 $\pm$ 0.2 & 73.1 $\pm$ 0.2 & 94.0 $\pm$ 0.2 & 48.9 $\pm$ 0.5 & 70.0 $\pm$ 0.3 & 68.6 $\pm$ 0.3 \\
& SignSparseSGD & $\times$ & $\checkmark$ & 66.8 $\pm$ 0.1 & 68.2 $\pm$ 0.3 & 29.4 $\pm$ 1.3 & 71.2 $\pm$ 3.7 & 69.1 $\pm$ 0.6 & 83.8 $\pm$ 0.5 & 91.3 $\pm$ 0.3 & 83.3 $\pm$ 0.5 & 93.7 $\pm$ 0.2 & 54.3 $\pm$ 0.6 & 75.7 $\pm$ 0.3 & 71.5 $\pm$ 0.8 \\
\cmidrule(lr){2-16}
& DynSignSparseSGD & $\checkmark$ & $\times$ & 70.0 $\pm$ 0.1 & 68.9 $\pm$ 0.1 & 27.6 $\pm$ 0.2 & 58.1 $\pm$ 1.2 & 68.2 $\pm$ 0.1 & 86.2 $\pm$ 0.1 & 90.0 $\pm$ 0.2 & 73.2 $\pm$ 0.2 & 93.9 $\pm$ 0.2 & 49.0 $\pm$ 0.5 & 70.1 $\pm$ 0.3 & 68.6 $\pm$ 0.3 \\
& DynSignSparseSGD & $\times$ & $\checkmark$ & 68.1 $\pm$ 0.1 & 69.8 $\pm$ 0.3 & 30.6 $\pm$ 1.2 & 73.7 $\pm$ 4.0 & 70.6 $\pm$ 0.4 & 85.1 $\pm$ 0.3 & 92.5 $\pm$ 0.2 & 84.7 $\pm$ 0.4 & 93.8 $\pm$ 0.1 & 54.8 $\pm$ 0.6 & 76.2 $\pm$ 0.4 & 72.7 $\pm$ 0.7 \\
\midrule
\multirow{4}{*}{2} & SignSparseSGD & $\checkmark$ & $\times$ & 70.5 $\pm$ 0.1 & 70.2 $\pm$ 0.1 & 29.5 $\pm$ 0.1 & 62.8 $\pm$ 0.3 & 69.7 $\pm$ 0.1 & 86.7 $\pm$ 0.1 & 90.3 $\pm$ 0.3 & 75.9 $\pm$ 0.2 & 94.1 $\pm$ 0.1 & 52.0 $\pm$ 0.2 & 72.4 $\pm$ 0.1 & 70.4 $\pm$ 0.2 \\
& SignSparseSGD & $\times$ & $\checkmark$ & 67.4 $\pm$ 0.4 & 69.7 $\pm$ 0.1 & 32.0 $\pm$ 1.4 & 79.5 $\pm$ 2.2 & 71.7 $\pm$ 0.2 & 84.1 $\pm$ 0.3 & 92.0 $\pm$ 0.3 & 87.2 $\pm$ 0.5 & 94.2 $\pm$ 0.3 & 59.5 $\pm$ 1.2 & 78.4 $\pm$ 1.0 & 74.2 $\pm$ 0.7 \\
\cmidrule(lr){2-16}
& DynSignSparseSGD & $\checkmark$ & $\times$ & 70.6 $\pm$ 0.0 & 70.3 $\pm$ 0.1 & 29.7 $\pm$ 0.1 & 62.8 $\pm$ 0.6 & 69.7 $\pm$ 0.2 & 86.7 $\pm$ 0.1 & 90.4 $\pm$ 0.2 & 75.9 $\pm$ 0.2 & 94.2 $\pm$ 0.1 & 52.0 $\pm$ 0.4 & 72.4 $\pm$ 0.2 & 70.4 $\pm$ 0.2 \\
& DynSignSparseSGD & $\times$ & $\checkmark$ & 68.6 $\pm$ 0.2 & 71.0 $\pm$ 0.1 & 33.8 $\pm$ 2.1 & 82.0 $\pm$ 1.4 & 73.7 $\pm$ 0.3 & 85.1 $\pm$ 0.1 & 91.7 $\pm$ 0.4 & 89.6 $\pm$ 0.2 & 94.3 $\pm$ 0.4 & 60.1 $\pm$ 1.2 & 79.7 $\pm$ 0.9 & 75.4 $\pm$ 0.7 \\
\midrule
\multirow{4}{*}{4} & SignSparseSGD & $\checkmark$ & $\times$ & 71.4 $\pm$ 0.1 & 72.1 $\pm$ 0.0 & 32.0 $\pm$ 0.1 & 65.2 $\pm$ 0.6 & 72.2 $\pm$ 0.1 & 87.0 $\pm$ 0.1 & 93.1 $\pm$ 0.2 & 81.0 $\pm$ 0.4 & 94.7 $\pm$ 0.2 & 54.8 $\pm$ 0.5 & 74.7 $\pm$ 0.1 & 72.6 $\pm$ 0.2 \\
& SignSparseSGD & $\times$ & $\checkmark$ & 68.5 $\pm$ 0.3 & 71.9 $\pm$ 0.2 & 37.3 $\pm$ 0.7 & 81.5 $\pm$ 1.2 & 75.7 $\pm$ 0.7 & 84.7 $\pm$ 0.3 & 92.5 $\pm$ 0.2 & 91.4 $\pm$ 1.1 & 95.4 $\pm$ 0.1 & 62.6 $\pm$ 0.7 & 81.1 $\pm$ 1.1 & 76.6 $\pm$ 0.6 \\
\cmidrule(lr){2-16}
& DynSignSparseSGD & $\checkmark$ & $\times$ & 71.6 $\pm$ 0.0 & 72.2 $\pm$ 0.1 & 32.2 $\pm$ 0.2 & 65.3 $\pm$ 0.6 & 72.3 $\pm$ 0.1 & 87.0 $\pm$ 0.1 & 93.0 $\pm$ 0.2 & 81.0 $\pm$ 0.4 & 94.7 $\pm$ 0.1 & 54.7 $\pm$ 0.7 & 75.0 $\pm$ 0.2 & 72.6 $\pm$ 0.2 \\
& DynSignSparseSGD & $\times$ & $\checkmark$ & 69.4 $\pm$ 0.2 & 72.8 $\pm$ 0.1 & 38.4 $\pm$ 1.1 & 84.1 $\pm$ 2.0 & 77.8 $\pm$ 0.3 & 85.4 $\pm$ 0.2 & 92.8 $\pm$ 0.3 & 93.2 $\pm$ 0.8 & 95.2 $\pm$ 0.1 & 64.5 $\pm$ 0.7 & 82.2 $\pm$ 1.2 & 77.8 $\pm$ 0.6 \\
\midrule
\multirow{4}{*}{8} & SignSparseSGD & $\checkmark$ & $\times$ & 72.3 $\pm$ 0.1 & 73.9 $\pm$ 0.2 & 35.1 $\pm$ 0.8 & 74.6 $\pm$ 0.5 & 75.1 $\pm$ 0.4 & 87.4 $\pm$ 0.1 & 93.5 $\pm$ 0.2 & 87.4 $\pm$ 0.2 & 95.2 $\pm$ 0.1 & 58.3 $\pm$ 0.2 & 78.4 $\pm$ 0.2 & 75.6 $\pm$ 0.3 \\
& SignSparseSGD & $\times$ & $\checkmark$ & 69.9 $\pm$ 0.3 & 73.7 $\pm$ 0.1 & 41.1 $\pm$ 1.7 & 84.7 $\pm$ 5.2 & 79.8 $\pm$ 0.7 & 84.8 $\pm$ 0.6 & 92.9 $\pm$ 0.4 & 94.0 $\pm$ 0.3 & 95.9 $\pm$ 0.2 & 66.5 $\pm$ 0.7 & 83.6 $\pm$ 0.6 & 78.8 $\pm$ 1.0 \\
\cmidrule(lr){2-16}
& DynSignSparseSGD & $\checkmark$ & $\times$ & 72.6 $\pm$ 0.1 & 73.9 $\pm$ 0.2 & 35.4 $\pm$ 0.5 & 74.5 $\pm$ 0.6 & 75.1 $\pm$ 0.3 & 87.4 $\pm$ 0.0 & 93.5 $\pm$ 0.2 & 87.4 $\pm$ 0.4 & 95.2 $\pm$ 0.1 & 58.7 $\pm$ 0.4 & 78.5 $\pm$ 0.1 & 75.6 $\pm$ 0.3 \\
& DynSignSparseSGD & $\times$ & $\checkmark$ & 70.5 $\pm$ 0.4 & 74.5 $\pm$ 0.2 & 45.2 $\pm$ 0.7 & 85.2 $\pm$ 4.0 & 82.4 $\pm$ 0.5 & 85.3 $\pm$ 0.3 & 93.4 $\pm$ 0.1 & 95.0 $\pm$ 0.2 & 95.8 $\pm$ 0.1 & 68.0 $\pm$ 1.3 & 84.7 $\pm$ 0.2 & 80.0 $\pm$ 0.7 \\
\midrule
\multirow{4}{*}{16} & SignSparseSGD & $\checkmark$ & $\times$ & 73.3 $\pm$ 0.1 & 75.8 $\pm$ 0.2 & 39.3 $\pm$ 0.5 & 82.2 $\pm$ 0.9 & 79.1 $\pm$ 0.1 & 87.7 $\pm$ 0.1 & 93.9 $\pm$ 0.2 & 94.1 $\pm$ 0.2 & 95.6 $\pm$ 0.1 & 65.3 $\pm$ 0.3 & 82.5 $\pm$ 0.4 & 79.0 $\pm$ 0.3 \\
& SignSparseSGD & $\times$ & $\checkmark$ & 71.6 $\pm$ 0.1 & 75.0 $\pm$ 0.2 & 45.6 $\pm$ 0.9 & 91.2 $\pm$ 1.1 & 83.6 $\pm$ 0.5 & 85.7 $\pm$ 0.1 & 93.6 $\pm$ 0.3 & 96.8 $\pm$ 0.2 & 96.2 $\pm$ 0.2 & 71.2 $\pm$ 0.3 & 85.8 $\pm$ 0.0 & 81.5 $\pm$ 0.4 \\
\cmidrule(lr){2-16}
& DynSignSparseSGD & $\checkmark$ & $\times$ & 73.9 $\pm$ 0.0 & 76.2 $\pm$ 0.1 & 39.6 $\pm$ 0.2 & 82.2 $\pm$ 0.9 & 79.2 $\pm$ 0.1 & 87.8 $\pm$ 0.1 & 94.0 $\pm$ 0.1 & 94.2 $\pm$ 0.1 & 95.6 $\pm$ 0.1 & 65.8 $\pm$ 0.2 & 82.5 $\pm$ 0.4 & 79.2 $\pm$ 0.2 \\
& DynSignSparseSGD & $\times$ & $\checkmark$ & 72.2 $\pm$ 0.1 & 75.7 $\pm$ 0.2 & 51.7 $\pm$ 0.6 & 91.8 $\pm$ 0.5 & 85.9 $\pm$ 0.3 & 86.1 $\pm$ 0.1 & 93.9 $\pm$ 0.1 & 97.7 $\pm$ 0.2 & 96.4 $\pm$ 0.3 & 72.2 $\pm$ 0.4 & 86.9 $\pm$ 0.1 & 82.8 $\pm$ 0.3 \\
\bottomrule
\end{tabular}
\end{adjustbox}
\end{table}

\begin{table}
\caption{Ablation on connection selection with CLIP (ViT-L/14) backbone. We compare SignSparseSGD and DynSignSparseSGD under random gradient pruning (Rand) and importance-based gradient pruning (Impt). Top-1 accuracy (\%) averaged over 3 fixed random seeds.}
\label{tab:ablation_selection_ViTL14_with_aug}
\centering
\footnotesize
\setlength{\tabcolsep}{1.8pt}
\renewcommand{\arraystretch}{1.00}
\begin{adjustbox}{max totalsize={\textwidth}{0.95\textheight},center}
\begin{tabular}{@{}c l *{14}{d}@{}}
\toprule
Shots & Method & \multicolumn{2}{c}{Selection} & ImageNet & SUN & Aircr. & EuroSAT & Cars & Food & Pets & Flow. & Caltech & DTD & UCF & Avg. \\ 
\cmidrule(lr){3-4}
 &  & Rand & Impt &  &  &  &  &  &  &  &  &  &  &  &  \\ \midrule
\multirow{1}{*}{0} & CLIP &  &  & 73.5 & 67.7 & 30.0 & 55.1 & 76.8 & 90.3 & 93.1 & 76.2 & 95.2 & 52.4 & 73.7 & 71.3 \\
\midrule
\multirow{4}{*}{1} & SignSparseSGD & $\checkmark$ & $\times$ & 76.2 $\pm$ 0.1 & 72.1 $\pm$ 0.2 & 36.3 $\pm$ 0.4 & 65.3 $\pm$ 2.6 & 79.5 $\pm$ 0.0 & 91.0 $\pm$ 0.1 & 94.2 $\pm$ 0.3 & 81.5 $\pm$ 0.5 & 95.6 $\pm$ 0.3 & 56.0 $\pm$ 0.2 & 77.3 $\pm$ 0.2 & 75.0 $\pm$ 0.4 \\
& SignSparseSGD & $\times$ & $\checkmark$ & 74.6 $\pm$ 0.5 & 73.3 $\pm$ 0.2 & 40.5 $\pm$ 1.7 & 71.7 $\pm$ 3.7 & 80.0 $\pm$ 0.3 & 90.2 $\pm$ 0.1 & 94.6 $\pm$ 0.2 & 91.5 $\pm$ 1.3 & 95.5 $\pm$ 0.3 & 61.7 $\pm$ 1.1 & 82.8 $\pm$ 0.5 & 77.8 $\pm$ 0.9 \\
\cmidrule(lr){2-16}
& DynSignSparseSGD & $\checkmark$ & $\times$ & 76.2 $\pm$ 0.1 & 72.2 $\pm$ 0.2 & 36.2 $\pm$ 0.4 & 65.3 $\pm$ 2.5 & 79.4 $\pm$ 0.0 & 91.1 $\pm$ 0.1 & 94.2 $\pm$ 0.2 & 81.6 $\pm$ 0.5 & 95.6 $\pm$ 0.3 & 56.1 $\pm$ 0.3 & 77.3 $\pm$ 0.3 & 75.0 $\pm$ 0.4 \\
& DynSignSparseSGD & $\times$ & $\checkmark$ & 75.6 $\pm$ 0.2 & 74.1 $\pm$ 0.2 & 41.3 $\pm$ 0.9 & 73.8 $\pm$ 4.1 & 80.9 $\pm$ 0.3 & 90.7 $\pm$ 0.2 & 94.7 $\pm$ 0.3 & 92.3 $\pm$ 1.0 & 95.5 $\pm$ 0.2 & 61.5 $\pm$ 0.7 & 82.9 $\pm$ 0.8 & 78.5 $\pm$ 0.8 \\
\midrule
\multirow{4}{*}{2} & SignSparseSGD & $\checkmark$ & $\times$ & 77.1 $\pm$ 0.0 & 74.0 $\pm$ 0.1 & 38.9 $\pm$ 0.5 & 68.3 $\pm$ 0.5 & 80.6 $\pm$ 0.1 & 91.3 $\pm$ 0.2 & 94.6 $\pm$ 0.1 & 83.9 $\pm$ 0.7 & 96.3 $\pm$ 0.4 & 59.3 $\pm$ 0.2 & 79.2 $\pm$ 0.3 & 76.7 $\pm$ 0.3 \\
& SignSparseSGD & $\times$ & $\checkmark$ & 75.3 $\pm$ 0.2 & 74.9 $\pm$ 0.1 & 41.9 $\pm$ 1.9 & 84.4 $\pm$ 0.7 & 82.2 $\pm$ 0.4 & 90.5 $\pm$ 0.2 & 94.3 $\pm$ 0.4 & 94.3 $\pm$ 1.0 & 97.0 $\pm$ 0.3 & 67.6 $\pm$ 1.2 & 84.7 $\pm$ 0.7 & 80.6 $\pm$ 0.6 \\
\cmidrule(lr){2-16}
& DynSignSparseSGD & $\checkmark$ & $\times$ & 77.1 $\pm$ 0.0 & 73.9 $\pm$ 0.1 & 39.0 $\pm$ 0.6 & 68.4 $\pm$ 0.5 & 80.6 $\pm$ 0.1 & 91.3 $\pm$ 0.2 & 94.6 $\pm$ 0.2 & 83.9 $\pm$ 0.6 & 96.3 $\pm$ 0.3 & 59.4 $\pm$ 0.2 & 79.1 $\pm$ 0.2 & 76.7 $\pm$ 0.3 \\
& DynSignSparseSGD & $\times$ & $\checkmark$ & 75.9 $\pm$ 0.3 & 75.6 $\pm$ 0.3 & 43.9 $\pm$ 2.6 & 82.6 $\pm$ 1.0 & 83.2 $\pm$ 0.6 & 90.8 $\pm$ 0.2 & 94.5 $\pm$ 0.3 & 95.2 $\pm$ 0.7 & 96.8 $\pm$ 0.6 & 68.5 $\pm$ 1.2 & 85.0 $\pm$ 0.6 & 81.1 $\pm$ 0.8 \\
\midrule
\multirow{4}{*}{4} & SignSparseSGD & $\checkmark$ & $\times$ & 78.0 $\pm$ 0.0 & 75.9 $\pm$ 0.0 & 42.2 $\pm$ 0.4 & 69.7 $\pm$ 0.1 & 82.6 $\pm$ 0.1 & 91.6 $\pm$ 0.0 & 95.2 $\pm$ 0.3 & 89.3 $\pm$ 0.2 & 96.7 $\pm$ 0.3 & 62.1 $\pm$ 0.1 & 81.9 $\pm$ 0.3 & 78.6 $\pm$ 0.2 \\
& SignSparseSGD & $\times$ & $\checkmark$ & 76.1 $\pm$ 0.2 & 76.2 $\pm$ 0.1 & 47.5 $\pm$ 0.1 & 85.0 $\pm$ 3.0 & 84.7 $\pm$ 0.7 & 90.6 $\pm$ 0.1 & 94.9 $\pm$ 0.0 & 96.5 $\pm$ 0.8 & 97.3 $\pm$ 0.1 & 69.4 $\pm$ 0.2 & 87.4 $\pm$ 1.0 & 82.3 $\pm$ 0.6 \\
\cmidrule(lr){2-16}
& DynSignSparseSGD & $\checkmark$ & $\times$ & 78.1 $\pm$ 0.1 & 75.9 $\pm$ 0.1 & 42.3 $\pm$ 0.4 & 69.8 $\pm$ 0.2 & 82.6 $\pm$ 0.2 & 91.6 $\pm$ 0.0 & 95.2 $\pm$ 0.3 & 89.3 $\pm$ 0.2 & 96.7 $\pm$ 0.3 & 62.1 $\pm$ 0.2 & 81.9 $\pm$ 0.4 & 78.7 $\pm$ 0.2 \\
& DynSignSparseSGD & $\times$ & $\checkmark$ & 76.6 $\pm$ 0.1 & 77.0 $\pm$ 0.1 & 50.5 $\pm$ 0.6 & 85.4 $\pm$ 2.5 & 86.0 $\pm$ 0.9 & 91.1 $\pm$ 0.2 & 95.0 $\pm$ 0.1 & 97.0 $\pm$ 1.0 & 97.5 $\pm$ 0.3 & 70.8 $\pm$ 0.6 & 87.6 $\pm$ 0.7 & 83.1 $\pm$ 0.6 \\
\midrule
\multirow{4}{*}{8} & SignSparseSGD & $\checkmark$ & $\times$ & 78.8 $\pm$ 0.2 & 77.9 $\pm$ 0.1 & 46.4 $\pm$ 0.4 & 77.6 $\pm$ 0.8 & 84.9 $\pm$ 0.2 & 92.1 $\pm$ 0.0 & 95.2 $\pm$ 0.2 & 95.0 $\pm$ 0.2 & 97.4 $\pm$ 0.3 & 67.1 $\pm$ 0.8 & 84.9 $\pm$ 0.0 & 81.6 $\pm$ 0.3 \\
& SignSparseSGD & $\times$ & $\checkmark$ & 76.8 $\pm$ 0.2 & 77.5 $\pm$ 0.0 & 56.0 $\pm$ 1.2 & 89.2 $\pm$ 2.7 & 87.6 $\pm$ 0.3 & 90.7 $\pm$ 0.2 & 95.4 $\pm$ 0.2 & 97.3 $\pm$ 0.3 & 97.5 $\pm$ 0.6 & 72.0 $\pm$ 0.6 & 88.6 $\pm$ 0.5 & 84.4 $\pm$ 0.6 \\
\cmidrule(lr){2-16}
& DynSignSparseSGD & $\checkmark$ & $\times$ & 79.0 $\pm$ 0.1 & 78.0 $\pm$ 0.1 & 46.6 $\pm$ 0.2 & 77.7 $\pm$ 0.9 & 84.9 $\pm$ 0.2 & 92.0 $\pm$ 0.1 & 95.2 $\pm$ 0.3 & 95.1 $\pm$ 0.2 & 97.4 $\pm$ 0.3 & 67.4 $\pm$ 1.0 & 84.8 $\pm$ 0.2 & 81.6 $\pm$ 0.3 \\
& DynSignSparseSGD & $\times$ & $\checkmark$ & 77.3 $\pm$ 0.1 & 78.3 $\pm$ 0.1 & 58.7 $\pm$ 0.6 & 88.9 $\pm$ 2.7 & 88.8 $\pm$ 0.2 & 91.2 $\pm$ 0.1 & 95.5 $\pm$ 0.3 & 97.8 $\pm$ 0.5 & 97.4 $\pm$ 0.5 & 73.1 $\pm$ 0.4 & 89.2 $\pm$ 0.4 & 85.1 $\pm$ 0.5 \\
\midrule
\multirow{4}{*}{16} & SignSparseSGD & $\checkmark$ & $\times$ & 79.8 $\pm$ 0.0 & 79.9 $\pm$ 0.1 & 52.1 $\pm$ 0.4 & 84.9 $\pm$ 0.8 & 87.6 $\pm$ 0.2 & 92.2 $\pm$ 0.0 & 95.7 $\pm$ 0.1 & 98.0 $\pm$ 0.2 & 97.5 $\pm$ 0.0 & 73.1 $\pm$ 0.4 & 87.6 $\pm$ 0.2 & 84.4 $\pm$ 0.2 \\
& SignSparseSGD & $\times$ & $\checkmark$ & 78.0 $\pm$ 0.1 & 79.3 $\pm$ 0.3 & 61.6 $\pm$ 0.9 & 92.7 $\pm$ 1.5 & 89.5 $\pm$ 0.3 & 90.8 $\pm$ 0.2 & 95.4 $\pm$ 0.3 & 98.5 $\pm$ 0.1 & 97.7 $\pm$ 0.0 & 76.5 $\pm$ 0.8 & 90.1 $\pm$ 0.2 & 86.4 $\pm$ 0.4 \\
\cmidrule(lr){2-16}
& DynSignSparseSGD & $\checkmark$ & $\times$ & 80.0 $\pm$ 0.0 & 80.0 $\pm$ 0.0 & 52.1 $\pm$ 0.4 & 85.0 $\pm$ 0.9 & 87.6 $\pm$ 0.1 & 92.2 $\pm$ 0.1 & 95.7 $\pm$ 0.1 & 98.0 $\pm$ 0.1 & 97.5 $\pm$ 0.1 & 73.2 $\pm$ 0.3 & 87.5 $\pm$ 0.3 & 84.4 $\pm$ 0.2 \\
& DynSignSparseSGD & $\times$ & $\checkmark$ & 78.3 $\pm$ 0.1 & 79.6 $\pm$ 0.1 & 65.0 $\pm$ 0.3 & 93.4 $\pm$ 0.9 & 90.3 $\pm$ 0.1 & 91.2 $\pm$ 0.0 & 95.6 $\pm$ 0.1 & 98.9 $\pm$ 0.2 & 97.4 $\pm$ 0.3 & 77.8 $\pm$ 0.7 & 91.0 $\pm$ 0.2 & 87.1 $\pm$ 0.3 \\
\bottomrule
\end{tabular}
\end{adjustbox}
\end{table}

Tables~\ref{tab:ablation_selection_ViTB32_with_aug}, \ref{tab:ablation_selection_ViTB16_with_aug}, and \ref{tab:ablation_selection_ViTL14_with_aug} isolate the effect of the support-selection rule in the sign-sparse optimizers. Across all three CLIP backbones, all shot levels, and both SignSparseSGD and DynSignSparseSGD, importance-based selection (i.e., selecting the top-$k$ coordinates by gradient magnitude) consistently outperforms random selection. Averaged over the 15 backbone--shot configurations, importance-based selection improves SignSparseSGD by 3.3 points and DynSignSparseSGD by 4.4 points. The gain ranges from 2.0 to 4.5 points for SignSparseSGD and from 2.7 to 6.0 points for DynSignSparseSGD, and it remains positive at 16 shots on every backbone. DynSignSparseSGD with importance-based selection is the best variant in all 15 backbone--shot configurations.

The comparison between static and dynamic masks reveals a second pattern. Under random selection, DynSignSparseSGD is nearly identical to SignSparseSGD. The average gap is below 0.1 points and never exceeds 0.2 points. Re-sampling a random sparse support, therefore, provides almost no benefit. Under importance-based selection, the dynamic mask consistently helps. DynSignSparseSGD improves over SignSparseSGD in all 15 cases, with gains between 0.5 and 1.8 points and an average gain of 1.2 points. The main effect, therefore, comes from choosing informative coordinates under a fixed sparsity budget, not from re-sampling a random support.

These results support the interpretation in the main paper. Gradient magnitude is useful for deciding where to update, but not for deciding how much to update. This is consistent with Fig.~\ref{fig:gradient_diagnostics}(c)--(d), where sign agreement is concentrated in the top-$|g|$ coordinates. Random sparsification alone is not sufficient. Strong performance is obtained when sparse sign updates are focused on large-gradient parameters, and it improves further when the active set is refreshed during training.

\section{Selection Scope}
\label{app:selection_scope}
\label{app:local_vs_global_selection}
\suppanchor{selection-scope}

\begin{table}
\caption{Effect of sparse-support selection scope for CLIP (ViT-B/32) with data augmentation. Tensor-wise selection applies $\Top_{k_r}$ independently to each trainable parameter tensor, as in Eq.~\eqref{eq:block_budgets}; model-wise selection applies one $\Top_k$ operation to the concatenated trainable parameters. Top-1 accuracy (\%) averaged over three fixed random seeds.}
\label{tab:ablation_selection_scope_ViTB32_with_aug}
\centering
\footnotesize
\setlength{\tabcolsep}{1.8pt}
\renewcommand{\arraystretch}{1.00}
\begin{adjustbox}{max totalsize={\textwidth}{0.95\textheight},center}
\begin{tabular}{@{}c l *{14}{d}@{}}
\toprule
Shots & Method & \multicolumn{2}{c}{Selection scope} & ImageNet & SUN & Aircraft & EuroSAT & Cars & Food & Pets & Flowers & Caltech & DTD & UCF & Average \\
\cmidrule(lr){3-4}
 &  & Tensor-wise & Model-wise & ImageNet & SUN & Aircraft & EuroSAT & Cars & Food & Pets & Flowers & Caltech & DTD & UCF & Average \\ \midrule
\multirow{1}{*}{0} & CLIP &  &  & 62.0 & 62.1 & 18.1 & 35.8 & 60.1 & 78.9 & 85.1 & 64.0 & 91.4 & 43.0 & 61.6 & 60.2 \\
\midrule
\multirow{4}{*}{1} & SignSparseSGD & $\checkmark$ & $\times$ & 60.6 $\pm$ 0.6 & 65.3 $\pm$ 0.1 & 21.6 $\pm$ 0.5 & 66.0 $\pm$ 5.2 & 61.3 $\pm$ 1.1 & 77.9 $\pm$ 0.5 & 88.5 $\pm$ 0.4 & 75.0 $\pm$ 0.8 & 92.3 $\pm$ 0.5 & 50.9 $\pm$ 1.5 & 71.5 $\pm$ 0.3 & 66.4 $\pm$ 1.0 \\
& SignSparseSGD & $\times$ & $\checkmark$ & 64.9 $\pm$ 0.3 & 67.4 $\pm$ 0.1 & 21.8 $\pm$ 0.2 & 55.8 $\pm$ 1.7 & 62.7 $\pm$ 0.2 & 80.9 $\pm$ 0.3 & 89.7 $\pm$ 0.3 & 72.0 $\pm$ 0.8 & 92.9 $\pm$ 0.2 & 49.9 $\pm$ 0.5 & 68.2 $\pm$ 0.8 & 66.0 $\pm$ 0.5 \\
\cmidrule(lr){2-16}
& DynSignSparseSGD & $\checkmark$ & $\times$ & 62.6 $\pm$ 0.3 & 67.0 $\pm$ 0.3 & 22.7 $\pm$ 0.8 & 69.1 $\pm$ 2.7 & 63.1 $\pm$ 0.4 & 79.7 $\pm$ 0.5 & 89.1 $\pm$ 0.2 & 77.7 $\pm$ 1.2 & 92.6 $\pm$ 0.7 & 51.9 $\pm$ 1.0 & 72.0 $\pm$ 0.4 & 67.9 $\pm$ 0.8 \\
& DynSignSparseSGD & $\times$ & $\checkmark$ & 60.5 $\pm$ 0.3 & 65.8 $\pm$ 0.3 & 21.6 $\pm$ 0.7 & 70.0 $\pm$ 2.4 & 62.8 $\pm$ 0.6 & 78.9 $\pm$ 0.6 & 86.9 $\pm$ 0.7 & 79.0 $\pm$ 1.3 & 92.6 $\pm$ 0.6 & 47.6 $\pm$ 1.1 & 71.6 $\pm$ 0.4 & 67.0 $\pm$ 0.8 \\
\midrule
\multirow{4}{*}{2} & SignSparseSGD & $\checkmark$ & $\times$ & 61.9 $\pm$ 0.3 & 66.9 $\pm$ 0.2 & 24.3 $\pm$ 0.6 & 79.1 $\pm$ 6.7 & 63.6 $\pm$ 0.1 & 78.0 $\pm$ 0.4 & 89.1 $\pm$ 0.4 & 78.7 $\pm$ 0.2 & 93.8 $\pm$ 0.3 & 56.4 $\pm$ 0.4 & 74.1 $\pm$ 0.8 & 69.6 $\pm$ 0.9 \\
& SignSparseSGD & $\times$ & $\checkmark$ & 65.4 $\pm$ 0.1 & 69.0 $\pm$ 0.2 & 23.2 $\pm$ 0.5 & 64.4 $\pm$ 4.1 & 64.3 $\pm$ 0.2 & 81.4 $\pm$ 0.1 & 90.0 $\pm$ 0.8 & 74.9 $\pm$ 0.6 & 93.6 $\pm$ 0.4 & 55.9 $\pm$ 0.9 & 70.5 $\pm$ 0.4 & 68.4 $\pm$ 0.7 \\
\cmidrule(lr){2-16}
& DynSignSparseSGD & $\checkmark$ & $\times$ & 63.2 $\pm$ 0.1 & 68.2 $\pm$ 0.3 & 25.5 $\pm$ 0.6 & 79.8 $\pm$ 4.2 & 66.0 $\pm$ 0.1 & 79.5 $\pm$ 0.4 & 89.2 $\pm$ 0.3 & 83.8 $\pm$ 1.4 & 94.0 $\pm$ 0.3 & 57.1 $\pm$ 1.1 & 74.8 $\pm$ 0.6 & 71.0 $\pm$ 0.9 \\
& DynSignSparseSGD & $\times$ & $\checkmark$ & 60.9 $\pm$ 0.2 & 67.3 $\pm$ 0.2 & 24.7 $\pm$ 0.9 & 78.6 $\pm$ 5.3 & 65.3 $\pm$ 0.2 & 78.6 $\pm$ 0.6 & 87.5 $\pm$ 0.9 & 84.8 $\pm$ 0.8 & 94.0 $\pm$ 0.2 & 55.0 $\pm$ 0.9 & 74.4 $\pm$ 0.6 & 70.1 $\pm$ 1.0 \\
\midrule
\multirow{4}{*}{4} & SignSparseSGD & $\checkmark$ & $\times$ & 63.3 $\pm$ 0.4 & 69.0 $\pm$ 0.4 & 25.2 $\pm$ 1.1 & 82.6 $\pm$ 1.0 & 66.9 $\pm$ 0.3 & 78.2 $\pm$ 0.4 & 89.3 $\pm$ 0.4 & 85.8 $\pm$ 1.2 & 94.7 $\pm$ 0.4 & 59.1 $\pm$ 1.0 & 75.9 $\pm$ 0.7 & 71.8 $\pm$ 0.7 \\
& SignSparseSGD & $\times$ & $\checkmark$ & 66.4 $\pm$ 0.2 & 70.8 $\pm$ 0.1 & 25.0 $\pm$ 0.4 & 66.0 $\pm$ 0.9 & 66.2 $\pm$ 0.2 & 81.6 $\pm$ 0.3 & 90.9 $\pm$ 0.1 & 81.4 $\pm$ 0.4 & 94.4 $\pm$ 0.3 & 57.7 $\pm$ 0.4 & 73.3 $\pm$ 0.3 & 70.3 $\pm$ 0.3 \\
\cmidrule(lr){2-16}
& DynSignSparseSGD & $\checkmark$ & $\times$ & 64.0 $\pm$ 0.2 & 69.9 $\pm$ 0.3 & 27.7 $\pm$ 0.6 & 83.9 $\pm$ 1.6 & 69.3 $\pm$ 0.1 & 79.6 $\pm$ 0.4 & 90.0 $\pm$ 0.4 & 88.9 $\pm$ 0.8 & 95.2 $\pm$ 0.2 & 61.1 $\pm$ 0.3 & 78.0 $\pm$ 0.9 & 73.4 $\pm$ 0.5 \\
& DynSignSparseSGD & $\times$ & $\checkmark$ & 62.3 $\pm$ 0.3 & 68.8 $\pm$ 0.1 & 28.0 $\pm$ 0.8 & 84.7 $\pm$ 1.7 & 69.1 $\pm$ 0.3 & 78.8 $\pm$ 0.4 & 88.8 $\pm$ 0.2 & 89.8 $\pm$ 1.5 & 95.0 $\pm$ 0.2 & 60.1 $\pm$ 0.1 & 77.8 $\pm$ 0.7 & 73.0 $\pm$ 0.6 \\
\midrule
\multirow{4}{*}{8} & SignSparseSGD & $\checkmark$ & $\times$ & 64.7 $\pm$ 0.1 & 70.9 $\pm$ 0.3 & 31.1 $\pm$ 0.8 & 85.9 $\pm$ 0.5 & 71.4 $\pm$ 0.5 & 79.0 $\pm$ 0.2 & 90.3 $\pm$ 0.1 & 89.3 $\pm$ 0.7 & 94.9 $\pm$ 0.3 & 62.3 $\pm$ 0.9 & 78.8 $\pm$ 0.7 & 74.4 $\pm$ 0.5 \\
& SignSparseSGD & $\times$ & $\checkmark$ & 67.4 $\pm$ 0.1 & 72.5 $\pm$ 0.2 & 28.3 $\pm$ 0.1 & 76.5 $\pm$ 1.5 & 68.8 $\pm$ 0.2 & 81.9 $\pm$ 0.0 & 91.6 $\pm$ 0.3 & 87.4 $\pm$ 0.6 & 94.8 $\pm$ 0.2 & 61.2 $\pm$ 0.3 & 76.3 $\pm$ 0.1 & 73.3 $\pm$ 0.3 \\
\cmidrule(lr){2-16}
& DynSignSparseSGD & $\checkmark$ & $\times$ & 65.0 $\pm$ 0.2 & 71.9 $\pm$ 0.1 & 34.8 $\pm$ 0.6 & 88.4 $\pm$ 1.0 & 75.1 $\pm$ 0.2 & 79.8 $\pm$ 0.1 & 91.0 $\pm$ 0.3 & 92.5 $\pm$ 0.1 & 95.5 $\pm$ 0.3 & 64.5 $\pm$ 1.0 & 79.7 $\pm$ 0.6 & 76.2 $\pm$ 0.4 \\
& DynSignSparseSGD & $\times$ & $\checkmark$ & 63.6 $\pm$ 0.3 & 71.2 $\pm$ 0.1 & 34.5 $\pm$ 0.5 & 87.8 $\pm$ 0.3 & 75.2 $\pm$ 0.1 & 79.3 $\pm$ 0.0 & 90.0 $\pm$ 0.5 & 92.7 $\pm$ 0.1 & 95.3 $\pm$ 0.3 & 63.9 $\pm$ 0.5 & 79.7 $\pm$ 0.5 & 75.7 $\pm$ 0.3 \\
\midrule
\multirow{4}{*}{16} & SignSparseSGD & $\checkmark$ & $\times$ & 66.3 $\pm$ 0.2 & 73.1 $\pm$ 0.3 & 36.4 $\pm$ 0.9 & 90.7 $\pm$ 1.6 & 76.4 $\pm$ 0.6 & 79.3 $\pm$ 0.5 & 89.9 $\pm$ 1.0 & 93.8 $\pm$ 1.0 & 95.8 $\pm$ 0.0 & 67.5 $\pm$ 0.2 & 81.7 $\pm$ 0.4 & 77.4 $\pm$ 0.6 \\
& SignSparseSGD & $\times$ & $\checkmark$ & 68.4 $\pm$ 0.1 & 74.2 $\pm$ 0.0 & 31.8 $\pm$ 0.1 & 83.0 $\pm$ 0.5 & 72.3 $\pm$ 0.3 & 82.4 $\pm$ 0.2 & 91.9 $\pm$ 0.2 & 91.9 $\pm$ 0.5 & 95.4 $\pm$ 0.1 & 66.8 $\pm$ 0.5 & 79.3 $\pm$ 0.5 & 76.1 $\pm$ 0.3 \\
\cmidrule(lr){2-16}
& DynSignSparseSGD & $\checkmark$ & $\times$ & 66.9 $\pm$ 0.2 & 73.5 $\pm$ 0.1 & 42.0 $\pm$ 1.1 & 92.0 $\pm$ 0.6 & 79.8 $\pm$ 0.4 & 80.3 $\pm$ 0.4 & 91.0 $\pm$ 0.3 & 95.7 $\pm$ 1.0 & 95.8 $\pm$ 0.0 & 69.2 $\pm$ 0.5 & 82.7 $\pm$ 0.5 & 79.0 $\pm$ 0.5 \\
& DynSignSparseSGD & $\times$ & $\checkmark$ & 66.1 $\pm$ 0.3 & 73.2 $\pm$ 0.1 & 41.9 $\pm$ 1.1 & 92.5 $\pm$ 0.5 & 79.7 $\pm$ 0.6 & 79.9 $\pm$ 0.4 & 90.6 $\pm$ 0.2 & 95.4 $\pm$ 0.9 & 95.8 $\pm$ 0.1 & 68.6 $\pm$ 0.6 & 82.5 $\pm$ 0.4 & 78.7 $\pm$ 0.5 \\
\bottomrule
\end{tabular}
\end{adjustbox}
\end{table}

\begin{table}
\caption{Effect of sparse-support selection scope for CLIP (ViT-B/16) with data augmentation. Tensor-wise selection applies $\Top_{k_r}$ independently to each trainable parameter tensor, as in Eq.~\eqref{eq:block_budgets}; model-wise selection applies one $\Top_k$ operation to the concatenated trainable parameters. Top-1 accuracy (\%) averaged over three fixed random seeds.}
\label{tab:ablation_selection_scope_ViTB16_with_aug}
\centering
\footnotesize
\setlength{\tabcolsep}{1.8pt}
\renewcommand{\arraystretch}{1.00}
\begin{adjustbox}{max totalsize={\textwidth}{0.95\textheight},center}
\begin{tabular}{@{}c l *{14}{d}@{}}
\toprule
Shots & Method & \multicolumn{2}{c}{Selection scope} & ImageNet & SUN & Aircraft & EuroSAT & Cars & Food & Pets & Flowers & Caltech & DTD & UCF & Average \\
\cmidrule(lr){3-4}
 &  & Tensor-wise & Model-wise & ImageNet & SUN & Aircraft & EuroSAT & Cars & Food & Pets & Flowers & Caltech & DTD & UCF & Average \\ \midrule
\multirow{1}{*}{0} & CLIP &  &  & 66.7 & 62.6 & 23.9 & 42.2 & 65.5 & 85.2 & 88.3 & 67.4 & 93.3 & 44.4 & 65.2 & 64.1 \\
\midrule
\multirow{4}{*}{1} & SignSparseSGD & $\checkmark$ & $\times$ & 66.8 $\pm$ 0.1 & 68.2 $\pm$ 0.3 & 29.4 $\pm$ 1.3 & 71.2 $\pm$ 3.7 & 69.1 $\pm$ 0.6 & 83.8 $\pm$ 0.5 & 91.3 $\pm$ 0.3 & 83.3 $\pm$ 0.5 & 93.7 $\pm$ 0.2 & 54.3 $\pm$ 0.6 & 75.7 $\pm$ 0.3 & 71.5 $\pm$ 0.8 \\
& SignSparseSGD & $\times$ & $\checkmark$ & 70.1 $\pm$ 0.1 & 70.1 $\pm$ 0.2 & 29.6 $\pm$ 0.6 & 67.0 $\pm$ 1.5 & 69.0 $\pm$ 0.1 & 86.2 $\pm$ 0.1 & 92.6 $\pm$ 0.1 & 78.2 $\pm$ 0.6 & 94.2 $\pm$ 0.2 & 52.6 $\pm$ 0.4 & 72.8 $\pm$ 0.1 & 71.1 $\pm$ 0.4 \\
\cmidrule(lr){2-16}
& DynSignSparseSGD & $\checkmark$ & $\times$ & 68.1 $\pm$ 0.1 & 69.8 $\pm$ 0.3 & 30.6 $\pm$ 1.2 & 73.7 $\pm$ 4.0 & 70.6 $\pm$ 0.4 & 85.1 $\pm$ 0.3 & 92.5 $\pm$ 0.2 & 84.7 $\pm$ 0.4 & 93.8 $\pm$ 0.1 & 54.8 $\pm$ 0.6 & 76.2 $\pm$ 0.4 & 72.7 $\pm$ 0.7 \\
& DynSignSparseSGD & $\times$ & $\checkmark$ & 66.5 $\pm$ 0.2 & 68.8 $\pm$ 0.4 & 29.4 $\pm$ 1.4 & 74.6 $\pm$ 2.3 & 69.8 $\pm$ 0.4 & 84.6 $\pm$ 0.2 & 91.4 $\pm$ 0.2 & 86.5 $\pm$ 0.7 & 93.7 $\pm$ 0.2 & 53.4 $\pm$ 0.2 & 75.7 $\pm$ 0.9 & 72.2 $\pm$ 0.6 \\
\midrule
\multirow{4}{*}{2} & SignSparseSGD & $\checkmark$ & $\times$ & 67.4 $\pm$ 0.4 & 69.7 $\pm$ 0.1 & 32.0 $\pm$ 1.4 & 79.5 $\pm$ 2.2 & 71.7 $\pm$ 0.2 & 84.1 $\pm$ 0.3 & 92.0 $\pm$ 0.3 & 87.2 $\pm$ 0.5 & 94.2 $\pm$ 0.3 & 59.5 $\pm$ 1.2 & 78.4 $\pm$ 1.0 & 74.2 $\pm$ 0.7 \\
& SignSparseSGD & $\times$ & $\checkmark$ & 70.7 $\pm$ 0.1 & 71.6 $\pm$ 0.2 & 31.8 $\pm$ 1.1 & 71.9 $\pm$ 2.2 & 70.6 $\pm$ 0.0 & 86.5 $\pm$ 0.1 & 92.9 $\pm$ 0.3 & 81.7 $\pm$ 0.4 & 94.3 $\pm$ 0.3 & 57.3 $\pm$ 1.6 & 75.1 $\pm$ 0.0 & 73.1 $\pm$ 0.6 \\
\cmidrule(lr){2-16}
& DynSignSparseSGD & $\checkmark$ & $\times$ & 68.6 $\pm$ 0.2 & 71.0 $\pm$ 0.1 & 33.8 $\pm$ 2.1 & 82.0 $\pm$ 1.4 & 73.7 $\pm$ 0.3 & 85.1 $\pm$ 0.1 & 91.7 $\pm$ 0.4 & 89.6 $\pm$ 0.2 & 94.3 $\pm$ 0.4 & 60.1 $\pm$ 1.2 & 79.7 $\pm$ 0.9 & 75.4 $\pm$ 0.7 \\
& DynSignSparseSGD & $\times$ & $\checkmark$ & 67.1 $\pm$ 0.3 & 70.0 $\pm$ 0.3 & 33.5 $\pm$ 2.0 & 82.3 $\pm$ 3.1 & 73.6 $\pm$ 0.3 & 84.5 $\pm$ 0.1 & 90.9 $\pm$ 0.1 & 90.3 $\pm$ 0.2 & 94.2 $\pm$ 0.3 & 59.7 $\pm$ 1.3 & 78.8 $\pm$ 0.5 & 75.0 $\pm$ 0.8 \\
\midrule
\multirow{4}{*}{4} & SignSparseSGD & $\checkmark$ & $\times$ & 68.5 $\pm$ 0.3 & 71.9 $\pm$ 0.2 & 37.3 $\pm$ 0.7 & 81.5 $\pm$ 1.2 & 75.7 $\pm$ 0.7 & 84.7 $\pm$ 0.3 & 92.5 $\pm$ 0.2 & 91.4 $\pm$ 1.1 & 95.4 $\pm$ 0.1 & 62.6 $\pm$ 0.7 & 81.1 $\pm$ 1.1 & 76.6 $\pm$ 0.6 \\
& SignSparseSGD & $\times$ & $\checkmark$ & 71.5 $\pm$ 0.0 & 73.4 $\pm$ 0.1 & 34.1 $\pm$ 0.4 & 74.5 $\pm$ 0.6 & 73.2 $\pm$ 0.5 & 86.8 $\pm$ 0.1 & 93.4 $\pm$ 0.2 & 87.9 $\pm$ 0.5 & 94.8 $\pm$ 0.2 & 59.6 $\pm$ 0.8 & 77.3 $\pm$ 0.2 & 75.1 $\pm$ 0.3 \\
\cmidrule(lr){2-16}
& DynSignSparseSGD & $\checkmark$ & $\times$ & 69.4 $\pm$ 0.2 & 72.8 $\pm$ 0.1 & 38.4 $\pm$ 1.1 & 84.1 $\pm$ 2.0 & 77.8 $\pm$ 0.3 & 85.4 $\pm$ 0.2 & 92.8 $\pm$ 0.3 & 93.2 $\pm$ 0.8 & 95.2 $\pm$ 0.1 & 64.5 $\pm$ 0.7 & 82.2 $\pm$ 1.2 & 77.8 $\pm$ 0.6 \\
& DynSignSparseSGD & $\times$ & $\checkmark$ & 68.2 $\pm$ 0.2 & 71.9 $\pm$ 0.1 & 38.4 $\pm$ 1.3 & 84.0 $\pm$ 2.0 & 77.7 $\pm$ 0.5 & 84.8 $\pm$ 0.2 & 92.6 $\pm$ 0.4 & 93.8 $\pm$ 0.6 & 95.1 $\pm$ 0.3 & 63.4 $\pm$ 1.0 & 81.6 $\pm$ 1.1 & 77.4 $\pm$ 0.7 \\
\midrule
\multirow{4}{*}{8} & SignSparseSGD & $\checkmark$ & $\times$ & 69.9 $\pm$ 0.3 & 73.7 $\pm$ 0.1 & 41.1 $\pm$ 1.7 & 84.7 $\pm$ 5.2 & 79.8 $\pm$ 0.7 & 84.8 $\pm$ 0.6 & 92.9 $\pm$ 0.4 & 94.0 $\pm$ 0.3 & 95.9 $\pm$ 0.2 & 66.5 $\pm$ 0.7 & 83.6 $\pm$ 0.6 & 78.8 $\pm$ 1.0 \\
& SignSparseSGD & $\times$ & $\checkmark$ & 72.5 $\pm$ 0.2 & 74.7 $\pm$ 0.2 & 36.8 $\pm$ 0.2 & 79.0 $\pm$ 0.9 & 76.3 $\pm$ 0.4 & 87.2 $\pm$ 0.1 & 93.7 $\pm$ 0.2 & 92.6 $\pm$ 0.3 & 95.2 $\pm$ 0.2 & 64.0 $\pm$ 0.5 & 80.2 $\pm$ 0.1 & 77.5 $\pm$ 0.3 \\
\cmidrule(lr){2-16}
& DynSignSparseSGD & $\checkmark$ & $\times$ & 70.5 $\pm$ 0.4 & 74.5 $\pm$ 0.2 & 45.2 $\pm$ 0.7 & 85.2 $\pm$ 4.0 & 82.4 $\pm$ 0.5 & 85.3 $\pm$ 0.3 & 93.4 $\pm$ 0.1 & 95.0 $\pm$ 0.2 & 95.8 $\pm$ 0.1 & 68.0 $\pm$ 1.3 & 84.7 $\pm$ 0.2 & 80.0 $\pm$ 0.7 \\
& DynSignSparseSGD & $\times$ & $\checkmark$ & 69.4 $\pm$ 0.3 & 73.8 $\pm$ 0.1 & 45.1 $\pm$ 0.5 & 85.8 $\pm$ 4.0 & 82.2 $\pm$ 0.2 & 85.0 $\pm$ 0.2 & 92.9 $\pm$ 0.1 & 95.7 $\pm$ 0.3 & 95.7 $\pm$ 0.4 & 67.2 $\pm$ 0.7 & 84.2 $\pm$ 0.3 & 79.7 $\pm$ 0.6 \\
\midrule
\multirow{4}{*}{16} & SignSparseSGD & $\checkmark$ & $\times$ & 71.6 $\pm$ 0.1 & 75.0 $\pm$ 0.2 & 45.6 $\pm$ 0.9 & 91.2 $\pm$ 1.1 & 83.6 $\pm$ 0.5 & 85.7 $\pm$ 0.1 & 93.6 $\pm$ 0.3 & 96.8 $\pm$ 0.2 & 96.2 $\pm$ 0.2 & 71.2 $\pm$ 0.3 & 85.8 $\pm$ 0.0 & 81.5 $\pm$ 0.4 \\
& SignSparseSGD & $\times$ & $\checkmark$ & 73.6 $\pm$ 0.1 & 76.2 $\pm$ 0.1 & 41.1 $\pm$ 0.4 & 85.5 $\pm$ 1.2 & 79.8 $\pm$ 0.2 & 87.5 $\pm$ 0.2 & 94.2 $\pm$ 0.1 & 95.8 $\pm$ 0.2 & 95.6 $\pm$ 0.2 & 68.8 $\pm$ 0.4 & 84.0 $\pm$ 0.2 & 80.2 $\pm$ 0.3 \\
\cmidrule(lr){2-16}
& DynSignSparseSGD & $\checkmark$ & $\times$ & 72.2 $\pm$ 0.1 & 75.7 $\pm$ 0.2 & 51.7 $\pm$ 0.6 & 91.8 $\pm$ 0.5 & 85.9 $\pm$ 0.3 & 86.1 $\pm$ 0.1 & 93.9 $\pm$ 0.1 & 97.7 $\pm$ 0.2 & 96.4 $\pm$ 0.3 & 72.2 $\pm$ 0.4 & 86.9 $\pm$ 0.1 & 82.8 $\pm$ 0.3 \\
& DynSignSparseSGD & $\times$ & $\checkmark$ & 71.6 $\pm$ 0.0 & 75.3 $\pm$ 0.1 & 52.6 $\pm$ 0.3 & 91.8 $\pm$ 0.6 & 86.2 $\pm$ 0.2 & 85.6 $\pm$ 0.1 & 93.8 $\pm$ 0.1 & 97.5 $\pm$ 0.2 & 96.2 $\pm$ 0.3 & 71.7 $\pm$ 0.5 & 86.2 $\pm$ 0.2 & 82.6 $\pm$ 0.2 \\
\bottomrule
\end{tabular}
\end{adjustbox}
\end{table}

\begin{table}
\caption{Effect of sparse-support selection scope for CLIP (ViT-L/14) with data augmentation. Tensor-wise selection applies $\Top_{k_r}$ independently to each trainable parameter tensor, as in Eq.~\eqref{eq:block_budgets}; model-wise selection applies one $\Top_k$ operation to the concatenated trainable parameters. Top-1 accuracy (\%) averaged over three fixed random seeds.}
\label{tab:ablation_selection_scope_ViTL14_with_aug}
\centering
\footnotesize
\setlength{\tabcolsep}{1.8pt}
\renewcommand{\arraystretch}{1.00}
\begin{adjustbox}{max totalsize={\textwidth}{0.95\textheight},center}
\begin{tabular}{@{}c l *{14}{d}@{}}
\toprule
Shots & Method & \multicolumn{2}{c}{Selection scope} & ImageNet & SUN & Aircraft & EuroSAT & Cars & Food & Pets & Flowers & Caltech & DTD & UCF & Average \\
\cmidrule(lr){3-4}
 &  & Tensor-wise & Model-wise & ImageNet & SUN & Aircraft & EuroSAT & Cars & Food & Pets & Flowers & Caltech & DTD & UCF & Average \\ \midrule
\multirow{1}{*}{0} & CLIP &  &  & 73.5 & 67.7 & 30.0 & 55.1 & 76.8 & 90.3 & 93.1 & 76.2 & 95.2 & 52.4 & 73.7 & 71.3 \\
\midrule
\multirow{4}{*}{1} & SignSparseSGD & $\checkmark$ & $\times$ & 74.6 $\pm$ 0.5 & 73.3 $\pm$ 0.2 & 40.5 $\pm$ 1.7 & 71.7 $\pm$ 3.7 & 80.0 $\pm$ 0.3 & 90.2 $\pm$ 0.1 & 94.6 $\pm$ 0.2 & 91.5 $\pm$ 1.3 & 95.5 $\pm$ 0.3 & 61.7 $\pm$ 1.1 & 82.8 $\pm$ 0.5 & 77.8 $\pm$ 0.9 \\
& SignSparseSGD & $\times$ & $\checkmark$ & 76.6 $\pm$ 0.1 & 74.1 $\pm$ 0.1 & 39.1 $\pm$ 0.4 & 70.7 $\pm$ 4.6 & 80.3 $\pm$ 0.4 & 91.0 $\pm$ 0.1 & 94.4 $\pm$ 0.2 & 86.2 $\pm$ 1.1 & 95.9 $\pm$ 0.6 & 59.3 $\pm$ 0.7 & 80.1 $\pm$ 0.1 & 77.1 $\pm$ 0.7 \\
\cmidrule(lr){2-16}
& DynSignSparseSGD & $\checkmark$ & $\times$ & 75.6 $\pm$ 0.2 & 74.1 $\pm$ 0.2 & 41.3 $\pm$ 0.9 & 73.8 $\pm$ 4.1 & 80.9 $\pm$ 0.3 & 90.7 $\pm$ 0.2 & 94.7 $\pm$ 0.3 & 92.3 $\pm$ 1.0 & 95.5 $\pm$ 0.2 & 61.5 $\pm$ 0.7 & 82.9 $\pm$ 0.8 & 78.5 $\pm$ 0.8 \\
& DynSignSparseSGD & $\times$ & $\checkmark$ & 73.0 $\pm$ 0.9 & 72.6 $\pm$ 0.4 & 39.9 $\pm$ 0.7 & 74.7 $\pm$ 4.5 & 80.3 $\pm$ 0.5 & 90.4 $\pm$ 0.2 & 93.8 $\pm$ 0.1 & 92.5 $\pm$ 0.5 & 95.4 $\pm$ 0.3 & 59.1 $\pm$ 1.4 & 82.8 $\pm$ 1.0 & 77.7 $\pm$ 1.0 \\
\midrule
\multirow{4}{*}{2} & SignSparseSGD & $\checkmark$ & $\times$ & 75.3 $\pm$ 0.2 & 74.9 $\pm$ 0.1 & 41.9 $\pm$ 1.9 & 84.4 $\pm$ 0.7 & 82.2 $\pm$ 0.4 & 90.5 $\pm$ 0.2 & 94.3 $\pm$ 0.4 & 94.3 $\pm$ 1.0 & 97.0 $\pm$ 0.3 & 67.6 $\pm$ 1.2 & 84.7 $\pm$ 0.7 & 80.6 $\pm$ 0.6 \\
& SignSparseSGD & $\times$ & $\checkmark$ & 77.3 $\pm$ 0.1 & 75.8 $\pm$ 0.1 & 41.5 $\pm$ 0.7 & 75.1 $\pm$ 1.8 & 81.7 $\pm$ 0.3 & 91.3 $\pm$ 0.1 & 94.6 $\pm$ 0.3 & 90.4 $\pm$ 0.7 & 96.5 $\pm$ 0.5 & 65.5 $\pm$ 1.0 & 81.9 $\pm$ 0.6 & 79.2 $\pm$ 0.6 \\
\cmidrule(lr){2-16}
& DynSignSparseSGD & $\checkmark$ & $\times$ & 75.9 $\pm$ 0.3 & 75.6 $\pm$ 0.3 & 43.9 $\pm$ 2.6 & 82.6 $\pm$ 1.0 & 83.2 $\pm$ 0.6 & 90.8 $\pm$ 0.2 & 94.5 $\pm$ 0.3 & 95.2 $\pm$ 0.7 & 96.8 $\pm$ 0.6 & 68.5 $\pm$ 1.2 & 85.0 $\pm$ 0.6 & 81.1 $\pm$ 0.8 \\
& DynSignSparseSGD & $\times$ & $\checkmark$ & 73.7 $\pm$ 0.9 & 74.3 $\pm$ 0.2 & 42.9 $\pm$ 2.1 & 82.4 $\pm$ 1.9 & 82.0 $\pm$ 0.6 & 90.3 $\pm$ 0.3 & 93.9 $\pm$ 0.3 & 95.6 $\pm$ 0.9 & 96.8 $\pm$ 0.4 & 66.7 $\pm$ 1.1 & 84.1 $\pm$ 1.2 & 80.2 $\pm$ 0.9 \\
\midrule
\multirow{4}{*}{4} & SignSparseSGD & $\checkmark$ & $\times$ & 76.1 $\pm$ 0.2 & 76.2 $\pm$ 0.1 & 47.5 $\pm$ 0.1 & 85.0 $\pm$ 3.0 & 84.7 $\pm$ 0.7 & 90.6 $\pm$ 0.1 & 94.9 $\pm$ 0.0 & 96.5 $\pm$ 0.8 & 97.3 $\pm$ 0.1 & 69.4 $\pm$ 0.2 & 87.4 $\pm$ 1.0 & 82.3 $\pm$ 0.6 \\
& SignSparseSGD & $\times$ & $\checkmark$ & 78.1 $\pm$ 0.0 & 77.3 $\pm$ 0.1 & 44.8 $\pm$ 0.2 & 77.9 $\pm$ 1.5 & 83.4 $\pm$ 0.3 & 91.6 $\pm$ 0.1 & 95.4 $\pm$ 0.1 & 95.3 $\pm$ 0.4 & 97.2 $\pm$ 0.4 & 67.8 $\pm$ 1.2 & 84.8 $\pm$ 0.4 & 81.2 $\pm$ 0.4 \\
\cmidrule(lr){2-16}
& DynSignSparseSGD & $\checkmark$ & $\times$ & 76.6 $\pm$ 0.1 & 77.0 $\pm$ 0.1 & 50.5 $\pm$ 0.6 & 85.4 $\pm$ 2.5 & 86.0 $\pm$ 0.9 & 91.1 $\pm$ 0.2 & 95.0 $\pm$ 0.1 & 97.0 $\pm$ 1.0 & 97.5 $\pm$ 0.3 & 70.8 $\pm$ 0.6 & 87.6 $\pm$ 0.7 & 83.1 $\pm$ 0.6 \\
& DynSignSparseSGD & $\times$ & $\checkmark$ & 74.5 $\pm$ 0.2 & 75.7 $\pm$ 0.2 & 48.8 $\pm$ 0.6 & 85.7 $\pm$ 2.6 & 85.6 $\pm$ 0.7 & 90.7 $\pm$ 0.2 & 94.5 $\pm$ 0.5 & 96.8 $\pm$ 1.1 & 97.3 $\pm$ 0.3 & 69.9 $\pm$ 0.6 & 87.1 $\pm$ 1.0 & 82.4 $\pm$ 0.7 \\
\midrule
\multirow{4}{*}{8} & SignSparseSGD & $\checkmark$ & $\times$ & 76.8 $\pm$ 0.2 & 77.5 $\pm$ 0.0 & 56.0 $\pm$ 1.2 & 89.2 $\pm$ 2.7 & 87.6 $\pm$ 0.3 & 90.7 $\pm$ 0.2 & 95.4 $\pm$ 0.2 & 97.3 $\pm$ 0.3 & 97.5 $\pm$ 0.6 & 72.0 $\pm$ 0.6 & 88.6 $\pm$ 0.5 & 84.4 $\pm$ 0.6 \\
& SignSparseSGD & $\times$ & $\checkmark$ & 78.8 $\pm$ 0.1 & 78.9 $\pm$ 0.2 & 49.6 $\pm$ 0.5 & 84.4 $\pm$ 0.4 & 86.2 $\pm$ 0.4 & 91.9 $\pm$ 0.1 & 95.4 $\pm$ 0.3 & 96.9 $\pm$ 0.2 & 97.5 $\pm$ 0.4 & 70.6 $\pm$ 0.6 & 86.7 $\pm$ 0.2 & 83.4 $\pm$ 0.3 \\
\cmidrule(lr){2-16}
& DynSignSparseSGD & $\checkmark$ & $\times$ & 77.3 $\pm$ 0.1 & 78.3 $\pm$ 0.1 & 58.7 $\pm$ 0.6 & 88.9 $\pm$ 2.7 & 88.8 $\pm$ 0.2 & 91.2 $\pm$ 0.1 & 95.5 $\pm$ 0.3 & 97.8 $\pm$ 0.5 & 97.4 $\pm$ 0.5 & 73.1 $\pm$ 0.4 & 89.2 $\pm$ 0.4 & 85.1 $\pm$ 0.5 \\
& DynSignSparseSGD & $\times$ & $\checkmark$ & 76.0 $\pm$ 0.1 & 77.3 $\pm$ 0.2 & 57.1 $\pm$ 0.2 & 87.4 $\pm$ 3.7 & 88.6 $\pm$ 0.5 & 90.8 $\pm$ 0.1 & 95.2 $\pm$ 0.3 & 98.2 $\pm$ 0.3 & 97.3 $\pm$ 0.4 & 72.0 $\pm$ 0.4 & 88.4 $\pm$ 0.3 & 84.4 $\pm$ 0.6 \\
\midrule
\multirow{4}{*}{16} & SignSparseSGD & $\checkmark$ & $\times$ & 78.0 $\pm$ 0.1 & 79.3 $\pm$ 0.3 & 61.6 $\pm$ 0.9 & 92.7 $\pm$ 1.5 & 89.5 $\pm$ 0.3 & 90.8 $\pm$ 0.2 & 95.4 $\pm$ 0.3 & 98.5 $\pm$ 0.1 & 97.7 $\pm$ 0.0 & 76.5 $\pm$ 0.8 & 90.1 $\pm$ 0.2 & 86.4 $\pm$ 0.4 \\
& SignSparseSGD & $\times$ & $\checkmark$ & 79.9 $\pm$ 0.1 & 80.3 $\pm$ 0.1 & 54.2 $\pm$ 0.6 & 88.1 $\pm$ 0.4 & 87.9 $\pm$ 0.2 & 92.2 $\pm$ 0.1 & 95.8 $\pm$ 0.1 & 98.4 $\pm$ 0.0 & 97.5 $\pm$ 0.2 & 75.1 $\pm$ 0.5 & 88.3 $\pm$ 0.4 & 85.2 $\pm$ 0.2 \\
\cmidrule(lr){2-16}
& DynSignSparseSGD & $\checkmark$ & $\times$ & 78.3 $\pm$ 0.1 & 79.6 $\pm$ 0.1 & 65.0 $\pm$ 0.3 & 93.4 $\pm$ 0.9 & 90.3 $\pm$ 0.1 & 91.2 $\pm$ 0.0 & 95.6 $\pm$ 0.1 & 98.9 $\pm$ 0.2 & 97.4 $\pm$ 0.3 & 77.8 $\pm$ 0.7 & 91.0 $\pm$ 0.2 & 87.1 $\pm$ 0.3 \\
& DynSignSparseSGD & $\times$ & $\checkmark$ & 77.4 $\pm$ 0.1 & 79.0 $\pm$ 0.1 & 64.2 $\pm$ 0.6 & 93.9 $\pm$ 0.6 & 90.2 $\pm$ 0.2 & 90.8 $\pm$ 0.0 & 95.2 $\pm$ 0.4 & 98.8 $\pm$ 0.2 & 97.4 $\pm$ 0.3 & 77.4 $\pm$ 0.4 & 90.3 $\pm$ 0.6 & 86.8 $\pm$ 0.3 \\
\bottomrule
\end{tabular}
\end{adjustbox}
\end{table}

Tables~\ref{tab:ablation_selection_scope_ViTB32_with_aug}, \ref{tab:ablation_selection_scope_ViTB16_with_aug}, and \ref{tab:ablation_selection_scope_ViTL14_with_aug} compare two scopes for magnitude-based support selection. The tensor-wise rule is exactly the blockwise rule in Eq.~\eqref{eq:block_budgets}, because each block $\mathcal I_r$ corresponds to one trainable parameter tensor. It applies $\Top_{k_r}$ independently within every tensor and unions the resulting supports. The model-wise alternative concatenates the same eligible trainable parameters and applies one $\Top_k$ operation at the same nominal density $\rho$. 

At the aggregate level, tensor-wise selection improves all 30 method--backbone--shot averages. For SignSparseSGD, the overall gain is $+1.12$, $+1.10$, and $+1.09$ points with ViT-B/16, ViT-B/32, and ViT-L/14, respectively. Averaged over the three backbones, the gains at 1, 2, 4, 8, and 16 shots are $+0.54$, $+1.22$, $+1.37$, $+1.17$, and $+1.22$ points. Therefore, the effect is not restricted to a particular backbone or shot level. It is smallest at 1 shot, peaks at 4 shots, and remains above one point from 2 to 16 shots.

DynSignSparseSGD also benefits from tensor-wise selection, but the gap is smaller. Its overall gains are $+0.35$, $+0.59$, and $+0.69$ points for ViT-B/16, ViT-B/32, and ViT-L/14. The corresponding backbone-averaged gains across shot levels are $+0.75$, $+0.73$, $+0.50$, $+0.47$, and $+0.27$ points. The advantage is strongest in the lowest-shot settings and generally decreases as the support set grows. Nevertheless, it remains positive for every backbone and shot count. Dynamic refresh also remains beneficial under either scope: DynSignSparseSGD exceeds SignSparseSGD in all 15 tensor-wise averages and all 15 model-wise averages.

The larger scope effect for SignSparseSGD is consistent with the difference between fixed and refreshed supports. Under static model-wise selection, one global ranking from the initial minibatch determines the allocation across all tensors for the complete trajectory. Any early concentration in a small set of tensors is therefore retained. Tensor-wise quotas prevent such cross-tensor competition by reserving $k_r$ coordinates for every tensor. Under dynamic model-wise selection, the global ranking is recomputed at each iteration. It can therefore reallocate the support as the gradient changes, which can partly mitigate an initially concentrated allocation. The accuracy comparison supports this interpretation, but it does not directly measure how selected coordinates are distributed across tensors.

The per-dataset results are heterogeneous. For SignSparseSGD, tensor-wise selection is higher in all 15 backbone--shot configurations on EuroSAT, Flowers102, DTD, and UCF101, and in 13 of 15 configurations on Aircraft. Using the printed per-dataset means, its average gains on these datasets are approximately $+7.71$, $+2.83$, $+1.69$, $+2.76$, and $+2.59$ points, respectively. Model-wise selection is consistently higher on ImageNet, SUN397, and Food101, with average differences of approximately $-2.65$, $-1.43$, and $-2.14$ points for tensor-wise minus model-wise selection. It is also higher in 13 of 15 configurations on Oxford Pets. Thus, the positive overall result does not imply uniform per-dataset dominance. It reflects larger gains on datasets that are especially sensitive to the allocation of a fixed sparse support.

The dynamic comparison is more uniform. Tensor-wise DynSignSparseSGD is higher in 133 of the 165 backbone--shot--dataset cells, equal in 8, and lower in 24. It is higher in all 15 configurations on ImageNet, SUN397, Food101, Oxford Pets, and DTD. Its largest dataset-level average gains are approximately $+1.56$ points on ImageNet and $+1.29$ on DTD. The main exceptions are Flowers102 and EuroSAT, where the mean differences are about $-0.49$ and $-0.14$ points. Reported seed deviations are of a similar order for the two dynamic scopes. For the static method, tensor-wise rows often have larger deviations, especially on datasets where the mean gains are large. Selection scope therefore changes average accuracy more consistently than it changes seed variability.

\enlargethispage{3\baselineskip}
Both rules use the same nominal density. Because Eq.~\eqref{eq:block_budgets} enforces at least one selected coordinate per tensor, floor and minimum-one effects can make the realized cardinalities differ slightly. The comparison evaluates the implemented per-tensor quota rule rather than a perfectly cardinality-matched relaxation. Tensor-wise selection is the stronger default. It improves every aggregate configuration and is most consequential when the support is fixed from one noisy initial gradient.

\section{Hardware and Software}
\label{sec:hardware_software}
\suppanchor{hardware_software}

All experiments were conducted on a Linux server using NVIDIA A100 GPUs on the Taylor cluster under a consistent software environment. Unless otherwise stated, all experiments were run with random seeds 1, 2, and 3. Table~\ref{tab:hardware_software} summarizes the hardware and software configuration.

\begin{table}[!h]
  \caption{Hardware and software environment.}
  \label{tab:hardware_software}
  \begin{center}
  \begin{small}
  \scalebox{0.9}{
  \begin{tabular}{|p{3.0cm}|c|}
    \hline
    \multicolumn{2}{|c|}{\textbf{Hardware}} \\ \hline
    \textbf{RAM} & 504 GB \\ \hline
    \textbf{CPU model} & Intel(R) Xeon(R) Silver 4310 CPU @ 2.10GHz \\ \hline
    \textbf{\# of CPUs} & 48 \\ \hline
    \textbf{GPU model} & NVIDIA A100 \\ \hline
    \textbf{GPU memory} & 40 GB \\ \hline
    \textbf{\# of GPUs} & 8 \\ \hline
    \multicolumn{2}{|c|}{\textbf{Software}} \\ \hline
    \textbf{Operating System} & Ubuntu 18.04.6 LTS \\ \hline
    \textbf{Python} & 3.10.16 \\ \hline
    \textbf{PyTorch} & 2.5.1 \\ \hline
    \textbf{CUDA} & 12.4 \\ \hline
    \multicolumn{2}{|c|}{\textbf{Reproducibility}} \\ \hline
    \textbf{Random seeds} & 1, 2, 3 \\ \hline
  \end{tabular}}
  \end{small}
  \end{center}
\end{table}
\FloatBarrier
\end{document}